Florentin Smarandache, Surapati Pramanik
*(Editors)*

# New Trends in Neutrosophic Theory and Applications

Florentin Smarandache, Surapati Pramanik (Editors)
New Trends in Neutrosophic Theory and Applications

Florentin Smarandache, Surapati Pramanik
*(Editors)*

# New Trends in Neutrosophic Theory and Applications

## Peer Reviewers

# TABLE OF CONTENTS

# Aims and Scope

Neutrosophic theory and applications have been expanding in all directions at an astonishing rate especially after the introduction the journal entitled "Neutrosophic Sets and Systems". New theories, techniques, algorithms have been rapidly developed. One of the most striking trends in the neutrosophic theory is the hybridization of neutrosophic set with other potential sets such as rough set, bipolar set, soft set, hesitant fuzzy set, etc. The different hybrid structure such as rough neutrosophic set, single valued neutrosophic rough set, bipolar neutrosophic set, single valued neutrosophic hesitant fuzzy set, etc. are proposed in the literature in a short period of time. Neutrosophic set has been a very important tool in all various areas of data mining, decision making, e-learning, engineering, medicine, social science, and some more.

The Book "New Trends in Neutrosophic Theories and Applications" focuses on theories, methods, algorithms for decision making and also applications involving neutrosophic information. Some topics deal with data mining, decision making, e-learning, graph theory, medical diagnosis, probability theory, topology, and some more.

*Florentin Smarandache, Surapati Pramanik*

# Preface

Neutrosophic set has been derived from a new branch of philosophy, namely Neutrosophy. Neutrosophic set is capable of dealing with uncertainty, indeterminacy and inconsistent information. Neutrosophic set approaches are suitable to modeling problems with uncertainty, indeterminacy and inconsistent information in which human knowledge is necessary, and human evaluation is needed.

Neutrosophic set theory was proposed in 1998 by Florentin Smarandache, who also developed the concept of single valued neutrosophic set, oriented towards real world scientific and engineering applications. Since then, the single valued neutrosophic set theory has been extensively studied in books and monographs introducing neutrosophic sets and its applications, by many authors around the world. Also, an international journal - *Neutrosophic Sets and Systems* started its journey in 2013.

Single valued neutrosophic sets have found their way into several hybrid systems, such as neutrosophic soft set, rough neutrosophic set, neutrosophic bipolar set, neutrosophic expert set, rough bipolar neutrosophic set, neutrosophic hesitant fuzzy set, etc. Successful applications of single valued neutrosophic sets have been developed in multiple criteria and multiple attribute decision making.

The present book starts by proposing an approach for data mining with single valued neutrosophic information from large amounts of data and then progresses to topics in decision making in neutrosophic environment and neutrosophic hybrid environment, e-learning, graph theory, medical diagnosis, neutrosophic models in sociology, topology, and some more.

The book collects thirty original research and application papers from different perspectives covering different areas of neutrosophic studies, such data mining, decision making, e-learning, graph theory, medical diagnosis, probability theory, topology, and some theoretical papers. This book shows examples applications of neutrosophic set and neutrosophic hybrid set in multiple criteria and multiple attribute decision making, medical diagnosis, etc.

The first chapter presents the two essential pillars of data mining: similarity measures and machine learning in single valued neutrosophic environment. It shows that neutrosophic logic can perform an important role in data mining method. It defines single valued neutrosophic score function (*SVNSF*) to aggregate attribute values of each alternative. It also presents an approach of data mining with single valued neutrosophic information from large amounts of data, and furnishes a numerical example for the proposed approach.

The second, third, and fourth chapter deal with decision making in neutrosophic hesitant fuzzy information.

The second chapter presents a class of distance measures for single-valued neutrosophic hesitant fuzzy sets and discusses their properties with parameter changing. It also provides multi-attribute decision making (MADM), an illustrative example, and a comparison with other existing methods.

The third chapter presents an axiomatic system of distance and similarity measures between single-valued neutrosophic hesitant fuzzy sets. It also proposes a class of distance and similarity measures based on three basic forms: the geometric distance model, the set-theoretic approach, and the matching functions. The distance measure between each alternative and the ideal alternative is used to establish a multiple attribute decision making method under single-valued neutrosophic hesitant fuzzy environment. It provides a numerical example of investment alternatives to show the effectiveness and the usefulness of the proposed approach.

The fourth chapter extends grey relational analysis (GRA) method for MADM by defining score value, accuracy value, certainty value, and normalized Hamming distance of single-valued neutrosophic hesitant fuzzy set (SVNHFS). It also defines the positive ideal solution (PIS) and the negative ideal solution (NIS) by score value and accuracy value. It proposes GRA method for multi-attribute decision making under single valued neutrosophic hesitant fuzzy set environment. It also provides an illustrative example to demonstrate the validity and the effectiveness of the proposed method.

The fifth chapter exposes TOPSIS method for MADM problems with bipolar neutrosophic information. The Hamming and the Euclidean distance functions are defined in order to determine the distance between bipolar neutrosophic numbers. The bipolar neutrosophic relative positive ideal solution (BNRPIS) and bipolar neutrosophic relative negative ideal solution (BNRNIS) are also characterized. The applicability of the proposed method is verified and a comparison with other existing methods is provided.

The sixth chapter presents TOPSIS approach for multi-attribute decision making in refined neutrosophic environment. An illustrative numerical example of tablet selection is provided to show the applicability of the proposed TOPSIS approach.

The seventh chapter presents several new similarity measures based on trigonometric Hamming similarity operators of rough neutrosophic sets and their applications in decision making. Some properties of the proposed similarity measures are established. Also, a numerical example is given to illustrate the applicability of the proposed similarity measures in decision making.

The eighth chapter develops a fuzzy single valued neutrosophic set with entropy weight based MADM technique. Its feasibility for automated guided vehicle (AGV) selecting and ranking of material handling systems for a given industrial application is examined.

The ninth chapter makes a connection between decision making in game and real life.

The tenth chapter develops two new methods for solving multiple attribute decision making (MADM) problems with interval valued neutrosophic assessments. It also discusses an alternative method to solve MADM problems based on the combination of angle cosine and projection method. Finally, an illustrative numerical example in Khadi institution is provided to verify the effectiveness of the proposed methods.

The eleventh chapter introduces improved weighted average geometric operator and define new score function. It establishes a multiple-attribute decision-making method based the proposed operator and newly defined score function.

The twelfth chapter presents modeling of logistics center location problem using the score and accuracy function, hybrid-score-accuracy function of SVNNs and linguistic variables under single-valued neutrosophic environment, where weight of the decision makers are completely unknown and the weight of criteria are incompletely known.

The thirteenth chapter reports about current trends to enhance e-learning process by using neutrosophic techniques to extract useful knowledge for selecting, evaluating, personalizing, and adapting the eLearning process.

The fourteenth chapter introduces certain types of single valued neutrosophic graphs, such as strong single valued neutrosophic graph, constant single valued neutrosophic graph and complete single valued neutrosophic graphs. It investigates some of their properties with proofs and examples.

The fifteenth chapter combines the concept of bipolar neutrosophic set and graph theory. It introduces the notions of bipolar single valued neutrosophic graphs, strong bipolar single valued neutrosophic graphs, complete bipolar single valued neutrosophic graphs, and regular bipolar single valued neutrosophic graphs. It also investigates some of their related properties.

The sixteenth chapter expounds two ways of determining the neutrosophic distance between neutrosophic vertex graphs. It proposes two neutrosophic distances based on the Haussdorff distance, and a robust modified variant of the Haussdorff distance. These distances satisfy the metric distance measure axioms. Furthermore, a similarity measure between neutrosophic edge graphs is explained, based on a probabilistic variant of Haussdorff distance.

The seventeenth chapter describes operations on interval valued neutrosophic graphs. It presents operations of Cartesian product, composition, union and join on interval valued neutrosophic graphs. It investigates some of their properties with proofs and examples.

The eighteenth chapter conveys the usefulness of neutrosophic theory in medical imaging, e.g. denoising and segmentation.

The nineteenth chapter delivers a theoretical framework of love dynamics in neutrosophic environment.

The twentieth chapter emphasizes the neutrosophic crisp probability theory and the decision making process by presenting some fundamental definitions and operations.

The $21^{st}$ and $22^{nd}$ chapters devote to study neutrosophic sets and neutrosophic topology.

Chapters from $23^{rd}$ to $30^{th}$ present theoretical improvements of neutrosophic set and its variants.

We hope that this book will offer a useful resource of ideas, techniques, methods, and approaches for additional researches on applications in different fields of neutrosophic sets and various neutrosophic hybrid sets.

We are grateful to our referees, whose valuable and highly appreciated reviews guided us in selecting the chapters in the book.

*Florentin Smarandache, Surapati Pramanik*

# DATA MINING

KALYAN MONDAL[1], SURAPATI PRAMANIK[2], BIBHAS C. GIRI[3]

[1]Department of Mathematics, Jadavpur University, West Bengal, India. E-mail: kalyanmathematic@gmail.com
[2]Department of Mathematics, Nandalal Ghosh B.T. College, Panpur, PO-Narayanpur, and District: North 24 Parganas, Pin Code: 743126, West Bengal, India. E-mail: sura_pati@yahoo.co.in
[3]Department of Mathematics, Jadavpur University, West Bengal, India. E-mail: bcgiri.jumath@gmail.com

# Role of Neutrosophic Logic in Data Mining

## Abstract

This paper presents a data mining process of single valued neutrosophic information. This approach gives a presentation of data analysis common to all applications. Data mining depends on two main elements, namely the concept of similarity and the machine learning framework. It describes a lot of real world applications for the domains namely mathematical, medical, educational, chemical, multimedia etc. There are two main types of indeterminacy in supervised learning: cognitive and statistical. Statistical indeterminacy deals with the random behavior of nature. All existing data mining techniques can handle the uncertainty that arises (or is assumed to arise) in the natural world from statistical variations or randomness. Cognitive uncertainty deals with human cognition. In real world problems for data mining, indeterminacy components may arise. Neutrosophic logic can handle this situation. In this paper, we have shown the role of single valued neutrosophic set logic in data mining. We also propose a data mining approach in single valued neutrosophic environment.

## 1. Introduction

Data mining [1] is actually assumed as “knowledge mining” from data. Data mining is an essential process where intelligent methods are applied to extract data patterns [2]. Data mining is a process that analyzes large amounts of data to find new and hidden information. In other words; it is the process of analyzing data from different perspectives and summarizing it into some useful information. The following are the different data mining techniques [3]: association, classification, clustering, and sequential patterns. E.Hullermeier [4] proposed fuzzy methods in data mining.

This paper focuses on real-world applications of single valued neutrosophic set [5] for data mining. Data mining decomposes into two main elements: the notion of similarity and the single valued neutrosophic machine learning techniques that are applied in the described applications. Indeed, similarity, or more generally comparison measures are used at all levels of the data mining and information retrieval tasks. At the lowest level, they are used for the matching between a query to a database and the elements it contains, for the extraction of relevant data. Then similarity and dissimilarity measures can be used in the process of cleaning and management of missing data to create a reasonable set of data. To generalize particular information contained in this reasonable set, dissimilarity measures are used in the case of inductive learning and similarity measures for case-based reasoning or clustering tasks. Eventually, similarities are used to interpret results of the learning process into an expressible form of knowledge through the definition of prototypes. Most of collective data for an investigation involves indeterminacy. Single valued neutrosophic set can handle his situation. So, there is an important role of single valued neutrosophic set in data mining.

This paper is arranged as follows. Section 2 presents some basic knowledge of single valued neutrosophic set. Section 3 considers the component of similarity, and machine learning techniques. Section 4 describes a methodical approach of data mining under single valued neutrosophic environment. Section 5 presents a numerical example for data mining. Section 6 presents concluding remarks.

## 2. Neutrosophic Preliminaries

### 2.1 Definition on neutrosophic sets [6]

The concept of neutrosophic set is originated from neutrosophy [6], a new branch of philosophy.

**Definition 1:**[6] Let $\xi$ be a space of points (objects) with generic element in $\xi$ denoted by $x$. Then a neutrosophic set $\alpha$ in $\xi$ is characterized by a truth membership function $T_\alpha$ an indeterminacy membership function $I_\alpha$ and a falsity membership function $F_\alpha$. The functions $T_\alpha$ and $F_\alpha$ are real standard or non-standard subsets of $]^-0, 1^+[$ that is $T_\alpha: \xi \to ]^-0, 1^+[$; $I_\alpha: \xi \to ]^-0, 1^+[$; $F_\alpha: \xi \to ]^-0, 1^+[$.

It should be noted that there is no restriction on the sum of $T_\alpha(x)$, $I_\alpha(x)$, $F_\alpha(x)$ i.e. $^-0 \le T_\alpha(x) + I_\alpha(x) + F_\alpha(x) \le 3^+$

**Definition 2:** [6] The complement of a single valued neutrosophic set α is denoted by $\alpha^c$ and is defined by

$$T_{\alpha^c}(x) = \{1^+\} - T_\alpha(x);\ I_{\alpha^c}(x) = \{1^+\} - I_\alpha(x)$$

$$F_{\alpha^c}(x) = \{1^+\} - F_\alpha(x)$$

**Definition 3: (Containment)** [6] A single valued neutrosophic set α is contained in the other single valued neutrosophic set β, $\alpha \subseteq \beta$ if and only if the following result holds.

$$\inf T_\alpha(x) \le \inf T_\beta(x),\ \sup T_\alpha(x) \le \sup T_\beta(x)$$

$$\inf I_\alpha(x) \ge \inf I_\beta(x),\ \sup I_\alpha(x) \ge \sup I_\beta(x)$$

$$\inf F_\alpha(x) \ge \inf F_\beta(x),\ \sup F_\alpha(x) \ge \sup F_\beta(x)$$

for all $x$ in $\xi$.

**Definition 4: (Single-valued single valued neutrosophic set)**[5] .

Let $\xi$ be a universal space of points (objects) with a generic element of $\xi$ denoted by $x$.

A single-valued single valued neutrosophic set *S* is characterized by a true membership function $T_s(x)$, an indeterminacy membership function $I_s(x)$, a falsity membership function $F_s(x)$ with $T_s(x)$, $I_s(x)$, $F_s(x) \in [0, 1]$ for all *x* in $\xi$. When $\xi$ is continuous a SNVS can be written as

$$S = \int_x \langle T_s(x), F_s(x), I_s(x) \rangle / x, \forall x \in \xi$$

and when $\xi$ is discrete a SVNSs *S* can be written as:

$$S = \sum \langle T_S(x), F_S(x), I_S(x) \rangle / x, \forall x \in \xi$$

It should be noted that for a SVNS *S*,

$$0 \le \sup T_S(x) + \sup F_S(x) + \sup I_S(x) \le 3, \forall x \in \xi$$

and for a single valued neutrosophic set, the following relation holds:

$$0 \le \sup T_S(x) + \sup F_S(x) + \sup I_S(x) \le 3, \forall x \in \xi$$

**Definition 5:** The complement of a single valued neutrosophic set *S* is denoted by $S^c$ and is defined by

$$T_S{}^c(x) = F_S(x); \quad I_S{}^c(x) = 1 - I_S(x); \quad F_S{}^c(x) = T_S(x)$$

**Definition 6:** A SVNS $S_\alpha$ is contained in the other SVNS $S_\beta$, denoted as $S_\alpha \subseteq S_\beta$ iff, $T_{S_\alpha}(x) \le T_{S_\beta}(x)$; $I_{S_\alpha}(x) \ge I_{S_\beta}(x)$; $F_{S_\alpha}(x) \ge F_{S_\beta}(x)$, $\forall x \in \xi$.

**Definition 7:** Two single valued single valued neutrosophic sets $S_\alpha$ and $S_\beta$ are equal, i.e. $S_\alpha = S_\beta$, if and only if $S_\alpha \subseteq S_\beta$ and $S_\alpha \supseteq S_\beta$

**Definition 8: (Union)** The union of two SVNSs $S_\alpha$ and $S_\beta$ is a SVNS $S_\gamma$, written as $S_\gamma = S_\alpha \cup S_\beta$.

Its truth membership, indeterminacy-membership and falsity membership functions are related to those of $S_\alpha$ and $S_\beta$ by

$$T_{S_\gamma}(x) = \max\left(T_{S_\alpha}(x), T_{S_\beta}(x)\right);$$

$$I_{S_\gamma}(x) = \max\left(I_{S_\alpha}(x), I_{S_\beta}(x)\right);$$

$F_{S_\gamma}(x) = \min\left(F_{S_\alpha}(x), F_{S_\beta}(x)\right)$ for all *x* in $\xi$

**Definition 9: (intersection)** The intersection of two SVNSs, $S_\alpha$ and $S_\beta$ is a SVNS $S_\delta$, written as $S_\delta = S_\alpha \cap S_\beta$. Its truth membership, indeterminacy-membership and falsity membership functions are related to those of $S_\alpha$ an $S_\beta$ as follows:

$$T_{S_\delta}(x) = \min\left(T_{S_\alpha}(x), T_{S_\beta}(x)\right);$$

$$I_{S_\delta}(x) = \max\left(I_{S_\alpha}(x), I_{S_\beta}(x)\right);$$

$$F_{S_\delta}(x) = \max\left(F_{S_\alpha}(x), F_{S_\beta}(x)\right), \forall x \in \xi$$

## 3. Data Mining [2]

In this section, we discuss the theoretical background common to the applications, considering successively the notion of similarity and machine learning techniques under single valued neutrosophic environment.

### 3.1 Similarity [2]

The notion of similarity or more generally of comparison measures, is central for all real-world applications: it aims at quantifying the extent to which two objects are similar, or dissimilar, one to another, providing a numerical value for this comparison. Similarities and dissimilarities between objects are generally evaluated from values of their attributes or variables characterizing these objects. Dissimilarities are classically defined from distances. Similarities and dissimilarities are often expressed from each other: the more similar two objects are, the less dissimilar they are, the smaller their distance. Weights can be associated with variables, according to the semantics of the application or the importance of the variables. It appears that some quantities are used in various environments, with different forms, based on the same principles. Most of the classic dissimilarity measures between two objects with continuous numerical attributes are the Euclidian distance, the Manhattan distance, and more generally Minkowski distances.

### 3.2 Neutrosophy Machine Learning [2]

The second part of the theoretical background common to all applications concerns the neutrosophy machine learning techniques that use the previous similarity measures. Machine learning is an important way to extract knowledge from sets of cases, especially in large scale databases. In this section, we consider only the neutrosophy machine learning methods (involving indeterminacy) that are used in the applications, leaving aside other techniques as for neutrosophy case-based reasoning or neutrosophy association rules. Three methods are successively considered: neutrosophy decision trees, neutrosophy prototypes and neutrosophy clustering. The first two belong to the supervised learning framework, i.e. they consider that each data point is associated with a category. Single valued neutrosophic set clustering belongs to the unsupervised learning framework, i.e. no decomposition of the data set with indeterminacy into categories is available.

#### 3.2.1. Single valued neutrosophic set Decision Trees [2]

Neutrosophy decision trees (NDT) particularly can be interesting for data mining and information retrieval because they enable the user to take into account indeterminacy descriptions of the cases, or heterogeneous values (symbolic, numerical, or neutrosophical) [5]. Moreover, they are appreciated for their interpretability, because they provide a linguistic description of the relations between descriptions of the cases and decision to make or class to assign. The rules obtained through NDT make it easier for the user to interact with the system or the expert to understand, confirm or amend his own knowledge. Another quality of NDT is their robustness, since a small variation of descriptions does not drastically change the decision or the class associated with a case, which guarantees a resistance to measurement errors and avoids sharp differences for close values of the descriptions.

**3.2.2. Single valued neutrosophic set Prototype Construction** [2]

Neutrosophy prototypes are another approaches to the characterization of data categories: they provide descriptions or interpretable summarizations of data sets, so as to help a user to better apprehend their contents: a prototype is an element chosen to represent a group of data, to summarize it and underline its most characteristic features. It can be defined from a statistical point of view, for instance as the data mean or the median; more complex representatives can also be used as the most typical value [7] for instance. The prototype notion was also studied from a cognitive science point of view, and specific properties were pointed out in [8]: it was shown that a prototype underlines the common features of the category members, but also their distinctive features as opposed to other categories, underlining the specificity of the group. Furthermore, prototypes were related to the typicality notion, i.e. the fact that all data do not have the same status as regards the group: some members of the group are better examples, more representative or more characteristic than others. It was also shown that the typicality of a point depends both on its resemblance to other members of the group (internal resemblance), and on its dissimilarity to members of other groups (external dissimilarity). More precisely, the method consists of computing internal resemblance and external dissimilarity for each data point. Internal resemblance and external dissimilarity are respectively defined as the aggregation (mean or median) of the resemblance to the other members of the group, and as the aggregation of the dissimilarity to members of other groups, for a given choice of the resemblance and dissimilarity measures.

## 4. Single Valued Neutrosophic Logic in Data Mining

The tools that have been proposed in single valued neutrosophic set (SVNS) have the potential to support all of the steps that neutralized a process of knowledge discovery. SVNS can be used in the data selection and preparation phase for data modeling. For any data analysis associated with an experiment or investigation, it is observed that much information involve indeterminacy. Single valued neutrosophic set logic is capable of dealing with this situation. So, for the case of data mining single valued neutrosophic set logic has an important role.

Standard methods of data analysis can be extended in a rather generic way by means of an extension principle. For example, the functional relation between the data points and the decision making function can be extended to the case of single valued neutrosophic data, where the observations are described in terms of single valued neutrosophic sets. If single valued neutrosophic data is not used in the data preparation phase, they can still be employed in a later stage in order to analyze the original data.

Various techniques are widely used for data mining from gathering data within a domain of expertise. Delphi method [9] and BIRCH method [10] are very popular for data mining. Rekha and Swapna [2] studied the role of fuzzy logic in data mining. Literature review reflects that there is no single valued neutrosophic approach for data mining till now.

### 4.1. Neutrosophic data mining method

Generally, there are many attributes in decision making problems, where some of them are important and others may not be so important. So it is crucial to select the proper attributes for decision-making situation. Now, we shall propose a methodical approach for data mining with single valued neutrosophic information to prepare a panel of attributes which are technically sound. All steps of this proposed approach are given as follows.

**Step 1: Problem field selection**

Consider a multi-attribute decision making problem with m alternatives and n attributes (large numbers of data). Let $A_1, A_2, ..., A_m$ and $C_1, C_2, ..., C_n$ denote the alternatives and attributes respectively. In decision making process, we have to select a finite but more important attributes from given n attributes. All attributes are expressed in single valued neutrosophic number.

**Table 1:** *Single valued neutrosophic set decision matrix*

$D = \langle d_{ij} \rangle_{m \times n} =$

$$\begin{array}{c|cccc} & C_1 & C_2 & \cdots & C_n \\ \hline A_1 & \langle d_{11} \rangle & \langle d_{12} \rangle & \cdots & \langle d_{1n} \rangle \\ A_2 & \langle d_{21} \rangle & \langle d_{22} \rangle & \cdots & \langle d_{2n} \rangle \\ . & \cdots & \cdots & \cdots & \cdots \\ . & \cdots & \cdots & \cdots & \cdots \\ A_m & \langle d_{m1} \rangle & \langle d_{m2} \rangle & \cdots & \langle d_{mn} \rangle \end{array} \quad (1)$$

Here, $d_{ij}$ ($i$ = 1, 2, …, $m$ and $j$ = 1, 2, …, $n$) are all single valued neutrosophic numbers.

**Step 2: Single valued neutrosophic set score matrix**

**Definition 10: Single valued neutrosophic score function (SVNSF)**

Single valued neutrosophic score function (*SVNSF*) corresponding to each attribute is defined as follows.

$$SVNSF(C_j) = \frac{1}{m} \sum_{r=1}^{m} \frac{2 + T_{rj} - I_{rj} - F_{rj}}{3} \quad (2)$$

Where, $j$ = 1, 2, …, $n$

Using equation (2) we calculate single valued neutrosophic score matrix as follows.

**Table:** *Single valued neutrosophic score matrix*

$$SVNSF(C_j) = \begin{array}{c|c} attributes & Single\ valued\ neutrosophic\ score\ value \\ \hline C_1 & SVNSF(C_1) \\ C_2 & SVNSF(C_2) \\ \vdots & \vdots \\ C_n & SVNSF(C_n) \end{array} \quad (3)$$

**Step 3: Selection zone**

Single valued neutrosophic score values are classified into three zones. These are described as follows.

**Definition 11:** *SVNSF* of all the attributes are classified in three categories and it is defined as follows

**Highly acceptable zone:** $0.50 \le SVNSF(C_j) \le 1$

**Tolerable acceptable zone:** $0.25 \le SVNSF(C_j) \le 0.50$

**Unacceptable acceptable zone:** $0.00 \le SVNSF(C_j) \le 0.25$

**Step 4: Ranking of attributes**

According to the single valued neutrosophic score values, we can set up a panel of all attributes in descending order and we can choose important attributes from large number of attributes into decision making process considering highly acceptable zone and tolerable acceptable zone

**Step 5:** End

## 5. Numerical Example

In this section we demonstrate a numerical problem for applicability and effectiveness of this proposed approach. The methodical steps are as follows.

**Step 1: Problem field selection**

Suppose a person who wants to purchase a SIM card for mobile connection. So, it is necessary to select suitable SIM card for his/her mobile connection. There is a panel with four possible alternatives (SIM cards) for mobile connection. The alternatives (SIM cards) are presented as follows:

$A_1$: Airtel

$A_2$: Vodafone

$A_3$: BSNL

$A_4$: IDEA.

For this purpose, the following attributes about SIM cards may be arise in decision making process. These are stated as follows.

1. Service quality of the corresponding company ($C_1$)
2. Cost ($C_2$)
3. Call rate per second ($C_3$)
4. Internet facilities ($C_4$)
5. Tower facility ($C_5$)
6. Call drops ($C_6$)
7. Risk factor ($C_7$)

**Table3**: Single valued neutrosophic decision matrix

$D = \langle d_{ij} \rangle_{4\times7} =$

| | $C_1$ | $C_2$ | $C_3$ | $C_4$ | $C_5$ | $C_6$ | $C_7$ |
|---|---|---|---|---|---|---|---|
| $A_1$ | $\langle 0.8, 0.3, 0.2 \rangle$ | $\langle 0.7, 0.3, 0.2 \rangle$ | $\langle 0.7, 0.2, 0.4 \rangle$ | $\langle 0.8, 0.1, 0.2 \rangle$ | $\langle 0.3, 0.5, 0.5 \rangle$ | $\langle 0.2, 0.3, 0.6 \rangle$ | $\langle 0.1, 0.6, 0.5 \rangle$ |
| $A_2$ | $\langle 0.8, 0.3, 0.3 \rangle$ | $\langle 0.7, 0.1, 0.2 \rangle$ | $\langle 0.7, 0.3, 0.4 \rangle$ | $\langle 0.8, 0.1, 0.1 \rangle$ | $\langle 0.3, 0.4, 0.5 \rangle$ | $\langle 0.2, 0.5, 0.6 \rangle$ | $\langle 0.1, 0.4, 0.5 \rangle$ |
| $A_3$ | $\langle 0.8, 0.2, 0.2 \rangle$ | $\langle 0.7, 0.3, 0.1 \rangle$ | $\langle 0.7, 0.4, 0.4 \rangle$ | $\langle 0.8, 0.2, 0.2 \rangle$ | $\langle 0.3, 0.5, 0.6 \rangle$ | $\langle 0.2, 0.1, 0.5 \rangle$ | $\langle 0.1, 0.6, 0.3 \rangle$ |
| $A_4$ | $\langle 0.8, 0.1, 0.2 \rangle$ | $\langle 0.7, 0.0, 0.2 \rangle$ | $\langle 0.7, 0.1, 0.4 \rangle$ | $\langle 0.8, 0.3, 0.2 \rangle$ | $\langle 0.3, 0.3, 0.3 \rangle$ | $\langle 0.2, 0.3, 0.3 \rangle$ | $\langle 0.1, 0.5, 0.4 \rangle$ |

(4)

**Step 2: Single valued neutrosophic score matrix**

Using equation (2) we calculate single valued neutrosophic score matrix as follows.

**Table 4:** Single valued neutrosophic set score matrix

$$SVNSF(C_j) = \begin{array}{c|c} attributes & Single\ valued\ neutrosophic\ score\ value \\ \hline C_1 & 0.7833 \\ C_2 & 0.7833 \\ C_3 & 0.6833 \\ C_4 & 0.8167 \\ C_5 & 0.4667 \\ C_6 & 0.4667 \\ C_7 & 0.3833 \end{array} \quad (5)$$

**Step 3: Selection zone**

Single valued neutrosophic score values are classified into three zones. These are described as follows.

**Definition 11:** *SVNSF* of all the attributes are classified in three categories and it is defined as follows

**Highly acceptable zone:** $0.50 \le SVNSF(C_j) \le 1$

**Tolerable acceptable zone:** $0.25 \le SVNSF(C_j) \le 0.50$

**Unacceptable acceptable zone:** $0.00 \le SVNSF(C_j) \le 0.25$

**Step 4: Ranking of attributes**

From equation (5) we can write single valued neutrosophic score values of all attributes in descending order as follows.

$$SVNSF(C_4) \succ SVNSF(C_1) \succ SVNSF(C_2) \succ SVNSF(C_3) \succ SVNSF(C_5) \succ SVNSF(C_6) \succ SVNSF(C_7)$$

So, attributes corresponding to single values neutrosophic score values (highly acceptable and tolerance zone) can be chosen as important attributes for decision making process.

**Step 5:** End

## 6. Conclusion

In this paper we briefly present first two of the essential pillars of data mining: similarity measures and machine learning in single valued neutrosophic environment. We showed that neutrosophic logic can perform an important role in data mining method. We define single valued neutrosophic score function (*SVNSF*) to aggregate attribute values of each alternative. We also propose an approach for data mining with single valued neutrosophic information from large amounts of data and furnish a numerical example for the proposed approach. In future this method can be extended in interval neutrosophic environment for data mining.

## References

1. Jiawei Han, Micheline Kamber, Jian Pei, Data mining concepts and techniques, 2012
2. M. Rekha and M. Swapna. Role of fuzzy logic in data Mining, International Journal of Advance Research in Computer Science and Management Studies, 2(12), 2014.
3. E.Hullermeier Fuzzy methods in data mining: status and prospects. Fuzzy Sets and Systems, 156(3), 387-406, 2005.
4. C.Z Janikow. Fuzzy decision trees: issues and methods IEEE transactions on systems Man and Cybernetics, 28(1), 1-14, 1998.
5. H. Wang, F. Smarandache, Y. Q. Zhang and R. Sunderraman. Single valued neutrosophic sets, Multispace and Multistructure, 4(2010), 410-413.
6. F. Smarandache. A unifying field in logics. neutrosophy: neutrosophic probability, set and logic. Rehoboth: American Research Press, 1998.
7. M.Friedman and M.Ming and A.Kandel. On the theory of typicality, International journal of Uncertainity, Fuzzyness and Knowledge-Based Systems, 3(2): 127-142, 1995.
8. E.Rosch and C.Mervis .Family Resemblance: Studies of internal structure of categories, Cognitive psychology, 7, 573-605, 1975.
9. C. C Hsu and B A. Sandford. The Delphi technique: making sense of consensus, practical assessment, Research and Evaluation, 12(10), 2007.
10. Y. Rani, Manju and H. Rohil. Comparative Analysis of BIRCH and CURE Hierarchical Clustering Algorithm using WEKA 3.6.9, The SIJ Transactions on Computer Science Engineering & its Applications (CSEA), 2 (1), 2014.

# DECISION MAKING

**Pranab Biswas**[1], **Surapati Pramanik**[2*], **Bibhas C. Giri**[3]

1Department of Mathematics, Jadavpur University, Kolkata, 700032, India. E-mail: paldam2010@gmail.com
2*Department of Mathematics, Nandalal Ghosh B.T. College, Panpur, P.O.-Narayanpr, District-North 24 Parganas, West Bengal, PIN-743126, India. Corresponding author's E-mail: sura_pati@yahoo.co.in
3Department of Mathematics, Jadavpur University, Kolkata,700032, India. E-mail: bcgiri.jumath@gmail.com

# Some Distance Measures of Single Valued Neutrosophic Hesitant Fuzzy Sets and Their Applications to Multiple Attribute Decision Making

## Abstract

Single-valued neutrosophic hesitant fuzzy set is a merged form of single-valued neutrosophic sets and hesitant fuzzy sets. This set is a useful tool to handle imprecise, incomplete and inconsistent information existing in multi-attribute decision making problems. In multi-attribute decision making, distance measures play an important role to take a decision regarding alternatives. In this paper we propose a variety of distance measures for single valued neutrosophic sets. Furthermore, we apply these measures to multi-attribute decision making problem with single-valued neutrosophic hesitant fuzzy set environment to find out the best alternative. We provide an illustrative example to validate and to show fruitfulness of the proposed approach. Finally, we compare the proposed approach with other existing methods for solving multi-attribute decision making problems.

## 1. Introduction

Distance and similarity measures are significant in a variety of scientific fields such as decision making, pattern recognition, and market prediction. Lots of studies have been done on fuzzy sets [1], intuitionistic fuzzy sets [2], and neutrosophic sets [3]. Among them the most widely used distance measure are Hamming distance and Euclidean distance. Generally when people make decision, they often hesitate to select for one thing or another to reach the final decision. Tora and Narukawa [4], Tora [5] introduced hesitant fuzzy set (HFS), an extension of fuzzy set, which allows the membership degree to assume a set of possible values. HFS can express the hesitant information compressively than other extensions of fuzzy sets. Xu and Xia [6] defined some distance measures on the basis of well-known Hamming distance and Euclidean distance and the Housdroff metric. They developed a class of hesitant distance measures and discussed some of their properties. Peng et al. [7] proposed the generalised hesitant fuzzy synergetic weighted distance measure and applied it to multi-attribute decision making (MADM) problem, where the best alternative. Having defined hesitancy degree, Li et al. [8] proposed some distance and similarity measures on HFSs and developed a TOPSIS method for MADM.

On the other hand, Zhu et al. [9] proposed a dual hesitant fuzzy set (DHFS) which consists of two parts – one is the membership hesitancy function and another is the non-membership hesitancy function. DHFS generalises fuzzy set (FS), intuitionistic fuzzy set (IFS), hesitant fuzzy set (HFS), and its membership degree and non-membership degree are presented by two set of possible values. Consequently, DHFS can represent imprecise and uncertain information existing in real decision making problem in more flexible way than FS, IFS, HFS. Singh [10] defined some distance and similarity measures of DHFSs on the basis of the geometric distance model, the set theoretic approach and the matching functions to study MADM with DHFSs.

However, HFSs and DHFSs cannot represent indeterminacy hesitant function for incomplete or inconsistent information. This type of function is an another issue to be considered in decision making and thus it should be included with membership hesitant and non-membership hesitant function to catch up imprecise, incomplete, inconsistent information found in decision making process. Ye [11] introduced single-valued neutrosophic hesitant fuzzy set (SVNHF) which consists of three parts – the truth membership hesitancy function, the indeterminacy membership hesitancy function, and falsity membership hesitancy function. This set can express imprecise, incomplete, inconsistent information with these three kinds of hesitancy functions in a more flexible way. In same discussions [11], Ye developed two aggregation operators for SVNHFS information and applied these operators to MADM problems. Sahin and Liu [12] defined correlation coefficient of SVNHFSs to solve MADM with SVNHFSs. Literature review suggests that the distance measures and similarity measures have not been studied, therefore we need to develop distance measures for SVNHFSs.

In this paper, we propose a class of distance measures for single-valued neutrosophic hesitant fuzzy sets and study their properties with variational parameters. We apply the weighted distance measures to calculate the distances between each alternative and ideal alternative in the MADM problems. With these distance values, we present the ranking order of alternatives for selecting the best one. We present an illustrative example to verify the proposed approach and to show its fruitfulness. Finally, we compare the proposed method with other existing methods for solving MADM under SVNHF environment.

The rest of the paper is organised as follows: Section 2 presents some basics of single-valued neutrosophic set and hesitant fuzzy sets and the existing distance measures for HFSs. Section 3 proposes Hamming distance measure, Euclidean distance measure, generalised distance measure, and Hausdroff distance. Section 4 devotes application of proposed distance measure to MADM with SVNHFS information. In Section 5, an illustrative example is given to validate and show effectiveness of the proposed approach. In Section 6, we present concluding remarks and future scope of research.

## 2. Preliminaries

In this section we review some basic definitions regarding single-valued neutrosophic sets and hesitant fuzzy sets to develop the present paper.

### 2.1. Single valued neutrosophic set

**Definition 1.** [13]

Let $X$ be a space of points (objects) with a generic element in $X$ denoted by $x$. A SVNS $A$ in $X$ is characterized by a truth membership function $T_A(\mathrm{x})$, an indeterminacy membership function $I_A(\mathrm{x})$, and a falsity membership function $F_A(\mathrm{x})$ and is denoted by

$$A=\{x,\langle T_A(\mathrm{x}),I_A(\mathrm{x}),F_A(\mathrm{x})\rangle \mid x\in X\}.$$

Here $T_A(\mathrm{x})$, $I_A(\mathrm{x})$ and $F_A(\mathrm{x})$ are real subsets of $[0,1]$ that is $T_A(\mathrm{x}):\mathrm{X}\to[0,1]$, $I_A(\mathrm{x}):\mathrm{X}\to[0,1]$ and $F_A(\mathrm{x}):\mathrm{X}\to[0,1]$. The sum of $T_A(\mathrm{x})$, $I_A(\mathrm{x})$ and $F_A(\mathrm{x})$ lies in $[0,3]$ that is $0\le T_A(\mathrm{x})+I_A(\mathrm{x})+F_A(\mathrm{x})\le 3$. For convenience, SVNS $A$ can be denoted by $\tilde{A}=\langle T_A(\mathrm{x}),I_A(\mathrm{x}),F_A(\mathrm{x})\rangle$ for all $x$ in $X$.

Now we mention some commonly used distance measures for two SNVS $A$ and $B$ on $X=\{x_1,x_2,\ldots,x_n\}$.

1. Normalized Hamming distance measure [14]:

$$D^N_{Ham}(A,B)=\frac{1}{3n}\sum_{i=1}^{n}\left(\left|T_A(x_i)-T_B(x_i)\right|+\left|I_A(x_i)-I_B(x_i)\right|+\left|F_A(x_i)-F_B(x_i)\right|\right) \quad (1)$$

2. Normalized Euclidean distance measure [14]:

$$D^N_{Euc}(A,B)=\sqrt{\frac{1}{3n}\sum_{i=1}^{n}\left(\left(T_A(x_i)-T_B(x_i)\right)^2+\left(I_A(x_i)-I_B(x_i)\right)^2+\left(F_A(x_i)-F_B(x_i)\right)^2\right)} \quad (2)$$

3. The Hausdroff metric [15]:

$$D^N_{Ham}(A,B)=\frac{1}{n}\sum_{i=1}^{n}\max\left\{\left|T_A(x_i)-T_B(x_i)\right|,\left|I_A(x_i)-I_B(x_i)\right|,\left|F_A(x_i)-F_B(x_i)\right|\right\} \quad (3)$$

### 2.2. Hesitant fuzzy sets

**Definition 2.** [4, 5, 16]

Let $X$ be a fixed set. A hesitant fuzzy set $A$ on $X$ is presented in terms of a function such that when applied to $X$ returns a subset of $[0,1]$, i.e.

$A=\{\langle x,h_A(x)\rangle \mid x\in X\}$, where $h_A(x)$ is a set of some different values in $[0,1]$, representing the possible membership degrees of the element $x\in X$ to $A$. For convenience, $h_A(x)$ is called a hesitant fuzzy element (HFE).

We have some well-known distance measures for two SNVS $A$ and $B$ on $X=\{x_1,x_2,\ldots,x_n\}$.

1. Generalized hesitant normalized distance:

$$D^N_G(A,B)=\left[\frac{1}{n}\sum_{i=1}^{n}\left(\frac{1}{l_{x_i}}\sum_{j=1}^{l_{x_i}}\left|h_A^{\sigma(j)}-h_B^{\sigma(j)}\right|^{\lambda}\right)\right]^{1/\lambda},\ \lambda>0 \quad (4)$$

2. Generalized hesitant normalized Hausdorff distance:

$$D^N_{Hau}(A,B)=\left[\frac{1}{n}\sum_{i=1}^{n}\max_j\left|h_A^{\sigma(j)}-h_B^{\sigma(j)}\right|^{\lambda}\right]^{1/\lambda},\lambda>0. \quad (5)$$

$l_{x_i}=\max\{l(h_A(x_i)),l(h_B(x_i))\}$ for each $x_i$ in $X$; $h_A^{\sigma(i)}(x_i)$ and $h_B^{\sigma(i)}(x_i)$ are the $j$th largest values in $h_A(x_i)$ and $h_B(x_i)$, respectively. $l(h_A(x_i))$ and $l(h_B(x_i))$ are the number of values in $h_A(x_i)$ and $h_B(x_i)$, respectively.

**Definition 3.** [16]

Let $A_1$, $A_2$ and $A_3$ be three HFSs on $X=\{\mathrm{x}_1,\mathrm{x}_2,\ldots,\mathrm{x}_n\}$, then the distance measure between $A_1$ and $A_2$ is defined as $d(A_1,A_2)$, which satisfies the following properties:

1. $0 \le d(A_1, A_2) \le 1$;
2. $d(A_1, A_2) = 0$ if and only if $A_1 = A_2$;
3. $d(A_1, A_2) = d(A_2, A_1)$;

The similarity measure between $A_1$ and $A_2$ is defined as $s(A_1, A_2)$, which satisfies the following properties:

1. $0 \le s(A_1, A_2) \le 1$;
2. $s(A_1, A_2) = 1$ if and only if $A_1 = A_2$;
3. $s(A_1, A_2) = s(A_2, A_1)$.

If $d(A_1, A_2)$ be the distance measure between two HFSs $A_1$ and $A_2$, then $s(A_1, A_2) = 1 - d(A_1, A_2)$ is the similarity measure between two HFSs $A_1$ and $A_2$. Similarly, if $s(A_1, A_2)$ be the similarity measure between two HFSs $A_1$ and $A_2$, then $d(A_1, A_2) = 1 - s(A_1, A_2)$ is the distance measure between two HFSs $A_1$ and $A_2$.

## 3. Distance measure of single valued neutrosophic sets

The neutrosophic set [3] theory pioneered by Smarandache has emerged as one of the research focus in many branches such as management sciences, engineering, applied mathematics. Neutrosophic set generalizes the concept of the crisp set, fuzzy set [1], interval valued fuzzy set [17], intuitionistic fuzzy set [2], and interval valued intuitionistic fuzzy set [18].

**Definition 4.** [11]

Let $X$ be a fixed set, then a single valued neutrosophic hesitant fuzzy set $N$ on $X$ is defined as follows:

$N = \{\langle x, t(x), i(x), f(x)\rangle \mid x \in X\}$, where, $t(x)$, $i(x)$, $f(x)$ are three sets of some values in $[0,1]$, denoting the respectively the possible truth, indeterminacy and falsity membership degrees of the element $x \in X$ to the set $N$. The membership degrees $t(x)$, $i(x)$ and $f(x)$ satisfy the following conditions:

$0 \le \delta, \gamma, \eta \le 1,\ 0 \le \delta + \gamma + \eta \le 3$

where, $\delta \in t(x),\ \gamma \in i(x),\ \eta \in f(x),\ \delta^+ \in t^+(x) = \bigcup_{\delta \in t(x)} \max t(x),\ \gamma^+ \in i^+(x) = \bigcup_{\gamma \in i(x)} \max i(x)$

and $\eta^+ \in f^+(x) = \bigcup_{\eta \in i(x)} \max f(x)$ for all $x \in X$.

For convenience of notation, the triple $n(x) = \langle t(x), i(x), f(x)\rangle$ is called a single valued neutrosophic hesitant fuzzy element (SVNHFE) and is denoted by $n = \langle t, i, f\rangle$. It is to be noted that the number of values for possible truth, indeterminacy and falsity membership degrees of the element in different SVNHFEs may be different.

**Definition 5.** [11]

Let $n_1 = \langle t_1, i_1, f_1\rangle$ and $n_2 = \langle t_2, i_2, f_2\rangle$ be two SVNHFEs, the following operational rules are defined as follows:

1. $n_1 \oplus n_2 = \left\langle \bigcup_{\delta_1 \in t_1, \gamma_1 \in i_1, \eta_1 \in f_1,\ \delta_2 \in t_2, \gamma_2 \in i_2, \eta_2 \in f_2} \{\{t_1 + t_2 - t_1 t_2\}, \{i_1 i_2\}, \{f_1, f_2\}\} \right\rangle$;

2. $n_1 \otimes n_2 = \left\langle \bigcup_{\delta_1 \in t_1, \gamma_1 \in i_1, \eta_1 \in f_1,\ \delta_2 \in t_2, \gamma_2 \in i_2, \eta_2 \in f_2} \{\{t_1 t_2\}, \{i_1 + i_2 - i_1 i_2\}, \{f_1 + f_2 - f_1 f_2\}\} \right\rangle;$

3. $\lambda n_1 = \left\langle \bigcup_{\delta_1 \in t_1, \gamma_1 \in i_1, \eta_1 \in f_1} \left\{\{1-(1-t_1)^\lambda\}, \{i_1^\lambda\}, \{f_1^\lambda\}\right\} \right\rangle, \lambda > 0$ ;

4. $n_1^\lambda = \left\langle \bigcup_{\delta_1 \in t_1, \gamma_1 \in i_1, \eta_1 \in f_1} \left\{\{t_1^\lambda\}, \{1-(1-i_1)^\lambda\}, \{1-(1-f_1)^\lambda\}\right\} \right\rangle, \lambda > 0.$

Motivating from the concept provided by Xu and Xia [16], we define a generalized single valued neutrosophic hesitant normalized distance:

$$D_G^N = \left[ \frac{1}{3n} \sum_{i=1}^{n} \left( \frac{1}{l_{x_i}} \left( \sum_{j=1}^{\# h_{x_i}} \left| \delta_A^{\sigma(j)}(x_i) - \delta_B^{\sigma(j)}(x_i) \right|^\lambda + \sum_{k=1}^{\# g_{x_i}} \left| \gamma_A^{\sigma(k)}(x_i) - \gamma_B^{\sigma(k)}(x_i) \right|^\lambda + \sum_{p=1}^{\# m_{x_i}} \left| \eta_A^{\sigma(p)}(x_i) - \eta_B^{\sigma(p)}(x_i) \right|^\lambda \right) \right) \right]^{1/\lambda}, \lambda > 0 \quad (6)$$

where $l_{x_i} = \# h_{x_i} + \# g_{x_i} + \# m_{x_i}$ ; $\# h_{x_i}$ , $\# g_{x_i}$ and $\# m_{x_i}$ are the number of elements $t$, $i$, and $f$, respectively.

If $\lambda = 1$, Eq.(9) reduces to single valued neutrosophic hesitant normalized Hamming distance:

$$D_{GHam}^N = \left[ \frac{1}{3n} \sum_{i=1}^{n} \left( \frac{1}{l_{x_i}} \left( \sum_{j=1}^{\# h_{x_i}} \left| \delta_A^{\sigma(j)}(x_i) - \delta_B^{\sigma(j)}(x_i) \right| + \sum_{k=1}^{\# g_{x_i}} \left| \gamma_A^{\sigma(k)}(x_i) - \gamma_B^{\sigma(k)}(x_i) \right| + \sum_{p=1}^{\# m_{x_i}} \left| \eta_A^{\sigma(p)}(x_i) - \eta_B^{\sigma(p)}(x_i) \right| \right) \right) \right] \quad (7)$$

If $\lambda = 2$, Eq.(9) reduces to single valued neutrosophic hesitant normalized Euclidean distance:

$$D_G^N = \left[ \frac{1}{3n} \sum_{i=1}^{n} \left( \frac{1}{l_{x_i}} \left( \sum_{j=1}^{\# h_{x_i}} \left| \delta_A^{\sigma(j)}(x_i) - \delta_B^{\sigma(j)}(x_i) \right|^2 + \sum_{k=1}^{\# g_{x_i}} \left| \gamma_A^{\sigma(k)}(x_i) - \gamma_B^{\sigma(k)}(x_i) \right|^2 + \sum_{p=1}^{\# m_{x_i}} \left| \eta_A^{\sigma(p)}(x_i) - \eta_B^{\sigma(p)}(x_i) \right|^2 \right) \right) \right]^{1/2} \quad (8)$$

However, if we consider, Hausdroff metric to the distance measure, then the generalized single valued neutrosophic hesitant normalized Hausdorff distance can be defined as follows:

$$D_{GHau}^N(A,B) = \left[ \frac{1}{n} \sum_{i=1}^{n} \max \left\{ \begin{array}{r} \max_j \left| \delta_A^{\sigma(j)}(x_i) - \delta_B^{\sigma(j)}(x_i) \right|^\lambda, \max_k \left| \gamma_A^{\sigma(k)}(x_i) - \gamma_B^{\sigma(k)}(x_i) \right|^\lambda, \\ \max_p \left| \eta_A^{\sigma(p)}(x_i) - \eta_B^{\sigma(p)}(x_i) \right|^\lambda \end{array} \right\} \right]^{1/\lambda}, \lambda > 0. \quad (9)$$

For $\lambda = 1$, Eq.(12) reduces to single valued neutrosophic hesitant normalized Hamming Hausdorff distance:

$$D_{GHau}^N(A,B) = \left[ \frac{1}{3n} \sum_{i=1}^{n} \max \left\{ \begin{array}{r} \max_j \left| \delta_A^{\sigma(j)}(x_i) - \delta_B^{\sigma(j)}(x_i) \right|, \max_k \left| \gamma_A^{\sigma(k)}(x_i) - \gamma_B^{\sigma(k)}(x_i) \right|, \\ \max_p \left| \eta_A^{\sigma(p)}(x_i) - \eta_B^{\sigma(p)}(x_i) \right| \end{array} \right\} \right]. \quad (10)$$

For $\lambda = 2$, Eq.(12) reduces to single valued neutrosophic hesitant normalized Euclidean Hausdorff distance:

$$D_{GHau}^N(A,B) = \left[ \frac{1}{3n} \sum_{i=1}^{n} \max \left\{ \begin{array}{r} \max_j \left| \delta_A^{\sigma(j)}(x_i) - \delta_B^{\sigma(j)}(x_i) \right|^2, \max_k \left| \gamma_A^{\sigma(k)}(x_i) - \gamma_B^{\sigma(k)}(x_i) \right|^2, \\ \max_p \left| \eta_A^{\sigma(p)}(x_i) - \eta_B^{\sigma(p)}(x_i) \right|^2 \end{array} \right\} \right]^{1/2}. \quad (11)$$

In some situations, we need weight of each element $x_i \in X$ , and then we present the following weighted distance measures for SVNHFs. Assume that the weight of the element $x_i \in X$ is $w_i (i = 1, 2, ..., n)$ with $w_i \in [0,1]$ and $\sum_{i=1}^{n} w_i = 1$, then we have a generalized SVNHF weighted distance:

$$D_G^N = \left[ \frac{1}{3} \sum_{i=1}^{n} \left( \frac{w_i}{l_{x_i}} \left( \sum_{j=1}^{\# h_{x_i}} \left| \delta_A^{\sigma(j)}(x_i) - \delta_B^{\sigma(j)}(x_i) \right|^\lambda + \sum_{k=1}^{\# g_{x_i}} \left| \gamma_A^{\sigma(k)}(x_i) - \gamma_B^{\sigma(k)}(x_i) \right|^\lambda + \sum_{p=1}^{\# m_{x_i}} \left| \eta_A^{\sigma(p)}(x_i) - \eta_B^{\sigma(p)}(x_i) \right|^\lambda \right) \right) \right]^{1/\lambda}, \lambda > 0 \quad (12)$$

and a generalized SVNH weighed Hausdroff distance:

$$D_{GHau}^{N}(A,B)=\left[\frac{1}{3}\sum_{i=1}^{n}\max\left\{\begin{matrix}\max_{j} w_i\left|\delta_A^{\sigma(j)}(x_i)-\delta_B^{\sigma(j)}(x_i)\right|^{\lambda},\max_{k} w_i\left|\gamma_A^{\sigma(k)}(x_i)-\gamma_B^{\sigma(k)}(x_i)\right|^{\lambda},\\ \max_{p} w_i\left|\eta_A^{\sigma(p)}(x_i)-\eta_B^{\sigma(p)}(x_i)\right|^{\lambda}\end{matrix}\right\}\right]^{1/\lambda},\lambda>0. \quad (13)$$

# 4. Application of proposed distance measure in multi-attribute decision making

In this section we use the proposed distance measures to find out the best alternative in multi-attribute decision making with single valued neutrosophic hesitant fuzzy environment.

For a multi-attribute decision making problem, assume that $A=\{A_1,A_2,...,A_m\}$ be the set of m alternatives, $C=\{C_1,C_2,...,C_n\}$ be the set of n attributes, whose weight vector $w=(w_1,w_2,...,w_n)^T$, satisfies $w_j>0(j=1,2,...,n)$ and $\sum_{j=1}^{n}w_j=1,$ where $w_j$ denotes the weight of the attribute $C_j$. The performance of the alternative $A_i$ with respect to the attribute $C_j$ is measured by an SVNHFE $n_{ij}=\{t_{ij},i_{ij},f_{ij}\}$, where $t_{ij}=\{\delta_{ij}\,|\,\delta_{ij}\in t_{ij},0\le\delta_{ij}\le1\}$, $i_{ij}=\{\gamma_{ij}\,|\,\gamma_{ij}\in i_{ij},0\le\gamma_{ij}\le1\}$, and $f_{ij}=\{\eta_{ij}\,|\,\eta_{ij}\in f_{ij},0\le\eta_{ij}\le1\}$ are the possible truth, indeterminacy and falsity membership degree, respectively such that $0\le\delta_{ij}^{+}+\gamma_{ij}^{+}+\eta_{ij}^{+}\le3,$ where, $\delta_{ij}^{+}=\bigcup_{\delta_{ij}\in t_{ij}}\max\{\delta_{ij}\}$, $\gamma_{ij}^{+}=\bigcup_{\gamma_{ij}\in i_{ij}}\max\{\gamma_{ij}\}$, and $\eta_{ij}^{+}=\bigcup_{\eta_{ij}\in f_{ij}}\max\{\eta_{ij}\}$.

All $n_{ij}=\{t_{ij},i_{ij},f_{ij}\}(i=1,2,...,m;j=1,2,...,n)$ are contained in SVNHF decision matrix $N=(n_{ij})_{m\times n}$ (See Table 1.)

Table 1. SVNHF decision matrix

| | $C_1$ | $C_1$ | ... | $C_n$ |
|---|---|---|---|---|
| $A_1$ | $n_{11}$ | $n_{11}$ | ... | $n_{11}$ |
| $A_1$ | $n_{11}$ | $n_{11}$ | ... | $n_{11}$ |
| $\vdots$ | $\vdots$ | $\vdots$ | $\ddots$ | $\vdots$ |
| $A_m$ | $n_{m1}$ | $n_{m1}$ | ... | $n_{m1}$ |

Basically attributes are two types:

1. benefit type attributes,
2. cost type attributes.

In such cases, we propose the rating values of ideal alternatives $A_j^*$ as $n_j^*=\{t_j^*,i_j^*,f_j^*\}$ for $j=1,2,...,n$, where,

$n_j^*=\{\{1\},\{0\},\{0\}\}$ for benefit type attributes and $n_j^*=\{\{0\},\{1\},\{1\}\}$ for cost type attributes.

Then to determine the best alternatives, we propose the following steps:

**Step 1.** Determine the distance between an alternative $A_j(j=1,2,...,n)$ and the ideal alternative $A^*$ using proposed distance measure according to the nature of attributes.

**Step 2.** Rank the alternative on the basis of distance measure values.

**Step 3.** Obtain the best alternative according to the minimum value of distance measure.

# 5. Numerical example

In this section we consider the example adopted from Ye [11] to illustrate the application of the proposed GRA method for MADM proposed in Section 4. Consider an investment company that wants to invest a sum of money in the best option. There is a panel with four possible alternatives: (1) $A_1$ is the car company; (2) $A_2$ is the food company; (3) $A_3$ is the computer company; (4) $A_4$ is the arms company. To take a decision, the investment company consider three attributes: (1) $C_1$ is the risk analysis; (2) $C_2$ is the growth analysis; (3) $C_3$ is the environmental impact analysis.

The attribute weight vector is given as $W=(0.35,0.25.0.40)^T$ . The four possible alternatives $\{A_1,A_2,A_3,A_4\}$ are evaluated by using SVNHFEs under three attributes $C_j(j=1,2,3)$. We can arrange the rating values in a matrix form namely a SVNHF decision matrix $X=(x_{ij})_{4\times 3}$ that is shown in Table-1.

Table 1. *Single valued neutrosophic hesitant fuzzy decision matrix*

| $C_1$ | $C_2$ | $C_3$ |
|---|---|---|
| $\{\{0.3,0.4,0.5\},\{0.1\},\{0.3,0.4\}\}$ | $\{\{0.5,0.6\},\{0.2,0.3\},\{0.3,0.4\}\}$ | $\{\{0.3,0.4,0.5\},\{0.1\},\{0.3,0.4\}\}$ |
| $\{\{0.6,0.7\},\{0.1,0.2\},\{0.2,0.3\}\}$ | $\{\{0.6,0.7\},\{0.1\},\{0.3\}\}$ | $\{\{0.3,0.4,0.5\},\{0.1\},\{0.3,0.4\}\}$ |
| $\{\{0.5,0.6\},\{0.4\},\{0.2,0.3\}\}$ | $\{\{0.6\},\{0.3\},\{0.4\}\}$ | $\{\{0.5,0.6\},\{0.1\},\{0.3\}\}$ |
| $\{\{0.7,0.8\},\{0.1\},\{0.1,0.2\}\}$ | $\{\{0.6,0.7\},\{0.1\},\{0.2\}\}$ | $\{\{0.3,0.5\},\{0.2\},\{0.1,0.2,0.3\}\}$ |

Now we consider the following steps, described in Section-4, to find the best alternatives.

**Step 1.**Using Eq.(14), we calculate the SVNH weighted distance measure between alternatives $A_i(i=1,2,3,4)$ and ideal alternative $A^*$ for $\lambda=1,2,5,10$ which are shown in Table 2.:

Table 2. *Weighted distant measures for different values of* $\lambda$ *'s*

| $\lambda$ | $A_1$ | $A_2$ | $A_3$ | $A_4$ | **Ranking** |
|---|---|---|---|---|---|
| $\lambda$ **=1** | 0.1361 | 0.0810 | 0.1089 | 0.0816 | $A_2 \succ A_4 \succ A_3 \succ A_1$ |
| $\lambda$ **=2** | 0.2687 | 0.1531 | 0.2065 | 0.1738 | $A_2 \succ A_4 \succ A_3 \succ A_1$ |
| $\lambda$ **=5** | 0.4487 | 0.2469 | 0.3192 | 0.3462 | $A_2 \succ A_3 \succ A_4 \succ A_1$ |
| $\lambda$ **=10** | 0.5677 | 0.3059 | 0.3841 | 0.4802 | $A_2 \succ A_3 \succ A_4 \succ A_1$ |

**Step 2.**Rank the alternative according to the value of SVNH weighted distance measure.

**Step 3.** Based on the minimum value of SVNH weighted distance measures for different values of $\lambda=1,2,5,10$, we conclude that $A_2$ as the best alternative, which is same as the results obtained in Ye [11] and Sahin and Liu [12].

From Table-2, we observe that ranking results change with different values of $\lambda$.Therefore taking the values of $\lambda$ according to decision maker's preference play a crucial role in ranking process. Ye [11] considered weighted cosine similarity measure of SVNHFs and Sahin and Liu [12] proposed weighted correlation co-efficient to determine the ranking order of alternatives. In both studies, we see that there is no option to consider different values of the attitudinal character $\lambda$ that can change the ranking result as we have seen in our study. Thus our method is more realistic and flexible over these two methods, furthermore our method is simple and effective.

## 6. Conclusion

In this paper, we develop a class of distance measures for single-valued neutrosophic hesitant fuzzy sets and discussed their properties with variational parameters. We apply the weighted distance measures to calculate the distances between each alternative and ideal alternative in the MADM problems. With these distance values, we obtain the ranking order of alternatives for selecting the best one. We provide an illustrative example to verify the proposed approach and to show its fruitfulness. Finally, we compared the proposed method with other existing methods for solving MADM under SVNHF environment. The proposed method is simple and effective to handle MADM under SVNHF. We hope that the proposed distance measures can be extended to interval neutrosophic hesitant fuzzy set, and can be applied in medical diagnosis, pattern recognition, and personal selection under neutrosophic hesitant fuzzy environment.

## References

1. L.A. Zadeh, Fuzzy sets, Information Control, 8(1965) 338–353.
2. K.T. Atanassov, Intuitionistic fuzzy sets, Fuzzy Sets and Systems 20(1986) 87–96.
3. F. Smarandache, A unifying field in logics, neutrosophy: neutrosophic probability, set and logic. American Research Press, Rehoboth, 1998.
4. V. Torra, Y. Narukawa, On hesitant fuzzy sets and decision in: The 18th IEEE International Conference on Fuzzy Systems, Jeju Island, Korea, 2009 1378-1382.
5. V. Torra, Hesitant fuzzy sets, International Journal of Intelligent Systems 25(2010) 529-539.
6. M.M. Xia, Z.S. Xu, Hesitant fuzzy information aggregation in decision making, International Journal of Approximate Reasoning 52(2011) 395-407.
7. D.H. Peng, C.Y. Gao, Z. F. Gao, Generalized hesitant fuzzy synergetic weighted distance measures and their application to multiple criteria decision-making, Applied Mathematical Modelling 37(8)(2013) 5837-5850.
8. L. Wang, S. Xu, Q. Wang, M. Ni. Distance and similarity measures of dual hesitant fuzzy sets with their applications to multiple attribute decision making. In Progress in Informatics and Computing (PIC), 2014 International Conference on 2014 May 16 (88-92). IEEE.
9. B. Zhu, Z.S. Xu, M.M. Xia, Dual hesitant fuzzy sets, Journal of Applied Mathematics (2012) doi: 10.1155/2012/879629.
10. P. Singh, Distance and similarity measures for multiple-attribute decision making with dual hesitant fuzzy sets. Computational and Applied Mathematics (2013) doi: 10.1007/s40314-015-0219-2.
11. J. Ye, Multiple-attribute decision making under a single-valued neutrosophic hesitant fuzzy environment, Journal of Intelligent Systems (2014) doi: 10.1515/jisys-2014-0001.
12. R. Sahin, P Liu, Correlation coefficient of single-valued neutrosophic hesitant fuzzy sets and its applications in decision making, Neural Computing and Applications (2016) doi: 10.1007/s00521-015-2163-x.
13. H.Wang, F. Smarandache, R. Sunderraman, Y.Q. Zhang, Single-valued neutrosophic sets, Multi space and Multi structure. 4(2010) 410–413.
14. P. Majumdar, S.K. Samanta, On similarity and entropy of neutrosophic sets, Journal of Intelligent and fuzzy Systems, 26(3)(2014) 1245-1252.
15. S. Broumi,F. Smarandache, Several similarity measures of neutrosophic sets. Neutrosophic Sets and Systems, 1(1)(2013), 54-62.
16. Z. Xu,M. Xia, Distance and similarity measures for hesitant fuzzy sets. Information Sciences, 181(11)(2011), 2128-2138.
17. M.B. Gorzalczany, A method of inference in approximate reasoning based on interval-valued fuzzy sets, Fuzzy Sets and Systems 21 (1987) 1–17.
18. K. Atanassov, G. Gargov, Interval valued intuitionistic fuzzy sets. Fuzzy sets and systems, 31(3) (1989), 343-349.

**RIDVAN ŞAHIN**[1], **PEIDE LIU**[2]

[1] Faculty of Education, Bayburt University, Bayburt, 69000, Turkey. E-mail: mat.ridone@gmail.com
[2] School of Management Science and Engineering, Shandong University of Finance and Economics, Jinan 250014, Shandong, China. E-mail:peide.liu@gmail.com

# Distance and Similarity Measures for Multiple Attribute Decision Making with Single-Valued Neutrosophic Hesitant Fuzzy Information

## Abstract

With respect to a combination of hesitant sets, and single-valued neutrosophic sets which are a special case of neutrosophic sets, the single valued neutrosophic hesitant sets (SVNHFS) have been proposed as a new theory set that allows the truth-membership degree, indeterminacy membership degree and falsity-membership degree including a collection of crisp values between zero and one, respectively. There is no consensus on the best way to determine the order of a sequence of single-valued neutrosophic hesitant fuzzy elements. In this paper, we first develop an axiomatic system of distance and similarity measures between single-valued neutrosophic hesitant fuzzy sets and also propose a class of distance and similarity measures based on three basic forms such that the geometric distance model, the set-theoretic approach, and the matching functions. Then we utilize the distance measure between each alternative and ideal alternative to establish a multiple attribute decision making method under single-valued neutrosophic hesitant fuzzy environment. Finally, a numerical example of investment alternatives is provided to show the effectiveness and usefulness of the proposed approach. The advantages of the proposed distance measure over existing measures have been discussed.

## 1. Introduction

Most of our traditional tools for formal modeling, reasoning and computing are crisp, deterministic and precise in character. However, there are many complicated problems in economics, engineering, environment, social science, medical science, etc., that involve data which are not always all crisp. Classical methods cannot successfully handle uncertainty, because the uncertainties appearing in these domains may be of various types. Zadeh (1965) introduced fuzzy sets (FS) and applied them in many fields including uncertainty. As a generalization of the fuzzy sets, Atanassov (1986) introduced the concept of intuitionistic fuzzy set (IFS). Then Atanassov and Gargov (1989) extended the concept of IFS to interval-valued intuitionistic fuzzy set (IVIFS). Current literature has very large number of distance and similarity measures for FSs and IFSs (Atanassov 1986; Xuecheng 1992; Chen et al. 1995; Liu et al. 2015; Farhadinia 2014; Wang 1997; Szmidt and Kacprzyk 2000; Khaleie and Fasanghari 2012; Grzegorzewski 2004; Wang and Xin 2005; Xu 2007; Hung and Yang 2007; Li 2007; Şahin 2015; Tan 2011).

Torra and Narukawa (2009) and Torra (2010) proposed the concept of hesitant fuzzy set (HFS), discussed the relationship between hesitant fuzzy set and intuitionistic fuzzy set and showed that the envelope of hesitant fuzzy set is an intuitionistic fuzzy set. The membership degree of an element in hesitant fuzzy set includes a set of possible values between zero and one. Since its appearance, the hesitant fuzzy information has been used to solve multiple attribute decision making problems. Xia and Xu (2011) defined some techniques for aggregating hesitant fuzzy information and utilized their performances in decision making. Based on the relationship between HFS and IFS, they proposed the set-theoretic laws of HFSs. Xu and Xia (2011) defined a collection of distance measures for HFSs and generated the similarity measures associated with the proposed distance measures.

Furthermore, Zhu et al. (2012) introduced dual hesitant fuzzy set (DHFS) as a generalization of FSs, IFSs, HFSs, and fuzzy multisets (FMSs) and presented some basic operations of DHFSs. A DHFS are characterized by two class of possible values, the membership degrees and nonmembership degrees. Therefore, DHFSs include FSs, IFSs, HFSs, and FMSs under certain conditions, and so they have the desirable performances and advantages of its own and appear to be a more favorable method than aforementioned sets because of considering much more information given by decision makers. Singh (2013) introduced a comprehensive family of distance measures and related similarity measures for DHFSs.

As a new branch of philosophy that combines the knowledge of logics, philosophy, set theory, and probability, Smarandache (1999, 2005) proposed the concept of neutrosophic sets (NSs) as a further generalization of uncertainty modeling tools. Unlike the aforementioned sets, a neutrosophic set consists of three membership functions such that the truth-membership function, the indeterminacy-membership function and the falsity membership function. Additionally, the uncertainty presented here, i.e. the indeterminacy factor, is independent on the truth and falsity values, whereas the incorporated uncertainty is dependent on the degrees of belongingness and non-belongingness of existing sets. The structure of NSs is not appropriate to apply to real-life situations. Therefore, Wang et al. (2005, 2010) developed single-valued neutrosophic sets (SVNSs) and interval neutrosophic sets (INSs), which are an extension of NSs. Şahin (2014) proposed a neutrosophic hierarchical clustering algorithm based on relationship between SVNSs. Şahin and Küçük (2014) defined a subsethood measure for SVNSs and applied it in a decision making problem. The correlation coefficients of SVNSs as well as a decision-making method using SVNSs were proposed by Ye (2013). In addition, Ye (2014b) investigated the concept of simplified neutrosophic sets (SNSs), which can be expressed by three real numbers in the real unit interval $[0,1]$, provided the set-theoretic operators of SNSs, and developed a multi criteria decision making (MCDM) method based on the aggregation operators of SNSs. But, Peng et al. (2015) showed that some operations of Ye (2014b) may also be unrealistic in special cases, and defined the novel operations and aggregation operators and applied them to MCDM problems. Also, Ye (2014a) proposed the single valued neutrosophic cross-entropy for solving multicriteria decision making (MCDM) problems with single valued neutrosophic information. Broumi and Smarandache (2013) extended the correlation coefficient to INSs. Zhang et al. (2014) developed a MCDM method based on aggregation operators within an interval neutrosophic environment. Furthermore, Majumdar and Samanta (2014) proposed the distance and similarity measures between SVNSs. Ye (2014d) extended these measures to INSs as based on the relationship between similarity measures and distances. Liu and Wang (2014) discussed a single-valued neutrosophic normalized weighted Bonferroni mean (SVNNWBM) operator based on Bonferroni mean, the weighted Bonferroni mean (WBM), and the normalized WBM. Peng et al. (2015) introduced the multi-valued neutrosophic sets (MVNSs) and developed the operations of multi-valued neutrosophic numbers (MVNNs) based on Einstein operations.

Recently, Ye (2015c) proposed the concept of single valued neutrosophic hesitant fuzzy set (SVNHFS) as a generalization of FSs, IFSs, HFSs, FMSs, and also SVNSs and discussed the basic operations and properties of SVNHFSs. SVNHFSs consist of three parts, first is the truth-membership hesitancy function, second is the indeterminacy-membership hesitancy function, and third is the falsity-membership hesitancy function. The current sets, including FSs, IFSs, HFSs, FMSs, and SVNSs can be regarded as special cases of SVNHFSs. In a SVNHFS, the truth-membership hesitancy degrees, indeterminacy-membership hesitancy degrees and falsity-membership hesitancy degrees are represented by three sets of possible values between zero and one, respectively. Therefore, it is not only more general than aforementioned set but only more suitable for solving MADM problems due to considering much more information provided by decision makers.

From above analysis, we cannot utilize the current measures for dealing with distance and similarity measure between SVNHFSs. Therefore, we need to develop new distance and similarity measures for SVNHFSs, because a SVNHFS consists of three basic membership function such that the truth-membership hesitancy function and indeterminacy-membership hesitancy function and falsity-membership hesitancy function. In this paper, we first define a compressive class of distance measures between SVNHFSs and then proposed the similarity measures based on the geometric distance model, the set-theoretic approach and the matching functions. Also, we show that the proposed measures satisfies the axiom definition of distance and similarity measures developed for SVNHFSs. Finally, we utilize the proposed distance measure to solve a MADM problem with single valued neutrosophic hesitant fuzzy information. The rest of this paper is organized as follows. In section 2, we introduce some basic concepts related to HFS, SVNS and SVNHFS, and some operational and theoretical laws. In Section 3, we propose a variety class of distance measures of SVNHFSs as a further generalization of the existing distance measure for HFSs, DHFSs, IFSs, and SVNSs. Based on the geometric distance model, the set-theoretic approach and the matching functions, we present some similarity measures between SVNHFSs. Section 4 develops a MADM method with single valued neutrosophic hesitant fuzzy information based on the proposed distance measure for SVNHFSs. In Section 6, an illustrative example is provided to demonstrate the application and effectiveness of the developed method. Section 7 gives related comparative analysis. Finally, conclusions and future work are given in Section 8.

## 2. Preliminaries

In this subsection, we give some concepts related to NSs and SVNSs.

### 2.1 Neutrosophic set

**Definition 1.** (Smarandache 2005) Let $X$ be a universe of discourse, then a neutrosophic set is defined as:

$$A = \{\langle x, T_A(x), I_A(x), F_A(x)\rangle : x \in X\}, \tag{1}$$

which is characterized by a truth-membership function $T_A: X \to ]0^-, 1^+[$, an indeterminacy-membership function $I_A: X \to ]0^-, 1^+[$and a falsity-membership function $F_A: X \to ]0^-, 1^+[$.

There is not restriction on the sum of $T_A(x)$, $I_A(x)$ and $F_A(x)$, so $0^- \leq \sup T_A(x) + \sup I_A(x) + \sup F_A(x) \leq 3^+$.

In the following, we adopt the representations $t_A(x)$, $\imath_A(x)$ and $f_A(x)$ instead of $T_A(x)$, $I_A(x)$ and $F_A(x)$, respectively.

Wang et al. (2010) defined the single valued neutrosophic set which is an instance of neutrosophic set.

*2.2.Single valued neutrosophic sets*

**Definition 2.** Wang et al. (2010) Let $X$ be a universe of discourse, then a single valued neutrosophic set is defined as:

$$A = \{\langle x, t_A(x), \imath_A(x), f_A(x)\rangle : x \in X\}, \tag{2}$$

where $t_A: X \to [0,1]$, $\imath_A: X \to [0,1]$ and $f_A: X \to [0,1]$ with $0 \leq t_A(x) + \imath_A(x) + f_A(x) \leq 3$ for all $x \in X$. The values $t_A(x), \imath(x)$ and $f_A(x)$ denote the truth-membership degree, the indeterminacy-membership degree and the falsity membership degree of $x$ to $A$, respectively.

*2.3.Hesitant fuzzy sets*

**Definition 3.** (Torra 2010) A hesitant fuzzy set $M$ on $X$ is defined in terms of a function $h_M$ when applied to $X$, which returns a finite subset of [0,1], i.e.,

$$M = \{\langle x, h_M(x)\rangle : x \in X\}, \tag{3}$$

where $h_M(x)$ is a set of some different values in $[0,1]$, representing the possible membership degrees of the element $x \in X$ to $M$.

*2.4.Single-valued neutrosophic hesitant sets*

**Definition 4.** (Ye 2014c) Let $X$ be a fixed set, then a single-valued neutrosophic hesitant fuzzy set $A$ on $X$ is defined as,

$$A = \left\{\langle x, \left(\tilde{t}_A(x), \tilde{\imath}_A(x), \tilde{f}_A(x)\right)\rangle : x \in X\right\} \tag{4}$$

in which $\tilde{t}_A(x), \tilde{\imath}_A(x)$, and $\tilde{f}_A(x)$ are three sets of some different values in [0,1], denoting the truth-membership hesitant degrees, indeterminacy-membership hesitant degrees, and falsity-membership hesitant degrees of the element $x \in X$ to $A$, respectively, with the conditions $0 \leq \gamma, \delta, \eta \leq 1$ and $0 \leq \gamma^+ + \delta^+ + \eta^+ \leq 3$, where $\gamma \in \tilde{t}_A(x)$, $\delta \in \tilde{\imath}(x)$, $\eta \in \tilde{f}_A(x)$, $\gamma^+ \in \tilde{t}_A^+(x) =$ $\cup_{\gamma\in\tilde{t}_A(x)}\max\{\gamma\}$, $\delta^+ \in \tilde{\iota}_A^+(x) = \cup_{\delta\in\tilde{\iota}_A(x)}\max\{\delta\}$, and $\eta^+ \in \tilde{f}_A^+(x) = \cup_{\eta\in\tilde{f}_A(x)}\max\{\eta\}$ for $x \in X$.

For convenience, the three tuple $A = \{ (\tilde{t}_A(x), \tilde{\iota}_A(x), \tilde{f}_A(x))\}$ is called a single-valued neutrosophic hesitant fuzzy element (SVNHFE) or a triple hesitant fuzzy element, which is denoted by the simplified symbol $A = \{ (\tilde{t}_A, \tilde{\iota}_A, \tilde{f}_A)\}$.

Now, we give the following definitions to propose the distance and similarity measures between SVNHFSs.

**Definition 5** Let $A$, $B$ and $C$ be three SVNHSs on $X = \{x_1, x_2, \ldots, x_n\}$, then the distance measure between $A$ and $B$ is defined as $\tilde{d}(A, B)$, which satisfies the following properties:

(1) $0 \leq \tilde{d}(A,B) \leq 1$;
(2) $\tilde{d}(A,B) = 0$ if and only if $A = B$;
(3) $\tilde{d}(A,B) = \tilde{d}(B,A)$.
(4) $\tilde{d}(A,B) \leq \tilde{d}(A,C)$ and $\tilde{d}(B,C) \leq \tilde{d}(A,C)$, if $A \subseteq B \subseteq C$.

**Definition 6**. Let $A$, $B$ and $C$ be three SVNHSs on $X = \{x_1, x_2, \ldots, x_n\}$, then the similarity measure between $A$ and $B$ is defined as $\tilde{s}(A, B)$, which satisfies the following properties:

(1) $0 \leq \tilde{s}(A,B) \leq 1$;
(2) $\tilde{s}(A,B) = 1$ if and only if $A = B$;
(3) $\tilde{s}(A,B) = \tilde{s}(B,A)$.
(4) $\tilde{s}(A,B) \geq \tilde{s}(A,C)$ and $\tilde{s}(B,C) \geq \tilde{s}(A,C)$, if $A \subseteq B \subseteq C$.

From Definitions 5 and 6, it is noted that $\tilde{s}(A,B) = 1 - \tilde{d}(A,B)$.
Similar to HFS, in most of the cases, the number of values in different SVNHFEs might be different, i.e., $l_{\tilde{t}_A}(x_i) \neq l_{\tilde{t}_B}(x_i)$ , $l_{\tilde{\iota}_A}(x_i) \neq l_{\tilde{\iota}_B}(x_i)$ and $l_{\tilde{f}_A}(x_i) \neq l_{\tilde{f}_B}(x_i)$ . Let $l_{\tilde{t}}(x_i) = \max\{l_{\tilde{t}_A}(x_i), l_{\tilde{t}_B}(x_i)\}$, $l_{\tilde{\iota}}(x_i) = \max\{l_{\tilde{\iota}_A}(x_i), l_{\tilde{\iota}_B}(x_i)\}$ and $l_{\tilde{f}}(x_i) = \max\{l_{\tilde{f}_A}(x_i), l_{\tilde{f}_B}(x_i)\}$ for each $x_i \in X$. We can make them have the same number of elements through adding some elements to the SVNHFE which has less number of elements. The selection of this operation mainly depends on the decision makers' risk preferences. Pessimists expect unfavorable outcomes and may add the minimum of the truth-membership degree and maximum value of indeterminacy-membership degree and falsity-membership degree. Optimists anticipate desirable outcomes and may add the maximum of the truth-membership degree and minimum value of indeterminacy-membership degree and falsity-membership degree That is, according to the pessimistic principle, if $l_{\tilde{t}_A}(x_i) < l_{\tilde{t}_B}(x_i)$, then the least value of $\tilde{t}_A(x_i)$ or $\tilde{t}_B(x_i)$ will be added to $\tilde{t}_A(x_i)$. Moreover, if $l_{\tilde{\iota}_A}(x_i) < l_{\tilde{\iota}_B}(x_i)$, then the largest value of $l_{\tilde{t}_A}(x_i)$ or $l_{\tilde{t}_B}(x_i)$ will be inserted in $\tilde{\iota}_A(x_i)$ for $x_i \in X$. Similarity, if $l_{\tilde{f}_A}(x_i) < l_{\tilde{f}_B}(x_i)$, then the largest value of $l_{\tilde{f}_A}(x_i)$ or $l_{\tilde{f}_B}(x_i)$ will be inserted in $\tilde{f}_A(x_i)$ for $x_i \in X$.

## 3. Some distance measures for SVNHFSs

In this section, we give some distance measures between two SVNHFSs.

Based on the geometric distance model for SVNHFSs, we define the following distance measures.

(1) Generalized single valued neutrosophic hesitant normalized distance (GN), for $\lambda > 0$;

$$\tilde{d}_{GN} = \left( \frac{1}{3n} \sum_{i=1}^{n} \left( \frac{1}{l_{\tilde{t}}(x_i)} \sum_{j=1}^{l_{\tilde{t}}(x_i)} \left| \tilde{t}_A^{\sigma(j)}(x_i) - \tilde{t}_B^{\sigma(j)}(x_i) \right|^{\lambda} + \frac{1}{l_{\tilde{\iota}}(x_i)} \sum_{j=1}^{l_{\tilde{\iota}}(x_i)} \left| \tilde{\iota}_A^{\sigma(j)}(x_i) - \tilde{\iota}_B^{\sigma(j)}(x_i) \right|^{\lambda} + \frac{1}{l_{\tilde{f}}(x_i)} \sum_{j=1}^{l_{\tilde{f}}(x_i)} \left| \tilde{f}_A^{\sigma(j)}(x_i) - \tilde{f}_B^{\sigma(j)}(x_i) \right|^{\lambda} \right) \right)^{\frac{1}{\lambda}}, \quad (5)$$

where $\tilde{t}_A^{\sigma(j)}(x_i), \tilde{t}_B^{\sigma(j)}(x_i)$; $\tilde{\iota}_A^{\sigma(j)}(x_i), \tilde{\iota}_B^{\sigma(j)}(x_i)$ and $\tilde{f}_A^{\sigma(j)}(x_i), \tilde{f}_B^{\sigma(j)}(x_i)$ are the $j$th largest values of truth-membership hesitant degrees, indeterminacy-membership hesitant degrees, and falsity-membership hesitant degrees of $A$ and $B$, respectively.

i. If λ = 1, Eq. (5) reduces a single valued neutrosophic hesitant normalized Hamming distance (NH):

$$\tilde{d}_{NH} = \frac{1}{3n} \sum_{i=1}^{n} \left( \frac{1}{l_{\tilde{t}}(x_i)} \sum_{j=1}^{l_{\tilde{t}}(x_i)} \left| \tilde{t}_A^{\sigma(j)}(x_i) - \tilde{t}_B^{\sigma(j)}(x_i) \right| + \frac{1}{l_{\tilde{\iota}}(x_i)} \sum_{j=1}^{l_{\tilde{\iota}}(x_i)} \left| \tilde{\iota}_A^{\sigma(j)}(x_i) - \tilde{\iota}_B^{\sigma(j)}(x_i) \right| + \frac{1}{l_{\tilde{f}}(x_i)} \sum_{j=1}^{l_{\tilde{f}}(x_i)} \left| \tilde{f}_A^{\sigma(j)}(x_i) - \tilde{f}_B^{\sigma(j)}(x_i) \right| \right). \quad (6)$$

ii. If λ = 2, Eq. (5) reduces a single valued neutrosophic hesitant normalized Euclidean distance (NE)

$$\tilde{d}_{NE} = \left( \frac{1}{3n} \sum_{i=1}^{n} \left( \frac{1}{l_{\tilde{t}}(x_i)} \sum_{j=1}^{l_{\tilde{t}}(x_i)} \left| \tilde{t}_A^{\sigma(j)}(x_i) - \tilde{t}_B^{\sigma(j)}(x_i) \right|^{2} + \frac{1}{l_{\tilde{\iota}}(x_i)} \sum_{j=1}^{l_{\tilde{\iota}}(x_i)} \left| \tilde{\iota}_A^{\sigma(j)}(x_i) - \tilde{\iota}_B^{\sigma(j)}(x_i) \right|^{2} + \frac{1}{l_{\tilde{f}}(x_i)} \sum_{j=1}^{l_{\tilde{f}}(x_i)} \left| \tilde{f}_A^{\sigma(j)}(x_i) - \tilde{f}_B^{\sigma(j)}(x_i) \right|^{2} \right) \right)^{\frac{1}{2}}. \quad (7)$$

Equation (5) can be viewed as a most generalized case of distance measures. We can see that if there is no indeterminacy in SVNHFS, then the indeterminacy-membership value of SVNHFS will disappear, hence, Eqs. (5), (6), and (7) are reduced to a generalized dual hesitant normalized distance, a dual hesitant normalized Hamming distance and a dual hesitant normalized Euclidean distance, respectively (i.e., the distance measures proposed by Singh 2013). In addition, if there is no both indeterminacy and nonmembership in SVNHFS, then both indeterminacy-membership value and falsity-membership value of SVNHFS will disappear, hence, Eqs. (5), (6), and (7) are reduced to a generalized hesitant normalized distance, a hesitant normalized Hamming distance and a hesitant normalized Euclidean distance, respectively (i.e., the distance measure proposed by Xu and Xia 2011).

If we apply the Hausdorff metric to the distance measure, we obtain that

(2)Generalized single valued neutrosophic hesitant normalized Hausdorff distance (GNH):

$$\tilde{d}_{GNH} = \left(\frac{1}{3n}\sum_{i=1}^{n}\max_j\left(\left|\tilde{t}_A^{\sigma(j)}(x_i) - \tilde{t}_B^{\sigma(j)}(x_i)\right|^{\lambda}, \left|\tilde{\imath}_A^{\sigma(j)}(x_i) - \tilde{\imath}_B^{\sigma(j)}(x_i)\right|^{\lambda}, \left|\tilde{f}_A^{\sigma(j)}(x_i) - \tilde{f}_B^{\sigma(j)}(x_i)\right|^{\lambda}\right)\right)^{\frac{1}{\lambda}}. \quad (8)$$

i. If $\lambda = 1$, Eq. (6) reduces a single valued neutrosophic hesitant normalized Hamming–Hausdorff distance (NHH):

$$\tilde{d}_{NHH} = \left(\frac{1}{3n}\sum_{i=1}^{n}\max_j\left(\left|\tilde{t}_A^{\sigma(j)}(x_i) - \tilde{t}_B^{\sigma(j)}(x_i)\right|, \left|\tilde{\imath}_A^{\sigma(j)}(x_i) - \tilde{\imath}_B^{\sigma(j)}(x_i)\right|, \left|\tilde{f}_A^{\sigma(j)}(x_i) - f_B^{\sigma(j)}(x_i)\right|\right)\right). \quad (9)$$

ii. If $\lambda = 2$, Eq. (6) reduces a single valued neutrosophic hesitant normalized Euclidean–Hausdorff distance (NEH):

$$\tilde{d}_{NEH} = \left(\frac{1}{3n}\sum_{i=1}^{n}\max_j\left(\left|\tilde{t}_A^{\sigma(j)}(x_i) - \tilde{t}_B^{\sigma(j)}(x_i)\right|^{2}, \left|\tilde{\imath}_A^{\sigma(j)}(x_i) - \tilde{\imath}_B^{\sigma(j)}(x_i)\right|^{2}, \left|\tilde{f}_A^{\sigma(j)}(x_i) - \tilde{f}_B^{\sigma(j)}(x_i)\right|^{2}\right)\right)^{\frac{1}{2}}. \quad (10)$$

In many practical situations, the weight of each element $x_i \in X$ should be taken into account. For instance, in MADM problems, the considered attribute usually has different importance, thus needs to be assigned with different weights. Since in SVNHFSs, we have three types of degree, one is truth-membership degree, other is indeterminacy-membership and final is falsity-membership degree. Since three degrees may have different importance, according to decision maker, different weights can be assigned to each element in each degree. Assume that the weights $\omega = (\omega_1, \omega_2, \dots \omega_n)^T$ with $\omega_j \in [0,1]$, $\sum_{i=1}^{n}\omega_i = 1$; $\psi = (\psi_1, \psi_2, \dots \psi_n)^T$ with $\psi_i \in [0,1]$, $\sum_{i=1}^{n}\psi_i = 1$ and $\phi = (\phi_1, \phi_2, \dots \phi_n)^T$ with $\phi_i \in [0,1]$, $\sum_{i=1}^{n}\phi_i = 1$ denote the weights assigned to truth-membership degree, indeterminacy-membership degree and falsity-membership degree, respectively, of SVNHFS.

Now, we present the following weighted distance measures for SVNHFSs.

(3) Generalized single valued neutrosophic hesitant weighted distance (GW):

$$\tilde{d}_{GW} = \left(\frac{1}{3}\sum_{i=1}^{n}\left(\omega_i\left(\frac{1}{l_{\tilde{t}}(x_i)}\sum_{j=1}^{l_{\tilde{t}}(x_i)}\left|\tilde{t}_A^{\sigma(j)}(x_i) - \tilde{t}_B^{\sigma(j)}(x_i)\right|^{\lambda}\right) + \psi_i\left(\frac{1}{l_{\tilde{\imath}}(x_i)}\sum_{j=1}^{l_{\tilde{\imath}}(x_i)}\left|\tilde{\imath}_A^{\sigma(j)}(x_i) - \tilde{\imath}_B^{\sigma(j)}(x_i)\right|^{\lambda}\right) + \phi_i\left(\frac{1}{l_{\tilde{f}}(x_i)}\sum_{j=1}^{l_{\tilde{f}}(x_i)}\left|\tilde{f}_A^{\sigma(j)}(x_i) - \tilde{f}_B^{\sigma(j)}(x_i)\right|^{\lambda}\right)\right)\right)^{\frac{1}{\lambda}}. \quad (11)$$

i. If $\lambda = 1$, then we get a single valued neutrosophic hesitant weighted Hamming distance (WH):

$$\tilde{d}_{WH} = \frac{1}{3}\sum_{i=1}^{n}\left(\omega_j\left(\frac{1}{l_{\tilde{t}}(x_i)}\sum_{j=1}^{l_{\tilde{t}}(x_i)}\left|\tilde{t}_A^{\sigma(j)}(x_i) - \tilde{t}_B^{\sigma(j)}(x_i)\right|\right) + \psi_j\left(\frac{1}{l_{\tilde{\imath}}(x_i)}\sum_{j=1}^{l_{\tilde{\imath}}(x_i)}\left|\tilde{\imath}_A^{\sigma(j)}(x_i) - \tilde{\imath}_B^{\sigma(j)}(x_i)\right|\right) + \phi_j\left(\frac{1}{l_{\tilde{f}}(x_i)}\sum_{j=1}^{l_{\tilde{f}}(x_i)}\left|\tilde{f}_A^{\sigma(j)}(x_i) - \tilde{f}_B^{\sigma(j)}(x_i)\right|\right)\right). \quad (12)$$

ii. If $\lambda = 2$, then we get a single valued neutrosophic hesitant weighted Euclidean distance (WE):

$$\tilde{d}_{WE} = \left(\frac{1}{3}\sum_{i=1}^{n}\left(\omega_j\left(\frac{1}{l_{\tilde{t}}(x_i)}\sum_{j=1}^{l_{\tilde{t}}(x_i)}\left|\tilde{t}_A^{\sigma(j)}(x_i) - \tilde{t}_B^{\sigma(j)}(x_i)\right|^2\right) + \psi_j\left(\frac{1}{l_{\tilde{\imath}}(x_i)}\sum_{j=1}^{l_{\tilde{\imath}}(x_i)}\left|\tilde{\imath}_A^{\sigma(j)}(x_i) - \tilde{\imath}_B^{\sigma(j)}(x_i)\right|^2\right) + \phi_j\left(\frac{1}{l_{\tilde{f}}(x_i)}\sum_{j=1}^{l_{\tilde{f}}(x_i)}\left|\tilde{f}_A^{\sigma(j)}(x_i) - \tilde{f}_B^{\sigma(j)}(x_i)\right|^2\right)\right)\right)^{\frac{1}{2}}. \quad (13)$$

(4) Generalized single valued neutrosophic hesitant weighted Hausdorff distance (GWH), for $\lambda > 0$;

$$\tilde{d}_{GWH} = \left(\frac{1}{3}\sum_{i=1}^{n}\max_j\left(\omega_i\left|\tilde{t}_A^{\sigma(j)}(x_i) - \tilde{t}_B^{\sigma(j)}(x_i)\right|^\lambda, \psi_j\left|\tilde{\imath}_A^{\sigma(j)}(x_i) - \tilde{\imath}_B^{\sigma(j)}(x_i)\right|^\lambda, \phi_j\left|\tilde{f}_A^{\sigma(j)}(x_i) - f_B^{\sigma(j)}(x_i)\right|^\lambda\right)\right)^{\frac{1}{\lambda}}.$$

(14)

i. $\lambda = 1$, then we get a single valued neutrosophic hesitant weighted Hamming–Hausdorff distance (WHH):

$$\tilde{d}_{WHH} = \left(\frac{1}{3}\sum_{i=1}^{n}\max_j\left(\omega_i\left|\tilde{t}_A^{\sigma(j)}(x_i) - \tilde{t}_B^{\sigma(j)}(x_i)\right|, \psi_i\left|\tilde{\imath}_A^{\sigma(j)}(x_i) - \tilde{\imath}_B^{\sigma(j)}(x_i)\right|, \phi_i\left|\tilde{f}_A^{\sigma(j)}(x_i) - \tilde{f}_B^{\sigma(j)}(x_i)\right|\right)\right).$$

(15)

ii. $\lambda = 2$, then we get a single valued neutrosophic hesitant weighted Euclidean–Hausdorff distance (WEH):

$$\tilde{d}_{WEH}=\left(\frac{1}{3}\sum_{i=1}^{n}\max_{j}\left(\omega_i\left(\left|\tilde{t}_A^{\sigma(j)}(x_i)-\tilde{t}_B^{\sigma(j)}(x_i)\right|^2\right),\psi_i\left(\left|\tilde{\imath}_A^{\sigma(j)}(x_i)-\tilde{\imath}_B^{\sigma(j)}(x_i)\right|^2\right),\phi_i\left(\left|\tilde{f}_A^{\sigma(j)}(x_i)-\tilde{f}_B^{\sigma(j)}(x_i)\right|^2\right)\right)\right)^{\frac{1}{2}}. \quad (16)$$

Next, we shall show that the proposed distance measures satisfy axiom definition of distance measure.

**Theorem 7.** Let $A$, $B$ and $C$ be any SVNHFSs, then $\tilde{d}_{NH}(A,B)$ is a distance measure.

**Proof.** We should prove that $\tilde{d}_{NH}(A,B)$ satisfies axioms (D1)-(D4).

(D1) Suppose that $A$ and $B$ are two SVNHFSs with $n$ attributes, then

$$\left|\tilde{t}_A^{\sigma(j)}(x_i)-\tilde{t}_B^{\sigma(j)}(x_i)\right|\geq 0,\ \left|\tilde{\imath}_A^{\sigma(j)}(x_i)-\tilde{\imath}_B^{\sigma(j)}(x_i)\right|\geq 0 \text{ and } \left|\tilde{f}_A^{\sigma(j)}(x_i)-\tilde{f}_B^{\sigma(j)}(x_i)\right|\geq 0$$

and so

$$\frac{1}{l_{\tilde{t}}(x_i)}\sum_{j=1}^{l_{\tilde{t}}(x_i)}\left|\tilde{t}_A^{\sigma(j)}(x_i)-\tilde{t}_B^{\sigma(j)}(x_i)\right|\geq 0,$$

$$\frac{1}{l_{\tilde{\imath}}(x_i)}\sum_{j=1}^{l_{\tilde{\imath}}(x_i)}\left|\tilde{\imath}_A^{\sigma(j)}(x_i)-\tilde{\imath}_B^{\sigma(j)}(x_i)\right|\geq 0 \text{ and } \frac{1}{l_{\tilde{f}}(x_i)}\sum_{j=1}^{l_{\tilde{f}}(x_i)}\left|\tilde{f}_A^{\sigma(j)}(x_i)-\tilde{f}_B^{\sigma(j)}(x_i)\right|\geq 0.$$

Thus, we have that

$$\frac{1}{3n}\sum_{i=1}^{n}\left(\frac{1}{l_{\tilde{t}}(x_i)}\sum_{j=1}^{l_{\tilde{t}}(x_i)}\left|\tilde{t}_A^{\sigma(j)}(x_i)-\tilde{t}_B^{\sigma(j)}(x_i)\right|+\frac{1}{l_{\tilde{\imath}}(x_i)}\sum_{j=1}^{l_{\tilde{\imath}}(x_i)}\left|\tilde{\imath}_A^{\sigma(j)}(x_i)-\tilde{\imath}_B^{\sigma(j)}(x_i)\right|+\frac{1}{l_{\tilde{f}}(x_i)}\sum_{j=1}^{l_{\tilde{f}}(x_i)}\left|\tilde{f}_A^{\sigma(j)}(x_i)-\tilde{f}_B^{\sigma(j)}(x_i)\right|\right)\geq 0$$

and $\tilde{d}_{NH}(A,B)\geq 0$.

On the other hand, since

$$\left|\tilde{t}_A^{\sigma(j)}(x_i)-\tilde{t}_B^{\sigma(j)}(x_i)\right|\leq 1,\ \left|\tilde{\imath}_A^{\sigma(j)}(x_i)-\tilde{\imath}_B^{\sigma(j)}(x_i)\right|\leq 1 \text{ and } \left|\tilde{\imath}_A^{\sigma(j)}(x_i)-\tilde{\imath}_B^{\sigma(j)}(x_i)\right|\leq 1,$$

we get

$$\frac{1}{3n}\sum_{i=1}^{n}\left(\frac{1}{l_{\tilde{t}}(x_i)}\sum_{j=1}^{l_{\tilde{t}}(x_i)}\left|\tilde{t}_A^{\sigma(j)}(x_i)-\tilde{t}_B^{\sigma(j)}(x_i)\right|+\frac{1}{l_{\tilde{\imath}}(x_i)}\sum_{j=1}^{l_{\tilde{\imath}}(x_i)}\left|\tilde{\imath}_A^{\sigma(j)}(x_i)-\tilde{\imath}_B^{\sigma(j)}(x_i)\right|+\frac{1}{l_{\tilde{f}}(x_i)}\sum_{j=1}^{l_{\tilde{f}}(x_i)}\left|\tilde{f}_A^{\sigma(j)}(x_i)-\tilde{f}_B^{\sigma(j)}(x_i)\right|\right)\leq 1$$

and so $\tilde{d}_{NH}(A,B)\leq 1$.

Then it implies that $0\leq\tilde{d}_{NH}(A,B)\leq 1$.

D2)

$$\tilde{d}_{NH}(A,B) = \frac{1}{3n}\sum_{i=1}^{n}\left(\frac{1}{l_{\tilde{t}}(x_i)}\sum_{j=1}^{l_{\tilde{t}}(x_i)}\left|\tilde{t}_A^{\sigma(j)}(x_i) - \tilde{t}_B^{\sigma(j)}(x_i)\right| + \frac{1}{l_{\tilde{\imath}}(x_i)}\sum_{j=1}^{l_{\tilde{\imath}}(x_i)}\left|\tilde{\imath}_A^{\sigma(j)}(x_i) - \tilde{\imath}_B^{\sigma(j)}(x_i)\right| + \frac{1}{l_{\tilde{f}}(x_i)}\sum_{j=1}^{l_{\tilde{f}}(x_i)}\left|\tilde{f}_A^{\sigma(j)}(x_i) - \tilde{f}_B^{\sigma(j)}(x_i)\right|\right)$$

$$= \frac{1}{3n}\sum_{i=1}^{n}\left(\frac{1}{l_{\tilde{t}}(x_i)}\sum_{j=1}^{l_{\tilde{t}}(x_i)}\left|\tilde{t}_B^{\sigma(j)}(x_i) - \tilde{t}_A^{\sigma(j)}(x_i)\right| + \frac{1}{l_{\tilde{\imath}}(x_i)}\sum_{j=1}^{l_{\tilde{\imath}}(x_i)}\left|\tilde{\imath}_B^{\sigma(j)}(x_i) - \tilde{\imath}_A^{\sigma(j)}(x_i)\right| + \frac{1}{l_{\tilde{f}}(x_i)}\sum_{j=1}^{l_{\tilde{f}}(x_i)}\left|\tilde{f}_B^{\sigma(j)}(x_i) - \tilde{f}_A^{\sigma(j)}(x_i)\right|\right)$$

$$= \tilde{d}_{NH}(B,A).$$

(D3) Consider $A = B$, then

$$A = B \Leftrightarrow \frac{1}{l_{\tilde{t}}(x_i)}\sum_{j=1}^{l_{\tilde{t}}(x_i)}\tilde{t}_A^{\sigma(j)}(x_i) + \frac{1}{l_{\tilde{\imath}}(x_i)}\sum_{j=1}^{l_{\tilde{\imath}}(x_i)}\tilde{\imath}_A^{\sigma(j)}(x_i) + \frac{1}{l_{\tilde{f}}(x_i)}\sum_{j=1}^{l_{\tilde{f}}(x_i)}\tilde{f}_A^{\sigma(j)}(x_i)$$

$$= \frac{1}{l_{\tilde{t}}(x_i)}\sum_{j=1}^{l_{\tilde{t}}(x_i)}\tilde{t}_B^{\sigma(j)}(x_i) + \frac{1}{l_{\tilde{\imath}}(x_i)}\sum_{j=1}^{l_{\tilde{\imath}}(x_i)}\tilde{\imath}_B^{\sigma(j)}(x_i) + \frac{1}{l_{\tilde{f}}(x_i)}\sum_{j=1}^{l_{\tilde{f}}(x_i)}\tilde{f}_B^{\sigma(j)}(x_i)$$

$$\Leftrightarrow \frac{1}{l_{\tilde{t}}(x_i)}\sum_{j=1}^{l_{\tilde{t}}(x_i)}\left|\tilde{t}_A^{\sigma(j)}(x_i) - \tilde{t}_B^{\sigma(j)}(x_i)\right| + \frac{1}{l_{\tilde{\imath}}(x_i)}\sum_{j=1}^{l_{\tilde{\imath}}(x_i)}\left|\tilde{\imath}_A^{\sigma(j)}(x_i) - \tilde{\imath}_B^{\sigma(j)}(x_i)\right| + \frac{1}{l_{\tilde{f}}(x_i)}\sum_{j=1}^{l_{\tilde{f}}(x_i)}\left|\tilde{f}_A^{\sigma(j)}(x_i) - \tilde{f}_B^{\sigma(j)}(x_i)\right| = 0$$

$$\Leftrightarrow \frac{1}{3n}\sum_{i=1}^{n}\left(\frac{1}{l_{\tilde{t}}(x_i)}\sum_{j=1}^{l_{\tilde{t}}(x_i)}\left|\tilde{t}_B^{\sigma(j)}(x_i) - \tilde{t}_A^{\sigma(j)}(x_i)\right| + \frac{1}{l_{\tilde{\imath}}(x_i)}\sum_{j=1}^{l_{\tilde{\imath}}(x_i)}\left|\tilde{\imath}_B^{\sigma(j)}(x_i) - \tilde{\imath}_A^{\sigma(j)}(x_i)\right| + \frac{1}{l_{\tilde{f}}(x_i)}\sum_{j=1}^{l_{\tilde{f}}(x_i)}\left|\tilde{f}_B^{\sigma(j)}(x_i) - \tilde{f}_A^{\sigma(j)}(x_i)\right|\right) = 0$$

$$\Leftrightarrow \tilde{d}_{NH}(A,B) = 0.$$

(D4) Since $A \subseteq B \subseteq C$, we have

$$\frac{1}{l_{\tilde{t}}(x_i)}\sum_{j=1}^{l_{\tilde{t}}(x_i)}\tilde{t}_A^{\sigma(j)}(x_i) \leq \frac{1}{l_{\tilde{t}}(x_i)}\sum_{j=1}^{l_{\tilde{t}}(x_i)}\tilde{t}_B^{\sigma(j)}(x_i) \leq \frac{1}{l_{\tilde{t}}(x_i)}\sum_{j=1}^{l_{\tilde{t}}(x_i)}\tilde{t}_C^{\sigma(j)}(x_i),$$

$$\frac{1}{l_{\tilde{\imath}}(x_i)}\sum_{j=1}^{l_{\tilde{\imath}}(x_i)}\tilde{\imath}_C^{\sigma(j)}(x_i) \leq \frac{1}{l_{\tilde{\imath}}(x_i)}\sum_{j=1}^{l_{\tilde{\imath}}(x_i)}\tilde{\imath}_B^{\sigma(j)}(x_i) \leq \frac{1}{l_{\tilde{\imath}}(x_i)}\sum_{j=1}^{l_{\tilde{\imath}}(x_i)}\tilde{\imath}_A^{\sigma(j)}(x_i),$$

$$\frac{1}{l_{\tilde{f}}(x_i)}\sum_{j=1}^{l_{\tilde{f}}(x_i)}\tilde{f}_C^{\sigma(j)}(x_i) \leq \frac{1}{l_{\tilde{f}}(x_i)}\sum_{j=1}^{l_{\tilde{f}}(x_i)}\tilde{f}_B^{\sigma(j)}(x_i) \leq \frac{1}{l_{\tilde{f}}(x_i)}\sum_{j=1}^{l_{\tilde{f}}(x_i)}\tilde{f}_A^{\sigma(j)}(x_i).$$

Then it follows that

$$\frac{1}{l_{\tilde{t}}(x_i)}\sum_{j=1}^{l_{\tilde{t}}(x_i)}\left|\tilde{t}_A^{\sigma(j)}(x_i)-\tilde{t}_B^{\sigma(j)}(x_i)\right| \le \frac{1}{l_{\tilde{t}}(x_i)}\sum_{j=1}^{l_{\tilde{t}}(x_i)}\left|\tilde{t}_A^{\sigma(j)}(x_i)-\tilde{t}_C^{\sigma(j)}(x_i)\right|,$$
$$\frac{1}{l_{\tilde{\imath}}(x_i)}\sum_{j=1}^{l_{\tilde{\imath}}(x_i)}\left|\tilde{\imath}_A^{\sigma(j)}(x_i)-\tilde{\imath}_B^{\sigma(j)}(x_i)\right| \le \frac{1}{l_{\tilde{\imath}}(x_i)}\sum_{j=1}^{l_{\tilde{\imath}}(x_i)}\left|\tilde{\imath}_A^{\sigma(j)}(x_i)-\tilde{\imath}_C^{\sigma(j)}(x_i)\right|,$$
$$\frac{1}{l_{\tilde{f}}(x_i)}\sum_{j=1}^{l_{\tilde{f}}(x_i)}\left|\tilde{f}_A^{\sigma(j)}(x_i)-\tilde{f}_B^{\sigma(j)}(x_i)\right| \le \frac{1}{l_{\tilde{f}}(x_i)}\sum_{j=1}^{l_{\tilde{f}}(x_i)}\left|\tilde{f}_A^{\sigma(j)}(x_i)-\tilde{f}_C^{\sigma(j)}(x_i)\right|$$

and so

$$\begin{aligned}
\tilde{d}_{NH}(A,B) &= \frac{1}{3n}\sum_{i=1}^{n}\left(\frac{1}{l_{\tilde{t}}(x_i)}\sum_{j=1}^{l_{\tilde{t}}(x_i)}\left|\tilde{t}_A^{\sigma(j)}(x_i)-\tilde{t}_B^{\sigma(j)}(x_i)\right| + \frac{1}{l_{\tilde{\imath}}(x_i)}\sum_{j=1}^{l_{\tilde{\imath}}(x_i)}\left|\tilde{\imath}_A^{\sigma(j)}(x_i)-\tilde{\imath}_B^{\sigma(j)}(x_i)\right|\right.\\
&\qquad \left. + \frac{1}{l_{\tilde{f}}(x_i)}\sum_{j=1}^{l_{\tilde{f}}(x_i)}\left|\tilde{f}_A^{\sigma(j)}(x_i)-\tilde{f}_B^{\sigma(j)}(x_i)\right|\right)\\
&\le \frac{1}{3n}\sum_{i=1}^{n}\left(\frac{1}{l_{\tilde{t}}(x_i)}\sum_{j=1}^{l_{\tilde{t}}(x_i)}\left|\tilde{t}_A^{\sigma(j)}(x_i)-\tilde{t}_C^{\sigma(j)}(x_i)\right| + \frac{1}{l_{\tilde{\imath}}(x_i)}\sum_{j=1}^{l_{\tilde{\imath}}(x_i)}\left|\tilde{\imath}_A^{\sigma(j)}(x_i)-\tilde{\imath}_C^{\sigma(j)}(x_i)\right|\right.\\
&\qquad \left. + \frac{1}{l_{\tilde{f}}(x_i)}\sum_{j=1}^{l_{\tilde{f}}(x_i)}\left|\tilde{f}_A^{\sigma(j)}(x_i)-\tilde{f}_C^{\sigma(j)}(x_i)\right|\right)\\
&= \tilde{d}_{NH}(A,C).
\end{aligned}$$

Similarly, we can prove $\tilde{d}_{NH}(B,C) \le \tilde{d}_{NH}(A,C)$.

**Theorem 8.** Let $A$ and $B$ be two SVNHSs, then $\tilde{d}_{GH}(A,B)$ and $\tilde{d}_{NE}(A,B)$ are two distance measures.

**Proof.** By the similar proof manner of Theorem 7, we can also give the proof of Theorem 8 (omitted).

**Theorem 9.** Let $A$, $B$ and $C$ be any SVNHFSs, then $\tilde{d}_{GNH}(A,B)$ is the distance measure.

**Proof.** We should prove that $\tilde{d}_{GNH}(A,B)$ satisfies axioms (D1)-(D4).

(D1) Suppose that $A$ and $B$ are two SVNHFSs with $n$ attributes, then

$\left|\tilde{t}_A^{\sigma(j)}(x_i)-\tilde{t}_B^{\sigma(j)}(x_i)\right| \ge 0$, $\left|\tilde{\imath}_A^{\sigma(j)}(x_i)-\tilde{\imath}_B^{\sigma(j)}(x_i)\right| \ge 0$ and $\left|\tilde{f}_A^{\sigma(j)}(x_i)-\tilde{f}_B^{\sigma(j)}(x_i)\right| \ge 0$

and so

$$\left(\frac{1}{3n}\sum_{i=1}^{n}\max_j\left(\left|\tilde{t}_A^{\sigma(j)}(x_i)-\tilde{t}_B^{\sigma(j)}(x_i)\right|^{\lambda},\left|\tilde{\imath}_A^{\sigma(j)}(x_i)-\tilde{\imath}_B^{\sigma(j)}(x_i)\right|^{\lambda},\left|\tilde{f}_A^{\sigma(j)}(x_i)-f_B^{\sigma(j)}(x_i)\right|^{\lambda}\right)\right)^{\frac{1}{\lambda}} \ge 0.$$

Thus, we have $\tilde{d}_{GNH}(A,B) \ge 0$.

(D2)

$$\begin{aligned}\tilde{d}_{GNH}(A,B) &= \left(\frac{1}{3n}\sum_{i=1}^{n}\max_{j}\left(\left|\tilde{t}_A^{\sigma(j)}(x_i)-\tilde{t}_B^{\sigma(j)}(x_i)\right|^{\lambda},\left|\tilde{\imath}_A^{\sigma(j)}(x_i)-\tilde{\imath}_B^{\sigma(j)}(x_i)\right|^{\lambda},\left|\tilde{f}_A^{\sigma(j)}(x_i)-\tilde{f}_B^{\sigma(j)}(x_i)\right|^{\lambda}\right)\right)^{\frac{1}{\lambda}}\\ &= \left(\frac{1}{3n}\sum_{i=1}^{n}\max_{j}\left(\left|\tilde{t}_B^{\sigma(j)}(x_i)-\tilde{t}_A^{\sigma(j)}(x_i)\right|^{\lambda},\left|\tilde{\imath}_B^{\sigma(j)}(x_i)-\tilde{\imath}_A^{\sigma(j)}(x_i)\right|^{\lambda},\left|\tilde{f}_B^{\sigma(j)}(x_i)-\tilde{f}_A^{\sigma(j)}(x_i)\right|^{\lambda}\right)\right)^{\frac{1}{\lambda}}\\ &= \tilde{d}_{GNH}(B,A)\end{aligned}$$

(D3) Let $\tilde{d}_{GNH}(A,B)=0$, then

$$\begin{aligned}&\Leftrightarrow\left(\frac{1}{3n}\sum_{i=1}^{n}\max_{j}\left(\left|\tilde{t}_A^{\sigma(j)}(x_i)-\tilde{t}_B^{\sigma(j)}(x_i)\right|^{\lambda},\left|\tilde{\imath}_A^{\sigma(j)}(x_i)-\tilde{\imath}_B^{\sigma(j)}(x_i)\right|^{\lambda},\left|\tilde{f}_A^{\sigma(j)}(x_i)-\tilde{f}_B^{\sigma(j)}(x_i)\right|^{\lambda}\right)\right)^{\frac{1}{\lambda}}=0\\ &\Leftrightarrow \tilde{t}_A^{\sigma(j)}(x_i)=\tilde{t}_B^{\sigma(j)}(x_i),\quad \tilde{\imath}_A^{\sigma(j)}(x_i)=\tilde{\imath}_B^{\sigma(j)}(x_i)\quad\text{and}\quad \tilde{f}_A^{\sigma(j)}(x_i)=\tilde{f}_B^{\sigma(j)}(x_i)\\ &\Leftrightarrow A=B.\end{aligned}$$

(D4) Since $A\subseteq B\subseteq C$, we have

$$\tilde{t}_A^{\sigma(j)}(x_i)\le\tilde{t}_B^{\sigma(j)}(x_i)\le\tilde{t}_C^{\sigma(j)}(x_i),\tilde{\imath}_C^{\sigma(j)}(x_i)\le\tilde{\imath}_B^{\sigma(j)}(x_i)\le\tilde{\imath}_A^{\sigma(j)}(x_i)\text{ and }\tilde{f}_C^{\sigma(j)}(x_i)\le\tilde{f}_B^{\sigma(j)}(x_i)\le\tilde{f}_A^{\sigma(j)}(x_i).$$

Then it follows that

$$\begin{aligned}&\max_{j}\left(\left|\tilde{t}_A^{\sigma(j)}(x_i)-\tilde{t}_B^{\sigma(j)}(x_i)\right|^{\lambda},\left|\tilde{\imath}_A^{\sigma(j)}(x_i)-\tilde{\imath}_B^{\sigma(j)}(x_i)\right|^{\lambda},\left|\tilde{f}_A^{\sigma(j)}(x_i)-\tilde{f}_B^{\sigma(j)}(x_i)\right|^{\lambda}\right)\\ &\le\max_{j}\left(\left|\tilde{t}_A^{\sigma(j)}(x_i)-\tilde{t}_C^{\sigma(j)}(x_i)\right|^{\lambda},\left|\tilde{\imath}_A^{\sigma(j)}(x_i)-\tilde{\imath}_C^{\sigma(j)}(x_i)\right|^{\lambda},\left|\tilde{f}_A^{\sigma(j)}(x_i)-\tilde{f}_C^{\sigma(j)}(x_i)\right|^{\lambda}\right)\end{aligned}$$

and so

$$\begin{aligned}\tilde{d}_{GNH}(A,B) &= \left(\frac{1}{3n}\sum_{i=1}^{n}\max_{j}\left(\left|\tilde{t}_A^{\sigma(j)}(x_i)-\tilde{t}_B^{\sigma(j)}(x_i)\right|^{\lambda},\left|\tilde{\imath}_A^{\sigma(j)}(x_i)-\tilde{\imath}_B^{\sigma(j)}(x_i)\right|^{\lambda},\left|\tilde{f}_A^{\sigma(j)}(x_i)-\tilde{f}_B^{\sigma(j)}(x_i)\right|^{\lambda}\right)\right)^{\frac{1}{\lambda}}\\ &\le \left(\frac{1}{3n}\sum_{i=1}^{n}\max_{j}\left(\left|\tilde{t}_A^{\sigma(j)}(x_i)-\tilde{t}_C^{\sigma(j)}(x_i)\right|^{\lambda},\left|\tilde{\imath}_A^{\sigma(j)}(x_i)-\tilde{\imath}_C^{\sigma(j)}(x_i)\right|^{\lambda},\left|\tilde{f}_A^{\sigma(j)}(x_i)-\tilde{f}_C^{\sigma(j)}(x_i)\right|^{\lambda}\right)\right)^{\frac{1}{\lambda}}\\ &= \tilde{d}_{GNH}(A,C).\end{aligned}$$

Similarly, we can prove $\tilde{d}_{GNH}(B,C)\le\tilde{d}_{GNH}(A,C)$.

**Theorem 10.** Let $A$ and $B$ be two SVNHSs, then $\tilde{d}_{NHH}(A,B)$ and $\tilde{d}_{NEH}(A,B)$ are two distance measures.

**Proof.** By the similar proof manner of Theorem 9, we can also give the proof of Theorem 10 (omitted).

## 4. Some similarity measures for SVNHFSs

In this section, we present some similarity measures based on the proposed distance measures between SVNHFSs.

### 4.1. The similarity measures based on geometric distance model for SVNHFSs

With respect to Eq. (5), the similarity measure can be defined as follows:

(1) Similarity measure based on generalized single valued neutrosophic hesitant normalized distance:

$\tilde{s}_{GN}(A,B) = 1 - \tilde{d}_{GN}(A,B) = 1 -$

$$\tilde{d}_{GN} = \left( \frac{1}{3n} \sum_{i=1}^{n} \left( \frac{1}{l_{\tilde{t}}(x_i)} \sum_{j=1}^{l_{\tilde{t}}(x_i)} \left| \tilde{t}_A^{\sigma(j)}(x_i) - \tilde{t}_B^{\sigma(j)}(x_i) \right|^{\lambda} + \frac{1}{l_{\tilde{\imath}}(x_i)} \sum_{j=1}^{l_{\tilde{\imath}}(x_i)} \left| \tilde{\imath}_A^{\sigma(j)}(x_i) - \tilde{\imath}_B^{\sigma(j)}(x_i) \right|^{\lambda} + \frac{1}{l_{\tilde{f}}(x_i)} \sum_{j=1}^{l_{\tilde{f}}(x_i)} \left| \tilde{f}_A^{\sigma(j)}(x_i) - \tilde{f}_B^{\sigma(j)}(x_i) \right|^{\lambda} \right) \right)^{\frac{1}{\lambda}} \quad (17)$$

Similarly, we give another similarity measures based on distance measure as follows:

(i) Similarity measure based on single valued neutrosophic hesitant normalized Hamming distance:

$$\tilde{s}_{NH}(A,B) = 1 - \tilde{d}_{NH}(A,B) \quad (18)$$

(ii) Similarity measure based on single valued neutrosophic hesitant normalized Euclidian distance:

$$\tilde{s}_{NE}(A,B) = 1 - \tilde{d}_{NE}(A,B) \quad (19)$$

(2) Similarity measure based on generalized single valued neutrosophic hesitant normalized Hausdorff distance:

$$\tilde{s}_{GNH}(A,B) = 1 - \tilde{d}_{GNH}(A,B) \quad (20)$$

(i) Similarity measure based on single valued neutrosophic hesitant normalized Hamming–Hausdorff distance:

$$\tilde{s}_{NHH}(A,B) = 1 - \tilde{d}_{NHH}(A,B) \quad (21)$$

(ii) Similarity measure based on single valued neutrosophic hesitant normalized Euclidian–Hausdorff distance:

$$\tilde{s}_{NEH}(A,B) = 1 - \tilde{d}_{NEH}(A,B) \quad (22)$$

(3) Similarity measure based on generalized single valued neutrosophic hesitant weighted Hausdorff distance:

$$\tilde{s}_{GW}(A,B) = 1 - \tilde{d}_{GW}(A,B) \quad (23)$$

(i) Similarity measure based on single valued neutrosophic hesitant weighted Hamming distance:

$$\tilde{s}_{WH}(A,B) = 1 - \tilde{d}_{WH}(A,B) \quad (24)$$

(ii) Similarity measure based on single valued neutrosophic hesitant weighted Euclidian distance:

$$\tilde{s}_{WE}(A,B) = 1 - \tilde{d}_{WE}(A,B) \quad (25)$$

**4.2. Similarity measure based on the set-theoretic approach**

Let A and B be two SVNHFSs, then we define a similarity measure from the point of set-theoretic view as follows:

$$\tilde{s}_{ST}(A,B) = \frac{1}{3n} \sum_{i=1}^{n} \frac{\sum_{j=1}^{l_{\tilde{t}}(x_i)} (\min \Delta\tilde{t}_{AB}(x_i)) + \sum_{j=1}^{l_{\tilde{\imath}}(x_i)} (\min \Delta\tilde{\imath}_{AB}(x_i)) + \sum_{j=1}^{l_{\tilde{f}}(x_i)} (\min \Delta\tilde{f}_{AB}(x_i))}{\sum_{j=1}^{l_{\tilde{t}}(x_i)} (\max \Delta\tilde{t}_{AB}(x_i)) + \sum_{j=1}^{l_{\tilde{\imath}}(x_i)} (\max \Delta\tilde{\imath}_{AB}(x_i)) + \sum_{j=1}^{l_{\tilde{f}}(x_i)} (\max \Delta\tilde{f}_{AB}(x_i))}, \quad (26)$$

where $\Delta\tilde{t}_{AB}(x_i) = \left(\tilde{t}_A^{\sigma(j)}(x_i), \tilde{t}_B^{\sigma(j)}(x_i)\right)$, $\Delta\tilde{\imath}_{AB}(x_i) = \left(\tilde{\imath}_A^{\sigma(j)}(x_i) - \tilde{\imath}_B^{\sigma(j)}(x_i)\right)$, $\Delta\tilde{f}_{AB}(x_i) = \left(\tilde{f}_A^{\sigma(j)}(x_i) - \tilde{f}_B^{\sigma(j)}(x_i)\right)$

By taking into account the weight of each element $x_i \in X$ for truth-membership function, indeterminacy-membership function and falsity membership function, we define a similarity measure as:

$$\tilde{s}_{WST}(A,B) = \frac{1}{3n}\sum_{i=1}^{n}\frac{\sum_{j=1}^{l_{\tilde{t}}(x_i)}\omega_j(\min\Delta\tilde{t}_{AB}(x_i)) + \sum_{j=1}^{l_{\tilde{\imath}}(x_i)}\psi_j(\min\Delta\tilde{\imath}_{AB}(x_i)) + \sum_{j=1}^{l_{\tilde{f}}(x_i)}\phi_j(\min\Delta\tilde{f}_{AB}(x_i))}{\sum_{j=1}^{l_{\tilde{t}}(x_i)}\omega_j(\max\Delta\tilde{t}_{AB}(x_i)) + \sum_{j=1}^{l_{\tilde{\imath}}(x_i)}\psi_j(\max\Delta\tilde{\imath}_{AB}(x_i)) + \sum_{j=1}^{l_{\tilde{f}}(x_i)}\phi_j(\max\Delta\tilde{f}_{AB}(x_i))} \quad (27)$$

### 4.3. Similarity measure based on matching function

The concept of similarity between FSs based on a matching function was defined by Chen et al. (1995). Then Xu (2007) extended the matching function to deal with the similarity measures for IFSs. In the following, we propose the similarity measure for SVNHFSs based on the matching function.

Suppose that A and B are two SVNHFSs, then we define a similarity measure based on the matching function as follows:

$$\tilde{s}_{MF}(A,B) = \frac{1}{3n}\sum_{i=1}^{n}\frac{\sum_{j=1}^{l_{\tilde{t}}(x_i)}\nabla\tilde{t}_{AB}(x_i) + \sum_{j=1}^{l_{\tilde{\imath}}(x_i)}\nabla\tilde{\imath}_{AB}(x_i) + \sum_{j=1}^{l_{\tilde{f}}(x_i)}\nabla\tilde{f}_{AB}(x_i)}{\max\left(\sum_{j=1}^{l_{\tilde{t}}(x_i)}\mathit{\Delta}\tilde{t}_{AB}(x_i), \sum_{j=1}^{l_{\tilde{t}}(x_i)}\mathit{\Delta}\tilde{\imath}_{AB}(x_i), \sum_{j=1}^{l_{\tilde{t}}(x_i)}\mathit{\Delta}\tilde{f}_{AB}(x_i)\right)} \quad (28)$$

where $\nabla\tilde{t}_{AB}(x_i) = \left(\tilde{t}_A^{\sigma(j)}(x_i)\times\tilde{t}_B^{\sigma(j)}(x_i)\right)$, $\nabla\tilde{\imath}_{AB}(x_i) = \left(\tilde{\imath}_A^{\sigma(j)}(x_i)\times\tilde{\imath}_B^{\sigma(j)}(x_i)\right)$ and $\nabla\tilde{f}_{AB}(x_i) = \left(\tilde{f}_A^{\sigma(j)}(x_i)\times \tilde{f}_B^{\sigma(j)}(x_i)\right)$, and $\mathit{\Delta}\tilde{t}_{AB}(x_i) = \left(\tilde{t}_A^{\sigma(j)}(x_i)\right)^2 + \left(\tilde{t}_B^{\sigma(j)}(x_i)\right)^2$, $\mathit{\Delta}\tilde{\imath}_{AB}(x_i) = \left(\tilde{\imath}_A^{\sigma(j)}(x_i)\right)^2 + \left(\tilde{\imath}_B^{\sigma(j)}(x_i)\right)^2$ and $\mathit{\Delta}\tilde{f}_{AB}(x_i) = \left(\tilde{f}_A^{\sigma(j)}(x_i)\right)^2 + \left(\tilde{f}_B^{\sigma(j)}(x_i)\right)^2$.

If we consider weight of each $x \in X$, then we get

$$\tilde{s}_{WMF}(A,B) = \frac{1}{3n}\sum_{i=1}^{n}\frac{\sum_{j=1}^{l_{\tilde{t}}(x_i)}\omega_j(\nabla\tilde{t}_{AB}(x_i)) + \sum_{j=1}^{l_{\tilde{\imath}}(x_i)}\psi_j(\nabla\tilde{\imath}_{AB}(x_i)) + \sum_{j=1}^{l_{\tilde{f}}(x_i)}\phi_j(\nabla\tilde{f}_{AB}(x_i))}{\max\left(\sum_{j=1}^{l_{\tilde{t}}(x_i)}\omega_j(\mathit{\Delta}\tilde{t}_{AB}(x_i)), \sum_{j=1}^{l_{\tilde{t}}(x_i)}\psi_j(\mathit{\Delta}\tilde{\imath}_{AB}(x_i)), \sum_{j=1}^{l_{\tilde{t}}(x_i)}\phi_j(\mathit{\Delta}\tilde{f}_{AB}(x_i))\right)} \quad (29)$$

It is clear that $\tilde{s}_{WMF}(A,B)$ satisfies all the properties described in Definition 6.

## 5. Decision-making method based on the single-valued neutrosophic hesitant fuzzy information

In this section, we apply the developed distance and similarity measures to a MADM problem with single-valued neutrosophic hesitant fuzzy information.

For the MADM problem, let $A = \{A_1, A_2, \dots, A_m\}$ be a set of alternatives, $C = \{C_1, C_2, \dots, C_n\}$ be a set of attributes. Suppose that $\omega = (\omega_1, \omega_2, \dots \omega_n)^T$, $\psi = (\psi_1, \psi_2, \dots \psi_n)^T$ and $\phi = (\phi_1, \phi_2, \dots \phi_n)^T$ are the potential weighting vector assigned to the truth-membership, the indeterminacy-membership and the falsity-membership, respectively, in each alternative, where $\omega_j \geq 0, \psi_j \geq 0$ and $\phi_j \geq 0, j = 1,2,\dots,n$, $\sum_{j=1}^{n}\omega_j = 1$, $\sum_{j=1}^{n}\psi_j = 1$ and $\sum_{j=1}^{n}\phi_j = 1$. If the decision makers provide several values for the alternative $A_i (i = 1,2,\dots,m)$ under the attribute $C_j$ $(j = 1,2,\dots,n)$, these values can be characterized as a SVNHFN $e_{ij} = \{t_{ij}, \imath_{ij}, f_{ij}\}$ $(j = 1,2,\dots,n; i = 1,2,\dots,m)$. Assume that $E = [e_{ij}]_{m\times n}$ is the decision matrix, where $e_{ij}$ is expressed by a single-valued neutrosophic hesitant fuzzy element.

In multiple attribute decision-making environments, we can utilize the concept of ideal point to determine the best alternative in the decision set. Although the ideal alternative does not exist in real world, it does provide a useful theoretical construct against which to evaluate alternatives. Therefore, we propose each ideal SVNHFN in the ideal alternative $A^* = \{\langle C_j, e_j^*\rangle : C_j \in C\}$ as $e_j^* = \{\tilde{t}_j^*, \tilde{\imath}_j^*, \tilde{f}_j^*\} = \{\{1\},\{0\},\{0\}\}$ $(j = 1,2,\dots,n)$.

Thus, we can develop a procedure for the decision maker to select the best choice with single valued neutrosophic hesitant fuzzy information, which can be given as follows:

***Step1.*** *Compute the distance (similarity) measure between an alternative* $A_i$ $(i = 1,2,\dots,m)$ *and the ideal alternative* $A^*$ *by using proposed distance (similarity) measure.*

***Step 2.*** *Rank all of the alternative with respect to the values of distance (similarity) measure.*

***Step 3.*** *Choose the best alternative with respect to the minimum value of distance (maximum value of similarity).*

***Step4.*** *End.*

## 6. Practical example

Here, an example for the multicriteria decision-making problem of alternatives is used as the demonstration of the application of the proposed decision-making method, as well as the effectiveness of the proposed method.

We take the example adopted from Ye (2014c) to illustrate the utility of the proposed distance and similarity measures. Also, we show that the results obtained using the proposed distance measure are more reasonable than the results obtained using Ye's (2014c) cosine similarity measure.

**Example 11.** Suppose that an investment company that wants to invest a sum of money in the best option. There is a panel with four possible alternatives in which to invest the money: (1) $A_1$ is a car company, (2) $A_2$ is a food company, (3) $A_3$ is a computer company, and (4) $A_4$ is an arms company. The investment company must make a decision according to the three attributes: (1) $C_1$ is the risk analysis; (2) $C_2$ is the growth analysis, and (3) $C_3$ is the environmental impact analysis. Suppose that $\omega = (0.35, 0.25, 0.40)$, $\psi = (0.35, 0.40, 0.25)$, and $\phi = (0.30, 0.40, 0.30)$ are the attribute weight vector for truth-membership degree, the indeterminacy-membership degree and the falsity membership degree, respectively. The four possible alternatives are to be evaluated under these three attributes and are presented in the form of single valued neutrosophic hesitant fuzzy information by decision maker according to three attributes $C_j$ $(j = 1,2,3)$, as expressed in the following single valued neutrosophic hesitant fuzzy decision matrix $E$:

**Table 1:** *Decision matrix E*

$$E = \begin{pmatrix} \{\{0.3,0.4,0.5\},\{0.1\},\{0.3,0.4\}\} & \{\{0.5,0.6\},\{0.2,0.3\},\{0.3,0.4\}\} & \{\{0.2,0.3\},\{0.1,0.2\},\{0.5,0.6\}\} \\ \{\{0.6,0.7\},\{0.1,0.2\},\{0.2,0.3\}\} & \{\{0.6,0.7\},\{0.1\},\{0.3\}\} & \{\{0.6,0.7\},\{0.1,0.2\},\{0.1,0.2\}\} \\ \{\{0.5,0.6\},\{0.4\},\{0.2,0.3\}\} & \{\{0.6\},\{0.3\},\{0.4\}\} & \{\{0.5,0.6\},\{0.1\},\{0.3\}\} \\ \{\{0.7,0.8\},\{0.1\},\{0.1,0.2\}\} & \{\{0.6,0.7\},\{0.1\},\{0.2\}\} & \{\{0.3,0.5\},\{0.2\},\{0.1,0.2,0.3\}\} \end{pmatrix}$$

To get the best alternative(s), the following steps are involved:

**Step 1.** Using Eq. (12), we can compute the single valued neutrosophic hesitant weighted Hamming distance between the alternatives and the ideal alternative as:

$$\tilde{d}_{NH}(A_1, A^*) = 0.4370, \tilde{d}_{NH}(A_2, A^*) = 0.2383, \tilde{d}_{NH}(A_3, A^*) = 0.3679, \tilde{d}_{NH}(A_4, A^*) = 0.2654.$$

**Step 2.** With respect to the values of weighted Hamming distance, we can rank the alternatives as $A_2 \succ A_4 \succ A_3 \succ A_1$.

**Step 3.** The alternative $A_2$ is the optimal choice according to the minimum value among weighted Hamming distances, which is not in agreement with the one obtained in Ye (2014c).

Above example clearly shows that the developed method is effective and applicable under single-valued neutrosophic hesitant fuzzy environment.

## 7. Related comparative analysis

**Case 1.** Ye (2014c) proposed a method based on single-valued neutrosophic hesitant fuzzy aggregation operators and cosine measure function to find the best alternative. This method lacks the decision makers' risk factor, which causes the distortion of similarity between an alternative and the ideal alternative and makes the proposed method more realistic. Therefore, the results of the proposed method don't coincide with the existing method Ye (2014c). In proposed method, we not only consider the decision makers' risk case but also the individual weighting vectors of truth-membership, indeterminacy-membership, and falsity-membership degrees of each element in decision space, separately. From Table 2, we can see that the rankings are changed according to different parameters $\lambda$, consequently, the proposed distance measure can provide a more flexible decision and more choice for decision makers because of the decision maker' risk factor and the individual weighting vector of membership degrees. Combining the analyses above, our method is more precise and reliable than the result produced in Ye (2014c).

**Table 2:** *Results obtained by Eq. (5) corresponding different $\lambda$ values*

| $\lambda$ | $A_1$ | $A_2$ | $A_3$ | $A_4$ | Ranking |
|---|---|---|---|---|---|
| $\lambda = 1$ | 0.4370 | 0.2383 | 0.3679 | 0.2654 | $A_2 \succ A_4 \succ A_3 \succ A_1$ |
| $\lambda = 2$ | 0.6611 | 0.5256 | 0.5942 | 0.5619 | $A_2 \succ A_4 \succ A_3 \succ A_1$ |
| $\lambda = 5$ | 0.8683 | 0.8102 | 0.8320 | 0.8455 | $A_2 \succ A_3 \succ A_4 \succ A_1$ |
| $\lambda = 10$ | 0.9395 | 0.9140 | 0.9214 | 0.9320 | $A_2 \succ A_3 \succ A_4 \succ A_1$ |

**Case 2.** In order to validate the feasibility of the proposed decision making method, we give another comparative study between our method and existing methods and use the concept of weighted Euclidian distance. Xu and Xia's method (2011) is used to rank the HFSs which are only characterized by a set of the membership degrees, whereas Singh's method (2013) is applied to DHFSs which are taken into account both the membership hesitant degree and the non-membership hesitant degree in decision making process.

**Table 3:** *Relationships between existing methods and proposed method*

| Methods | Rankings | The best alternative(s) | The worst alternative(s) |
|---|---|---|---|
| Xu and Xia's method (2011) | $A_2 \succ A_3 \succ A_4 \succ A_1$ | $A_2$ | $A_1$ |
| Singh's method (2015) | $A_2 \succ A_4 \succ A_3 \succ A_1$ | $A_2$ | $A_1$ |
| Ye's method (2005) | $A_4 \succ A_2 \succ A_3 \succ A_1$ | $A_4$ | $A_1$ |
| Our method | $A_2 \succ A_4 \succ A_3 \succ A_1$ | $A_2$ | $A_1$ |

On the other hand, we also utilize the weighted Euclidian distance to determine the final ranking order of all the alternatives associated with SVNHFS, which are expressed by the truth-membership hesitant degree, indeterminacy-membership hesitant degree, and falsity-membership hesitant degree, to calculate the distance measures and to rank all of the alternatives according to these values. Using the MCDM problem in Example 11, the results with different methods are shown in Table 3.

According to the results presented in Table 3, if the distance methods in Xu and Xia (2011) and the Singh (2013) are used, then the best alternatives are $A_2$ and the worst one is $A_1$, respectively. Ye's method (2014c) say that the best ones are $A_4$ and the worst one is $A_1$. With respect to proposed method in this paper, the best one is $A_2$ and $A_1$ is the worst one. But, there are some small differences in the ranking of the alternatives due to definition of set theories. Additionally, from results of Table 3, we can say that the concept of distance measure is more remarkable and more useable than cosine measure to determine the order of the alternatives.

As mentioned above, the single valued neutrosophic hesitant fuzzy set is a generalization of FSs, IFSs, HFSs, FMSs, DHFSs and also SVNSs. Therefore a SVNHFS (truth-membership hesitant degree, indeterminacy-membership hesitant degree, and falsity-membership hesitant degree) contains more information than the HFS (membership hesitant degree), the IFS (both membership degree and nonmembership degree), the DHFS (membership hesitant degree and nonmembership hesitant degree), and also SVNS (truth-membership degree, indeterminacy-membership degree, and falsity-membership degree). Then, the proposed distance and similarity measures of SVNHFSs is a further generalization of the distance and similarity measures of FSs, IFSs, HFSs, FMSs, DHFSs and also SVNSs. In other words, the distance and similarity measures of FSs, IFSs, HFSs, FMSs, DHFSs and also SVNSs are special cases of the distance and similarity measures of SVNHFSs proposed in this paper. Therefore, the discrimination measures for SVNHFSs can be used to solve not only distance and similarity measures with SVNHFSs but also the problems of fuzzy environment, hesitant fuzzy environment, intuitionistic fuzzy environment, dual hesitant fuzzy environment and single valued neutrosophic environment, whereas the methods in Xu and Xia (2011), Xu (2007), Singh (2013) and Majumdar and Samanta (2014) are only sustainable for problems with HFSs, IFSs, DHFSs, and SVNSs, respectively. Moreover, since SVNHFSs include the aforementioned fuzzy sets, the decision-making method using the proposed distance and similarity measures is more general and more feasibility than existing decision-making methods in fuzzy setting, intuitionistic fuzzy setting, hesitant fuzzy setting, dual hesitant fuzzy setting, and single-valued neutrosophic setting.

## 8. Conclusions

Based on the combination of both HFSs and SVNSs as a further generalization of fuzzy concepts, the SVNHFS contains more information because it takes into account the information of its truth-membership hesitant degree, indeterminacy-membership hesitant degree, and falsity-membership hesitant degree, whereas the HFS only contains the information of its membership hesitant degrees and DHFS contains the information of its membership hesitant degree and nonmembership hesitant degree. Therefore, it has the desirable characteristics and advantages of its own, appears to be a more flexible method than the existing methods and include much more information given by decision makers. Based on the geometric distance model, the set-theoretic approach, and the matching functions, this paper proposed some distance and similarity measures between SVHNSs as a new extension of discrimination measures between fuzzy sets, hesitant fuzzy sets, dual hesitant fuzzy sets and the single-valued neutrosophic sets. In a multiple attribute decision making process with single-valued neutrosophic hesitant fuzzy information, the proposed distance measure between each alternative and the ideal alternative was used to rank the alternatives and determine the best one(s) according to the measure values. Finally, a numeric example was given to verify the proposed approach and to show its practicality and favorable. The developed method has useable and effective calculation, and presents a new model for handling decision-making problems under the single-valued neutrosophic hesitant fuzzy environment. In the future, we shall further develop more discrimination measures such as correlation coefficient, entropy and cross-entropy for SVNHFSs and apply them to solve practical applications in these areas, such as group decision making, expert system, clustering, information fusion system, fault diagnoses, and medical diagnoses.

## References

1. K. Atanassov, Intuitionistic fuzzy sets. Fuzzy Sets and Systems, (20) (1986) 87–96.
2. K. Atanassov and G. Gargov (1989) Interval-valued intuitionistic fuzzy sets Fuzzy Sets and Systems, 31(3) 343–349.
3. S. Broumi, F. Smarandache (2013) Correlation coefficient of interval neutrosophic set, Applied Mechanics and Materials, 436 511–517.
4. S.M. Chen, S.M. Yeh and P.H. Hsiao (1995) A comparison of similarity measures of fuzzy values. Fuzzy Sets and Systems, 72:79–89.
5. P. Grzegorzewski (2004) Distances between intuitionistic fuzzy sets and/or interval-valued fuzzy sets based on the Hausdorff metric. Fuzzy Sets and Systems, 148, 319–328.
6. W.L. Hung and M.S. Yang (2007) Similarity measures of intuitionistic fuzzy sets based on L p metric. International Journal of Approximate Reasoning, 46:120–136
7. W.L. Hung and M.S. Yang (2004) Similarity measures between type-2 fuzzy sets. International Journal of Uncertainty, Fuzziness and Knowledge-Based Systems, 12, 827–841.
8. P. Liu and Y. Wang, Multiple attribute decision-making method based on single-valued neutrosophic normalized weighted Bonferroni mean, Neural Computing and Applications, 25 (7-8) (2014) 2001-2010.
9. Y.H. Li, D.L. Olson and Z. Qin (2007) Similarity measures between intuitionistic fuzzy (vague) sets: a comparative analysis. Pattern Recognition Letters, 28:278–285
10. Z.Z. Liang, P.F. Shi (2003) Similarity measures on intuitionistic fuzzy sets. Pattern Recognition Letters, 24:2687–2693.
11. R. Şahin (2015) Fuzzy multicriteria decision making method based on the improved accuracy function for interval valued intuitionistic fuzzy sets. Soft Computing, DOI: 10.1007/s00500-015-1657-x
12. R. Şahin (2014), Neutrosophic hierarchical clustering algorithms, Neutrosophic Sets and Systems, 2,18-24.
13. R Şahin and A. Küçük (2014) Subsethood measures for single valued neutrosophic sets, Journal of Intelligent & Fuzzy Systems DOI: 10.3233/IFS-141304.
14. F. Smarandache (2005) A generalization of the intuitionistic fuzzy set. International journal of Pure and Applied Mathematics, 24 287-297.
15. F. Smarandache, A unifying field in logics. neutrosophy: Neutrosophic probability, set and logic, American Research Press, Rehoboth, 1999.
16. H. Wang, F. Smarandache, Y.Q. Zhang and R. Sunderraman, (2005) Interval neutrosophic sets and logic: Theory and applications in computing, Hexis, Phoenix, AZ.

17. H. Wang, F. Smarandache, Y.Q. Zhang and R. Sunderraman, (2010) Single valued neutrosophic sets, Multispace and Multistructure 4, 410–413.
18. J. Ye, (2013) Multicriteria decision-making method using the correlation coefficient under single-valued neutrosophic environment, International Journal of General Systems 42(4) 386–394.
19. J. Ye, (2014a) Single valued neutrosophic cross-entropy for multicriteria decision making problems, Applied Mathematical Modelling 38, 1170–1175.
20. J. Ye, (2014b) A multicriteria decision-making method using aggregation operators for simplified neutrosophic sets, Journal of Intelligent & Fuzzy Systems 26, 2459–2466.
21. J. Ye (2014c) Multiple-attribute Decision-Making Method under a single-valued neutrosophic hesitant fuzzy environment, Journal of Intelligent Systems, DOI: 10.1515/jisys-2014-0001.
22. J. Ye, (2014d) Similarity measures between interval neutrosophic sets and their applications in multicriteria decision-making, Journal of Intelligent & Fuzzy Systems, 26 (1) 165–172.
23. J.J, Peng, J.Q. Wang, J. Wang, H.Y. Zhang and X.H. Chen, (2015) Simplified neutrosophic sets and their applications in multi-criteria group decision-making problems. International Journal of Systems Science. DOI: 10.1080/00207721.2014.994050.
24. B. Farhadinia (2014) An efficient similarity measure for intuitionistic fuzzy sets. Soft Computing, 18(1):85–94
25. P. Majumdar and S.K. Samanta, On similarity and entropy of neutrosophic sets, Journal of Intelligent & Fuzzy Systems, 26 (3) (2014) 1245–1252.
26. J.J. Peng, J.Q. Wang, X.H. Wu, J. Wang and X.H. Chen, (2015) Multi-valued neutrosophic sets and power aggregation operators with their applications in multi-criteria group decision-making problems. International Journal of Computational Intelligence Systems 8(2) 345-363.
27. P. Singh (2013) Distance and similarity measures for multiple-attribute decision making with dual hesitant fuzzy sets, Computational Applied Mathematics. DOI: 10.1007/s40314-015-0219-2.
28. S. Khaleie and M. Fasanghari (2012) An intuitionistic fuzzy group decision making method using entropy and association coefficient. Soft Computing, 16(7):1197–1211
29. E. Szmidt and J. Kacprzyk (2000) Distances between intuitionistic fuzzy sets. Fuzzy Sets and Systems, 114:505–518
30. V. Torra (2010) Hesitant fuzzy sets. International Journal of Intelligent Systems, 25:529–539
31. V. Torra and Y. Narukawa (2009) On hesitant fuzzy sets and decision. In: The 18th IEEE international conference on fuzzy systems, Jeju Island, Korea, 2009, pp 1378–1382
32. W.Q. Wang (1997) New similarity measure on fuzzy sets and on elements. Fuzzy Sets and Systems, 85:305–309
33. W.Q. Wang and X.L. Xin (2005) Distance measure between intuitionistic fuzzy sets. Pattern Recognition Letters, 26:2063–2069.
34. M.M. Xia and Z.S. Xu (2011) Hesitant fuzzy information aggregation in decision making, International Journal of Approximate Reasoning, 52 (3) 395–407.
35. Z. Liang and P. Shi (2003) Similarity measures on intuitionistic fuzzy sets. Pattern Recognition Letters, 24 2687–2693.
36. Z. Xu (2007) Some similarity measures of intuitionistic fuzzy sets and their applications to multiple attribute decision making. Fuzzy Optimization and Decision Making, 6:109–121.
37. Z. Xu and M. Xia (2011) Distance and similarity measures for hesitant fuzzy sets. Information Science, 181:2128–2138
38. L. Xuecheng, (1992) Entropy, distance measure and similarity measure of fuzzy sets and their relations, Fuzzy Sets and Systems, 52 (3) 305–318.
39. C. Tan (2011) Generalized intuitionistic fuzzy geometric aggregation operator and its application to multi-criteria group decision making. Soft Computing, 15(5):867–876
40. L.A. Zadeh (1965) Fuzzy sets. Information Control, 8:338–353.

41. H.Y. Zhang, J.Q. Wang, X.H. Chen, (2014) Interval neutrosophic sets and their application in multicriteria decision making problems, The Scientific World Journal, 645953.
42. HC. Liu, JX. You, MM. Shan, LN. Shao (2015) Failure mode and effects analysis using intuitionistic fuzzy hybrid TOPSIS approach. Soft Computing, 19(4):1085–1098
43. B. Zhu, Z. Xu and M. Xia (2012) Dual hesitant fuzzy sets. Journal of Applied Mathematics, 2012:1–13

PRANAB BISWAS[1], SURAPATI PRAMANIK[2*], BIBHAS C. GIRI[3]

1 Department of Mathematics, Jadavpur University, Kolkata, 700032, India. E-mail: paldam2010@gmail.com
2* Department of Mathematics, Nandalal Ghosh B.T. College, Panpur, P.O.-Narayanpr, District-North 24 Parganas, West Bengal, PIN-743126, India. Corresponding author's E-mail: sura_pati@yahoo.co.in
3 Department of Mathematics, Jadavpur University, Kolkata, 700032, India.Email: bcgiri.jumath@gmail.com

# GRA Method of Multiple Attribute Decision Making with Single Valued Neutrosophic Hesitant Fuzzy Set Information

## Abstract

Single valued neutrosophic hesitant fuzzy set has three independent parts, namely the truth membership hesitancy function, indeterminacy membership hesitancy function, and falsity membership hesitancy function, which are in the form of sets that assume values in the unit interval [0, 1]. Single valued neutrosophic hesitant fuzzy set is considered as a powerful tool to express uncertain, incomplete, indeterminate and inconsistent information in the process of multi attribute decision making problems. In this paper we study multi attribute decision making problems in which the rating values are expressed with single valued neutrosophic hesitant fuzzy set information. Firstly, we define score value and accuracy value to compare single valued neutrosophic hesitant fuzzy sets and then define normalised Hamming distance between the single valued neutrosophic hesitant fuzzy sets. Secondly, we propose the grey relational analysis method for multi attribute decision making under single valued neutrosophic hesitant fuzzy set environment. Finally, we provide an illustrative example to demonstrate the validity and effectiveness of the proposed method.

## 1. Introduction

Multi-attribute decision making (MADM) used in human activities is a useful process for selecting the best alternative that has the highest degree of satisfaction from a set of feasible alternatives with respect to the attributes. Because the real world is fuzzy rather than precise in nature, the rating values of alternative with respect to attribute considered in MADM problems are often imprecise or incomplete in nature. This has led to the development of the fuzzy set theory proposed by Zadeh [1]. Fuzzy set theory has been proved to be an effective tool in MADM process [2-6]. However, fuzzy set can represent imprecise information with membership degree only. The intuitionistic fuzzy set (IFS) proposed by Attanasov [7], a generalisation of fuzzy sets, is characterized by membership and non-membership functions where non-membership is independent. Recently, IFS has been successfully applied in many decision making problems, especially in MADM problems [8-12].

However IFS can handle incomplete information and but it cannot express indeterminate and inconsistent information with membership and non-membership functions. Smarandache [13] introduced the neutrosophic set (NS) from philosophical point of view to deal with uncertain, imprecise, incomplete and inconsistent information that exist in real world. NS is characterised with truth membership, indeterminacy and falsity membership degree, which are independent in nature. This set generalises the concept of crisp set, fuzzy set, intuitionistic fuzzy set, paraconsistent set, dialetheist set, paradoxist set, and tautological set. Since the introduction of NS and single-valued neutrosophic set proposed by Wang et al. [14] in 2010, the model of decision making under neutrosophic environment has been received much attention to the researchers. Many methods of MADM such as TOPSIS method [15, 16], grey relational analysis (GRA) method [17,18], distance and similarity measure method [19-23], and outranking method [24] were developed under neutrosophic environment.

However, in a decision making process sometimes decision maker may feel hesitate to take decision among the set of possible values instead of single value. Tora [25], Tora and Narukawa [26] introduced the hesitant fuzzy set (HF), which permits the membership degree of an element to a given set to be represented by the set of possible numerical values in [0,1]. HF, an extension of fuzzy set, is useful to deal uncertain information in the process of MADM. Xia and Xu [27] proposed some aggregation operators for hesitant fuzzy information and applied them to MADM problem in hesitant fuzzy environment. Wei [28] studied some models for hesitant fuzzy MADM problem by developing some prioritized aggregation operators for hesitant fuzzy information. Xu and Zhang [29] developed TOPSIS method for hesitant fuzzy MADM with incomplete weight information.

Decision maker does not consider the non-membership degrees of rating values in hesitant fuzzy MADM. However, non-membership degrees play an important role to express incomplete information. Zhu et al. [30] gave the idea of the dual hesitant fuzzy set (DHFS), in which membership degrees and non-membership degrees are in the form of sets of values in [0,1]. DHFS generalizes the HF sets and expresses incomplete information effectively. Ye [31] and Chen et al.[32] proposed co-relation method between DHFSs and applied the method to MADM with hesitant fuzzy information. Singh [33] defined and applied distance and similarity measure between DHFSs in MADM. However in a decision making process, indeterminate type information cannot be captured with DHFS.

In 2014, Ye [34] introduced single-valued neutrosophic hesitant fuzzy set (SVNHFS) by coordinating HFS and SVNS. SVNHFS generalises the FS, IFS, HFS, DHFS and SVNS, and can represent uncertain, imprecise, incomplete and inconsistent information. SVNHFSs are characterized by truth hesitancy, indeterminacy hesitancy and falsity-hesitancy membership functions which are independent. Therefore SVNHFS can express the three kinds of hesitancy information that exist in MADM in real situations. Ye [34] developed single valued neutrosophic hesitant fuzzy weighted averaging and single valued neutrosophic hesitant fuzzy weighted geometric operators for SVNHFS information and applied these two operators in MADM. Liu and Shi [35] proposed hybrid weighted average operator for interval neutrosophic hesitant fuzzy set in which the truth hesitancy, indeterminacy hesitancy and falsity-hesitancy membership functions are in the form of sets of interval values contained in [0, 1]. Sahin and Liu [36] defined co-relation co-efficient between SVNHFSs and used it for MADM.

Grey relational analysis (GRA)[37], a part of grey system theory, is successfully applied in solving a variety of MADM problems in intuitionistic fuzzy environment [38-42], neutrosophic environment [43], interval neutrosophic environment [44, 45, 46], neutrosophic soft set environment [47-49], rough neutrosophic environment [50] respectively. However, literature review reflects that GRA method of MADM with SVNHFS has not been studied in the literature. Therefore we need attention for this issue. The aim of the paper is to extend the concept of GRA method for solving MADM problem in which the rating values of the alternatives over the attributes are considered with SVNHFSs.

The rest of the paper is organised as follows: Section 2 presents some basic concept related to SVNHFSs. In Section 3, we propose GRA method for MADM problems, where rating values are considered with SVNHFSs. In Section 4, we illustrate our proposed method with an example. Section 5 presents concluding remarks of the study.

## 2. Preliminaries

In this section we recall some basic definitions of hesitant fuzzy set, single valued neutrosophic hesitant fuzzy set, score function accuracy function of triangular fuzzy intuitionistic fuzzy numbers.

**Definition 1.** [25]Let $X$ be a fixed set, then a hesitant fuzzy set (HFS) $A$ on $X$ is in terms of a function that when applied to $X$ returns a subset of $[0,1]$, i.e.,

$A=\{\langle x,h_A(x)\rangle \mid x\in X\}$, where, $h_A(x)$ is a set of some different values in $[0,1]$, representing the possible membership degrees of the element $x\in X$ to $A$. For convenience, $h_A(x)$ is called a hesitant fuzzy element (HFE).

**Definition 2.** [34] Let $X$ be fixed set, then a single valued hesitant fuzzy element (SVHFE) $N$ on $X$ is defined as $N=\{\langle x,t(x),i(x),f(x)\rangle \mid x\in X\}$ (1) where $t(x)$, $i(x)$ and $f(x)$ represent three sets of values in $[0,1]$, denoting respectively the possible truth, indeterminacy and falsity membership degree of the element $x\in X$ to the set $N$. The membership degrees $t(x)$, $i(x)$ and $f(x)$ satisfy the following conditions:

$$0\le \delta,\gamma,\eta\le 1;\ 0\le \delta^+ +\gamma^+ +\eta^+\le 3 \tag{2}$$

where, $\delta\in t(x),\gamma\in i(x),\eta\in f(x)$, $\delta^+\in t^+(x)=\bigcup_{\delta\in t(x)}\max t(x)$, $\gamma^+\in i^+(x)=\bigcup_{\delta\in t(x)}\max i(x)$, $\eta^+\in f^+(x)=\bigcup_{\delta\in t(x)}\max f(x)$ for all $x\in X$.

For convenience, the triplet $n(x)=\langle t(x),i(x),f(x)\rangle$ is called a SVNHFE denoted by $n=\langle t,i,f\rangle$. Note that the number of values for possible truth, indeterminacy and falsity membership degrees of the element in different SVNHFEs may be different.

**Definition 3.** [34] Let $n_1=\langle t_1,i_1,f_1\rangle$ and $n_2=\langle t_2,i_2,f_2\rangle$ be two SVNHFEs, the following operational rules are defined as follows:

7. $n_1\oplus n_2=\left\langle \bigcup_{\delta_1\in t_1,\gamma_1\in i_1,\eta_1\in f_1,\ \delta_2\in t_2,\gamma_2\in i_2,\eta_2\in f_2}\{\{t_1+t_2-t_1t_2\},\{i_1i_2\},\{f_1,f_2\}\}\right\rangle$;

8. $n_1\otimes n_2=\left\langle \bigcup_{\delta_1\in t_1,\gamma_1\in i_1,\eta_1\in f_1,\ \delta_2\in t_2,\gamma_2\in i_2,\eta_2\in f_2}\{\{t_1t_2\},\{i_1+i_2-i_1i_2\},\{f_1+f_2-f_1f_2\}\}\right\rangle$;

9. $\lambda n_1=\left\langle \bigcup_{\delta_1\in t_1,\gamma_1\in i_1,\eta_1\in f_1}\{\{1-(1-t_1)^\lambda\},\{i_1^\lambda\},\{f_1^\lambda\}\}\right\rangle,\lambda>0$ ;

10. $n_1^\lambda=\left\langle \bigcup_{\delta_1\in t_1,\gamma_1\in i_1,\eta_1\in f_1}\{\{t_1^\lambda\},\{1-(1-i_1)^\lambda\},\{1-(1-f_1)^\lambda\}\}\right\rangle,\lambda>0.$

**Definition 4.** Let $n_i = \langle t_i, i_i, f_i \rangle$ $(i = 1,2,...,n)$ be a collection of SVNHFEs, then the score function $S(n_i)$, and accuracy function $A(n_i)$ of $n_i (i = 1,2,...,n)$ can be defined as follows:

1. $$S(n_i) = \frac{1}{3}\left[2 + \frac{1}{l_t}\sum_{\delta \in t}\delta - \frac{1}{l_i}\sum_{\gamma \in i}\gamma - \frac{1}{l_f}\sum_{\eta \in f}\eta\right] \quad (3)$$
2. $$A(n_i) = \frac{1}{l_t}\sum_{\delta \in t}\delta - \frac{1}{l_f}\sum_{\eta \in f}\eta; \quad (4)$$

where, $l_t$, $l_i$, and $l_f$, are the numbers of values of $t_i$, $i_i$, and $f_i$ respectively in $n_i$.

**Definition 5.** Let $n_1 = \langle t_1, i_1, f_1 \rangle$ and $n_2 = \langle t_2, i_2, f_2 \rangle$ be two SVNHFEs, the following rules can be defined for comparison purposes:

1. If $S(n_1) > S(n_2)$, then $n_1$ is greater than $n_2$ and denoted by $n_1 \succ n_2$;
2. If $S(n_1) = S(n_2)$ and $A(n_1) > A(n_2)$, then $n_1 \succ n_2$;
3. If $S(n_1) = S(n_2)$ and $A(n_1) = A(n_2)$, then $n_1 \approx n_2$.

**Definition 6.** Let $n_1 = \langle t_1, i_1, f_1 \rangle$ and $n_2 = \langle t_2, i_2, f_2 \rangle$ be two SVNHFEs, the normalised Hamming distance is defined as

$$D(n_1,n_2) = \frac{1}{3}\left(\left|\frac{1}{l_{t_1}}\sum_{\delta_1 \in t_1}\delta_1 - \frac{1}{l_{t_2}}\sum_{\delta_2 \in t_2}\delta_2\right| + \left|\frac{1}{l_{i_1}}\sum_{\gamma_1 \in i_1}\gamma_1 - \frac{1}{l_{i_2}}\sum_{\gamma_2 \in i_2}\gamma_2\right| + \left|\frac{1}{l_{f_1}}\sum_{\eta_1 \in f_1}\eta_1 - \frac{1}{l_{f_2}}\sum_{\eta_2 \in f_2}\eta_2\right|\right) \quad (5)$$

where $l_{t_k}$, $l_{i_k}$, and $l_{f_k}$ are the possible membership values in $n_k$ for $k = 1,2$, respectively.

The distance function $D(n_1, n_2)$ of two SVNHFEs $n_1$ and $n_2$ satisfies the following properties:

1. $0 \le D(n_1, n_2) \le 1;$
2. $D(n_1, n_2) = 0$ if and only if $n_1 = n_2;$
3. $D(n_1, n_2) = D(n_2, n_1);$
4. If $n_1 \le n_2 \le n_3$, and $n_3$ is an SVNHFE on $X$, then $D(n_1,n_2) \le D(n_1,n_3)$ and $D(n_2,n_3) \le D(n_1,n_3)$.

## 3. GRA method for multi-attribute decision making with SVNHFS information

In this section, we propose GRA based approach to find out the best alternative in multi-attribute decision making problem in SVNHFS environment. Assume that $A = \{A_1, A_2, ..., A_m\}$ be the discrete set of $m$ alternatives and $C = \{C_1, C_2, ..., C_n\}$ be the set of $n$ attributes for a multi-attribute decision making problem. Suppose that the rating values of the $i-$th alternative $A_i (i = 1,2,...,m)$ over the attribute $C_j (j = 1,2,...,n)$ are expressed in terms of SVNHFSs $x_{ij} = \langle t_{ij}, i_{ij}, f_{ij} \rangle$, where $t_{ij} = \{\delta_{ij} \mid \delta_{ij} \in t_{ij}, 0 \le \delta_{ij} \le 1\}$, $i_{ij} = \{\gamma_{ij} \mid \gamma_{ij} \in i_{ij}, 0 \le \gamma_{ij} \le 1\}$, and $f_{ij} = \{\eta_{ij} \mid \eta_{ij} \in f_{ij}, 0 \le \eta_{ij} \le 1\}$ are the possible truth, indeterminacy and falsity membership degrees, respectively. With these rating values, we can construct a decision matrix $X = (x_{ij})_{m \times n}$, where the entries of this matrix are SVNHFSs. The decision matrix can be presented as follows:

$$X = \begin{bmatrix} x_{11} & x_{12} & \dots x_{1n} \\ x_{21} & x_{22} & \dots x_{2n} \\ \vdots & \vdots & \ddots \vdots \\ x_{m1} & x_{m2} & \dots x_{mn} \end{bmatrix} \quad (6)$$

We develop the GRA method using the following steps by considering the weight vector $W = (w_1, w_2, ..., w_n)^T$ of attributes where $w_j \in [0,1]$ and $\sum_{j=1}^{n} w_j = 1.$

**Step 1.** Determine the single valued neutrosophic hesitant fuzzy positive ideal solution (SVNHFPIS) $A^{+}$ and the single valued neutrosophic hesitant fuzzy negative ideal solution (SVNHFNIS) $A^{-}$ of alternatives in the decision matrix $X$ by the following equations, respectively:

$$A^{+}=\begin{cases}\max\limits_{1\le i\le m}(x_{i1}),\max\limits_{1\le i\le m}(x_{i2}),...,\max\limits_{1\le i\le m}(x_{in})\text{ for benefit type attribute;}\\ \min\limits_{1\le i\le m}(x_{i1}),\min\limits_{1\le i\le m}(x_{i2}),...,\min\limits_{1\le i\le m}(x_{in})\text{ for cost type attribute}\end{cases} \tag{7}$$
$$=\left(A_{i1}^{+},A_{i2}^{+},...,A_{in}^{+}\right)$$

$$A^{-}=\begin{cases}\min\limits_{1\le i\le m}(x_{i1}),\min\limits_{1\le i\le m}(x_{i2}),...,\min\limits_{1\le i\le m}(x_{in})\text{ for benefit type attribute;}\\ \max\limits_{1\le i\le m}(x_{i1}),\max\limits_{1\le i\le m}(x_{i2}),...,\max\limits_{1\le i\le m}(x_{in})\text{ for cost type attribute}\end{cases} \tag{8}$$
$$=\left(A_{i1}^{-},A_{i2}^{-},...,A_{in}^{-}\right)$$

The rating values $x_{ij}$ can be compared by the score function $S(x_{ij})$ and accuracy function $A(x_{ij})$ defined in Definition 3.

**Step 2.** Determine the grey relational co-efficient of each alternative from $A^{+}$ and $A^{-}$ by the following equations:

$$\xi_{ij}^{+}=\frac{\min\limits_{1\le i\le m}\min\limits_{1\le i\le m}D(x_{ij},A_j^{+})+\max\limits_{1\le i\le m}\max\limits_{1\le i\le m}D(x_{ij},A_j^{+})}{D(x_{ij},A_j^{+})+\rho\max\limits_{1\le i\le m}\max\limits_{1\le i\le m}D(x_{ij},A_j^{+})} \tag{9}$$

$$\xi_{ij}^{-}=\frac{\min\limits_{1\le i\le m}\min\limits_{1\le i\le m}D(x_{ij},A_j^{-})+\max\limits_{1\le i\le m}\max\limits_{1\le i\le m}D(x_{ij},A_j^{-})}{D(x_{ij},A_j^{-})+\rho\max\limits_{1\le i\le m}\max\limits_{1\le i\le m}D(x_{ij},A_j^{-})} \tag{10}$$

where the identification co-efficient is considered as $\rho=0.5$.

**Step 3.** Calculate the degree of grey relational coefficient of each alternative $A_i(i=1,2,...,m)$ from $A^{+}$ and $A^{-}$ by the following equations:

$$\xi_i^{+}=\sum_{j=1}^{n}w_j\xi_{ij}^{+} \tag{11}$$

$$\xi_i^{-}=\sum_{j=1}^{n}w_j\xi_{ij}^{-} \tag{12}$$

**Step 4.** Calculate the relative closeness co-efficient $\xi_i$ for each alternative $A_i(i=1,2,..,m)$ with respect to the positive ideal solution $A^{+}$ as

$$\xi_i=\frac{\xi_i^{+}}{\xi_i^{+}+\xi_i^{-}}\text{ for } i=1,2,..,m. \tag{13}$$

**Step 5.** Rank the alternative according the relative closeness co-efficient $\xi_i(i=1,2,..,m)$.

## 4. A Numerical Example

In this section we consider the example adopted from Ye [34] to illustrate the application of the proposed GRA method for MADM proposed in Section 4. Consider an investment company that wants to invest a sum of money in the best option. The following four possible alternatives are considered to invest the money:

1. $A_1$ is the car company;
2. $A_2$ is the food company;
3. $A_3$ is the computer company;
4. $A_4$ is the arms company.

The investment company must take a decision according to the following three attributes:

1. $C_1$ is the risk analysis;
2. $C_2$ is the growth analysis;
3. $C_3$ is the environmental impact analysis.

The attribute weight vector is given as $W=(0.35,0.25.0.40)^T$ . The four possible alternatives $\{A_1,A_2,A_3,A_4\}$ are evaluated using SVNHFEs under three attributes $C_j(j=1,2,3)$. We can arrange the rating values in a matrix form called a SVNHF decision matrix $X=(x_{ij})_{4\times3}$ (see Table-1).

Table 1. *Single valued neutrosophic hesitant fuzzy decision matrix*

| $C_1$ | $C_2$ | $C_3$ |
|---|---|---|
| $\{\{0.3,0.4,0.5\},\{0.1\},\{0.3,0.4\}\}$ | $\{\{0.5,0.6\},\{0.2,0.3\},\{0.3,0.4\}\}$ | $\{\{0.3,0.4,0.5\},\{0.1\},\{0.3,0.4\}\}$ |
| $\{\{0.6,0.7\},\{0.1,0.2\},\{0.2,0.3\}\}$ | $\{\{0.6,0.7\},\{0.1\},\{0.3\}\}$ | $\{\{0.3,0.4,0.5\},\{0.1\},\{0.3,0.4\}\}$ |
| $\{\{0.5,0.6\},\{0.4\},\{0.2,0.3\}\}$ | $\{\{0.6\},\{0.3\},\{0.4\}\}$ | $\{\{0.5,0.6\},\{0.1\},\{0.3\}\}$ |
| $\{\{0.7,0.8\},\{0.1\},\{0.1,0.2\}\}$ | $\{\{0.6,0.7\},\{0.1\},\{0.2\}\}$ | $\{\{0.3,0.5\},\{0.2\},\{0.1,0.2,0.3\}\}$ |

Now we apply the proposed method to find out the best alternative, which can be described as follows:

**Step 1.** Comparing the attribute values by score function and accuracy function of SVNHFEs, we can determine the neutrosophic hesitant fuzzy positive ideal solution (SVNHFPIS) $A^+$ by the Eq.(7) as follows:

$$\begin{aligned} A^+ &= \left[\{\{0.7,0.8\},\{0.1\},\{0.1,0.2\}\},\{\{0.6,0.7\},\{0.1\},\{0.2\}\},\{\{0.6,0.7\},\{0.1,0.2\},\{0.1,0.2\}\}\right] \\ &= \left[A_1^+,A_2^+,A_3^+\right] \end{aligned} \quad (14)$$

Similarly, we can determine the neutrosophic hesitant fuzzy negative ideal solution (SVNHFPIS) $A^-$ by the Eq.(8) as follows:

$$\begin{aligned} A^- &= \left[\{\{0.5,0.6\},\{0.4\},\{0.2,0.3\}\},\{\{0.6\},\{0.3\},\{0.4\}\},\{\{0.2,0.3\},\{0.1,0.2\},\{0.5,0.6\}\}\right] \\ &= \left[A_1^-,A_2^-,A_3^-\right] \end{aligned} \quad (15)$$

**Step 2.** Calculate the grey relational co-efficient of each alternative from positive ideal solutions $A^+$ and negative ideal solutions $A^-$ by equations (9) and (10) for $\rho=0.5$ , respectively.

$$\xi_{ij}^+ = \begin{bmatrix} 0.4218 & 0.5010 & 0.3333 \\ 0.6166 & 0.8018 & 1.0000 \\ 0.4003 & 0.4709 & 0.5717 \\ 1.0000 & 1.0000 & 0.5350 \end{bmatrix} \quad (16)$$

$$\xi_{ij}^+ = \begin{bmatrix} 0.4218 & 0.7275 & 1.0000 \\ 0.5329 & 0.5329 & 0.3333 \\ 1.0000 & 1.0000 & 0.4218 \\ 0.4003 & 0.4709 & 0.4218 \end{bmatrix}$$

(17)

Here, we consider $i=1,2,3,4$ and $j=1,2,3.$

**Step 3.**Calculate the degree of grey relational co-efficient of each alternative from $A^+$ and $A^-$ by Eqs. (11) and (12), respectively.

$\xi_1^+ = 0.4062$ $\xi_2^+ = 0.8162$ $\xi_3^+ = 0.4865$ $\xi_4^+ = 0.8140$ (18)

$\xi_1^- = 0.7295$ $\xi_2^- = 0.4531$ $\xi_3^- = 0.7687$ $\xi_4^- = 0.4265$ (19)

**Step 4.**Calculate the relative closeness coefficient $\xi_i$ for each alternative $A_i (i = 1,2,3,4)$ by Eq.(13).

$\xi_1 = 0.3577$, $\xi_2 = 0.6430$, $\xi_3 = 0.3875$, and $\xi_4 = 0.6561$.

**Step 5.** Rank the alternative according to the relative closeness coefficient $\xi_i (i = 1,2,3,4)$.

Therefore $A_4 \succ A_2 \succ A_3 \succ A_1$ indicates that the most desirable alternative is $A_4$.

We notice that the ranking order obtained by the proposed method is indifferent with the ranking of the alternative obtained by Ye's method [34].

## 5. Conclusions

In general, the information of rating values considered in MADM problems is imprecise, indeterminate, incomplete and inconsistent in nature. SVNHFS is a useful tool that can capture all these type of information in MADM process. In this paper we investigate MADM problem in which rating values are considered with SVNHFSs. To extend the GRA method for MADM, we first define score value, accuracy value, certainty value, and normalised Hamming distance of SVNHFS. Having defined the positive ideal solution (PIS) and the negative ideal solution (NIS) by score value and accuracy value, we calculate the grey relational degree between each alternative and ideal alternatives (PIS and NIS). Then we determine a relative relational degree to obtain the ranking order of all alternatives by calculating the degree of grey relation to both the positive and negative ideal solution simultaneously. Finally, we provide an illustrative example to show the validity and effectiveness of the proposed approach. The proposed approach is compared with other existing methods to show that our approach is straightforward and can be applied effectively with other decision making problems under SVNHF environment. In future, we will extend the proposed approach to MADM under SVNHFS environment with unknown weight information and MADM with interval valued neutrosophic hesitant fuzzy environment.

## References

1. L.A. Zadeh, Fuzzy sets, Information Control, 8(1965) 338–353.
2. R. Bellman, L.A. Zadeh, Decision making in a fuzzy environment, Management Science 17B (4)(1970) 141-164.
3. C.L Hwang, K. Yoon, Multiple attribute decision making: Methods and Applications, Springer-Verlag, Berlin, 1981.
4. S.J. Chen, C.L Hwang, Fuzzy multiple attribute decision making: Methods and Applications, Springer-Verlag, Berlin, 1992.
5. L. Zeng, Expected value method for fuzzy multiple attribute decision making, Tsinghua Science and Technology 11(2006) 102-106.
6. C.T. Chen, Extension of TOPSIS for group decision-making under fuzzy environment, Fuzzy Sets and Systems 114(2000) 1-9.
7. K.T. Atanassov, Intuitionistic fuzzy sets, Fuzzy Sets and Systems 20(1986) 87–96.
8. E. Szmidt, J. Kacprzyk, Using intuitionistic fuzzy sets in group decision making, Control and Cybernetics 31(2002) 1037-1053.
9. Z. Xu, Intuitionistic preference relations and their applications in group decision making, Information Sciences 117(2007) 2363-2379.
10. DF, Li, YC, Wang, S, Liu, F, Shan. Fractional programming methodology for multi-attribute group decision making using IFS, Applied Soft Computing 9(2009) 219-225.
11. G.W. Wei, Gray relational analysis method for intuitionistic fuzzy multiple attribute decision making, Expert Systems and Applications 38(2011) 11671-11677.

12. S. Pramanik, D. Mukhopadhyaya. Grey relational analysis based intuitionistic fuzzy multi criteria group decision-making approach for teacher selection in higher education. International Journal of Computer Applications 34(10) (2011):21-29.
13. F. Smarandache, A unifying field in logics, neutrosophy: neutrosophic probability, set and logic. American Research Press, Rehoboth, 1998.
14. H.Wang, F. Smarandache, R. Sunderraman, Y.Q. Zhang, Single-valued neutrosophic sets, Multi space and Multi structure. 4(2010) 410–413.
15. P. Biswas P, S. Pramanik, B.C. Giri, TOPSIS method for multi-attribute group decision-making under single-valued neutrosophic environment, Neural Computing and Applications 2015, doi: 10.1007/s00521-015-1891-2.
16. P. Chi, P. Liu, An extended TOPSIS method for the multiple attribute decision making problems based on interval neutrosophic set, Neutrosophic Sets and Systems 1(1)(2013) 63-70.
17. P. Biswas, S, Pramanik, B.C. Giri , Entropy based grey relational analysis method for multi-attribute decision making under single valued neutrosophic assessments, Neutrosophic Sets and Systems 2(2014) 102–110.
18. P. Biswas P, S. Pramanik, B.C. Giri, A new methodology for neutrosophic multi-attribute decision making with unknown weight information, Neutrosophic Sets and Systems 3(2014) 42–52.
19. S. Broumi, F. Smarandache, Several similarity measures of neutrosophic sets, Neutrosophic Sets and Systems1(2013) 54–62.
20. J. Ye, Similarity measures between interval neutrosophic sets and their multi-criteria decision- making method, Journal of Intelligent & Fuzzy Systems 26(2014) 165-172.
21. S. Pramanik, P. Biswas, B. Giri, Hybrid vector similarity measures and their applications to multi-attribute decision making under neutrosophic environment, Neural Computing and Applications 2015, 1–14. doi: 10.1007/s00521-015-2125-3.
22. P. Biswas, S, Pramanik, B.C. Giri, Cosine similarity measure based multi-attribute decision-making with trapezoidal fuzzy neutrosophic numbers, Neutrosophic Sets and Systems 8(2015) 47–57.
23. K. Mondal, S. Pramanik, Neutrosophic refined similarity measure based on cotangent function and its application to multi-attribute decision making, Global Journal of Advanced Research 2(2)(2015) 486-496.
24. J. Peng, J. Wang, H. Zhang, X. Chen, An outranking approach for multi-criteria decision-making problems with simplified neutrosophic sets, Applied Soft Computing 25(2014) 336-346.
25. V. Torra, Hesitant fuzzy sets, International Journal of Intelligent Systems 25(2010) 529-539.
26. V. Torra, Y. Narukawa, On hesitant fuzzy sets and decision in: The 18th IEEE International Conference on Fuzzy Systems, Jeju Island, Korea, 2009 1378-1382.
27. M.M. Xia, Z.S. Xu, Hesitant fuzzy information aggregation in decision making, International Journal of Approximate Reasoning 52(2011) 395-407.
28. G.W. Wei, Hesitant fuzzy prioritized operators and their application to multi-attribute decision making, Knowledge-Based Systems 31(2012) 176-182.
29. Z.S. Xu, X. Zhang, Hesitant fuzzy multi attribute decision making based on TOPSIS with incomplete weight information, Knowledge-Based Systems 52(2013) 53-64.
30. B. Zhu, Z.S. Xu, M.M. Xia, Dual hesitant fuzzy sets, Journal of Applied Mathematics (2012) doi: 10.1155/2012/879629.
31. J. Ye, Correlation coefficient of dual hesitant fuzzy sets and its application to multiple attribute decision making, Applied Mathematical Modelling 38(2014) 659-666.
32. Y.F. Chen, X.D. Peng, G.H. Guan, H.D. Jiang, Approaches to multiple attribute decision making based on the correlation coefficient with dual hesitant fuzzy information, Journal of Intelligent and Fuzzy Systems 26(2014) 2547-2556.
33. P. Singh, Distance and similarity measures for multiple attribute decision making with dual hesitant fuzzy sets, Comp. Appl. Math. (2015) doi: 10.1007/s40314-015-0219-2.
34. J. Ye, Multiple-attribute decision making under a single-valued neutrosophic hesitant fuzzy environment, Journal of Intelligent Systems (2014) doi: 10.1515/jisys-2014-0001.

35. P. Liu, L Shi, The generalized hybrid weighted average operator based on interval neutrosophic hesitant set and its application to multiple attribute decision making, Neural Computing and Applications26 (2015) 457-471.
36. R. Sahin, P Liu, Correlation coefficient of single-valued neutrosophic hesitant fuzzy sets and its applications in decision making, Neural Computing and Applications (2016) doi: 10.1007/s00521-015-2163-x.
37. J.L. Deng, Introduction to grey systems theory, The Journal of Grey Systems 1(1) (1989) 1-24.
38. G. Wei, GRA method for multiple attribute decision making with incomplete weight information in intuitionistic fuzzy setting, Knowledge-Based Systems 23(3) (2010) 243-247.
39. X. Zhang, F. Jin, P. Liu, A grey relational projection method for multi-attribute decision making based on intuitionistic trapezoidal fuzzy number, Applied Mathematical Modelling 37(5)(2013) 3467-3477.
40. S.F Zhang, S.Y. Liu, A GRA-based intuitionistic fuzzy multi-criteria group decision making method for personal selection, Expert Systems With Applications 38(9)(2011) 11401-11405.
41. K. Mondal, S. Pramanik, Intuitionistic fuzzy multicriteria group decision making approach to quality-brick selection problem, Journal of Applied Quantitative Methods 9(2) (2014) 35-50.
42. P.P. Dey, S. Pramanik, B.C. Giri, Multi-criteria group decision making in intuitionistic fuzzy environment based on grey relational analysis for weaver selection in Khadi institution, Journal of Applied and Quantitative Methods 10(4) (2015) 1-14.
43. K. Mondal, S. Pramanik, Neutrosophic decision making model for clay-brick selection in construction field based on grey relational analysis, Neutrosophic Sets and Systems 9 (2015) 64-71.
44. S. Pramanik, K. Mondal, Interval neutrosophic multi-attribute decision-making based on grey relational analysis, Neutrosophic Sets and Systems 9 (2015)13-22.
45. P.P. Dey, S. Pramanik, & B.C. Giri, An extended grey relational analysis based multiple attribute decision making in interval neutrosophic uncertain linguistic setting, Neutrosophic Sets and Systems 11 (2016) 21-30.
46. P.P Dey, S. Pramanik, B.C. Giri, An extended grey relational analysis based interval neutrosophic multi-attribute decision making for weaver selection, Journal of New Theory 9 (2015) 82-93.
47. S. Pramanik, S. Dalapati, GRA based multi criteria decision making in generalized neutrosophic soft set environment, Global Journal of Engineering Science and Research Management 3(5) (2016) 153-169.
48. P.P. Dey, S. Pramanik, & B.C. Giri, Neutrosophic soft multi-attribute group decision making based on grey relational analysis method, Journal of New Results in Science 10 (2016) 25-37.
49. P.P. Dey, S. Pramanik, & B.C. Giri, Neutrosophic soft multi-attribute decision making based on grey relational projection method, Neutrosophic Sets and Systems 11 (2016) 98-106.
50. K. Mondal, S. Pramanik, Rough neutrosophic multi-attribute decision-making based on grey relational analysis, Neutrosophic Sets and Systems 7 (2015) 8-17.

**PARTHA PRATIM DEY[1], SURAPATI PRAMANIK[2, *], BIBHAS C. GIRI[3]**

1, 3 Department of Mathematics, Jadavpur University, Kolkata-700032, West Bengal, India.
2* Department of Mathematics, Nandalal Ghosh B.T. College, Panpur, P.O.-Narayanpur, District North 24 Parganas, Pin code-743126, West Bengal, India. Corresponding author's E-mail: sura_pati@yahoo.co.in

# TOPSIS for Solving Multi-Attribute Decision Making Problems under Bi-Polar Neutrosophic Environment

## Abstract

The paper investigates a technique for order preference by similarity to ideal solution (TOPSIS) method to solve multi-attribute decision making problems with bipolar neutrosophic information. We define Hamming distance function and Euclidean distance function to determine the distance between bipolar neutrosophic numbers. In the decision making situation, the rating of performance values of the alternatives with respect to the attributes are provided by the decision maker in terms of bipolar neutrosophic numbers. The weights of the attributes are determined using maximizing deviation method. We define bipolar neutrosophic relative positive ideal solution (BNRPIS) and bipolar neutrosophic relative negative ideal solution (BNRNIS). Then, the ranking order of the alternatives is obtained by TOPSIS method and most desirable alternative is selected. Finally, a numerical example for car selection is solved to demonstrate the applicability and effectiveness of the proposed approach and comparison with other existing method is also provided.

## 1. Introduction

Zadeh [1] introduced the concept of fuzzy set to deal with problems with imprecise information in 1965. However, Zadeh [1] considers one single value to express the grade of membership of the fuzzy set defined in a universe. But, it is not always possible to represent the grade of membership value by a single point. In order to overcome the difficulty, Turksen [2] incorporated interval valued fuzzy sets. In 1986, Atanassov [3] extended the concept of fuzzy sets [1] and defined intuitionistic fuzzy sets which are characterized by grade of membership and non-membership functions. Later, Lee [4, 5] introduced the notion of bipolar fuzzy sets by extending the concept of fuzzy sets where the degree of membership is expanded from [0, 1] to [-1, 1]. In a bipolar fuzzy set, if the degree of membership is zero then we say the element is unrelated to the corresponding property, the membership degree (0, 1] of an element specifies that the element somewhat satisfies the property, and the membership degree $[-1, 0)$ of an element implies that the element somewhat satisfies the implicit counter-property [6]. Zhou and Li [7] incorporated the notion of bipolar fuzzy semirings and investigated relative properties using positive t- cut, negative s- cut and equivalence relation. Smarandache [8, 9, 10, 11] incorporated indeterminacy membership function as independent component and defined neutrosophic set on three components truth, indeterminacy and falsehood. However, from practical point of view, Wang et al. [12] defined single valued neutrosophic sets (SVNSs) where degree of truth membership, indeterminacy membership and falsity membership $\in [0, 1]$. Deli et al. [13] introduced the notion of bipolar neutrosophic sets (BNSs) which is a generalization of the fuzzy sets, bipolar fuzzy sets, intuitionistic fuzzy sets, neutrosophic sets. Pramanik and Mondal defined rough bipolar neutrosophic set [14].

Zhang and Wu [15] presented a TOPSIS [16] method for solving single valued neutrosophic multi-criteria decision making with incomplete weight information. Chi and Liu [17] proposed an extended TOPSIS method for MADM problems where the attribute weights are unknown and the attribute values are expressed in terms of interval neutrosophic numbers. Biswas et al. [18] developed a new TOPSIS based approach for solving multi-attribute group decision making problem with simplified neutrosophic information. Broumi et al. [19] extended TOPSIS method for multiple attribute decision making based on interval neutrosophic uncertain linguistic variables. In neutrosophic hybrid environment, Pramanik et al. [20] extended TOPSIS method for singled valued soft expert set based multi-attribute decision making problems. Dey et al. [21] presented TOPSIS method for generalized neutrosophic soft multi-attribute group decision making. Mondal et al. [22] presented TOPSIS in rough neutrosophic environment and provided illustrative example.

Deli et al. [13] investigated a bipolar neutrosophic multi-criteria decision making approach based on bipolar neutrosophic weighted average and geometric operators and the score, certainty and accuracy functions. Uluçay et al. [23] studied similarity measures of bipolar neutrosophic sets and their application to multiple criteria decision making. Literature review suggests that TOPSIS method in bipolar neutrosophic environment is yet to appear. Therefore this issue needs to be addressed.

In this paper, we define Hamming distances and Euclidean distances between two BNSs and develop a new TOPSIS based method for solving MADM problems under bipolar neutrosophic assessments.

The content of the paper is organized as follows. Section 2 presents some basic definitions concerning neutrosophic sets, SVNSs, BNSs which are helpful for the construction of the paper. Hamming and Euclidean distances between two bipolar neutrosophic numbers (BNNs) are also defined in the Section 2. Section 3 is devoted to present TOPSIS method for MADM problems under bipolar neutrosophic environment. A car selection problem is solved in Section 4 to illustrate the applicability of the proposed method. Sectin 5 presents conclusion.

## 2. Preliminaries

In this Section, we provide basic definitions regarding neutrosophic sets, SVNSs, BNSs.

**2.1 Neutrosophic Sets** [8, 9, 10, 11]

Consider $U$ be a space of objects with a generic element of $U$ denoted by $x$. Then, a neutrosophic set $N$ on $U$ is defined as follows:

$$N = \{x, \langle \mathrm{T}_N(x), \mathrm{I}_N(x), \mathrm{F}_N(x) \rangle \mid x \in U\}$$

where, $\mathrm{T}_N(x)$ , $\mathrm{I}_N(x)$ , $\mathrm{F}_N(x)$ : $U \rightarrow ]^-0, 1^+[$ represent respectively the degrees of truth-membership, indeterminacy-membership, and falsity-membership of a point $x \in U$ to the set $N$ with the condition $^-0 \leq \mathrm{T}_N(x) + \mathrm{I}_N(x) + \mathrm{F}_N(x) \leq 3^+$.

**2.2 Single valued neutrosophic Sets** [12]

Let $U$ be a universal space of points with a generic element of $X$ denoted by $x$, then a SVNS $S$ is presented as follows:

$$S = \{x, \langle \mathrm{T}_S(x), \mathrm{I}_S(x), \mathrm{F}_S(x) \rangle \mid x \in U\}$$

where, $\mathrm{T}_S(x)$, $\mathrm{I}_S(x)$, $\mathrm{F}_S(x)$: $U \rightarrow [0, 1]$ and $0 \leq \mathrm{T}_S(x) + \mathrm{I}_S(x) + \mathrm{F}_S(x) \leq 3$ for each point $x \in U$.

**2.3 Bipolar Neutrosophic Set** [13]

**Definition 1.** Let $U$ be a universal space of points, then a BNS $B$ in $U$ is defined as follows

$$B = \{x, \langle \mathrm{T}_B^+(x), \mathrm{I}_B^+(x), \mathrm{F}_B^+(x), \mathrm{T}_B^-(x), \mathrm{I}_B^-(x), \mathrm{F}_B^-(x) \rangle \mid x \in U\},$$

where $\mathrm{T}_B^+(x)$, $\mathrm{I}_B^+(x)$, $\mathrm{F}_B^+(x)$: $U \rightarrow [0, 1]$ and $\mathrm{T}_B^-(x)$, $\mathrm{I}_B^-(x)$, $\mathrm{F}_B^-(x)$: $U \rightarrow [-1, 0]$.

The positive membership degrees $\mathrm{T}_B^+(x)$, $\mathrm{I}_B^+(x)$, and $\mathrm{F}_B^+(x)$ represent the truth membership, indeterminate membership, and false membership of an element $x \in U$ corresponding to a bipolar neutrosophic set $B$ and the negative membership degrees $\mathrm{T}_B^-(x)$, $\mathrm{I}_B^-(x)$, and $\mathrm{F}_B^-(x)$ represent the truth membership, indeterminate membership, and false membership of an element $x \in U$ to some implicit counter property corresponding to a bipolar neutrosophic set $B$. For convenience, a bipolar neutrosophic number is represented by $\tilde{b} = <\mathrm{T}_B^+, \mathrm{I}_B^+, \mathrm{F}_B^+, \mathrm{T}_B^-, \mathrm{I}_B^-, \mathrm{F}_B^->$.

Example: Consider $U = \{u_1, u_2, u_3, u_4\}$. Then

$B$ = {< $u_1$, 0.6, 0.2, 0.1, -0.7, -0.1, -0.04>; < $u_2$, 0.4, 0.3, 0.1, -0.5, -0.09, -0.4>; < $u_3$, 0.8, 0.5, 0.4, -0.3, -0.01, -0.5>; < $u_4$, 0.3, 0.6, 0.7, -0.2, -0.3, -0.7>]

is a bipolar neutrosophic subset of $U$.

**Definition 2.** Let, $B_1 = \{x, \langle \mathrm{T}_{B_1}^+(x), \mathrm{I}_{B_1}^+(x), \mathrm{F}_{B_1}^+(x), \mathrm{T}_{B_1}^-(x), \mathrm{I}_{B_1}^-(x), \mathrm{F}_{B_1}^-(x) \rangle \mid x \in U\}$ and $B_2 = \{x, \langle \mathrm{T}_{B_2}^+(x), \mathrm{I}_{B_2}^+(x), \mathrm{F}_{B_2}^+(x), \mathrm{T}_{B_2}^-(x), \mathrm{I}_{B_2}^-(x), \mathrm{F}_{B_2}^-(x) \rangle \mid x \in U\}$ be two BNSs. Then $B_1 \subseteq B_2$ if and only if

$\mathrm{T}_{B_1}^+(x) \leq \mathrm{T}_{B_2}^+(x)$, $\mathrm{I}_{B_1}^+(x) \leq \mathrm{I}_{B_2}^+(x)$, $\mathrm{F}_{B_1}^+(x) \geq \mathrm{F}_{B_2}^+(x)$; $\mathrm{T}_{B_1}^-(x) \geq \mathrm{T}_{B_2}^-(x)$, $\mathrm{I}_{B_1}^-(x) \geq \mathrm{I}_{B_2}^-(x)$, $\mathrm{F}_{B_1}^-(x) \leq \mathrm{F}_{B_2}^-(x)$ for all $x \in U$.

**Definition 3.** Consider, $B_1 = \{x, \langle T^+_{B_1}(x), I^+_{B_1}(x), F^+_{B_1}(x), T^-_{B_1}(x), I^-_{B_1}(x), F^-_{B_1}(x) \rangle \mid x \in U\}$ and $B_2 = \{x, \langle T^+_{B_2}(x), I^+_{B_2}(x), F^+_{B_2}(x), T^-_{B_2}(x), I^-_{B_2}(x), F^-_{B_2}(x) \rangle \mid x \in U\}$ be two BNSs. Then $B_1 = B_2$ if and only if

$T^+_{B_1}(x) = T^+_{B_2}(x)$, $I^+_{B_1}(x) = I^+_{B_2}(x)$, $F^+_{B_1}(x) = F^+_{B_2}(x)$; $T^-_{B_1}(x) = T^-_{B_2}(x)$, $I^-_{B_1}(x) = I^-_{B_2}(x)$, $F^-_{B_1}(x) = F^-_{B_2}(x)$ for all $x \in U$.

**Definition 4.** Consider, $B = \{x, \langle T^+_B(x), I^+_B(x), F^+_B(x), T^-_B(x), I^-_B(x), F^-_B(x) \rangle \mid x \in U\}$ be a BNS. The complement of $B$ is denoted by $B^c$ and is defined by

$T^+_{B^c}(x) = \{1^+\} - T^+_B(x)$, $I^+_{B^c}(x) = \{1^+\} - I^+_B(x)$, $F^+_{B^c}(x) = \{1^+\} - F^+_B(x)$;

$T^-_{B^c}(x) = \{1^-\} - T^-_B(x)$, $I^-_{B^c}(x) = \{1^-\} - I^-_B(x)$, $F^-_{B^c}(x) = \{1^-\} - F^-_B(x)$ for all $x \in U$.

**Definition 5.** Consider, $B_1 = \{x, \langle T^+_{B_1}(x), I^+_{B_1}(x), F^+_{B_1}(x), T^-_{B_1}(x), I^-_{B_1}(x), F^-_{B_1}(x) \rangle \mid x \in U\}$ and $B_2 = \{x, \langle T^+_{B_2}(x), I^+_{B_2}(x), F^+_{B_2}(x), T^-_{B_2}(x), I^-_{B_2}(x), F^-_{B_2}(x) \rangle \mid x \in U\}$ be two BNSs. Then their union $B_1 \cup B_2$ is defined as follows:

$$B_1 \cup B_2 = \{\text{Max}\,(T^+_{B_1}(x), T^+_{B_2}(x)), \frac{I^+_{B_1}(x) + I^+_{B_2}(x)}{2}, \text{Min}\,(F^+_{B_1}(x), F^+_{B_2}(x)), \text{Min}\,(T^-_{B_1}(x), T^-_{B_2}(x)), \frac{I^-_{B_1}(x) + I^-_{B_2}(x)}{2}, \text{Max}\,(F^-_{B_1}(x), F^-_{B_2}(x))\} \text{ for all } x \in U.$$

**Definition 6.** Consider, $B_1 = \{x, \langle T^+_{B_1}(x), I^+_{B_1}(x), F^+_{B_1}(x), T^-_{B_1}(x), I^-_{B_1}(x), F^-_{B_1}(x) \rangle \mid x \in U\}$ and $B_2 = \{x, \langle T^+_{B_2}(x), I^+_{B_2}(x), F^+_{B_2}(x), T^-_{B_2}(x), I^-_{B_2}(x), F^-_{B_2}(x) \rangle \mid x \in U\}$ be two BNSs. Then their intersection $B_1 \cap B_2$ is defined as follows:

$$B_1 \cap B_2 = \{\text{Min}\,(T^+_{B_1}(x), T^+_{B_2}(x)), \frac{I^+_{B_1}(x) + I^+_{B_2}(x)}{2}, \text{Max}\,(F^+_{B_1}(x), F^+_{B_2}(x)), \text{Max}\,(T^-_{B_1}(x), T^-_{B_2}(x)), \frac{I^-_{B_1}(x) + I^-_{B_2}(x)}{2}, \text{Min}\,(F^-_{B_1}(x), F^-_{B_2}(x))\} \text{ for all } x \in U.$$

**Definition 7.** Suppose $\tilde{b}_1 = \langle T^+_{B_1}, I^+_{B_1}, F^+_{B_1}, T^-_{B_1}, I^-_{B_1}, F^-_{B_1} \rangle$ and $\tilde{b}_2 = \langle T^+_{B_2}, I^+_{B_2}, F^+_{B_2}, T^-_{B_2}, I^-_{B_2}, F^-_{B_2} \rangle$ are two BNNs, then

i. $\alpha . \tilde{b}_1 = \langle 1 - (1 - T^+_{B_1})^{\alpha}, (I^+_{B_1})^{\alpha}, (F^+_{B_1})^{\alpha}, -(-T^-_{B_1})^{\alpha}, -(-I^-_{B_1})^{\alpha}, -(1 - (1 - (-F^-_{B_1}))^{\alpha}) \rangle$;

ii. $(\tilde{b}_1)^{\alpha} = \langle (T^+_{B_1})^{\alpha}, 1 - (1 - I^+_{B_1})^{\alpha}, 1 - (1 - F^+_{B_1})^{\alpha}, -(1 - (1 - (-T^-_{B_1}))^{\alpha}), -(-I^-_{B_1})^{\alpha}, -(-F^-_{B_1}))^{\alpha} \rangle$;

iii. $\tilde{b}_1 + \tilde{b}_2 = \langle T^+_{B_1} + T^+_{B_2} - T^+_{B_1} . T^+_{B_2}, I^+_{B_1} . I^+_{B_2}, F^+_{B_1} . F^+_{B_2}, -T^-_{B_1} . T^-_{B_2}, -(-I^-_{B_1} - I^-_{B_2} - I^-_{B_1} . I^-_{B_2}), -(-F^-_{B_1} - F^-_{B_2} - F^-_{B_1} . F^-_{B_2}) \rangle$;

iv. $\widetilde{b}_1 . \widetilde{b}_2 = < T^+_{B_1} . T^+_{B_2}, I^+_{B_1} + I^+_{B_2} - I^+_{B_1} . I^+_{B_2}, F^+_{B_1} + F^+_{B_2} - F^+_{B_1} . F^+_{B_2}, (- T^-_{B_1} - T^-_{B_2} - T^-_{B_1} . T^-_{B_2}), - I^-_{B_1} . I^-_{B_2}, - F^-_{B_1} . F^-_{B_2} >$,

where $\alpha > 0$.

### 2.4. The distance between two BNNs

In this sub-section, we propose the distance between two BNNs.

Consider $B_1 = \sum_{i=1}^{m} (x_i, < T^+_{B_1}(x_i), I^+_{B_1}(x_i), F^+_{B_1}(x_i), T^-_{B_1}(x_i), I^-_{B_1}(x_i), F^-_{B_1}(x_i) >)$, $B_2 = \sum_{i=1}^{m} (x_i, < T^+_{B_2}(x_i), I^+_{B_2}(x_i), F^+_{B_2}(x_i), T^-_{B_2}(x_i), I^-_{B_2}(x_i), F^-_{B_2}(x_i) >)$ be two BNNs then,

(1). The Hamming distance between two BNNs is defined as follows:

$$D_H(B_1, B_2) = \sum_{i=1}^{m} \{|(T^+_{B_1}(x_i) - T^+_{B_2}(x_i))| + |(I^+_{B_1}(x_i) - I^+_{B_2}(x_i))| + |(F^+_{B_1}(x_i) - F^+_{B_2}(x_i))| + |(T^-_{B_1}(x_i) - T^-_{B_2}(x_i))| + |(I^-_{B_1}(x_i) - I^-_{B_2}(x_i))| + |(F^-_{B_1}(x_i) - F^-_{B_2}(x_i))|\} \quad (1)$$

(2). The normalized Hamming distance between two BNNs is defined as follows:

$$^N D_H(B_1, B_2) = \frac{1}{6m} \sum_{i=1}^{m} \{|(T^+_{B_i}(x_i) - T^+_{B_2}(x_i))| + |(I^+_{B_1}(x_i) - I^+_{B_2}(x_i))| + |(F^+_{B_1}(x_i) - F^+_{B_2}(x_i))| + |(T^-_{B_1}(x_i) - T^-_{B_2}(x_i))| + |(I^-_{B_1}(x_i) - I^-_{B_2}(x_i))| + |(F^-_{B_1}(x_i) - F^-_{B_2}(x_i))|\} \quad (2)$$

(3). The Euclidean distance between two BNNs is defined as follows:

$$E_H(B_1, B_2) = \sqrt{\sum_{i=1}^{m} \left\{ \begin{array}{l} (T^+_{B_1}(x_i) - T^+_{B_2}(x_i))^2 + (I^+_{B_1}(x_i) - I^+_{B_2}(x_i))^2 + (F^+_{B_1}(x_i) - F^+_{B_2}(x_i))^2 + \\ (T^-_{B_1}(x_i) - T^-_{B_2}(x_i))^2 + (I^-_{B_1}(x_i) - I^-_{B_2}(x_i))^2 + (F^-_{B_1}(x_i) - F^-_{B_2}(x_i))^2 \end{array} \right\}} \quad (3)$$

(4). The normalized Euclidean distance between two BNNs is defined as follows:

$$^N E_H(B_1, B_2) = \sqrt{\frac{1}{6m} \sum_{i=1}^{m} \left\{ \begin{array}{l} (T^+_{B_1}(x_i) - T^+_{B_2}(x_i))^2 + (I^+_{B_1}(x_i) - I^+_{B_2}(x_i))^2 + (F^+_{B_1}(x_i) - F^+_{B_2}(x_i))^2 + \\ (T^-_{B_1}(x_i) - T^-_{B_2}(x_i))^2 + (I^-_{B_1}(x_i) - I^-_{B_2}(x_i))^2 + (F^-_{B_1}(x_i) - F^-_{B_2}(x_i))^2 \end{array} \right\}} \quad (4)$$

with the following properties:

(1). $0 \le D_H(B_1, B_2) \le 6m$

(2). $0 \le {}^N D_H(B_1, B_2) \le 1$

(3). $0 \le E_H(B_1, B_2) \le \sqrt{6m}$

(4). $0 \le {}^N E_H(B_1, B_2) \le 1$.

## 3. TOPSIS method for MADM with bipolar neutrosophic information

In this Section, we present an approach based on TOPSIS method to deal with MADM problems under bipolar neutrosophic environment.

Let $A$ = {$A_1, A_2, \ldots, A_m$}, ($m \geq 2$) be a discrete set of $m$ feasible alternatives, $C$ = {$C_1, C_2, \ldots, C_n$}, ($n \geq 2$) be a set of attributes under consideration and $w = (w_1, w_2, \ldots, w_n)^T$ be the unknown weight vector of the attributes with $0 \leq w_j \leq 1$ and $\sum_{j=1}^{n} w_j = 1$. The rating of performance value of alternative $A_i$, ($i = 1, 2, \ldots, m$) with respect to the predefined attribute $C_j$, ($j = 1, 2, \ldots, n$) is presented by the decision maker (DM) and they can be expressed by BNNs. Therefore, the proposed approach is presented using the following steps:

**Step 1. Construction of decision matrix with BNNs**

The rating of performance value of alternative $A_i$ ($i = 1, 2, \ldots, m$) with respect to the attribute $C_j$, ($j = 1, 2, \ldots, n$) is expressed by BNNs and they can be presented in the decision matrix as follows:

$$\left\langle \tilde{r}_{ij} \right\rangle_{m \times n} = \begin{bmatrix} r_{11} & r_{12} & \ldots & r_{1n} \\ r_{21} & r_{22} & \ldots & r_{2n} \\ . & . & \ldots & . \\ . & . & \ldots & . \\ r_{m1} & r_{m2} & \ldots & r_{mn} \end{bmatrix}$$

Here, we have $r_{ij} = (T_{ij}^{+}, I_{ij}^{+}, F_{ij}^{+}, T_{ij}^{-}, I_{ij}^{-}, F_{ij}^{-})$ with $T_{ij}^{+}, I_{ij}^{+}, F_{ij}^{+}, -T_{ij}^{-}, -I_{ij}^{-}, -F_{ij}^{-} \in [0, 1]$ and $0 \leq T_{ij}^{+} + I_{ij}^{+} + F_{ij}^{+} - T_{ij}^{-} - I_{ij}^{-} - F_{ij}^{-} \leq 6$ for $i = 1, 2, \ldots, m$; $j = 1, 2, \ldots, n$.

**Step 2. Determination of weights of the attributes**

We assume that the weights of the attributes are not equal and they are fully unknown to the DM. Therefore, in this paper, maximizing deviation method [24] is used to find the unknown weights. The main idea of maximizing deviation method can be expressed as follows. If the attribute values $r_{ij}$ ($j = 1, 2, \ldots, n$) in the attribute $C_j$ have small differences between the alternatives, then $C_j$ has a small significance in ranking of all alternatives and a small weight is assigned for the attribute. If the attribute values $r_{ij}$ ($j = 1, 2, \ldots, n$) in the attribute $C_j$ are same, then $C_j$ has no effect in the ranking results and zero is assigned to the weight of the attribute. However, if the attribute values $r_{ij}$ ($j = 1, 2, \ldots, n$) over the attribute $C_j$ have big differences, then $C_j$ will play a key role in ranking of all alternatives and we will allocate a big weight for the attribute. The deviation values of alternative $A_i$ ($i = 1, 2, \ldots, m$) to all other alternatives under the attribute $C_j$ ($j = 1, 2, \ldots, n$) can be defined as $Z_{ij}(w_j) = \sum_{k=1}^{m} z(r_{ij}, r_{kj}) w_j$, then $Z_j(w_j) = \sum_{i=1}^{m} Z_{ij} w_j = \sum_{i=1}^{m}\sum_{k=1}^{m} z(r_{ij}, r_{kj}) w_j$ presents the total deviation values of all alternatives to the other alternatives for the attribute $C_j$ ($j = 1, 2, \ldots, n$). Now $Z(w_j) = \sum_{j=1}^{n} Z_j(w_j) = \sum_{j=1}^{n} \sum_{i=1}^{m}\sum_{k=1}^{m} z(r_{ij}, r_{kj}) w_j$ presents the total deviation of all attributes to the other alternatives with respect to all alternatives. Now we construct the non-linear optimizing model based on above analysis to obtain unknown attribute weight $w_j$ as follows:

$$\text{Max } Z(w_j) = \sum_{j=1}^{n} \sum_{i=1}^{m}\sum_{k=1}^{m} z(r_{ij}, r_{kj}) w_j \tag{5}$$

Subject to $\sum_{j=1}^{n} w_j^2 = 1$, $w_j \geq 0$, $j = 1, 2, \ldots, n$.

We now formulate the Lagrange multiplier function, and obtain

$$L(w_j,\rho) = \sum_{j=1}^{n} \sum_{i=1}^{m}\sum_{k=1}^{m} z(r_{ij}, r_{kj}) w_j + \rho \left(\sum_{j=1}^{q} w_j^2 - 1\right)$$

where $\rho$ is the Lagrange multiplier.

Then, we calculate the partial derivatives of L with respect to $w_j$ and $\rho$ respectively as follows:

$$\frac{\partial L(w_j,\rho)}{\partial w_j} = \sum_{i=1}^{m}\sum_{k=1}^{m} z(r_{ij}, r_{kj}) w_j + \rho \left(\sum_{j=1}^{n} w_j^2 - 1\right) = 0,$$

$$\frac{\partial L(w_j,\rho)}{\partial \rho} = \sum_{j=1}^{n} w_j^2 - 1 = 0.$$

Therefore, the weight of the attribute $C_j$ is obtained as

$$w_j = \frac{\sum_{i=1}^{m}\sum_{k=1}^{m} z(r_{ij}, r_{kj})}{\sqrt{\sum_{j=1}^{n}\left(\sum_{i=1}^{m}\sum_{k=1}^{m} z(r_{ij}, r_{kj})\right)^2}} \tag{6}$$

and the normalized weight of the attribute $C_j$ is given by

$$w_j^* = \frac{\sum_{i=1}^{m}\sum_{k=1}^{m} z(r_{ij}, r_{kj})}{\sum_{j=1}^{n}\left(\sum_{i=1}^{m}\sum_{k=1}^{m} z(r_{ij}, r_{kj})\right)}. \tag{7}$$

**Step 3. Construction of weighted decision matrix**

We find aggregated weighted decision matrix by multiplying weights [25] of the attributes and the aggregated decision matrix $\langle r_{ij}^{w_j} \rangle_{m\times n}$ is constructed as follows:

$$\langle r_{ij} \rangle_{m\times n} \otimes w_j = \langle r_{ij}^{w_j} \rangle_{m\times n} = \begin{bmatrix} r_{11}^{w_1} & r_{12}^{w_2} & \dots & r_{1n}^{w_n} \\ r_{21}^{w_1} & r_{22}^{w_2} & \dots & r_{2n}^{w_n} \\ . & . & \dots & . \\ . & . & \dots & . \\ r_{m1}^{w_1} & r_{m2}^{w_2} & \dots & r_{mn}^{w_n} \end{bmatrix}$$

$r_{ij}^{w_j} = (T_{ij}^{w_j+}, I_{ij}^{w_j+}, F_{ij}^{w_j+}, T_{ij}^{w_j-}, I_{ij}^{w_j-}, F_{ij}^{w_j-})$ with $T_{ij}^{w+}, I_{ij}^{w+}, F_{ij}^{w+}, T_{ij}^{w-}, I_{ij}^{w-}, F_{ij}^{w-} \in [0, 1]$ and $0 \le T_{ij}^{w+} + I_{ij}^{w+} + F_{ij}^{w+} - T_{ij}^{w-} - I_{ij}^{w-} - F_{ij}^{w-} \le 6$ for i = 1, 2, …, m; j = 1, 2, …, n.

**Step 4. Identify the bipolar neutrosophic relative positive ideal solution (BNRPIS) and bipolar neutrosophic relative negative ideal solution (BNRNIS)**

In real life decision making, we confront two types of attributes namely, benefit type attributes ($\beta_1$) and cost type attributes ($\beta_2$). In bipolar neutrosophic environment, assume that $Q_{BNRPIS}^{w+}$ and $Q^{w-}_{BNRNIS}$ be the bipolar neutrosophic relative positive ideal solution (BNRPIS) and bipolar neutrosophic relative negative ideal solution (BNRNIS). Then, $Q^{w+}_{BNRPIS}$ and $Q^{w-}_{BNRNIS}$ are defined as follows:

$$Q^{w+}_{BNRPIS} = \left( \left\langle {}^{+}T_1^{w_1+}, {}^{+}I_1^{w_1+}, {}^{+}F_1^{w_1+}, {}^{+}T_1^{w_1-}, {}^{+}I_1^{w_1-}, {}^{+}F_1^{w_1-}, \right\rangle, \left\langle {}^{+}T_2^{w_2+}, {}^{+}I_2^{w_2+}, {}^{+}F_2^{w_2+}, {}^{+}T_2^{w_2-}, {}^{+}I_2^{w_2-}, {}^{+}F_2^{w_2-} \right\rangle, \ldots, \left\langle {}^{+}T_n^{w_n+}, {}^{+}I_n^{w_n+}, {}^{+}F_n^{w_n+}, {}^{+}T_n^{w_n-}, {}^{+}I_n^{w_n-}, {}^{+}F_n^{w_n-} \right\rangle \right) \quad (8)$$

$$Q^{w-}_{BNRNIS} = \left( \left\langle {}^{-}T_1^{w_1+}, {}^{-}I_1^{w_1+}, {}^{-}F_1^{w_1+}, {}^{-}T_1^{w_1-}, {}^{-}I_1^{w_1-}, {}^{-}F_1^{w_1-}, \right\rangle, \left\langle {}^{-}T_2^{w_2+}, {}^{-}I_2^{w_2+}, {}^{-}F_2^{w_2+}, {}^{-}T_2^{w_2-}, {}^{-}I_2^{w_2-}, {}^{-}F_2^{w_2-} \right\rangle, \ldots, \left\langle {}^{-}T_n^{w_n+}, {}^{-}I_n^{w_n+}, {}^{-}F_n^{w_n+}, {}^{-}T_n^{w_n-}, {}^{-}I_n^{w_n-}, {}^{-}F_n^{w_n-} \right\rangle \right) \quad (9)$$

where

$\left\langle {}^{+}T_j^{w_j+}, {}^{+}I_j^{w_j+}, {}^{+}F_j^{w_j+}, {}^{+}T_j^{w_j-}, {}^{+}I_j^{w_j-}, {}^{+}F_j^{w_j-} \right\rangle$ = <[{ $\underset{i}{Max}(T_{ij}^{w_j+}) \mid j \in \beta_1$ }; { $\underset{i}{Min}(T_{ij}^{w_j+}) \mid j \in \beta_2$ }], [{ $\underset{i}{Min}(I_{ij}^{w_j+}) \mid j \in \beta_1$ }; { $\underset{i}{Max}(I_{ij}^{w_j+}) \mid j \in \beta_2$ }], [{ $\underset{i}{Min}(F_{ij}^{w_j+}) \mid j \in \beta_1$ }; { $\underset{i}{Max}(F_{ij}^{w_j+}) \mid j \in \beta_2$ }], [{ $\underset{i}{Min}(T_{ij}^{w_j-}) \mid j \in \beta_1$ }; { $\underset{i}{Max}(T_{ij}^{w_j-}) \mid j \in \beta_2$ }], [{ $\underset{i}{Max}(I_{ij}^{w_j-}) \mid j \in \beta_1$ }; { $\underset{i}{Min}(I_{ij}^{w_j-}) \mid j \in \beta_2$ }],[{ $\underset{i}{Max}(F_{ij}^{w_j-}) \mid j \in \beta_1$ }; { $\underset{i}{Min}(F_{ij}^{w_j-}) \mid j \in \beta_2$ }]>, j = 1, 2, …, n;

$\left\langle {}^{-}T_j^{w_j+}, {}^{-}I_j^{w_j+}, {}^{-}F_j^{w_j+}, {}^{-}T_j^{w_j-}, {}^{-}I_j^{w_j-}, {}^{-}F_j^{w_j-} \right\rangle$ = < [{ $\underset{i}{Min}(T_{ij}^{w_j+}) \mid j \in \beta_1$ }; { $\underset{i}{Max}(T_{ij}^{w_j+}) \mid j \in \beta_2$ }], [{ $\underset{i}{Max}(I_{ij}^{w_j+}) \mid j \in \beta_1$ }; { $\underset{i}{Min}(I_{ij}^{w_j+}) \mid j \in \beta_2$ }], [{ $\underset{i}{Max}(F_{ij}^{w_j+}) \mid j \in \beta_1$ }; { $\underset{i}{Min}(F_{ij}^{w_j+}) \mid j \in \beta_2$ }], [{ $\underset{i}{Max}(T_{ij}^{w_j-}) \mid j \in \beta_1$ }; { $\underset{i}{Min}(T_{ij}^{w_j-}) \mid j \in \beta_2$ }], [{ $\underset{i}{Min}(I_{ij}^{w_j-}) \mid j \in \beta_1$ }; { $\underset{i}{Max}(I_{ij}^{w_j-}) \mid j \in \beta_2$ }], [{ $\underset{i}{Min}(F_{ij}^{w_j-}) \mid j \in \beta_1$ }; { $\underset{i}{Max}(F_{ij}^{w_j-}) \mid j \in \beta_2$ }] >, j = 1, 2, …, n.

**Step 5. Calculation of distance of each alternative from BNRPIS and BNRNIS**

The normalized Euclidean distance of each alternative $\left\langle T_{ij}^{w_j+}, I_{ij}^{w_j+}, F_{ij}^{w_j+}, T_{ij}^{w_j-}, I_{ij}^{w_j-}, F_{ij}^{w_j-} \right\rangle$ from the BNRPIS $\left\langle {}^{+}T_j^{w_j+}, {}^{+}I_j^{w_j+}, {}^{+}F_j^{w_j+}, {}^{+}T_j^{w_j-}, {}^{+}I_j^{w_j-}, {}^{+}F_j^{w_j-} \right\rangle$ for i = 1, 2, …, m; j = 1, 2, …., n can be defined as follows:

$$Euc_N^{i+} = \sqrt{\frac{1}{6n} \sum_{j=1}^{n} \left\{ \begin{matrix} (T_{ij}^{w_j+} - {}^{+}T_{ij}^{w_j+})^2 + (I_{ij}^{w_j+} - {}^{+}I_{ij}^{w_j+})^2 + (F_{ij}^{w_j+} - {}^{+}F_{ij}^{w_j+})^2 + \\ T_{ij}^{w_j-} - {}^{+}T_{ij}^{w_j-})^2 + (I_{ij}^{w_j-} - {}^{+}I_{ij}^{w_j-})^2 + (F_{ij}^{w_j-} - {}^{+}F_{ij}^{w_j-})^2 \end{matrix} \right\}} \quad (10)$$

Similarly, normalized Euclidean distance of each alternative $\left\langle T_{ij}^{w_j+}, I_{ij}^{w_j+}, F_{ij}^{w_j+}, T_{ij}^{w_j-}, I_{ij}^{w_j-}, F_{ij}^{w_j-} \right\rangle$ from the BNRNIS $\left\langle {}^{-}T_j^{w_j+}, {}^{-}I_j^{w_j+}, {}^{-}F_j^{w_j+}, {}^{-}T_j^{w_j-}, {}^{-}I_j^{w_j-}, {}^{-}F_j^{w_j-} \right\rangle$ for i = 1, 2, …, m; j = 1, 2, …., n can be written as follows:

$$Euc_N^{i-} = \sqrt{\frac{1}{6n} \sum_{j=1}^{n} \left\{ \begin{array}{l} (T_{ij}^{w_j+} - {}^-T_{ij}^{w_j+})^2 + (I_{ij}^{w_j+} - {}^-I_{ij}^{w_j+})^2 + (F_{ij}^{w_j+} - {}^-F_{ij}^{w_j+})^2 + \\ T_{ij}^{w_j-} - {}^-T_{ij}^{w_j-})^2 + (I_{ij}^{w_j-} - {}^-I_{ij}^{w_j-})^2 + (F_{ij}^{w_j-} - {}^-F_{ij}^{w_j-})^2 \end{array} \right\}} \tag{11}$$

**Step 6. Evaluate the relative closeness co-efficient**

The relative closeness co-efficient of each alternative $A_i$, (i = 1, 2, …, m) with respect to the BNRPIS $Q_{BNRPIS}^{w+}$ is defined as follows:

$$cc_i^* = \frac{Euc_N^{i-}}{Euc_N^{i+} + Euc_N^{i-}} \tag{12}$$

where, $0 \le cc_i^* \le 1$, i = 1, 2, …, m.

**Step 7. Rank the alternatives**

Rank the alternatives according to the descending order of the alternatives and select the best alternative with maximum value of $cc_i^*$.

## 4. A numerical example

We consider the problem [13] where a customer wants to buy a car. There are four types cars (alternatives) $A_i$, i = 1, 2, 3, 4 are available. The customer considers four attributes namely Fuel economy ($C_1$), Aerod ($C_2$), Comfort ($C_3$), Safety $C_4$ to assess the alternatives. Now we solve the problem with bipolar neutrosophic information based on TOPSIS method to select most desirable car for the customer. Then, the proposed TOPSIS approach for solving the problem is presented in the following steps:

**Step 1: Formulation of decision matrix**

We construct the decision matrix with bipolar neutrosophic information presented by the DM as given below (see Table 1).

Table 1. *The decision matrix provided by the DM*

| | $C_1$ | $C_2$ | $C_3$ | $C_4$ |
|---|---|---|---|---|
| $A_1$ | (0.5, 0.7, 0.2, -0.7, -0.3, -0.6) | (0.4, 0.4, 0.5, -0.7, -0.8, -0.4) | (0.7, 0.7, 0.5, -0.8, -0.7, -0.6) | (0.1, 0.5, 0.7, -0.5, -0.2, -0.8) |
| $A_2$ | (0.9, 0.7, 0.5, -0.7, -0.7, -0.1) | (0.7, 0.6, 0.8, -0.7, -0.5, -0.1) | (0.9, 0.4, 0.6, -0.1, -0.7, -0.5) | (0.5, 0.2, 0.7, -0.5, -0.1, -0.9) |
| $A_3$ | (0.3, 0.4, 0.2, -0.6, -0.3, -0.7) | (0.2, 0.2, 0.2, -0.4, -0.7, -0.4) | (0.9, 0.5, 0.5, -0.6, -0.5, -0.2) | (0.7, 0.5, 0.3, -0.4, -0.2, -0.2) |
| $A_4$ | (0.9, 0.7, 0.2, -0.8, -0.6, -0.1) | (0.3, 0.5, 0.2, -0.5, -0.5, -0.2) | (0.5, 0.4, 0.5, -0.1, -0.7, -0.2) | (0.4, 0.2, 0.8, -0.5, -0.5, -0.6) |

**Step 2. Calculation of the weights of the attributes**

We use normalized Hamming distance and obtain the weights of the attributes by maximizing deviation method as follows:

$w_1 = 0.2585$, $w_2 = 0.2552$, $w_3 = 0.2278$, $w_4 = 0.2585$, where $\sum_{j=1}^{4} w_j = 1$.

**Step 3. Construction of weighted decision matrix**

The weighted decision matrix is obtained by multiplying weights to decision matrix as given below (see Table 2)

Table 2. *The weighted decision matrix*

| | $C_1$ | $C_2$ | $C_3$ | $C_4$ |
|---|---|---|---|---|
| $A_1$ | (0.164, 0.912, 0.66, -0.912, -0.732, -0.211) | (0.122, 0.791, 0.838, -0.913, -0.945, -0.122) | (0.24, 0.922, 0.854, -0.95, -0.922, -0.208) | (0.027, 0.836, 0.912, -0.836, -0.66, -0.337) |
| $A_2$ | (0.488, 0.912, 0.836, -0.912, -0.912, -0.027) | (0.264, 0.874, 0.945, -0.913, -0.838, -0.026) | (0.408, 0.812, 0.89, -0.592, -0.922, -0.162) | (0.164, 0.66, 0.912, -0.836, -0.551, -0.444) |
| $A_3$ | (0.088, 0.789, 0.66, -0.876, -0.732, -0.267) | (0.055, 0.663, 0.663, -0.791, -0.913, -0.122) | (0.408, 0.854, 0.854, -0.89, -0.854, -0.055) | (0.267, 0.836, 0.088, -0.789, -0.66, -0.055) |
| $A_4$ | (0.448, 0.912, 0.66, -0.944, -0.876, -0.027) | (0.087, 0.838, 0.663, -0.838, -0.838, -0.055) | (0.146, 0.812, 0.854, -0.592, -0.922, -0.055) | (0.124, 0.66, 0.944, -0.836, -0.836, -0.208) |

**Step 4. Recognize the BNRPIS and BNRNIS**

The BNRPIS ($R^{w+}_{BPRPIS}$) and BNRNIS ($R^{w-}_{BPRNIS}$) are obtained from the weighted decision matrix as follows:

$R^{w+}_{BPRPIS}$= < (0.448, 0.789, 0.66, -0.944, -0.732, -0.027); (0.264, 0.663, 0.663, -0.913, -0.838, -0.026); (0.408, 0.812, 0.854, -0.89, -0.854, -0.055); (0.267, 0.66, 0.88, -0.836, -0.551, -0.055) >;

$R^{w-}_{BPRNIS}$= < (0.088, 0.912, 0.836, -0.876, -0.912, -0.267); (0.055, 0.878, 0.945, -0.791, -0.945, -0.122); (0.146, 0.922, 0.89, -0.592, -0.922, -0.208); (0.027, 0.836, 0.912, -0.789, -0.836, -0.444) >.

**Step 5. Distance measures of each alternative from the BNRPISs and BNRNISs**

The normalized Euclidean distances of each alternative from the BNRPISs are computed as follows:

$Euc_N^{1+} = 0.0479$, $Euc_N^{2+} = 0.0161$, $Euc_N^{3+} = 0.013$, $Euc_N^{4+} = 0.0469$.

Similarly, the normalized Euclidean distances of each alternative from the BNRNISs are computed as follows:

$Euc_N^{1-} = 0.0123$, $Euc_N^{2-} = 0.0247$, $Euc_N^{3-} = 0.0548$, $Euc_N^{4-} = 0.0192$.

**Step 6. Calculation of the relative closeness coefficient**

We determine the relative closeness co-efficient $cc_i^*$, (i = 1, 2, 3, 4) using Eq. (12).

$cc_1^* = 0.2043$, $cc_2^* = 0.6054$, $cc_3^* = 0.8082$, $cc_4^* = 0.2905$.

**Step 7.** Rank the alternatives

The ranking order of the cars is presented according to the relative closeness coefficient as given below.

$A_3 \succ A_2 \succ A_4 \succ A_1$

Consequently, $A_3$ is the most preferable alternative.

**Note 1:** Deli et al. [13] consider the weight vector of the attributes as $w = (\frac{1}{2}, \frac{1}{4}, \frac{1}{8}, \frac{1}{8})$ for car selection. However, if we take weight vector of the attributes as $w = (\frac{1}{2}, \frac{1}{4}, \frac{1}{8}, \frac{1}{8})$, then relative closeness co-efficient $cc_i^*$, (i = 1, 2, 3, 4) are computed as given below.

$cc_1^* = 0.3746$, $cc_2^* = 0.5761$, $cc_3^* = 0.4716$, $cc_4^* = 0.6944$.

Therefore, the ranking order of the cars can be represented as follows:

$A_4 \succ A_2 \succ A_3 \succ A_1$

So, $A_4$ would be the most suitable alternative.

## 5. Conclusion

In this paper, we present a TOPSIS method for solving MADM problem with bipolar neutrosophic information. We define Hamming distance function and Euclidean distance function to determine the distance between BNNs. In the decision making situation, the rating of performance values of the alternatives with respect to the attributes are provided by the DM in terms of BNNs. The weights of the attributes are obtained by maximizing deviation method and we construct the weighted decision matrix. We also define BNRPIS and BNRNIS. Euclidean distance measure is employed to compute the distances of each alternative from BNRPISs as well as BNRNISs. Relative closeness coefficients are calculated to rank the alternative and to obtain the best alternative. Finally, the proposed method is applied to solve a car selection problem to verify the applicability of the proposed method and comparison with other existing method is also provided.

## References

1. L.A. Zadeh. Fuzzy sets, Information and Control, 8 (1965), 338-353.
2. I.B.Turksen. Interval valued fuzzy sets based on normal forms, Fuzzy Sets and Systems, 20(2) (1986), 191-210.
3. K.T. Atanassov. Intuitionistic fuzzy sets, Fuzzy Sets and Systems, 20 (1986), 87-96.
4. K.M. Lee. Bipolar-valued fuzzy sets and their operations, In Proc. Int. Conf. on Intelligent Technologies, Bangkok, Thailand (2000), 307-312.
5. K.M. Lee. Bipolar fuzzy subalgebras and bipolar fuzzy ideals of BCK/BCI- algebras, Bull. Malays. Math. Sci. Soc., 32(3) (2009), 361-367.
6. J. Chen, S. Li, M. Ma, X. Wang. m-Polar fuzzy sets: an extension of bipolar fuzzy sets, The Scientific World Journal, 2014. http://dx.doi.org/10.1155/2014/416530.
7. M. Zhou, S. Li. Application of bipolar fuzzy sets in semirings, Journal of Mathematical Research with Applications, 34(1) (2014), 61-72.
8. F. Smarandache. A unifying field of logics. Neutrosophy: neutrosophic probability, set and logic, American Research Press, Rehoboth, 1998.
9. F. Smarandache. Linguistic paradoxes and tautologies. Libertas Mathematica, University of Texas at Arlington, IX (1999), 143-154.
10. F. Smarandache. Neutrosophic set – a generalization of intuitionistic fuzzy sets, International Journal of Pure and Applied Mathematics, 24(3) (2005), 287-297.
11. F. Smarandache. Neutrosophic set – a generalization of intuitionistic fuzzy set, Journal of Defence Resources Management, 1(1) (2010), 107-116.
12. H. Wang, F. Smarandache, Y. Zhang, and R. Sunderraman. Single valued neutrosophic sets, Multi-space and Multi-Structure, 4 (2010), 410-413.
13. Deli, M. Ali, F. Smarandache. Bipolar neutrosophic sets and their application based on multi-criteria decision making problems, Proceedings of the 2015 International Conference on Advanced Mechatronic Systems, Beiging, China, August, 20-24, 2015, 249-254.
14. S. Pramanik, K.Mondal. Rough bipolar neutrosophic set. Global Journal of Engineering Science and Research Management 3(6) (2016), 71-81.
15. Z. Zhang, C. Wu. A novel method for single-valued neutrosophic multi-criteria decision making with incomplete weight information, Neutrosophic Sets and Systems, 4 (2014), 35-49.
16. C.L. Hwang, K. Yoon. Multiple attribute decision making: methods and applications, Springer, New York, 1981.
17. P. Chi, P. Liu. An extended TOPSIS method for the multiple attribute decision making problems based on interval neutrosophic set, Neutrosophic Sets and Systems, 1 (2013), 63-70.
18. P. Biswas, S. Pramanik, B.C. Giri. TOPSIS method for multi-attribute group decision making under single-valued neutrosophic environment, Neural Computing and Applications, (2015), DOI: 10.1007/s00521-015-1891-2.
19. S. Broumi, J. Ye. F. Smnarandache. An extended TOPSIS method for multiple attribute decision making based on interval neutrosophic uncertain linguistic variables. Neutrosophic Sets and Systems 8 (2015), 22-31.
20. S. Pramanik, P. P. Dey, B. C. Giri. TOPSIS for singled valued soft expert set based multi-attribute decision making problems. Neutrosophic Sets and Systems 10 (2015), 88-95.
21. P.P. Dey, S. Pramanik, B.C. Giri. Generalized neutrosophic soft multi-attribute group decision making based on TOPSIS. Critical Review 11 (2015), 41-55.
22. K. Mondal, S. Pramanik, F. Smarandache. TOPSIS in rough neutrosophic environment. Neutrosophic Sets and Systems 13 (2016). In Press.

23. V. Uluçay, I. Deli, M. Şahin, Similarity measures of bipolar neutrosophic sets and their application to multiple criteria decision making. Neural Computing and Applications (2016). doi 10.1007/s00521-016-2479-1.
24. Y.M. Yang. Using the method of maximizing deviations to make decison for multi-indices, System Engineering and Electronics, 7 (1998) 24-31.
25. J.J. Peng, J.Q. Wang, J. Wang, H. Zhang, X. Chen. Simplified neutrosophic sets and their applications in multi-criteria group decision-making problems, International Journal of Systems Science, (2014), DOI:10.1080/00207721.2014.994050.

**SURAPATI PRAMANIK[1*], DURGA BANERJEE[2], B. C. GIRI[3]**

1* Department of Mathematics, Nandalal Ghosh B.T. College, Panpur, P.O.-Narayanpur, District –North 24 Parganas, , West Bengal, India-743126. E-mail: sura_pati@yahoo.co.in
2 Ranaghat Yusuf Institution, P. O. Ranaghat,Dist. Nadia, India-741201. E-mail: dbanerje3@gmail.com
3 Department of Mathematics, Jadavpur University, Jadavpur, West Bengal, India-700032.
E-mail: bcgiri.jumath@gmail.com

# TOPSIS Approach for Multi Attribute Group Decision Making in Refined Neutrosophic Environment

## Abstract

This paper presents TOPSIS approach for multi attribute decision making in refined neutrosophic environment. The weights of each decision makers are considered as a single valued neutrosophic numbers. The attribute weights for every decision maker are also considered as a neutrosophic numbers. Aggregation operator is used to combine all decision makers' opinion into a single opinion for rating between attributes and alternatives. Euclidean distances from positive ideal solution and negative ideal solution are calculated to construct relative closeness coefficients. Lastly, an illustrative example of tablet selection is provided to show the applicability of the proposed TOPSIS approach.

## 1. Introduction

Decision making in neutrosophic environment is a developing area of research. Florentin Smarandache [1] introduced neutrosophic set which is the generalization of fuzzy set (FS) introduced by L.A. Zadeh [2] and intuitionistic fuzzy set (IFS) proposed by K. T. Atanassov [3]. Florentin Smarandache and his colleagues [4] presented an instance of single valued neutrosophic set called single valued neutrosophic set (SVNS) and their set theoretic operations. FS only considers membership function to represent imprecise data. IFS is characterized by membership and non-membership degrees, which are independent but the sum of degrees of membership and non-membership is less than unity. Both FS and IFS are unable to deal with indeterminacy in real decision making problem. Indeterminacy plays an important role in decision making situation. For example, in an application form there are three options 'YES / NO/ N. A.' for gender M / F / Others. So, different kinds of uncertainty and vagueness with indeterminacy cannot be explained by the fuzzy concept or intuitionistic fuzzy concept. Florentin Smarandache [1] first focused on indeterminacy of the imprecise data and introduced the concept of neutrosophic set consisting of three membership functions namely truth, indeterminacy and falsity membership functions which are independent.

Hawang and Yoon [5] introduced a technique for order preference by similarity to ideal solution (TOPSIS). TOPSIS for multi criteria decision making (MCDM) problem in fuzzy environment has been proposed by Chen [6]. Boran et al. [7] applied TOPSIS approach to multi attribute group decision making (MAGDM) in intuitionistic fuzzy environment. Multicriteria decision - making method using the correlation coefficient under single valued neutrosophic environment has been proposed by Ye [8]. Ye [9] further established single valued neutrosophic cross entropy for MCDM. Biswas et al. [10] presented entropy based grey relational analysis method for multi-attribute decision - making under single valued neutrosophic assessments. Biswas et al. [11] proposed MCDM with unknown weight information. Pramanik et al. [12] developed hybrid vector similarity measures and their applications to multi-attribute decision making under neutrosophic environment. Zhang et al. [13] presented interval neutrosophic MCDM. Pramanik and Mondal [14] presented interval neutrosophic multi-attribute decision-making based on grey relational analysis. Ye [15] applied aggregation operator for MCDM problem for simplified neutrosophic sets. Some important approaches in neutrosophic decision making problems can be found in [16-32]. Biswas et al. [33] proposed TOPSIS method for MAGDM for under single valued neutrosophic environment. Chai and Liu [34] applied TOPSIS method for MCDM with interval neutrosophic set. Broumi et al. [35] presented extended TOPSIS method for multiple attribute decision making based on interval neutrosophic uncertain linguistic variables. In neutrosophic hybrid environment, Pramanik et al. [36] presented TOPSIS for singled valued soft expert set based multi-attribute decision making problems. Dey et al. [37] studied generalized neutrosophic soft multi-attribute group decision making based on TOPSIS. Dey et al. [38] proposed TOPSIS for solving multi-attribute decision making problems under bi-polar neutrosophic environment. Mondal et al. [39] presented TOPSIS in rough neutrosophic environment and provided an illustrative example.

Yager [40] introduced the concept of multiset in 1986. Sebastian and Ramakrishnan [41] developed the concept of multi fuzzy set and studied some of their properties. Shinoj and John [42] presented intuitionistic fuzzy multiset. Ye and Ye [43] presented Dice similarity measure between single valued neutrosophic multisets and its application in medical diagnosis Smarandache [44] proposed n- valued refined neutrosophic logic and its application. Broumi and Smarandache [45] defined neutrosophic refined similarity measure based on cosine function. Mondal and Pramanik [46] proposed neutrosophic refined similarity measure using tangent function and applied it to multi attribute decision making. Mondal and Pramanik [47] also defined neutrosophic refined similarity measure and its application based on cotangent function. Pramanik et al. [48] recently presented MCGDM in neutrosophic refined environment and its application in teacher selection. Nadaban and Dzitac [49] discussed the general view in neutrosophic TOPSIS and presented a very brief survey on the applications of neutrosophic sets in MCDM problems.

The present paper is devoted to extend TOPSIS approach for MAGDM in refined neutrosophic environment. An aggregation operator due to Jun Ye [15] is used in refined neutrosophic environment. The relative closeness coefficients for all attributes are calculated and the alternative with least value of relative closeness coefficient is selected as the best alternative.

The rest of the paper has been framed as follows:

In section 2, we recall some relevant definitions and properties. General TOPSIS approach is discussed in section 3. TOPSIS for MAGDM is stepwise proposed in section 4. A numerical example is described and solved in section 5. Section 6 presents conclusions and future scope of research.

## 2. Some well established definitions and properties

In this section, we recall some established definitions and properties which are connected in the present article.

**2.1. Neutrosophic set (NS)**[1]

Let Y be a space of points (objects) with generic element y in Y. A neutrosophic set A in Y is denoted by

A= {<y: $T_A(y)$, $I_A(y)$, $F_A(y)$>: $y \in Y$ } where $T_A$, $I_A$, $F_A$ represent membership, indeterminacy and non-membership function respectively. $T_A$, $I_A$, $F_A$ are defined as follows:

$T_A : Y \to ]^-0, 1^+[$

$I_A : Y \to ]^-0, 1^+[$

$F_A : Y \to ]^-0, 1^+[$

Here, $T_A(y)$, $I_A(y)$, $F_A(y)$ are the real standard or non-standard subset of $]^-0, 1^+[$ and

$^-0 \le T_A(y)+I_A(y)+F_A(y) \le 3^+$

**2.2. Single valued neutrosophic set (SVNS)** [4]

Let Y be a space of points with generic element in $y \in Y$. A single valued neutrosophic set A in Y is characterized by a truth-membership function $T_A(y)$, an indeterminacy-membership function $I_A(y)$ and a falsity-membership function $F_A(y)$, for each point y in Y, $T_A(y)$, $I_A(y)$, $F_A(y) \in [0, 1]$, when Y is continuous then single-valued neutrosophic set A can be written as

$A = \int_A < T_A(y), I_A(y), F_A(y) > / y, y \in Y$

When A is discrete, single-valued neutrosophic set can be written as $\sum_{i=1}^{n} \langle T_A(y_i), I_A(y_i), F_A(y_i) > / y_i, y_i \in Y$

**2.3. Complement of neutrosophic set** [1]

The complement of a neutrosophic set A is denoted by A′and defined as

A′= {<y: $T_{A'}(y)$, $I_{A'}(y)$, $F_{A'}(y)$>, $y \in Y$ }

$T_{A'}(y) = \{1^+\} - T_A(y)$

$I_{A'}(y) = 1^+\} - I_A(y)$

$F_{A'}(y) = \{1^+\} - F_A(y)$

**2.4 Properties**

Let A and B be two SVNSs, then the following properties [1] hold good:

$1. A \oplus B = \langle T_A(x) + T_B(x) - T_A(x).T_B(x), I_A(x).I_B(x), F_A(x).F_B(x)\rangle, \forall x \in X$

$2. A \otimes B = \langle T_A(x).T_B(x), I_A(x).+ I_B(x) - I_A(x).I_B(x), F_A(x) + F_B(x) - F_A(x).F_B(x)\rangle, \forall x \in X$

$3. A \cup B = \langle \max(T_A(x), T_B(x)), \min(I_A(x), I_B(x)), \min(F_A(x), F_B(x))\rangle$

$4. A \cap B = \langle \min(T_A(x), T_B(x)), \max(I_A(x), I_B(x)), \max(F_A(x), F_B(x))\rangle$

**2.5 Euclidean distance between two SVNSs [50]**

Let $A = \{\langle x_i : T_A(x_i), I_A(x_i), F_A(x_i)\rangle, i = 1,2,...,n\}$, and $B = \{\langle x_i : T_B(x_i), I_B(x_i), F_B(x_i)\rangle, i = 1, 2, ..., n\}$ be SVNSs. Then the Euclidean distance between two SVNSs A and B can be defined as follows:

$$E(A,B) = \sqrt{\sum_{i=1}^{n}((T_A(x_i) - T_A(x_i))^2 + (I_A(x_i) - I_B(x_i))^2 + (F_A(x_i) - F_B(x_i))^2)} \tag{1}$$

The normalized Euclidean distance between two SVNSs A and B can be defined as follows:

$$E_N(A,B) = \sqrt{\frac{1}{3n}\sum_{i=1}^{n}((T_A(x_i) - T_A(x_i))^2 + (I_A(x_i) - I_B(x_i))^2 + (F_A(x_i) - F_B(x_i))^2)} \tag{2}$$

**2.6 Neutrosophic refined set [44]**

Let A be a neutrosophic refined set.

A= $\{<x, T_A^1(x_i), T_A^2(x_i), ..., T_A^m(x_i)), (I_A^1(x_i), I_A^2(x_i), ..., I_A^m(x_i)), (F_A^1(x_i), F_A^2(x_i), ..., F_A^m(x_i))>: x \in X\}$ where, $T_A^j(x_i)$ : $X \in [0,1]$, $I_A^j(x_i)$ : $X \in [0,1]$, $F_A^j(x_i)$ : $X \in [0,1]$, j = 1, 2, …,m such that $0 \le \sup T_A^j(x_i) + \sup I_A^j(x_i) + \sup F_A^j(x_i) \le 3$, for j = 1, 2, …, m for any $x \in X$. Now, $(T_A^j(x_i), I_A^j(x_i), F_A^j(x_i))$ is the truth-membership sequence, indeterminacy-membership sequence and falsity-membership sequence of the element x, respectively. Also, m is called the dimension of neutrosophic refined sets A.

**2.7 Crispfication of a Neutrosophic set [33]**

Let $A_j = \{\langle x_i : T_{A_j}(x_i), I_{A_j}(x_i), F_{A_j}(x_i)\rangle, j = 1,2,...,n\}$ be n SVNSs. The equivalent crisp number of each $A_j$ can be defined as

$$A_j^c = \frac{1 - \sqrt{((1 - T_{A_j}(x_i))^2 + (I_{A_j}(x_i))^2 + (F_{A_j}(x_i))^2)/3}}{\sum_{j=1}^{n}\left\{1 - \sqrt{((1 - T_{A_j}(x_i))^2 + (I_{A_j}(x_i))^2 + (F_{A_j}(x_i))^2)/3}\right\}}. \tag{3}$$

**2.8 Aggregation operator [15]**

In the present problem, there are p alternatives. The aggregation operator [15] applied to neutrosophic refined set is defined as follows:

$$F(D_1, D_2, ..., D_r) = \langle \prod_{i=1}^{r}(T_{ij}^k)^{w_i}, \prod_{i=1}^{r}(I_{ij}^k)^{w_i}, \prod_{i=1}^{r}(F_{ij}^k)^{w_i}\rangle$$

$$\tilde{d}_{kj} = \langle \prod_{i=1}^{r}(T_{ij}^k)^{w_i}, \prod_{i=1}^{r}(I_{ij}^k)^{w_i}, \prod_{i=1}^{r}(F_{ij}^k)^{w_i}\rangle, \tag{4}$$

or $\tilde{d}_{kj} = \langle \tilde{T}_{kj}, \tilde{I}_{kj}, \tilde{F}_{kj}\rangle$ where $i = 1,2,...,r;\ j = 1,2,...,q$ and $k = 1,2,...,p$

**Proof:** For the proof see [15].

**Properties**

The three main properties of aggregation operator are given below:

**i) Idempotency:**

Let $D_1 = D_2 = \ldots = D_r = D$ where $D = \langle T, I, F\rangle$, then $F(D_1, D_2, ..., D_r) = D$

$$F(D_1, D_2, ..., D_r) = \langle \prod_{i=1}^{r} (T_{ij}^k)^{w_i}, \prod_{i=1}^{r} (I_{ij}^k)^{w_i}, \prod_{i=1}^{r} (F_{ij}^k)^{w_i} \rangle$$

$D_1=D_2=\ldots=D_r=D$ in other words $T_{ij}^k = T, I_{ij}^k = I, F_{ij}^k = F$

$$F(D_1, D_2, ..., D_r) = F(D, D, ..., D) = \langle T^{\sum_{i=1}^{r} w_i}, I^{\sum_{i=1}^{r} w_i}, F^{\sum_{i=1}^{r} w_i} \rangle = \langle T, I, F \rangle = D \text{ since, } \sum_{i=1}^{r} w_i = 1 \qquad (4.1)$$

**ii) Boundedness:**

Since, $0 \le w_i \le 1$ and $0 \le (T_{ij}^k)^{w_i} \le 1$, $0 \le (I_{ij}^k)^{w_i} \le 1$, $0 \le (F_{ij}^k)^{w_i} \le 1$

then $0 \le \prod_{i=1}^{r} (T_{ij}^k)^{w_i} \le 1, 0 \le \prod_{i=1}^{r} (I_{ij}^k)^{w_i} \le 1, 0 \le \prod_{i=1}^{r} (F_{ij}^k)^{w_i} \le 1$

$$\text{therefore, } \langle 0, 1, 1 \rangle \le F(D_1, D_2, ..., D_r) \le \langle 1, 0, 0 \rangle \qquad (4.2)$$

**iii) Monotonicity:**

Let us suppose, $D_j \le D_j^* \ \forall j = 1, 2, ..., r.$

Then $(T_{ij}^k)^{w_i} \le (T_{ij}^{*k})^{w_i}$, $(I_{ij}^k)^{w_i} \ge (I_{ij}^{*k})^{w_i}$, $(F_{ij}^k)^{w_i} \ge (F_{ij}^{*k})^{w_i}$ which implies $\prod_{i=1}^{r} (T_{ij}^k)^{w_i} \le \prod_{i=1}^{r} (T_{ij}^{*k})^{w_i}$

$$, \prod_{i=1}^{r} (I_{ij}^k)^{w_i} \ge \prod_{i=1}^{r} (I_{ij}^{*k})^{w_i} \ \prod_{i=1}^{r} (F_{ij}^k)^{w_i} \ge \prod_{i=1}^{r} (F_{ij}^{*k})^{w_i} \text{ i.e. } F(D_1, D_2, ..., D_r) \subset F(D_1^*, D_2^*, ..., D_r^*) \qquad (4.3)$$

## 3. TOPSIS approach

TOPSIS approach is employed to identify the best alternative based on the concept of compromise solution. The best compromise solution reflects the shortest Euclidean distance from the positive ideal solution and the farthest Euclidean distance from the negative ideal solution. TOPSIS approach can be presented as follows:

Assume that $A = \{A_1, A_2, ..., A_m\}$ be the set of alternatives with the set C of q attributes, namely, $C = \{C_1, C_2, ..., C_q\}$, $D = (d_{ij})_{m \times q}$ be the decision matrix and $W = \{w_1, w_2, ..., w_q\}$ be the weight vector of attributes.

**3.1 Normalize and weighted normalized form of decision matrix**

**i) For the profit matrix**

Let $d_j^+ = \max_i(d_{ij})$ and $d_j^- = \min_i(d_{ij})$, then the normalized value of $d_{ij}$ becomes $d_{ij}^N = \dfrac{d_{ij} - d_j^-}{d_j^+ - d_j^-}$ (5)

**ii) For the cost matrix**

Let $d_j^+ = \max_i(d_{ij})$ and $d_j^- = \min_i(d_{ij})$, then the normalized value of $d_{ij}$ becomes $d_{ij}^N = \dfrac{d_j^+ - d_{ij}}{d_j^+ - d_j^-}$ (6)

**iii)** The weighted normalized decision matrix is defined as $d_{ij}^W = d_{ij}^N \times w_j$ (7)

Here, $i = 1, 2, ..., m; j = 1, 2, ..., q$, $w_j \ge 0$, and $\sum_{j=1}^{q} w_j = 1$

**3.2 Positive Ideal Solution (PIS) and Negative Ideal Solution (NIS)**

i)The PIS for the profit matrix can be written as PIS = $\{d_1^{w+}, d_2^{w+}, ..., d_q^{w+}\} = \max_i d_{ij}^w$

ii) The PIS for the cost matrix can be written as PIS = $\{d_1^{w+}, d_2^{w+}, ..., d_q^{w+}\} = \min_i d_{ij}^w$

iii) The NIS for the profit matrix can be written as NIS = $\{d_1^{w-}, d_2^{w-}, ..., d_q^{w-}\} = \min_i d_{ij}^w$

iv)The NIS for the cost matrix can be written as NIS = $\{d_1^{w-}, d_2^{w-}, ..., d_q^{w-}\} = \max_i d_{ij}^w$
$i = 1,2,...,m; j = 1,2,...,q$

**3.3 Euclidean distances from PIS and NIS**

The deviational values from PIS and NIS can be respectively calculated as:

$$E_i^+ = \sqrt{\sum_{j=1}^{q} (d_{ij}^w - d_j^{w+})^2} \quad i = 1,2,...,m \tag{8}$$

$$E_i^- = \sqrt{\sum_{j=1}^{q} (d_{ij}^w - d_j^{w-})^2} \quad i = 1,2,...,m \tag{9}$$

**3.4 Determination of relative closeness coefficients**

The relative closeness coefficient for each alternative can be written as

$$E_i = \frac{E_i^+}{E_i^+ + E_i^-} \quad i = 1,2,...,m \tag{10}$$

**3.5 Ranking of alternatives**

Using relative closeness coefficients, the ranking has been made in the ascending order.

# 4. TOPSIS approach for MAGDM with neutrosophic refined set

A systematic approach to extend the TOPSIS to the refined neutrosophic environment has been proposed in this section. This method is very suitable for solving the group decision-making problem under the refined neutrosophic environment.

**Step 1:**

Let us consider a group of r decision makers ($D_1$, $D_2$…,$D_r$) and q attributes ($C_1$, $C_2$…,$C_q$). The decision matrix (see Table 1) can be presented as follows:

**Table 1:** *Decision matrix*

| | $C_1$ | $C_2$ | ... | $C_q$ |
|---|---|---|---|---|
| $D_1$ | $\langle T_{11}^1, I_{11}^1, F_{11}^1\rangle, A_1$<br>$\langle T_{11}^2, I_{11}^2, F_{11}^2\rangle, A_2$<br>.................,<br>.................,<br>$\langle T_{11}^p, I_{11}^p, F_{11}^p\rangle A_p$ | $\langle T_{12}^1, I_{12}^1, F_{12}^1\rangle, A_1$<br>$\langle T_{12}^2, I_{12}^2, F_{12}^2\rangle, A_2$<br>.................,<br>.................,<br>$\langle T_{12}^p, I_{12}^p, F_{12}^p\rangle A_p$ | ... | $\langle T_{1q}^1, I_{1q}^1, F_{1q}^1\rangle, A_1$<br>$\langle T_{1q}^2, I_{1q}^2, F_{1q}^2\rangle, A_2$<br>.................,<br>.................,<br>$\langle T_{1q}^p, I_{1q}^p, F_{1q}^p\rangle A_p$ |
| $D_2$ | $\langle T_{21}^1, I_{21}^1, F_{21}^1\rangle, A_1$<br>$\langle T_{21}^2, I_{21}^2, F_{21}^2\rangle, A_2$<br>.................,<br>.................,<br>$\langle T_{21}^p, I_{21}^p, F_{21}^p\rangle A_p$ | $\langle T_{22}^1, I_{22}^1, F_{22}^1\rangle, A_1$<br>$\langle T_{22}^2, I_{22}^2, F_{22}^2\rangle, A_2$<br>.................,<br>.................,<br>$\langle T_{22}^p, I_{22}^p, F_{22}^p\rangle A_p$ | ... | $\langle T_{2q}^1, I_{2q}^1, F_{2q}^1\rangle, A_1$<br>$\langle T_{2q}^2, I_{2q}^2, F_{2q}^2\rangle, A_2$<br>.................,<br>.................,<br>$\langle T_{2q}^p, I_{2q}^p, F_{2q}^p\rangle A_p$ |
| ... | ... | ... | ... | ... |
| | $\langle T_{r1}^1, I_{r1}^1, F_{r1}^1\rangle, A_1$<br>$\langle T_{r1}^2, I_{r1}^2, F_{r1}^2\rangle, A_2$ | $\langle T_{r2}^1, I_{r2}^1, F_{r2}^1\rangle, A_1$<br>$\langle T_{r2}^2, I_{r2}^2, F_{r2}^2\rangle, A_2$ | | $\langle T_{rq}^1, I_{rq}^1, F_{rq}^1\rangle, A_1$<br>$\langle T_{rq}^2, I_{rq}^2, F_{rq}^2\rangle, A_2$ |

$$\begin{array}{c|cccc} & \begin{Bmatrix} \ldots\ldots\ldots, \\ T_{21}^{p}, I_{21}^{p}, F_{21}^{p}\ A_p \end{Bmatrix} & \begin{Bmatrix} \ldots\ldots\ldots, \\ T_{22}^{p}, I_{22}^{p}, F_{22}^{p}\ A_p \end{Bmatrix} & & \begin{Bmatrix} \ldots\ldots\ldots, \\ T_{2q}^{p}, I_{2q}^{p}, F_{2q}^{p}\ A_p \end{Bmatrix} \\ \ldots & \ldots & \ldots & \ldots & \ldots \\ D_r & \begin{Bmatrix} T_{r1}^{1}, I_{r1}^{1}, F_{r1}^{1}\ , A_1 \\ \backslash T_{r1}^{2}, I_{r1}^{2}, F_{r1}^{2} /, A_2 \\ \ldots\ldots\ldots, \\ \ldots\ldots\ldots, \\ \langle T_{r1}^{p}, I_{r1}^{p}, F_{r1}^{p} \rangle A_p \end{Bmatrix} & \begin{Bmatrix} T_{r2}^{1}, I_{r2}^{1}, F_{r2}^{1}\ , A_1 \\ \backslash T_{r2}^{2}, I_{r2}^{2}, F_{r2}^{2} /, A_2 \\ \ldots\ldots\ldots, \\ \ldots\ldots\ldots, \\ \langle T_{r2}^{p}, I_{r2}^{p}, F_{r2}^{p} \rangle A_p \end{Bmatrix} & \ldots & \begin{Bmatrix} T_{rq}^{1}, I_{rq}^{1}, F_{rq}^{1}\ , A_1 \\ \backslash T_{rq}^{2}, I_{rq}^{2}, F_{rq}^{2} /, A_2 \\ \ldots\ldots\ldots, \\ \ldots\ldots\ldots, \\ \langle T_{rq}^{p}, I_{rq}^{p}, F_{rq}^{p} \rangle A_p \end{Bmatrix} \end{array} \quad (11)$$

**Step 2**

**Crispfication of neutrosophic weights**

The r decision makers have their own neutrosophic decision weights $(w_1, w_2, \ldots w_r)$. Each $w_k = \langle T_k, I_k, F_k \rangle$ is represented by a neutrosophic number. The equivalent crisp weight can be obtained using the equation (3)

$$w_k^c = \frac{1 - \sqrt{((1-T_k)^2 + (I_k)^2 + (F_k)^2))/3}}{\sum_{k=1}^{r}\left\{1 - \sqrt{((1-T_k)^2 + (I_k)^2 + (F_k)^2)/3}\right\}}, \text{ and}$$

$$w_k^c \geq 0, \ \sum_{k=1}^{r} w_k^c = 1 \quad (12)$$

**Step 3**

**Construction of aggregated decision matrix**

The aggregated neutrosophic decision matrix (see Table 2) can be constructed as follows:

**Table 2:** *Aggregated decision matrix*

| | $C_1$ | $C_2$ | ... | $C_q$ |
|---|---|---|---|---|
| $A_1$ | $\tilde{d}_{11}$ | $\tilde{d}_{12}$ | ... | $\tilde{d}_{1q}$ |
| $A_2$ | $\tilde{d}_{21}$ | $\tilde{d}_{22}$ | ... | $\tilde{d}_{2q}$ |
| ... | ... | ... | ... | ... |
| $A_p$ | $\tilde{d}_{p1}$ | $\tilde{d}_{p2}$ | ... | $\tilde{d}_{pq}$ |

(13)

**Step 4**

**Description of weights of attributes**

In decision making situation, decision makers would not like to give equal importance to all attributes. Thus each DM would have different opinion regarding the weights of attribute. For grouped opinion, all DMs' opinions need to be aggregated by the aggregation operator for a particular attribute. The weight matrix (see Table 3) can be written as follows:

**Table 3:** *Weight matrix of attributes*

| | $C_1$ | $C_2$ | ... | $C_q$ |
|---|---|---|---|---|
| $D_1$ | $w'_{11}$ | $w'_{12}$ | ... | $w'_{1q}$ |
| $D_2$ | $w'_{21}$ | $w'_{22}$ | ... | $w'_{2q}$ |
| ... | ... | ... | ... | ... |
| $D`_r$ | $w'_{r1}$ | $w'_{r2}$ | ... | $w'_{rq}$ |

(14)

Here $w'_{ij} = \langle T'_{ij}, I'_{ij}, F'_{ij} \rangle$

The aggregated weight [15] for the attribute $C_j$ is defined as follows:

$$\overline{w}_j = \langle \prod_{i=1}^{r} T'_{ij}, \prod_{i=1}^{r} I'_{ij}, \prod_{i=1}^{r} F'_{ij} \rangle = \langle \overline{T}_j, \bar{I}_j, \overline{F}_j \rangle \ \ j=1,2,\ldots,q \quad (15)$$

**Step 5**

**Construction of aggregated weighted decision matrix**

The aggregated weighted neutrosophic decision matrix (see Table 4) can be formed as:

**Table 4:** *Aggregated weighted decision matrix*

$$\begin{array}{c|cccc} & C_1 & C_2 & \dots & C_q \\ \hline A_1 & \bar{w}_1\tilde{d}_{11} & \bar{w}_2\tilde{d}_{12} & \dots & \bar{w}_q\tilde{d}_{1q} \\ A_2 & \bar{w}_1\tilde{d}_{21} & \bar{w}_2\tilde{d}_{22} & \dots & \bar{w}_q\tilde{d}_{2q} \\ \dots & \dots & \dots & \dots & \dots \\ A_p & \bar{w}_1\tilde{d}_{p1} & \bar{w}_2\tilde{d}_{p2} & \dots & \bar{w}_q\tilde{d}_{pq} \end{array} \quad (16)$$

$$= \begin{array}{c|cccc} & C_1 & C_2 & \dots & C_q \\ \hline A_1 & \langle T^w_{11}, I^w_{11}, F^w_{11}\rangle & \langle T^w_{12}, I^w_{12}, F^w_{12}\rangle & \dots & \langle T^w_{1q}, I^w_{1q}, F^w_{1q}\rangle \\ A_2 & \langle T^w_{21}, I^w_{21}, F^w_{21}\rangle & \langle T^w_{22}, I^w_{22}, F^w_{22}\rangle & \dots & \langle T^w_{2q}, I^w_{2q}, F^w_{2q}\rangle \\ \dots & \dots & \dots & \dots & \dots \\ A_p & \langle T^w_{p1}, I^w_{p1}, F^w_{p1}\rangle & \langle T^w_{p2}, I^w_{p2}, F^w_{p2}\rangle & \dots & \langle T^w_{pq}, I^w_{pq}, F^w_{pq}\rangle \end{array} \quad (17)$$

$$\text{or } \bar{w}_j\tilde{d}_{kj} = \langle \bar{T}_j, \bar{I}_j, \bar{F}_j\rangle \otimes \langle \tilde{T}_{kj}, \tilde{I}_{kj}, \tilde{F}_{kj}\rangle = \langle \bar{T}_j.\tilde{T}_{kj}, \bar{I}_j + \tilde{I}_{kj} - \bar{I}_j.\tilde{I}_{kj}, \bar{F}_j + \tilde{F}_{kj} - \bar{F}_j.\tilde{F}_{kj}\rangle = \langle T^w_{kj}, I^w_{kj}, F^w_{kj}\rangle = (d^w_{kj})_{p\times q} \quad (18)$$

where $k = 1, 2, \dots, p$ and $j = 1, 2, \dots, q$.

**Step 6**

**Relative positive ideal solution (RPIS) and relative negative ideal solution (RNIS)**

In this step, we find out relative positive ideal solution (RPIS) $(S^+_N)$ and the relative negative ideal solution (RNIS) $(S^-_N)$ for the above aggregated neutrosophic decision matrix. The RPIS is defined as $S^+_N = \{d^{w+}_1, d^{w+}_2, \dots, d^{w+}_q\}$, where $d^{w+}_j = \langle T^{w+}_j, I^{w+}_j, F^{w+}_j\rangle$ and

$$\langle T^{w+}_j, I^{w+}_j, F^{w+}_j\rangle = \langle \max_k T^w_{kj}, \min_k I^w_{kj}, \min_k F^w_{kj}\rangle \text{ (for profit type attribute)} \quad (19)$$

Or

$$\langle T^{w+}_j, I^{w+}_j, F^{w+}_j\rangle = \langle \min_k T^w_{kj}, \max_k I^w_{kj}, \max_k F^w_{kj}\rangle \text{ (for cost type attribute)} \quad (20)$$

The RNIS is defined as $S^-_N = \{d^{w-}_1, d^{w-}_2, \dots, d^{w-}_q\}$, where $d^{w-}_j = \langle T^{w-}_j, I^{w-}_j, F^{w-}_j\rangle$ and

$$\langle T^{w-}_j, I^{w-}_j, F^{w-}_j\rangle = \langle \min_k T^w_{kj}, \max_k I^w_{kj}, \max_k F^w_{kj}\rangle \text{ (for profit type attribute)} \quad (21)$$

Or

$$\langle T^{w-}_j, I^{w-}_j, F^{w-}_j\rangle = \langle \max_k T^w_{kj}, \min_k I^w_{kj}, \min_k F^w_{kj}\rangle \text{ (for cost type attribute)} \quad (22)$$

**Step 7**

**Determination of distances of each alternative from the RPIS and the RNIS**

The normalized Euclidean distance between $\langle T^w_{kj}, I^w_{kj}, F^w_{kj}\rangle$ and $\langle T^{w+}_j, I^{w+}_j, F^{w+}_j\rangle$ can be written as below:

$$Eu^+_k = \sqrt{\frac{1}{3q}\sum_{j=1}^{q}\left(\left(T^w_{kj} - T^{w+}_j\right)^2 + (I^w_{kj} - I^{w+}_j)^2 + (F^w_{kj} - F^{w+}_j)\right)^2} \quad (23)$$

$$Eu^-_k = \sqrt{\frac{1}{3q}\sum_{j=1}^{q}\left(\left(T^w_{kj} - T^{w-}_j\right)^2 + (I^w_{kj} - I^{w-}_j)^2 + (F^w_{kj} - F^{w-}_j)\right)^2} \quad (24)$$

**Step 8**

**Calculation of relative closeness coefficient**

The relative closeness coefficient for each alternative $A_k$ with respect to $S^+_N$ is defined as:

$$R_k = \frac{Eu^+_k}{Eu^+_k + Eu^-_k} \quad (25)$$

where $0 \le R_k \le 1$

**Step 9**
**Ranking of alternatives**
The alternative, for which the closeness coefficient is least, has become the best alternative.

## 5. Numerical Example

The stepwise description of a numerical example is presented as below:

**Step 1**

Suppose that the owner of a small shop wants to buy a tab. After initial screening,three tabs from three different companies $A_1$, $A_2$, $A_3$ remain for further evaluation. A committee comprising of four decision makers, namely, $D_1$, $D_2$, $D_3$, $D_4$, has been formed in order to buy the most suitable tablet with respect to five main attributes, $C_1$, $C_2$, $C_3$, $C_4$, $C_5$. The five attributes have been described below:

i. technical specifications ($C_1$)
ii. quality ($C_2$)
iii. supply chain reliability ($C_3$),
iv. finances ($C_4$)) and
v. ecology ($C_5$)

In the present problem, r = 4, q = 1, 2, …, 5, p = 1, 2, 3.

**Step 1**

The profit type decision matrix (see Table 5) can be written as:

**Table 5:** *Decision matrix*

| | | $C_1$ | $C_2$ | $C_3$ | $C_4$ | $C_5$ |
|---|---|---|---|---|---|---|
| $D_1$ | $A_1$ | (0.7, 0.2, 0.1) | (0.8, 0.3, 0.3) | (0.4, 0.1, 0.2) | (0.5, 0.1, 0.1) | (0.6,0.4,0.1) |
| | $A_2$ | (0.6, 0.2, 0.1) | (0.7, 0.4, 0.2) | (0.3, 0.2, 0.1) | (0.3, 0.1, 0.2) | (0.8,0.2, 0.2) |
| | $A_3$ | (0.7, 0.1, 0.2) | (0.6, 0.2, 0.2) | (0.4, 0.4, 0.4) | (0.6, 0.1, 0.1) | (0.7, 0.1, 0.1) |
| $D_2$ | $A_1$ | (0.8,0.2,0.1) | (0.7, 0.1, 0.2) | (0.5, 0.1, 0.1) | (0.6, 0.2, 0.3) | (0.5, 0.6, 0.1) |
| | $A_2$ | (0.7, 0.3, 0.2) | (0.6, 0.1, 0.1) | (0.6, 0.2, 0.3) | (0.5, 0.1, 0.2) | (0.4, 0.5, 0.2) |
| | $A_3$ | (0.6, 0.2, 0.2) | (0.8, 0.2, 0.1) | (0.6, 0.1, 0.2) | (0.7, 0.1, 0.1) | (0.5, 0.5, 0.1) |
| $D_3$ | $A_1$ | (0.9, 0.1, 0.1) | (0.5, 0.3, 0.2) | (0.6, 0.4, 0.1) | (0.2, 0.5, 0.3) | (0.4,0.4,0.4) |
| | $A_2$ | (0.8, 0.2, 0.1) | (0.6, 0.3, 0.1) | (0.5, 0.4, 0.1) | (0.4, 0.2, 0.1) | (0.5,0.3, 0.2) |
| | $A_3$ | (0.8, 0.1, 0.2) | (0.7, 0.1, 0.1) | (0.6, 0.3, 0.2) | (0.4, 0.1, 0.1) | (0.6,0.1, 0.2) |
| $D_4$ | $A_1$ | (0.6, 0.1, 0.1) | (0.8,0.2, 0.1) | (0.9, 0.2, 0.3) | (0.7, 0.4,0.3) | (0.7, 0.3,0.4) |
| | $A_2$ | (0.7, 0.2, 0.1) | (0.7, 0.1,0.3) | (0.7, 0.3, 0.1) | (0.6,0.5,0.1) | (0.6,0.2,0.4) |
| | $A_3$ | (0.7, 0.1, 0.2) | (0.6, 0.1, 0.2) | (0.6, 0.2, 0.1) | (0.7, 0.1,0.3) | (0.7, 0.3,0.2) |

**Step 2**
The neutrosophic weights of decision makers are considered as {(0.8, 0.1, 0.1), (0.9, 0.2, 0.1), (0.5, 0.4, 0.1), (0.8, 0.2, 0.2)}. Using the equation (10), the equivalent crisp weights are {0.27317, 0.27317, 0.19912, 0.25453}.
**Step 3**
The aggregated decision matrix can be determined by applying the aggregated operator (4) and calculated as below:

**Table 6:** *Aggregated decision matrix*

| | $C_1$ | $C_2$ | $C_3$ | $C_4$ | $C_5$ |
|---|---|---|---|---|---|
| $A_1$ | (0.734, 0.146, 0.1) | (0.702, 0.201, 0.187) | (0.567, 0.157, 0.16) | (0.477, 0.237, 0.222) | (0.548, 0.415, 0.188) |
| $A_2$ | (0.689, 0.224, 0.121) | (0.651, 0.182, 0.16) | (0.498 0.255, 0.135) | (0.436, 0.173, 0.146) | (0.5603, 0.279, 0.239) |
| $A_3$ | (0.689, 0.121, 0.2) | (0.669, 0.146, 0.144) | (0.537, 0.217, 0.217) | (0.6, 0.1, 0.132) | (0.619, 0.205, 0.137) |

**Step 4**

The weight matrix (see Table 7) of attributes as described in (14) can be displayed as follows:

**Table 7:** *Weight matrix of attributes*

| | $C_1$ | $C_2$ | $C_3$ | $C_4$ | $C_5$ |
|---|---|---|---|---|---|
| $D_1$ | (0.9, 0.1, 0.2) | (0.8, 0.2, 0.3) | (0.5, 0.4, 0.3) | (0.5, 0.2, 0.15) | (0.5, 0.4, 0.4) |
| $D_2$ | (0.8, 0.2, 0.1) | (0.7, 0.1, 0.3) | (0.6, 0.3, 0.3) | (0.8, 0.25, 0.1) | (0.6, 0.3, 0.4) |
| $D_3$ | (0.6, 0.3, 0.2) | (0.5, 0.3, 0.2) | (0.8, 0.2, 0.1) | (0.7, 0.2, 0.1) | (0.4, 0.4, 0.4) |
| $D_4$ | (0.6, 0.1, 0.2) | (0.6, 0.1, 0.2) | (0.6, 0.2, 0.3) | (0.5, 0.1, 0.2) | (0.3, 0.2, 0.1) |

The aggregated weights for all attributes are presented below:

$$\overline{w} = \{(0.725, 0.15, 0.166), (0.653, 0.15, 0.25), (0.604, 0.27, 0.241), (0.608, 0.178, 0.133), (0.444, 0.31, 0.281)\}.$$

**Step 5**

The aggregated weighted neutrosophic decision matrix (see Table 8) can be formed as:

**Table 8:** *The aggregated weighted neutrosophic decision matrix*

| | $C_1$ | $C_2$ | $C_3$ | $C_4$ | $C_5$ |
|---|---|---|---|---|---|
| $A_1$ | (0.532, 0.274, 0.249) | (0.458, 0.321, 0.390) | (0.342, 0.385, 0.362) | (0.29, 0.373, 0.325) | (0.243, 0.596, 0.416) |
| $A_2$ | (0.4995, 0.340, 0.2669) | (0.425, 0.305, 0.37) | (0.301, 0.456, 0.343) | (0.265, 0.32, 0.2596) | (0.249, 0.502, 0.453) |
| $A_3$ | (0.4995, 0.253, 0.333) | (0.437, 0.274, 0.358) | (0.324, 0.428, 0.406) | (0.365, 0.260, 0.247) | (0.275, 0.451, 0.3795) |

**Step 6**

Since the present problem is to make decision to buy a tablet, the decision matrix is profit type matrix. Using (19), the RPIS is presented below:

$$S_N^{+} = \{(0.532, 0.253, 0.249), (0.45, 0.274, 0.358), (0.342, 0.385, 0.343), (0.365, 0.26, 0.247), (0.275, 0.451, 0.3795)\}.$$

Using (21) the RNIS is presented below:

$$S_N^{-} = \{(0.4995, 0.340, 0.333), (0.425, 0.321, 0.39), (0.301, 0.456, 0.406), (0.265, 0.373, 0.325), (0.243, 0.596, 0.453)\}.$$

**Step 7**

The normalized Euclidean distance from RPIS by using (22) is given below:

$Eu_1^{+} = 0.0588$, $Eu_2^{+} = 0.0518$, $Eu_3^{+} = 0.0313$.

The normalized Euclidean distance from RNIS by using (23) is given below:

$Eu_1^{-} = 0.0401$, $Eu_2^{-} = 0.0408$, $Eu_3^{-} = 0.0676$.

**Step 8**

The relative closeness coefficient (24) for each alternative has been presented in the table 9.

**Table 9:** *Ranking of alternatives*

| Alternatives | $R_k = \frac{Eu_k^+}{Eu_k^+ + Eu_k^-}$ | Ranking |
|---|---|---|
| $A_1$ | 0.594 | 3 |
| $A_2$ | 0.559 | 2 |
| $A_3$ | 0.316 | 1 |

**Step 9**

Table 9 reflects that $A_3$ is the most suitable tablet for purchasing.

## 6. Conclusion

This paper presents TOPSIS approach for MAGDM for refined neutrosophic environment. This is the first attempt to propose TOPSIS in refined neutrosophic environment. The proposed approach can be applied to other real MAGDM problem in refined neutrosophic environment such as project management in IT sectors, banking system, etc. The Authors hope that this proposed approach will enlighten a new path for MAGDM in refined neutrosophic environment.

## References

1. F. Smarandache. A unifying field in logics: neutrosophic logic, neutrosophy, neutrosophic set, neutrosophic probability, and neutrosophic statistics, Rehoboth: American research Press (1998).
2. L.A. Zadeh. Fuzzy sets. Information and Control 8(3) (1965), 338-3534.
3. K. T. Atanassov. Intuitionistic fuzzy sets. Fuzzy Sets and Systems 20 (1986), 87-96.
4. H. Wang, F. Smarandache, Y. Q. Zhang, R. Sunderraman. Single valued neutrosophic sets. Multispace and Multi structure 4 (2010), 410–413.
5. C. L. Hwang, K. Yoon. Multiple attribute decision making: methods and applications. Springer, New York (1981).
6. C. T. Chen. Extensions of TOPSIS for group decision making under fuzzy environment. Fuzzy Sets and Systems 114 (2000), 1- 9.
7. F. E. Boran, S. Genc, M. Kurt, D. Akay. A multui-criteria intuitionistic fuzzy group decision making for supplier selection with TOPSIS method. Expert System Application 36 (8) (2009), 11363 - 11368.
8. J. Ye. Multicriteria decision – making method using the correlation coefficient under single valued neutrosophic environment. International Journal of General Systems 42 (4) (2013), 386-394.
9. J. Ye. Single valued neutrosophic cross entropy for multi-criteria decision making problems. Applied Mathematical Modelling 38 (3) (2013), 1170- 1175.
10. P. Biswas, S. Pramanik, B.C. Giri. Entropy based grey relational analysis method for multi-attribute decision - making under single valued neutrosophic assessments. Neutrosophic Sets and Systems 2 (2014), 102 - 110.
11. P. Biswas, S. Pramanik, B.C. Giri. A new methodology for neutrosophic multi-attribute decision making with unknown weight information. Neutrosophic Sets and Systems 3 (2014), 42 - 52.
12. S. Pramanik, P. Biswas, B.C. Giri. Hybrid vector similarity measures and their applications to multi-attribute decision making under neutrosophic environment. Neural Computing and Applications (2015). DOI: 10.1007/s00521-015-2125-3.
13. H. Y. Zhang, J. Q. Wang, X. H. Chen. Interval neutrosophic sets and their application in multi – criteria decision making problems. The Scientific World Journal 2014 (2014).http://dx.doi.org/10.1155/2014/645953.
14. S. Pramanik, K. Mondal. Interval neutrosophic multi-Attributed-making based on grey relational analysis. Neutrosophic Sets and Systems 9 (2015), 13-22.

15. J. Ye. A multicriteria decision - making method using aggregation operators for simplifiedneutrosophicsets. Journal of Intelligent and Fuzzy Systems 26 (2014) 2459 – 2466.
16. Z. Tian, J. Wang, J. Wang, and H. Zhang. Simplified neutrosophic linguistic multi-criteria group decision-making approach to green product development. Group Decision and Negotiation (2016). DOI 10.1007/s10726-016-9479-5.
17. Z. Tian, J. Wang, J. Wang, and H. Zhang. An improved MULTIMOORA approach for multi-criteria decision-making based on interdependent inputs of simplified neutrosophic linguistic information. Neural Computing and Applications (2016). DOI 10.1007/s00521-016-2378-5.
18. J. Peng, J. Wang, X. Wu. An extension of the ELECTRE approach with multi-valued neutrosophic information. Neural Computing and Applications (2016). DOI:10.1007/s00521-016-2411-8.
19. P. Ji, H. Zhang, J.Wang. A projection-based TODIM method under multi-valued neutrosophic environments and its application in personnel selection. Neural Computing and Applications (2016). DOI 10.1007/s00521-016-2436-z.
20. H. Zhang, J. Wang, and X. Chen. An outranking approach for multi-criteria decision-making problems with interval-valued neutrosophic sets. Neural Computing and Applications 27(3) (2016), 615–627.
21. P. Biswas, S. Pramanik, B.C. Giri**.** Value and ambiguity index based ranking method of single-valued trapezoidal neutrosophic numbers and its application to multi-attribute decision making. Neutrosophic Sets and Systems 12 (2016), 127-138.
22. P. Biswas, S. Pramanik, B.C. Giri. Aggregation of triangular fuzzy neutrosophic set information and its application to multi-attribute decision making. Neutrosophic Sets and Systems 12 (2016), 20-40.
23. P.P. Dey, S. Pramanik, B.C. Giri. An extended grey relational analysis based multiple attribute decision making in interval neutrosophic uncertain linguistic setting. Neutrosophic Sets and Systems 11 (2016), 21-30.
24. K. Mondal,S. Pramanik. Neutrosophic decision making model for clay-brick selection in construction field based on grey relational analysis. Neutrosophic Sets and Systems 9 (2015), 64-71.
25. K. Mondal, S. Pramanik. Neutrosophic tangent similarity measure and its application to multiple attribute decision making. Neutrosophic Sets and Systems 9 (2015), 85-92.
26. P. Biswas, S. Pramanik, and B.C. Giri. Cosine similarity measure based multi-attribute decision-making with trapezoidal fuzzy neutrosophic numbers. Neutrosophic Sets and Systems 8 (2015), 47-57.
27. J. Ye. Simplified neutrosophic harmonic averaging projection-based method for multiple attribute decision-making problems. International Journal of Machine Learning and Cybernetics (2015). DOI 10.1007/s13042-015-0456-0.
28. K. Mondal, S. Pramanik. Neutrosophic decision making model of school choice. Neutrosophic Sets and Systems, 7 (2015), 62-68.
29. H. Zhang, P. Ji, J., Wang, X.Chen. An improved weighted correlation coefficient based on integrated weight for interval neutrosophic sets and its application in multi-criteria decision making problems. International Journal of Computational Intelligence Systems 8(6) (2015), 1027–1043.
30. J. J. Peng, J. Q. Wang, H. Y. Zhang, X. H. Chen. An outranking approach for multi-criteria decision-making problems with simplified neutrosophic sets. Applied Soft Computing 25 (2014), 336–346.
31. J. Ye. Bidirectional projection method for multiple attribute group decision making with neutrosophic numbers. Neural Computing and Applications (2016). DOI: 10.1007/s00521-015-2123-5.
32. K. Mondal,S. Pramanik. Multi-criteria group decision making approach for teacher recruitment in higher education under simplified neutrosophic environment. Neutrosophic Sets and Systems 6 (2014), 28-34.
33. P. Biswas, S. Pramanik, B.C. Giri.TOPSIS method for multi-attribute group decision making under single-valued neutrosophic environment. Neural Computing and Applications (2015). DOI: 10.1007/s00521-015-1891-2.
34. P. Chi, P. Liu. An extended TOPSIS method for multi-attribute decision making problems on interval neutrosophic set. Neutrosophic Sets and Systems 1 (2013), 63 -70.

35. S. Broumi, J. Ye. F. Smnarandache. An extended TOPSIS method for multiple attribute decision making based on interval neutrosophic uncertain linguistic variables. Neutrosophic Sets and Systems 8 (2015), 22-31.
36. S. Pramanik, P. P. Dey, B. C. Giri. TOPSIS for singled valued soft expert set based multi-attribute decision making problems. Neutrosophic Sets and Systems 10 (2015), 88-95.
37. P.P. Dey, S. Pramanik, B.C. Giri. Generalized neutrosophic soft multi-attribute group decision making based on TOPSIS. Critical Review 11 (2015), 41-55.
38. P. P. Dey, S. Pramanik, B. C. Giri. TOPSIS for solving multi-attribute decision making problems under bi-polar neutrosophic environment. New Trends in Neutrosophic Theories and Applications (2016). In Press.
39. K. Mondal, S. Pramanik, F. Smarandache. TOPSIS in rough neutrosophic environment. Neutrosophic Sets and Systems 13(2016). In Press.
40. R. R. Yager. On the theory of bags (multi sets). International Journal of General Systems 13(1986), 23-37.
41. S. Sebastian, T. V. Ramakrishnan. Multi fuzzy sets: an extension of fuzzy sets. Fuzzy Information Engineering 3(1) (2011), 35-43.
42. T. K. Shinoj, S. J. John. Intuitionistic fuzzy multi-sets and its application in medical diagnosis. WorldAcademy of Science, Engine Technology 61 (2012), 1178-1181.
43. S. Ye, J. Ye. Dice Similarity measure between single valued neutrosophic multisets and its application in medical diagnosis. Neutrosophic Sets and Systems 6 (2014), 50-55.
44. F. Smarandache. n-Valued refined neutrosophic logic and its applications in physics, Progress in Physics 4 (2013), 143-146.
45. S. Broumi, F. Smarandache. Neutrosophic refined similarity measure based on cosine function. Neutrosophic Sets and Systems 6 (2014), 42-48.
46. K. Mondal, S. Pramanik. Neutrosophic refined similarity measure based on tangent function and itsapplication to multi attribute decision making. Journal of New theory 8 (2015), 41-50.
47. K. Mondal, S. Pramanik. Neutrosophic refined similarity measure based on cotangent function and its application to multi-attribute decision making, Global Journal of Advanced Research 2(2) (2015), 486-494.
48. S. Pramanik, D. Banerjee, B.C. Giri. Multi – criteria group decision making model in neutrosophic refined set and its application. Global Journal of Engineering Science and Research Management 3 (6) (2016), 1- 10.
49. S. Nadaban, S. Dzitac. Neutrosophic TOPSIS: a general view. 6 th International Conference on Computers Communications and Control (ICCCC) (2016) 1- 4.
50. P. Majumder, S. K. Samanta. On similarity and entropy of neutrosophic sets. Journal of Intelligent and Fuzzy Systems 26(2014), 1245–1252.

**Kalyan Mondal**[1], **Surapati Pramanik**[2*], **Florentin Smarandache**[3]

1 Department of Mathematics, Jadavpur University, West Bengal, India Email:kalyanmathematic@gmail.com
2 Department of Mathematics, Nandalal Ghosh B.T. College, Panpur, PO-Narayanpur, and District: North 24 Parganas, Pin Code: 743126, West Bengal, India. Corresponding author's E-mail: sura_pati@yahoo.co.in
3 Mathematics & Science Department, University of New Mexico, 705 Gurley Ave., Gallup, NM 87301, USA. E-mail: fsmarandache@gmail.com

# Several Trigonometric Hamming Similarity Measures of Rough Neutrosophic Sets and their Applications in Decision Making

## Abstract

In 2014, Broumi et al. (S. Broumi, F. Smarandache, M. Dhar, Rough neutrosophic sets, Italian Journal of Pure and Applied Mathematics, 32 (2014), 493-502.) introduced the notion of rough neutrosophic set by combining neutrosophic sets and rough sets, which has been a mathematical tool to deal with problems involving indeterminacy and incompleteness. The real world is full of indeterminacy. Naturally, real world decision making problem involves indeterminacy. Rough neutrosophic set is capable of describing and handling imprecise, indeterminate and inconsistent and incomplete information. This paper is devoted to propose several new similarity measures based on trigonometric hamming similarity operators of rough neutrosophic sets and their applications in decision making. We prove the required properties of the proposed similarity measures. To illustrate the applicability of the proposed similarity measures in decision making, an illustrative problem is solved.

## 1. Introduction

L. A. Zadeh [1] introduced the degree of membership in 1965 and defined the concept of fuzzy set to deal with uncertainty. K. T. Atanassov [2] introduced the degree of non-membership as independent component in 1986 and defined the intuitionistic fuzzy set. F. Smarandache [3, 4] introduced the degree of indeterminacy as independent component and defined the neutrosophic set in 1998.

To use the concept of neutrosophic set in practical fields such as real scientific and engineering applications, Wang et al. [5] presented an instance of neutrosophic set, called single valued neutrosophic set (SVNS).

In many applications, due to lack of knowledge or data about the problem domains, the decision information may be provided with intervals, instead of real numbers. To deal with the situation Wang et al. [6] introduced interval valued neutrosophic sets (IVNS), which is characterized by a membership function, non-membership function and an indeterminacy function, whose values are intervals rather than real numbers. Also, the interval valued neutrosophic set can represent uncertain, imprecise, incomplete and inconsistent information which exist in the real world.

In 2014, Broumi et al. [7, 8] introduced the concept of rough neutrosophic set (RNS). It is derived by hybridizing the concepts of rough set proposed by Pawlak [9] and neutrosophic set originated by F. Smarandache [3, 4]. Neutrosophic sets and rough sets are both capable of dealing with uncertainty and partial information. Rough neutrosophic set [7, 8] is the generalization of rough fuzzy sets [10], [11] and rough intuitionistic fuzzy sets [12].

Mondal and Pramanik [13] applied the concept of rough neutrosophic set in multi-attribute decision making based on grey relational analysis in 2015. S. Pramanik and K. Mondal [14] also studied cosine similarity measure of rough neutrosophic sets and its application in medical diagnosis in 2015. Mondal and Pramanik [15] proposed multi attribute decision making using rough accuracy score function. Pramanik and Mondal [16] proposed cotangent similarity measure under rough neutrosophic environment. Pramanik and Mondal [17] further proposed some similarity measures namely Dice similarity measure and Jaccard similarity measure in rough neutrosophic environment. Mondal et al. [18] proposed rough neutrosophic variational coefficient similarity measure and presented its application in multi attribute decision making. Mondal et al. [19] presented rough neutrosophic TOPSIS for multi-attribute group decision making problem. Mondal and Pramanik [20] studied tri-complex rough neutrosophic similarity measure and its application in multi-attribute decision making. Mondal et al. [21] further proposed rough neutrosophic hyper-complex set and its application to multi-attribute decision making.

Literature review reflects that no studies have been made on multi-attribute decision making using trigonometric Hamming similarity measures under rough neutrosophic environment. In this paper, we propose cosine, sine and cotangent Hamming similarity measures under rough neutrosophic environment. We also present a numerical example to show the effectiveness and applicability of the proposed similarity measures.

## 2. Mathematical Preliminaries

**2.1 Neutrosophic set** [3, 4]

Let $U$ be a universe of discourse. Then the neutrosophic set $A$ is presented in the form:

$A = \{< x: T_A(x), I_A(x), F_A(x)>, x \in U\}$, where the functions $T$, $I$, $F$: $U \rightarrow ]^{-}0,1^{+}[$ represent respectively the degree of membership, the degree of indeterminacy, and the degree of non-membership of the element $x \in U$ to the set $P$ satisfying the following the condition.

$^{-}0 \leq \sup T_A(x) + \sup I_A(x) + \sup F_A(x) \leq 3^{+}$

**2.2 Single valued neutrosophic sets** [6]

**Definition 2.2** [6]

Wang et al. [6] mentioned that the neutrosophic set assumes the value from real standard or non-standard subsets of $]^{-}0, 1^{+}[$. So instead of $]^{-}0, 1^{+}[$ Wang et al. [6] consider the interval [0, 1] for technical applications, because $]^{-}0, 1^{+}[$ is difficult to apply in the real applications such as scientific and engineering problems.

Assume that $X$ be a space of points (objects) with generic elements in $X$ denoted by $x$. A SVNS $A$ in $X$ is characterized by a truth-membership function $T_A(x)$, an indeterminacy-membership function $I_A(x)$, and a falsity membership function $F_A(x)$, for each point $x$ in $X$, $T_A(x)$, $I_A(x)$, $F_A(x) \in [0, 1]$. When $X$ is continuous, a SVNS $A$ can be written as follows:

$$A = \int_x \frac{<T_A(x), I_A(x), F_A(x)>}{x} : x \in X$$

When $X$ is discrete, a SVNS $A$ can be written as follows:

$$A = \sum_{i=1}^{n} \frac{<T_A(x_i), I_A(x_i), F_A(x_i)>}{x_i} : x_i \in X.$$

For two SVNSs, $A_{SVNS} = \{<x: T_A(x), I_A(x), F_A(x)> \mid x \in X\}$ and $B_{SVNS} = \{<x, T_B(x), I_B(x), F_B(x)> \mid x \in X\}$, $A_{SVNS} \subseteq B_{SVNS}$ and $A_{SVNS} = B_{SVNS}$ are defined as follows:

(1) $A_{SVNS} \subseteq B_{SVNS}$ if and only if $T_A(x) \leq T_B(x)$, $I_A(x) \geq I_B(x)$, $F_A(x) \geq F_B(x)$

(2) $A_{SVNS} = B_{SVNS}$ if and only if $T_A(x) = T_B(x)$, $I_A(x) = I_B(x)$, $F_A(x) = F_B(x)$ for any $x \in X$

**2.3 Hamming distance** [17]

Hamming distance [17] between two neutrosophic sets $A(T_A(x), I_A(x), F_A(x))$ and $B(T_B(x), I_B(x), F_B(x))$ is defined as

$$H(A,B) = \frac{1}{2} \sum_{i=1}^{n} (|T_A(x) - T_B(x)| + |I_A(x) - I_B(x)| + |F_A(x) - F_B(x)|) \quad (1)$$

**2.4 Rough neutrosophic set (RNS)**

**Definition 2.2.1** [1], [2]: Let $Z$ be a non-null set and $R$ be an equivalence relation on $Z$. Let A be a neutrosophic set in $Z$ with the membership function $T_A$, indeterminacy function $I_A$ and non-membership function $F_A$. The lower and the upper approximations of $A$ in the approximation ($Z$, $R$) denoted by $\underline{N}(A)$ and $\overline{N}(A)$ are respectively defined as follows:

$$\underline{N}(A) = \left\langle < x, T_{\underline{N}(A)}(x), I_{\underline{N}(A)}(x), F_{\underline{N}(A)}(x) > / z \in [x]_R, x \in Z \right\rangle$$

$$\overline{N}(A) = \left\langle < x, T_{\overline{N}(A)}(x), I_{\overline{N}(A)}(x), F_{\overline{N}(A)}(x) > / z \in [x]_R, x \in Z \right\rangle \quad (2)$$

where, $T_{\underline{N}(A)}(x) = \wedge_{z \in [x]_R} T_A(z)$, $I_{\underline{N}(A)}(x) = \wedge_{z \in [x]_R} I_A(z)$, $F_{\underline{N}(A)}(x) = \wedge_{z \in [x]_R} F_A(z)$,

$T_{\overline{N}(A)}(x) = \vee_{z \in [x]_R} T_A(z)$, $I_{\overline{N}(A)}(x) = \vee_{z \in [x]_R} T_A(z)$, $F_{\overline{N}(A)}(x) = \vee_{z \in [x]_R} I_A(z)$.

So, $0 \leq T_{\underline{N}(A)}(x) + I_{\underline{N}(A)}(x) + F_{\underline{N}(A)}(x) \leq 3$ and $0 \leq T_{\overline{N}(A)}(x) + I_{\overline{N}(A)}(x) + F_{\overline{N}(A)}(x) \leq 3$ hold. Here $\vee$ and $\wedge$ denote "max" and "min'' operators respectively. $T_A(z)$, $I_A(z)$ and $F_A(z)$ are the membership, indeterminacy and non-membership degrees of z with respect to A. $\underline{N}(A)$ and $\overline{N}(A)$ are two neutrosophic sets in Z.

Thus, NS mappings $\underline{N}$, $\overline{N}$ : $N(Z) \rightarrow N(Z)$ denote respectively the lower and upper rough NS approximation operators, and the pair $(\underline{N}(A), \overline{N}(A))$ is called the rough neutrosophic set in (Z, $R$).

Based on the above mentioned definition, it is observed that $\underline{N}(A)$ and $\overline{N}(A)$ have constant membership on the equivalence class of $R$, if $\underline{N}(A) = \overline{N}(A)$; i.e. $T_{\underline{N}(A)}(x) = T_{\overline{N}(A)}(x)$, $I_{\underline{N}(A)}(x) = I_{\overline{N}(A)}(x)$, $F_{\underline{N}(A)}(x) = F_{\overline{N}(A)}(x)$.

For any $x$ belongs to $Z$, $P$ is said to be a definable neutrosophic set in the approximation ($Z$, $R$). Obviously, zero neutrosophic set ($0_N$) and unit neutrosophic sets ($1_N$) are definable neutrosophic sets.

**Definition 2.2.2** [1], [2]: Let $N(A) = (\underline{N}(A), \overline{N}(A))$ is a rough neutrosophic set in (Z, $R$). The rough complement of $N(A)$ is denoted by $\sim N(A) = (\underline{N}(A)^c, \overline{N}(A)^c)$, where $\underline{N}(A)^c, \overline{N}(A)^c$ are the complements of neutrosophic sets of $\underline{N}(A), \overline{N}(A)$ respectively.

$$\underline{N}(A)^c = \langle < x, F_{\underline{N}(A)}(x), 1 - I_{\underline{N}(A)}(x), T_{\underline{N}(A)}(x) > /, x \in Z \rangle, \text{ and}$$

$$\overline{N}(A)^c = \langle < x, F_{\underline{N}(A)}(x), 1 - I_{\overline{N}(A)}(x), T_{\overline{N}(A)}(x) > /, x \in Z \rangle \quad (3)$$

**Definition 2.2.3** [1], [2]: Let $N(A)$ and $N(B)$ are two rough neutrosophic sets respectively in $Z$, then the following definitions hold good:

$$N(A) = N(B) \Leftrightarrow \underline{N}(A) = \underline{N}(B) \wedge \overline{N}(A) = \overline{N}(B)$$

$$N(A) \subseteq N(B) \Leftrightarrow \underline{N}(A) \subseteq \underline{N}(B) \wedge \overline{N}(A) \subseteq \overline{N}(B)$$

$$N(A) \cup N(B) = < \underline{N}(A) \cup \underline{N}(B), \overline{N}(A) \cup \overline{N}(B) >$$

$$N(A) \cap N(B) = < \underline{N}(A) \cap \underline{N}(B), \overline{N}(A) \cap \overline{N}(B) >$$

$$N(A) + N(B) = < \underline{N}(A) + \underline{N}(B), \overline{N}(A) + \overline{N}(B) >$$

$$N(A) \,.\, N(B) = < \underline{N}(A) \,.\, \underline{N}(B), \overline{N}(A) \,.\, \overline{N}(B) >$$

If $A$, $B$, $C$ are the rough neutrosophic sets in ($Z$, $R$), then the following propositions can be stated from definitions.

**Proposition 1** [1], [2]:

1. $\sim A(\sim A) = A$
2. $A \cup B = B \cup A, A \cap B = B \cap A$
3. $(A \cup B) \cup C = A \cup (B \cup C)$, $(A \cap B) \cap C = A \cap (B \cap C)$
4. $(A \cup B) \cap C = (A \cup B) \cap (A \cup C)$, $(A \cap B) \cup C = (A \cap B) \cup (A \cap C)$

**Proposition 2** [1], [2]:

De Morgan's Laws are satisfied for rough neutrosophic sets $N(A)$ and $N(B)$

1. $\sim (N(A) \cup N(B)) = (\sim N(A)) \cap (\sim N(B))$
2. $\sim (N(A) \cap N(B)) = (\sim N(A)) \cup (\sim N(B))$

For the proofs of the propositions, see [1, 2]

**Proposition 3**[1], [2]:

If $A$ and $B$ are two neutrosophic sets in $U$ such that $A \subseteq B, then\ N(A) \subseteq N(B)$

1. $N(A \cap B) \subseteq N(A) \cap N(B)$
2. $N(A \cup B) \supseteq N(A) \cup N(B)$

For the proofs of the propositions, see [1, 2]

**Proposition 4** [1], [2]:

1. $\underline{N}(A) = \sim \overline{N}(\sim A)$
2. $\overline{N}(A) = \sim \underline{N}(\sim A)$
3. $\underline{N}(A) \subseteq \overline{N}(A)$

For the proofs of the propositions, see [1, 2]

## 3. Cosine Hamming Similarity Measures of RNS

Assume that $A = \langle (\underline{T}_A(x_i), \underline{I}_A(x_i), \underline{F}_A(x_i)), (\overline{T}_A(x_i), \overline{I}_A(x_i), \overline{F}_A(x_i)) \rangle$ and $B = \langle (\underline{T}_B(x_i), \underline{I}_B(x_i), \underline{F}_B(x_i)), (\overline{T}_B(x_i), \overline{I}_B(x_i), \overline{F}_B(x_i)) \rangle$ in $X$ = {$x_1$, $x_2$, …, $x_n$} be any two rough neutrosophic sets. A cosine Hamming similarity operator between rough neutrosophic sets $A$ and $B$ is defined as follows:

$C_{CHSO}(A, B)=$

$$\frac{1}{n}\sum_{i=1}^{n}\cos\left(\frac{\pi}{6}\left(\left|\Delta T_A(x_i)-\Delta T_B(x_i)\right|+\left|\Delta I_A(x_i)-\Delta I_B(x_i)\right|+\left|\Delta F_A(x_i)-\Delta F_B(x_i)\right|\right)\right) \tag{4}$$

Here, $\Delta T_A(x_i)=\left(\frac{\underline{T}_A(x_i)+\overline{T}_A(x_i)}{2}\right)$, $\Delta T_B(x_i)=\left(\frac{\underline{T}_B(x_i)+\overline{T}_B(x_i)}{2}\right)$,

$\Delta I_A(x_i)=\left(\frac{\underline{I}_A(x_i)+\overline{I}_A(x_i)}{2}\right)$, $\Delta I_B(x_i)=\left(\frac{\underline{I}_B(x_i)+\overline{I}_B(x_i)}{2}\right)$,

$\Delta F_A(x_i)=\left(\frac{\underline{F}_A(x_i)+\overline{F}_A(x_i)}{2}\right)$, $\Delta F_B(x_i)=\left(\frac{\underline{F}_B(x_i)+\overline{F}_B(x_i)}{2}\right)$.

Also, $[\Delta T_A(x), \Delta I_A(x), \Delta F_A(x)] \neq [0, 0, 0]$ and $[\Delta T_B(x), \Delta I_B(x), \Delta F_B(x)] \neq [0, 0, 0]$, $i = 1, 2, \ldots, n$.

**Proposition** 3.1

The defined rough neutrosophic cosine hamming similarity operator $C_{CHSO}(A, B)$ between RNSs $A$ and $B$ satisfies the following properties:

1. $0 \leq C_{RCHSO}(A, B) \leq 1$
2. $C_{CHSO}(A, B) = 1$ if and only if $A = B$
3. $C_{CHSO}(A, B) = C_{CHSO}(B, A)$

Proof of the property 1.

Since the functions $\Delta T_A(x)$, $\Delta I_A(x)$, $\Delta F_A(x)$, $\Delta T_B(x)$, $\Delta I_B(x)$, and $\Delta F_B(x)$, and the value of the cosine function are within [0,1], the similarity measure based on rough neutrosophic cosine hamming similarity function also lies within [ 0,1].

Hence $0 \leq C_{CHSO}(A, B) \leq 1$.

This completes thee proved.

Proof of the property 2.

For any two RNSs $A$ and $B$, if $A = B$, then the following relations hold $\Delta T_A(x_i)=\Delta T_B(x_i)$, $\Delta I_A(x_i)=\Delta I_B(x_i)$, $\Delta F_A(x_i)=\Delta F_B(x_i)$. Hence

$\left|\Delta T_A(x_i)-\Delta T_B(x_i)\right|=0$, $\left|\Delta I_A(x_i)-\Delta I_B(x_i)\right|=0$, $\left|\Delta F_A(x_i)-\Delta F_B(x_i)\right|=0$.

Thus $C_{CHSO}(A, B) = 1$

Conversely,

If $C_{CHSO}(A, B) = 1$, then $\left|\Delta T_A(x_i)-\Delta T_B(x_i)\right|=0$, $\left|\Delta I_A(x_i)-\Delta I_B(x_i)\right|=0$, $\left|\Delta F_A(x_i)-\Delta F_B(x_i)\right|=0$. since cos(0) = 1. So we can write $T_A(x_i)=T_B(x_i)$, $I_A(x_i)=I_B(x_i)$, $F_A(x_i)=F_B(x_i)$

Hence $A = B$.

## 4. Sine Hamming Similarity Measures of RNS

Assume that $A=\langle(\underline{T}_A(x_i),\underline{I}_A(x_i),\underline{F}_A(x_i)),(\overline{T}_A(x_i),\overline{I}_A(x_i),\overline{F}_A(x_i))\rangle$ and $B=\langle(\underline{T}_B(x_i),\underline{I}_B(x_i),\underline{F}_B(x_i)),(\overline{T}_B(x_i),\overline{I}_B(x_i),\overline{F}_B(x_i))\rangle$ in $X$ = {$x_1$, $x_2$, …, $x_n$} be any two rough neutrosophic sets. A sine Hamming similarity operator between two rough neutrosophic sets $A$ and $B$ is defined as follows:

$S_{CHSO}(A, B)$=

$$1-\left[\frac{1}{n}\sum_{i=1}^{n}\sin\left(\frac{\pi}{6}\left(|\Delta T_A(x_i)-\Delta T_B(x_i)|+|\Delta I_A(x_i)-\Delta I_B(x_i)|+|\Delta F_A(x_i)-\Delta F_B(x_i)|\right)\right)\right] \quad (4)$$

Also, [ $\Delta T_A(x)$, $\Delta I_A(x)$, $\Delta F_A(x)$ ] $\neq$ [0, 0, 0] and [ $\Delta T_B(x)$, $\Delta I_B(x)$, $\Delta F_B(x)$ ] $\neq$ [0, 0, 0], $i$ = 1, 2, …, $n$.

**Proposition 4.1**

The defined rough neutrosophic sine Hamming similarity operator $S_{CHSO}(A, B)$ between RNSs $A$ and $B$ satisfies the properties 4, 5, 6 as follows.

1. $0 \leq S_{CHSO}(A, B) \leq 1$
1. $S_{CHSO}(A, B) = 1$ if and only if $A = B$
2. $S_{CHSO}(A, B) = S_{CHSO}(B, A)$

Proof of the property 1.

Since the functions $\Delta T_A(x)$, $\Delta I_A(x)$, $\Delta F_A(x)$, $\Delta T_B(x)$, $\Delta I_B(x)$, and $\Delta F_B(x)$, and the value of the sine function are within [0 ,1], the similarity measure based on rough neutrosophic sine hamming similarity function also lies within [ 0,1].

Hence $0 \leq S_{CHSO}(A, B) \leq 1$.

Proof of the property 2.

For any two RNSs $A$ and $B$ if $A$ = $B$, then the following relations hold $\Delta T_A(x_i)=\Delta T_B(x_i)$, $\Delta I_A(x_i)=\Delta I_B(x_i)$, $\Delta F_A(x_i)=\Delta F_B(x_i)$. Hence

$|\Delta T_A(x_i)-\Delta T_B(x_i)|=0$, $|\Delta I_A(x_i)-\Delta I_B(x_i)|=0$, $|\Delta F_A(x_i)-\Delta F_B(x_i)|=0$. Thus $S_{CHSO}(A, B) = 1$

Conversely,

If $S_{CHSO}(A, B) = 1$, then $|\Delta T_A(x_i)-\Delta T_B(x_i)|=0$, $|\Delta I_A(x_i)-\Delta I_B(x_i)|=0$, $|\Delta F_A(x_i)-\Delta F_B(x_i)|=0$. since sin(0) = 0. So we can write $T_A(x_i)=T_B(x_i)$, $I_A(x_i)=I_B(x_i)$, $F_A(x_i)=F_B(x_i)$

Hence $A = B$.

Proof of the property 3.

This proof is obvious.

## 5. Cotangent Hamming Similarity Measures of RNS

Assume that $A=\langle(\underline{T}_A(x_i),\underline{I}_A(x_i),\underline{F}_A(x_i)),(\overline{T}_A(x_i),\overline{I}_A(x_i),\overline{F}_A(x_i))\rangle$ and $B=\langle(\underline{T}_B(x_i),\underline{I}_B(x_i),\underline{F}_B(x_i)),(\overline{T}_B(x_i),\overline{I}_B(x_i),\overline{F}_B(x_i))\rangle$ in $X = \{x_1, x_2, \ldots, x_n\}$ be any two rough neutrosophic sets. A cotangent Hamming similarity operator between two rough neutrosophic sets *A* and *B* can be defined as follows:

$$COT_{CHSO}(A, B)= \frac{1}{n}\sum_{i=1}^{n}\cot\left(\frac{\pi}{4}+\frac{\pi}{12}\left(|\Delta T_A(x_i)-\Delta T_B(x_i)|+|\Delta I_A(x_i)-\Delta I_B(x_i)|+|\Delta F_A(x_i)-\Delta F_B(x_i)|\right)\right)$$

(5)

Also, $[\Delta T_A(x), \Delta I_A(x), \Delta F_A(x)]\neq[0, 0, 0]$ and $[\Delta T_B(x), \Delta I_B(x), \Delta F_B(x)]\neq[0, 0, 0]$, i = 1, 2, …, n.

**Proposition 5.1**

The defined rough neutrosophic cotangent Hamming similarity operator $COT_{CHSO}(A, B)$ between RNSs A and B satisfies the properties 7, 8, 9.

1. $0\leq COT_{CHSO}(A, B)\leq 1$
2. $COT_{CHSO}(A, B) = 1$ if and only if $A = B$
3. $COT_{CHSO}(A, B) = COT_{CHSO}(B, A)$

**Proof of the property 1:**

Proof: Since the functions $\Delta T_A(x)$, $\Delta I_A(x)$, $\Delta F_A(x)$, $\Delta T_B(x)$, $\Delta I_B(x)$, and $\Delta F_B(x)$, and the value of the cotangent function are within [0 ,1], the similarity measure based on rough neutrosophic cotangentHamming similarity function also lies within [ 0,1].

Hence $0\leq COT_{CHSO}(A, B)\leq 1$

**Proof of the property 2:**

For any two RNSs *A* and *B* if *A* = *B*, we have

$\Delta T_A(x_i)=\Delta T_B(x_i)$, $\Delta I_A(x_i)=\Delta I_B(x_i)$, $\Delta F_A(x_i)=\Delta F_B(x_i)$.

Hence

$|\Delta T_A(x_i)-\Delta T_B(x_i)|=0$, $|\Delta I_A(x_i)-\Delta I_B(x_i)|=0$, $|\Delta F_A(x_i)-\Delta F_B(x_i)|=0$. Thus $COT_{CHSO}(A, B) = 1$

Conversely,

If $COT_{CHSO}(A, B) = 1$, then $|\Delta T_A(x_i)-\Delta T_B(x_i)|=0$, $|\Delta I_A(x_i)-\Delta I_B(x_i)|=0$, $|\Delta F_A(x_i)-\Delta F_B(x_i)|=0$.

Since $\cot(\frac{\pi}{4}) = 1$, we can write $T_A(x_i)=T_B(x_i)$, $I_A(x_i)=I_B(x_i)$, $F_A(x_i)=F_B(x_i)$

Hence A = B.

**Proof of the property 3:**

This proof is obvious.

## 6. Decision making under trigonometric rough neutrosophic Hamming similarity measures

In this section, we apply rough cosine, sine and cotangent Hamming similarity measures between RNSs to the multi-criteria decision making problem. Assume that $S = \{S_1, S_2, \dots, S_m\}$ be a set of alternatives and $A = \{A_1, A_2, \dots, A_n\}$ be a set of attributes.

The proposed decision making method is described using the following steps.

**Step 1**: **Construction of the decision matrix with rough neutrosophic number**

Decision maker considers the decision matrix with respect to m alternatives and n attributes in terms of rough neutrosophic numbers as follows.

**Table1:** *Rough neutrosophic decision matrix*

$$D = \langle \underline{d}_{ij}, \overline{d}_{ij} \rangle_{m\times n} = \begin{array}{c|cccc} & A_1 & A_2 & \cdots & A_n \\ \hline S_1 & \langle \underline{d}_{11}, \overline{d}_{11} \rangle & \langle \underline{d}_{12}, \overline{d}_{12} \rangle & \dots & \langle \underline{d}_{1n}, \overline{d}_{1n} \rangle \\ S_2 & \langle \underline{d}_{21}, \overline{d}_{21} \rangle & \langle \underline{d}_{22}, \overline{d}_{22} \rangle & \dots & \langle \underline{d}_{2n}, \overline{d}_{2n} \rangle \\ . & \dots & \dots & \dots & \dots \\ . & \dots & \dots & \dots & \dots \\ S_m & \langle \underline{d}_{m1}, \overline{d}_{m1} \rangle & \langle \underline{d}_{m2}, \overline{d}_{m2} \rangle & \dots & \langle \underline{d}_{mn}, \overline{d}_{mn} \rangle \end{array} \quad (6)$$

Here $\langle \underline{d}_{ij}, \overline{d}_{ij} \rangle$ is the rough neutrosophic number according to the $i$-th alternative and the $j$-th attribute.

**Step 2**: **Determination of the weights of attribute**

Assume that the weight of the attributes $A_j$ ($j = 1, 2, \dots, n$) considered by the decision-maker be $w_j$ (($j = 1, 2, \dots, n$)) such that $\forall w_j \in [0, 1]$ ($j = 1, 2, \dots, n$) and $\sum_{j=1}^{n} w_j = 1$.

**Step 3**: **Determination of the benefit type attribute and cost type attribute**

Generally, the evaluation attribute can be categorized into two types: benefit type attribute and cost type attribute. Let $K$ be a set of benefit type attributes and $M$ be a set of cost type attributes. In the proposed decision-making method, an ideal alternative can be identified by using a maximum operator for the benefit type attribute and a minimum operator for the cost type attribute to determine the best value of each criterion among all alternatives. We define an ideal alternative S* as follows:

S* = $\{S_1^*, S_2^*, \dots, S_m^*\}$, where benefit attribute is presented as

$$S_j^* = \left[\max_i T_{A_j}^{(S_i)}, \min_i I_{A_j}^{(S_i)}, \min_i F_{A_j}^{(S_i)}\right]$$

and cost type attribute is presented as

$$S_j^* = \left[\min_i T_{A_j}^{(S_i)}, \max_i I_{A_j}^{(S_i)}, \max_i F_{A_j}^{(S_i)}\right].$$

**Step 4**: **Determination of the overall weighted rough trigonometric neutrosophic Hamming similarity function (WRTNHSF) of the alternatives**

We define weighted rough trigonometric neutrosophic similarity function as follows.

$$C_{WCHSO}(A, B) = \sum_{j=1}^{n} w_j C_{CHSO}(A, B) \quad (7)$$

$$S_{WCHSO}(A, B) = \sum_{j=1}^{n} w_j S_{CHSO}(A, B) \quad (8)$$

$$COT_{WCHSO}(A, B) = \sum_{j=1}^{n} w_j COT_{CHSO}(A, B) \quad (9)$$

where $\sum_{j=1}^{n} w_j = 1$ , $j = 1, 2, \ldots, n$.

**Step 5**: **Ranking the alternatives**

Using the weighted rough trigonometric neutrosophic similarity measure between each alternative and the ideal alternative, the ranking order of all alternatives can be determined and the best alternative can be selected with the highest similarity value.

**Step 6:** End

## 7. Numerical Example

Assume that a decision maker intends to select the most suitable smart phone for rough use from the three initially chosen smart phones ($S_1$, $S_2$, $S_3$) by considering four attributes namely: features $A_1$, reasonable price $A_2$, customer care $A_3$, risk factor $A_4$. Based on the proposed approach discussed in section 5, the considered problem is solved using the following steps:

**Step 1**: **Construction of the decision matrix with rough neutrosophic numbers**

The decision maker forms a decision matrix with respect to three alternatives and four attributes in terms of rough neutrosophic numbers (see the Table 2).

Table 2. *Decision matrix with rough neutrosophic number*

$d_S = \langle \underline{N}(P), \overline{N}(P) \rangle_{3\times 4} =$ (10)

| | $A_1$ | $A_2$ | $A_3$ | $A_4$ |
|---|---|---|---|---|
| $S_1$ | $\langle (0.6, 0.3, 0.3), (0.8, 0.1, 0.1) \rangle$ | $\langle (0.6, 0.4, 0.4), (0.8, 0.2, 0.2) \rangle$ | $\langle (0.6, 0.4, 0.4), (0.8, 0.2, 0.2) \rangle$ | $\langle (0.7, 0.4, 0.4), (0.9, 0.2, 0.2) \rangle$ |
| $S_2$ | $\langle (0.7, 0.3, 0.3), (0.9, 0.1, 0.3) \rangle$ | $\langle (0.6, 0.3, 0.3), (0.8, 0.3, 0.3) \rangle$ | $\langle (0.6, 0.2, 0.2), (0.8, 0.4, 0.2) \rangle$ | $\langle (0.7, 0.3, 0.3), (0.9, 0.3, 0.3) \rangle$ |
| $S_3$ | $\langle (0.6, 0.2, 0.2), (0.8, 0.0, 0.2) \rangle$ | $\langle (0.7, 0.3, 0.3), (0.9, 0.1, 0.1) \rangle$ | $\langle (0.7, 0.4, 0.6), (0.9, 0.2, 0.4) \rangle$ | $\langle (0.6, 0.3, 0.2), (0.8, 0.1, 0.2) \rangle$ |

**Step 2: Determination of the weights of the attributes**

The weight vectors considered by the decision maker are 0.32, 0.28, 0.28 and 0.12 respectively.

**Step 3: Determination of the benefit attribute and cost attribute**

Here three benefit types attributes $A_1$, $A_2$, $A_3$ and one cost type attribute $A_4$.

$S^* = [(0.8, 0.1, 0.2), (0.8, 0.2, 0.2), (0.8, 0.3, 0.3), (0.0.7, 0.3, 0.3)]$

**Step 4: Determination of the overall weighted rough trigonometric neutrosophic Hamming similarity function (WRHNHSF) of the alternatives**

We calculate weighted rough trigonometric neutrosophic Hamming similarity values as follows.

$C_{WCHSO}(S_1, S^*) = 0.99554$, $C_{WCHSO}(S_2, S^*) = 0.99253$, $C_{WCHSO}(S_3, S^*) = 0.99799$

$S_{WCHSO}(S_1, S^*) = 0.89455$, $S_{WCHSO}(S_2, S^*) = 0.89233$, $S_{WCHSO}(S_3, S^*) = 0.91729$

$COT_{WCHSO}(S_1, S^*) = 0.92114$, $COT_{WCHSO}(S_2, S^*) = 0.90322$, $COT_{WCHSO}(S_3, S^*) = 0.93009$

**Step 5: Ranking the alternatives**

Ranking the alternatives is prepared based on the descending order of similarity measures. Highest value reflects the best alternative.

Here,

$C_{WCHSO}(S_3, S^*) \succ C_{WCHSO}(S_1, S^*) \succ C_{WCHSO}(S_2, S^*)$

$S_{WCHSO}(S_3, S^*) \succ S_{WCHSO}(S_1, S^*) \succ S_{WCHSO}(S_2, S^*)$

$COT_{WCHSO}(S_3, S^*) \succ COT_{WCHSO}(S_1, S^*) \succ COT_{WCHSO}(S_2, S^*)$

Hence, the smartphone $S_3$ is the best alternative for rough use.

**Step 6:** End

**7.1 Comparison**

All the three similarity measures provided the same ranking order.

## 8. Conclusion

In this paper, we propose rough trigonometric Hamming similarity measures based multi-attribute decision making of rough neutrosophic environment and prove some of their basic properties. We provide an application, namely selection of the most suitable smart phone for rough use. We also present comparison with the three rough neutrosophic similarity measures. The concept presented in this paper can be applied other multiple attribute decision making problems in rough neutrosophic environment.

## References

1. L. A. Zadeh, Fuzzy sets, Information and Control, 8(3) (1965), 338–353.
2. K. Atanassov, Intuitionistic fuzzy sets. Fuzzy Sets Systems, 20(1986), 87-96.
3. F. Smarandache, A unifying field in logics, neutrosophy: neutrosophic probability, set and logic, American Research Press, Rehoboth, 1998.
4. F. Smarandache, Neutrosophic set–a generalization of intuitionistic fuzzy sets, International Journal of Pure and Applied Mathematics, 24(3) (2005), 287-297.
5. H. Wang, F. Smarandache, Y. Q. Zhang, R. Sunderraman, Single valued neutrosophic sets, Multispace and Multistructure, 4 (2010), 410–413.
6. H. Wang, F. Smarandache, Y. Q. Zhang, R. Sunderraman, Interval neutrosophic sets and logic: theory and applications in computing, Hexis, Phoenix, AZ, 2005.
7. S. Broumi, F. Smarandache, M. Dhar, Rough neutrosophic sets, Italian Journal of Pure and Applied Mathematics, 32 (2014), 493-502.
8. S. Broumi, F. Smarandache, M. Dhar, Rough neutrosophic sets, Neutrosophic Sets and Systems, 3 (2014), 60-66.
9. Z. Pawlak, Rough sets, International Journal of Information and Computer Sciences, 11(5) (1982), 341-356.
10. D. Dubios, H. Prade, Rough fuzzy sets and fuzzy rough sets, International Journal of General System, 17(1990), 191-208.
11. S. Nanda, S. Majumdar. Fuzzy rough sets, Fuzzy Sets and Systems, 45 (1992), 157–160.
12. K. V. Thomas, L. S. Nair, Rough intuitionistic fuzzy sets in a lattice, International Mathematics Forum, 6(27) (2011),1327–1335.
13. K. Mondal, S. Pramanik, Rough neutrosophic multi-attribute decision-making based on grey relational analysis, Neutrosophic Sets and Systems, 7(2015), 8-17.
14. S. Pramanik, K. Mondal, Cosine similarity measure of rough neutrosophic sets and its application in medical diagnosis, Global Journal of Advanced Research, 2(1)(2015), 212-220.
15. K. Mondal, S. Pramanik, Rough neutrosophic multi-attribute decision-making based on rough accuracy score function, Neutrosophic Sets and Systems, 8(2015), 16-22.
16. S. Pramanik, K. Mondal, Cotangent similarity measure of rough neutrosophic sets and its application to medical diagnosis, Journal of New Theory, 4(2015), 90-102.
17. S. Pramanik, K. Mondal, Some rough neutrosophic similarity measure and their application to multi attribute decision making, Global Journal of Engineering Science and Research Management, 2(7)(2015), 61-74.
18. K. Mondal, S. Pramanik, F. Smarandache. Multi-attribute decision making based on rough neutrosophic variational coefficient similarity measure, Neutrosophic Sets and Systems 13 (2016) (In press).
19. K. Mondal, S. Pramanik, F. Smarandache. Rough neutrosophic TOPSIS for multi-attribute group decision making, Neutrosophic Sets and Systems13 (2016) (In press).

20. K. Mondal, S. Pramanik. (2015). Tri-complex rough neutrosophic similarity measure and its application in multi-attribute decision making. Critical Review, 11 (2015), 26-40.
21. K. Mondal, S. Pramanik, F. Smarandache. Rough neutrosophic hyper-complex set and its application to multi-attribute decision making. Critical Review 13 (2016) (In press).

NITAL P. NIRMAL[1*], MANGAL G. BHATT[2]

1Production Engineering Department, Shantilal Shah Govt. Engineering College, Bhavnagar, Gujarat India- 364060. Email: nirmal_nital@gtu.edu.in
2Shantilal Shah Govt. Engineering College, Bhavnagar, Gujarat, India- 364060.
Email: mangalbhatt15@gmail.com

# Selection of Automated Guided Vehicle using Single Valued Neutrosophic Entropy Based Novel Multi Attribute Decision Making Technique

## Abstract

Selection of material handling equipment for typical conditions and handling environment is one of the multi attribute decision making problem. The objective of the research paper is to implement and validate multi attribute selection of automated guided vehicle for material handling purpose. The present paper proposes a single valued neutrosophic set with entropy weight based multi attribute decision making technique. A proposed technique also works with more uncertainty, imprecise, indeterminate and inconsistent information. The proposed methodology follows with the example for selection and ranking of automated guided vehicle and in validation and sensitivity analysis of the novel multi attribute decision making technique carried out. The result of the study builds assurance in suitability of single valued neutrosophic set entropy based novel multi attribute decision making for selection of automated guided vehicle alternatives.

## 1. Introduction

Material Handling Equipment (MHE) is playing a vital role in today's manufacturing system and also improving productivity in the small, medium or large scale manufacturing industries. MHE is a very essential task for the manufacturing sectors because of the considerable capital investment required(Onut, Kara, & Mert, 2009). Saputro et al. (2015)) reviewed 42 papers for MHE selection and established ranking to appropriate MHE for complex selection problems. Right MHE selection and good design of the MHE can increase productivity and reduce investment and operation's costs.

Karande & Chakraborty (2013) investigated the various functions performed by MHE are as follows:

a. Transportation and logistics (for moving material form one point to another, i.e. conveyors, cranes, industrial trucks, etc.)
b. Positioning (for aid machining operation like, robots, index tables, rotary tables, etc.)

c. Unit formation (for holding or carrying purpose pallets, skids, containers, bins, etc.)
d. Storage (For store/ inventory automatic storage and retrieval system (AS/RS), pallets, etc.).

On the real difficulties in developing and using selection methods is due to the natural vagueness associated with the inputs to the model (Deb, Bhattacharyya, & Sorkhel, 2002).

## 2. Literature Survey

Literature survey is carried out with two elements one for selection methodology for MHE using multi attribute decision making (MADM) and the other for literature on single valued neutrosophic set theory.

### 1.1 Literature survey of selection methodology used for selection of MHE using MADM techniques

Since last three decades researchers pay more attention in finding and implementing different MADM techniques with different criteria (attributes). Onut et al. ( 2009) implemented fuzzy Analytic Network Process (ANP) for assigning weights to the attributes for MHE selection and fuzzy Technique for Order Preference by Similarity to Ideal Solution (TOPSIS) is used to ranking solution. Maniya & Bhatt (2011) implemented and validated modified grey relational analysis (M-GRA) method combined with AHP for multi attribute selection of Automated Guided Vehicle (AGV) for the material handling. Sawant et al.(2011) worked on Preference Selection Index (PSI) method for AGV selection in manufacturing environment. Chakraborty & Banik (2006) worked on Analytic Hierarchy Process (AHP) for material handling equipment selection model. Kulak (2005) investigated a fuzzy multi attribute selection of material handling equipment which consist of a database, rule based system and multi attribute fuzzy information axiom approach for selecting MHE. Nguyen et al. (2016) worked for fuzzy AHP and fuzzy additive ratio assessment for conveyor evaluation ranking and selection process. Mirhosseyni & Webb, (2009) presents fuzzy knowledge based expert system and then genetic algorithm (GA) for efficient selection and assignment of MHE. Eko Saputro & Daneshvar Rouyendegh (2016) investigated a hybrid approach for selecting MHE in a ware house by using entropy based hierarchical fuzzy TOPSIS and Multi Objective Mixed Integer Linear Programming (MOMILP) for ranking and selecting best alternatives. Anand et al. (2011) investigated MHE selection with ANP for complex decision making problem. Biswas et al. (2016) proposed TOPSIS approach to SVNS and applied the approach for multi attribute group decision making problem.

### 1.2 Literature survey of single valued neutrosophic set

In classical MADM approach, input variables are crisp sets but in the real world decision problem input variables are expressed in terms of qualitative information. Qualitative information provided by decision makers (DMs)/experts can be easily expressed by linguistic variables.

Sometimes due to lack of time-pressure, limited knowledge about public domain, decision maker may prefer linguistic variables (Zadeh (1975)) to deal with imprecise data. To cover up the limitation of fuzzy set, Atanassov (1986) proposed the Intuitionistic Fuzzy Set (IFS) by adding truth membership Ta (x) and falsity Membership Fa (x). Further Atanassov (1986) proposed the Interval Valued Intuitionistic Fuzzy Set (IVIFS). However, drawback of IFS and IVIFS is that they cannot handle indeterminate and inconsistent information. In real application, information of input data is often incomplete, indeterminate and inconsistent ( Chi & Liu (2013)). The limitation of above sets is covered up with Neutrosophic Set (NS) (Smarandache (2002)) with degree of truth, indeterminacy and falsity, where all membership function is completely independent. Single Valued Neutrosophic Set (SVNS) is an instance of NS, which can handle uncertainty, imprecise, indeterminate and inconsistent information (Wang et al. (2010). Majumdar (2015) established uncertain data processing with NS and further generalized and combined with soft sets in decision making process. Ye (2013) worked on correlation and correlation coefficient of SVNS based on the extension of the correlation of IFS. Ye (2014a) worked on single valued neutrosophic cross-entropy for MADM techniques. Zhang et al. (2014) applied interval neutrosophic set applied to multi criteria decision making for investment problem. Biswas et al. (2014) presented neutrosophic MADM with unknown weight information methodology. Pramanik et al. (2015) presented hybrid vector similarity measures and their applications to multi-attribute decision making under neutrosophic environment. Ye (2014b) worked on vector similarity measures of simplified NS with investigating money case study.

## 3. SVNS Entropy based MADM Methodology

Steps of SNVS entropy based novel MADM as follows.

**Step 1:** Define the goal of MADM problem such as ranking/ evaluation/ sorting/ selection of various alternatives involved in decision making procedure.

**Step 2:** Identify the possible alternative with attributes (criteria's).

**Step 3:** Prepare the decision matrix.

Let, $A = \{A_i,\ for\ i = 1, 2, 3, \ldots\ldots..m\}$ be a set of alternative while, $C = \{C_i,\ for\ j = 1, 2, 3, \ldots\ldots..n\}$ be a set of attributes (criteria). The different values of criteria's may be quantitative and/or qualitative in nature.

**Step 4:** Convert qualitative information into fuzzy numbers. Normalization of matrix is shown in Table 1.

Table 1: *Matrix Normalization Techniques*

| Name of Normalization Methods | Normalized Value | |
|---|---|---|
| | Benefit Values | Non- Beneficial Values |
| Linear Scale Transformation, Max Method (LSTMM) | $R_{ij} = \frac{X_{ij}}{X_{iMax}}$ | $R_{ij} = \frac{X_{iMin}}{X_{ij}}$ |
| Linear Scale Transformation, Max-Min Method (LSTMMM) | $R_{ij} = \frac{X_{ij} - MinX_{ij}}{MaxX_{ij} - MinX_{ij}}$ | $R_{ij} = \frac{MaxX_{ij} - X_{ij}}{MaxX_{ij} - MinX_{ij}}$ |
| Linear Scale Transformation Sum Method (LSTSM) | $R_{ij} = \frac{X_{ij}}{\sum_{i=1}^{m} X_i}$ | $R_{ij} = 1 - \frac{X_{ij}}{\sum_{i=1}^{m} X_i}$ |
| Vector Normalization Method (VNM) | $R_{ij} = \frac{X_{ij}}{\sqrt{\sum_{i=1}^{m} X_{ij}^{2}}}$ | $R_{ij} = 1 - \frac{X_{ij}}{\sqrt{\sum_{i=1}^{m} X_{ij}^{2}}}$ |

**Step 5:** Conversion classic set/ fuzzy set to SVNS

To validate proposed MADM method with other MADM techniques, we propose the conversion rule to use the input matrix in classic or fuzzy set to SVNS for beneficial and non-beneficial criteria.

(i) **Beneficial criteria:** (higher value of performance measures of selection criteria is desirable i.e., profit, quality, etc.): Considering positive ideal solution (PIS) as $< T_{\max}{}^{*}(x), I_{\min}{}^{*}(x), F_{\min}{}^{*}(x) >$; normalized input matrix beneficial criteria are considered as degree of truthness $T_{A}(x)$, while degree of indeterminacy and degree of falsehood as $I_{A}(x) = F_{A}(x) = 1 - T_{A}(x)$ respectively.

(ii) **Non beneficial criteria:** (Lower value of performance measure of selection criteria is desirable i.e. cost): Considering negative ideal solution (NIS) as $< T_{\min}{}^{*}(x), I_{\max}{}^{*}(x), F_{\max}{}^{*}(x) >$; normalized input matrix non beneficial criteria are considered as degree of indeterminacy and falsehood as $I_{A}(x) = F_{A}(x)$ while degrees of truthness is considered as $T_{A}(x) = 1 - I_{A}(x) = 1 - F_{A}(x)$.

(iii) Find the entropy value for attribute with equation no (1).

$$E_j = 1 - {}^{1}\!/_{n} \sum_{i=1}^{m} (T_{ij}(x_i) + F_{ij}(x_i)) \left|2(I_{ij}(x_i)) - 1\right| \quad (1)$$

**Step 6:** We find entropy weight for attribute using the method proposed by Wang and Zhang (2009).

$$W_j = \frac{1-E_j}{\sum_{j=1}^{n}(1-E_j)} \quad (2)$$

We get weight vector $W = (w_1, w_{2,} w_{3,\ldots\ldots\ldots\ldots\ldots} w_n)^T$ of attributes,

$C = \{C_j for j = 1, 2, 3, \ldots \ldots .. n\}$ with $W_j \geq 0$ and $\sum_{j=1}^{n} W_j = 1$.

**Step 7:** Calculate the alternative value with following equation (3).

$$A_w = \sum_{j=1}^{n} W_j * \left((T_{ij}(x) * T_{ij}{}^{*}(x)) + (I_{ij}(x) * I_{ij}{}^{*}(x) + (F_{ij}(x) * F_{ij}{}^{*}(x))\right) \quad (3)$$

Where, beneficial attribute PIS=$< T_{\max}{}^{*}(x) I_{\min}{}^{*}(x), F_{\min}{}^{*}(x) > = < 1, 0, 0 >$,

For non-beneficial attribute NIS=$< T_{\min}{}^{*}(x) I_{\max}{}^{*}(x), F_{\max}{}^{*}(x) > = < 0, 1, 1 >$.

**Step 8:** Ranking of alternatives after calculation is performed according to ascending order.

## 4. Case Study

An example is considered to show and validate the SVNS entropy based novel MADM method for selection of an AGV for an industrial application. The detailed rationalization of steps involved in the application of novel MADM for section of AGV is explained below.

**Step 1:** The objective is to ranking and selection of the best AGV for a given industrial application.

**Step 2:** In the present work eight alternatives of AGV and six attributes (Criteria) are considered, the same as (K. D. Maniya & Bhatt, 2011). Criteria are: controllability (C1), accuracy (C2), cost (C3), range (C4), reliability (C5) and flexibility (C6). Here cost (C3) from the given attributes is given as non-beneficial attribute indicate with (-) sign in decision matrix; while other attributes are the beneficial attribute indicate with (+) sign in decision matrix.

**Step 3:** Here, in the AGV alternative and attributes and their values are presented in matrix format. The crisp data for AGV selection adopted from (K. D. Maniya & Bhatt, 2011) is shown in Table 2.

**Table 2:** *The crisp data for AGV selection attributes (decision matrix)[adopted from (K. D. Maniya & Bhatt, 2011)]*

| | C1 (+) | C2 (+) | C3 (-) | C4 (+) | C5 (+) | C6 (+) |
|---|---|---|---|---|---|---|
| A1 | 0.895 | 0.495 | 0.695 | 0.495 | 0.895 | 0.295 |
| A2 | 0.115 | 0.895 | 0.895 | 0.895 | 0.495 | 0.495 |
| A3 | 0.115 | 0.115 | 0.895 | 0.115 | 0.695 | 0.895 |
| A4 | 0.295 | 0.895 | 0.115 | 0.495 | 0.495 | 0.895 |
| A5 | 0.895 | 0.495 | 0.115 | 0.695 | 0.295 | 0.495 |
| A6 | 0.495 | 0.495 | 0.895 | 0.115 | 0.695 | 0.695 |
| A7 | 0.115 | 0.295 | 0.895 | 0.115 | 0.895 | 0.895 |
| A8 | 0.115 | 0.495 | 0.695 | 0.495 | 0.495 | 0.695 |

**Step 4:** Normalization of decision matrix

In the proposed case study we use Linear Scale Transformation, Max Method (LSTMM) as shown in Table 3.

**Table 3:** *Normalized decision matrix with (linear scale transformation, max method)*

| | C1 (+) | C2 (+) | C3 (-) | C4 (+) | C5 (+) | C6 (+) |
|---|---|---|---|---|---|---|
| A1 | 1.0000 | 0.5531 | 0.1655 | 0.5531 | 1.0000 | 0.3296 |
| A2 | 0.1285 | 1.0000 | 0.1285 | 1.0000 | 0.5531 | 0.5531 |
| A3 | 0.1285 | 0.1285 | 0.1285 | 0.1285 | 0.7765 | 1.0000 |
| A4 | 0.3296 | 1.0000 | 1.0000 | 0.5531 | 0.5531 | 1.0000 |
| A5 | 1.0000 | 0.5531 | 1.0000 | 0.7765 | 0.3296 | 0.5531 |
| A6 | 0.5531 | 0.5531 | 0.1285 | 0.1285 | 0.7765 | 0.7765 |
| A7 | 0.1285 | 0.3296 | 0.1285 | 0.1285 | 1.0000 | 1.0000 |
| A8 | 0.1285 | 0.5531 | 0.1655 | 0.5531 | 0.5531 | 0.7765 |

**Step 5:** Convert crisp normalized matrix into SVNS decision matrix with $< T_{ij}(x), I_{ij}(x), F_{ij}(x) >$ degree of truthness, indeterminate and falsehood. As shown in Table 4.

**Step 6:** Find the entropy weight using equation (2) with $W_j \geq 0$ and $\sum_{j=1}^{n} W_j = 1$ as shown in Table 4.

**Step 7:** Calculate the alternative value with following equation (3) as shown in Table 4.

**Step 8:** Ranking or selections of alternative: The alternatives are ranked according to ascending order as shown in Table 4.

**Table 4:** *SVNS entropy based decision matrix*

| | C1 (+) | C2 (+) | C3 (-) | C4 (+) | C5 (+) | C6 (+) | $A_w$ | Rank |
|---|---|---|---|---|---|---|---|---|
| A1 | <1.0000, 0.0000, 0.0000> | <0.5531, 0.4469, 0.4469> | <0.8345, 0.1655, 0.1655> | <0.5531, 0.4469, 0.4469> | <1.0000, 0.0000, 0.0000> | <0.3296, 0.6704, 0.6704> | 0.6235 | 3 |
| A2 | <0.1285, 0.8715, 0.8715 > | <1.0000, 0.0000, 0.0000> | <0.8715, 0.1285, 0.1285> | <1.0000, 0.0000, 0.0000> | <0.5531, 0.4469, 0.4469> | <0.5531, 0.4469, 0.4469> | 0.5354 | 4 |
| A3 | <0.1285, 0.8715, 0.8715> | <0.1285, 0.8715, 0.8715> | <0.8715, 0.1285, 0.1285> | <0.1285, 0.8715, 0.8715> | <0.7765, 0.2235, 0.2235> | <1.0000, 0.0000, 0.0000> | 0.3969 | 8 |
| A4 | <0.3296, 0.6704, 0.6704> | <1.0000, 0.0000, 0.0000> | <0.0000, 1.0000, 1.0000> | <0.5531, 0.4469, 0.4469> | <0.5531, 0.4469, 0.4469> | <1.0000, 0.0000, 0.0000> | 0.9313 | 2 |
| A5 | <1.0000, 0.0000, 0.0000> | <0.5531, 0.4469, 0.4469> | <0.0000, 1.0000, 1.0000> | <0.7765, 0.2235, 0.2235> | <0.3296, 0.6704, 0.6704> | <0.5531, 0.4469, 0.4469> | 0.9335 | 1 |
| A6 | <0.5531, 0.4469,0.4469> | <0.5531, 0.4469, 0.4469> | <0.8715, 0.1285, 0.1285> | <0.1285, 0.8715, 0.8715> | <0.7765, 0.2235, 0.2235> | <0.7765, 0.2235, 0.2235> | 0.4996 | 5 |
| A7 | <0.1285, 0.8715, 0.8715> | <0.3296, 0.6704, 0.6704> | <0.8715, 0.1285, 0.1285> | <0.1285, 0.8715, 0.8715> | <1.0000, 0.0000, 0.0000> | <1.0000, 0.0000, 0.0000> | 0.4547 | 7 |
| A8 | <0.1285, 0.8715, 0.8715 > | <0.5531, 0.4469, 0.4469> | <0.8345, 0.1655, 0.1655> | <0.5531, 0.4469, 0.4469> | <0.5531, 0.4469, 0.4469> | <0.7765, 0.2235, 0.2235> | 0.4619 | 6 |
| $A^*$ | <1,0, 0> | <1,0,0 > | <0,1,1 > | <1,0,0 > | <1,0,0 > | <1,0, 0> | | |
| $E_j$ | 0.3226 | 0.5615 | 0.3362 | 0.4874 | 0.5293 | 0.4176 | | |
| $W_j$ | 0.2025 | 0.1311 | 0.1984 | 0.1532 | 0.1407 | 0.1741 | $\sum_{j=1}^{n} W_j=1$ | |

## 5. Validation with Sensitivity Analysis

Sensitivity analysis of proposed methodology regarding ranking of alternatives with various normalization methods for same input data is shown in Table 5.

The graphical representation of sensitivity analysis for mentioned MADM techniques has been shown in the figure 1. It proves that SVNS entropy based novel MADM technique with different normalizing techniques shows negligible effect on final ranking order of AGV as compared to with PSI technique.

**Table: 5** *Ranking comparison with different normalization methods for F-SVNS Novel MADM and PSI MADM Technique.*

| | **F-SVNS Entropy based Novel MADM Technique** | | | | **PSI (K. D. Maniya & Bhatt, 2011)** | | | |
|---|---|---|---|---|---|---|---|---|
| | **LST MM** | **LST MMM** | **LST SM** | **VNM** | **LST MM** | **LST MMM** | **LST SM** | **VNM** |
| **A1** | 3 | 3 | 3 | 3 | 4 | 4 | 4 | 4 |
| **A2** | 4 | 6 | 4 | 4 | 3 | 3 | 5 | 5 |
| **A3** | 8 | 8 | 8 | 8 | 8 | 8 | 8 | 8 |
| **A4** | 2 | 1 | 2 | 2 | 1 | 1 | 3 | 1 |
| **A5** | 1 | 2 | 1 | 1 | 6 | 6 | 1 | 2 |
| **A6** | 5 | 5 | 6 | 5 | 7 | 7 | 2 | 3 |
| **A7** | 7 | 7 | 7 | 7 | 5 | 5 | 6 | 6 |
| **A8** | 6 | 4 | 5 | 6 | 2 | 2 | 7 | 7 |

**Figure 1:** *Graphic representation of Sensitivity Analysis with different Normalization Method F-SVNS Entropy based Novel MADM and PSI Method*

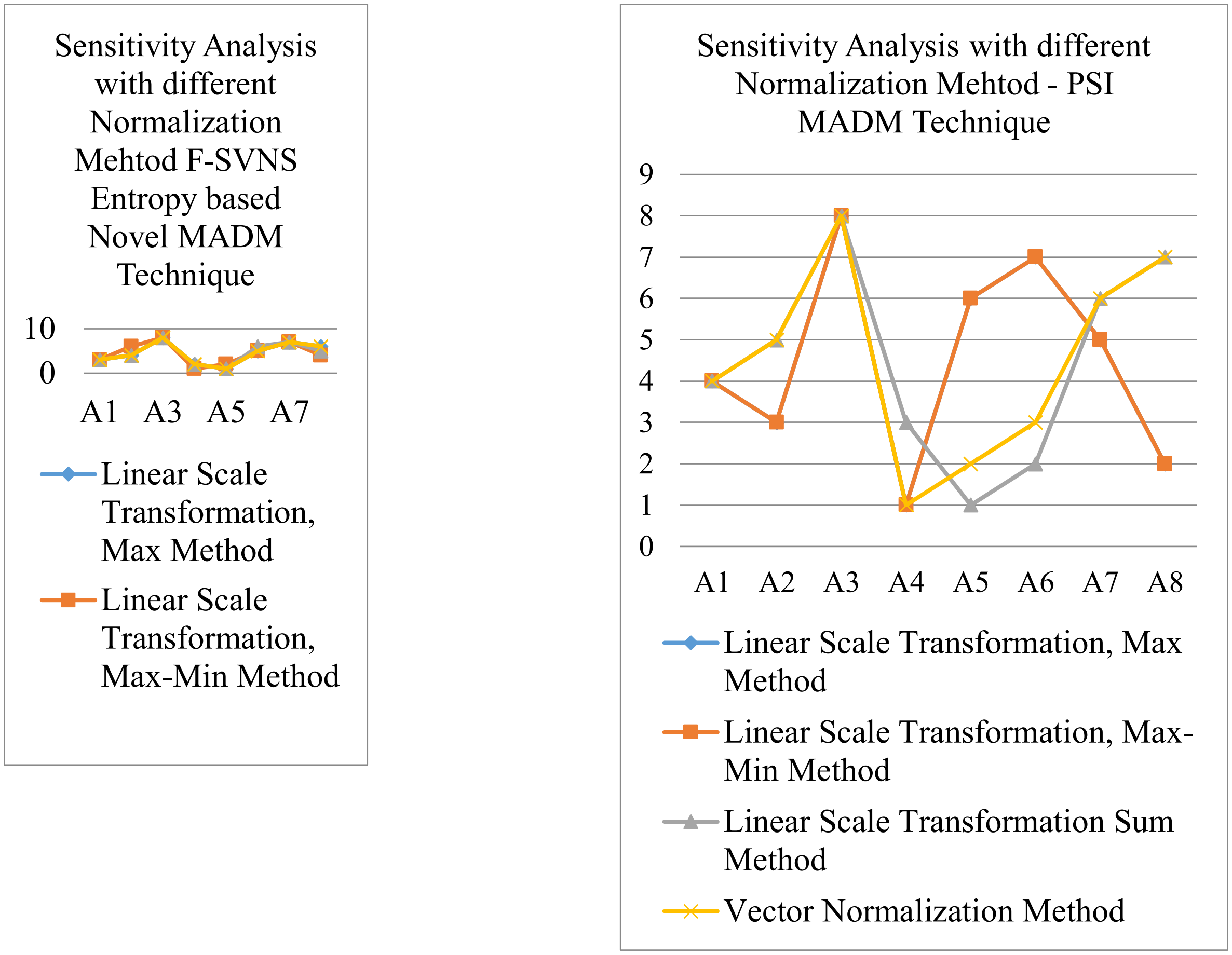

## 6. Result and Discussions

In this paper, SVNS entropy weight MADM technique is developed and implemented to examine its feasibility for selecting and ranking of AGV for material handling system for a given industrial application. The main concluding remarks of proposed technique are listed below:

- The proposed methodology of ranking or selection of alternatives is suitable to decision making under incomplete information, indeterminate and inconsistent information.
- The proposed SVNS entropy weight MADM technique gives more efficient and compromise selection of best alternative.
- During calculation and normalization there is no loss of information; no single attribute has become zero.
- A sensitivity analysis also shows negligible effect on final ranking order and selection of AGV.
- Proposed methodology is capable to converting decision maker’s crisp information or fuzzy information into SVNS form, which makes more efficient ranking solution.

# References

1. Anand, G., Kodali, R., & Kumar, B. S. (2011). Development of analytic network process for the selection of material handling systems in the design of flexible manufacturing systems (FMS). Journal of Advances in Management Research, 8(1), 123-147. doi:10.1108/09727981111129336.
2. Atanassov, K. T. (1986). Intuitionistic fuzzy sets. Fuzzy Sets and Systems, 20(1), 87-96.
3. Biswas, P., Pramanik, S., Giri, B.C. (2016). TOPSIS method for multi-attribute group decision making under single-valued neutrosophic environment. Neural computing and Applications, 27(3) 727-737.
4. Biswas, P., Pramanik, S., & Giri, B. C. (2014). A new methodology for neutrosophic multi-attribute decision making with unknown weight information. Neutrosophic Sets and Systems, 3, 42-52.
5. Chakraborty, S., & Banik, D. (2006). Design of a material handling equipment selection model using analytic hierarchy process. The International Journal of Advanced Manufacturing Technology, 28(11), 1237-1245. doi:10.1007/s00170-004-2467-y.
6. Chi, P., & Liu, P. (2013). An extended TOPSIS method for the multiple attribute decision making problems based on interval neutrosophic set. Neutrosophic Sets and Systems, 1(1), 63-70.
7. Deb, S. K., Bhattacharyya, B., & Sorkhel, S. K. (2002). Material handling equipment selection by fuzzy multi-criteria decision making methods. In N. R. Pal & M. Sugeno (Eds.), Advances in Soft Computing — AFSS 2002: 2002 AFSS International Conference on Fuzzy Systems Calcutta, India, February 3–6, 2002 Proceedings (pp. 99-105). Berlin, Heidelberg: Springer Berlin Heidelberg.
8. Eko Saputro, T., & Daneshvar Rouyendegh, B. (2016). A hybrid approach for selecting material handling equipment in a warehouse. International Journal of Management Science and Engineering Management, 11(1), 34-48.
9. Karande, P., & Chakraborty, S. (2013). Material handling equipment selection using weighted utility additive theory. Journal of Industrial Engineering, 2013, 9. doi:10.1155/2013/268708.
10. Kulak, O. (2005). A decision support system for fuzzy multi-attribute selection of material handling equipments. Expert Systems with Applications, 29(2), 310-319.
11. Majumdar, P. (2015). Neutrosophic sets andits applications to decision making. In D. P. Acharjya, S. Dehuri, & S. Sanyal (Eds.), Computational Intelligence for Big Data Analysis: Frontier Advances and Applications (pp. 97-115). Cham: Springer International Publishing.
12. Maniya, K., & Bhatt, M. G. (2010). A selection of material using a novel type decision-making method: Preference selection index method. Materials & Design, 31(4), 1785-1789.
13. Maniya, K. D., & Bhatt, M. G. (2011). A multi-attribute selection of automated guided vehicle using the AHP/M-GRA technique. International Journal of Production Research, 49(20), 6107-6124.
14. Mirhosseyni, S. H. L., & Webb, P. (2009). A hybrid fuzzy knowledge-based expert system and genetic algorithm for efficient selection and assignment of material handling equipment. Expert Systems with Applications, 36(9), 11875-11887.
15. Nguyen, H.-T., Md Dawal, S. Z., Nukman, Y., P. Rifai, A., & Aoyama, H. (2016). An integrated MCDM model for conveyor equipment evaluation and selection in an FMC based on a fuzzy AHP and fuzzy ARAS in the presence of vagueness. PLoS ONE, 11(4), e0153222. doi:10.1371/journal.pone.0153222.
16. Onut, S., Kara, S. S., & Mert, S. (2009). Selecting the suitable material handling equipment in the presence of vagueness. The International Journal of Advanced Manufacturing Technology, 44(7), 818-828. doi:10.1007/s00170-008-1897-3.
17. Pramanik, S, Biswas, P., & Giri, B.C. (2015). Hybrid vector similarity measures and their applications to multi-attribute decision making under neutrosophic environment. Neural Computing and Applications. DOI 10.1007/s00521-015-2125-3
18. Saputro, T. E., Masudin, I., & Daneshvar Rouyendegh, B. (2015). A literature review on MHE selection problem: levels, contexts, and approaches. International Journal of Production Research, 53(17), 5139-5152.

19. Sawant, V. B., Mohite, S. S., & Patil, R. (2011). A decision-making methodology for automated guided vehicle selection problem using a preference selection index method. In K. Shah, V. R. Lakshmi Gorty, & A. Phirke (Eds.), Technology Systems and Management: First International Conference, ICTSM 2011, Mumbai, India, February 25-27, 2011. Selected Papers (pp. 176-181). Berlin, Heidelberg: Springer Berlin Heidelberg.
20. Smarandache, F. (2002). Neutrosophic set–a generalization of the intuitionistic fuzzy set. Paper presented at the University of New Mexico.
21. Wang, H., Smarandache, F., Zhang, Y., & Sunderraman, R. (2010). Single valued neutrosophic sets. Multispace and Multistructure 4 (2010), 410-413.
22. Wang, J., & Zhang, Z. (2009). Multi-criteria decision-making method with incomplete certain information based on intuitionistic fuzzy number. Control and Decision, 24(2), 226-230.
23. Ye, J. (2013). Multicriteria decision-making method using the correlation coefficient under single-valued neutrosophic environment. International Journal of General Systems, 42(4), 386-394.
24. Ye, J. (2014a). Single valued neutrosophic cross-entropy for multicriteria decision making problems. Applied Mathematical Modelling, 38(3), 1170-1175.
25. Ye, J. (2014b). Vector similarity measures of simplified neutrosophic sets and their application in multicriteria decision making. International Journal of Fuzzy Systems, 16(2), 204-215.
26. Zadeh, L. A. (1975). The concept of a linguistic variable and its application to approximate reasoning-I. Information Sciences 8, 199-249.
27. Zhang, H.-y., Wang, J.-q., & Chen, X.-h. (2014). Interval neutrosophic sets and their application in multicriteria decision making problems. The Scientific World Journal, 2014. http://dx.doi.org/10.1155/2014/645953.

**Florentin Smarandache**[1], **Mirela Teodorescu**[2]

1 Mathematics & Science Department, University of New Mexico, USA. E-mail: fsmarandache@gmail.com
2 Neutrosophic Science International Association, New Mexico, USA. E-mail : mirela.teodorescu@yahoo.co.uk

# From Linked Data Fuzzy to Neutrosophic Data Set Decision Making in Games vs. Real Life

## Abstract

In our lives, reality becomes a game, and in the same way, the game becomes reality, the game is an exercise, simulation of real life on a smaller scale, then it extends itself into reality. This article aims to make a connection between decision making in game which comprises all the issues that intervene in the process and further making a connection with real life. The method for identification involved, detected or induced uncertainties is a jointing process from linked data fuzzy to neutrosophic data set on a case study, EVE Online game. This analysis is useful for psychologists, sociologists, economic analysis, process management, business area, also for researchers of games domain.

## 1. Introduction

The aim of this study is to offer a method of refining the uncertainties, neutral states appeared in a process being a game reflected in the real life, through neutrosophic theory.

In higher forms concerning us, we can associate the function of play as derived from two basic aspects met by us: “as a contest for something or a representation of something”, as asserts Huizinga (Huizinga,1980, p.13).

The games, in their configuration, structure, follow the rules, procedures, concepts defined by game theory. There are three categories of games: games of skills, games of chance and games of strategy.

Games of chance type face uncertainty and risk in decision making process (Janis, Mann, 1977). These decisions are evaluated, analyzed and taken in accordance with game theory, according to the social system involved.

Neutrosophic theory applied in decision making for solving the uncertainties matches with game theory requirements (Von Neumann, Morgenstern, 1944).

From the multitude of games we chose to study neutrosophic making decisions, for the case of the EVE online, a complex game both as structure and players, involving complex criteria of the decisions making mechanism.

We have to take into consideration that Dr. Eyjólfur Guðmundsson as economist of the game EVE Online, applied the concept of Vernon Smith, Nobel Laureate for experimental economics, asserting: "This would be any economist's dream, because this is not just an experiment, this is more like a simulation. More like a fully-fledged system where you can input to see what happens" (http://www.ibtimes.co.uk/eve-online-meet-man-controlling-18-million-space-economy-1447437).

Our opinion is that the game is a precious source of ideas, energy, adrenaline, a simulator, an exercise for real life that promotes success but also decay through addiction, tolerance and thus it can be treated just like drugs. But we want to discuss only the positive side of the game.

This game covers both linked data and social media practices, in this context, social media representing computer-mediated tools that allow people or entities to create, share, or exchange information, emotions, feelings, ideas, pictures/videos in virtual communities and networks and on the other side, to provide linked data as method of publishing structured data, interlinked and to become more useful through semantic queries. It builds upon standard Web technologies (such as HTTP, RDF and URIs). It extends them to share information in a way that can be read automatically by computers.

## 2. Background

We are surrounded by data characterized by the performance of our activities, the fuel efficiency of our cars, a multitude of products from different vendors, the values of the air parameters, or the way our taxes are spent. It helps us to make better decision; this data is playing an increasingly central role in our lives, driving the emergence of data economy. Increasing numbers of individuals and organizations are contributing to this deluge by choosing to share their data with others. Availability of data is very important in evaluating, analysis, making decision process (Heath, Bizer, 2011).

### 2.1 Fundamentals of Neutrosophic Theory

Uncertainty represents an unsolved situation, it defines a fuzziness state. Uncertainty is an actant's subjective state related to a phenomenon, or decision making, and it becomes objective when it is inserted in a probability calculus system or into an algorithm.

It is mentioned in specialty literature that Zadeh introduced the degree of membership/truth (t), the rest would be (1-t) equal to f, their sum being 1, so he defined the fuzzy set in 1965 (Zadeh, 1965). Further, Atanassov introduced the degree of non- membership /falsehood (f) and he defined the intuitionistic fuzzy set (Atanassov, 1986), asserting: if $0 <= t + f <= 1$ and $0 <= 1 - t - f$, it would be interpreted as indeterminacy $t + f <= 1$. In this case, the indeterminacy state, as proposition, cannot be described in fuzzy logic, is missing the uncertainty state; the neutrosophic logic helps to make a distinction between a "relative truth" and an "absolute truth", while fuzzy logic does not. As novelty to previous theory, Smarandache introduced and defined explicitly the degree of indeterminacy/ neutrality (i) as independent component $^{-}0 <= t + i + f <= 3^{+}$ . In neutrosophy set, the three components t, i, f are independent because it is possible from a source to get (t), from another independent source to get (i) and from the third source to get (f). Smarandache goes further; he refined the range (Smarandache, 2005).

**Neutrosophic Set**: Let U be a universe of discourse, and M a set included in U. An element x from U is noted with respect to the set M as x(T, I, F) and belongs to M in the following way: it is t% true in the set, i% indeterminate (unknown if it is) in the set, and f% false, where t varies in T, i varies in I, f varies in F (Smarandache, 2005).

Statically T, I, F are subsets, but dynamically T, I, F are functions/operators depending on many known or unknown parameters. Neutrosophic set generalizes the fuzzy set (especially intuitionistic fuzzy set), paraconsistent set, intuitionistic set, etc.

### 2.2 Applicability of Neutrosophic Theory

Applicability of neutrosophic theory is large, from social sciences such as sociology, philosophy, literature, arts (Smarandache, Vlăduțescu, 2014; Smarandache, 2015; Păun, Teodorescu, 2014; Opran, Voinea, Teodorescu,2014; Smarandache, Gîfu, Teodorescu, 2015 ) to sciences such as physics, artificial intelligence, mathematics (Smarandache, Vlădăreanu, 2014).

There are some remarkable results of netrosophic theory applied in practical applications such as artificial intelligence (Gal et al, 2014), in robotics there are confirmed results of neutrosofics logics applied to make decisions for uncertainty situations (Okuyama el al 2013; Smarandache, 2011), also for the real-time adaptive networked control of the robot movement on surface with uncertainties (Smarandache, 2014).

Athar Kharal has also a contribution to multi criteria making decision (MCDM) developing an algorithm of uncertainty criteria selection using neutrosophic sets. The proposed method allows the degree of satisfiability (t), non-satisfiability (f) and indeterminacy (i) mentioning a set of criteria represented by neutrosophic sets (Kharal, 2014).

There is no instant game, or instant action; if they existed, it would involve a very limited time fund, if they were instant, we could calculate the uncertainty, we should not have too many variables. If the time is longer, more variables appear, more uncertainties. We evaluate the situation “1” according to what every social actor wants, it is the sustained decision. The state "0" represents the decision that is rejected by social actors. Between “0” and “1” remain states of uncertainty, neutrality, uncertain decisions. In this manner we extend the fuzzy theory to neutrosophic theory. In fact, the novelty of neutrosophy consists of approaching the indeterminacy status. (Smarandahe, 2005).

Starting of this point, we are confidence that neutrosophic theory can help to analyze, evaluate and make the right decision in process analysis taking into account all sources that can generate uncertainty, from human being (not appropriate skill), logistics concept, lack of information, programming automation process according to requirements, etc.

## 3. Games, elements of culture, double articulated

Probably, Ludwig Wittgenstein was the first academic philosopher who addressed the definition of the word *game*. In his *work, Philosophical Investigations*, Wittgenstein argued that the "elements of games, such as play, rules, and competition", all contribute to define what games are, but not totally (Wittgenstein, 1953).

Jean Piaget suggested in his work, *Genetic Epistemology*, "that children think differently than adults and proposed a stage theory of cognitive development". He was the first one to note that children play an active role in gaining knowledge of the world, playing games; children can be thought of as "little scientists" who are actively constructing their knowledge and understanding of the world (Piaget, 1970; Piaget, 1983).

***Argument 1.***

Games are culture related, everything is repeated by as many people as possible, and it becomes acceptable to most, spread and cultivated (e.g. internet; it shows a minimal know-how). Culture is a complex principle of behavior, spiritual and material values created by mankind, beliefs, tradition and art, passed down from generation to generation. The sense of culture finds its significance in the life of an individual and society. In this context, "any authentic creation is a gift to the future." – asserts Albert Camus.

For humans, culture is the specific environment of existence. It defines an existential field, characterized by a synthesis between objective and subjective, between real and ideal. Culture defines a synthetic human way of existence and it is the symbol of man's creative force. It represents a real value system.

***Argument 2.***

Games produce culture, the Internet being unlimited, is a huge catalyst of desire. It is a sublime achievement in economic terms. For example, the Internet can offer so many texts about Kant that you never got around to finish in silence any of his "Critiques". The time of assimilation is now dedicated to search. More than ever, McLuhan's equation says it all about the Internet: "The medium is the message", says a voice that is heard beyond any meaning of utterances made. Only the pleasure, the voice, the search on Internet are now authentic. Time is limited, not space. From time only desire can provide the intensity necessary to forget this ontological asymmetry (Luhan, 1967).

The Internet allows many people to discover their identity more easily. Some people who were shy or lonely or feel unattractive, discover that they can socialize more successfully and express themselves more freely in an online environment.

Being able to pretend you are someone else is an important mental skill that the child acquires if he is involved in such games. The same thing is experienced by an adult on Internet games, on Facebook, for example, a doubling of personality, a place where you can be different, without constraints, where you are at your own free will, where decision belongs entirely to you, where only uncertainties hinder you. You can think what you want but you can never think of everything that can be thought. If it were possible for every man to think all that is conceivable and with a consistent content, there would be no freedom of thought or thoughts individualized particular to each topic. Mentally anticipating the future, one can access one's individual mental states of the distant future; one can get art work depicting thoughts, theories or advanced and complicated scientific technologies, currently unimaginable.

## 4. Case study

**EVE Online** is a “massively multiplayer online game set 23,000 years in the future. As an elite pilot of one of the four controlling races, the player will explore, build, and dominate across an universe of over 7,000 star systems”[1], see Figure 1. In EVE Online the possibilities are endless. Eve Online is a peculiar concept, it is a simulation, it is an experiment which mirrors the social interactions and communications of the real world, just like the real world, it has a fully functioning economy. In fact, “it has an economy which could be used and studied in order to help what we do in the real world, according to the man charged with overseeing how the $18 million economy operates”[2].

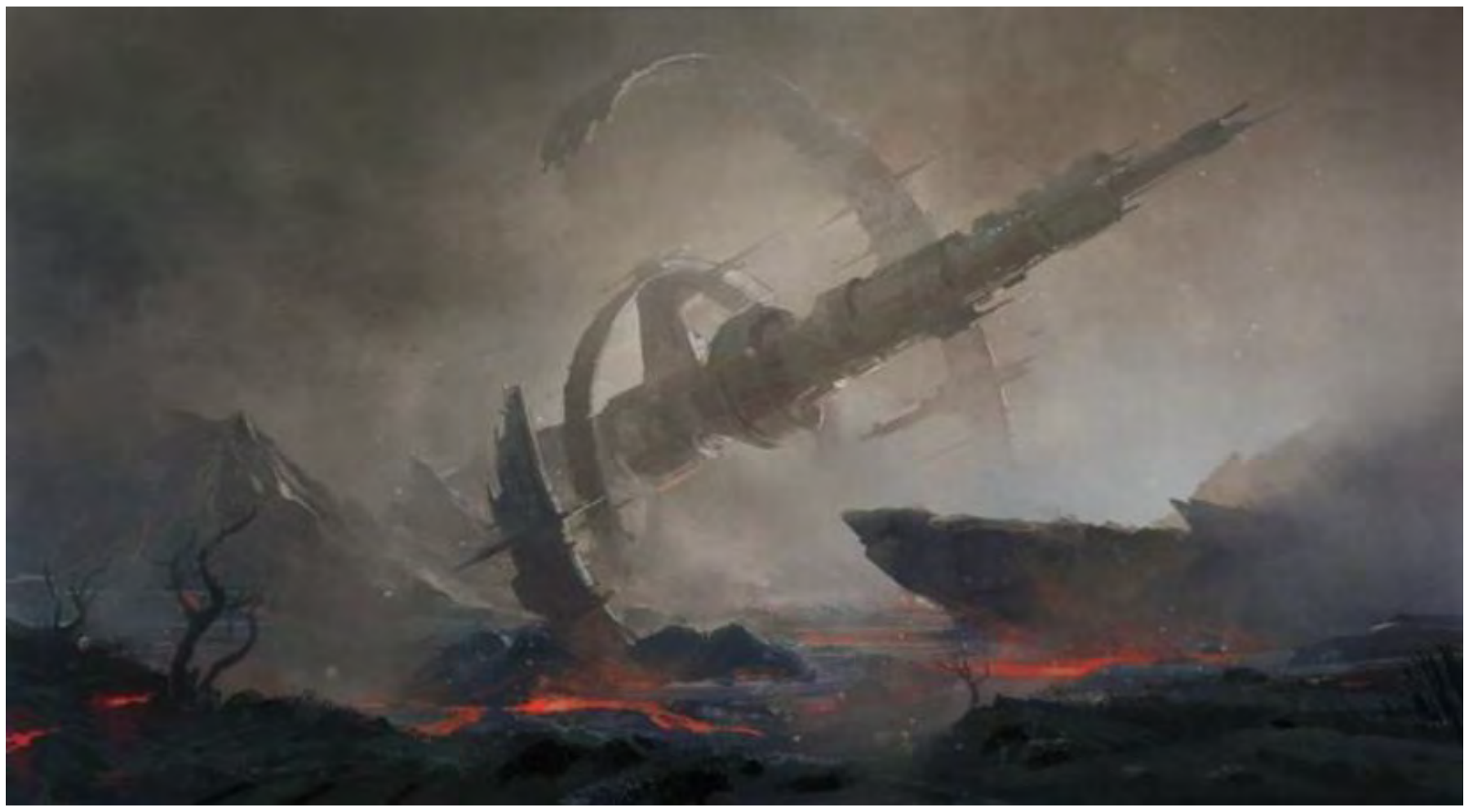

Figure 1.**Very real implications**

While what is happening in universe environment of EVE planets in the far off star systems may not have much relevance in today's world, but the way the EVE economy functions could have very real implications: "We try to have a relative balance of money coming in and money coming out and the increase per month should represent the net increase in economic value”; “We function as a national economics institute, statistics office and central bank giving advice to government, with the government being the developers and us being the monitoring authority"[3].

Considering data of EVE Online game, the complexity of environment, we can estimate some causes that can generate uncertainties such as: unknown universe; cohesion of the team members; alliance trust; financial system crisis; equipment reliability.

1 https://www.eveonline.com
2 David Gilbert, Eve Online: Meet the Man Controlling the $18 Million Space Economy, International Business Time, May 6, 2014
3 http://www.ibtimes.co.uk/eve-online-meet-man-controlling-18-million-space-economy-1447437

For each of these causes the occurrences in a time unit are analyzed (e.g.: 1 week), the space M assimilated to the environment of universe, the governance is poorly defined in this space: the control of universe, defeat the forces of evil, building a stable system. According to this conditions we can simulate the situation by a Pareto Chart, see Figure 2:

| Causes | Frequency | Relative frequency % | Cumulated frequency % |
|---|---|---|---|
| Unknown Universe | 18 | 35,29% | 35,29% |
| Team cohesion | 13 | 25,49% | 60,78% |
| Alliance trust | 10 | 19,61% | 80,39% |
| Financial stability | 6 | 11,76% | 92,16% |
| Equipment Reliability | 4 | 7,84% | 100,00% |
| | 51 | | |

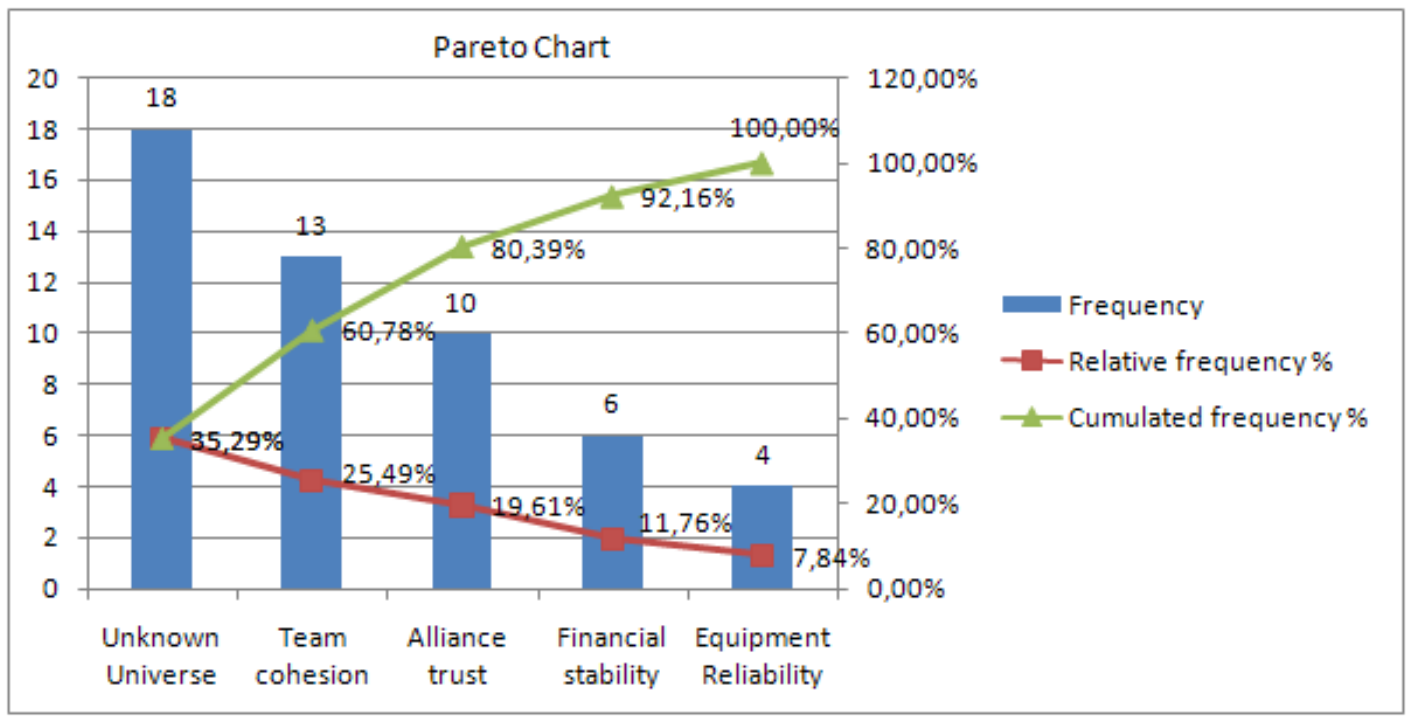

Figure 2. *Pareto Chart Step 1*

Pareto analysis is a creative way of evaluation causes of problems because it helps to stimulate the processes, thinking and organize thoughts, assessing the causes that lead to system instability through neutrosophic theory.

In this context, we define a space M consisting of 5 elements, where **t** means true, **i** means uncertainty and **f** means false:

$$M = \{ a_1 (t_1, i_1, f_1), a_2 (t_2, i_2, f_2), a_3 (t_3, i_3, f_3), a_4 (t_4, i_4, f_4), a_5 (t_5, i_5, f_5)\}$$

- Unknown universe: $a_1(t_1, i_1, f_1)$
- Team cohesion: $a_2(t_2, i_2, f_2)$
- Alliance trust: $a_3(t_3, i_3, f_3)$
- Financial stability: $a_4(t_4, i_4, f_4)$
- Equipment reliability: $a_5 (t_5, i_5, f_5)$

According to Pareto Chart, we established the rate for each space element percentage for the set (t, i, f).

Analyzing the content of elements data, we can establish that relative frequency of events means uncertainty and events solving means true, respectively non solving, false. The process is revealed in Table 1.

| Parameter | Space M | Frequency | T /action | T% | I% | F/action | F% |
|---|---|---|---|---|---|---|---|
| Unknown Universe | $a_{11}$ | 18 | 10 | 55,56 | 35,29 | 8 | 44,44 |
| Team cohesion | $a_{21}$ | 13 | 7 | 53,85 | 25,49 | 6 | 46,15 |
| Alliance trust | $a_{31}$ | 10 | 8 | 80,00 | 19,61 | 2 | 20,00 |
| Financial stability | $a_{41}$ | 6 | 5 | 83,33 | 11,76 | 1 | 16,67 |
| Equipment Reliability | $a_{51}$ | 4 | 3 | 75,00 | 7,84 | 1 | 25,00 |

Table 1. *Determination of T, I, F consistency - Step 1*

Neutrosophic interpretation gives an ordered list of alternatives of uncertainties, depending on us, which is the most preferred element.

We will stop at $a_2$component that generates uncertainty for the cause of "team cohesion". Pareto Chart says that by addressing the cause 20%, it determines 80% uncertainty and can also solve 80% of problems of the system stability. We have to concentrate on uncertainty of $a_2$ component, to reduce its value,

$a_{21}(t_{21}, i_{21}, f_{21})$ → $a_{21}$(53,85%, **25,49%,** 46,15%) in the first step of the process.

The refining process, step 2, can be seen in the next set of data presented in Figure 3.

| Causes | Frequency | Relative frequency % | Cumulated frequency % |
|---|---|---|---|
| Unknown Universe | 16 | 41,03% | 41,03% |
| Alliance trust | 9 | 23,08% | 64,10% |
| Team cohesion | 6 | 15,38% | 79,49% |
| Equipment Reliability | 4 | 10,26% | 89,74% |
| Financial stability | 4 | 10,26% | 100,00% |
| | 39 | | |

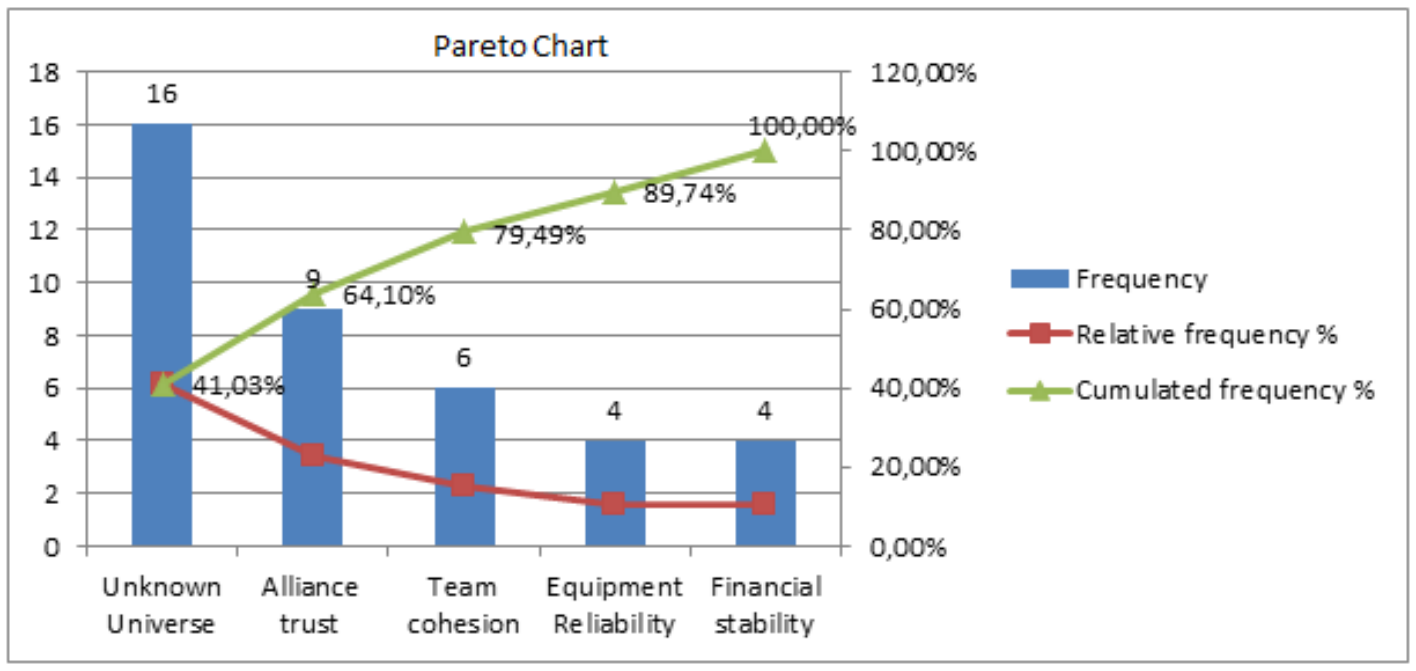

Figure 3. *Pareto Chart Step 2*

The relative dataset for the step 2 is shown in Table 2, where we follow up the element $a_{22}$.

| Parameter | Space M | Frequency | T /action | T% | I% | F/action | F% |
|---|---|---|---|---|---|---|---|
| Unknown Universe | $a_{12}$ | 16 | 12 | 75,00 | 41,03 | 4 | 25,00 |
| Alliance trust | $a_{32}$ | 9 | 7 | 77,78 | 23,08 | 2 | 22,22 |
| Team cohesion | $a_{22}$ | 6 | 4 | 66,67 | 15,38 | 2 | 33,33 |
| Equipment Reliability | $a_{52}$ | 4 | 3 | 75,00 | 10,26 | 1 | 25,00 |
| Financial stability | $a_{42}$ | 4 | 3 | 75,00 | 10,26 | 1 | 25,00 |

Table 2. *Determination of T, I, F consistency - Step 2*

$a_{22}(t_{22}, i_{22}, f_{22})$ → $a_{22}$(66,67%, **15,38%,** 33,33%) the second step of the process.

On the third step of the refining process, the data set is presented in Table 3.

| Parameter | Space M | Frequency | T /action | T% | I% | F/action | F% |
|---|---|---|---|---|---|---|---|
| Unknown Universe | $a_{13}$ | 16 | 12 | 75,00 | 43,24 | 4 | 25,00 |
| Alliance trust | $a_{33}$ | 9 | 7 | 77,78 | 24,32 | 2 | 22,22 |
| Equipment Reliability | $a_{53}$ | 4 | 3 | 75,00 | 10,81 | 1 | 25,00 |
| Financial stability | $a_{43}$ | 4 | 3 | 75,00 | 10,81 | 1 | 25,00 |
| Team cohesion | $a_{23}$ | 4 | 3 | 75,00 | 10,81 | 1 | 25,00 |

Table 3. *Determination of T, I, F consistency - Step 3*

$a_{23}(t_{23}, i_{23}, f_{23})$ → $a_{23}$(75%, **10.81%,** 25%) the third step of the process.

$a_{24}(t_{24}, i_{24}, f_{24})$ → $a_{24}$(100%, **8,33%,** 0%) the last step of the process.

The data set of last step of the refining process, is shown in Table 4.

| Parameter | Space M | Frequency | T /action | T% | I% | F/action | F% |
|---|---|---|---|---|---|---|---|
| Unknown Universe | $a_{14}$ | 16 | 12 | 75,00 | 44,44 | 4 | 25,00 |
| Alliance trust | $a_{34}$ | 9 | 7 | 77,78 | 25 | 2 | 22,22 |
| Equipment Reliability | $a_{54}$ | 4 | 3 | 75,00 | 11,11 | 1 | 25,00 |
| Financial stability | $a_{44}$ | 4 | 3 | 75,00 | 11,11 | 1 | 25,00 |
| Team cohesion | $a_{24}$ | 3 | 3 | 100,00 | 8,33 | 0 | 0,00 |

Table 4. *Determination of T, I, F consistency - Step 4*

We show below the effects of refining process in 4 steps, until the value of F is zero. The representation of "team cohesion" evolution is shown in Figure 4.

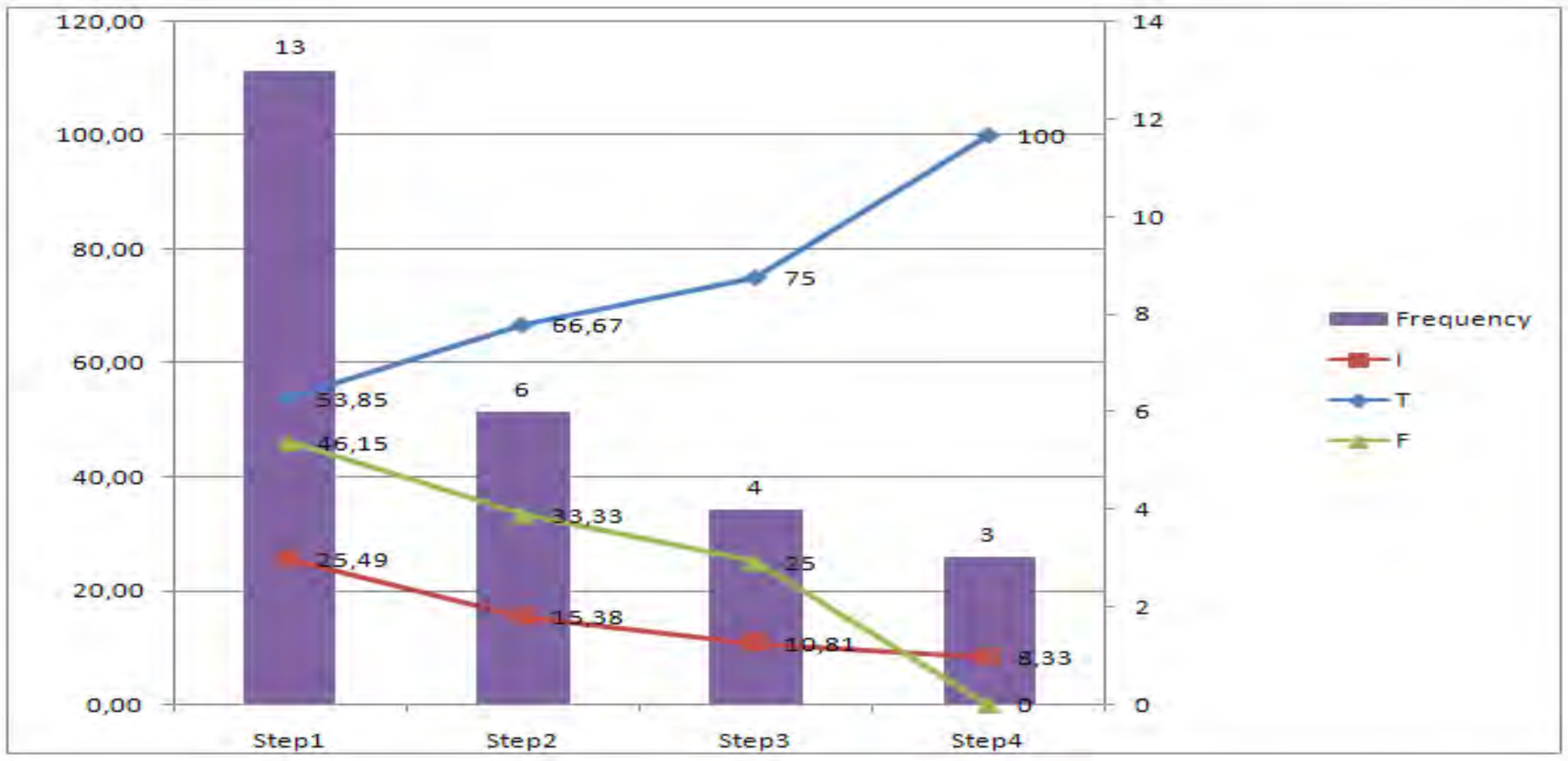

Figure 4. *Refining the process for "Team cohesion", element of space M*

The process of The Uncertainty Risk Management will take into consideration:

- Uncertainty management is a creative process, it involves identifying, evaluation and mitigation of the impact of the uncertainties in the process;
- Uncertainty management can be very formal with defined work process, or informal with no defined processes or methods;
- Uncertainty evaluation prioritizes the identified uncertainties by the likelihood and the potential impact if the event happens;
- Uncertainty mitigation is the development and deployment of a plan to avoid, transferring, sharing and reducing the process uncertainties.

When we try to make a good decision, a person must weigh the positives, negatives and uncertainty of each option, and to consider all the alternatives. For effective decision making, a person must be able to forecast the outcome of each option as well, and based on all these items, to determine which option is the best for that particular situation.

Decision-making is identified as a cognitive process that results in the selection of a belief or an action among several alternative possibilities. Decision-making is a complex process of identifying, analyzing and choosing alternatives based on the values and preferences of the decision maker. Decision-making is one of the central activities of management and it is an important part of any implementation process (Kahneman, Tversky, 2000).

Usually, in our daily lives, we implicitly compare multiple criteria and we want to be comfortable with the consequences of such decisions that are mostly made based only on intuition. On the other hand, when we confront with high stakes, it is important to structure the problem and to evaluate multiple criteria. In decision making process based on multiple criteria (Multi Criteria Decision Making) of whether to do an important issue or not, there are involved not only very complex multiple criteria, there are also inferred multiple parties who are deeply affected from the consequences, because present decisions, act in the future.

Decisions making related to games area, shows its similitude with real life, can be easily transferred to the real world. It is our choice whether to do this or not.

## 5. Conclusions

Decisions making is a complex act including variables related to uncertainty, with implications for the future work. Uncertainty, in turn, involves classification criteria based on methods that may be applied for determining the degree of uncertainty and settlement. Establishing the types of variables that influence uncertainty, it makes possible the identification of the decisions that we are referring, that will influence, will constrain the process on the one hand, and will be influenced and constrained by a specific decision on the other hand.

Problem solving and decision-making are important skills for business and life. Problem-solving often involves decision-making, and decision-making is especially important for management and leadership. Between them there is the correspondence: identification of the problem vs. frame of the decision; exploring the alternatives vs. improve to address needs and identify alternatives; select an alternative vs. decision and commitment to act; implementation of the solution vs. management of the consequences; evaluation of the situation vs. management of the consequences and frame the related decisions.

What we deduced on the basis of this study is that the game is reality and reality is game. We build reality through the game, we take risks that include uncertainties, the game becomes a training and an experimentation place for many specialists, proving that the school becomes life. Here is how the neutrosophic theory, guide us to be closer to solve uncertainties, transforming them into true or false, stable and controllable states of the systems.

## References

1. Adler, A. (2011).The practice and theory of individual psychology. Martino Fine Books.
2. Aumann, R.,Peleg, P.(1960). Von Neumann-Morgenstern solutions to cooperative games without side payments, Bull. Am. Math. Sot.66, 173-179.
3. Aumann, R., Hart,S.ed. (1992, 2002). Handbook of game theory with economic applications v. 1. Elsevier/North-Holland, Amsterdam.
4. Berne, E. (1964).Games people play: the basic hand book of transactional analysis. New York: Ballantine Books.
5. Erikson, E. (1993).Childhood and society. W.W. Norton & Company, Inc.
6. Gal, I.A., Vlădăreanu, L., Smarandache,F., H. Yu, M. Deng. (2014). Neutrosophic logic approaches applied to" rabot" real time control ,Vol 1,pp. 55-60,EuropaNova, Bruxelles.
7. Heath,T., Bizer,C.,(2011). Linked data: evolving the web into a global data space (1st edition). Synthesis lectures on the semantic web: theory and technology, 1:1, 1-136. Morgan & Claypool.
8. Huizinga, J.(1980). Homo ludens, Redwood Burn Ltd Trowbridge & Esher Jenkins, h.interactive audiences? the 'collective intelligence' of media fans, http://web.mit.edu/21fms/People/henry3/collectiveintelligence.html
9. Janis I., Mann, L. (1977). Decision making: a psychological analysis of conflict, choice, and commitment. New York: Free Press.
10. Kharal, A. (2014). A neutrosophic multi-criteria decision making method. New Mathematics and Natural Computation, 10 (2),143–162. DOI: 10.1142/S1793005714500070
11. Kahneman,D.,Tversky, A. (2000). Choices, values, and frames. New York; Cambridge, UK: Russell Sage Foundation; Cambridge University Press.
12. Mann, L., Harmoni, R., Power, C. (1991). The GOFER course in decision making, In Baron, Jonathan; Brown, Rex V. Teaching decision making to adolescents. Hillsdale, NJ: Lawrence Erlbaum Associates. pp. 61–78.
13. McLuhan, M. (1967).The Medium is the Massage: An Inventory of Effects, Penguin Books.
14. Okuyama,K.,MohdAnasri, Smarandache, F., Kroumov, V., (2013). Mobile robot navigation using artificial landmarks and GPS. Bulletin of the Research Institute of Technology, Okayama University of Science, Japan, 31, 46-51.
15. Opran, E.,Voinea, D.,Teodorescu, M., (2015). Art and being in neutrosophic communication. International Letters of Social and Humanistic Sciences, 6(1), 16-27.
16. Păun, M.G., Teodorescu, M., (2014). Hermeneutics can make beauty and ugly as neutral (as neutrosophic). Social Sciences and Education Research Review, 2, 252-61.
17. Popper, K.(2011).The Open Society and Its Enemies, Publisher George Routledge And Sons Limited.
18. Piaget, J. (1983). Piaget's theory. In P. Mussen (ed) Handbook of Child Psychology. 4th edition.Vol. 1. New York: Wiley.
19. Piaget, J. (1970). Genetic epistemology. New York: Norton.
20. Smarandache, F. (2015).Neutrosophic social structures specificities. Social Sciences and Education Research Review, 2(1), 13-20.

21. Smarandache, F., (2005). A unifying field in logics: neutrosophic logic, neutrosophy, neutrosophic set, neutrosophic probability and statistics. American Research Press, Rehoboth.
22. Smarandache, F.,Vlădăreanu, L.,(2014). Applications of neutrosophic logic on robots. Neutrosophic Theory and its Applications,Collected papers Vol.1,Brussels, 2014.
23. Smarandache, F. (2015). Thesis-antithesis-neutrothesis, and neutrosynthesis. Neutrosophic Sets and Systems, 8, 64-67.
24. Smarandache, F. (2015). (T, I, F)-neutrosophic and I-neutrosophic structures. Neutrosophic Sets and Systems, 8, 3-10.
25. Smarandache, F., Vlăduțescu, S.,(2014). Communicative universal convertibility matter-energy-information. Social Sciences and Education Research Review,1, 44-62.
26. Smarandache, F.,Gîfu, D., Teodorescu, M., (2015). Neutrosophic elements in discourse. Social Sciences and Education Research Review, (2), 125-32.
27. Smarandache, F., Gîfu, D., Teodorescu, M. (2015). Neutrosophy, a possible method of process analysis uncertainties solving , chapter in book- Uncertainty Communication Solution in Neutrosophic Key-Florentin Smarandache, Bianca Teodorescu, Mirela Teodorescu (Editors). ISBN: 978-1-59973-371-5
28. Thompson, J. (2006). "Violent video games feed unhealthy ideas to young kids". Tacoma News Tribune, January 8.
29. Şahin, R., Multi-criteria neutrosophic decision making method based on score and accuracy functions under neutrosophic environment. http://arxiv.org/abs/1412.5202
30. Vlădăreanu, L. , Gal, I.A. , H. Yu, M. Deng, (2015).Robot control intelligent interfaces using the DSmT and the neutrosophic logic. International Journal of Advanced Mechatronic Systems, 6(2-3), 128-135.
31. Von Neumann, J., Morgenstern,O., (1944).Theory of games and economic behavior. Princeton University Press.
32. Wittgenstein, L. (1953). Philosophical investigations. Oxford: Blackwell.

PARTHA PRATIM DEY[1], SURAPATI PRAMANIK[2,*], BIBHAS C. GIRI[3]

1, 3 Department of Mathematics, Jadavpur University, Kolkata-700032, West Bengal, India
*2 Department of Mathematics, Nandalal Ghosh B.T. College, Panpur, P.O.-Narayanpur, District –North 24 Parganas, Pin code-743126,West Bengal, India. Corresponding Author's E-mail: sura_pati@yahoo.co.in

# Extended Projection Based Models for Solving Multiple Attribute Decision Making Problems with Interval Valued Neutrosophic Information

## Abstract

The paper develops two new methods for solving multiple attribute decision making problems with interval – valued neutrosophic assessments. In the decision making situation, the rating of alternatives with respect to the predefined attributes is described by linguistic variables, which can be represented by interval - valued neutrosophic sets. We assume that the weight of the attributes are not equal in the decision making process and they are obtained by using maximizing deviation method. We define weighted projection measure and propose a method to rank the alternatives. Furthermore, we also develop an alternative method to solve multiple attribute decision making problems based on the combination of angle cosine and projection method. Finally, an illustrative numerical example in Khadi institution is solved to verify the effectiveness of the proposed methods.

## 1. Introduction

Multiple attribute decision making (MADM) is one of the most significant parts of modern decision science and it is a well known method for selecting the most desirable alternative from a set of all feasible alternatives with respect to some predefined attributes. However, the information about the attributes is generally incomplete, indeterminate and inconsistent in nature due to the complexity of real world problems. Smarandache [1-4] grounded the concept of neutrosophic sets (NSs) from philosophical point of view by incorporating the degree of indeterminacy or neutrality as independent component to deal with problems involving imprecise, indeterminate and inconsistent information and the concept of NSs has been applied to different fields such as decision sciences, social sciences, humanities, etc. From scientific and realistic point of view, Wang et al. [5] defined single valued NSs (SVNSs) and then presented the set theoretic operators and various properties of SVNSs. Wang et al. [6] also developed the notion of interval neutrosophic sets (INSs) characterized by membership, non-membership and falsity-membership functions, whose values are interval rather than real numbers.

In 2013, Chi and Liu [7] first discussed a novel approach for solving MADM problems based on extended TOPSIS method under interval neutrosophic environment. Zhang et al. [8] defined some operators for INSs and established a multi-criteria decision making method. Broumi and Smarandache [9] defined cosine similarity measure between two INSs and applied the concept to medical diagnosis problem. Ye [10] proposed some similarity measures between two IVNSs based on the relationship of similarity measures and distance measures and utilized the developed method to solve a multi-criteria decision making problem. Sahin and Liu [11] developed maximizing deviation method for solving MADM problems having incomplete weight information. They employed single valued neutrosophic weighted averaging operator and interval neutrosophic weighted averaging operator in order to aggregate the neutrosophic information corresponding to each alternative and the most desirable alternatives are obtained based on score and accuracy functions. Pramanik and Mondal [12] discussed interval neutrosophic MADM based on grey relational analysis (GRA) method where the unknown attribute weights are derived from information entropy method. Later, Dey et al. [13] studied an extended GRA based interval neutrosophic MADM for weaver selection in Khadi institution. Mondal and Pramanik [14] proposed cosine, Dice and Jaccard similarity measures of interval rough neutrosophic set for solving MADM problems. Recently, Dey et al. [15] investigated an extended GRA method for MADM problem with interval neutrosophic uncertain linguistic information.

Projection measure is useful device for solving decision making problems because it takes into account the distance as well as the included angle between points evaluated [16]. Xu and Hu [17] provided projection models for dealing with intuitionistic fuzzy MADM problems. Zeng et al. [18] demonstrated weighted projection algorithms for multiple arttribute group decision problems under intuitionistic fuzzy and interval – valued intuitionistic fuzzy environment. Yue [19-20] presented a projection method to obtain weights of the experts in a group decision making problem. Yue [21] proposed a projection based approach for partner selection in a group decision making problem with linguistic values and intuitionistic fuzzy information. Ju and Wang [22] investigated a methodology to multicriteria group decision problems with incomplete weight information in linguistic setting based on projection method. Yang and Du [23] developed a straightforward method for obtaining the weights of the decision makers based on angle cosine and projection method. Ye [24] discussed a simplified neutrosophic harmonic averaging projection based method to solve MADM problems. Ye [25] provided a decision making method based on credibility-induced interval neutrosophic weighted arithmetic averaging operator and credibility-induced interval neutrosophic weighted geometric averaging operator and the projection measure-based ranking method to solve MADM problems with interval neutrosophic information and credibility information.

In this paper, we define weighted projection measure for interval – neutrosophic information and develop a method for solving MADM problems based on weighted projection method. We also investigate a method for MADM under interval - valued neutrosophic environment based on the combination of angle cosine and projection method.

Rest of the paper is prepared as follows: Sec. 2 presents several definitions. An interval - valued neutrosophic MADM based on weighted projection method is discussed in Sec. 3. Subsection 3.1 presents the algorithm for MADM problems with interval valued neutrosophic information based on weighted projection method. Subsection 3.2 presents the approach for solving interval - valued neutrosophic MADM problems based on angle cosine and projection method. Subsection 3.3 presents the algorithm for MADM problem with interval valued neutrosophic information based on angle cosine and projection method In Sec. 4, we solve a numerical example to show the applicability and feasibility of the proposed method. Sec. 5 provides conslusion and future scope of research.

## 2. Preliminaries

In this Section, we briefly present some basic definitions which will be useful for the formulation of the paper.

### 2.1 Neutrosophic set

**Definition 2.1.1** [1-4]: Consider $U$ be a universal space of points with generic element in $U$ denoted by $x$. Then a NS $A$ is defined as follows:

$$A = \{x, \langle \mathrm{T}_A(x), \mathrm{I}_A(x), \mathrm{F}_A(x) \rangle \mid x \in U\} \tag{2.1}$$

where, $\mathrm{T}_A(x)$ , $\mathrm{I}_A(x)$ , $\mathrm{F}_A(x)$ : $U \to\ ]^-0, 1^+[$ are the truth-membership, indeterminacy-membership, and falsity-membership functions, respectively with $^-0 \le \sup \mathrm{T}_A(x) + \sup \mathrm{I}_A(x) + \sup \mathrm{F}_A(x) \le 3^+$.

**Definition 2.1.2.** [5] Let $U$ be a universal space of points with generic element in $U$ represented by $x$. Then, a SVNS $S \subset U$ is defined as follows:

$$S = \{x, \langle \mathrm{T}_S(x), \mathrm{I}_S(x), \mathrm{F}_S(x) \rangle \mid x \in U\} \tag{2.2}$$

where $\mathrm{T}_S(x)$ , $\mathrm{I}_S(x)$ and $\mathrm{F}_S(x)$ denote truth-membership, indeterminacy-membership and falsity-membership functions, respectively. For each point $x \in U$, we have, $\mathrm{T}_S(x), \mathrm{I}_S(x), \mathrm{F}_S(x)$: $U \to [0, 1]$ and $0 \le \sup \mathrm{T}_S(x) + \sup \mathrm{I}_S(x) + \sup \mathrm{F}_S(x) \le 3$.

**Definition 2.1.3.** [6] Let $U$ be a universe of discourse, with a generic element in $U$ represented by $x$. An interval valued neutrosophic set $N$ is represented as follows:

$$N = \{x, \langle \mathrm{T}_N(x), \mathrm{I}_N(x), \mathrm{F}_N(x) \rangle \mid x \in U\} \tag{2.3}$$

where $\mathrm{T}_N(x)$ , $\mathrm{I}_N(x)$ , $\mathrm{F}_N(x)$ are the truth-membership function, indeterminacy-membership function, and falsity-membership function, respectively. For each point $x \in U$, $\mathrm{T}_N(x)$ , $\mathrm{I}_N(x)$ , $\mathrm{F}_N(x) \subseteq [0, 1]$ and $0 \le \sup \mathrm{T}_N(x) + \sup \mathrm{I}_N(x) + \sup \mathrm{F}_N(x) \le 3$.

For convenience, if $\mathrm{T}_N(x) = [\mathrm{T}_N^L(x), \mathrm{T}_N^U(x)]$; $\mathrm{I}_N(x) = [\mathrm{I}_N^L(x), \mathrm{I}_N^U(x)]$; $\mathrm{F}_N(x) = [\mathrm{F}_N^L(x), \mathrm{F}_N^U(x)]$, then

$$N = \{x, \langle [\mathrm{T}_N^L(x), \mathrm{T}_N^U(x)], [\mathrm{I}_N^L(x), \mathrm{I}_N^U(x)], [\mathrm{F}_N^L(x), \mathrm{F}_N^U(x)], \rangle \mid x \in U\} \tag{2.4}$$

with the condition $0 \le \sup T_N^U(x) + \sup I_N^U(x) + \sup F_N^U(x) \le 3$.

For convenience, an interval valued neutrosophic number (IVNN) $\tilde{p}$ is represented by

$$\tilde{p} = \langle [T^-, T^+], [I^-, I^+], [F^-, F^+] \rangle. \tag{2.5}$$

## 2.2 Projection method

**Definition 2.2.1 [16, 26]:** Let $e = (e_1, e_2, \ldots, e_q)$ be a vector, then norm of $e$ is defined by

$$\| e \| = \sqrt{\sum_{j=1}^{q} e_j^2} \tag{2.6}$$

**Definition 2.2.2 [16, 26]:** Let $e = (e_1, e_2, \ldots, e_q)$ and $f = (f_1, f_2, \ldots, f_q)$ be two vectors, then angle cosine between $e$ and $f$ is defined as follows:

$$\text{Cos}\,(e, f) = \frac{\sum_{j=1}^{q} (e_j f_j)}{\sqrt{\sum_{j=1}^{q} e_j^2} \times \sqrt{\sum_{j=1}^{q} f_j^2}} \tag{2.7}$$

Obviously, $0 < \text{Cos}\,(e, f) \le 1$, and $\text{Cos}\,(e, f)$ denotes the closeness between $e$ and $f$ only in direction.

**Definition 2.2.3 [27]:** Consider $p_1 = \langle [T_1^-, T_1^+], [I_1^-, I_1^+], [F_1^-, F_1^+] \rangle$ and $p_2 = \langle [T_2^-, T_2^+], [I_2^-, I_2^+], [F_2^-, F_2^+] \rangle$ be two IVNNs. Then the angle cosine of the included angle between $p_1$ and $p_2$ is defined as follows:

$$Cos\ (p_1,\ p_2) = \frac{(T_1^- T_2^- + T_1^+ T_2^+ + I_1^- I_2^- + I_1^+ I_2^+ + F_1^- F_2^- + F_1^+ F_2^+)}{\sqrt{((T_1^-)^2 + (T_1^+)^2 + (I_1^-)^2 + (I_1^+)^2 + (F_1^-)^2 + (F_1^+)^2)}\sqrt{((T_2^-)^2 + (T_2^+)^2 + (I_2^-)^2 + (I_2^+)^2 + (F_2^-)^2 + (F_2^+)^2)}} \tag{2.8}$$

**Definition 2.2.4 [16, 26]:** Let $e = (e_1, e_2, \ldots, e_q)$ and $f = (f_1, f_2, \ldots, f_q)$ be two vectors, then the projection of vector $e$ onto vector $f$ can be defined as follows:

$$Proj\ (e)_f = \| e \| \,\text{Cos}\,(e, f) = \sqrt{\sum_{j=1}^{q} e_j^2} \times \frac{\sum_{j=1}^{q} (e_j f_j)}{\sqrt{\sum_{j=1}^{q} e_j^2} \times \sqrt{\sum_{j=1}^{q} f_j^2}} = \frac{\sum_{j=1}^{q} (e_j f_j)}{\sqrt{\sum_{j=1}^{q} f_j^2}} \tag{2.9}$$

where, $Proj\ (e)_f$ indicates that the closeness of $e$ and $f$ in magnitude.

**Definition 2.2.5 [25]:** Consider $U = (u_1, u_2, \ldots, u_m)$ be a finite universe of discourse and $A$, $B$ be two IVNSs in $U$, then

$$Proj\ (A)_B = \frac{1}{\| B \|} \sum_{j=1}^{q} (e_j f_j) = \frac{1}{\| B \|} \sum_{j=1}^{q} (T_{\alpha_j}^- T_{\beta_j}^- + T_{\alpha_j}^+ T_{\beta_j}^+ + I_{\alpha_j}^- I_{\beta_j}^- + I_{\alpha_j}^+ I_{\beta_j}^+ + F_{\alpha_j}^- F_{\beta_j}^- + F_{\alpha_j}^+ F_{\beta_j}^+) \tag{2.10}$$

is called the projection of $A$ on $B$, where $\alpha_j = \langle [T_{\alpha_j}^-, T_{\alpha_j}^+], [I_{\alpha_j}^-, I_{\alpha_j}^+], [F_{\alpha_j}^-, F_{\alpha_j}^+] \rangle$ and $\beta_j = \langle [T_{\beta_j}^-, T_{\beta_j}^+], [I_{\beta_j}^-, I_{\beta_j}^+], [F_{\beta_j}^-, F_{\beta_j}^+] \rangle$ are the i-th IVNNs of $A$ and $B$ respectively. Especially, when q = 1, we obtain the projection of $\alpha_1$ on $\beta_1$ as follows:

$$Proj\ (\alpha_1)_{\beta_1} = \frac{1}{\| \beta_1 \|} (T_{\alpha_j}^- T_{\beta_j}^- + T_{\alpha_j}^+ T_{\beta_j}^+ + I_{\alpha_j}^- I_{\beta_j}^- + I_{\alpha_j}^+ I_{\beta_j}^+ + F_{\alpha_j}^- F_{\beta_j}^- + F_{\alpha_j}^+ F_{\beta_j}^+) \tag{2.11}$$

**Definition 2.2.6:** Consider $U$ = ($u_1$, $u_2$, …, $u_m$) be a finite universe of discourse and $A$ be an IVNS in $U$, then

$$\| A \| = \sqrt{\sum_{j=1}^{m} \alpha_j^2} \tag{2.12}$$

is called the modulus of $A$, where $\alpha_j = \left\langle [T^-_{\alpha_j}, T^+_{\alpha_j}], [I^-_{\alpha_j}, I^+_{\alpha_j}], [F^-_{\alpha_j}, F^+_{\alpha_j}] \right\rangle$ .

**Definition 2.2.7:** Consider $U$ = ($u_1$, $u_2$, …, $u_m$) be a finite universe of discourse and $A$ be an IVNS in $U$, then

$$\| A \|_w = \sqrt{\sum_{j=1}^{m} (w_j \alpha_j)^2} \tag{2.13}$$

is said to be the weighted modulus of $A$, where $\alpha_j = \left\langle [T^-_{\alpha_j}, T^+_{\alpha_j}], [I^-_{\alpha_j}, I^+_{\alpha_j}], [F^-_{\alpha_j}, F^+_{\alpha_j}] \right\rangle$ and $w = \{w_1, w_2, …, w_m\}$ be the weight vector assigned for $\beta_j$, where $0 \le w_j \le 1$ with $\sum_{j-1}^{m} w_j = 1$.

**Definition 2.2.8:** Consider $U$ = ($u_1$, $u_2$, …, $u_m$) be a finite universe of discourse and $A$, $B$ be any two IVNSs in $U$, then

$$Proj_{\ w}(A)_B = \frac{1}{\| B \|_w} \sum_{j=1}^{q} (e_j f_j) = \frac{1}{\| B \|_w} \sum_{j=1}^{q} (T^-_{\alpha_j} T^-_{\beta_j} + T^+_{\alpha_j} T^+_{\beta_j} + I^-_{\alpha_j} I^-_{\beta_j} + I^+_{\alpha_j} I^+_{\beta_j} + F^-_{\alpha_j} F^-_{\beta_j} + F^+_{\alpha_j} F^+_{\beta_j})$$
(2.14)

is said to be the weighted projection of $A$ on $B$, where $\alpha_j = \left\langle [T^-_{\alpha_j}, T^+_{\alpha_j}], [I^-_{\alpha_j}, I^+_{\alpha_j}], [F^-_{\alpha_j}, F^+_{\alpha_j}] \right\rangle$ and $\beta_j = \left\langle [T^-_{\beta_j}, T^+_{\beta_j}], [I^-_{\beta_j}, I^+_{\beta_j}], [F^-_{\beta_j}, F^+_{\beta_j}] \right\rangle$ are the i-th IVNNs of $A$ and $B$ respectively. Consider $w = \{w_1, w_2, …, w_q\}$ be the weight vector assigned for $\beta_j$, where $0 \le w_j \le 1$ with $\sum_{j-1}^{q} w_j = 1$.

**Definition 2.2.7 [28]:** Consider $\alpha$ = ([$T_1^-, T_1^+$], [$I_1^-, I_1^+$],[$F_1^-, F_1^+$]) and $\beta$ = ([$T_2^-, T_2^+$], [$I_2^-$, $I_2^+$],[$F_2^-, F_2^+$]) be any two IVNNs, then Hamming distance between $\alpha$ and $\beta$ is defined as follows:

$$\Re_{Ham}(\alpha, \beta) = 1/6(|T_1^- - T_2^-| + |T_1^+ - T_2^+| + |I_1^- - I_2^-| + |I_1^+ - I_2^+| + |F_1^- - F_2^-| + |F_1^+ - F_2^-|). \tag{2.15}$$

**Definition 2.2.8 [28]:** Consider $\alpha$ = ([$T_1^-, T_1^+$], [$I_1^-, I_1^+$],[$F_1^-, F_1^+$]) and $\beta$ = ([$T_2^-, T_2^+$], [$I_2^-$, $I_2^+$],[$F_2^-, F_2^+$]) be any two IVNNs, then the Euclidean distance between $\alpha$ and $\beta$ is defined as given below.

$$\Re_{Euc}(\alpha, \beta) = \sqrt{1/6\left((T_1^- - T_2^-)^2 + (T_1^+ - T_2^+)^2 + (I_1^- - I_2^-)^2 + (I_1^+ - I_2^+)^2 + (F_1^- - F_2^-)^2 + (F_1^+ - F_2^+)^2\right)}$$
(2.16)

**Definition 2.2.9 [28]:** Let $A$ = ([$\dot{T}_i^-, \dot{T}_i^+$], [$\dot{I}_i^-, \dot{I}_i^+$], [$\dot{F}_i^-, \dot{F}_i^+$]), (i = 1, 2, ..., m) and $B$ = ([$\hat{T}_i^-, \hat{T}_i^+$], [$\hat{I}_i^-, \hat{I}_i^+$], [$\hat{F}_i^-, \hat{F}_i^+$]), (i = 1, 2, ..., m) be any two IVNSs, then the Hamming distance between $A$ and $B$ is presented as given below.

$$\Re_{Ham}(A, B) = \frac{1}{6m} \sum_{i=1}^{m} (|\dot{T}_i^- - \hat{T}_i^-| + |\dot{T}_i^+ - \hat{T}_i^+| + |\dot{I}_i^- - \hat{I}_i^-| + |\dot{I}_i^+ - \hat{I}_i^+| + |\dot{F}_i^- - \hat{F}_i^-| + |\dot{F}_i^+ - \hat{F}_i^+|) \tag{2.17}$$

**Definition 2.2.10 [28]:** Consider $A$ = ([$\dot{T}_i^-, \dot{T}_i^+$], [$\dot{I}_i^-, \dot{I}_i^+$], [$\dot{F}_i^-, \dot{F}_i^+$]), (i = 1, 2, ..., m) and B = ([$\hat{T}_i^-, \hat{T}_i^+$], [$\hat{I}_i^-, \hat{I}_i^+$], [$\hat{F}_i^-, \hat{F}_i^+$]), (i = 1, 2, ..., m) be any two IVNSs, then the Euclidean distance between $A$ and $B$ is defined as given below.

$$\Re_{Euc}(A, B) = \sqrt{1/6m \sum_{i=1}^{m} \left( (\dot{T}_i^- - \hat{T}_i^-)^2 + (\dot{T}_i^+ - \hat{T}_i^+)^2 + (\dot{I}_i^- - \hat{I}_i^-)^2 + (\dot{I}_i^+ - \hat{I}_i^+)^2 + (\dot{F}_i^- - \hat{F}_i^-)^2 + (\dot{F}_i^+ - \hat{F}_i^+)^2 \right)} \quad (2.18)$$

## 2.3 Conversion between linguistic variables and IVNNs

A variable whose values can be represented in terms of words or sentences in a natural language is said to be a linguistic variable. The performance values of the alternatives with respect to attributes can be expressed by linguistic variables such as extreme good, very good, good, and medium good, etc. Linguistic variables can be transformed into IVNNs as given below [15].

**Table 1.** *Transformation between the linguistic variables and IVNNs*

| Linguistic variables | IVNNs |
|---|---|
| Extreme good (EG) | ([0.95, 1], [0.05, 0. 1], [0, 0.1]) |
| Very good (VG) | ([0.75, 0.95], [0.1, 0.15], [0.1, 0.2]) |
| Good (G) | ([0.6, 0.75], [0.1, 0.2], [0.2, 0.25]) |
| Medium Good (MG) | ([0.5, 0.6], [0.2, 0.25], [0.25, 0.35]) |
| Medium (M) | ([0.4, 0.5], [0.2, 0.3], [0.35, 0.45]) |
| Medium low (ML) | ([0.3, 0.4], [0.15, 0.25], [0.45, 0.5]) |
| Low (L) | ([0.2, 0.3], [0.1, 0.2], [0.5, 0.65]) |
| Very low (VL) | ([0.05, 0.2], [0.1, 0.15], [0.65, 0.8]) |
| Extreme low (EL) | ([0, 0.05], [0.05, 0. 1], [0.8, 0.95]) |

# 3. An interval - valued neutrosophic MADM based on weighted projection method

Assume that H = {$h_1, h_2, \ldots, h_m$}, (m≥2) be a discrete set of alternatives and K ={$k_1, k_2, \ldots, k_n$}, (n≥2) be a set of attributes under consideration in a MADM problem. The rating of performance value of alternative $h_i$, i = 1, 2, …, m with respect to the predefined attribute $k_j$, j = 1, 2, …, n is represented by linguistic variables. The linguistic variables can be expressed by IVNNs. Assume $w = \{w_1, w_2, \ldots, w_n\}$ be the unknown weight vector of the attributes, where $0 \le w_j \le 1$ with $\sum_{j=1}^{n} w_j = 1$. The weighted projection method for solving MADM problem with interval -valued neutrosophic information is described by using the following steps:

**Step 1. Formulation of decision matrix with IVNNs**

The evaluation value of the alternative $h_i$, i = 1, 2, …, m with respect to the attribute $k_j$, j = 1, 2, …, n is presented by the expert in terms of linguistic variables that can be expressed by IVNNs. Therefore, interval – valued neutrosophic decision matrix $D_{\tilde{N}}$ is presented as given below.

$$D_{\tilde{N}} = \left\langle r_{\tilde{N}_{ij}} \right\rangle_{m\times n} = \begin{bmatrix} r_{11} & r_{12} & \cdots & r_{1n} \\ r_{21} & r_{22} & \cdots & r_{2n} \\ . & . & \cdots & . \\ . & . & \cdots & . \\ r_{m1} & r_{m2} & \cdots & r_{mn} \end{bmatrix} \tag{3.1}$$

Here $r_{ij} = <[T^-_{ij}, T^+_{ij}], [I^-_{ij}, I^+_{ij}], [F^-_{ij}, F^+_{ij}]>$; $T^-_{ij}, T^+_{ij}, I^-_{ij}, I^+_{ij}, F^-_{ij}, F^+_{ij} \in [0, 1]$ and $0 \leq \sup T^+_{ij} + \sup I^+_{ij} + \sup F^+_{ij} \leq 3$, $i = 1, 2, \ldots, m$; $j = 1, 2, \ldots, n$. Here, $[T^-_{ij}, T^+_{ij}]$ represents the degree that the alternative $h_i$ satisfies the attribute $k_j$. $[I^-_{ij}, I^+_{ij}]$ denotes the degree that the alternative $h_i$ is indeterminacy on the attribute $k_j$. $[F^-_{ij}, F^+_{ij}]$ indicates the degree that the alternative $h_i$ does not satisfies the attribute $k_j$.

**Step 2. Standardize the decision matrix**

Generally, two types of attributes are encountered in practical decision making problems such as benefit type attribute where bigger value of the attribute reflects better alternative and cost type attribute where bigger value of the attribute reflects worse alternative. However, in order to remove the influence of different physical dimensions to decision results, we require to standardize the decision matrix. The standardize decision matrix $S = [s_{ij}]_{m \times n}$ owing to Chi and Liu [7] is formulated as follows:

$$D_{\hat{S}} = \left\langle t_{\hat{S}_{ij}} \right\rangle_{m\times n} = \begin{bmatrix} t_{11} & t_{12} & \cdots & t_{1n} \\ t_{21} & t_{22} & \cdots & t_{2n} \\ . & . & \cdots & . \\ . & . & \cdots & . \\ t_{m1} & t_{m2} & \cdots & t_{mn} \end{bmatrix} \tag{3.2}$$

where $t_{ij} = ([\ddot{T}^-_{ij}, \ddot{T}^+_{ij}], [\ddot{I}^-_{ij}, \ddot{I}^+_{ij}], [\ddot{F}^-_{ij}, \ddot{F}^+_{ij}])$, ($i = 1, 2, ..., m$; $j = 1, 2, ..., n$). Here, we have

$t_{ij} = r_{ij}$, if j is benefit type, (3.3)

$t_{ij} = \breve{r}_{ij}$, if j is cost type (3.4)

where, $\breve{r}_{ij}$ is the complement of $t_{ij}$.

**Step 3. Determination of unknown weights of the attributes**

In the decision making environment, we assume that the weights of the attributes are unknown to the expert and generally they are not identical. We use maximizing deviation method of Wang [29] to derive unknown attribute weights. The concept of maximizing deviation method is presented as follows. If an attribute has a small effect on the alternatives then the attribute value should be assigned with a small weight and the attribute which creates bigger deviation should be assigned with a bigger weight. However, if an attribute has very small or no effect on the alternatives then the weight of such attribute may be taken as zero [29].

The deviation values of alternatives $h_i$ to all other alternatives with respect to attribute $k_j$ can be formulated as $\Psi_{ij}(w_j) = \sum_{s=1}^{m} \Re(t_{ij}, t_{sj})\, w_j$, then $\Psi_j(w_j) = \sum_{i=1}^{m} \Re_{ij}(w_j) = \sum_{i=1}^{m}\sum_{s=1}^{m} \Re(t_{ij}, t_{sj})\, w_j$ represents total deviation of all alternatives to the other alternatives for the attribute $k_j$. $\Psi(w_j) = \sum_{j=1}^{n} \Re_j(w_j) =$ $\sum_{j=1}^{n}\sum_{i=1}^{m}\sum_{s=1}^{m}\Re(t_{ij},t_{sj})w_j$ denotes the deviation of all attributes for all alternatives to the other alternatives. Now we construct the optimization model [29] as given below.

Maximize $\Psi(w_j) = \sum_{j=1}^{n}\sum_{i=1}^{m}\sum_{s=1}^{m}\Re(t_{ij},t_{sj})w_j$

Subject to $\sum_{j=1}^{n}w_j^2 = 1$, $w_j \geq 0$, j = 1, 2, …, n. (3.5)

Solving the above model, we obtain attribute weight [29] as follows:

$$w_j = \frac{\sum_{i=1}^{m}\sum_{s=1}^{m}\Re(t_{ij},t_{sj})}{\sqrt{\sum_{j=1}^{n}\sum_{i=1}^{m}\sum_{s=1}^{m}\Re^2(t_{ij},t_{sj})}}, \; j = 1, 2, \ldots, n. \quad (3.6)$$

Then, the normalized attribute weight is obtained as

$$w_j = \frac{\sum_{i=1}^{m}\sum_{s=1}^{m}\Re(t_{ij},t_{sj})}{\sum_{j=1}^{n}\sum_{i=1}^{m}\sum_{s=1}^{m}\Re(t_{ij},t_{sj})}, \; j = 1, 2, \ldots, n. \quad (3.7)$$

**Step 4. Determination of interval - valued neutrosophic ideal solution**

We determine the interval - valued neutrosophic ideal solution $Z^* = (z_1^*, z_2^*, \ldots, z_n^*)$ [7] as given below.

$$z_j^* = ([1, 1], [0, 0], [0, 0]), \; j = 1, 2, \ldots, n. \quad (3.8)$$

The virtual interval - valued neutrosophic ideal solution $Z^* = (\mu_1^*, \mu_2^*, \ldots, \mu_n^*)$ [7] can also be obtained by identifying the best values for each attribute from all alternatives as shown below.

$$\mu_j^* = (\gamma_j^*, \delta_j^*, \lambda_j^*) \quad (3.9)$$

where, $\gamma_j^* = [\gamma_j^{L*}, \gamma_j^{U*}] = [\underset{i}{Max}\,\ddot{T}_{ij}^-, \underset{i}{Max}\,\ddot{T}_{ij}^+]$; $\delta_j^* = [\delta_j^{L*}, \delta_j^{U*}] = [\underset{i}{Min}\,\ddot{I}_{ij}^-, \underset{i}{Min}\,\ddot{I}_{ij}^+]$; $\lambda_j^* = [\lambda_j^{L*}, \lambda_j^{U*}] = [\underset{i}{Min}\,\ddot{F}_{ij}^-, \underset{i}{Min}\,\ddot{F}_{ij}^+]$.

**Step 5. Calculation of the weighted projection**

The weighted projection of the alternative $h_i$ (i = 1, 2, …, m) on the ideal solution $Z^*$ is defined as follows:

$$Proj_w(g_i)_{Z^*} = \frac{1}{\|Z^*\|_w}\sum_{j=1}^{n}w_j^2(\ddot{T}_{ij}^-\gamma_j^{L*} + \ddot{T}_{ij}^+\gamma_j^{U*} + \ddot{I}_{ij}^-\delta_j^{L*} + \ddot{I}_{ij}^+\delta_j^{U*} + \ddot{F}_{ij}^-\lambda_j^{L*} + \ddot{F}_{ij}^+\lambda_j^{U*})$$

$$= \frac{\sum_{j=1}^{n}w_j^2(\ddot{T}_{ij}^-\gamma_j^{L*} + \ddot{T}_{ij}^+\gamma_j^{U*} + \ddot{I}_{ij}^-\delta_j^{L*} + \ddot{I}_{ij}^+\delta_j^{U*} + \ddot{F}_{ij}^-\lambda_j^{L*} + \ddot{F}_{ij}^+\lambda_j^{U*})}{\sqrt{\sum_{j=1}^{n}w_j^2((\gamma_j^{L*})^2 + (\gamma_j^{U*})^2 + (\delta_j^{L*})^2 + (\delta_j^{U*})^2 + (\lambda_j^{L*})^2 + (\lambda_j^{U*})^2)}} \quad (3.10)$$

**Step 6. Ranking of the alternatives**

Rank the alternatives $h_i$ ( i = 1, 2, …, m) according to the weighted projection $Proj_w(h_i)_{Z^*}$ and bigger value of $Proj_w(h_i)_{Z^*}$ reflects the better alternative.

### 3.1 Algorithm 1.

An algorithm for MADM problems with interval valued neutrosophic information based on weighted projection method is provided in the following steps:

**Step 1.** The expert provides his/ her interval – valued neutrosophic decision matrix $D_{\tilde{N}}$ by Eq. (3.1).

**Step 2.** The decision matrix $D_{\tilde{N}}$, in Eq. (3.1) is standardized as shown, $D_{\hat{S}} = \langle t_{\hat{S}_{ij}} \rangle_{m\times n}$ in Eq. (3.2) by using Eqs. (3.3) – (3.4).

**Step 3.** The unknown weight of the attribute $w_j$, (j = 1, 2, …, n) is obtained by utilizing Eq. (3.7).

**Step 4.** The interval – valued neutrosophic ideal solution $Z^*$ is determined from the standardize decision matrix in Eq. (3.2).

**Step 5.** Determine the weighted projection $Proj_{\,w}(h_i)_{Z^*}$ using Eq. (3.10).

**Step 6.** Rank the alternatives $h_i$ (i = 1, 2, …, m) based on $Proj_{\,w}(h_i)_{Z^*}$ and select the best one.

**Step 7.** End.

### 3.2 Extension

**An approach for solving interval - valued neutrosophic MADM problems based on angle cosine and projection method**

The angle cosine [27] between the alternative $h_i$ (i = 1, 2, …, m) and the ideal solution $Z^*$ is defined as follows:

$$Cos\,(h_i, Z^*) = \frac{\sum_{j=1}^{n}(\ddot{T}_{ij}^{-}\gamma_j^{L*} + \ddot{T}_{ij}^{+}\gamma_j^{U*} + \ddot{I}_{ij}^{-}\delta_j^{L*} + \ddot{I}_{ij}^{+}\delta_j^{U*} + \ddot{F}_{ij}^{-}\lambda_j^{L*} + \ddot{F}_{ij}^{+}\lambda_j^{U*})}{\sqrt{\sum_{j=1}^{n}((\ddot{T}_{ij}^{-})^2 + (\ddot{T}_{ij}^{+})^2 + (\ddot{I}_{ij}^{-})^2 + (\ddot{I}_{ij}^{+})^2 + (\ddot{F}_{ij}^{-})^2 + (\ddot{F}_{ij}^{+})^2)}\sqrt{\sum_{j=1}^{n}((\gamma_j^{L*})^2 + (\gamma_j^{U*})^2 + (\delta_j^{L*})^2 + (\delta_j^{U*})^2 + (\lambda_j^{L*})^2 + (\lambda_j^{U*})^2)} \tag{3.11}$$

Now we propose the direction indicator $\xi$ $(0 \le \xi \le 1)$ to convert the direction closeness and magnitude closeness into relative closeness $\rho_i$. If the DM gives more interest on direction, then he or she provides bigger value to $\xi$. Otherwise, smaller value of $\xi$ is provided if the magnitude is much more important to the DM [23].

Therefore, the relative closeness [23] for selecting the best alternative is given as follows:

$$\rho_i = \xi\, Cos\,(h_i, Z^*) + (1 - \xi)\, Proj\,(h_i)_{Z^*} \tag{3.12}$$

The bigger value of $\rho_i$ gives the better alternative.

### 3.3 Algorithm 2.

An algorithm for MADM problem with interval valued neutrosophic information based on angle cosine and projection method can be demonstrated as follows:

**Step 1.** The expert presents the decision matrix $D_{\tilde{N}}$ as shown in Eq. (3.1).

**Step 2.** Utilize Eqs. (3.3) – (3.4) to standardize $D_{\tilde{N}}$ into $D_{\hat{S}} = \langle t_{\hat{S}_{ij}} \rangle_{m\times n}$.

**Step 3.** Define the ideal solution $Z^*$.

**Step 4.** Determine the angle cosine between the individual decision and the ideal decision $Z^*$ by utilizing Eq. (3.11).

**Step 5.** Find the projection measure of individual decision and the ideal decision $Z^*$ by using Eq. (3.10).

**Step 6.** Calculate the relative closeness $\rho_i$ in Eq. (3.12) by combining angle cosine and projection with direction indicator $\xi$.

**Step 7.** Rank the alternatives according to the decreasing order of the relative closeness $\rho_i$ and choose the most suitable alternative (s).

**Step 8.** End.

## 4. A numerical example

In this Section, we adapt an illustrative example from Dey et al. [13] for weaver selection in Khadi Institution where the information about attributes is expressed by linguistic variables. Consider a Khadi Institution wants to recruit two most competent weavers from a panel of three weavers $h_1$, $h_2$, $h_3$. Seven main attributes for weaver selection are: Skill ($k_1$); Previous experience ($k_2$); Honesty ($k_3$); Physical fitness ($k_4$); Locality of the weaver ($k_5$); Personality ($k_6$); Economic condition of the weaver ($k_7$) [30]. The Khadi Institution hire a Khadi expert to choose the desirable weavers based on the seven attributes. The evaluation information of an alternative $h_i$ ($i = 1, 2, 3$) with respect to seven attributes are provided by the Khadi expert in terms of linguistic variables as shown in the Table 2. It is to be noted that the seven attributes are of benefit type and the weights of the attributes are calculated by using maximizing deviation method.

### 4.1 Method 1

The procedure for weaver selection based on weighted projection method is presented by the following steps:

**Step 1:** We transform the linguistic decision matrix as shown in Table 2 into interval – valued neutrosophic decision matrix by means of Table 1.

**Table 2.** *Linguistic decision matrix*

| | $k_1$ | $k_2$ | $k_3$ | $k_4$ | $k_5$ | $k_6$ | $k_7$ |
|---|---|---|---|---|---|---|---|
| $h_1$ | G | G | VG | VG | VG | M | MG |
| $h_2$ | VG | VG | MG | G | VG | MG | ML |
| $h_3$ | G | VG | G | MG | G | G | G |

**Step 2:** Then the linguistic decision matrix is transformed into interval – valued neutrosophic decision matrix by using Table 1 as given below (see Table 3).

**Table 3.** *Interval – valued neutrosophic decision matrix*

C=

| | | | | | | |
|---|---|---|---|---|---|---|
| [[0.6,0.75],[0.1,0.2],[0.2,0.25]] | [[0.6,0.75],[0.1,0.2],[0.2,0.25]] | [[0.75,0.95],[0.1,0.15],[0.1,0.2]] | [[0.75,0.95],[0.1,0.15],[0.1,0.2] | [[0.75,0.95],[0.1,0.15],[0.1,0.2] | [[0.4,0.5],[0.2,0.3],[0.35,0.45]] | [[0.5,0.6],[[0..2,0.25],[0.25,0..35]] |
| [[0.75,0.95],[0.1,0.15],[0.1,0.2]] | [[0.75,0.95],[0.1,0.15],[0.1,0.2]] | [[0.5,0.6],[0.2,0.25],[0.25,0.35] | [[0.6,0.75],[0.1,0.2],[0.2,0.25]] | [[0.75,0.95],[0.1,0.15],[0.1,0.2]] | [[0.5,0.6],[0.2,0.25],[0.25,0.35]] | [[0.3,0.4],[0.15,0.2],[0.45,0.55] |
| [[0.6,0.75],[0.1,0.2],[0.2,0.25]] | [[0.75,0.95],[0.1,0.15],[0.1,0.2]] | [[0.6,0.75],[0.1,0.2],[0.2,0.25]] | [[0.5,0.6],[0.2,0.25],[0.25,0.35] | [[0.6,0.75],[0.1,0.2],[0.2,0.25]] | [[0.6,0.75],[0.1,0.2],[0.2,0.25]] | [[0.6,0.75],[0.1,0.2],[0.2,0.25]] |

**Step 3:** We employ Euclidean distance measure to get $\Re(t_{ij}, t_{sj})$, i = t = 1, 2, …, m; j = 1, 2, …, n and the normalized weights of the attributes are obtained as given below.

$w_1 = w_2 = 0.096$, $w_3 = w_4 = 0.176$, $w_5 = 0.096$, $w_6 = 0.151$, $w_7 = 0.207$ such that $\sum_{j=1}^{7} w_j = 1$, $w_j \geq 0$, j = 1, 2, …, 7.

**Step 4:** The virtual interval - valued neutrosophic ideal solution are obtained as given below.

$\mu_1^+$ = ([0.75, 0.95], [0.1, 0.15], [0.1, 0.2]); $\mu_2^+$ = ([0.75, 0.95], [0.1, 0.15], [0.1, 0.2]); $\mu_3^+$ = ([0.75, 0.95], [0.1, 0.15], [0.1, 0.2]); $\mu_4^+$ = ([0.75, 0.95], [0.1, 0.15], [0.1, 0.2]); $\mu_5^+$ = ([0.75, 0.95], [0.1, 0.15], [0.1, 0.2]); $\mu_6^+$ = ([0.6, 0.75], [0.1, 0.2], [0.2, 0.25]); $\mu_7^+$ = ([0.6, 0.75], [0.1, 0.2], [0.2, 0.25]).

**Step 5:** The weighted projection $Proj_{\,w}(h_i)_{Z^*}$ of the alternative $h_i$ (i = 1, 2, 3) on $Z^*$ is calculated as follows:

$Proj_{\,w}(h_1)_{Z^*} = 0.4255$, $Proj_{\,w}(h_2)_{Z^*} = 0.3730$, $Proj_{\,w}(h_3)_{Z^*} = 0.3972$.

**Step 6:** We rank the alternatives (weavers) according to the descending order of $Proj_{\,w}(h_i)_{Z^*}$ (i = 1, 2, 3). Here, we observe that

$$Proj_{\,w}(h_1)_{Z^*} > Proj_{\,w}(h_3)_{Z^*} > Proj_{\,w}(h_2)_{Z^*}$$

Consequently, $h_1$, $h_3$ are the most desirable alternatives for the Khadi Institution.

**Note 1:** We now compare our proposed weighted projection method with the methods investigated by Ye [25], Dey et al. [13], and Chi and Liu [7] and the obtained results are presented in the Table below.

*Table Results of different measure methods*

| Method | Measure value | Ranking order |
|---|---|---|
| Proposed method | Proj $_{w}(h_1)_{Z^*}$ = 0.4255, Proj $_{w}(h_2)_{Z^*}$ = 0.3730, Proj $_{w}(h_3)_{Z^*}$ = 0.3972 | $h_1 > h_3 > h_2$ |
| Ye [25] | Proj $(h_1)_{Z^*}$ = 2.87, Proj $(h_2)_{Z^*}$ = 2.777, Proj $(h_3)_{Z^*}$ = 2.739 | $h_1 > h_2 > h_3$ |
| Dey et al. [13] | $R_1$ = 0.077209, $R_2$ = 0.056516, $R_3$ = 0.056571 | $h_1 > h_3 > h_2$ |
| Chi and Liu [7] | $RCC_1$ = 0.6119, $RCC_2$ = 0.4231, $RCC_3$ = 0.4621 | $h_1 > h_3 > h_2$ |

### 4.2 Method 2

The procedure to get most desirable weaver(s) based on the combination of angle cosine and projection method is described by the following steps:

**Step 1:** Same as Step 1 of Method 1.

**Step 2:** Same as Step 2 of Method 1.

**Step 3:** Same as Step 3 of method 1.

**Step 4:** Same as Step 4 of method 1.

**Step 5:** The angle cosine between the alternative $h_i$ (i = 1, 2, 3) and the ideal solution $Z^*$ is calculated using Eq. (3.11) as given below.

$$Cos\,(h_1, Z^*) = 0.981,\ Cos\,(h_1, Z^*) = 0.962,\ Cos\,(h_1, Z^*) = 0.98.$$

**Step 6:** The projection measure between the alternative $h_i$ (i = 1, 2, 3) and the ideal solution $Z^*$ is calculated as follows.

$$Proj\ (h_1)_{Z^*} = 2.87,\ Proj\ (h_2)_{Z^*} = 2.777,\ Proj\ (h_3)_{Z^*} = 2.739.$$

**Step 7.** Combining angle cosine and projection measure with direction indicator $\xi$ = 0.5, the relative closeness $\rho_i$ (i = 1, 2, 3) is obtained as

$$\rho_1 = 1.926,\ \rho_2 = 1.87,\ \rho_3 = 1.86.$$

**Step 8:** The ranking order of the alternatives (weavers) is obtained as given below.

$$\rho_1 > \rho_2 > \rho_3$$

Therefore, $h_1$, $h_2$ are the most desirable weavers for Khadi Institution.

**Note 2:** However, if we take different direction indicators, the ranking order of the alternatives are obtained as given in Table 5.

Table 5. *Ranking order of the alternatives based on different direction indicators*

| Alternative | $\xi = 0$ | | $\xi = 0.25$ | | $\xi = 0.5$ | | $\xi = 0.75$ | | $\xi = 1$ | |
|---|---|---|---|---|---|---|---|---|---|---|
| | $\rho_i$ | Ranking | $\rho_i$ | Ranking | $\rho_i$ | Ranking | $\rho_i$ | Ranking | $\rho_i$ | Ranking |
| $h_1$ | 2.870 | 1 | 2.398 | 1 | 1.926 | 1 | 1.453 | 1 | 0.981 | 1 |
| $h_2$ | 2.777 | 2 | 2.323 | 2 | 1.870 | 2 | 1.416 | 3 | 0.962 | 3 |
| $h_3$ | 2.739 | 3 | 2.299 | 3 | 1.860 | 3 | 1.420 | 2 | 0.980 | 2 |

## 5. Conclusion

The paper is devoted to propose two new models for MADM problems with interval – valued neutrosophic information. In the decision making process, the rating of alternatives with respect to attributes are described by linguistic variables that can be represented by IVNNs. Since the weights of the attributes are fully unknown to the expert, we use maximization deviation method to find them. Then, we determine interval - valued neutrosophic ideal solutions. Finally, we develop weighted projection method to rank the alternatives. In this paper, we also propose an algorithm for MADM problems under interval neutrosophic environment via angle cosine and projection method. An illustrative example for weaver selection is solved to demonstrate the applicability of the proposed models. We also compare the obtained results with other existing approaches. In future, we will extend the concept to solve multi-attribute group decision making problems with interval – valued neutrosophic assessment. The authors hope that the proposed approach can be effective for dealing with diverse practical problems such as medical diagnosis, pattern recognition, management system, school choice, teacher selection, etc.

## References

1. F. Smarandache, A unifying field of logics. Neutrosophy: neutrosophic probability, set and logic, Rehoboth, American Research Press, 1998.
2. F. Smarandache, Linguistic paradoxes and tautologies, Libertas Mathematica, University of Texas at Arlington, IX (1999): 143-154.
3. F. Smarandache, Neutrosophic set – a generalization of intuitionistic fuzzy sets, International Journal of Pure and Applied Mathematics 24(3) (2005): 287-297.
4. F. Smarandache, Neutrosophic set – a generalization of intuitionistic fuzzy set, Journal of Defence Resources Management 1(1) (2010): 107-116.
5. H. Wang, F. Smarandache, Y.Q. Zhang, and R. Sunderraman, Single valued neutrosophic sets, Multispace and Multistructure 4 (2010): 410-413.
6. H. Wang, F. Smarandache, Y.Q. Zhang, and R. Sunderraman, Interval Neutrosophic Sets and Logic, Hexis, Arizona, 2005.
7. P. Chi, and P. Liu, An extended TOPSIS method for the multiple attribute decision making problems based on interval neutrosophic set, Neutrosophic Sets and Systems 1 (2013): 63-70.
8. H. Zhang, J. Wang, and X. Chen, Interval neutrosophic sets and their application in multicriteria decision making problem, The Scientific World Journal 2014, Article ID 645953, 15 pages, http://dx.doi.org/10.1155/2014/645953 .
9. S. Broumi, and F. Smarandache, Cosine similarity measure of interval neutrosophic sets, Neutrosophic Sets and Systems 5 (2014): 15-21.

10. J. Ye, Similarity measures between interval neutrosophic sets and their applications in multicriteria decision making, Journal of Intelligent & Fuzzy Systems 26 (2014): 165-172.
11. R. Sahin, and P. Liu, Maximizing deviation method for neutrosophic multiple attribute decision making with incomplete weight information, Neural Computing and Applications (2015), DOI: 10.1007/s00521-015-1995-8.
12. S. Pramanik, and K. Mondal, Interval neutrosophic multi-attribute decision – making based on grey relational analysis, Neutrosophic Sets and Systems 9 (2015): 13-22.
13. P.P. Dey, S. Pramanik, and B.C. Giri, An extended grey relational analysis based interval neutrosophic multi-attribute decision making for weaver selection, Journal of New Theory (9) (2015): 82-93.
14. S. Pramanik, and K. Mondal, Decision making based on some similarity measure under interval rough neutrosophic environment, Neutrosophic Sets and Systems 10 (2015): 46-57.
15. P.P. Dey, S. Pramanik, and B.C. Giri, An extended grey relational analysis based multiple attribute decision making in interval neutrosophic uncertain linguistic setting, Neutrosophic Sets and Systems 11 (2016): 21-30.
16. Z.S. Xu, and Q. L. Da, Projection method for uncertain multi-attribute decision making with preference information on alternatives, International Journal of Information Technology & Decision Making 3 (2004): 429-434.
17. Z. Xu, and H. Hu, Projection models for intuitionistic fuzzy multiple attribute decision making, International Journal of Information Technology & Decision Making 9(2) (2010): 267-280.
18. S. Zeng, T. Baležentis, J. Chen and G. Luo, A projection method for multiple attribute group decision making with intuitionistic fuzzy information, Informatica 24(3) (2013): 485-503.
19. Z. Yue, Approach to group decision making based on determining the weights of experts by using projection method, Applied Mathematical Modelling 36 (2012): 2900-2910.
20. Z. Yue, Application of the projection method to determine weights of decision makers for group decision making, Scientia Iranica 19(3) (2012): 872-878.
21. Z. Yue, An intuitionistic fuzzy projection-based approach for partner selection, Applied Mathematical Modelling 37 (2013): 9538-9551.
22. Y. Ju, and A. Wang, Projection method for multiple criteria group decision making with incomplete weight information in linguistic setting, Applied Mathematical Modelling 37 (2013): 9031-9040.
23. Q. Yang and P.A. Du, A straightforward approach for determining the weights of decision makers based on angle cosine and projection method, International Journal of Social, Behavioral, Educational, Economic, Business and Industrial Engineering 9(10) (2015): 3127-3133.
24. J. Ye, Simplified neutrosophic harmonic averaging projection-based method for multiple attribute decision-making problems, International Journal of Machine Learning & Cybernatics (2015), DOI: 10.1007/s13042-015-0456-0.
25. J. Ye, Interval neutrosophic multiple attribute decision – making method with credibility information, International Journal of Fuzzy Systems (2016), DOI: 10.1007/s40815-015-0122-4.
26. Z.S Xu, Theory method and Applications for multiple attribute decision-making with uncertainty, Tsinghua University Press, Beijing, 2004.
27. J. Ye, Vector similarity measures of simplified neutrosophic sets and their application in multicriteria decision making, International Journal of Fuzzy Systems 16(2) (2014): 204-211.
28. P. Majumder, and S.K. Samanta, On similarity and entropy of neutrosophic sets, Journal of Intelligent and Fuzzy Systems (2013), DOI: 10.3233/IFS-130810.
29. Y.M. Wang, Using the method of maximizing deviations to make decision for multi-indices, System Engineering and Electronics 7 (1998) 24-31.
30. P.P. Dey, S. Pramanik, and B.C. Giri, Multi-criteria group decision making in intuitionistic fuzzy environment based on grey relational analysis for weaver selection in Khadi Institution. Journal of Applied Quantitative Methods 10(4) (2015): 1-14.

**KANIKA MANDAL, KAJLA BASU**

Department of Mathematics, NIT Durgapur, West Bengal 713209, India.
E-mails: boson89@yahoo.com, kajla.basu@gmail.com

# Multi Criteria Decision Making Method in Neutrosophic Environment Using a New Aggregation Operator, Score and Certainty Function

## Abstract

Neutrosophic sets, being generalization of classic sets, fuzzy sets and intuitionistic fuzzy sets, can simultaneously represent uncertain, imprecise, incomplete, and inconsistent information existing in the real world. Neutrosophic theory has been developed in twenty first century and not much of arithmetic has been developed for this set. To solve any problem using neutrosophic data, it is desirable to have suitable operators, score function etc. Some operators like single valued neutrosophic weighted averaging (SVNWA) operator, single valued neutrosophic weighted geometric (SVNWG) operator are already defined in neutrosophic set (NS). In this paper an improved weighted average geometric (IWAG) operator which produces more meaningful results has been introduced to aggregate some real numbers and the same has been extended in neutrosophic environment. We further generalize this to include a wide range of aggregation operators for both real numbers and neutrosophic numbers. A new score function and certainty function have been defined which have some benefit compared to the existing ones. Further comparative study highlighting the benefit of this new approach of ranking in neutrosophic set has been presented. A multiple-attribute decision-making method is established on the basis of the proposed operator and newly defined score function.

## 1 INTRODUCTION

Neutrosophic set (NS), the generalization of classic set, fuzzy set, intutionistic fuzzy set, was first introduced by Smarandache. Smarandache [1] defined the degree of indeterminacy/neutrality as independent component in 1995 (published in 1998). NS can express uncertain, imprecise, incomplete and inconsistent information more precisely. How to aggregate information is an important problem in real management and decision process. Due to the complexity of management environments and decision problems, decision makers may provide their ratings or judgments to some certain degree, but it is possible that they are not so sure about their judgments. Namely, there may exist some uncertain, imprecise, incomplete, and inconsistent information, which are very important factors to be taken into account when trying to construct really adequate models and solutions of decision problems. Such kind of information is suitably expressed with neutrosophic fuzzy sets rather than exact numerical values, fuzzy or intuitionistic fuzzy. Thus, how to aggregate neutrosophic fuzzy information becomes an important part of multi-attribute decision-making with neutrosophic fuzzy sets.

In [17] Zhang-peng Tian et al. solved green product design selection problems using neutrosophic linguistic information. Xiao-hui Wu et al. [13] established ranking methods for simplified neutrosophic sets based on prioritized aggregation operators and cross-entropy measures to solve multi criteria decision making (MCDM) problem. Interval neutrosophic linguistic aggregation operators were developed and applied to the medical treatment selection process [16] by Yin-xiang Ma et al. In [20] Hong-yu Zhang, Jian-qiang Wang and Xiao-hong Chen defined some reliable operations for interval-valued neotrosophic sets. Based on those operators they also developed two aggregation operators which were applied to solve a MCDM problem. Jun Ye introduced in [4] single-valued neutrosophic hesitant fuzzy weighted averaging *(SVNHFWA)* operator and a single-valued neutrosophic hesitant fuzzy weighted geometric *(SVNHFWG)* operator and using those operator a multiple-attribute decision-making method was established. Peide Liu, Yanchang Chu, Yanwei Li, and Yubao Chen [7] presented some operational laws for neutrosophic numbers (NNs) based on Hamacher operations and proposed several averaging operators and applied them to group decision making. Broumi, Smarandache defined operations based on the arithmetic mean, geometrical mean and harmonic mean on interval-valued neutrosophic sets in [8].

In this paper we introduce an improved aggregating operator named improved weighted averaging geometric mean *(IWAGM)* for real numbers which produces more meaningful results and extend it for single valued neutrosophic set (SVNS) as improved single valued weighted averaging geometric *(ISVWAG)* operator. We further generalize the *IWAGM* operator and introduce generalized improved weighted averaging geometric mean *(GIWAGM)* which includes a wide range of weighted average geometric operators. Also we extend the *GIWAGM* for single valued neutrosophic numbers. We introduce a new score function and certainty function which are illustrated using simple examples and applied to numerical example. A comparative study highlighting the benefit of this new approach of ranking in NS has been discussed. An algorithm has been given to find optimum solution of multi-criteria decision making problem and a numerical illustration for a network problem has been presented.

The operators developed in this paper are original and have been developed for SVNS for the first time. Very few works have been done on averaging operator in SVNS. The score function and certainty function newly introduced in this paper have some benefits compared to the existing ones. The score function and certainty function defined earlier give same score function value for different neutrosophic numbers easily. But the newly proposed ones can remove this difficulty.

The rest of the paper is structured as follows: Section 2 introduces some concepts of neutrosophic sets and simplified neutrosophic sets. Section 3 describes weighted average mean *(WAM)* and weighted geometric mean *(WGM)* for real numbers and their limitations. In section 4, we define a new $IWAGM$ for real numbers and compare the results with the existing ones highlighting the improvement over the $WAM$ and *WGM*. $IWAGM$ has been extended in neutrosophic environment in section 5. In section 6, we generalize the $IWAGM$ introduced in section 4 and extend the generalization for neutrosophic numbers. Section 7 introduces a new approach defining a new score function and certainty function to compare the neutrosophic numbers. Why the approach is more realistic and meaningful is discussed in this section. Section 8 presents the algorithm for finding optimum alternative among alternatives in a decision making problem in neutrosophic environment using the introduced operator $IWAG$ in subsection 5.2 and the comparison approach defined in section 7. In section 9, a numerical example demonstrates the application and effectiveness of the proposed aggregation operator and comparison rules in decision-making problems. We conclude the paper in section 10.

# 2 NEUTROSOPHIC SETS

## 2.1 Definition

Let $U$ be an universe of discourse then the neutrosophic set $A$ is defined as $A = \{\langle x : T_A(x), I_A(x), F_A(x)\rangle, x \in U\}$, where the functions $T, I, F: U \rightarrow ]^{-}0, 1^{+}[$ define respectively the degree of membership (or Truth), the degree of indeterminacy and the degree of non-membership (or falsehood) of the element $x \in U$ to the set $A$ with the condition $^{-}0 \leq T_A(x) + I_A(x) + F_A(x) \leq 3^{+}$.

To apply neutrosophic set to science and technology, we consider the neutrosophic set which takes the value from the subset of $[0,1]$ instead of $]^{-}0,1^{+}[$; i.e., we consider SNS as defined by Ye in [3].

### 2.2 Simplified Neutrosophic Set

Let $X$ be a space of points (objects) with generic elements in $X$ denoted by $x$. A neutrosophic set $A$ in $X$ is characterized by a truth-membership function $T_A(x)$, an indeterminacy membership function $I_A(x)$, and a falsity-membership function $F_A(x)$, if the functions $T_A(x), I_A(x), F_A(x)$ are singleton subintervals/subsets in the real standard $[0,1]$, i.e., $T_A(x) : X \to [0,1]$, $I_A(x) : X \to [0,1]$ and $F_A(x) : X \to [0,1]$. Then a simplification of the neutrosophic set $A$ is denoted by $A = \{\langle x, T_A(x), I_A(x), F_A(x)\rangle, x \in X\}$.

### 2.3 Simplified neutrosophic set(SVNS)

Let $X$ be a space of points (objects) with generic elements in $X$ denoted by $x$. A SVNS $A$ in $X$ is characterized by a truth-membership function $T_A(x)$, an indeterminacy membership function $I_A(x)$ and a falsity-membership function $F_A(x)$, for each point $x \in X$, $T_A(x), I_A(x), F_A(x) \in [0,1]$. Therefore, a SVNS $A$ can be written as $A_{SVNS} = \{\langle x, T_A(x), I_A(x), F_A(x)\rangle, x \in X\}$. For two SVNS, $A_{SVNS} = \{\langle x, T_A(x), I_A(x), F_A(x)\rangle, x \in X\}$ and $B_{SVNS} = \{\langle x, T_B(x), I_B(x), F_B(x)\rangle, x \in X\}$, the following expressions are defined in [12] as follows:
$A_{NS} \subseteq B_{NS}$ if and only if $T_A(x) \leq T_B(x)$, $I_A(x) \geq I_B(x)$, $F_A(x) \geq F_B(x)$.
$A_{NS} = B_{NS}$ if and only if $T_A(x) = T_B(x)$, $I_A(x) = I_B(x)$, $F_A(x) = F_B(x)$.
$A^c = \langle x, F_A(x), 1 - I_A(x), T_A(x)\rangle$.

For convenience, a SVNS $A$ is denoted by $A = \langle T_A(x), I_A(x), F_A(x)\rangle$ for any $x$ in $X$. For two SVNSs $A$ and $B$, the operational relations (1), (2), (3) are defined by [3] and (4) by [2]
(1) $A + B = \langle\ T_A(x) + T_B(x) - T_A(x)T_B(x),\ I_A(x) + I_B(x) - I_A(x)I_B(x), F_A(x) + F_B(x) - F_A(x)F_B(x)\ \rangle$.
(2) $A.B = \langle T_A(x).T_B(x), I_A(x).I_B(x), F_A(x).F_B(x)\rangle$
(3) $A^\lambda = \langle T_A^\lambda(x), I_A^\lambda(x), F_A^\lambda(x)\rangle$.
(4) For any scalar $\lambda > 0$, $\lambda A = \langle \min(\lambda T_A(x), 1), \min(\lambda I_A(x), 1), \min(\lambda F_A(x), 1)\rangle$.

## 3 AGGREGATION OPERATORS

Aggregation operators are mathematical functions that are used to combine information. That is, they are used to combine $N$ data (for example, $N$ numerical values) in a single datum. In classical algebra $WAM$ and $WGM$ are very useful to combine n real numbers $a_1, a_2, \ldots, a_n$.
The WAM of n real numbers $a_1, a_2, \ldots, a_n$ with associated weights $w_1, w_2, \ldots, w_n$ respectively, $w_i \in [0,1]$ and $\sum w_i = 1$, is defined by $\sum_{i=1}^{n} a_i w_i$.
The $WGM$ of $n$ real numbers $a_1, a_2, \ldots, a_n$ with associated weights $w_1, w_2, \ldots, w_n$ respectively, $w_i \in [0,1]$ and $\sum w_i = 1$, is defined by $\prod_{i=1}^{n} a_i^{w_i}$.

## 3.1 Some limitations of $WAM$ and $WGM$

The result of an aggregation operator is meaningful if its value tends to one or some number(s) (among those to be combined) whose weight(s) is on the higher side. They do not correctly aggregate the information, if the aggregated value does not tend towards maximum arguments or does not lie between the maximum and minimum arguments. Let us consider some cases.
Example. Case 1: Take two real numbers 0.0001 and 1 with their weights $w_1 = 0.9$, $w_2 = 0.1$ respectively. Then $WAM = 0.10009$, $WGM = 0.000251$. Case 2: Again take 0.0001 and 1 with their weights $w_1 = 0.1$, $w_2 = 0.9$ respectively. Then $WAM = 0.90001$, $WGM = 0.398107$.
From these results we observe that from the first case the value of $WGM$ is more close to the number whose weight is maximum than $WAM$. So in this case, $WGM$ aggregates the numbers more close to the highest weighted number. On the other side in the second case the value of $WAM$ is nearest to the maximum weighted number whereas $WGM$ is close to 0.0001, the minimum weighted number. Here $WAM$ value is more meaningful. The examples show that $WAM$ and $WGM$ operators may not simultaneously give meaningful result while aggregating the information. Now we propose a new aggregation operator that always gives a moderate value close to the maximum weighted number.

# 4 THE NEWLY PROPOSED WEIGHTED MEAN

Let $a_1, a_2, \ldots, a_n$ are $n$ real numbers with associated weights $w_1, w_2, \ldots, w_n$ respectively, $w_i \in [0,1]$ and $\sum w_i = 1$. Then we define improved weighted average geometric mean ($IWAGM$) as

$$IWAGM(a_1, a_2, \ldots, a_n) = \sum_{i=1}^{n} a_i^{\frac{1}{2}} w_i \prod_{i=1}^{n} a_i^{\frac{w_i}{2}} \tag{1}$$

## 4.1 Properties

Let $a_1, a_2, \ldots, a_n$ are $n$ real numbers. Then the aggregated result of the $IWAGM$ operator clearly satisfies desired properties of an aggregation operator :

(1). **Idempotency:** let $a_i$ $(i = 1, 2, \ldots, n)$ be a collection of real numbers. If all $a_i$ $(i = 1, 2, \ldots, n)$ are equal, that is, $a_i = a$, for all$(i = 1, 2, \ldots, n)$, then $IWAGM(a_1, a_2, \ldots, a_n) = a^{\frac{1}{2}} \sum_{i=1}^{n} w_i \prod_{i=1}^{n} a^{\frac{w_i}{2}} = a$

(2). **Boundedness:** If $a^- = \min_i a_i$ and $a^+ = \max_i a_i$, $\sum_{i=1}^{n} (a^-)^{\frac{1}{2}} w_i \prod_{i=1}^{n} (a^-)^{\frac{w_i}{2}} \leq \sum_{i=1}^{n} a_i^{\frac{1}{2}} w_i \prod_{i=1}^{n} a_i^{\frac{w_i}{2}} \leq \sum_{i=1}^{n} (a^+)^{\frac{1}{2}} w_i \prod_{i=1}^{n} (a^+)^{\frac{w_i}{2}}$,
i.e. $a^- \leq IWAGM(a_1, a_2, \ldots, a_n) \leq a^+$.

(3). **Symmetry or commutativity:** The order of the arguments has no influence on the result. For every permutation $\sigma$ of $1, 2, \ldots, n$ the operator satisfies $IWAGM(a_{\sigma(1)}, a_{\sigma(2)}, \ldots, a_{\sigma(n)}) = IWAGM(a_1, a_2, \ldots, a_n)$

(4). **Monotonicity :** If $a_i \leq a_i^*$ for all $(i = 1, 2, \ldots, n)$,
$\sum_{i=1}^{n} a_i^{\frac{1}{2}} w_i \prod_{i=1}^{n} a_i^{\frac{w_i}{2}} \leq \sum_{i=1}^{n} (a_i^*)^{\frac{1}{2}} w_i \prod_{i=1}^{n} (a_i^*)^{\frac{w_i}{2}}$.
So $IWAGM(a_1, a_2, \ldots, a_n) \leq IWAGM(a_1^*, a_2^*, \ldots, a_n^*)$.

## 4.2 Theorem

For $n$ real numbers $a_1, a_2, \ldots, a_n$,
$WGM(a_1, a_2, \ldots, a_n) \leq IWAGM(a_1, a_2, \ldots, a_n) \leq WAM(a_1, a_2, \ldots, a_n)$.

**Proof:** We know $WAM$ of some real numbers always greater than or equal to $WGM$ of those real numbers.
So if we consider $n$ numbers $a_1^{\frac{1}{2}}, a_2^{\frac{1}{2}}, \ldots, a_n^{\frac{1}{2}}$ with their weights $w_1, w_2, \ldots, w_n$ respectively, then
$\sum_{i=1}^{n} a_i^{\frac{1}{2}} w_i \geq \prod_{i=1}^{n} a_i^{\frac{w_i}{2}}$.
Now, $\frac{\sum_{i=1}^{n} a_i^{\frac{1}{2}} w_i \prod_{i=1}^{n} a_i^{\frac{w_i}{2}}}{\prod_{i=1}^{n} a_i^{w_i}} = \frac{\sum_{i=1}^{n} a_i^{\frac{1}{2}} w_i}{\prod_{i=1}^{n} a_i^{\frac{w_i}{2}}} \geq 1$.
i.e.,

$$WGM(a_1, a_2, \ldots, a_n) \leq IWAGM(a_1, a_2, \ldots, a_n) \tag{2}$$

Again we know if $a_1, a_2, \ldots, a_n$ be positive real numbers, not all equal,
$w_1, w_2, \ldots, w_n$ be positive real numbers such that $\sum_{i=1}^{n} w_i = 1$ and m is rational,

lies between 0 and 1, $\sum_{i=1}^{n} w_i a_i^m \leq (\sum_{i=1}^{n} w_i a_i)^m$.
Taking $m = \frac{1}{2}$,

$$\sum_{i=1}^{n} w_i a_i^{\frac{1}{2}} \leq (\sum_{i=1}^{n} w_i a_i)^{\frac{1}{2}} \tag{3}$$

Also, $\prod_{i=1}^{n} a_i^{w_i} \leq \sum_{i=1}^{n} a_i w_i$
Taking square root in both side,

$$\prod_{i=1}^{n} a_i^{\frac{w_i}{2}} \leq (\sum_{i=1}^{n} a_i w_i)^{\frac{1}{2}} \tag{4}$$

Multiplying (3) and (4), we get

$$IWAGM(a_1, a_2, \ldots, a_n) \leq WAM(a_1, a_2, \ldots, a_n) \tag{5}$$

So combining (2) and (5), we get our proposed result.

### 4.3 Meaningful advantage of the proposed operator

Using the newly introduced operator, the aggregated results of the numbers with their weightage given in subsection 3.1, are given below: For case 1, $IWAGM(0.0001, 1) = 0.001728$ , and for case 2, $IWAGM(0.0001, 1) = 0.5684$. So in both the cases the newly introduced operator gives a moderate value close to the maximum weighted number. $WAM$ and $WGM$ may not simultaneously give meaningful result for all the numbers; but result from the proposed operator is meaningful since it holds the relation (2) and (5). In fact the new operator improves both the $WAM$ and $WGM$ and gives a moderate, meaningful value.

## 5 WEIGHTED AGGREGATION OPERATORS IN NEUTROSOPHIC ENVIRONMENT

### 5.1 Extension of $WAM$ and $WGM$ of classical algebra in neutrosophic set

In neutrosophic environment $SVNWA$ and $SVNWG$, the most well known aggregation operators, are the extension of $WAM$ and $WGM$ of classical algebra.

#### 5.1.1 Definition I

Let $A_i = (T_{A_i}(x), I_{A_i}(x), F_{A_i}(x))$ $(i = 1, 2, \ldots, n)$ be a collection of SVNSs. A mapping $F_w : SVNS^n \to SVNS$ is called single valued neutrosophic weighted averaging operator of dimension $n$ if it satisfies $F_w(A_1, A_2, \ldots, A_n) = \sum_{i=1}^{n} w_i A_i$, where $w = (w_1, w_2, \ldots, w_n)^T$ is the weight vector of $A_i$ $(i = 1, 2, \ldots, n)$, $w_i \in [0, 1]$ and $\sum w_i = 1$.

#### 5.1.2 Definition II

Let $A_i = (T_{A_i}(x), I_{A_i}(x), F_{A_i}(x))$ $(i = 1, 2, \ldots, n)$ be a collection of SVNSs. A mapping $F_w : SVNS^n \to SVNS$ is called SVNG operator of dimension $n$ if it satisfies $F_w(A_1, A_2, \ldots, A_n) = \prod_{i=1}^{n} A_i^{w_i}$.

### 5.2 Extension of proposed aggregation operator in neutrosophic set

Let $A_i = (T_A(i), I_A(i), F_A(i))$ $(i = 1, 2, \ldots, n)$ be a collection of SVNSs. Then we define improved single valued weighted averaging geometric $(ISVWAG)$ operator as

$$ISVWAG(A_1, A_2, \ldots, A_n) = \sum_{i=1}^{n} A_i^{\frac{1}{2}} w_i \prod_{i=1}^{n} A_i^{\frac{w_i}{2}} \tag{6}$$

#### 5.2.1 Properties

Let $A_i = (T_{A_i}(x), I_{A_i}(x), F_{A_i}(x))$ $(i = 1, 2, \ldots, n)$ be a collection of SVNSs. Then the aggregated result of the $ISVNWAG$ operator is also a single valued neutrosophic number (SVNN) and satisfies the desired properties of an aggregation operator.
To prove the properties we first prove a lemma.

#### 5.2.2 Lemma 1

Let $A_1 = (T_{A_1}(x), I_{A_1}(x), F_{A_1}(x))$, $A_2 = (T_{A_2}(x), I_{A_2}(x), F_{A_2}(x))$, $B_1 = (T_{B_1}(x), I_{B_1}(x), F_{B_1}(x))$, $B_2 = (T_{B_2}(x), I_{B_2}(x), F_{B_2}(x))$ are SVNNs such that $A_1 \supseteq B_1$, $A_2 \supseteq B_2$. Then $(A_1 + A_2) \supseteq (B_1 + B_2)$. i.e., $(T_{A_1} + T_{A_2} - T_{A_1}T_{A_2}) \geq (T_{B_1} + T_{B_2} - T_{B_1}T_{B_2})$, $(I_{A_1} + I_{A_2} - I_{A_1}I_{A_2}) \leq (I_{B_1} + I_{B_2} - I_{B_1}I_{B_2})$, $(F_{A_1} + F_{A_2} - F_{A_1}F_{A_2}) \leq (F_{B_1} + F_{B_2} - F_{B_1}F_{B_2})$.

**Proof:** Let X be the universe. For each point $x \in X$, $T_A(x), I_A(x), F_A(x) \in [0,1]$. Now it is given that $A_1 \supseteq B_1$, $A_2 \supseteq B_2$, i.e., for each value of $x \in X$, $T_{A_1}(x) \geq T_{B_1}(x)$ and $T_{A_2}(x) \geq T_{B_2}(x)$. Let $x_1$ be an arbitrary point in X. So $T_{A_1}(x_1) \geq T_{B_1}(x_1)$ and $T_{A_2}(x_1) \geq T_{B_2}(x_1)$. Also $T_{A_1}(x_1), T_{A_2}(x_1), T_{B_1}(x_1), T_{B_2}(x_1) \in [0,1]$. $\cos x$ for $x \in [0, \frac{\pi}{2}]$ is a continuous function. i.e., $\cos x$ assumes every value in $[0,1]$. So we can consider $T_{A_1}(x_1) = \cos\phi_1$, $T_{A_2}(x_1) = \cos\phi_2$, $T_{B_1}(x_1) = \cos\theta_1$, $T_{B_2}(x_1) = \cos\theta_2$, for some $\phi_1, \phi_2, \theta_1, \theta_2 \in [0, \frac{\pi}{2}]$.
Now $T_{A_1}(x_1)+T_{A_2}(x_1)-T_{A_1}(x_1)T_{A_2}(x_1) = \cos\phi_1+\cos\phi_2-\cos\phi_1\cos\phi_2 = \cos\phi_1+(1-\cos\phi_1)\cos\phi_2 = 1-2\sin^2\frac{\phi_1}{2}+2\sin^2\frac{\phi_1}{2}\cos\phi_2 = 1-2\sin^2\frac{\phi_1}{2}(1-\cos\phi_2) = 1-2\sin^2\frac{\phi_1}{2}.2\sin^2\frac{\phi_2}{2} = 1-4\sin^2\frac{\phi_1}{2}\sin^2\frac{\phi_2}{2}$.
Similarly, $T_{B_1}(x_1)+T_{B_2}(x_1)-T_{B_1}(x_1)T_{B_2}(x_1) = 1-4\sin^2\frac{\theta_1}{2}\sin^2\frac{\theta_2}{2}$.
Since $T_{A_1}(x_1) \geq T_{B_1}(x_1), T_{A_2}(x_1) \geq T_{B_2}(x_1)$,
$\cos\phi_1 \geq \cos\theta_1$. i.e., $-\cos\phi_1 \leq -\cos\theta_1$, $1-\cos\phi_1 \leq 1-\cos\theta_1$, i.e., $2\sin^2\frac{\phi_1}{2} \leq 2\sin^2\frac{\theta_1}{2}$.
Similarly, $2\sin^2\frac{\phi_2}{2} \leq 2\sin^2\frac{\theta_2}{2}$.
i.e., $4\sin^2\frac{\phi_1}{2}\sin^2\frac{\phi_2}{2} \leq 4\sin^2\frac{\theta_1}{2}\sin^2\frac{\theta_2}{2}$.
i.e., $1-4\sin^2\frac{\phi_1}{2}\sin^2\frac{\phi_2}{2} \geq 1-4\sin^2\frac{\theta_1}{2}\sin^2\frac{\theta_2}{2}$.
i.e.,

$$T_{A_1}(x_1)+T_{A_2}(x_1)-T_{A_1}(x_1)T_{A_2}(x_1) \geq T_{B_1}(x_1)+T_{B_2}(x_1)-T_{B_1}(x_1)T_{B_2}(x_1) \quad (7)$$

Since (7) is true for any $x_1 \in X$, $T_{A_1}(x)+T_{A_2}(x)-T_{A_1}(x)T_{A_2}(x) \geq T_{B_1}(x)+T_{B_2}(x)-T_{B_1}(x)T_{B_2}(x)$.
So it has been shown that $T_{A_1} \geq T_{B_1}$ and $T_{A_2} \geq T_{B_2}$ imply $[T_{A_1}+T_{A_2}-T_{A_1}T_{A_2}] \geq [T_{B_1}+T_{B_2}-T_{B_1}T_{B_2}]$. In the same way,
$I_{A_1} \leq I_{B_1}$ and $I_{A_2} \leq I_{B_2}$ imply $(I_{A_1}+I_{A_2}-I_{A_1}I_{A_2}) \leq (I_{B_1}+I_{B_2}-I_{B_1}I_{B_2})$ also $F_{A_1} \leq F_{B_1}$ and $F_{A_2} \leq F_{B_2}$ imply $(F_{A_1}+F_{A_2}-F_{A_1}F_{A_2}) \leq (F_{B_1}+F_{B_2}-F_{B_1}F_{B_2})$. The proof is generic as it is true for each and every value of the truth, indeterminacy and falsity membership functions and does not depend on the types of the functions (triangular, trapezoidal, piecewise linear or Gaussian).

On the basis of the basic operations of SVNSs described in subsection 2.3, the value of the truth, indeterminacy and falsity membership function in aggregated result belongs to $[0,1]$. So the aggregated operator is also a SVNN. We will prove that the $ISVNWAG$ operator has the following desired properties:

(1) **Idempotency:** let $A_i$ $(i = 1,2,\ldots,n)$ be a collection of SVNNs. If all $A_i$ $(i = 1,2,\ldots,n)$ are equal, that is, $A_i = A$, for all$(i = 1,2,\ldots,n)$, then $ISVWAG(A_1, A_2, \ldots, A_n) = A$

(2) **Boundedness:** Let $A^- = (T_{A^-}(x), I_{A^-}(x), F_{A^-}(x))$, and $A^+ = (T_{A^+}(x), I_{A^+}(x), F_{A^+}(x))$, where $T_{A^-}(x) = \min_i T_{A_i}(x)$, $I_{A^-}(x) = \max_i I_{A_i}(x)$, $F_{A^-}(x) = \max_i F_{A_i}(x)$ and $T_{A^+}(x) = \max_i T_{A_i}(x)$, $I_{A^+}(x) = \min_i I_{A_i}(x)$, $F_{A^+}(x) = \min_i F_{A_i}(x)$. So $A^- \subseteq A_i \subseteq A^+$ for all $(i = 1, 2, \ldots, n)$. Also $(T_{A^-}(x))^{0.5} w_i \leq (T_{A_i}(x))^{0.5} w_i \leq (T_{A^+}(x))^{0.5} w_i$, $(I_{A^-}(x))^{0.5} w_i \geq (I_{A_i}(x))^{0.5} w_i \geq (I_{A^+}(x))^{0.5} w_i$ and $(F_{A^-}(x))^{0.5} w_i \geq (F_{A_i}(x))^{0.5} w_i \geq (F_{A^+}(x))^{0.5} w_i$. So $(A^-)^{0.5} w_i \subseteq (A_i)^{0.5} w_i \subseteq (A^+)^{0.5} w_i$. By using lemma 1, $\sum_{i=1}^{n} (A^-)^{0.5} w_i \subseteq \sum_{i=1}^{n} A_i^{\frac{1}{2}} w_i \subseteq \sum_{i=1}^{n} (A^+)^{0.5} w_i$, i.e.,

$$(A^-)^{\frac{1}{2}} \subseteq \sum_{i=1}^{n} A_i^{\frac{1}{2}} w_i \subseteq (A^+)^{\frac{1}{2}} \tag{8}$$

Again similarly, $(A^-)^{\frac{w_i}{2}} \subseteq A_i^{\frac{w_i}{2}} \subseteq (A^+)^{\frac{w_i}{2}}$, i.e.,

$$(A^-)^{\frac{1}{2}} \subseteq \prod_{i=1}^{n} A_i^{\frac{w_i}{2}} \subseteq (A^+)^{\frac{1}{2}} \tag{9}$$

From (8)

$$T_{(A^-)^{\frac{1}{2}}} \leq T_{\sum_{i=1}^{n} A_i^{\frac{1}{2}} w_i} \leq T_{(A^+)^{\frac{1}{2}}} \tag{10}$$

From (9)

$$T_{(A^-)^{\frac{1}{2}}} \leq T_{\prod_{i=1}^{n} A_i^{\frac{w_i}{2}}} \leq T_{(A^+)^{\frac{1}{2}}} \tag{11}$$

Multiplying (10) and (11) we get, $T_{A^-} \leq T_{ISVWAG(A_1, A_2, \ldots, A_n)} \leq T_{A^+}$.
similarly, $I_{A^-} \geq I_{ISVWAG(A_1, A_2, \ldots, A_n)} \geq I_{A^+}$ and $F_{A^-} \geq F_{ISVWAG(A_1, A_2, \ldots, A_n)} \geq F_{A^+}$.
Thus $A^- \subseteq ISVWAG(A_1, A_2, \ldots, A_n) \subseteq A^+$.

(3) **Symmetry or commutativity:** The order of the arguments has no influence on the result. For every permutation $\sigma$ of $1, 2, \ldots, n$ the operator satisfies $ISVWAG(A_{\sigma(1)}, A_{\sigma(2)}, \ldots, A_{\sigma(n)}) = ISVWAG(A_1, A_2, \ldots, A_n)$

(4) **Monotonicity :**Let $A_i \subseteq A_i^*$ for all $(i = 1, 2, \ldots, n)$, then $T_{A_i}(x) \leq T_{A_i^*}(x)$, $I_{A_i}(x) \geq I_{A_i^*}(x)$ and $F_{A_i}(x) \geq F_{A_i^*}(x)$. i.e., $(T_{A_i}(x))^{0.5} w_i \leq (T_{A_i^*}(x))^{0.5} w_i$, $(I_{A_i}(x))^{0.5} w_i \geq (I_{A_i^*}(x))^{0.5} w_i$ and $(F_{A_i}(x))^{0.5} w_i \geq (F_{A_i^*}(x))^{0.5} w_i$. So $(A_i)^{0.5} w_i \subseteq (A_i^*)^{0.5} w_i$. Therefore from the lemma 1 $\sum_{i=1}^{n} A_i^{0.5} w_i \subseteq \sum_{i=1}^{n} (A_i^*)^{0.5} w_i$ and also since $\prod_{i=1}^{n} A_i^{\frac{w_i}{2}} \subseteq \prod_{i=1}^{n} A_i^{*\frac{w_i}{2}}$, $ISVWAG(A_1, A_2, \ldots, A_n)$
$\subseteq ISVWAG(A_1^*, A_2^*, \ldots, A_n^*)$.

# 6 GENERALIZATION OF $IWAGM$ AND ITS EXTENSION IN NEUTROSOPHIC ENVIRONMENT

We formulate a general operator in case of real numbers and extend it to neutrosophic set also.
Let $a_1, a_2, \ldots, a_n$ are $n$ real numbers with associated weights $w_1, w_2, \ldots, w_n$ respectively, $w_i \in [0,1]$ and $\sum w_i = 1$. Then we define generalized improved weighted averaging geometric mean ($GIWAGM$) as

$$GIWAGM(a_1, a_2, \ldots, a_n) = (\sum_{i=1}^{n} a_i^{\frac{1}{k}} w_i \prod_{i=1}^{n} a_i^{\frac{w_i}{k}})^{k/2} \tag{12}$$

where k is any real number. The equation (12) satisfy the desired properties of aggregation operator:

(1) **Idempotency:** let $a_i$ $(i = 1, 2, \ldots, n)$ be a collection of real numbers. If all $a_i$ $(i = 1, 2, \ldots, n)$ are equal, that is, $a_i = a$, for all$(i = 1, 2, \ldots, n)$, then $GIWAGM(a_1, a_2, \ldots, a_n) = (a^{\frac{1}{k}} \sum_{i=1}^{n} w_i \prod_{i=1}^{n} a^{\frac{w_i}{k}})^{k/2} = a$

(2) **Boundedness:** If $a^- = \min_i a_i$ and $a^+ = \max_i a_i$,
$(\sum_{i=1}^{n} (a^-)^{\frac{1}{k}} w_i \prod_{i=1}^{n} (a^-)^{\frac{w_i}{k}})^{k/2} \leq (\sum_{i=1}^{n} a_i^{\frac{1}{k}} w_i \prod_{i=1}^{n} a_i^{\frac{w_i}{k}})^{k/2} \leq (\sum_{i=1}^{n} (a^+)^{\frac{1}{k}}$ $w_i \prod_{i=1}^{n} (a^+)^{\frac{w_i}{k}})^{k/2}$, i.e. $a^- \leq GIWAGM(a_1, a_2, \ldots, a_n) \leq a^+$.

(3) **Symmetry or commutativity:** The order of the arguments has no influence on the result. For every permutation $\sigma$ of $1, 2, \ldots, n$ the operator satisfies $GIWAGM(a_{\sigma(1)}, a_{\sigma(2)}, \ldots, a_{\sigma(n)}) = GIWAGM(a_1, a_2, \ldots, a_n)$

(4) **Monotonicity :**If $a_i \leq a_i^*$ for all $(i = 1, 2, \ldots, n)$,
$(\sum_{i=1}^{n} a_i^{\frac{1}{k}} w_i \prod_{i=1}^{n} a_i^{\frac{w_i}{k}})^{k/2} \leq (\sum_{i=1}^{n} (a_i^*)^{\frac{1}{k}} w_i \prod_{i=1}^{n} (a_i^*)^{\frac{w_i}{k}})^{k/2}$.
So $GIWAGM(a_1, a_2, \ldots, a_n) \leq GIWAGM(a_1^*, a_2^*, \ldots, a_n^*)$.
And for neutrosophic sets let $A_i = (T_A(i), I_A(i), F_A(i))$ $(i = 1, 2, \ldots, n)$ be a collection of SVNSs with associated weights $w_1, w_2, \ldots, w_n$ respectively, $w_i \in [0,1]$ and $\sum w_i = 1$.. Then we define generalized improved single valued weighted averaging geometric (GISVWAG) operator as

$$GISVWAG(A_1, A_2, \ldots, A_n) = (\sum_{i=1}^{n} A_i^{\frac{1}{k}} w_i \prod_{i=1}^{n} A_i^{\frac{w_i}{k}})^{k/2} \tag{13}$$

where k is any real number. The equation (13) satisfy the desired properties of aggregation operator:

(1) **Idempotency:** let $A_i$ $(i = 1, 2, \dots, n)$ be a collection of SVNNs. If all $A_i$ $(i = 1, 2, \dots, n)$ are equal, that is, $A_i = A$, for all$(i = 1, 2, \dots, n)$, then $ISVWAG(A_1, A_2, \dots, A_n) = A$

(2) **Boundedness:** Let $A^- = (T_{A^-}(x), I_{A^-}(x), F_{A^-}(x))$, and $A^+ = (T_{A^+}(x), I_{A^+}(x), F_{A^+}(x))$, where $T_{A^-}(x) = \min_i T_{A_i}(x)$, $I_{A^-}(x) = \max_i I_{A_i}(x)$, $F_{A^-}(x) = \max_i F_{A_i}(x)$ and $T_{A^+}(x) = \max_i T_{A_i}(x)$, $I_{A^+}(x) = \min_i I_{A_i}(x)$, $F_{A^+}(x) = \min_i F_{A_i}(x)$. So $A^- \subseteq A_i \subseteq A^+$ for all $(i = 1, 2, \dots, n)$. Also $(T_{A^-}(x))^{1/k} w_i \le (T_{A_i}(x))^{1/k} w_i \le (T_{A^+}(x))^{1/k} w_i$, $(I_{A^-}(x))^{1/k} w_i \ge (I_{A_i}(x))^{1/k} w_i \ge (I_{A^+}(x))^{1/k} w_i$ and $(F_{A^-}(x))^{1/k} w_i \ge (F_{A_i}(x))^{1/k} w_i \ge (F_{A^+}(x))^{1/k} w_i$. So $(A^-)^{1/k} w_i \subseteq (A_i)^{1/k} w_i \subseteq (A^+)^{1/k} w_i$. By using lemma 1, $\sum_{i=1}^n (A^-)^{1/k} w_i \subseteq \sum_{i=1}^n A_i^{1/k} w_i \subseteq \sum_{i=1}^n (A^+)^{1/k} w_i$, i.e.,

$$(A^-)^{1/k} \subseteq \sum_{i=1}^{n} A_i^{1/k} w_i \subseteq (A^+)^{1/k} \tag{14}$$

Again similarly, $(A^-)^{\frac{w_i}{k}} \subseteq A_i^{\frac{w_i}{k}} \subseteq (A^+)^{\frac{w_i}{k}}$, i.e.,

$$(A^-)^{\frac{1}{k}} \subseteq \prod_{i=1}^{n} A_i^{\frac{w_i}{k}} \subseteq (A^+)^{\frac{1}{k}} \tag{15}$$

From (14)

$$T_{(A^-)^{\frac{1}{k}}} \le T_{\sum_{i=1}^n A_i^{\frac{1}{k}} w_i} \le T_{(A^+)^{\frac{1}{k}}} \tag{16}$$

From (15)

$$T_{(A^-)^{\frac{1}{k}}} \le T_{\prod_{i=1}^n A_i^{\frac{w_i}{k}}} \le T_{(A^+)^{\frac{1}{k}}} \tag{17}$$

Multiplying (16) and (17) we get, $T_{(A^-)^{\frac{2}{k}}} \le T_{\sum_{i=1}^n A_i^{\frac{1}{k}} w_i} T_{\prod_{i=1}^n A_i^{\frac{w_i}{k}}} \le T_{(A^+)^{\frac{2}{k}}}$.
i.e., $T_{A^-} \le T_{GISVWAG(A_1, A_2, \dots, A_n)} \le T_{A^+}$.
similarly, $I_{A^-} \ge I_{GISVWAG(A_1, A_2, \dots, A_n)} \ge I_{A^+}$ and $F_{A^-} \ge F_{GISVWAG(A_1, A_2, \dots, A_n)} \ge F_{A^+}$. So $A^- \subseteq GISVWAG(A_1, A_2, \dots, A_n) \subseteq A^+$.

(3) **Symmetry or commutativity:** The order of the arguments has no influence on the result. For every permutation $\sigma$ of $1, 2, \dots, n$ the operator satisfies $GISVWAG(A_{\sigma(1)}, A_{\sigma(2)}, \dots, A_{\sigma(n)}) = GISVWAG(A_1, A_2, \dots, A_n)$

(4) **Monotonicity :**Let $A_i \subseteq A_i^*$ for all $(i = 1, 2, \ldots, n)$, then $T_{A_i}(x) \leq T_{A_i^*}(x)$, $I_{A_i}(x) \geq I_{A_i^*}(x)$ and $F_{A_i}(x) \geq F_{A_i^*}(x)$. i.e., $(T_{A_i}(x))^{1/k}w_i \leq (T_{A_i^*}(x))^{1/k}w_i$, $(I_{A_i}(x))^{1/k}w_i \geq (I_{A_i^*}(x))^{1/k}w_i$ and $(F_{A_i}(x))^{1/k}w_i \geq (F_{A_i^*}(x))^{1/k}w_i$. So $(A_i)^{1/k}w_i \subseteq (A_i^*)^{1/k}w_i$. Therefore from the lemma 1, $\sum_{i=1}^{n} A_i^{1/k}w_i \subseteq \sum_{i=1}^{n}(A_i^*)^{1/k}w_i$ and also since $\prod_{i=1}^{n} A_i^{\frac{w_i}{k}} \subseteq \prod_{i=1}^{n} A_i^{*\frac{w_i}{k}}$, $GISVWAG(A_1, A_2, \ldots, A_n) \subseteq GISVWAG(A_1^*, A_2^*, \ldots, A_n^*)$.

Now if we put $k = 2$, (12) and (13) reduce to (1) and (6) respectively, i.e., the newly proposed operator for real numbers given in (1) is one of the particular cases of generalized operator (12) and similar for the case (6) and (13) also. For different values of $k$ it is possible to study these families individually.

# 7 COMPARISON APPROACH

## 7.1 Definition [20], [7]

Let $A$ and $B$ are two SVNN. Then the comparison approach based on score function (s), accuracy function (a) and certainty function (c) is given as follows:
(1) If $s(A) > s(B)$, then $A > B$.
(2) If $s(A) = s(B)$ and $a(A) > a(B)$, then $A > B$.
(3) If $s(A) = s(B)$ also $a(A) = a(B)$, but $c(A) > c(B)$, then $A > B$.
(4) If $s(A) = s(B)$, $a(A) = a(B)$ and $c(A) = c(B)$, then $A = B$.

## 7.2 Proposed score and certainty function

We introduce a new score function, accuracy function and certainty function to compare neutrosophic fuzzy numbers. According to the definition of score function as defined in [20], the larger the $T_A$ is, the greater the neutrosophic number is; the smaller the $I_A$ is, the greater the neutrosophic number is and the same holds for $F_A$ also. Based on the definition we give a new score function. Let $A = (T_A(x), I_A(x), F_A(x))$ be a neutrosophic number. The **score function of** $A$ is given by $s(A) = T_A(x)(1 + \sin(T_A(x)\frac{\pi}{2})) + \frac{1}{2(1+I_A(x))}(\cos(I_A(x)\frac{\pi}{2})) + \frac{1}{1+F_A(x)}(\cos(F_A(x)\frac{\pi}{2}))$. The accuracy function as defined in [7] is $a(A) = T_A(x) - F_A(x)$.
We define a new **certainty function** $c(A) = \frac{|cosT_A(x)\pi|+|cosI_A(x)\pi|+|cosF_A(x)\pi|}{3}$. In [7] Liu et al. gave the formula of score, accuracy function for a SVNN, A, as follows: Score function $s_1(A) = 2 + T_A(x) - I_A(x) - F_A(x)$, accuracy function $a_1(A) = T_A(x) - F_A(x)$. With these formulas in [20] Hong-yu Zhang, Jian-qiang

Wang, and Xiao-hong Chen added certainty function as $c_1(A) = T_A(x)$. The score function in [7] gives same value when $I_A(x)$ and $F_A(x)$ of a neutrosophic number is interchanged, i.e., different neutrosophic numbers can exist easily for which the given score function gives same value. For example $(0.2, 0.9, 0.1)$ and $(0.2, 0.1, 0.9)$ have same score function according [7] and [20]. But the newly introduced score function based on trigonometric function does not give same value for these neutrosophic sets. From another point of view as discussed in [5] the uncertainty is maximum (=1) at $(0.5, 0.5, 0.5)$, i.e., the certainty should be minimum (=0) at $(0.5, 0.5, 0.5)$ and the value of certainty increases if we increase or decrease any of truth, indeterminacy and falsity membership grade. But this property is not satisfied by the certainty function given in [20], whereas the newly proposed certainty function in sec 7 gives realistic result.

## 7.3 Comparison analysis using different examples

Table 1: Comparison analysis using different examples

| *Neutrosophic numbers* | *Method* | *Score value* | *Accuracy value* | *Certainty Value* | *Ranking order* |
|---|---|---|---|---|---|
| $A = (0.4, 0.3, 0.2)$ | *Existing* | $s_1(A) = 1.9$ | | | $A > B$ |
| $B = (0.4, 0.5, 0.6)$ | | $s_1(B) = 1.3$ | | | |
| | *Proposed* | $s(A) = 1.77$ | | | $A > B$ |
| | | $s(B) = 1.23$ | | | |
| $A = (1, 0, 1)$ | *Existing* | $s_1(A) = 2$ | | | $A > B$ |
| $B = (1, 1, 0.473)$ | | $s_1(B) = 1.527$ | | | |
| | *Proposed* | $s(A) = 2.5$ | $-$ | | $A < B$ |
| | | $s(B) = 2.5$ | $a(B) = 0.527$ | | |
| $A = (0.0867, 0.2867, 0.0867)$ | *Existing* | $s_1(A) = 1.7133$ | | | $A > B$ |
| $B = (0.3096, 0.5096, 0.0.3096)$ | | $s_1(B) = 1.4904$ | | | |
| | *Proposed* | $s(A) = 1.3599$ | $-$ | $c(A) = 0.8491$ | $A > B$ |
| | | $s(B) = 1.3599$ | $-$ | $c(B) = 0.38548$ | |

# 8 ALGORITHM FOR FINDING OPTIMUM ALTERNATIVE IN A MULTI-CRITERIA DECISION MAKING PROBLEM

Let $A_i, (i = 1, 2, \ldots, m)$ be m alternatives and $C_j, (j = 1, 2, \ldots, n)$ are n criteria. Assume that the weight of the criteria $C_j (j = 1, 2, \ldots, n)$, given by the decision-maker, is $w_j$, $w_j \in [0, 1]$ and $\sum_{j=1}^{n} w_j = 1$. The m options according to the n criterion are given below:

| | $C_1$ | $C_2$ | $C_3$ | $\ldots$ | $C_n$ |
|---|---|---|---|---|---|
| $A_1$ | $C_1^1$ | $C_2^1$ | $C_3^1$ | $\ldots$ | $C_n^1$ |
| $A_2$ | $C_1^2$ | $C_2^2$ | $C_3^2$ | $\ldots$ | $C_n^2$ |
| $A_3$ | $C_1^3$ | $C_2^3$ | $C_3^3$ | $\ldots$ | $C_n^3$ |
| $\vdots$ | $\vdots$ | $\vdots$ | $\vdots$ | $\vdots$ | $\vdots$ |
| $A_m$ | $C_1^m$ | $C_2^m$ | $C_3^m$ | $\ldots$ | $C_n^m$ |

where each $C_j^i$, $(i = 1, 2, \ldots, m)$ and $(j = 1, 2, \ldots, n)$ are in neutrosophic form and $C_j^i = \left\{T_{C_j}^i, I_{C_j}^i, F_{C_j}^i\right\}$ We propose a method to derive optimum alternative among the given alternatives through the algorithm given below:

**Step 1:** use the $ISVWAG$ operator given in (6) to combine n criteria for each alternative.

**Step 2:** calculate the score, accuracy and certainty function to compare the neutrosophic number as defined in section 7.

**Step 3:** Rank the alternatives.

# 9 NUMERICAL EXAMPLE

In a certain network, there are four options to go from one node to the other. Which path to be followed will be impacted by two benefit criteria $C_1$, $C_2$ and one cost criteria $C_3$ and the weight vectors are 0.35, 0.25 and 0.40 respectively. A decision maker evaluates the four options according to the three criteria mentioned above. We compare the proposed method with the existing methods in table 3 using the newly introduced approach to obtain the most desirable alternative from the decision matrix given in table 2.

Table 2: Decision matrix (information given by DM)

| | $c_1$ | $c_2$ | $c_3$ |
|---|---|---|---|
| $A_1$ | (0.4,0.2,0.3) | (0.4,0.2,0.3) | (0.2,0.2,0.5) |
| $A_2$ | (0.6,0.1,0.2) | (0.6,0.1,0.2) | (0.5,0.2,0.2) |
| $A_3$ | (0.3,0.2,0.3) | (0.5,0.2,0.3) | (0.5,0.3,0.2) |
| $A_4$ | (0.7,0,0.1) | (0.6,0.1,0.2) | (0.4,0.3,0.2) |

## 9.1 Comparison of aggregation operators using cosine similarity measure

To measure the similarity between two neutrosophic numbers we consider the cosine similarity measure as discussed by Jun Ye in [9] as follows:

Table 3: Result Comparison: the proposed method with the existing methods

| *Aggregation Operator* | *Aggregated Result* | *Score using existing method* | *Score using proposed formula* | *Ranking order in both approach* |
|---|---|---|---|---|
| *Single valued weighted average* | $SVWA(C_1^{(1)}, C_2^{(1)}, C_3^{(1)}) = (0.287, 0.187, 0.337)$ | $s_1(A_1) = 1.76$ | $s(A_1) = 1.46$ | $A_4 > A_2 > A_3 > A_1$ |
| | $SVWA(C_1^{(2)}, C_2^{(2)}, C_3^{(2)}) = (0.462, 0.134, 0.187)$ | $s_1(A_2) = 2.14$ | $s(A_2) = 2.007$ | |
| | $SVWA(C_1^{(3)}, C_2^{(3)}, C_3^{(3)}) = (0.373, 0.222, 0.238)$ | $s_1(A_3) = 1.912$ | $s(A_3) = 1.716$ | |
| | $SVWA(C_1^{(4)}, C_2^{(4)}, C_3^{(4)}) = (0.460, 0.142, 0.156)$ | $s_1(A_4) = 2.16$ | $s(A_4) = 2.03$ | |
| *Single valued weighted geometric* | $SVWG(C_1^{(1)}, C_2^{(1)}, C_3^{(1)}) = (0.303143, 0.2, 0.368011)$ | $s_1(A_1) = 1.735$ | $s(A_1) = 1.450532$ | $A_4 > A_2 > A_3 > A_1$ |
| | $SVWG(C_1^{(2)}, C_2^{(2)}, C_3^{(2)}) = (0.5578, 0.131951, 0.2)$ | $s_1(A_2) = 2.22$ | $s(A_2) = 2.211256$ | |
| | $SVWG(C_1^{(3)}, C_2^{(3)}, C_3^{(3)}) = (0.418141, 0.235216, 0.255085)$ | $s_1(A_3) = 1.92$ | $s(A_3) = 1.7845$ | |
| | $SVWG(C_1^{(4)}, C_2^{(4)}, C_3^{(4)}) = (0.538451, 0, 0.156917)$ | $s_1(A_4) = 2.38$ | $s(A_4) = 2.2798$ | |
| *Improved single valued weighted average geometric* | $ISVWAG(C_1^{(1)}, C_2^{(1)}, C_3^{(1)}) = (0.254226, 0.172108, 0.303)$ | $s(A_1) = 1.77$ | $s(A_1) = 1.44$ | $A_4 > A_2 > A_3 > A_1$ |
| | $ISVWAG(C_1^{(2)}, C_2^{(2)}, C_3^{(2)}) = (0.432056, 0.118963, 0.17)$ | $s_1(A_2) = 2.14$ | $s(A_2) = 1.96$ | |
| | $ISVWAG(C_1^{(3)}, C_2^{(3)}, C_3^{(3)}) = (0.338061, 0.201253, 0.21)$ | $s_1(A_3) = 1.92$ | $s(A_3) = 1.68$ | |
| | $ISVWAG(C_1^{(4)}, C_2^{(4)}, C_3^{(4)}) = (0.421219, 0, 0.13)$ | $s_1(A_4) = 2.28$ | $s(A_4) = 2.03$ | |

Let X be the universe and $A = \{\langle x_i, T_A(x_i), I_A(x_i), F_A(x_i)\rangle / x_i \in X\}$ and $B = \{\langle x_i, T_B(x_i), I_B(x_i), F_B(x_i)\rangle / x_i \in X\}$ are two SVNSs, then cosine similarity measure between A and B is

$$C(A,B) = \frac{1}{n}\sum_{i=1}^{n} \frac{T_A(x_i)T_B(x_i)+I_A(x_i)I_B(x_i)+F_A(x_i)F_B(x_i)}{\sqrt{(T_A(x_i))^2+I_A(x_i))^2+F_A(x_i))^2}\sqrt{(T_B(x_i))^2+I_B(x_i))^2+F_B(x_i))^2}}$$

Using the similarity measure formula comparison of aggregation operators are given in table 4:

## 9.2 Result discussion

The results given in table 4 show that all the aggregated results are more or less close to the corresponding maximum weighted neutrosophic number as similarity measure values are nearer to 1. Also it is observed that the proposed method gives almost same similarity measure value as the other existing methods as discussed in table 4. In other words, the newly introduced operator gives moderate and meaningful value similar to existing methods and close to the maximum weighted neutrosophic number.

Table 4: Comparison of aggregation operators using similarity measure

| *Alternative* | *Aggregation operator* | *Aggregated result* | *Corresponding maximum weighted number* | *Similarity measure value* |
|---|---|---|---|---|
| | $SVWA$ | $(0.287, 0.187, 0.337)$ | | 0.978 |
| $A_1$ | $SVWG$ | $(0.303143, 0.2, 0.368011)$ | $(0.4, 0.2, 0.3)$ | 0.975 |
| | $ISVWAG$ | $(0.254226, 0.172108, 0.303)$ | | 0.977 |
| | $SVWA$ | $(0.46, 0.13, 0.18)$ | | 0.993 |
| $A_2$ | $SVWG$ | $(0.55, 0.13, 0.2)$ | $(0.6, 0.1, 0.2)$ | 0.997 |
| | $ISVWAG$ | $(0.43, 0.11, 0.17)$ | | 0.995 |
| | $SVWA$ | $(0.373, 0.222, 0.238)$ | | 0.9806 |
| $A_3$ | $SVWG$ | $(0.418, 0.23, 0.25)$ | $(0.3, 0.2, 0.3)$ | 0.974 |
| | $ISVWAG$ | $(0.33, 0.2, 0.21)$ | | 0.9803 |
| | $SVWA$ | $(0.46, 0.14, 0.15)$ | | 0.946 |
| $A_2$ | $SVWG$ | $(0.54, 0, 0.16)$ | $(0.7, 0, 0.1)$ | 0.989 |
| | $ISVWAG$ | $(0.42, 0, 0.13)$ | | 0.987 |

# 10 CONCLUSION

At first we introduced a new aggregation operator ($IWAGM$) to combine n real numbers. We proved that the result using this operator always lies between $WAM$ and $WGM$ operator and the result will be meaningful in all the cases. Then we extended the operator in neutrosophic environment and it has also been shown that the extended operator ($ISVWAG$) gives meaningful result in neutrosophy. Next we introduced a trigonometric function based score function. Further we proposed a certainty function as well which gives realistic results comparison to the existing ones. A numerical problem has been solved using the proposed operator and the newly defined score function.

# ACKNOWLEDGMENT

The authors are very grateful to Florentin Smarandache, Surapati Pramanik and Pranab Biswas for their insightful and constructive comments and suggestions, which have been very helpful in improving the paper.

# References

[1] Florentin Smarandache, Neutrosophy/Neutrosophic Probability, Set and Logic, *Amer. Res.Press, Rehoboth, USA*, (1998).

[2] Haibin Wang, Florentin Smarandache, Yanqing Zhang and Rajshekhar Sunderraman, Interval Neutrosophic Sets and Logic: Theory and Application in Computing, *Hexis, Neutrosophic Book Series*, (2005).

[3] Jun Ye, A Multi-criteria Decision Making Method Using Aggregation Operators for Simplified Neutrosophic Sets, *Journal of Intelligent & Fuzzy Systems*, **26** (2014), 2459-2466, doi 10.3233/IFS-130916.

[4] Jun Ye, Multiple-attribute Decision Making Method under a Single Valued Neutrosophic Hesitant Fuzzy Environment, *Journal of Intelligent Systems*, **24** (2014), 23-36, doi 10.1515/jisys-2014-0001.

[5] Pinaki Majumdar and S.K. Samanta, On Similarity and Entropy of Neutrosophic Sets, *Journal of Intelligent and Fuzzy Systems*, **26** (2014), 1245–1252, doi 10.3233/IFS-130810.

[6] Lotfi Zadeh, Fuzzy Entropy and Conditioning, *Information Sciences*, **40** (1986), 165–174.

[7] Peide Liu, Yanchang Chu, Yanwei Li and Yubao Chen, Some Generalized Neutrosophic Number Hamacher Aggregation Operators and Their Application to Group Decision Making, *International Journal of Fuzzy systems*, **16** (2014), 242–255.

[8] Said Broumi and Florentin Smarandache, New Operations on Interval Neutrosophic Sets, *Journal of New Theory*, **1** (2015), 24–27.

[9] Jun Ye, Vector Similarity Measures of Simplified Neutrosophic Sets and Their Application in Multicriteria Decision Making, *International Journal of Fuzzy Systems*, **16** (2014), 204–215.

[10] Wenkai Zhang, Xia Li and Yanbing Ju, Some Aggregation Operators Based on Einstein Operations under Interval-Valued Dual Hesitant Fuzzy Setting and Their Application, *Mathematical Problems in Engineering*, **2014** (2014), 958927–9589248, doi 10.1155/2014/958927.

[11] Said Broumi and Florentin Smarandache, Cosine Similarity Measure of Interval Valued Neutrosophic Sets, *Neutrosophic Sets and Systems*, **5** (2014), 15–20.

[12] Haibin Wang, Florentin Smarandache, Yanqing Zhang and Rajshekhar Sunderraman, Single Valued Neutrosophic Sets, *in Multispace and Multistructure*, **4**, (2010), 410–413. http://fs.gallup.unm.edu/SingleValuedNeutrosophicSets.pdf.

[13] Zhang-peng Tian, Jing Wang, Hong-yu Zhang and Jian-qiang Wang, Multi-criteria Decision Making Based on Generalized Prioritized Aggregation Operators under Simplified Neutrosophic Uncertain Linguistic Environment, *Int. J. Mach. Learn. & Cyber*, (2016), 1–17, doi 10.1007/s13042-016-0552-9.

[14] Juan-juan Peng, Jian-qiang Wang, Hong-yu Zhang and Xiao-hong Chen, An Outranking Approach for Multi-criteria Decision Making Problems with Simplified Neutrosophic Sets, *Applied Soft Computing*, **25** (2014), 336–346.

[15] Zhang-peng Tian, Hong-yu Zhang, Jing Wang, Jian-qiang Wang and Xiao-hong Chen, Multi-criteria Decision Making Method Based on a Cross-entropy with Interval Neutrosophic Sets, *International Journal of Systems Science*, **47** (2016), 3598–3608, doi 10.1080/00207721.2015.1102359.

[16] Yin-xiang Ma, Jian-qiang Wang, Jing Wang and Xiaohui Wu, An Interval Neutrosophic Linguistic Multi-criteria Group Decision Making Method and Its Application in Selecting Medical Treatment Options, *Neural Computer & Application*, (2016), 1–21, doi 10.1007/s00521-016-2203-1.

[17] Zhang-peng Tian, Jing Wang, Jian-qiang Wang and Hong-yu Zhang, Simplified Neutrosophic Linguistic Multicriteria Group Decision-Making Approach to Green Product Development, *Group Decision and Negotiation*, **4** (2016), doi 10.1007/s10726-016-9479-5.

[18] Hong-yu Zhang, Pu Ji, Jian-qiang Wang and Xiao-hong Chen, A Neutrosophic Normal Cloud and Its Application in Decision Making, *Cognitive Computation*, **8** (2016), 649–669, doi 10.1007/s12559-016-9394-8,2016.

[19] Zhang-peng Tian, Jing Wang, Jian-qiang Wang and Hong-yu Zhang, A Likelyhood-based Qualitative Flexible Approach with Hesitant Fuzzy Linguistic Information, *Cognitive Computation*, **8** (2016), 670–683, doi 10.1007/s12559-016- 9400-1,2016.

[20] Hong-yu Zhang, Jian-qiang Wang and Xiao-hong Chen, Interval Neutrosophic Sets and Their Application in Multi-criteria Decision Making Problems, *The Scientic World Journal*, **2014** (2014), 645953-645967, doi 10.1155/2014/645953.

SURAPATI PRAMANIK[1*], SHYAMAL DALAPATI[2], TAPAN KUMAR ROY[3]

1* Department of Mathematics, Nandalal Ghosh B.T. College, Panpur, P.O.-Narayanpur, District –North 24 Parganas, Pin code-743126, West Bengal, India. Corresponding author's E-mail: sura_pati@yahoo.co.in
2,3 Indian Institute of Engineering Science and Technology, Department of Mathematics,Shibpur,Pin-711103, West Bengal, India. E-mails:dalapatishyamal30@gmail.com, roy_t_k@yahoo.co.in

# Logistics Center Location Selection Approach Based on Neutrosophic Multi-Criteria Decision Making

## Abstract

As an important and interesting topic in supply chain management, the concept of fuzzy set theory has been widely used in logistics center location in order to improve the reliability and suitability of the logistics center location with respect to the impacts of both qualitative and quantitative factor. However fuzzy set cannot deal with the indeterminacy involving with the problem. So the concept of single – valued neutrosophic set due to Wang et al. (2010) is very helpful to deal with the problem. A neutrosophic approach is a more general and suitable approach in order to deal with neutrosophic information than fuzzy set. Logistics center location selection is a multi-criteria decision making process involving subjectivity, impresion and fuzziness that can be easily represented by single-valued neutrosophic sets. In this paper, we use the score and accuracy function and hybrid score accuracy function of single- valued neutrosophic number and ranking method for single- valued neutrosophic numbers to model logistics center location problem. Finally, a numerical example has been presented to illustrate the proposed approach.

## 1. Introduction

Logistics systems have become essential for economic development and the normal function of the society, and suitable site selection for the logistics center has direct impact on the efficiency of logistics systems. So it is necessary to adopt a scientific approach for site selection. The logistic center location selection problem can be considered as multi-criteria decision making (MCDM) problem. Classical MCDM [1, 2, 3] problems deal with crisp numbers that is the ratings and the weights of the criteria are represented by crisp numbers. However, it is not always possible to present the information by crisp numbers. In order to deal this situation fuzzy set (FS) introduced by Zadeh [4] in 1965 can be used. It is very useful for many real life problems involving uncertainty. In 1986, Atanassov [5] grounded the notion of intuitionistic fuzzy set (IFS) by introducing non-membership function as independent component. However, it cannot handle indeterminacy part of the real life problems that exist in many real applications. Then in 1998, Smarandache proposed the neutrosophic set (NS) theory [6,7, 8, 9] which is the generalization of FS and IFS.

From scientific or engineering point of view, the neutrosophic set and set- theoretic view, operators need to be specified. Otherwise, it will be difficult to apply in the real applications. Therefore, Wang et al. [10] defined a single valued neutrosophic set (SVNS) and then provided the set theoretic operations and various properties of SVNS. The works on SVNS and their hybrid structure in theories and application have been progressing rapidly. Hence it is most important to conduct researches on MCDM approach based on SVNS environment. Biswas et al. [11] presented entropy based grey relational analysis method for multi-attribute decision making under SVNS. Ye [12] proposed the co-relation co-efficient of SVNSs for single-valued neutrosophic MCDM problem. While selecting the location for the logistics center not only quantitative factors likes costs, distances but also qualitative factors. Such as environmental impacts and governmental regulations should be taken into consideration. Tuzkaya et al. [13] presented an analytic network process approach to deal locating undesirable facilities. Badri [14] studied a method combinjing analytical hierarchy process (AHP) and goal programming model approach for international facility location problem. Chang and Chung [15] proposed a multi-criteria genetic optimization for distibution network problems. Liang and Wang [16] proposed a fuzzy multicriteria decision making method for facility site selection. Chu [17] proposed facility location selection using fuzzy TOPSIS under group decision. Recently, Pramanik and Dalapati [18] presented generalized neutrosophic soft multi criteria decision making based on grey relational analysis by introducing generalized neutrosophic soft weighted average operator. In this paper we present logistic center location model using score and accuracy function and hybrid score accuracy function of single-valued neutrosophic number due to Ye [19]. Finally, a numerical example has been provided to illustrate the proposed approach.

Rest of the paper has been organized in the following way. Section 2 presents preliminaries of neutrosophic sets and section 3 presents criteria for logistic center location selection. Section 4 is devoted to present modeling of logistic center location seclection problem. Section 5 presents a numerical example of the logistic center location selection problem. In Section 6 we presents conclusion.

## 2. Mathematical preliminaries

In this section, we will recall some basic definitions and concepts that are useful to develop the paper.

**Definition 1: Neutrosophic sets [6, 7, 8, 9]**

Let $\mathcal{P}$ be a universe of discourse with a generic element in $\mathcal{P}$ denoted by p. A neutrosophic set Z in $\mathcal{P}$ is characterized by a truth-membership function $t_Z(p)$, an indeterminacy-membership function $i_Z(p)$ and a falsity-membership function $f_Z(p)$ and defined by $Z = \{ <p, t_Z(p), i_Z(p), f_Z(p)>: p \in \mathcal{P}\}$.The function $t_Z(p)$ , $i_Z(p)$ and $f_Z(p)$ are real standard or nonstandard subsets of $]^-0, 1^+[$ i.e., $t_Z(p) : \mathcal{P} \rightarrow ]^-0, 1^+[$,

$i_Z(u): \mathcal{P} \rightarrow ]^-0, 1^+[$, and $f_Z(u) : \mathcal{P} \rightarrow ]^-0, 1^+[$ Hence, there is no restriction on the sum of $t_Z(p)$, $i_Z(p)$ and $f_Z(p)$ and $^-0 \le t_Z(p) + i_Z(p) + f_Z(p) \le 3^+$.

**Definition 2: Single valued neutrosophic sets [10].**

Let $\mathcal{P}$ be a universe of discourse with a generic element in $\mathcal{P}$ denoted by p. A single valued neutrosophic set M in $\mathcal{P}$ is characterized by a truth-membership function $t_M(p)$, an indeterminacy-membership function $i_M(p)$ and a falsity-membership function $f_M(p)$. It can be expressed as $M = \{< p, (t_M(p), i_M(p), f_M(p)) >: p \in \mathcal{P}, t_M(p), i_M(p), f_M(p) \in [0, 1]\}$. It should be noted that there is no restriction on the sum of $t_M(p)$, $i_M(p)$ and $f_M(p)$. Therefore, $0 \le t_M(p) + i_M(p) + f_M(p) \le 3$.

**Definition 3**: **Single valued neutrosophic number (SVNN) [19]**

Let $\mathcal{P}$ be a universe of discourse with generic element in $\mathcal{P}$ denoted by p. A SVNS M in $\mathcal{P}$ is characterized by a truth-membership function $t_M(p)$, a indeterminacy-membership function $i_M(p)$ and a falsity-membership function $f_M(p)$. Then, a SVNS M can be written as follows: $M = \{<p, t_M(p), i_M(p), f_M(p)>: p \in \mathcal{P}\}$, where $t_M(p), i_M(p), f_M(p) \in [0, 1]$ for each point p in $\mathcal{P}$. Since no restriction is imposed in the sum of $t_M(p)$, $i_M(p)$ and $f_M(p)$, it satisfies $0 \le t_M(p) + i_M(p) + f_M(p) \le 3$. For a SVNS M in $\mathcal{P}$ the triple $< t_M(p), i_M(p), f_M(p) >$ is called single valued neutrosophic number (SVNN).

**Definition 4: Complement of a SVNS [10]**

The complement of a single valued neutrosophic set M is denoted by M′and defined as

$M' = \{<p: t_{M'}(p), i_{M'}(p), f_{M'}(p)>, p \in \mathcal{P}\}$,

where $t_{M'}(p) = f_M(p)$, $i_{M'}(p) = \{1\} - i_M(p)$, $f_{M'}(p) = t_M(p)$.

For two SVNSs $M_1$ and $M_2$ in $\mathcal{P}$, $M_1$ is contained in the $M_2$, i.e. $M_1 \subseteq M_2$, if and only if $t_{M_1}(p) \le t_{M_2}(p)$, $i_{M_1}(p) \ge i_{M_2}(p)$, $f_{M_1}(p) \ge f_{M_2}(p)$ for any p in $\mathcal{P}$.

Two SVNSs $M_1$ and $M_2$ are equal, written as $M_1 = M_2$, if and only if $M_1 \subseteq M_2$ and $M_2 \subseteq M_1$.

**2.1. Conversion between linguistic variables and single valued neutrosophic numbers**

A linguistic variable simply presents a variable whose values are represented by words or sentences in natural or artificial languages. Importance of the decision makers may be differential in the decision making process. Ratings of criteria can be exressed by using linguistic variables such as very poor (VP), poor (P), good (G), very good (VG), excellent (EX), etc. Linguistic variables can be transformed into single valued neutrosophic numbers as given in the Table- 1.

**2.2 Ranking methods for SVNNs**

Now we recall the definition of the score function, accuracy function, and hybrid score-accuracy function of a SVNN, and the ranking method for SVNNs.

**Definition 5**: **Score function and accuracy function [19]**

The score function and accuracy function of the SVNN b= (t(b), i(b), f(b)) can be expressed as follows:

$S(b) = (1+t(b) - f(b))/2$ for $s(b) \in [0, 1]$ (1)

$ac(b) = (2 + t(b) - f(b) - i(b))/3$ for $h(b) \in [0, 1]$ (2)

For the score function of a SVNN b, the truth membership t(b) is bigger and the falsity-membership f(b) are smaller, than the score value of the SVNN a is greater. For the accuracy function of a SVNN b if the sum of t(b), 1-i(b) and 1-f(b) is bigger, then the statement is more affirmative, i.e., the accuracy of the SVNN b is higher. Based on score and accuracy functions for SVNNs, two theorems are stated below.

**Theorem 1**.

For any two SVNNs $b_1$ and $b_2$, if $b_1 > b_2$, then $s(b_1) > s(b_2)$.

Proof: See [19].

**Theorem 2**.

For any two SVNNs $b_1$ and $b_2$, if $s(b_1) = s(b_2)$ and $b_1 \geq b_2$, then $ac(b_1) \geq ac(b_2)$.

Prof: See [19]

Based on theorems 1 and 2, a ranking method between SVNNs can be given by the following definition.

**Definition 6: [19]**

Let $b_1$ and $b_2$ be two SVNNs. Then, the ranking method can be defined as follows:

1. If $s(b_1) > s(b_2)$, then $b_1 > b_2$;
2. If $s(b_1) = s(b_2)$ and $ac(b_1) \geq ac(b_2)$, then $b_1 \geq b_2$;

## 3. Criteria for logistics center location selection

We choose the most appropriate location on the basis of six criteria adapted from the study [20], namely, cost ($K_1$), distance to suppliers ($K_2$), dsistance to customers ($K_3$), conformance to governmental regulations and laws ($K_4$), quality of service ($K_5$) and environmental impact ($K_6$).

## 4. MCGDM method based on hybrid – score accuracy functions under single-valued neutrosophic environment

Assume that $B = \{B_1, \ldots, B_n\}$ $(n \geq 2)$ be the set of logistics centers, $K = \{K_1, K_2, \ldots, K_\rho\}$ $(\rho \geq 2)$ be the set of criteria and $E = \{E_1, E_2, \ldots, E_m\}$ $(m \geq 2)$ be the set of decision makers or experts. The weights of the decision makers and criteria are not previously assigned, where the information about the weights of the decision- makers is completely unknown and information about the weights of the criteria is incompletely known in the group decision making problem. In such a case, we develop a method based on the hybrid score – accuracy function for MCDM problem with unknown weights under single-valued neutrosophic environment using linguistic variables. The steps for solving MCGDM by proposed approach have been presented below.

**Step – 1**

**Formation of the decision matrix**

In the group decision process, if m decision makers or experts are required in the evaluation process, then the s-th (s = 1, 2,…, m) decision maker can provide the evaluation information of the alternative $B_i$ (i = 1, ..., n) on the criterion $K_j$ (j = 1, ..., $\rho$) in linguistic variables that can be expressed by the SVNN ( see Table 1). A MCGDM problem can be expressed by the following decision matrix:

$$M_s = (b^s_{ij})_{n\times\rho} = \begin{pmatrix} & K_1 & K_2 & \dots & K_\rho \\ B_1 & b^s_{11} & b^s_{12} & \dots & b^s_{1\rho} \\ B_2 & b^s_{21} & b^s_{22} & & b^s_{2\rho} \\ . & . & . & \dots & . \\ B_n & b^s_{n1} & b^s_{n2} & \dots & b^s_{n\rho} \end{pmatrix} \quad (3)$$

$B^s_{ij} = \{(K_j, t^s_{B_i}(K_j), i^s_{B_i}(K_j), f^s_{B_i}(K_j) / K_j \in K\}$

Here $0 \le t^s_{B_i}(K_j) + i^s_{B_i}(K_j) + f^s_{B_i}(K_j) \le 3$

$t^s_{B_i}(K_j) \in [0, 1], i^s_{B_i}(K_j) \in [0, 1], f^s_{B_i}(K_j) \in [0, 1]$

For s = 1, 2, ..., m, j =1, 2, … $\rho$, i = 1, 2, …n,

For convenience $b^s_{ij} = (t^s_{ij}, i^s_{ij}, f^s_{ij})$ is denoted as a SVNN in the SVNS $B^s_{ij}$ (s =1, 2, ..., m, i = 1, ..., n, j = 1, ..., $\rho$)

**Step – 2**

**Calculate hybrid score – accuracy matrix**

The hybrid score – accuracy matrix $H^s = (h^s_{ij})_{n\times\rho}$ (s = 1, 2……, m; i = 1, 2, ..., n; j = 1, 2, …, $\rho$) can be obtained from the decision matrix $M_s = (b^s_{ij})_{n\times\rho}$. The hybrid score-accuracy matrix $H^s$ expressed as

$$H^s = (h^s_{ij})_{n\times\rho} = \begin{pmatrix} & K_1 & K_2 & \dots\dots & K_\rho \\ B_1 & h^s_{11} & h^s_{12} & \dots & h^s_{1\rho} \\ B_2 & h^s_{21} & h^s_{22} & & h^s_{2\rho} \\ . & . & \dots & . & \\ B_n & h^s_{n1} & h^s_{n2} & \dots & h^s_{n\rho} \end{pmatrix} \quad (4)$$

$$h^s_{ij} = \frac{1}{2}\alpha\ (1 + t^s_{ij} - f^s_{ij}) + \frac{1}{3}(1 - \alpha)(2 + t^s_{ij} - i^s_{ij} - f^s_{ij}) \quad (5)$$

Where $\alpha \in [0, 1]$. When $\alpha = 1$ the equation (3) reduces to equation (1) and when $\alpha = 0$, the equation reduces to equation (2).

**Step – 3**

**Calculate the average matrix**

Form the obtained hybrid-score–accuracy matrix, the average matrix $H^{*}= (h^{*}_{ij})_{n \times \rho}$ (s =1, 2, ..., m; i = 1, 2, ..., n; j = 1, 2, ..., $\rho$) is

$$\text{expressed by } H^{*} = (h^{*}_{ij})_{n\times\rho} = \begin{pmatrix} & K_1 & K_2 & \ldots\ldots & K_\rho \\ B_1 & h^{*}_{11} & h^{*}_{12}\ldots & & h^{*}_{1\rho} \\ B_2 & h^{*}_{21} & h^{*}_{22} & & h^{*}_{2\rho} \\ . & . & \ldots & . & \\ B_n & h^{*}_{n1} & h^{*}_{n2}\ldots & & .h^{*}_{n\rho} \end{pmatrix} \quad (6)$$

$$\text{Here} \quad h^{*}_{ij} = \frac{1}{m}\sum_{s=1}^{m}(h^{s}_{ij}) \quad (7)$$

Ye [19] defined the collective correlation co-efficient between $H^{s}$ (s = 1, 2, ..., m) and $H^{*}$ as follows.

$$\Omega_s = \sum_{i=1}^{n} \frac{\sum_{j=1}^{\rho} h^{s}_{ij} h^{*}_{ij}}{\sqrt{\sum_{j=1}^{\rho}(h^{s}_{ij})^2}\sqrt{\sum_{j=1}^{\rho}(h^{*}_{ij})^2}} \quad (8)$$

**Step – 4**

**Determination decision maker's weights**

In decision making situation, the decision makers may exhibit personal biases and offer unduly high or low preference values with respect to their preferred or repugnant objects. In order to deal such cases, very low weights to these false or biased opinions can be assigned. Since the "mean value" reflects the distributing center of all elements of the set, the average matrix $H^{*}$ represents the maximum compromise among all individual decisions of the group. In this sense, a hybrid score accuracy matrix $H^{s}$ is closer to the average one $H^{*}$. Then the preference value of the s-th decision maker is closer to the average value and his/her evaluation is more reasonable and more important. Therefore, the weight of the s-th decision maker is bigger. Ye [19] defined weight model for decision maker as follows:

$$\gamma_s = \frac{\Omega_s}{\sum_{s=1}^{m}\Omega_s}, \quad 0 \le \gamma_s \le 1, \ \sum_{s=1}^{m}\gamma_s = 1 \text{ for } s = 1, 2, \ldots, m. \quad (9)$$

**Step – 5**

**Calculate collective hybrid score – accuracy matrix**

For the weight vector $\gamma = (\gamma_1, \gamma_2, \ldots, \gamma_m)^T$ of decision makers obtained from equation (6),

Ye [19] accumulates all individual hybrid score – accuracy matrix $H^s = (h^s_{ij})_{n\times\rho}$ (s = 1, 2,..., m; i = 1, 2, ..., n; j = 1, 2, ..., ρ) into a collective hybrid score accuracy matrix

$$H=(h_{ij})_{n\times\rho} = \begin{pmatrix} & K_1 & K_2 & \ldots\ldots & K_\rho \\ B_1 & h_{11} & h_{12}\ldots & & h_{1\rho} \\ B_2 & h_{21} & h_{22} & & h_{2\rho} \\ . & . & & \ldots & . \\ B_n & h_{n1} & h_{n2}\ldots & & .h_{n\rho} \end{pmatrix} \quad (10)$$

Here $h_{ij} = \Sigma_{s=1}^{m} \gamma_s h^s_{ij}$ (11)

**Step – 6**

**Weight model for criteria**

To deal decision making problem, the weights of the criteria can be given in advance in the form of partially known subset corresponding to the weight information of the criteria.

To determine weights of the critria Ye [19] introduced the following optimization model :

$$\text{Max } \omega = \frac{1}{n} \Sigma_{i=1}^{n} \Sigma_{j=1}^{\rho} \omega_j h_{ij} \quad (12)$$

Subject to

$\Sigma_{j=1}^{\rho} \omega_j = 1$

$\omega_j > 0$

Solving the linear programming problem (12), the weight vector of the criteria $\omega = (\omega_1, \omega_2, ..., \omega_n)^T$ can be easily determined.

**Step 7**

**Ranking of alternatives**

In order to rank alternatives, all values can be summed in each row of the collective hybrid score-accuracy matrix corresponding to the criteria weights by the overall weight hybrid score-accuracy value of each alternative $B_i$ (i = 1, 2, . . . , n):

$$\Psi(B_i) = \Sigma_{j=1}^{\rho} \omega^i h^{ij} \quad (13)$$

Based on the values of $\Psi(B_i)$ (i = 1, 2, ...., n), we can rank alternatives $B_i$ (i = 1, 2, ..., n) in descending order and choose the best alternative.

**Step – 8**

End

## 5. Example of the Logistics Center Location Selection Problem

Assume that a new modern logistic center is required in a town. There are four location $B_1$, $B_2$, $B_3$, $B_4$. A committee of four decision makers or experts namely, $E_1$, $E_2$, $E_3$, $E_4$ has been formed to select the most appropriate location on the basis of six criteria adopted from the study [20], namely, cost ($K_1$), distance to suppliers ($K_2$), distance to customers ($K_3$), conformance to government regulation and laws ($K_4$), quality of service ($K_5$) and environmental impact ($K_6$). Thus the four decision makers use linguistic variables (see Table 1) to rate the alternatives with respect to the criterion and construct the decision matrices ( see Table 2-5) as follows:

Table 1: *Conversion between linguistic variable and SVNNs*

| | *Linguistic term* | *SVNNs* |
|---|---|---|
| *1* | *Very Poor(VP)* | *(.05,.95,.95)* |
| *2* | *Poor (P)* | *(.25,.75,.75)* |
| *3* | *Good (G)* | *(.50,.50,.50)* |
| *4* | *Very Good (VG)* | *(.75,.25,.25)* |
| *5* | *Excellent (EX)* | *(.95,.05,.05)* |

Table 2: *Decision matrix for $E_1$ in the form of linguistic term.*

| $B_i$ | $K_1$ | $K_2$ | $K_3$ | $K_4$ | $K_5$ | $K_6$ |
|---|---|---|---|---|---|---|
| $B_1$ | *VG* | *EX* | *VG* | *G* | *G* | *P* |
| $B_2$ | *VG* | *G* | *G* | *EX* | *VG* | *VG* |
| $B_3$ | *G* | *EX* | *EX* | *G* | *VG* | *G* |
| $B_4$ | *EX* | *VG* | *G* | *EX* | *VG* | *VG* |

Table 3: *Decision matrix for $E_2$ in the form of linguistic term.*

| $B_i$ | $K_1$ | $K_2$ | $K_3$ | $K_4$ | $K_5$ | $K_6$ |
|---|---|---|---|---|---|---|
| $B_1$ | VG | VG | VG | G | VG | P |
| $B_2$ | EX | VG | VG | VG | P | P |
| $B_3$ | P | EX | EX | VG | G | G |
| $B_4$ | G | G | EX | VG | EX | EX |

Table 4: *Decision matrix for $E_3$ in the form of linguistic term.*

| $B_i$ | $K_1$ | $K_2$ | $K_3$ | $K_4$ | $K_5$ | $K_6$ |
|---|---|---|---|---|---|---|
| $B_1$ | VG | VG | EX | VG | VG | G |
| $B_2$ | EX | G | EX | VG | EX | VG |
| $B_3$ | P | EX | EX | VG | G | VG |
| $B_4$ | G | G | VG | EX | EX | EX |

Table 5: *Decision matrix for $E_4$ in the form of linguistic term.*

| $B_i$ | $K_1$ | $K_2$ | $K_3$ | $K_4$ | $K_5$ | $K_6$ |
|---|---|---|---|---|---|---|
| $B_1$ | EX | VP | P | VG | VG | VG |
| $B_2$ | G | G | EX | VG | G | EX |
| $B_3$ | P | EX | VG | G | VG | VG |
| $B_4$ | VG | VG | G | G | VG | G |

*Step-1*

**Formation of the decision matrix**

*Decision matrix for $E_1$ in the form of SVNN*

$M_1$ =

$$\begin{pmatrix} & K1 & K2 & K3 & K\,4 & K5 & K6 \\ B1 & (.75, .25, .25) & (.95, .05, .05) & (.75, .25, .25) & (.50, .50, .50) & (.50, .50, .50) & (.25,.75,.75) \\ B2 & (.75, .25, .25) & (.50, .50, .50) & (.50, .50, .50) & (.95,.05, .05) & (.75, .25, .25) & (.75, .25, .25) \\ B3 & (.50, .50, .50) & (.95, .05, .05) & (.95, .05, .05) & (.50, .50, .50) & (.75, .25, .25) & (.50, .50, .50) \\ B4 & (.95, .05, .05) & (.75, .25, .25) & (.50, .50, .50) & (.95, .05, .05) & (.75, .25, .25) & (.75, .25, .25) \end{pmatrix}$$

*Decision matrix for $E_2$ in the form of SVNN*

$M_2$ =

| | K1 | K2 | K3 | K4 | K5 | K6 |
|---|---|---|---|---|---|---|
| B1 | (.75,.25,.25) | (.75,.25,.25) | (.75,.25,.25) | (.50,.50,.50) | (.75,.25,.25) | (.25,.75,.75) |
| B2 | (.95,.05,.05) | (.75,.25,.25) | (.75,.25,.25) | (.75,.25,.25) | (.25,.75,.75) | (.25,.75,.75 |
| B3 | (.25,.75,.75) | (.95,.05,.25) | (.95,.05,.05) | (.75,.25,.25) | (.50,.50,.50) | (.50,.50,.50) |
| B4 | (.50,.50,.50) | (.50,.50,.50) | (.95,.05,.05) | (.75,.25,.25) | (.95,.05,.05) | (.95,.05,.05) |

*Decision matrix for $E_3$ in the form of SVNN*

$M_3$ =

| | K1 | K2 | K3 | K4 | K5 | K6 |
|---|---|---|---|---|---|---|
| B1 | (.75,.25,.25) | (.75,.25,.25) | (.95,.05,.05) | (.75,.25,.25) | (.75,.25,.25) | (.50,.50,.50) |
| B2 | (.95,.05,.05) | (.50,.50,.50) | (.95,.05,.05) | (.75,.25,.25) | (.95,.05,.05) | (.75,.25,.25) |
| B3 | (.25,.75,.75) | (.95,.05,.05) | (.95,.05,.05) | (.75,.25,.25) | (.50,.50,.50) | (.75,.25,.25) |
| B4 | (.50,.50,.50) | (.50,.50,.50) | (.75,.25,.25) | (.95,.05,.05) | (.95,.05,.05) | (.95,.05,.05) |

*Decision matrix for $E_4$ in the form of SVNN*

$M_4$ =

| | K1 | K2 | K 3 | K4 | K5 | K6 |
|---|---|---|---|---|---|---|
| B1 | (.95,.05,.05) | (.05,.95,.95) | (.25,.75,.75) | (.75,.25,.25) | (.75,.25,.25) | (.75,.25,.25) |
| B2 | (.50,.50,.50) | (.50,.50,.50) | (.95,.05,.05) | (.75,.25,.25) | (.50,.50,.50) | (.95,.05,.05) |
| B3 | (.25,.75,.75) | (.95,.05,.05) | (.75,.25,.25) | (.50,.50,.50) | (.75,.25,.25) | (.75,.25,.25) |
| B4 | (.75,.25,.25) | (.75,.25,.25) | (.50,.50,.50) | (.50,.50,.50) | (.75,.25,.25) | (.50,.50,.50) |

*Now we use the above method for single valued neutrophic group decision making to choice appropriate location. We take* $\alpha$ *= 0.5 for demonstrating the computing procedure*

*Step 2*

**Calculate hybrid score – accuracy matrix**

Hybrid score- accuracy matrix can be obtained from above decision matrix using equation (5) are given below respectively.

*Hybrid score-accuracy matrix for $M_1$*

$H^1$ =

| | K1 | K2 | K3 | K4 | K5 | K6 |
|---|---|---|---|---|---|---|
| B1 | .75 | .95 | .75 | .50 | .50 | .25 |
| B2 | .75 | .50 | .50 | .95 | .75 | .75 |
| B3 | .50 | .95 | .95 | .50 | .75 | .50 |
| B4 | .95 | .75 | .50 | .95 | .75 | .75 |

*Hybrid score-accuracy matrix for* $M_2$

$H^2 =$

$$\begin{pmatrix} & K1 & K2 & K3 & K4 & K5 & K6 \\ B1 & .75 & .75 & .75 & .50 & .75 & .25 \\ B2 & .95 & .75 & .75 & .75 & .25 & .25 \\ B3 & .25 & .95 & .95 & .75 & .50 & .50 \\ B4 & .50 & .50 & .95 & .75 & .95 & .95 \end{pmatrix}$$

*Hybrid score-accuracy matrix for* $M_3$

$H^3 =$

$$\begin{pmatrix} & K1 & K2 & K3 & K4 & K5 & K6 \\ B1 & .75 & .75 & .95 & .75 & .75 & .50 \\ B2 & .95 & .50 & .95 & .75 & .95 & .75 \\ B3 & .25 & .95 & .95 & .75 & .50 & .75 \\ B4 & .50 & .50 & .75 & .95 & .95 & .95 \end{pmatrix}$$

*Hybrid score-accuracy matrix for* $M_4$

$H^4 =$

$$\begin{pmatrix} & K1 & K2 & K3 & K4 & K5 & K6 \\ B1 & .95 & .05 & .25 & .75 & .75 & .75 \\ B2 & .50 & .50 & .95 & .75 & .50 & .95 \\ B3 & .25 & .95 & .75 & .50 & .75 & .75 \\ B4 & .75 & .75 & .50 & .50 & .75 & .50 \end{pmatrix}$$

**Step – 3**

**Calculate the average matrix**

*Form the above hybrid score-accuracy matrix by using euation(7). We form the average matrix* $H^*$

*The average matrix*

$H^* =$

$$\begin{pmatrix} & K1 & K2 & K3 & K4 & K5 & K6 \\ B1 & .8000 & 0.625 & 0.675 & 0.625 & 0.6875 & 0.4375 \\ B2 & .7875 & .5625 & .7875 & .8000 & 0.6125 & 0.6750 \\ B3 & .3125 & .9500 & .9000 & .6250 & .6250 & 0.6250 \\ B4 & .6750 & .6250 & .6750 & .7875 & .8500 & .7875 \end{pmatrix}$$

*The collective correlation co-efficent between* $H^s$ *and* $H^*$ *express follows by equation (8) :-*

$$\Omega_s = \sum_{i=1}^{4} \frac{\sum_{j=1}^{6} h_{ij}^s h_{ij}^*}{\sqrt{\sum_{j=1}^{6} (h_{ij}^s)^2}\sqrt{\sum_{j=1}^{6} (h_{ij}^*)^2}}$$

$\Omega_1 = 3.907$

$\Omega_2 = 3.964$

$\Omega_3 = 4.124$

$\Omega_4 = 3.800$

**Step – 4**

**Determination decision maker's weights**

*From the equation (9) we determine the weight of the four decision makers as follows :-*

$\Omega_1 + \Omega_2 + \Omega_3 + \Omega_4 = 15.79509754$

$$\gamma_1 = \frac{\Omega_1}{\Omega_1 + \Omega_2 + \Omega_3 + \Omega_4}$$

$$= \frac{3.90705306}{15.79509754} = .247$$

$\gamma_2 = .251$

$\gamma_3 = .261$

$\gamma_4 = .240$

**Step – 5**

**Calculate collective hybrid score – accuracy matrix**

*Hence the hybrid score-accuracy values of the different decision makers choice are aggregated by eq. (11) and the collective hybrid score-accuracy matrix can be formulated as follows:*

H =

| | K1 | K2 | K3 | K4 | K5 | K6 |
|---|---|---|---|---|---|---|
| B1 | .798 | .631 | .682 | .625 | .688 | .436 |
| B2 | .792 | .563 | .788 | .799 | .616 | .673 |
| B3 | .312 | .950 | .902 | .628 | .622 | .625 |
| B4 | .671 | .622 | .678 | .792 | .852 | .792 |

**Step – 6**

**Weight model for criteria**

*Assume that the information about criteria weights is incompletely known given as follows: weight vectors,*

$0.1 \le \omega_1 \le 0.2$, $0.1 \le \omega_2 \le 0.2$,

$0.1 \le \omega_3 \le 0.25$, $0.1 \le \omega_4 \le 0.2$,

$0.1 \le \omega_5 \le 0.2$, $0.1 \le \omega_6 \le 0.2$

*Using the linear programming model (12), we obtain the weight vector of the criteria as* $\omega = [0.1, 0.1, 0.25, 0.2, 0.15, 0.2]^T$.

**Step 7**

**Ranking of alternatives**

*Using euation (13) we calculate the over all hybrid score-accuracy values*

$\Psi(B_i)$ *(i = 1, 2, 3, 4):*

$\Psi(B_1) = .06288$

$\Psi(B_2) = .7193$

$\Psi(B_3) = .6956$

$\Psi(B_4) = .7434$

*Based on the above values of* $\Psi(B_i)$ *(i = 1, 2, 3, 4) the ranking order of the locations are as follows:*

$B_4 > B_2 > B_3 > B_1$

*Therefore the location* $B_4$ *is the best location.*

**Step – 8**

*End*

## 6. Conclusion

In this paper we have presented modeling of logistics center location problem using the score and accuracy function, hybrid-score-accuracy function of SVNNs and linguistic variables under single-valued neutrosophic environment, where weight of the decision makers are completely unknown and the weight of criteria are incompletely known.

## References

1. R. D. Arvey and J. E. Campion. The employment interview: a summary and review of recent research, Personnel Psychology, 35(2) (1982), 281-322.
2. L. Hwang and M. J. Lin. Group decision making under multiple criteria: methods and applications, Springer Verlag, Heidelberg, (1987).
3. M. A. Campion, E. D. Pursell and B. k. Brown. Structured interviewing; raising the psychometric properties of the employment interview, Personnel psychology 41(1) (1988), 25-42.
4. L. A. Zadeh. Fuzzy sets, Information and Control, 8 (1965), 338-353.
5. K. Atanassov. Intuitionistic fuzzy sets, Fuzzy sets and systems, 20 (1986), 87-96.
6. F. Smarandache. Neutrosophic set – a generalization of intuitionistic fuzzy set, Journal of Defence Resources Management, 1(1) (2010), 107-116.
7. F. Smarandache. Neutrosophic set – a generalization of intuitionistic fuzzy sets, International Journal of Pure and Applied Mathematics, 24(3) (2005), 287-297.
8. F. Smarandache. Linguistic paradoxes and tautologies, Libertas Mathematica, University of Texas at Arlington, IX (1999), 143-154.
9. F. Smarandache. A unifying field of logics. Neutrosophy: Neutrosophic probability, set and logic, American Research Press, Rehoboth, (1998).
10. H. Wang, F. Smarandache, Y. Zhang and R. Sunderraman. Single valued Neutrosophic Sets, Multi-space and Multi-structure, 4 (2010), 410-413.
11. P. Biswas, S. Pramanik and B. C. Giri. Entropy based grey relational analysis method for multi attribute decision making under single valued neutrosophic assessment, Neutrosophic set and systems, 2 (2014), 105-113.
12. J. Ye. Multi-criteria decision making method using the correlation co-efficient under single-valued neutrosophic environment, International Journal of General Systems, 42(4) (2013), 386-394.
13. G. Tuzkaya, S. Onut and B. Gulsan. An analysis network process approach for locating undesirable facilities: an example from Istanbul Turkey, Environ. Manag, 88(4) (2008), 970-983.
14. A. Masood Badri. Combining the analytic hierarchy process and goal programming for global facility location-allocation problem production Economics, 62 (5) (1999), 237–248.

15. F. Chang and S. Chung. Multi-criteria genetic optimization for distribution network problems, International Journal of Advanced Manufacturing Technology, 24 (7-8) (2004), 517–532.
16. GS. Liang and M. J. J. Wang. A fuzzy multi-criteria decision making method for facility site selection, International Journal of Production Research, 29(11) (1991), 231-233.
17. T. C. Chu. Facility location selection using fuzzy TOPSIS under group decisions. International Journal of uncertainty, Fuzziness and Knowledge based systems, 10(6) (2002), 687-701.
18. S. Pramanik and S. Dalapati. GRA based multi criteria decision making in generalized neutrosophic soft set environment, Global Journal of Engineering Science and Research Management, 3(5) (2016),153-169.
19. J. Ye. Multiple attribute group decision-making methods with unknown weights based on hybrid score-accuracy function under simplified neutrosophic environment, Un-published work.
20. H. Xiong, B. Wang, S. Wang and M. Li. A multi-criteria decision making approach based on fuzzy theory and credibility mechanism for logistics center location selection, Wuhan international Conference on e-Business, Summer (2014),89-96.

# E-LEARNING

**NOURAN M. RADWAN**

Assistant Lecturer, Information Systems Dept., Sadat Academy, Egypt
PhD Student, Information Systems Dept., Mansoura university, Egypt
E-mail: radwannouran@yahoo.com

# Neutrosophic Applications in E-learning: Outcomes, Challenges and Trends

## Abstract

There has been a sudden increase in the usage of elearning to support learner's learning process in higher education. Educational institutions are working in an increasingly competitive environment as many studies in elearning are implemented under complete information, while in the real world many uncertainty aspects do exist. This has resulted in emerging various approaches to handle uncertainty. Neutrosophic logic has been used to overcome the uncertainty of concepts that are associated with human expert judgments. This paper presents current trends to enhance elearning process by using neutrosophic to extract useful knowledge for selecting, evaluating, personalizing, and adapting elearning process.

## 1. Introduction

The word uncertainty is dealing with vague data, incomplete information, and imprecise knowledge regardless of what is the reason [1]. One of the significant problems of artificial intelligence is modeling uncertainty for solving real life situations [2]. Previous researches presented various models that handle uncertainty by simulating the process of human thinking [3,4]. Managing uncertainties is a goal for decision makers including indefinite cases where it is not true or false [5]. This leads to emerging approaches such as fuzzy, intuitionistic fuzzy, vague and neutrosophic models to give better attribute interpretations. Fuzzy, intuitionistic fuzzy, and vague models are limited as they cannot represent contradiction which are a feature of human thinking [6].

Smarandache [7] proposed neutrosophic logic as an extension of fuzzy logic of which variable x is described by triple values x= (t, i, f) where t is the degree of truth, f is the degree of false and i is the level of indeterminacy. Neutrosophic logic is capable to deal with contradictions which are true and false as the sum of components any number between 0 and $3^+$. An example of neutrosophic logic is as following; the argument "Tomorrow it will be sunny" does not mean a constant-valued components structure; this argument may be 60% true, 40% indeterminate and 35% false at a time; but at in a second time may change at 55% true, 40% indeterminate, and 45 % false according to new indications, provenances, etc. [8].

The usage of elearning has been increased in the last recent years in which learners have started using smart devices to access eLearning content. Also many elearning applications have been emerged to support universities in spreading educational resources to the learners [9]. Previous studies [10] in e-learning are implemented under complete information, while the real environment has uncertainty aspects. That is why traditional evaluation methods may not be virtuous. This leads us to suggest neutrosophic logic to give better attribute interpretations to enhance elearning.

Considering the above facts, this paper is organized as follows: Related work is described in Section 2. In Section 3, the need and outcome of using neutrosophic logic is discussed. Section 4 gives the trends and challenges of applying Neutrosophic to elearning.

## 2. Related Work

Neutrosophy is originated from "neuter" and "Sophia". Neuter means neutral in Latin and Sophia means wisdom in Greek. Neutrosophy means neutral thought knowledge [7]. Neutrosophic Logic was developed to represent mathematical model of uncertainty including vagueness, ambiguity, imprecision, and inconsistency. Expert systems, decision support system, belief system, and information fusion tend to depend not only on truth value, but also on false and indeterminacy values. So current systems which are dedicated to simulate human brain are constrained with strict conditions, whereas, Neutrosophic logic has its chance to simulate human thinking and to be utilized for real environment executions [8].

Aggarwal et al. in 2010 [11] propose block diagram of neutrosophic inference system to illustrate designing of neutrosophic classifier which is more flexible to get more accurate results. Aggarwal et al. in 2011 [12] suggest the possibility of extending the capabilities of the fuzzy systems by applying neutrosophic systems and incorporating neutrosophic logic in medical domain. Vagueness, imprecision, ambiguity, and inconsistency, should be presented in medical systems as medical diagnosis depends on available data and expert recognition, and avoiding uncertainty leads to misplaced accurate interpretation. Neutrosophic Cognitive Maps (NCM) for investigating the effect of critical factors of breast cancer is presented in [13]. A neutrosophic lung segmentation method was developed by [14] to improve the expectation-maximization analysis and morphological operations for our computer-aided detection segmentation. This method facilitates image analysis tasks and computer aided applications for lung abnormalities and improve the accuracy of lung segmentation, mostly for the cases affected by lung diseases.

Neutrosophic is used in many multiple criteria decision-making problems in real life such as personal choice in academia, project assessment, supplier selection, industry systems, and others areas of management systems [15] [16] [17] to support taking a correct decision from the available alternatives in uncertain environment. Researchers tend to use neutrosophic sets in various decision making applications as traditional crisp multiple criteria decision-making methods are not enough to handle uncertainty in real world cases because of the ambiguity of people thinking, it is more reasonable to simulate human thinking with handling contradictions which are true and false at the same time [18]. In 2016, Ye [19] suggests computation of trust value by integrating a neutrosophic logic with the proposed fuzzy based trust model that considers all the factors affecting the trust in ecommerce. As traditional models of trust fall in representing the indeterminacy values involved while capturing the perception of human. As it is concluded that imprecision of systems could be due to the deficiency of knowledge that received from human in the real world.

A proposed social Learning Management System (LMS) that integrates social activities in e-learning, and utilize a new set theory called the neutrosophic set to analyze social networks data conducted through learning activities is presented in [20]. A new approach based on neutrosophy is presented to provide better interpretation of the assessment results of the e-learning systems that are described by uncertainty aspects [21].

Another study [22] concerns the importance of social networks in e-learning systems. Recommender system has a significant role in e-learning as it supports e-learners in choosing among different learning objects and activities using different algorithms: C4.5, K-Means, support vector machine, and Apriori algorithms.

Neutrosophic sets [23] is proposed in order to evaluate the quality of learning objects based on the multi-criteria approach. Neutrosophic way of thinking help experts to represent their opinion in degrees of truth, false, and indeterminacy.

## 3. Outcomes of using Neutrosophic Logic

Neutrosophic idea is based on indeterminacy set that can deal with vagueness, imprecision; ambiguity and inconsistent information existing in real world. An example of neutrosophic problems is as follows: a vote with three symbols which are A, B and D ballots is occurred, in which some votes are indistinct, and it can't be determined if it is A, B or D. These indeterminate votes can be expressed with neutrosophic logic.

Therefore, indeterminacy can be handled by neutrosophic logic while other approaches neglect this point [7].

Fuzzy sets represent the membership without expressing the corresponding degree of non-membership so it provides an imperfect expression of uncertain information. The degree of non-membership in fuzzy sets is the complement of membership for fuzzy sets, Therefore the non-membership is not independent.

Intuitionistic fuzzy sets, as well as vague sets, are suitable in simulating the impreciseness of human understanding in decision making by representing degree of membership and non-membership, but it also cannot express indeterminacy degree which is the ignorance value between truth and false.

Indeterminate can be handled by neutrosophic logic which has the truth, indeterminacy and false membership functions as shown in Table 1[6].

Table 1. *Multivalued Logic Membership Function*

| | Fuzzy set | Intuitionistic Fuzzy | Vague | Neutrosophic |
|---|---|---|---|---|
| **Membership Function** | Degree of belonging | Degree of membership function and non-membership function | Degree of membership function and non-membership function | Degree of membership function, indeterminacy and non-membership function |
| | 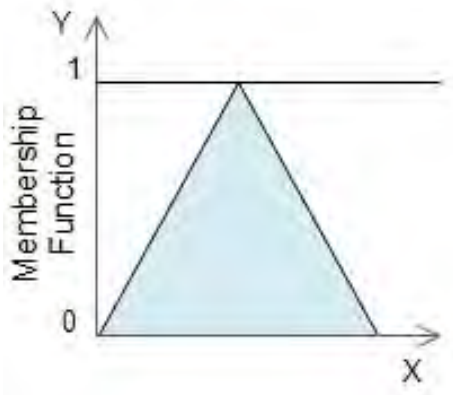  Fig 1. Type1 fuzzy membership function [1] | 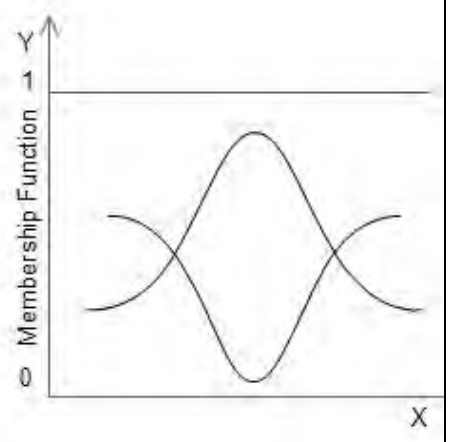  Figure 2. Intuitionistic Fuzzy Set [24] | 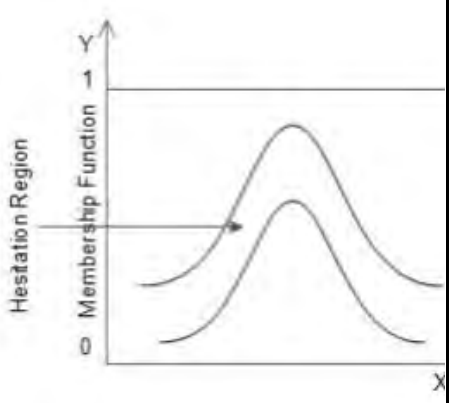  Figure 3. Vague Set [24] | 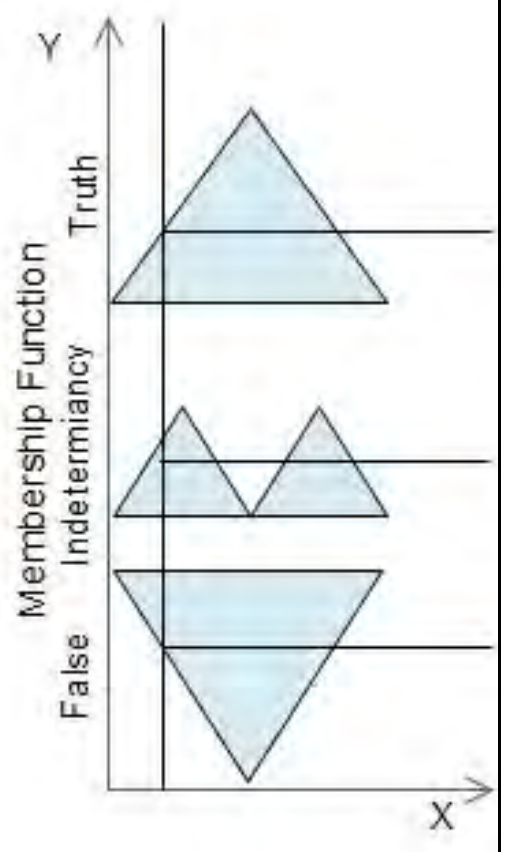  Figure 4. Neutrosophic Set [11] |

A better understanding of the need and the outcome of using neutrosophic logic is presented in this section. The uncertainties types include vagueness, imprecision, ambiguity, and inconsistency. *Vagueness* when available information is normally having a degrees of attribute; for example: "This man is nearly tall". *Imprecision* when information is not a definite value; for example: "The student performance for a task is between 80-85% ".  when available information has more than one meaning or refer to more than one subject; for example: "The flower color may be yellow or red". *Inconsistency* when obtainable information is conflicted or contradicted; for example:"the chance of raining tomorrow is 80%", it does not mean that the chance of not raining is 20%, since there might be hidden weather factors that is not aware of.

Fuzzy set describes vagueness, Intuitionistic fuzzy set is an extension of fuzzy sets which describes vagueness and imprecision by a range of membership values. Neutrosophic set describes vagueness, imprecision; ambiguity and inconsistent information that exists in real environment. Therefore, Neutrosophic logic handles indeterminacy of information while other approaches neglect this point.

## 4. Challenges and Trends of Neutrosophic Applications in E-learning

In the recent years, there has been increasing demand in incorporating of new technologies into educational processes. Effectiveness of elearning becomes a big challenge as elearning process is currently conducted in highly controlled way. Elearning challenges can be categorized according to their focus into: individual, course, technology and contextual as shown in Figure 5[9]. It is expected that neutrosophic logic utilized for enhancing eLearning environment as following:

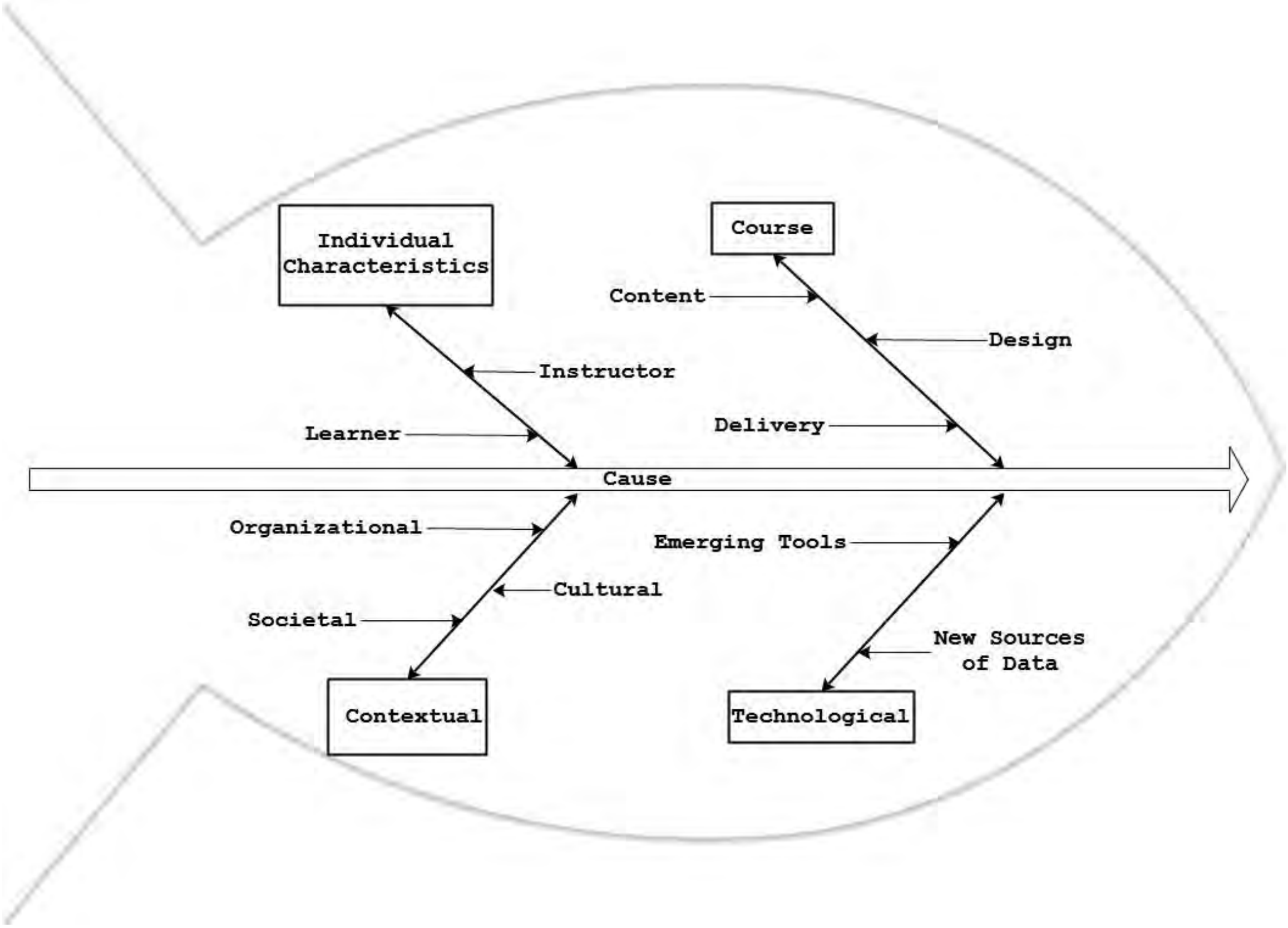

Figure 5. *E-Learning Challenges*

### 4.1 Neutrosophic Cognitive Map for E-learning Success Factors

According to the researchers' insights in the matter, elearning offer advantages over the traditional learning methods. The study of critical success factors helps decision makers to extract from the learning process the core activities that are essential for success. The investigation of the success factors from different perspectives such as learner's, instructor's and organization is needed [25]. Previous researches presented fuzzy cognitive maps and intuitionistic fuzzy logic by considering the expert's hesitancy in the determination of the relations between the concepts of a domain [26]. Further studies are needed to analyze and build a neutrosophic cognitive map for modelling critical success factors in elearning. Neutrosophic cognitive map extends the aggregation of the information from different resources under uncertain environment.

### 4.2 Neutrosophic Multi-Criteria Decision Making Methods for Elearning Software Selection

There are hundreds of elearning software available in the marketplace. Selecting the most suitable elearning that meets specific requirements is a problem of decision making. Many studies in elearning selection are implemented under complete information, while in the real world many uncertainty aspects do exist. As these systems were described by decision makers with vague, imprecise, ambiguous and inconsistent terms, it is understandable that traditional multi criteria decision making methods may not be effective [27] [28]. Further studies deal with presenting a hybrid neutrosophic multi criteria decision making method to handle indeterminacy of information.

### 4.3 Neutrosophic Expert System for Evaluating E-learning Applications

Expert system aims to represent the problem of uncertainty in knowledge to draw conclusion with the same level of accuracy as would a human expert do. Different evaluation models for e-learning quality attributes developed under the condition of the availability of complete information. Real environment is characterized by vagueness, imprecision, ambiguity, and inconsistency information, this problem leads researchers to use approaches that deals with uncertainty like fuzzy logic, intuitionistic fuzzy logic. This section suggests neutrosophic expert system for evaluation of elearning applications [6], [29].

### 4.4 Neutrosophic Logic Based System for E-learning Personalization

Generating the content according to learner's intellect is a current challenge in e-learning systems. Most of the e-learning systems evaluate the learner’s intellect level according to tests crisp responses that are taken during the learning process. However, many factors lead to uncertainty about the evaluation process. Further work will present a novel approach using neutrosophic logic to build an intelligent system that handles imprecision, vagueness, ambiguity, and inconsistence information about the learner’s assessment to personalize the learning material according to learner’s level is needed [30].

### 4.5 Adaptive test sheet generation in e-learning

Successful test sheet refers to the ability of questions to check the learner's cognitive skills in the most efficient manner. Designing and developing a test assignment for an adaptive e-learning system depends on organization of questions, concept, activity and learner level. Test adaptation can be done by utilizing one of learning styles models like Myers Brigs Type Indicator, Bloom’s Taxonomy, and David Kolb’s Model e-test classify in an e-learning environment. Adaptive test works toward providing electronic test sheet generation according to learner’s style in a customized surroundings. The incorporation of neutrosophic logic and learning style model provides a suitable way of assessment that has an important role in enhancing learner recognition, and performance. Traditional test includes questions with different difficulty levels to get the overall view about the learner’s ability despite of adaptive test that designed to ensure that learner is above a special ability value as it includes questions with definite difficulty level [31].

## 5. Conclusion

With the major changes in e-learning technology, there is a need to take into considerations the current trends and challenges of neutrosophic logic in elearning to add benefits to learners.

The chapter presents recent challenges and trends in neutrosophic applications. The neutrosophic logic has many achievements in different applications such as medical, decision making, ecommerce, and elearning. The outcome of neutrosophic logic is handling different uncertainty types vagueness, imprecision; ambiguity and inconsistent information exist in real world. Therefore, human thinking indeterminacy can be handled by neutrosophic logic while other approaches neglect this point. Furthermore, the study provides insights of neutrosophic applications challenges and trends in elearning. Future work will deal with talent elearning system to recommend training courses suitable for learner's talent in which neutrosophic is needed to identify learner's needs and skills. The integration of talent management and elearning system, improve the learner's task related skills.

## References

1. Uusitalo, L., Lehikoinen, A., Helle, I., & Myrberg, K. (2015). An overview of methods to evaluate uncertainty of deterministic models in decision support. Environmental Modelling & Software, 63, 24-31.
2. Motro, A., & Smets, P. (Eds.). (2012). Uncertainty management in information systems: from needs to solutions. Springer Science & Business Media,25-31.
3. Booker, J. M., & Ross, T. J. (2011). An evolution of uncertainty assessment and quantification. Scientia Iranica, 18(3), 669-676.
4. Liu, S., Sheng, K., & Forrest, J. (2012). On uncertain systems and uncertain models. Kybernetes, 41(5/6), 548-558.
5. Olson, D. L., & Wu, D. (2005). Decision making with uncertainty and data mining. In International Conference on Advanced Data Mining and Applications. Springer Berlin Heidelberg, 1-9.
6. Radwan, N. M., Senousy, M. B., & Alaa El Din, M. R. (2016). Approaches for managing uncertainty in learning management systems. Egyptian Computer Science Journal, 40(2),1-10.
7. Smarandache, Florentin. (1999). A unifying field in logics: neutrosophic logic. Philosophy,1-141.
8. Ansari, A. Q., Biswas, R., & Aggarwal, S. (2013). Neutrosophic classifier: an extension of fuzzy classifer. Applied Soft Computing, 13(1), 563-573.
9. Tsinakos, A. (2013). State of mobile learning around the world. Global Mobile Learning Implementations and Trends, 4-44.
10. Yigit, T., Isik, A. H., & Ince, M. (2014). Web-based learning object selection software using analytical hierarchy process. IET Software, 8(4), 174-183.
11. Aggarwal, S., Biswas, R., & Ansari, A. Q. (2010). Neutrosophic modeling and control. In Computer and Communication Technology (ICCCT), 2010 International Conference on IEEE, 718-723.
12. Ansari, A. Q., Biswas, R., & Aggarwal, S. (2011). Proposal for applicability of neutrosophic set theory in medical AI. International Journal of Computer Applications, 27(5), 5-11.
13. William, M. A., Devadoss, A. V., & Sheeba, J. J. (2013). A study on Neutrosophic cognitive maps (NCMs) by analyzing the Risk Factors of Breast Cancer. International Journal of Scientific & Engineering Research, 4(2), 1-4.
14. Guo, Y., Zhou, C., Chan, H. P., Chughtai, A., Wei, J., Hadjiiski, L. M., & Kazerooni, E. A. (2013). Automated iterative neutrosophic lung segmentation for image analysis in thoracic computed tomography. Medical physics, 40(8), 081912.
15. Mondal, K., & Pramanik, S. (2014). Multi-criteria group decision making approach for teacher recruitment in higher education under simplified neutrosophic environment. Neutrosophic Sets and Systems, 6, 28-34.

16. Mondal, K., & Pramanik, S. (2015). Neutrosophic decision making model of school choice. Neutrosophic Sets and Systems, 7, 62-68.
17. Biswas, P., Pramanik, S., & Giri, B. C. (2016). TOPSIS method for multi-attribute group decision-making under single-valued neutrosophic environment. Neural Computing and Applications, 27(3), 727-737.
18. Ye, J. (2015). An extended TOPSIS method for multiple attribute group decision making based on single valued neutrosophic linguistic numbers. Journal of Intelligent & Fuzzy Systems, 28(1), 247-255.
19. Aggarwal, S., & Bishnoi, A. (2016). Neutrosophic Trust Evaluation Model in B2C E-Commerce. In Hybrid Soft Computing Approaches, Springer India, 405-427.
20. Salama, A. A., Haitham, A., Manie, A. M., & Lotfy, M. M. (2014). Utilizing neutrosophic set in social network analysis e-learning systems. International Journal of Information Science and Intelligent Systems, 3(4), 61-72.
21. Albeanu, G., & Vlada, M. (2014). Neutrosophic approaches in e-learning assessment. In The International Scientific Conference eLearning and Software for Education " Carol I" National Defence University, 3, 435-441.
22. Salama, A. A., Eisa, M., ELhafeez, S. A., & Lotfy, M. M. (2015). Review of recommender systems algorithms utilized in social networks based e-learning systems & neutrosophic system. Neutrosophic Sets and Systems, 8, 32-40.
23. Madsen, H., Albeanu, G., Burtschy, B., & Popentiu-Vladicescu, F. (2015). Neutrosophic logic applied to decision making. In Intelligent Computing, Communication and Devices. Springer India, 1-7.
24. Lu, A., & Ng, W. (2005). Vague sets or intuitionistic fuzzy sets for handling vague data: which one is better?. In International Conference on Conceptual Modeling. Springer Berlin Heidelberg, 401-416.
25. Salmeron, J. L. (2009). Augmented fuzzy cognitive maps for modelling LMS critical success factors. Knowledge-based systems, 22(4), 275-278.
26. Kang, B., Deng, Y., Sadiq, R., & Mahadevan, S. (2012). Evidential cognitive maps. Knowledge-Based Systems, 35, 77-86.
27. Cavus, N. (2010). The evaluation of Learning Management Systems using an artificial intelligence fuzzy logic algorithm. Advances in Engineering Software, 41(2), 248-254.
28. Şahin, R., & Yiğider, M. (2014). A Multi-criteria neutrosophic group decision making metod based TOPSIS for supplier selection. arXiv preprint arXiv:1412.5077.
29. Radwan, N., Senousy, M. B., & Alaa El Din, M. (2016). Neutrosophic logic approach for evaluating learning management systems. Neutrosophic Sets and Systems, 3, 3-7.
30. Goyal, M., Yadav, D., & Tripathi, A. (2015). Intuitionistic fuzzy approach for adaptive presentation in an E-learning environment. In 2015 IEEE International Conference on Research in Computational Intelligence and Communication Networks (ICRCICN) IEEE, 108-113.
31. Goyal, M., Yadav, D., & Choubey, A. (2012). Fuzzy logic approach for adaptive test sheet generation in e-learning. In Technology Enhanced Education (ICTEE), 2012 IEEE International Conference on IEEE,1-4.

# GRAPH THEORY

SAID BROUMI[1*], MOHAMED TALEA[2], ASSIA BAKALI[3], FLORENTIN SMARANDACHE[4]

1, 2 Laboratory of Information processing, Faculty of Science Ben M'Sik, University Hassan II, B.P 7955, Sidi Othman, Casablanca, Morocco. E-mails: broumisaid78@gmail.com, taleamohamed@yahoo.fr
3 Ecole Royale Navale-Boulevard Sour Jdid, B.P 16303 Casablanca-Morocco. E-mail: assiabakali@yahoo.fr
4 Department of Mathematics, University of New Mexico, 705 Gurley Avenue, Gallup, NM 87301, USA.
E-mail: fsmarandache@gmail.com

# Single Valued Neutrosophic Graphs

## Abstract

The notion of single valued neutrosophic sets is a generalization of fuzzy sets, intuitionistic fuzzy sets. We apply the concept of single valued neutrosophic sets, an instance of neutrosophic sets, to graphs. We introduce certain types of single valued neutrosophic graphs (SVNG) and investigate some of their properties with proofs and examples.

## 1. Introduction

Neutrosophic sets (NSs) proposed by Smarandache [12, 13] is a powerful mathematical tool for dealing with incomplete, indeterminate and inconsistent information in real world. they are a generalization of the theory of fuzzy sets [24], intuitionistic fuzzy sets [21, 23] and interval valued intuitionistic fuzzy sets [22]. The neutrosophic sets are characterized by a truth-membership function (t), an indeterminacy-membership function (i) and a falsity-membership function (f) independently, which are within the real standard or nonstandard unit interval $]^{-}0, 1^{+}[$. In order to practice NS in real life applications conveniently, Wang et al. [16] introduced the concept of a single-valued neutrosophic sets (SVNS), a subclass of the neutrosophic sets. The SVNS is a generalization of intuitionistic fuzzy sets, in which three membership functions are independent and their value belong to the unit interval [0, 1]. Some more work on single valued neutrosophic sets and their extensions may be found on [2, 3, 4, 5,15, 17, 19, 20, 27, 28, 29, 30].

Graph theory has now become a major branch of applied mathematics and it is generally regarded as a branch of combinatorics. Graph is a widely used tool for solving a combinatorial problem in different areas such as geometry, algebra, number theory, topology, optimization, and computer science. Most important thing which is to be noted is that, when we have uncertainty regarding either the set of vertices or edges or both, the model becomes a fuzzy graph.

Lots of works on fuzzy graphs and intuitionistic fuzzy graphs [6, 7, 8, 25, 27] have been carried out and all of them have considered the vertex sets and edge sets as fuzzy and /or intuitionistic fuzzy sets. But, when the relations between nodes (or vertices) in problems are indeterminate, the fuzzy graphs and intuitionistic fuzzy graphs are failed. For this purpose, Samarandache [9, 10, 11, 14, 34] have defined four main categories of neutrosophic graphs, two based on literal indeterminacy (I), which called them; I-edge neutrosophic graph and I-vertex neutrosophic graph, these concepts are studied deeply and has gained popularity among the researchers due to its applications via real world problems [1, 33, 35]. The two others graphs are based on (t, i, f) components and called them; The (t, i, f)-Edge neutrosophic graph and the (t, i, f)-vertex neutrosophic graph, these concepts are not developed at all. In the literature the study of single valued neutrosophic graphs (SVN-graph) is still blank, we shall focus on the study of single valued neutrosophic graphs in this paper.

In this paper, some certain types of single valued neutrosophic graphs are developed and some interesting properties are explored.

## 2. Preliminaries

In this section, we mainly recall some notions related to neutrosophic sets, single valued neutrosophic sets, fuzzy graph and intuitionistic fuzzy graph relevant to the present work. See especially [6, 7, 12, 13, 16] for further details and background.

**Definition 2.1 [12]**. Let X be a space of points (objects) with generic elements in X denoted by x; then the neutrosophic set A (NS A) is an object having the form $A = \{< x: T_A(x), I_A(x), F_A(x)>, x \in X\}$, where the functions T, I, F: $X \rightarrow ]^-0,1^+[$ define respectively the a truth-membership function, an indeterminacy-membership function, and a falsity-membership function of the element $x \in X$ to the set A with the condition:

$$^-0 \leq T_A(x)+ I_A(x)+ F_A(x)\leq 3^+. \quad (1)$$

The functions $T_A(x)$, $I_A(x)$ and $F_A(x)$ are real standard or nonstandard subsets of $]^-0,1^+[$.

Since it is difficult to apply NSs to practical problems, Wang et al. [16] introduced the concept of a SVNS, which is an instance of a NS and can be used in real scientific and engineering applications.

**Definition 2.2 [16].** Let X be a space of points (objects) with generic elements in X denoted by x. A single valued neutrosophic set A (SVNS A) is characterized by truth-membership function $T_A(x)$, an indeterminacy-membership function $I_A(x)$, and a falsity-membership function $F_A(x)$. For each point x in X $T_A(x), I_A(x), F_A(x) \in [0, 1]$. A SVNS A can be written as

$$A = \{< x: T_A(x), I_A(x), F_A(x)>, x \in X\} \quad (2)$$

**Definition 2.3[6].** A fuzzy graph is a pair of functions G = (σ, μ) where σ is a fuzzy subset of a non empty set V and μ is a symmetric fuzzy relation on σ. i.e $\sigma: V \rightarrow [0,1]$ and

$\mu: VxV\rightarrow[0,1]$ such that $\mu(uv) \leq \sigma(u) \wedge \sigma(v)$ for all u, v ∈ V where uv denotes the edge between u and v and $\sigma(u) \wedge \sigma(v)$ denotes the minimum of σ(u) and σ(v). σ is called the fuzzy vertex set of V and μ is called the fuzzy edge set of E.

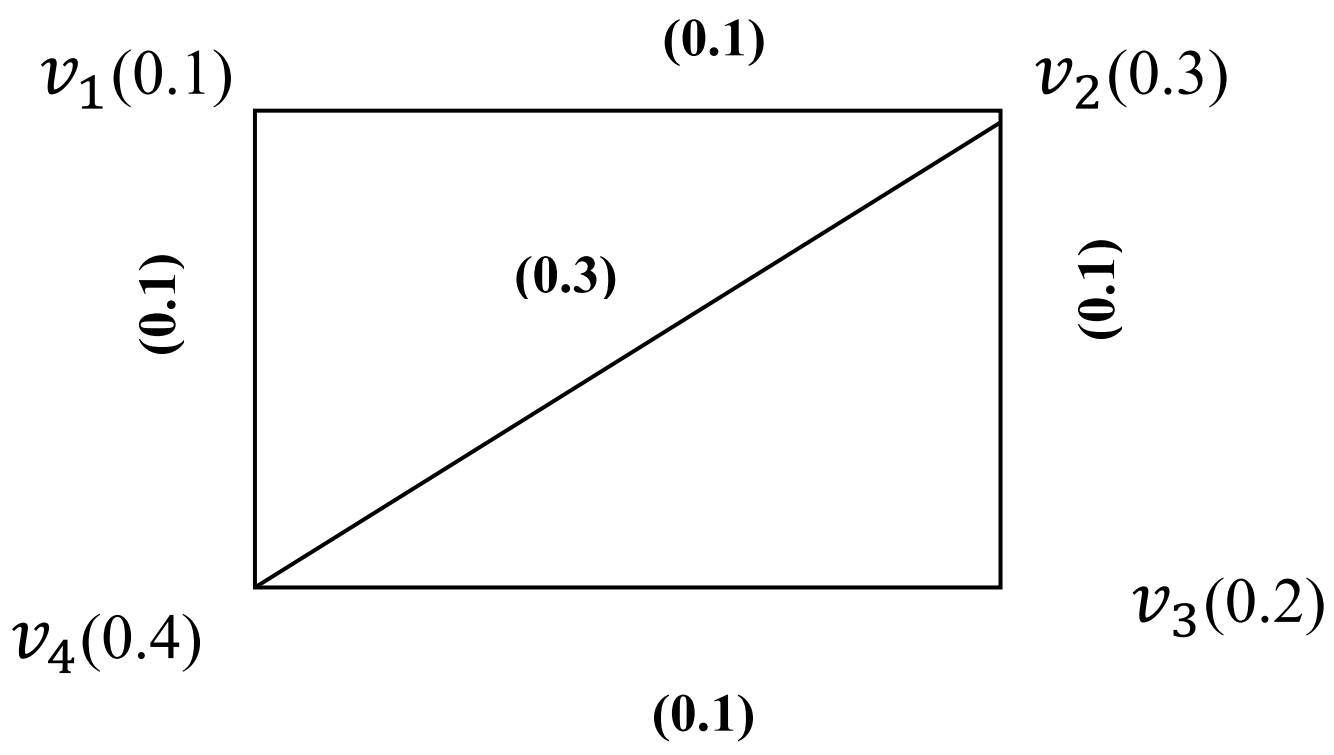

Figure 1: *Fuzzy Graph*

**Definition 2.4 [6].** The fuzzy subgraph H = (τ, ρ) is called a fuzzy subgraph of G = (σ, μ)

If τ(u) ≤ σ(u) for all u ∈ V and ρ (u, v) ≤ μ(u, v) for all u, v ∈ V.

**Definition 2.5 [7].** An Intuitionistic fuzzy graph is of the form G = (V, E) where

i. V={$v_1$, $v_2$,…., $v_n$} such that $\mu_1$: V→ [0,1] and $\gamma_1$: V → [0,1] denote the degree of membership and nonmembership of the element $v_i$ ∈ V, respectively, and 0 ≤ $\mu_1(v_i)$ + $\gamma_1(v_i)$) ≤ 1 for every $v_i$ ∈ V, (i = 1, 2, ……. n),

ii. E ⊆ V x V where $\mu_2$: VxV→[0,1] and $\gamma_2$: VxV→ [0,1] are such that

$\mu_2(v_i, v_j) \leq \min [\mu_1(v_i), \mu_1(v_j)]$ and $\gamma_2(v_i, v_j) \geq \max [\gamma_1(v_i), \gamma_1(v_j)]$

and $0 \leq \mu_2(v_i, v_j) + \gamma_2(v_i, v_j) \leq 1$ for every ($v_i$, $v_j$) ∈ E, (i, j = 1,2, ……. n)

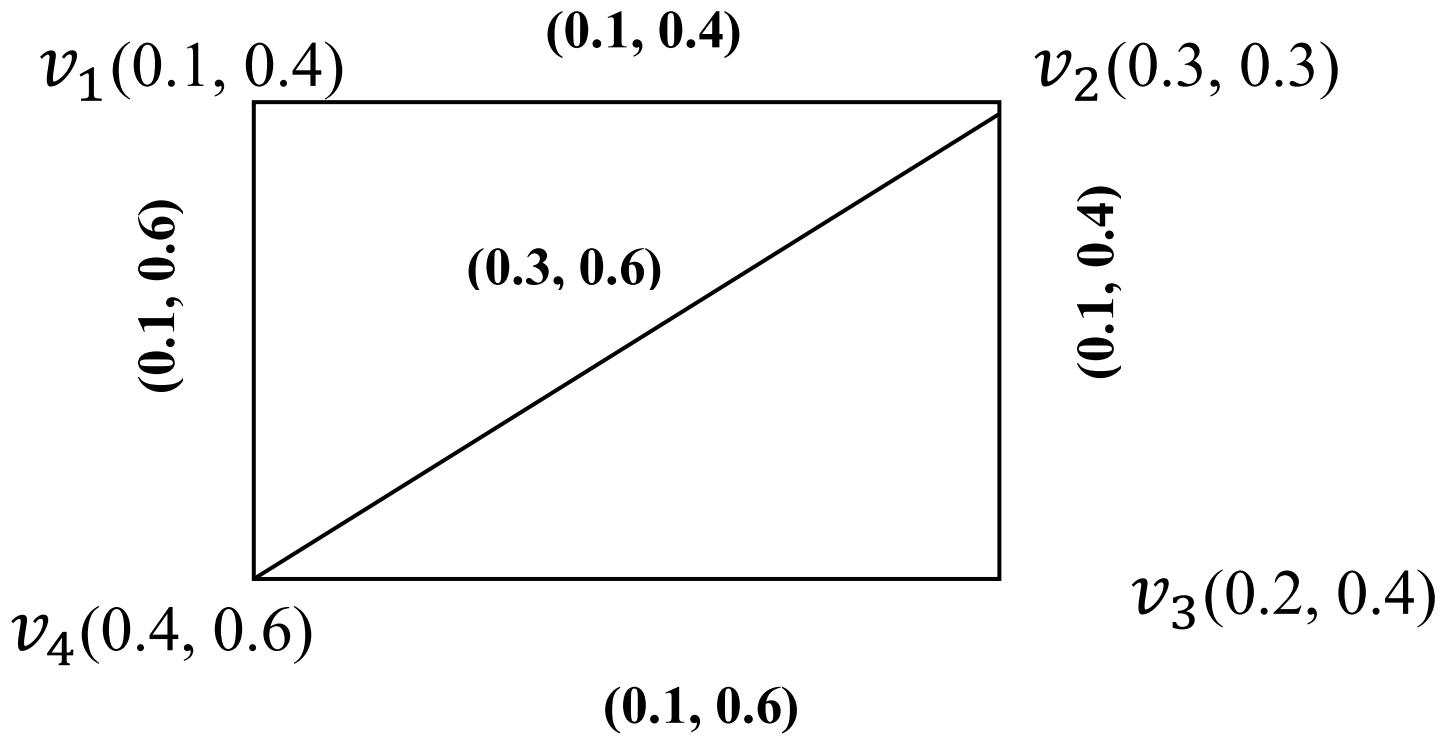

Figure 2: *Intuitionistic Fuzzy Graph*

**Definition 2.6 [31].** Let $A = (T_A, I_A, F_A)$ and $B = (T_B, I_B, F_B)$ be single valued neutrosophic sets on a set *X*. If $A = (T_A, I_A, F_A)$ is a single valued neutrosophic relation on a set *X*, then $A = (T_A, I_A, F_A)$ is called a single valued neutrosophic relation on $B = (T_B, I_B, F_B)$ if

$T_B(x, y) \leq \min(T_A(x), T_A(y))$

$I_B$(x, y) ≥ max($I_A$(x), $I_A$(y)) and

$F_B$(x, y) ≥ max($F_A$x), $F_A$(y)) for all x, y ∈ X.

A single valued neutrosophic relation $A$ on $X$ is called symmetric if $T_A(x, y) = T_A(y, x)$, $I_A(x, y) = I_A(y, x)$, $F_A(x, y) = F_A(y, x)$ and $T_B(x, y) = T_B(y, x)$, $I_B(x, y) = I_B(y, x)$ and $F_B(x, y) = F_B(y, x)$, for all $x, y \in X$.

## 3. Single Valued Neutrosophic Graphs

Throught this paper, we denote $G^*$ = (V, E) a crisp graph, and G = (A, B) a single valued neutrosophic graph

**Definition 3.1.** A single valued neutrosophic graph (SVN-graph) with underlying set V is defined to be a pair G= (A, B) where

1.The functions $T_A$:V→[0, 1], $I_A$:V→[0, 1] and $F_A$:V→[0, 1] denote the degree of truth-membership, degree of indeterminacy-membership and falsity-membership of the element $v_i$ ∈ V, respectively, and

$$0 \leq T_A(v_i) + I_A(v_i) + F_A(v_i) \leq 3 \text{ for all } v_i \in V \text{ (i=1, 2, …, n)}$$

2. The functions $T_B$: E ⊆ V x V →[0, 1], $I_B$:E ⊆ V x V →[0, 1] and $F_B$: E ⊆ V x V →[0, 1] are defined by

$T_B(\{v_i, v_j\}) \leq \min [T_A(v_i), T_A(v_j)]$,

$I_B(\{v_i, v_j\}) \geq \max [I_A(v_i), I_A(v_j)]$ and

$F_B(\{v_i, v_j\}) \geq \max [F_A(v_i), F_A(v_j)]$

Denotes the degree of truth-membership, indeterminacy-membership and falsity-membership of the edge $(v_i, v_j)$ ∈ E respectively, where

$$0 \leq T_B(\{v_i, v_j\}) + I_B(\{v_i, v_j\}) + F_B(\{v_i, v_j\}) \leq 3 \text{ for all } \{v_i, v_j\} \in E \text{ (i, j = 1, 2, …, n)}$$

We call A the single valued neutrosophic vertex set of V, B the single valued neutrosophic edge set of E, respectively, Note that B is a symmetric single valued neutrosophic relation on A. We use the notation $(v_i, v_j)$ for an element of E Thus, G = (A, B) is a single valued neutrosophic graph of G*= (V, E) if

$T_B(v_i, v_j) \leq \min [T_A(v_i), T_A(v_j)]$,

$I_B(v_i, v_j) \geq \max [I_A(v_i), I_A(v_j)]$ and

$F_B(v_i, v_j) \geq \max [F_A(v_i), F_A(v_j)]$ for all $(v_i, v_j) \in E$

**Example 3.2.** Consider a graph $G^*$ such that V= $\{v_1, v_2, v_3, v_4\}$, E=$\{v_1v_2, v_2v_3, v_3v_4, v_4v_1\}$. Let A be a single valued neutrosophic subset of V and let B a single valued neutrosophic subset of E denoted by

| | $v_1$ | $v_2$ | $v_3$ | $v_4$ |
|---|---|---|---|---|
| $T_A$ | 0.5 | 0.6 | 0.2 | 0.4 |
| $I_A$ | 0.1 | 0.3 | 0.3 | 0.2 |
| $F_A$ | 0.4 | 0.2 | 0.4 | 0.5 |

| | $v_1v_2$ | $v_2v_3$ | $v_3v_4$ | $v_4v_1$ |
|---|---|---|---|---|
| $T_B$ | 0.2 | 0.3 | 0.2 | 0.1 |
| $I_B$ | 0.3 | 0.3 | 0.3 | 0.2 |
| $F_B$ | 0.4 | 0.4 | 0.4 | 0.5 |

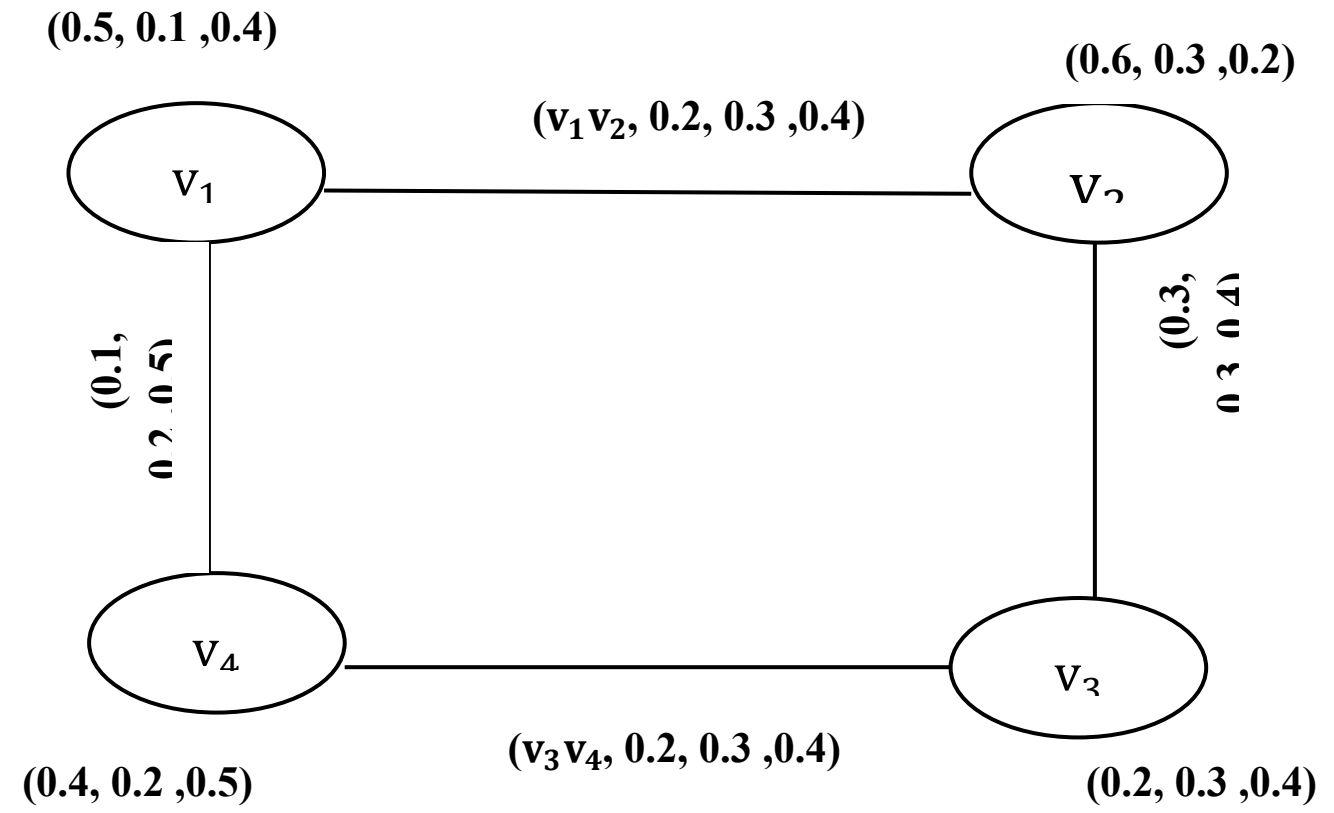

Figure 3: G: *Single valued neutrosophic graph*

In figure 3, (i) ($v_1$,0.5, 0.1,0.4) is a single valued neutrosophic vertex or SVN-vertex

(ii) ($v_1v_2$, 0.2, 0.3, 0.4) is a single valued neutrosophic edge or SVN-edge

(iii) ($v_1$, 0.5, 0.1, 0.4) and ($v_2$, 0.6, 0.3, 0.2) are single valued neutrosophic adjacent vertices.

(iv) ($v_1v_2$, 0.2, 0.3, 0.4) and ($v_1v_4$, 0.1, 0.2, 0.5) are a single valued neutrosophic adjacent edge.

**Note 1.** (i) When $T_{Bij} = I_{Bij} = F_{Bij}$ for some i and j, then there is no edge between $v_i$ and $v_j$ .

Otherwise there exists an edge between $v_i$ and $v_j$ .

(ii)If one of the inequalities is not satisfied in (1) and (2), then G is not an SVNG

The single valued neutrosophic graph G depicted in figure 3 is represented by the following adjacency matrix $\boldsymbol{M_G}$

$$\boldsymbol{M_G} = \begin{bmatrix} (0.5, 0.1, 0.4) & (0.2, 0.3, 0.4) & (0, 0, 0) & (0.1, 0.2, 0.5) \\ (0.2, 0.3, 0.4) & (0.6, 0.3, 0.2) & (0.3, 0.3, 0.4) & (0, 0, 0) \\ (0, 0, 0) & (0.3, 0.3, 0.4) & (0.2, 0.3, 0.4) & (0.2, 0.3, 0.4) \\ (0.1, 0.2, 0.5) & (0, 0, 0) & (0.2, 0.3, 0.4) & (0.4, 0.2, 0.5) \end{bmatrix}$$

**Definition 3.3.** A partial SVN-subgraph of SVN-graph G= (A, B) is a SVN-graph H = ( $\boldsymbol{V'}$, $\boldsymbol{E'}$) such that

(i) $\boldsymbol{V'} \subseteq \boldsymbol{V}$, where $\boldsymbol{T'_{Ai}} \leq \boldsymbol{T_{Ai}}$, $\mathbf{I'_{Ai}} \geq \mathbf{I_{Ai}}$, $\boldsymbol{F'_{Ai}} \geq \boldsymbol{F_{Ai}}$ for all $\boldsymbol{v_i} \in \boldsymbol{V}$.

(ii) $\boldsymbol{E'} \subseteq \boldsymbol{E}$, where $\boldsymbol{T'_{Bij}} \leq \boldsymbol{T_{Bij}}$, $\mathbf{I'_{Bij}} \geq \mathbf{I_{Bij}}$, $\boldsymbol{F'_{Bij}} \geq \boldsymbol{F_{Bij}}$ for all $(\boldsymbol{v_i}\ \boldsymbol{v_j}) \in \boldsymbol{E}$.

**Definition 3.4.** A SVN-subgraph of SVN-graph G= (V, E) is a SVN-graph H = ( $\boldsymbol{V'}$, $\boldsymbol{E'}$) such that

(i) $\boldsymbol{V'} = \boldsymbol{V}$, where $\boldsymbol{T'_{Ai}} = \boldsymbol{T_{Ai}}$, $\mathbf{I'_{Ai}} = \mathbf{I_{Ai}}$, $\boldsymbol{F'_{Ai}} = \boldsymbol{F_{Ai}}$ for all $\boldsymbol{v_i}$ in the vertex set of $\boldsymbol{V'}$.

(ii) $\boldsymbol{E'} = \boldsymbol{E}$, where $\boldsymbol{T'_{Bij}} = \boldsymbol{T_{Bij}}$, $\mathbf{I'_{Bij}} = \mathbf{I_{Bij}}$, $\boldsymbol{F'_{Bij}} = \boldsymbol{F_{Bij}}$ for every $(\boldsymbol{v_i}\ \boldsymbol{v_j}) \in \boldsymbol{E}$ in the edge set of $\boldsymbol{E'}$.

**Example 3.5.** $\mathbf{G_1}$ in Figure 4 is a SVN-graph. $\mathbf{H_1}$ in Figure 5 is a partial SVN-subgraph and

$\mathbf{H_2}$ in Figure 6 is a SVN-subgraph of $\mathbf{G_1}$

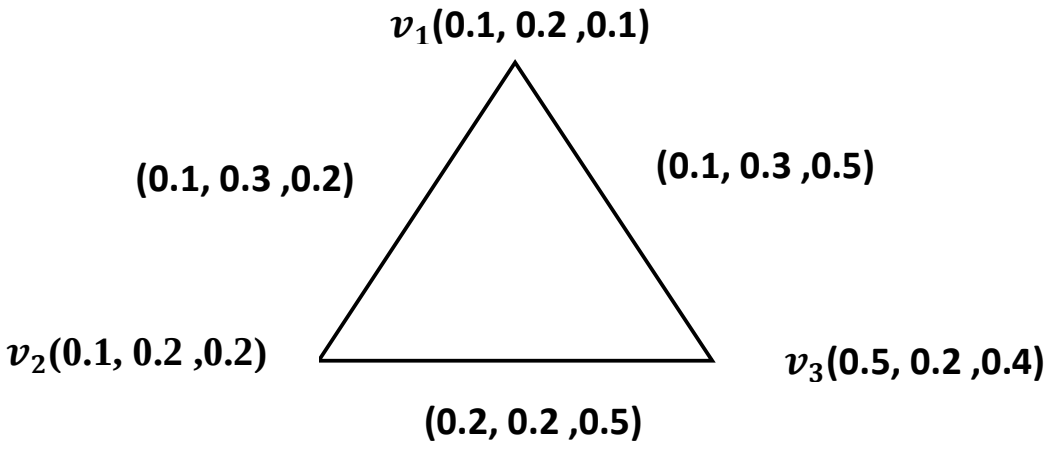

**Figure 4:** $G_1$, a single valued neutrosophic graph

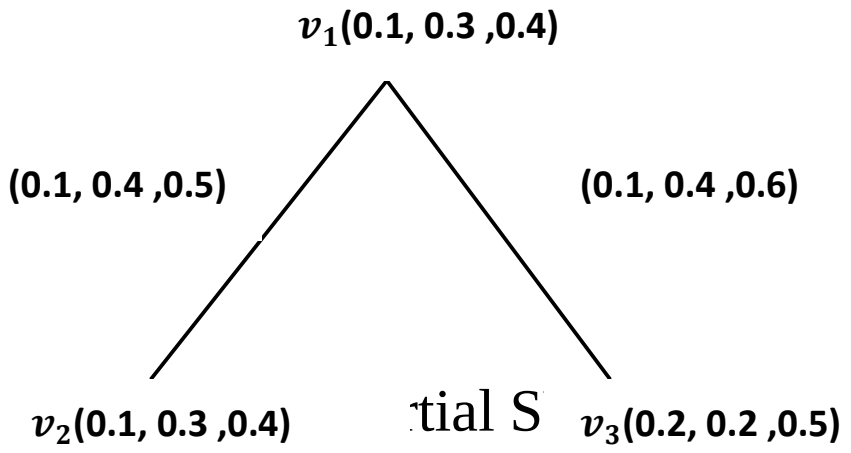

**F** tial S $G_1$

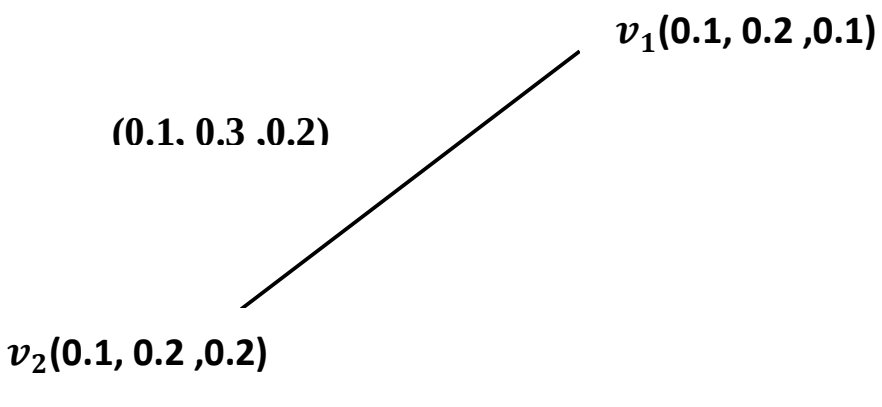

**Figure 6:** $H_2$, a SVN-subgraph of $G_1$.

**Definition 3.6**. The two vertices are said to be adjacent in a single valued neutrosophic graph G= (**A**, **B**) if $T_B(v_i, v_j) = \min\,[T_A(v_i), T_A(v_j)]$, $I_B(v_i, v_j) = \max\,[I_A(v_i), I_A(v_j)]$ and $F_B(v_i, v_j) = \max[F_A(v_i), F_A(v_j)]$. In this case, $v_i$ and $v_j$ are said to be neighbours and ($v_i$, $v_j$) is incident at $v_i$ and $v_j$ also.

**Definition 3.7**. A path P in a single valued neutrosophic graph G= (**A**, **B**) is a sequence of distinct vertices $v_0, v_1, v_3, \ldots v_n$ such that $T_B(v_{i-1}, v_i) > 0$, $I_B(v_{i-1}, v_i) > 0$ and $F_B(v_{i-1}, v_i) > 0$ for $0 \le i \le 1$. Here n ≥ 1 is called the length of the path P. A single node or vertex $v_i$ may also be considered as a path. In this case the path is of the length (0, 0, 0). The consecutive pairs $(v_{i-1}, v_i)$ are called edges of the path. We call P a cycle if $v_0 = v_n$ and n ≥3.

**Definition 3.8.** A single valued neutrosophic graph G= (A, B) is said to be connected if every pair of vertices has at least one single valued neutrosophic path between them, otherwise it is disconnected.

**Definition 3.9.** A vertex $v_j \in V$ of single valued neutrosophic graph G= (**A**, **B**) is said to be an isolated vertex if there is no effective edge incident at $v_j$.

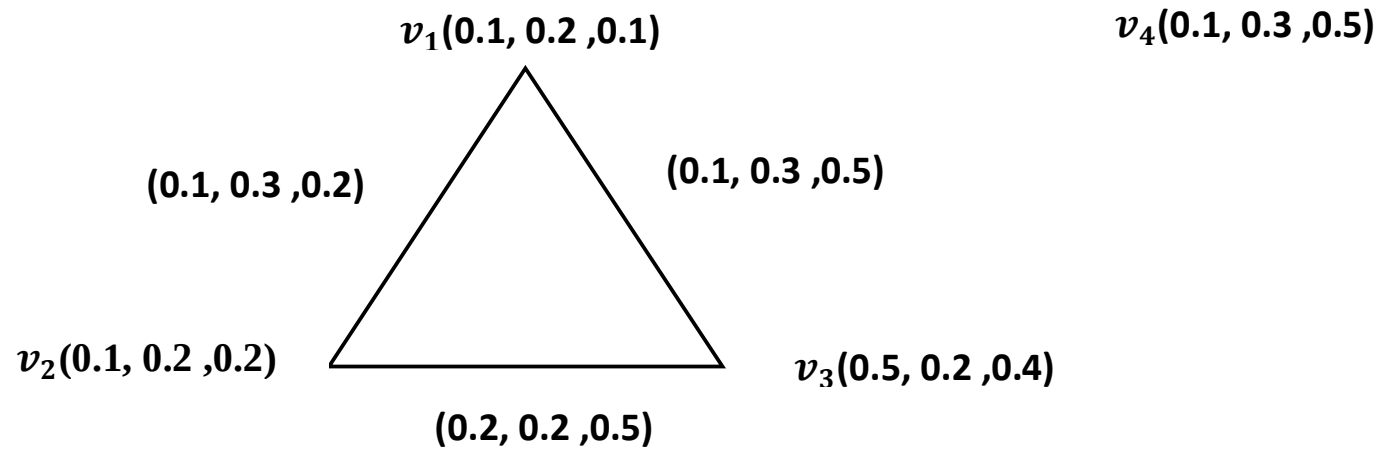

**Figure 7: Example of single valued neutrosophic graph**

In figure 7, the single valued neutrosophic vertex $v_4$ is an isolated vertex.

**Definition 3.10.** A vertex in a single valued neutrosophic G= (**A**, **B**) having exactly one neighbor is called ***a pendent vertex***. Otherwise, it is called ***non-pendent vertex***. An edge in a single valued neutrosophic graph incident with a pendent vertex is called a ***pendent edge***. Otherwise it is called ***non-pendent edge***. A vertex in a single valued neutrosophic graph adjacent to the pendent vertex is called **a support** of the pendent edge

**Definition 3.11.** A single valued neutrosophic graph G= (A, B) that has neither self loops nor parallel edge is called **simple single valued neutrosophic graph.**

**Definition 3.12.** When a vertex $\mathbf{v_i}$ is end vertex of some edges ($\mathbf{v_i}$, $\mathbf{v_j}$) of any SVN-graph G = (A, B). Then $\mathbf{v_i}$ and ($\mathbf{v_i}$, $\mathbf{v_j}$) are said to be **incident** to each other.

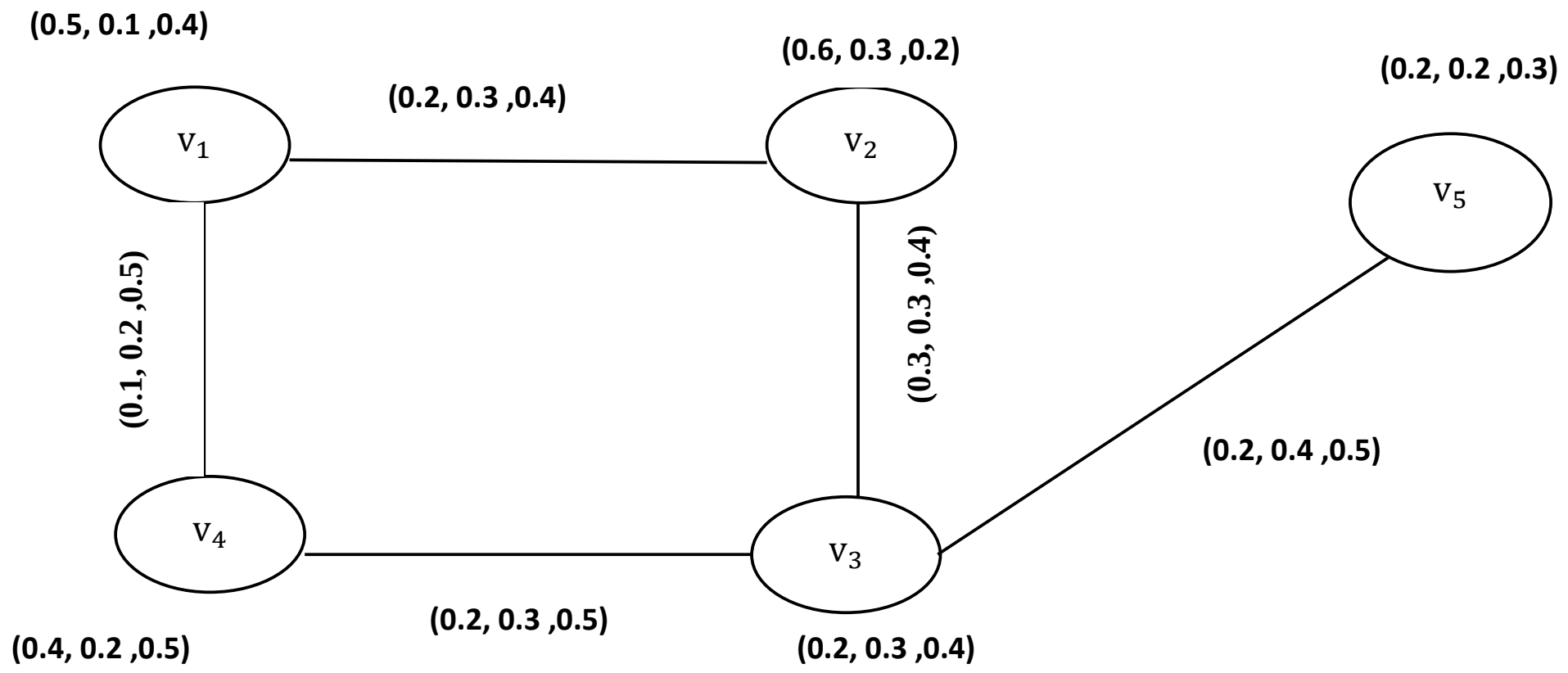

**Figure 8: Incident SVN-graph.**

In this graph $v_2v_3$, $v_3v_4$ and $v_3v_5$ are incident on $v_3$.

**Definition 3.13.** Let G= (**A**, **B**) be a single valued neutrosophic graph. Then the degree of any vertex **v** is sum of degree of truth-membership, sum of degree of indeterminacy-membership and sum of degree of falsity-membership of all those edges which are incident on vertex **v** denoted by d(v)= ($d_T(v)$, $d_I(v)$, $d_F(v)$) where

$d_T(v)=\sum_{u\neq v} T_B(u,v)$ denotes degree of truth-membership vertex.

$d_I(v)=\sum_{u\neq v} I_B(u,v)$ denotes degree of indeterminacy-membership vertex.

$d_F(v)=\sum_{u\neq v} F_B(u,v)$ denotes degree of falsity-membership vertex.

**Example 3.14.** Let us consider a single valued neutrosophic graph G= (A, B) of $G^*$ = (V, E) where V = {$v_1$, $v_2$, $v_3$, $v_4$} and E= {$v_1v_2$, $v_2v_3$, $v_3v_4$ , $v_4v_1$}.

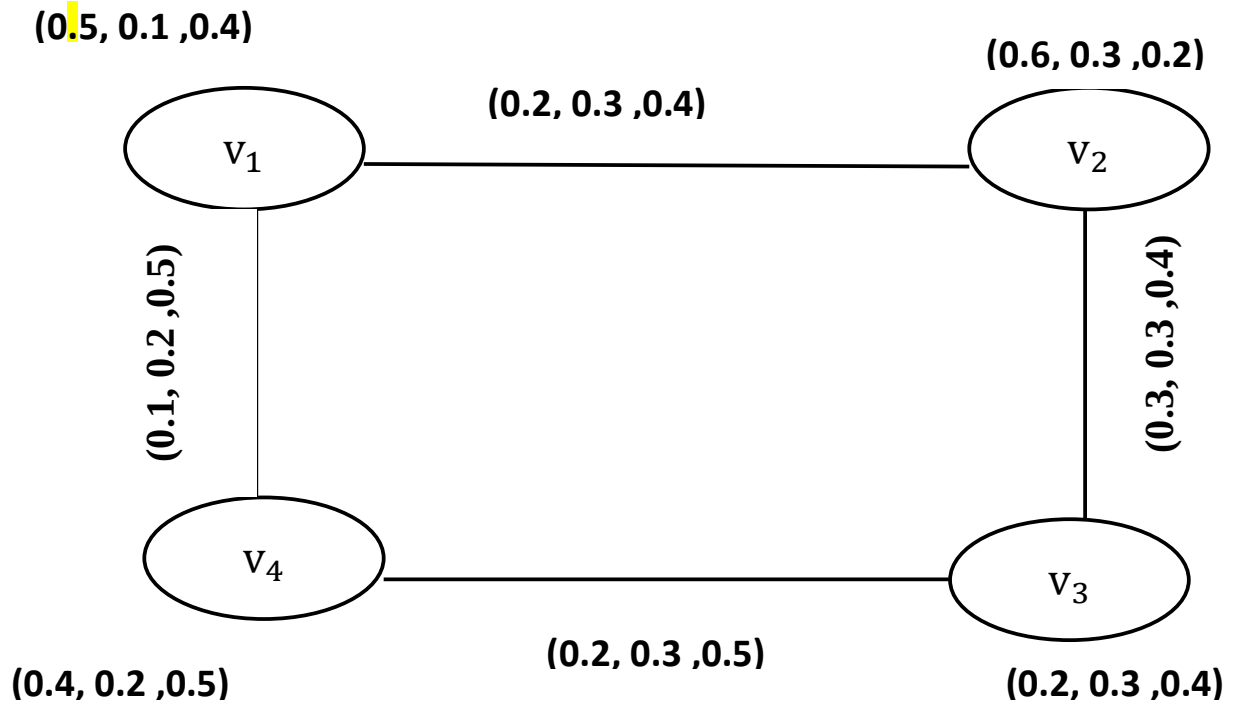

**Figure 9: Degree of vertex of single valued neutrosophic graph**

We have, the degree of each vertex as follows:

$d(v_1)$= ( 0.3, 0.5, 0.9), $d(v_2)$= ( 0.5, 0.6, 0.8), $d(v_3)$= ( 0.5, 0.6, 0.9), $d(v_4)$= ( 0.3, 0.5, 1)

**Definition 3.15.** A single valued neutrosophic graph G= (A, B) is called constant if degree of each vertex is k = ($k_1$, $k_2$, $k_3$). That is, d ($v$) = ($k_1$, $k_2$, $k_3$) for all $v \in$ V.

**Example 3.16.** Consider a single valued neutrosophic graph G such that V ={$v_1$, $v_2$, $v_3$, $v_4$} and E={$v_1v_2$, $v_2v_3$, $v_3v_4$ , $v_4v_1$}.

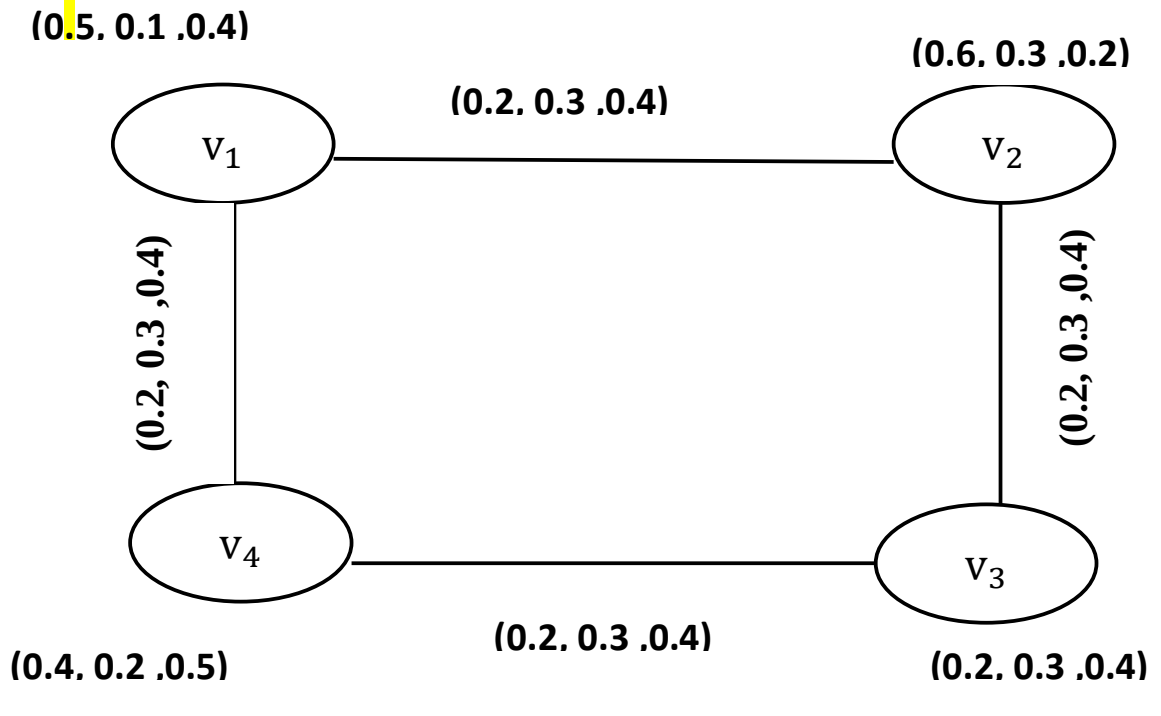

**Figure 10:** Constant SVN-graph.

Clearly, G is constant SVN-graph since the degree of $\mathbf{v_1}$, $\mathbf{v_2}$, $\mathbf{v_3}$ and $\mathbf{v_4}$ is (0.4, 0.6, 0.8).

**Definition 3.17.** A single valued neutrosophic graph G = (A, B) of $G^*$= (V, E) is called strong single valued neutrosophic graph if

$T_B(v_i, v_j) = \min\,[T_A(v_i),\ T_A(v_j)]$

$I_B(v_i, v_j) = \max\,[I_A(v_i),\ I_A(v_j)]$

$F_B(v_i, v_j) = \max\,[F_A(v_i), F_A(v_j)]$

For all $(v_i, v_j) \in$ E.

**Example 3.18.** Consider a graph $G^*$ such that V= $\{v_1, v_2, v_3, v_4\}$, E= $\{v_1v_2, v_2v_3, v_3v_4, v_4v_1\}$. Let A be a single valued neutrosophic subset of V and let B a single valued neutrosophic subset of E denoted by

| | $v_1$ | $v_2$ | $v_3$ | $v_4$ |
|---|---|---|---|---|
| $T_A$ | 0.5 | 0.6 | 0.2 | 0.4 |
| $I_A$ | 0.1 | 0.3 | 0.3 | 0.2 |
| $F_A$ | 0.4 | 0.2 | 0.4 | 0.5 |

| | $v_1v_2$ | $v_2v_3$ | $v_3v_4$ | $v_4v_1$ |
|---|---|---|---|---|
| $T_B$ | 0.5 | 0.2 | 0.2 | 0.4 |
| $I_B$ | 0.3 | 0.3 | 0.3 | 0.2 |
| $F_B$ | 0.4 | 0.4 | 0.5 | 0.5 |

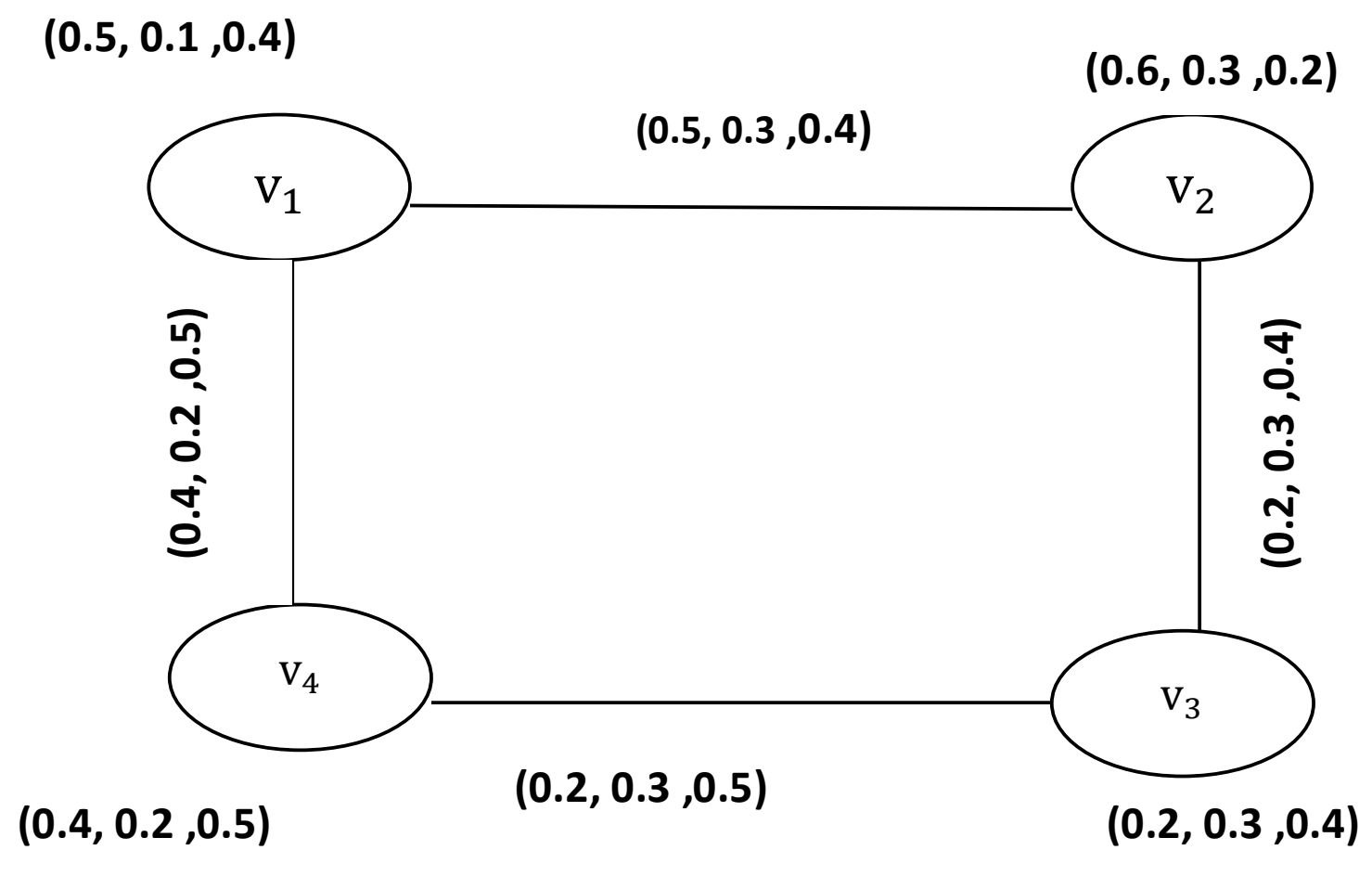

**Figure 11**: Strong SVN-graph

By routing computations, it is easy to see that G is a strong single valued neutrosophic of $G^*$.

**Proposition 3.19.** A single valued neutrosophic graph is the generalization of fuzzy graph

**Proof:** Suppose G= (V, E) be a single valued neutrosophic graph. Then by setting the indeterminacy- membership and falsity- membership values of vertex set and edge set equals to zero reduces the single valued neutrosophic graph to fuzzy graph.

**Proposition 3.20.** A single valued neutrosophic graph is the generalization of intuitionistic fuzzy graph.

**Proof:** Suppose G = (V, E) be a single valued neutrosophic graph. Then by setting the indeterminacy- membership value of vertex set and edge set equals to zero reduces the single valued neutrosophic graph to intuitionistic fuzzy graph.

**Definition 3.21.** The complement of a single valued neutrosophic graph G (A, B) on $G^*$ is a single valued neutrosophic graph $\bar{G}$ on $G^*$ where:

1. $\bar{A}$ =A

2. $\overline{T_A}(v_i)= T_A(v_i)$, $\overline{I_A}(v_i)= I_A(v_i)$, $\overline{F_A}(v_i) = F_A(v_i)$, for all $v_j$ ∈ V.

3. $\overline{T_B}(v_i, v_j)$= min $[T_A(v_i), T_A(v_j)]$ -$T_B(v_i, v_j)$

$\overline{I_B}(v_i, v_j)$= max $[I_A(v_i), I_A(v_j)]$ -$I_B(v_i, v_j)$ and

$\overline{F_B}(v_i, v_j)$= max $[F_A(v_i), F_A(v_j)]$ -$F_B(v_i, v_j)$, For all $(v_i, v_j)$ ∈ E

**Remark 3.22.** if G= (V, E) is a single valued neutrosophic graph on $G^*$. Then from above definition, it follow that $\bar{\bar{G}}$ is given by the single valued neutrosophic graph $\bar{\bar{G}}$ =($\bar{\bar{V}}$ , $\bar{\bar{E}}$ ) on $G^*$ where $\bar{\bar{V}}$ =V and $\overline{\overline{T_B}}(v_i, v_j)$= min $[T_A(v_i), T_A(v_j)]$-$T_B(v_i, v_j)$,

$\overline{\overline{I_B}}(v_i, v_j)$= min $[I_A(v_i), I_A(v_j)]$-$I_B(v_i, v_j)$,and

$\overline{\overline{F_B}}(v_i, v_j)$ = min $[F_A(v_i), F_A(v_j)]$-$F_B(v_i, v_j)$ For all $(v_i, v_j)$ ∈ E.

Thus $\overline{\overline{T_B}}$ =$T_B$, $\overline{\overline{I_B}}$ =$I_B$, and $\overline{\overline{F_B}}$ =$F_B$ on V, where E =( $T_B$, $I_B$, $F_B$) is the single valued neutrosophic relation on V. For any single valued neutrosophic graph G, $\bar{G}$ is strong single valued neutrosophic graph and G ⊆ $\bar{G}$.

**Proposition 3.23.** G= $\bar{\bar{G}}$ if and only if G is a strong single valued neutrosophic graph.

**Proof.** it is obvious.

**Definition 3.24.** A strong single valued neutrosophic graph G is called self complementary if G≅ $\bar{G}$. Where $\bar{G}$ is the complement of single valued neutrosophic graph G.

**Example 3.25.** Consider a graph $G^*$ = (V, E) such that V = {$v_1$, $v_2$, $v_3$, $v_4$}, E= {$v_1v_2$, $v_2v_3$, $v_3v_4$, $v_1v_4$}. Consider a single valued neutrosophic graph G.

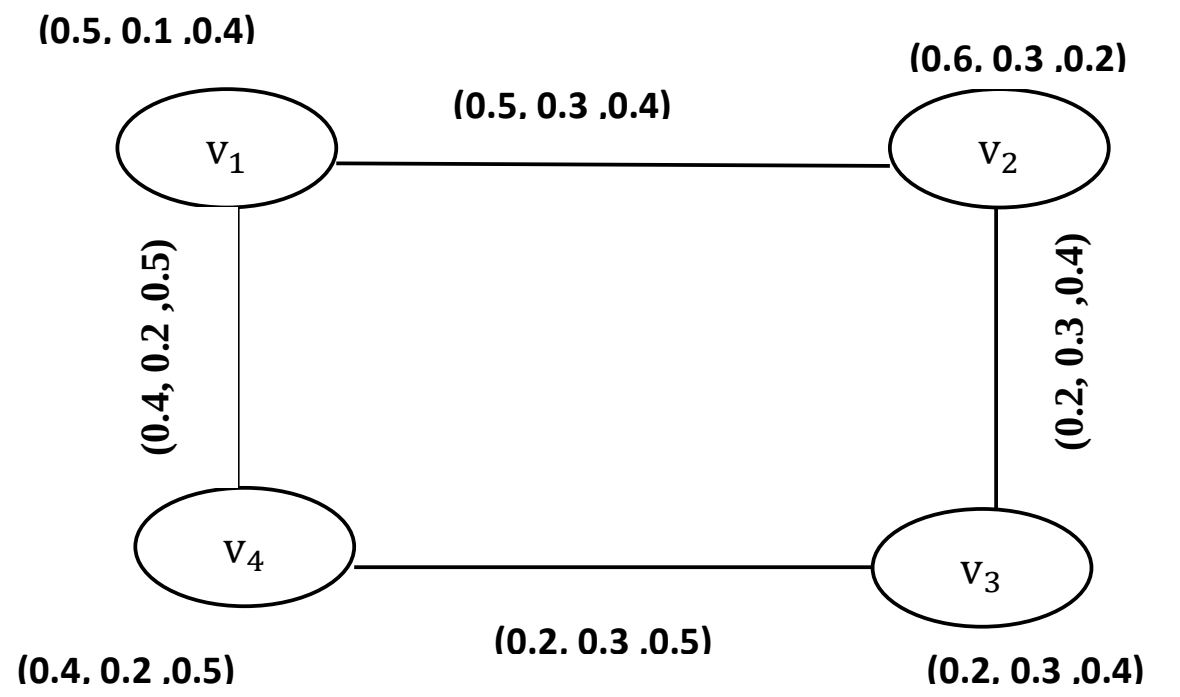

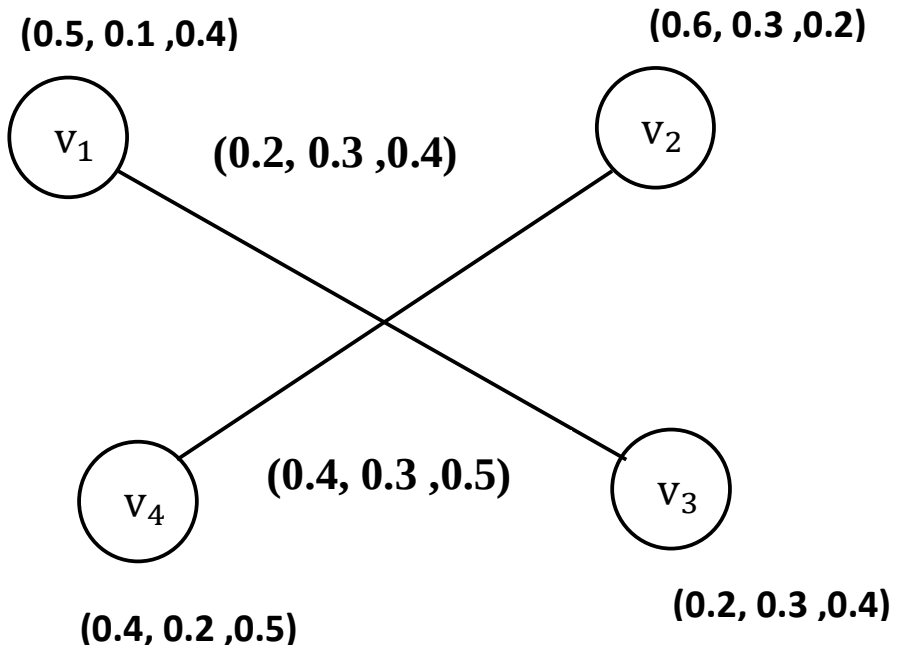

**Figure 12:** G: Strong SVN- graph

**Figure 13:** $\bar{G}$ Strong SVN- graph

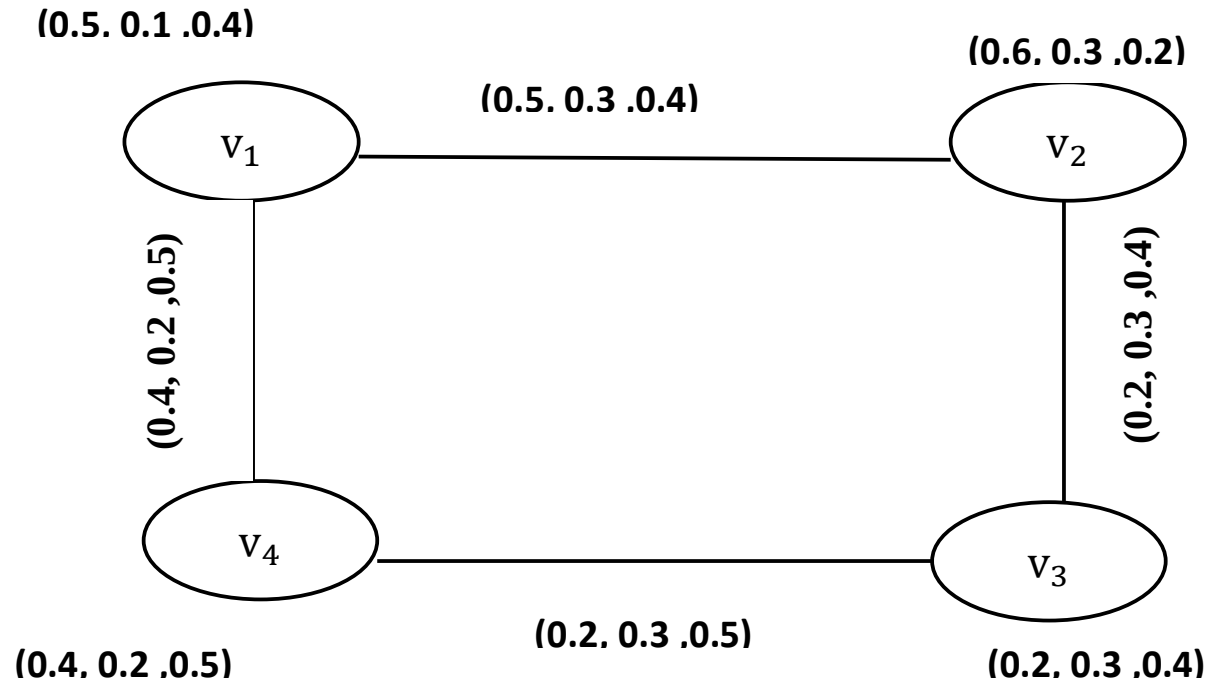

**Figure 14**: $\bar{\bar{G}}$ Strong SVN- graph

Clearly, G$\cong$ $\bar{\bar{G}}$ . Hence G is self complementary.

**Proposition 3.26.** Let G = (A, B) be a ***strong*** single valued neutrosophic graph. If

$T_B(v_i, v_j)$ = min $[T_A(v_i), T_A(v_j)]$,

$I_B(v_i, v_j)$ = max $[I_A(v_i), I_A(v_j)]$ and

$F_B(v_i, v_j)$ = max $[F_A(v_i), F_A(v_j)]$ for all $v_i, v_j$ ∈ V. Then G is self complementary.

**Proof.** Let G= (A, B) be a strong single valued neutrosophic graph such that

$T_B(v_i, v_j)$ = min $[T_A(v_i), T_A(v_j)]$

$I_B(v_i, v_j)$ = max $[I_A(v_i), I_A(v_j)]$

$F_B(v_i, v_j)$ = max $[F_A(v_i), F_A(v_j)]$

For all $v_i, v_j$ ∈ V. Then G$\approx$ $\bar{\bar{G}}$ under the identity map *I*: V →V. Hence G is self complementary.

**Proposition 3.27.** Let G be a self complementary single valued neutrosophic graph. Then

$\sum_{v_i \neq v_j} T_B(v_i, v_j) = \frac{1}{2} \sum_{v_i \neq v_j} \min [T_A(v_i), T_A(v_j)]$

$\sum_{v_i \neq v_j} I_B(v_i, v_j) = \frac{1}{2} \sum_{v_i \neq v_j} \max [I_A(v_i), I_A(v_j)]$

$\sum_{v_i \neq v_j} F_B(v_i, v_j) = \frac{1}{2} \sum_{v_i \neq v_j} \max [F_A(v_i), F_A(v_j)]$

**Proof**

If G be a self complementary single valued neutrosophic graph. Then there exist an isomorphism $f$: $V_1 \rightarrow V_1$ satisfying

$\overline{T_{V_1}}(f(v_i)) = T_{V_1}(f(v_i)) = T_{V_1}(v_i)$

$\overline{I_{V_1}}(f(v_i)) = I_{V_1}(f(v_i)) = I_{V_1}(v_i)$

$\overline{\overline{F_{V_1}}}(f(v_i)) = F_{V_1}(f(v_i)) = F_{V_1}(v_i)$ for all $v_i \in V_1$. And

$\overline{T_{E_1}}(f(v_i), f(v_j)) = T_{E_1}(f(v_i), f(v_j)) = T_{E_1}(v_i, v_j)$

$\overline{I_{E_1}}(f(v_i), f(v_j)) = I_{E_1}(f(v_i), f(v_j)) = I_{E_1}(v_i, v_j)$

$\overline{F_{E_1}}(f(v_i), f(v_j)) = F_{E_1}(f(v_i), f(v_j)) = F_{E_1}(v_i, v_j)$ for all $(v_i, v_j) \in E_1$

We have

$\overline{T_{E_1}}(f(v_i), f(v_j)) = \min [\overline{T_{V_1}}(f(v_i)), \overline{T_{V_1}}(f(v_j))] - T_{E_1}(f(v_i), f(v_j))$

i.e, $T_{E_1}(v_i, v_j) = \min [T_{V_1}(v_i), T_{V_1}(v_j)] - T_{E_1}(f(v_i), f(v_j))$

$T_{E_1}(v_i, v_j) = \min [T_{V_1}(v_i), T_{V_1}(v_j)] - T_{E_1}(v_i, v_j)$

That is

$\sum_{v_i \neq v_j} T_{E_1}(v_i, v_j) + \sum_{v_i \neq v_j} T_{E_1}(v_i, v_j) = \sum_{v_i \neq v_j} \min [T_{V_1}(v_i), T_{V_1}(v_j)]$

$\sum_{v_i \neq v_j} I_{E_1}(v_i, v_j) + \sum_{v_i \neq v_j} I_{E_1}(v_i, v_j) = \sum_{v_i \neq v_j} \max [I_{V_1}(v_i), I_{V_1}(v_j)]$

$\sum_{v_i \neq v_j} F_{E_1}(v_i, v_j) + \sum_{v_i \neq v_j} F_{E_1}(v_i, v_j) = \sum_{v_i \neq v_j} \max [F_{V_1}(v_i), F_{V_1}(v_j)]$

$$2 \sum_{v_i \neq v_j} T_{E_1}(v_i, v_j) = \sum_{v_i \neq v_j} \min [T_{V_1}(v_i), T_{V_1}(v_j)]$$
$$2 \sum_{v_i \neq v_j} I_{E_1}(v_i, v_j) = \sum_{v_i \neq v_j} \max [I_{V_1}(v_i), I_{V_1}(v_j)]$$
$$2 \sum_{v_i \neq v_j} F_{E_1}(v_i, v_j) = \sum_{v_i \neq v_j} \max [F_{V_1}(v_i), F_{V_1}(v_j)]$$

From these equations, Proposition 3.27 holds

**Proposition 3.28.** Let $G_1$ and $G_2$ be strong single valued neutrosophic graph, $\overline{G_1} \approx \overline{G_2}$ (isomorphism)

**Proof**. Assume that $G_1$ and $G_2$ are isomorphic, there exist a bijective map $f$: $V_1 \rightarrow V_2$ satisfying

$T_{V_1}(v_i) = T_{V_2}(f(v_i))$,

$I_{V_1}(v_i) = I_{V_2}(f(v_i))$,

$F_{V_1}(v_i) = F_{V_2}(f(v_i))$ for all $v_i \in V_1$. And

$T_{E_1}(v_i, v_j) = T_{E_2}(f(v_i), f(v_j))$,

$I_{E_1}(v_i, v_j) = I_{E_2}(f(v_i), f(v_j))$,

$F_{E_1}(v_i, v_j) = F_{E_2}(f(v_i), f(v_j))$ for all $(v_i, v_j) \in E_1$

By definition 3.21, we have

$\overline{T_{E_1}}(v_i, v_j) = \min [T_{V_1}(v_i), T_{V_1}(v_j)] - T_{E_1}(v_i, v_j)$

$= \min [T_{V_2}(f(v_i)),\ T_{V_2}(f(v_j))] - T_{E_2}(f(v_i), f(v_j)),$

$= \overline{T_{E_2}}(f(v_i), f(v_j)),$

$\overline{I_{E_1}}(v_i, v_j) = \max [I_{V_1}(v_i),\ I_{V_1}(v_j)] - I_{E_1}(v_i, v_j)$

$= \max [I_{V_2}(f(v_i)), I_{V_2}(f(v_j))] - I_{E_2}(f(v_i), f(v_j)),$

$= \overline{I_{E_2}}(f(v_i), f(v_j)),$

$\overline{F_{E_1}}(v_i, v_j) = \min [F_{V_1}(v_i), F_{V_1}(v_j)] - F_{E_1}(v_i, v_j)$

$= \min [F_{V_2}(f(v_i)), F_{V_2}(f(v_j))] - F_{E_2}(f(v_i), f(v_j)),$

$= \overline{F_{E_2}}(f(v_i), f(v_j)),$

For all $(v_i, v_j) \in E_1$. Hence $\overline{G_1} \approx \overline{G_2}$. The converse is straightforward.

## 4. COMPLETE SINGLE VALUED NEUTROSOPHIC GRAPHS

For the sake of simplicity we denote $T_A(v_i)$ by $T_{Ai}$ , $I_A(v_i)$ by $I_{Ai}$, and $I_A(v_i)$ by $I_{Ai}$. Also $T_B(v_i, v_j)$ by $T_{Bij}$ , $I_B(v_i, v_j)$ by $I_{Bij}$ and $F_B(v_i, v_j)$ by $F_{Bij}$.

**Definition 4.1.** A single valued neutrosophic graph G= (A, B) is called complete if

$T_{Bij}$= min ($T_{Ai}$, $T_{Aj}$), $I_{Bij}$= max ($I_{Ai}$, $I_{Aj}$) and $F_{Bij}$= max ($F_{Ai}$, $F_{Aj}$) for all $v_i, v_j \in$ V.

**Example 4.2.** Consider a graph $G^*$ = (V, E) such that V = {$v_1$, $v_2$, $v_3$, $v_4$}, E={$v_1v_2$, $v_1v_3$ , $v_2v_3$, $v_1v_4$, $v_3v_4$ , $v_2v_4$}. Then G= (A, B) is a complete single valued neutrosophic graph of $G^*$.

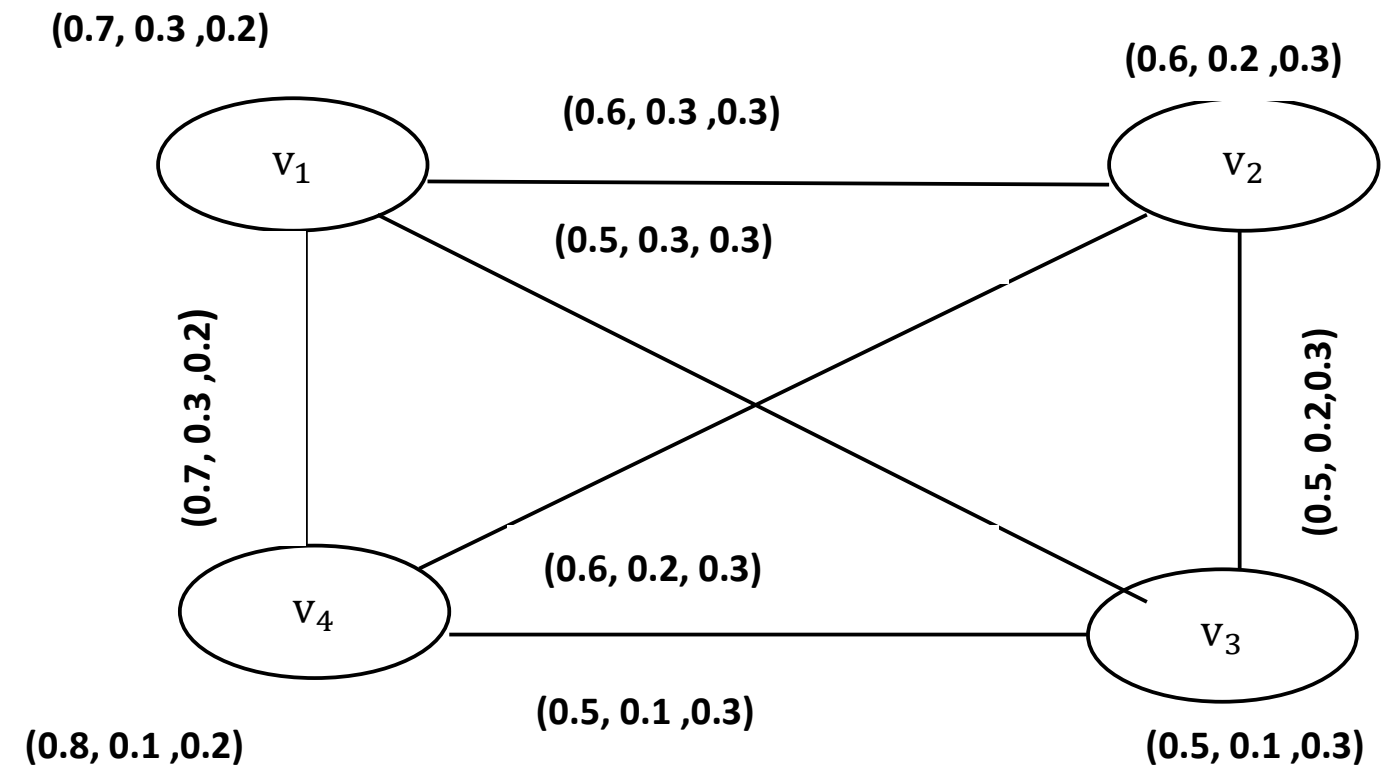

**Figure 13:** Complete single valued neutrosophic graph

**Definition 4.3.** The complement of a complete single valued neutrosophic graph G = (A, B) of $G^*$= (V, E) is a single valued neutrosophic complete graph $\bar{G}$= ($\bar{A}$, $\bar{B}$) on $G^*$= ($V$, $\bar{E}$) where

1. $\bar{V}$ =V

2. $\overline{T_A}(v_i) = T_A(v_i)$, $\overline{I_A}(v_i) = I_A(v_i)$, $\overline{F_A}(v_i) = F_A(v_i)$, for all $v_j \in$ V.

3. $\overline{T_B}(v_i, v_j) = \min [T_A(v_i), T_A(v_j)] - T_B(v_i, v_j)$

$\overline{I_B}(v_i, v_j) = \max [I_A(v_i), I_A(v_j)] - I_B(v_i, v_j)$ and

$\overline{F_B}(v_i, v_j) = \max [F_A(v_i), F_A(v_j)] - F_B(v_i, v_j)$ for all $(v_i, v_j) \in$ E

**Proposition 4.4:**

The complement of complete SVN-graph is a SVN-graph with no edge. Or if G is a complete then in $\bar{G}$ the edge is empty.

**Proof**

Let G= (A, B) be a complete SVN-graph.

So $T_{Bij}$= min ($T_{Ai}$, $T_{Aj}$), $I_{Bij}$= max ($I_{Ai}$, $I_{Aj}$) and $F_{Bij}$= max ($F_{Ai}$, $F_{Aj}$) for all $v_i, v_j \in$ V

Hence in $\bar{G}$,

$\bar{T}_{Bij}$= min $[T_{Ai}, T_{Aj}] - T_{Aij}$ for all i, j, ….., n

= min $[T_{Ai}, T_{Aj}] -$ min $[T_{Ai}, T_{Aj}]$ for all i, j,…..,n

= 0 for all i, j,…..,n

and

$\bar{I}_{Bij}$= max $[I_{Ai}, I_{Aj}] - I_{Bij}$ for all i, j,…..,n

= max $[I_{Ai}, I_{Aj}] -$ max $[I_{Ai}, I_{Aj}]$ for all i, j,…..,n

= 0 for all i, j,…..,n

Also

$\bar{F}_{Bij}$= max $[F_{Ai}, F_{Aj}] - F_{Bij}$ for all i, j,…..,n

= max $[F_{Ai}, F_{Aj}] -$ max $[F_{Ai}, F_{Aj}]$ for all i, j,…..,n

= 0 for all i, j,…..,n

Thus ($\bar{T}_{Bij}$, $\bar{I}_{Bij}$, $\bar{F}_{Bij}$) = (0 , 0, 0)

Hence, the edge set of $\bar{G}$ is empty if G is a complete SVN-graph.

## 4. CONCLUSION

Neutrosophic sets is a generalization of the notion of fuzzy sets and intuitionistic fuzzy sets. Neutrosophic models gives more precisions, flexibility and compatibility to the system as compared to the classical, fuzzy and/or intuitionistic fuzzy models. In this paper, we have introduced certain types of single valued neutrosophic graphs, such as strong single valued neutrosophic graph, constant single valued neutrosophic graph and complete single valued neutrosophic graphs. In future study, we plan to extend our research to regular and irregular single valued neutrosophic graphs, bipolar single valued neutrosophic graphs, interval valued neutrosophic graphs, strong interval valued neutrosophic, regular and irregular interval valued neutrosophic.

## ACKNOWLEDGEMENT

The authors are greatful to the editor of "Journal of New Theory" for giving permission to publish the article "Single valued neutrosophic graphs. *Journal of New Theory, 10(2016), (86-101).*" in the book **"New Trends in Neutrosophic Theories and Applications".**

# References

1. A. V. Devadoss, A. Rajkumar & N. J. P. Praveena. A Study on Miracles through Holy Bible using Neutrosophic Cognitive Maps (NCMS). International Journal of Computer Applications, 69(3) (2013).
2. A. Aydoğdu, On Similarity and Entropy of Single Valued Neutrosophic Sets, Gen. Math. Notes, Vol. 29, No. 1, July 2015, pp. 67-74.
3. Q. Ansari, R. Biswas & S. Aggarwal, (2012). Neutrosophic classifier: An extension of fuzzy classifier. Elsevier- Applied Soft Computing, 13 (2013) 563-573, http://dx.doi.org/10.1016/j.asoc.2012.08.002
4. A. Q. Ansari, R. Biswas & S. Aggarwal. (Poster Presentation) Neutrosophication of Fuzzy Models, IEEE Workshop On Computational Intelligence: Theories, Applications and Future Directions (hostedby IIT Kanpur), 14th July'13.
5. A. Q. Ansari, R. Biswas & S. Aggarwal Extension to fuzzy logic representation: Moving towards neutrosophic logic - A new laboratory rat, Fuzzy Systems (FUZZ), 2013 IEEE International Conference, 1–8, DOI:10.1109/FUZZ-IEEE.2013.6622412.
6. A. Nagoor Gani. and M. Basheer Ahamed, Order and Size in Fuzzy Graphs, Bulletin of Pure and Applied Sciences, Vol 22E (No.1) 2003; p.145-148.
7. A. Nagoor Gani. and S. Shajitha Begum, Degree, Order and Size in Intuitionistic Fuzzy Graphs, International Journal of Algorithms, Computing and Mathematics, (3)3 (2010).
8. A. Nagoor Gani and S.R Latha, On Irregular Fuzzy Graphs, Applied Mathematical Sciences, Vol.6, 2012, no.11,517-523.
9. F. Smarandache. Refined Literal Indeterminacy and the Multiplication Law of Sub-Indeterminacies, Neutrosophic Sets and Systems, Vol. 9, 58-2015,
10. F. Smarandache, Types of Neutrosophic Graphs and neutrosophic Algebraic Structures together with their Applications in Technology, seminar, Universitatea Transilvania din Brasov, Facultatea de Design de Produs si Mediu, Brasov, Romania 06 June 2015.
11. F. Smarandache, Symbolic Neutrosophic Theory, Europanova asbl, Brussels, 195p, 2015.
12. F. Smarandache , Neutrosophic set - a generalization of the intuitionistic fuzzy set, Granular Computing, 2006 IEEE International Conference, 38 – 42,2006,DOI: 10.1109/GRC.2006.1635754.
13. F. Smarandache, A geometric interpretation of the neutrosophic set — A generalization of the intuitionistic fuzzy set Granular Computing (GrC), 2011 IEEE International Conference , 602 – 606, 2011, DOI 10.1109/GRC.2011.6122665.
14. Gaurav Garg, Kanika Bhutani, Megha Kumar and Swati Aggarwal, Hybrid model for medical diagnosis using Neutrosophic Cognitive Maps with Genetic Algorithms, FUZZ-IEEE 2015 (IEEE International conference onfuzzy systems).
15. H .Wang,. Y. Zhang, R. Sunderraman, Truth-value based interval neutrosophic sets, Granular Computing, 2005 IEEE International Conference , P274 - 277 Vol. 1,2005DOI: 10.1109/GRC.2005.1547284.
16. H. Wang, F. Smarandache, Y. Zhang, and R. Sunderraman, Single valued Neutrosophic Sets, Multisspace and Multistructure 4 (2010) 410-413.
17. A. Deli, M. Ali, F. Smarandache , Bipolar neutrosophic sets and their application based on multi-criteria decision making problems, Advanced Mechatronic Systems (ICAMechS), 2015 International Conference, 249 - 254, DOI: 10.1109/ICAMechS.2015.7287068.
18. A. Turksen, Interval valued fuzzy sets based on normal forms, Fuzzy Sets and Systems, vol. 20, p. 191-210 (1986).
19. J. Ye, vector similarity measures of simplified neutrosophic sets and their application in multicriteria decision making, International Journal of Fuzzy Systems, Vol. 16, No. 2, p.204-211 (2014).
20. J. Ye, Single-Valued Neutrosophic Minimum Spanning Tree and Its Clustering Method, Journal of Intelligent Systems 23(3): 311–324, (2014).
21. K. Atanassov, "Intuitionistic fuzzy sets," Fuzzy Sets and Systems, vol. 20, p. 87-96 (1986).

22. K. Atanassov and G. Gargov, "Interval valued intuitionistic fuzzy sets," Fuzzy Sets and Systems, vol.31, pp. 343-349 (1989).
23. K. Atanassov. Intuitionistic fuzzy sets: theory and applications. Physica, New York, 1999.
24. L. Zadeh, Fuzzy sets, Inform and Control, 8(1965), 338-353
25. P. Bhattacharya, Some remarks on fuzzy graphs, Pattern Recognition Letters 6: 297-302, 1987.
26. R. Parvathi and M. G. Karunambigai, Intuitionistic Fuzzy Graphs, Computational Intelligence, Theory and applications, International Conference in Germany, Sept 18 -20, 2006.
27. R. Rıdvan, A. Küçük, Subsethood measure for single valued neutrosophic sets, Journal of Intelligent & Fuzzy Systems, vol. 29, no. 2, pp. 525-530, 2015,DOI: 10.3233/IFS-141304
28. S. Aggarwal, R. Biswas, A. Q. Ansari, Neutrosophic modeling and control , Computer and Communication Technology (ICCCT), 2010 International Conference , 718 – 723, DOI:10.1109/ICCCT.2010.5640435
29. S. Broumi, F. Smarandache, New distance and similarity measures of interval neutrosophic sets, Information Fusion (FUSION), 2014 IEEE 17th International Conference, 2014,p 1 – 7.
30. S. Broumi, F. Smarandache, Single valued neutrosophic trapezoid linguistic aggregation operators based multi-attribute decision making, Bulletin of Pure & Applied Sciences- Mathematics and Statistics,(2014)135-155, DOI : 10.5958/2320-3226.2014.00006.X
31. S. Broumi, M. Talea, F. Smarandache, Single Valued Neutrosophic Graphs: Degree, Order and Size, IEEE WCCI 2016, under process.
32. Y. Hai-Long, G. She, Yanhonge, L. Xiuwu, On single valued neutrosophic relations, Journal of Intelligent & Fuzzy Systems, vol. Preprint, no. Preprint, p. 1-12 (2015)
33. W. B. Vasantha Kandasamy and F. Smarandache, Fuzzy Cognitive Maps and Neutrosophic Congtive Maps,2013.
34. W. B. Vasantha Kandasamy, K. Ilanthenral and Florentin Smarandache, Neutrosophic Graphs: A New Dimension to Graph Theory, Kindle Edition, 2015.
35. W.B. Vasantha Kandasamy and F. Smarandache "Analysis of social aspects of migrant laborers living with HIV/AIDS using Fuzzy Theory and Neutrosophic Cognitive Maps", Xiquan, Phoenix (2004).

**Said Broumi** [1*], **Mohamed Talea** [2], **Assia Bakali** [3], **Florentin Smarandache** [4]

1, 2 Laboratory of Information processing, Faculty of Science Ben M'Sik, University Hassan II, B.P 7955, Sidi Othman, Casablanca, Morocco. E-mails: broumisaid78@gmail.com, taleamohamed@yahoo.fr
3 Ecole Royale Navale-Boulevard Sour Jdid, B.P 16303 Casablanca-Morocco. Email: assiabakali@yahoo.fr
4 Department of Mathematics, University of New Mexico, 705 Gurley Avenue, Gallup, NM 87301, USA.
E-mail : fsmarandache@gmail.com

# On Bipolar Single Valued Neutrosophic Graphs

## Abstract

In this article, we combine the concept of bipolar neutrosophic set and graph theory. We introduce the notions of bipolar single valued neutrosophic graphs, strong bipolar single valued neutrosophic graphs, complete bipolar single valued neutrosophic graphs, regular bipolar single valued neutrosophic graphs and investigate some of their related properties.

## 1. Introduction

Zadeh [32] coined the term 'degree of membership' and defined the concept of fuzzy set in order to deal with uncertainty. Atanassov [29, 31] incorporated the degree of non-membership in the concept of fuzzy set as an independent component and defined the concept of intuitionistic fuzzy set. Smarandache [12, 13] grounded the term 'degree of indeterminacy as an independent component and defined the concept of neutrosophic set from the philosophical point of view to deal with incomplete, indeterminate and inconsistent information in real world. The concept of neutrosophic sets is a generalization of the theory of fuzzy sets, intuitionistic fuzzy sets. Each element of a neutrosophic sets has three membership degrees including a truth membership degree, an indeterminacy membership degree, and a falsity membership degree which are within the real standard or nonstandard unit interval]−0, 1+[. Therefore, if their range is restrained within the real standard unit interval [0, 1], the neutrosophic set is easily applied to engineering problems. For this purpose, Wang et al. [17] introduced the concept of a single valued neutrosophic set (SVNS) as a subclass of the neutrosophic set. Recently, Deli et al. [23] defined the concept of bipolar neutrosophic as an extension of the fuzzy sets, bipolar fuzzy sets, intuitionistic fuzzy sets and neutrosophic sets studied some of their related properties including the score, certainty and accuracy functions to compare the bipolar neutrosophic sets. The neutrosophic sets theory of and their extensions have been applied in various part [1, 2, 3, 16, 18, 19, 20, 21, 25, 26, 27, 41, 42, 50, 51, 53].

A graph is a convenient way of representing information involving relationship between objects. The objects are represented by vertices and the relations by edges. When there is vagueness in the description of the objects or in its relationships or in both, it is natural that we need to designe a fuzzy graph Model. The extension of fuzzy graph theory [4, 6, 11] have been developed by several researchers including intuitionistic fuzzy graphs [5, 35, 44] considered the vertex sets and edge sets as intuitionistic fuzzy sets. Interval valued fuzzy graphs [32, 34] considered the vertex sets and edge sets as interval valued fuzzy sets. Interval valued intuitionistic fuzzy graphs [8, 52] considered the vertex sets and edge sets as interval valued intuitionstic fuzzy sets. Bipolar fuzzy graphs [6, 7, 40] considered the vertex sets and edge sets as bipolar fuzzy sets. M-polar fuzzy graphs [39] considered the vertex sets and edge sets as m-polar fuzzy sets. Bipolar intuitionistic fuzzy graphs [9] considered the vertex sets and edge sets as bipolar intuitionistic fuzzy sets. But, when the relations between nodes (or vertices) in problems are indeterminate, the fuzzy graphs and their extensions are failed. For this purpose, Samarandache [10, 11] have defined four main categories of neutrosophic graphs, two based on literal indeterminacy (I), which called them; I-edge neutrosophic graph and I-vertex neutrosophic graph, these concepts are studied deeply and has gained popularity among the researchers due to its applications via real world problems [7, 14, 15, 54, 55, 56]. The two others graphs are based on (t, i, f) components and called them; The (t, i, f)-Edge neutrosophic graph and the (t, i, f)-vertex neutrosophic graph, these concepts are not developed at all. Later on, Broumi et al. [46] introduced a third neutrosophic graph model. This model allows the attachment of truth-membership (t), indeterminacy–membership (i) and falsity-membership degrees (f) both to vertices and edges, and investigated some of their properties. The third neutrosophic graph model is called single valued neutrosophic graph (SVNG for short). The single valued neutrosophic graph is the generalization of fuzzy graph and intuitionistic fuzzy graph. Also the same authors [45] introduced neighborhood degree of a vertex and closed neighborhood degree of vertex in single valued neutrosophic graph as a generalization of neighborhood degree of a vertex and closed neighborhood degree of vertex in fuzzy graph and intuitionistic fuzzy graph. Also, Broumi et al. [47] introduced the concept of interval valued neutrosophic graph as a generalization fuzzy graph, intuitionistic fuzzy graph, interval valued fuzzy graph, interval valued intuitionistic fuzzy graph and single valued neutrosophic graph and have discussed some of their properties with proof and examples. In addition Broumi et al [48] have introduced some operations such as cartesian product, composition, union and join on interval valued neutrosophic graphs and investigate some their properties. On the other hand, Broumi et al [49] have discussed a sub class of interval valued neutrosophic graph called strong interval valued neutrosophic graph, and have introduced some operations such as, cartesian product, composition and join of two strong interval valued neutrosophic graph with proofs. In the literature the study of bipolar single valued neutrosophic graphs (BSVN-graph) is still blank, we shall focus on the study of bipolar single valued neutrosophic graphs in this paper. In the present paper, bipolar neutrosophic sets are employed to study graphs and give rise to a new class of graphs called bipolar single valued neutrosophic graphs. We introduce the notions of bipolar single valued neutrosophic graphs, strong bipolar single valued neutrosophic graphs, complete bipolar single valued neutrosophic graphs, regular bipolar single valued neutrosophic graphs and investigate some of their related properties. This paper is organized as follows;

In section 2, we give all the basic definitions related bipolar fuzzy set, neutrosophic sets, bipolar neutrosophic set, fuzzy graph, intuitionistic fuzzy graph, bipolar fuzzy graph, N-graph and single valued neutrosophic graph which will be employed in later sections. In section 3, we introduce certain notions including bipolar single valued neutrosophic graphs, strong bipolar single valued neutrosophic graphs, complete bipolar single valued neutrosophic graphs, the complement of strong bipolar single valued neutrosophic graphs, regular bipolar single valued neutrosophic graphs and illustrate these notions by several examples, also we described degree of a vertex, order, size of bipolar single valued neutrosophic graphs. In section 4, we give the conclusion.

## 2. Preliminaries

In this section, we mainly recall some notions related to bipolar fuzzy set, neutrosophic sets, bipolar neutrosophic set, fuzzy graph, intuitionistic fuzzy graph, bipolar fuzzy graph, ***N***-graph and single valued neutrosophic graph relevant to the present work. The readers are referred to [9, 12, 17, 35, 36, 38, 43, 46, 57] for further details and background.

**Definition 2.1** [12]. Let U be an universe of discourse; then the neutrosophic set A is an object having the form A = {< x: $T_A(x)$, $I_A(x)$, $F_A(x)$>, x ∈ U}, where the functions T, I, F: U→]$^-0,1^+$[ define respectively the degree of membership, the degree of indeterminacy, and the degree of non-membership of the element x ∈ U to the set A with the condition:

$$^-0 \leq T_A(x)+ I_A(x)+ F_A(x) \leq 3^+. \quad (1)$$

The functions $T_A(x)$, $I_A(x)$ and $F_A(x)$ are real standard or nonstandard subsets of ]−0,1+[.

Since it is difficult to apply NSs to practical problems, Wang et al. [16] introduced the concept of a SVNS, which is an instance of a NS and can be used in real scientific and engineering applications.

**Definition 2.2 [17].** Let X be a space of points (objects) with generic elements in X denoted by x. A single valued neutrosophic set A (SVNS A) is characterized by truth-membership function $T_A(x)$, an indeterminacy-membership function $I_A(x)$, and a falsity-membership function $F_A(x)$. For each point x in X $T_A(x)$, $I_A(x)$, $F_A(x)$ ∈ [0, 1]. A SVNS A can be written as

$$A = \{< x: T_A(x), I_A(x), F_A(x)>, x \in X\} \quad (2)$$

**Definition 2.3 [9]**. A bipolar neutrosophic set A in X is defined as an object of the form

A= {<x, $T^P$(x), $I^P$(x), $F^P$(x), $T^N$(x), $I^N$(x), $F^N$(x)>: x ∈ X}, where

$T^P$, $I^P$, $F^P$:X→ [1, 0] and $T^N$, $I^N$, $F^N$: X→ [-1, 0] .The Positive membership degree $T^P$(x), $I^P$(x), $F^P$(x) denotes the truth membership, indeterminate membership and false membership of an element ∈ X corresponding to a bipolar neutrosophic set A and the negative membership degree $T^N$(x), $I^N$(x), $F^N$(x) denotes the truth membership, indeterminate membership and false membership of an element ∈ X to some implicit counter-property corresponding to a bipolar neutrosophic set A.

**Example 2.4** Let X = $\{x_1, x_2, x_3\}$

$$A=\begin{Bmatrix} <x_1,0.5,0.3,0.1,-0.6,-0.4,-0.05> \\ <x_2,0.3,0.2,0.7,-0.02,-0.3,-0.02> \\ <x_3,0.8,0.05,0.4,-0.6,-0.6,-0.03> \end{Bmatrix}$$

is a bipolar neutrosophic subset of X

**Definition 2.5[9]**. Let $A_1$={<x, $T_1^P$(x), $I_1^P$(x), $F_1^P$(x), $T_1^N$(x), $I_1^N$(x), $F_1^N$(x)>} and $A_2$={<x, $T_2^P$(x), $I_2^P$(x), $F_2^P$(x), $T_2^N$(x), $I_2^N$(x), $F_2^N$(x) >} be two bipolar neutrosophic sets . Then $A_1 \subseteq A_2$ if and only if

$T_1^P(x) \le T_2^P(x)$ , $I_1^P(x) \le I_2^P(x)$, $F_1^P(x) \ge F_2^P(x)$ and $T_1^N(x) \ge T_2^N(x)$ , $I_1^N(x) \ge I_2^N(x)$ , $F_1^N(x) \le F_2^N(x)$ for all x ∈ X.

**Definition 2.6[9]**. Let $A_1$={<x, $T_1^P$(x), $I_1^P$(x), $F_1^P$(x), $T_1^N$(x), $I_1^N$(x), $F_1^N$(x)>} and $A_2$={<x, $T_2^P$(x), $I_2^P$(x), $F_2^P$(x), $T_2^N$(x), $I_2^N$(x), $F_2^N$(x) >} be two bipolar neutrosophic sets . Then $A_1 = A_2$ if and only if

$T_1^P(x) = T_2^P(x)$ , $I_1^P(x) = I_2^P(x)$, $F_1^P(x) = F_2^P(x)$ and $T_1^N(x) = T_2^N(x)$ , $I_1^N(x) = I_2^N(x)$ , $F_1^N(x) = F_2^N(x)$ for all x ∈ X

**Definition 2.7 [9].** Let $A_1$={<x, $T_1^P$(x), $I_1^P$(x), $F_1^P$(x), $T_1^N$(x), $I_1^N$(x), $F_1^N$(x)>} and $A_2$={<x, $T_2^P$(x), $I_2^P$(x), $F_2^P$(x), $T_2^N$(x), $I_2^N$(x), $F_2^N$(x) >} be two bipolar neutrosophic sets . Then their union is defined as:

$$(A_1 \cup A_2)(x) = \begin{pmatrix} \max(T_1^P(x),T_2^p(x)), \frac{I_1^P(x)+ I_2^P(x)}{2}, \min(T_1^P(x),T_2^P(x)) \\ \min(T_1^N(x),T_2^N(x)), \frac{I_1^N(x)+ I_2^N(x)}{2}, \max(T_1^N(x),T_2^N(x)) \end{pmatrix} \text{ for all x} \in \text{X}.$$

**Definition 2.8 [9]**. Let $A_1$={<x, $T_1^P$(x), $I_1^P$(x), $F_1^P$(x), $T_1^N$(x), $I_1^N$(x), $F_1^N$(x)>} and $A_2$={<x, $T_2^P$(x), $I_2^P$(x), $F_2^P$(x), $T_2^N$(x), $I_2^N$(x), $F_2^N$(x) >} be two bipolar neutrosophic sets . Then their intersection is defined as:

$$(A_1 \cap A_2)(x) = \begin{pmatrix} \min(T_1^P(x),T_2^P(x)), \frac{I_1^P(x)+ I_2^P(x)}{2}, \max(T_1^P(x),T_2^P(x)) \\ \max(T_1^N(x),T_2^N(x)), \frac{I_1^N(x)+ I_2^N(x)}{2}, \min(T_1^N(x),T_2^N(x)) \end{pmatrix} \text{ for all x} \in \text{X}.$$

**Definition 2.9 [9].** Let $A_1$= {<x, $T_1^P$(x), $I_1^P$(x), $F_1^P$(x), $T_1^N$(x), $I_1^N$(x), $F_1^N$(x)>: x ∈ X} be a bipolar neutrosophic set in X. Then the complement of A is denoted by $A^c$ and is defined by

$T_{A^c}^P$(x)={$1^P$}-$T_A^P$(x), $I_{A^c}^P$(x)={$1^P$}-$I_A^P$(x), $F_{A^c}^P$(x)={$1^P$}-$F_A^+$(x)

And

$T_{A^c}^N$(x)={$1^N$}-$T_A^N$(x), $I_{A^c}^N$(x)={$1^N$}-$I_A^N$(x), $F_{A^c}^N$(x)={$1^N$}-$F_A^N$(x)

**Definition 2.10 [43].** A fuzzy graph is a pair of functions G = (σ, μ) where σ is a fuzzy subset of a non empty set V and μ is a symmetric fuzzy relation on σ. i.e σ: V → [ 0,1] and μ: VxV→[0,1] such that μ(uv) ≤ σ(u) ∧ σ(v) for all u, v ∈ V where uv denotes the edge between u and v and σ(u) ∧ σ(v) denotes the minimum of σ(u) and σ(v). σ is called the fuzzy vertex set of V and μ is called the fuzzy edge set of E.

**Definition 2.11[38]:** By a *N*-graph G of a graph $G^*$, we mean a pair G= ($\mu_1$, $\mu_2$) where $\mu_1$ is an *N*-function in V and $\mu_2$ is an *N*-relation on E such that $\mu_2$(u, v) ≥ max ($\mu_1$(u), $\mu_1$(v)) all u, v ∈ V.

**Definition 2.12[35]:** An Intuitionistic fuzzy graph is of the form G = (V, E) where

iii. V= {$v_1$, $v_2$,…., $v_n$} such that $\mu_1$: V→ [0,1] and $\gamma_1$: V → [0,1] denote the degree of membership and non-membership of the element $v_i$ ∈ V, respectively, and

0 ≤ $\mu_1(v_i) + \gamma_1(v_i)$) ≤ 1 for every $v_i$ ∈ V, (i = 1, 2, ……. n),

iv. E ⊆ V x V where $\mu_2$: VxV→[0,1] and $\gamma_2$: VxV→ [0,1] are such that

$\mu_2(v_i, v_j) \le \min [\mu_1(v_i), \mu_1(v_j)]$ and $\gamma_2(v_i, v_j) \ge \max [\gamma_1(v_i), \gamma_1(v_j)]$

and $0 \le \mu_2(v_i, v_j) + \gamma_2(v_i, v_j) \le 1$ for every ($v_i$, $v_j$) ∈ E, (i, j = 1,2, ……. n)

**Definition 2.13 [57]**. Let X be a non-empty set. A bipolar fuzzy set A in X is an object having the form A = {(x, $\mu_A^P$(x), $\mu_A^N$(x)) | x ∈ X}, where $\mu_A^P$(x): X → [0, 1] and $\mu_A^N$(x): X → [−1, 0] are mappings.

**Definition 2.14[ 57]** Let X be a non-empty set. Then we call a mapping A = ($\mu_A^P$, $\mu_A^N$): X × X → [−1, 0] × [0, 1] a bipolar fuzzy relation on X such that $\mu_A^P$(x, y) ∈ [0, 1] and $\mu_A^N$(x, y) ∈ [−1, 0].

**Definition 2.15[36].** Let $A = (\mu_A^P,\ \mu_A^N)$ and $B = (\mu_B^P,\ \mu_B^N)$ be bipolar fuzzy sets on a set *X*. If $A = (\mu_A^P,\ \mu_A^N)$ is a bipolar fuzzy relation on a set *X*, then $A = (\mu_A^P,\ \mu_A^N)$ is called a bipolar fuzzy relation on $B = (\mu_B^P,\ \mu_B^N)$ if $\mu_B^P(x, y) \leq \min(\mu_A^P(x), \mu_A^P(y))$ and $\mu_B^N(x, y) \geq \max(\mu_A^N(x), \mu_A^N(y))$ for all *x, y* ∈ *X*.

A bipolar fuzzy relation *A* on *X* is called symmetric if $\mu_A^P(x, y) = \mu_A^P(y, x)$ and $\mu_A^N(x, y) = \mu_A^N(y, x)$ for all *x, y* ∈ *X*.

**Definition 2.16[36].** A bipolar fuzzy graph of a graph $G^*$= (V, E) is a pair G = (A,B), where A = ($\mu_A^P$, $\mu_A^N$) is a bipolar fuzzy set in V and B = ($\mu_B^P$, $\mu_B^N$) is a bipolar fuzzy set on E ⊆ V x V such that $\mu_B^P$(xy) ≤ min{$\mu_A^P$(x), $\mu_A^P$(y)} for all xy ∈ $E$, $\mu_B^N$(xy) ≥ min{$\mu_A^N$(x), $\mu_A^N$(y)} for all xy ∈ $E$ and $\mu_B^P$(xy) = $\mu_B^N$(xy) = 0 for all xy ∈ $\tilde{V}^2$ –E. Here A is called bipolar fuzzy vertex set of V, B the bipolar fuzzy edge set of E.

**Definition 2.17[46]** A single valued neutrosophic graph (SVNG) of a graph $G^*$= (V, E) is a pair G = (A, B), where

1.V= $\{v_1, v_2,\ldots, v_n\}$ such that $T_A$:V→[0, 1], $I_A$:V→[0, 1] and $F_A$:V→[0, 1] denote the degree of truth-membership, degree of indeterminacy-membership and falsity-membership of the element $v_i$ ∈ V, respectively, and

0≤ $T_A(v_i) + I_A(v_i) + F_A(v_i)$ ≤3 for every $v_i$ ∈ V (i=1, 2, …,n)

2. E ⊆ V x V where $T_B$:V x V →[0, 1], $I_B$:V x V →[0, 1] and $F_B$:V x V →[0, 1] are such that $T_B(v_i, v_j) \leq \min[T_A(v_i), T_A(v_j)]$, $I_B(v_i, v_j) \geq \max[I_A(v_i), I_A(v_j)]$ and $F_B(v_i, v_j) \geq \max[F_A(v_i), F_A(v_j)]$ and

0≤ $T_B(v_i, v_j) + I_B(v_i, v_j) + F_B(v_i, v_j)$ ≤3 for every $(v_i, v_j)$ ∈ E (i, j = 1, 2,…, n)

**Definition 2.18[46]:** Let G = (V, E) be a single valued neutrosophic graph. Then the degree of a vertex v is defined by d(v)= $(d_T(v), d_I(v), d_F(v))$ where

$d_T(v)=\sum_{u\neq v} T_B(u, v)$, $d_I(v)=\sum_{u\neq v} I_B(u, v)$ and $d_F(v)=\sum_{u\neq v} F_B(u, v)$

## 3. Bipolar Single Valued Neutrosophic Graph

**Definition 3.1.** Let X be a non-empty set. Then we call a mapping A = (x, $T^P$(x), $I^P$(x), $F^P$(x), $T^N$(x), $I^N$(x), $F^N$(x)):X × X → [−1, 0] × [0, 1] a bipolar single valued neutrosophic relation on X such that $T_A^P$(x, y) ∈ [0, 1], $I_A^P$(x, y) ∈ [0, 1], $F_A^P$(x, y) ∈ [0, 1], and $T_A^N$(x, y) ∈ [−1, 0], $I_A^N$(x, y) ∈ [−1, 0], $F_A^N$(x, y) ∈ [−1, 0].

**Definition 3.2.** Let $A = (T_A^P, I_A^P, F_A^P, T_A^N, I_A^N, F_A^N)$ and $B = (T_B^P, I_B^P, F_B^P, T_B^N, I_B^N, F_B^N)$ be bipolar single valued neutrosophic graph on a set *X*. If $B = (T_B^P, I_B^P, F_B^P, T_B^N, I_B^N, F_B^N)$ is a bipolar single valued neutrosophic relation on $A = (T_A^P, I_A^P, F_A^P, T_A^N, I_A^N, F_A^N)$ then

$T_B^P(x, y) \leq \min(T_A^P(x), T_A^P(y))$, $T_B^N(x, y) \geq \max(T_A^N(x), T_A^N(y))$
$I_B^P(x, y) \geq \max(I_A^P(x), I_A^P(y))$, $I_B^N(x, y) \leq \min(I_A^N(x), I_A^N(y))$
$F_B^P(x, y) \geq \max(F_A^P(x), F_A^P(y))$, $F_B^N(x, y) \leq \min(F_A^N(x), F_A^N(y))$ for all *x, y* ∈ *X*.

A bipolar single valued neutrosophic relation *B* on *X* is called symmetric if $T_B^P(x, y) = T_B^P(y, x)$, $I_B^P(x, y) = I_B^P(y, x)$, $F_B^P(x, y) = F_B^P(y, x)$ and $T_B^N(x, y) = T_B^N(y, x)$, $I_B^N(x, y) = I_B^N(y, x)$, $F_B^N(x, y) = F_B^N(y, x)$, for all *x, y* ∈ *X*.

**Definition 3.3.** A bipolar single valued neutrosophic graph of a graph $G^{*}=$ (V, E) is a pair G = (A, B), where A = $(T_A^P, I_A^P, F_A^P, T_A^N, I_A^N, F_A^N)$ is a bipolar single valued neutrosophic set in V and B = $(T_B^P, I_B^P, F_B^P, T_B^N, I_B^N, F_B^N)$ is a bipolar single valued neutrosophic set in $\tilde{V}^2$ such that

$T_B^P(\boldsymbol{v_i}, \boldsymbol{v_j}) \leq \min(T_A^P(\boldsymbol{v_i}), T_A^P(\boldsymbol{v_j}))$

$I_B^P(\boldsymbol{v_i}, \boldsymbol{v_j}) \geq \max(I_A^P(\boldsymbol{v_i}), I_A^P(\boldsymbol{v_j}))$

$F_B^P(\boldsymbol{v_i}, \boldsymbol{v_j}) \geq \max(F_A^P(\boldsymbol{v_i}), F_A^P(\boldsymbol{v_j}))$

And

$T_B^N(\boldsymbol{v_i}, \boldsymbol{v_j}) \geq \max(T_A^N(\boldsymbol{v_i}), T_A^N(\boldsymbol{v_j}))$

$I_B^N(\boldsymbol{v_i}, \boldsymbol{v_j}) \leq \min(I_A^N(\boldsymbol{v_i}), I_A^N(\boldsymbol{v_j}))$

$F_B^N(\boldsymbol{v_i}, \boldsymbol{v_j}) \leq \min(F_A^N(\boldsymbol{v_i}), F_A^N(\boldsymbol{v_j}))$ for all $\boldsymbol{v_i v_j} \in \tilde{V}^2$.

**Notation**: An edge of BSVNG is denoted by $e_{ij} \in$ E or $\boldsymbol{v_i v_j} \in$ E

Here the sextuple ($v_i$, $T_A^P(v_i)$, $I_A^P(v_i)$, $F_A^P(v_i)$, $T_A^N(v_i)$, $I_A^N(v_i)$, $F_A^N(v_i)$) denotes the positive degree of truth-membership, the positive degree of indeterminacy-membership, the positive degree of falsity-membership, the negative degree of truth-membership, the negative degree of indeterminacy-membership, the negative degree of falsity- membership of the vertex vi.

The sextuple ($e_{ij}$, $T_B^P, I_B^P, F_B^P, T_B^N, I_B^N, F_B^N$) denotes the positive degree of truth-membership, the positive degree of indeterminacy-membership, the positive degree of falsity-membership, the negative degree of truth-membership, the negative degree of indeterminacy-membership, the negative degree of falsity- membership of the edge relation $e_{ij}$ = $(\boldsymbol{v_i}, \boldsymbol{v}_j)$ on V× V.

**Note 1**. (i) When $T_A^P = I_A^P = F_A^P = 0$ and $T_A^N = I_A^N = F_A^N = 0$ for some i and j, then there is no edge between $v_i$ and $v_j$ .

Otherwise there exists an edge between $v_i$ and $v_j$ .

(ii) If one of the inequalities is not satisfied in (1) and (2), then G is not an BSVNG

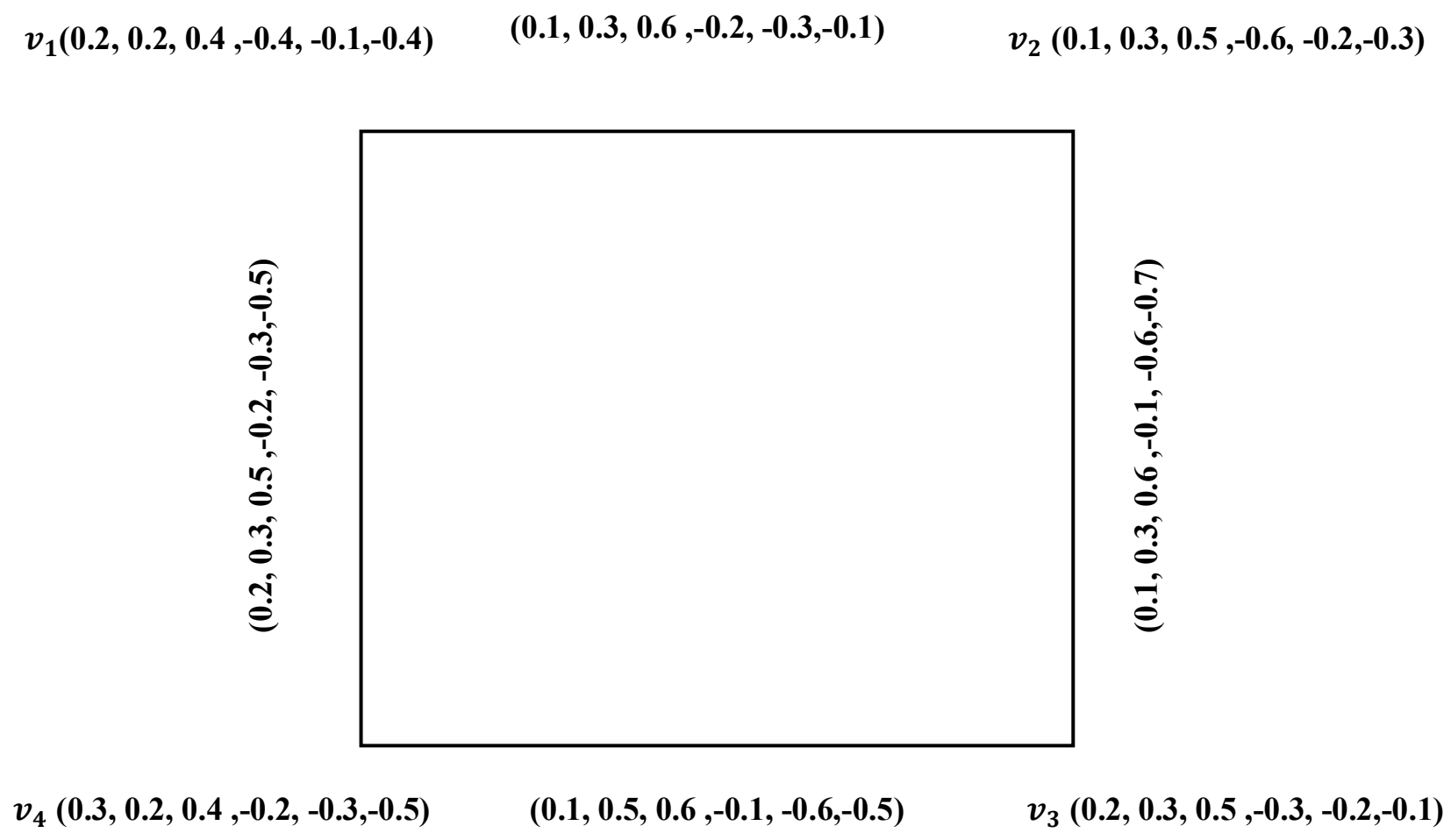

Fig. 1: *Bipolar single valued neutrosophic graph.*

**Proposition 3.5:** A bipolar single valued neutrosophic graph is the generalization of fuzzy graph

**Proof:** Suppose G= (A, B) be a bipolar single valued neutrosophic graph. Then by setting the positive indeterminacy-membership, positive falsity-membership and negative truth-membership, negative indeterminacy-membership, negative falsity-membership values of vertex set and edge set equals to zero reduces the bipolar single valued neutrosophic graph to fuzzy graph.

**Example 3.6:**

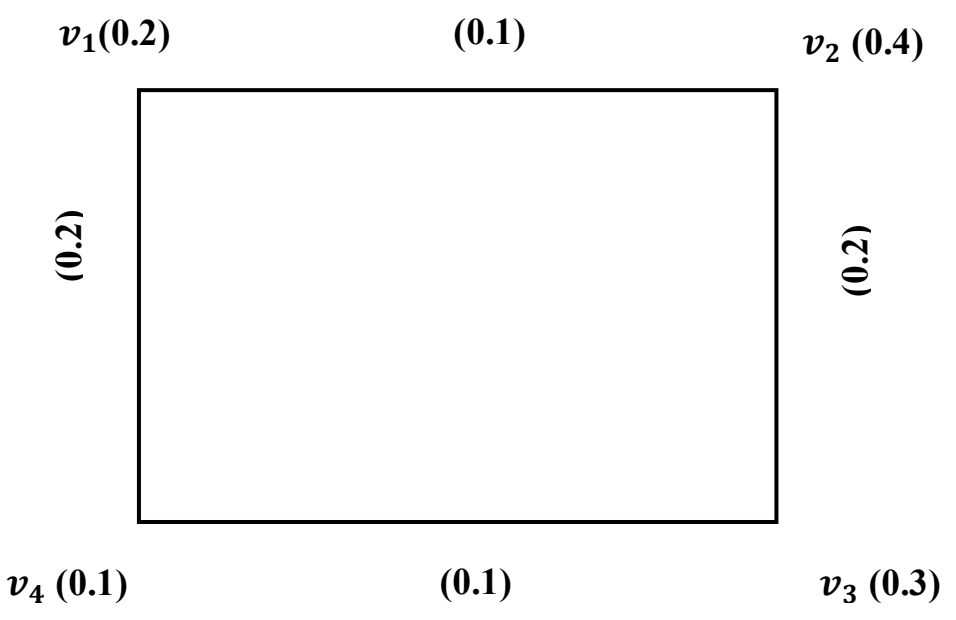

Fig.2: *Fuzzy graph*

**Proposition 3.7:** A bipolar single valued neutrosophic graph is the generalization of intuitionistic fuzzy graph

**Proof:** Suppose G= (A, B) be a bipolar single valued neutrosophic graph. Then by setting the positive indeterminacy-membership, negative truth-membership, negative indeterminacy-membership, negative falsity-membership values of vertex set and edge set equals to zero reduces the bipolar single valued neutrosophic graph to intuitionistic fuzzy graph.

**Example 3.8**

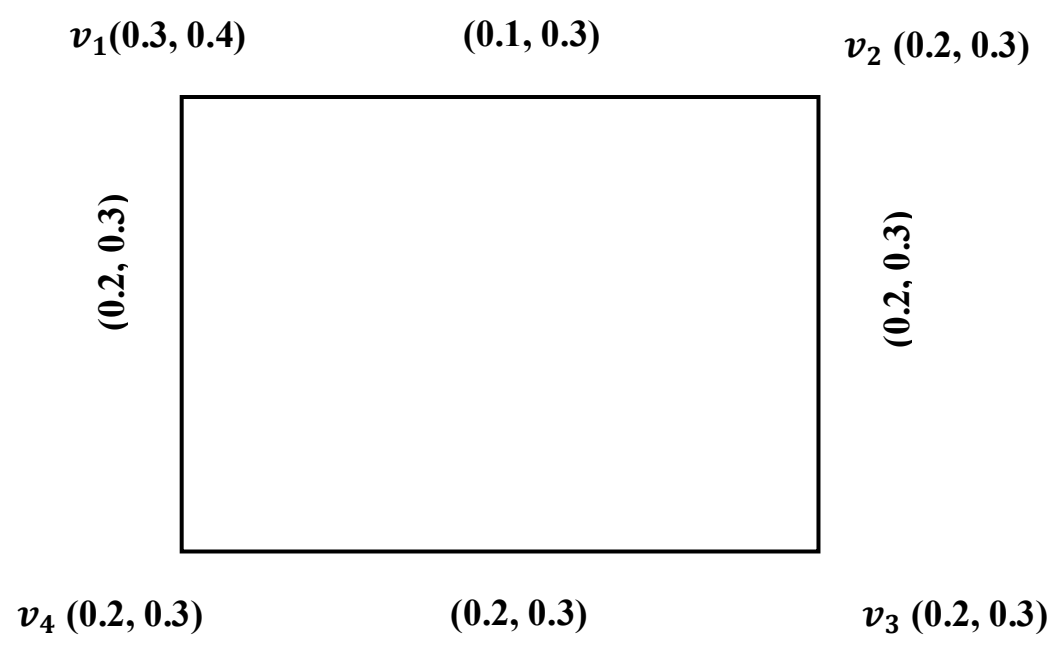

Fig.3: *Intuitionistic fuzzy graph*

**Proposition 3.9:** A bipolar single valued neutrosophic graph is the generalization of single valued neutrosophic graph

**Proof:** Suppose G= (A, B) be a bipolar single valued neutrosophic graph. Then by setting the negative truth-membership, negative indeterminacy-membership, negative falsity-membership values of vertex set and edge set equals to zero reduces the bipolar single valued neutrosophic graph to single valued neutrosophic graph.

**Example 3.10**

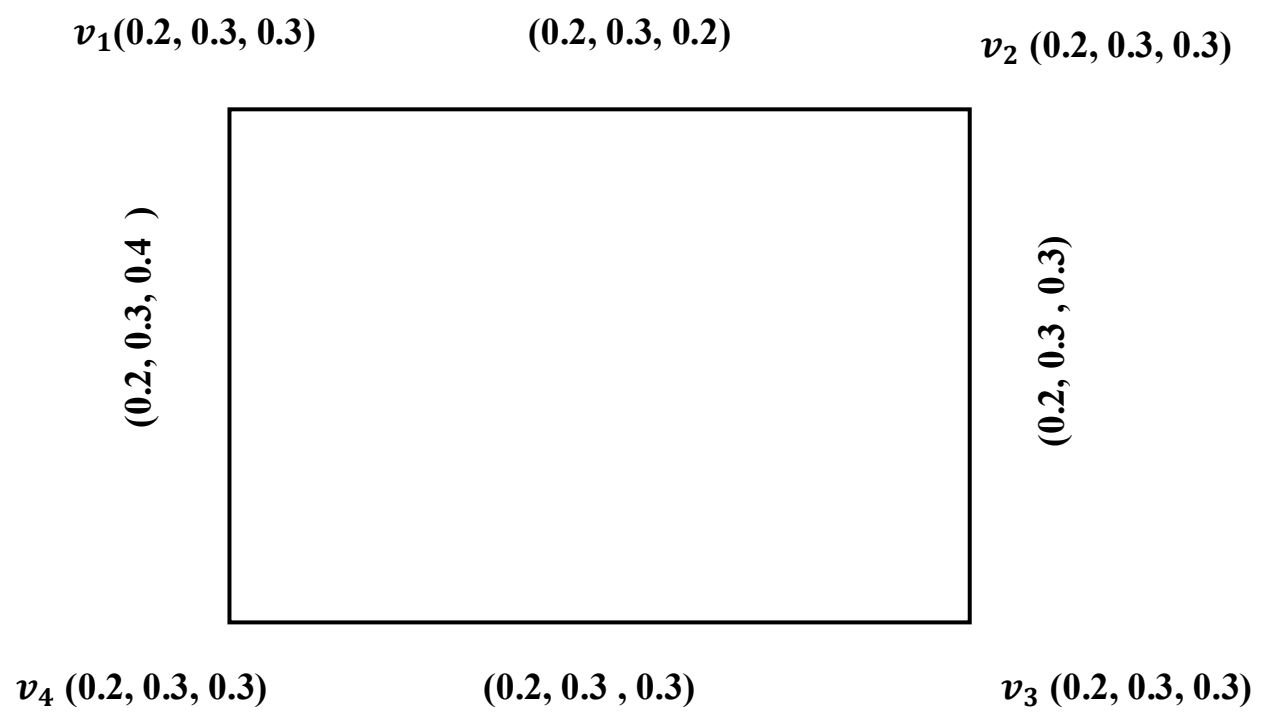

Fig.4: *Single valued neutrosophic graph*

**Proposition 3.11:** A bipolar single valued neutrosophic graph is the generalization of bipolar intuitionstic fuzz graph

**Proof:** Suppose G= (A, B) be a bipolar single valued neutrosophic graph. Then by setting the positive indeterminacy-membership, negative indterminacy-membership values of vertex set and edge set equals to zero reduces the bipolar single valued neutrosophic graph to bipolar intuitionstic fuzzy graph

**Example 3.12**

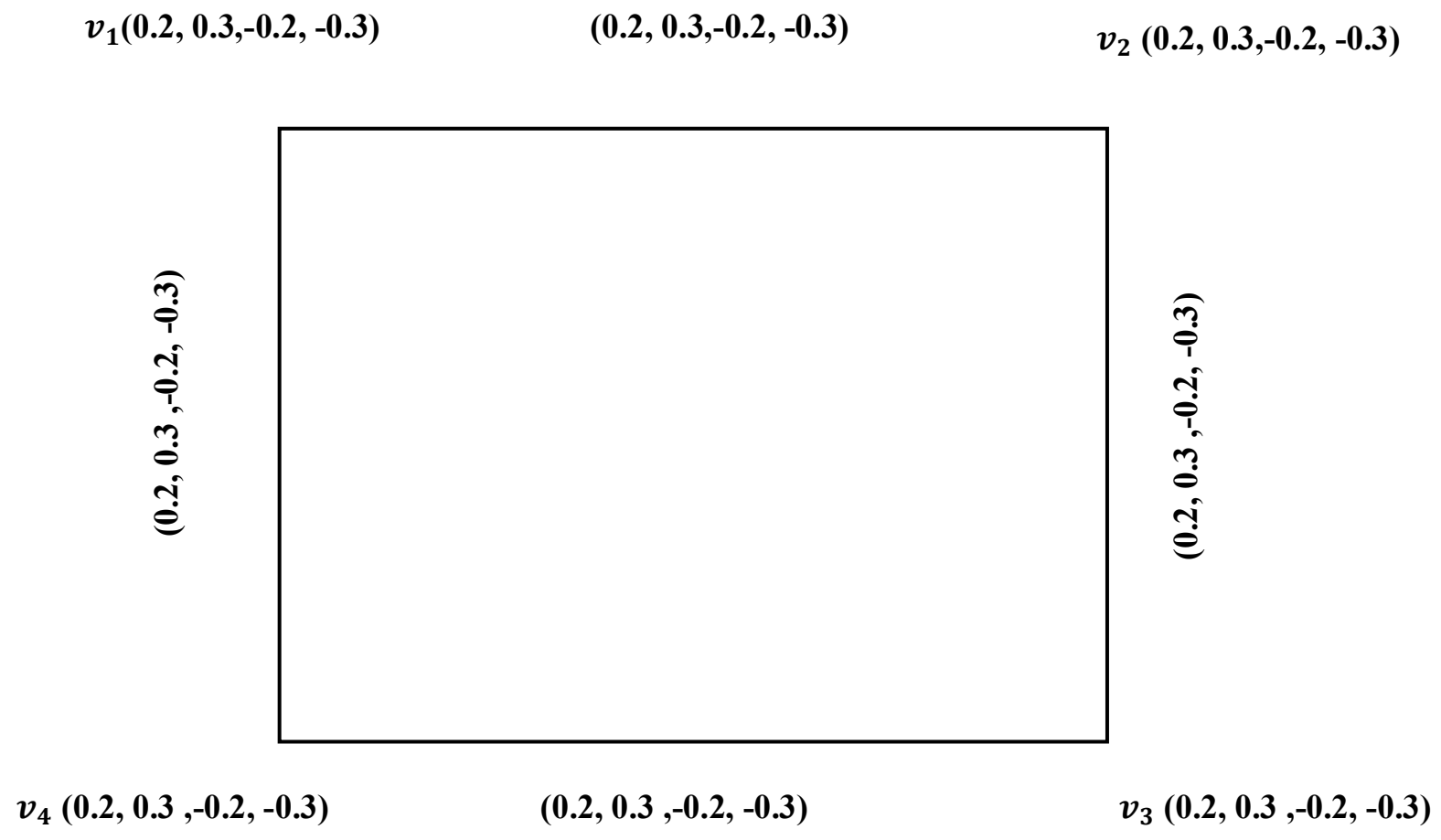

Fig. 5: *Bipolar intuitionistic fuzzy graph.*

**Proposition 3.13:** A bipolar single valued neutrosophic graph is the generalization of *N*-graph

**Proof:** Suppose G= (A, B) be a bipolar single valued neutrosophic graph. Then by setting the positive degree membership such truth-membership, indeterminacy- membership, falsity-membership and negative indeterminacy-membership, negative falsity-membership values of vertex set and edge set equals to zero reduces the single valued neutrosophic graph to *N*-graph.

**Example 3.14**:

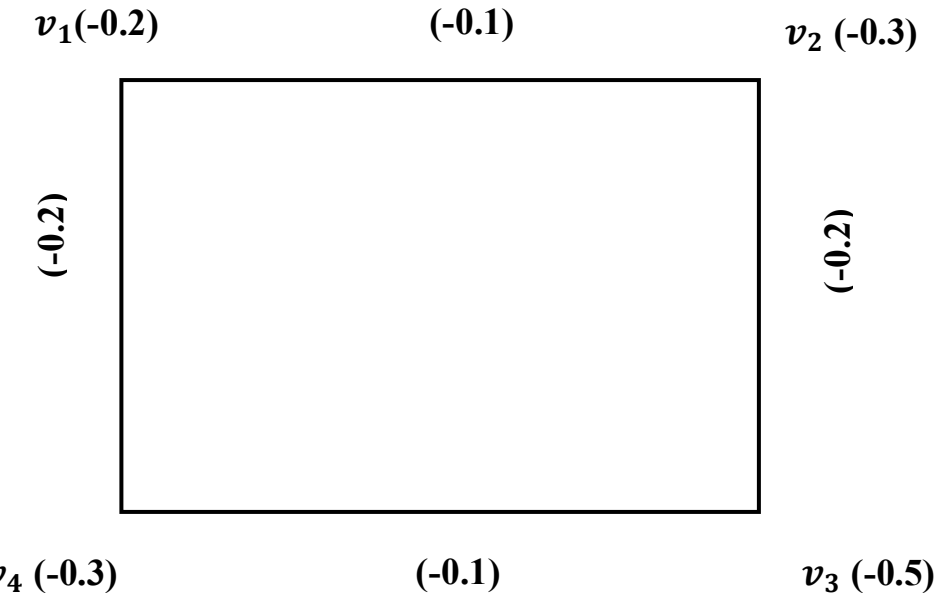

Fig. 6: *N*- graph

**Definition 3.15.** A bipolar single valued neutrosophic graph that has neither self loops nor parallel edge is called simple bipolar single valued neutrosophic graph.

**Definition 3.16.** A bipolar single valued neutrosophic graph is said to be connected if every pair of vertices has at least one bipolar single valued neutrosophic graph between them, otherwise it is disconnected.

**Definition 3.17.** When a vertex $\mathbf{v_i}$ is end vertex of some edges ($\mathbf{v_i}$, $\mathbf{v_j}$) of any BSVN-graph G= (A, B). Then $\mathbf{v_i}$ and ($\mathbf{v_i}$, $\mathbf{v_j}$) are said to be **incident** to each other.

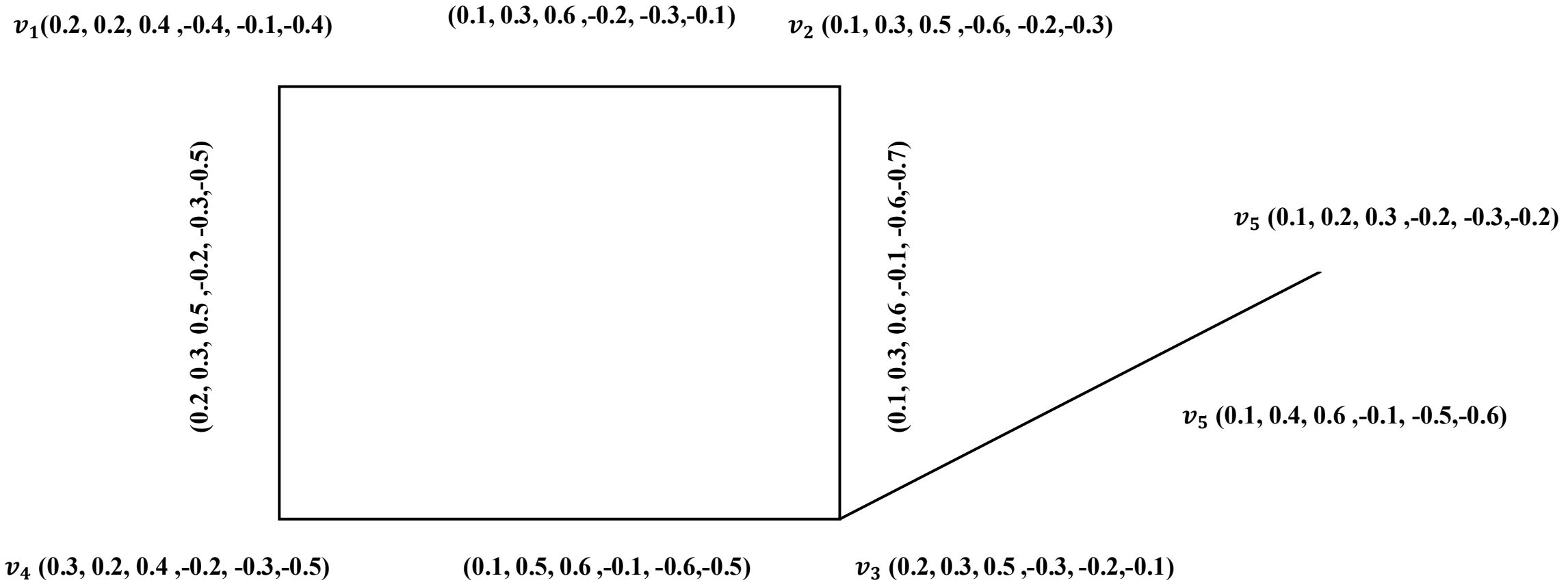

Fig. 7: *Incident BSVN-graph*

In this graph $v_2v_3$, $v_3v_4$ and $v_3v_5$ are incident on $v_3$.

**Definition 3.18** Let G= (V, E) be a bipolar single valued neutrosophic graph. Then the degree of any vertex **v** is sum of positive degree of truth-membership, positive sum of degree of indeterminacy-membership, positive sum of degree of falsity-membership, negative degree of truth-membership, negative sum of degree of indeterminacy-membership, and negative sum of degree of falsity-membership of all those edges which are incident on vertex **v** denoted by d(v)= $(d_T^P(v), d_I^P(v), d_F^P(v), d_T^N(v), d_I^N(v), d_F^N(v))$ where

$d_T^P(v) = \sum_{u \neq v} T_B^P(u, v)$ denotes the positive T- degree of a vertex v,

$d_I^P(v) = \sum_{u \neq v} I_B^P(u, v)$ denotes the positive I- degree of a vertex v,

$d_F^P(v)$ =$\sum_{u\neq v} F_B^P(u,v)$ denotes the positive F- degree of a vertex v,
$d_T^N(v)$ =$\sum_{u\neq v} T_B^N(u,v)$ denotes the negative T- degree of a vertex v,
$d_I^N(v)$=$\sum_{u\neq v} I_B^N(u,v)$ denotes the negative I- degree of a vertex v,
$d_F^N(v)$ =$\sum_{u\neq v} F_B^N(u,v)$ denotes the negative F- degree of a vertex v

**Definition 3.19:** The minimum degree of G is

$\delta(G) = (\delta_T^P(G), \delta_I^P(G), \delta_F^P(G), \delta_T^N(G), \delta_I^N(G), \delta_F^N(G))$ , where

$\delta_T^P(G)= \wedge \{d_T^P(v) \mid v \in V\}$ denotes the minimum positive T- degree,

$\delta_I^P(G)= \wedge \{d_I^P(v) \mid v \in V\}$ denotes the minimum positive I- degree,

$\delta_F^P(G)= \wedge \{d_F^P(v) \mid v \in V\}$ denotes the minimum positive F- degree,

$\delta_T^N(G)= \wedge \{d_T^N(v) \mid v \in V\}$ denotes the minimum negative T- degree,

$\delta_I^N(G)= \wedge \{d_I^N(v) \mid v \in V\}$ denotes the minimum negative I- degree,

$\delta_F^N(G)= \wedge \{d_F^N(v)\mid v \in V\}$ denotes the minimum negative F- degree

**Definition 3.20:** The maximum degree of G is

$\Delta(G) = (\Delta_T^P(G), \Delta_I^P(G), \Delta_F^P(G), \Delta_T^N(G), \Delta_I^N(G), \Delta_F^N(G))$, where

$\Delta_T^P(G)= \vee \{d_T^P(v) \mid v \in V\}$ denotes the maximum positive T- degree,

$\Delta_I^P(G)= \vee \{d_I^P(v) \mid v \in V\}$ denotes the maximum positive I- degree,

$\Delta_F^P(G)= \vee \{d_F^P(v) \mid v \in V\}$ denotes the maximum positive F- degree,

$\Delta_T^N(G)= \vee \{d_T^N(v) \mid v \in V\}$ denotes the maximum negative T- degree,

$\Delta_I^N(G)= \vee \{d_I^N(v) \mid v \in V\}$ denotes the maximum negative I- degree,

$\Delta_F^N(G)= \vee \{d_F^N(v)\mid v \in V\}$ denotes the maximum negative F- degree

**Example 3.21**. Let us consider a bipolar single valued neutrosophic graph G= (A, B) of $G^*$ = (V, E), such that V = $\{v_1, v_2, v_3, v_4\}$, E = $\{(v_1, v_2), (v_2, v_3), (v_3, v_4), (v_4, v_1)\}$

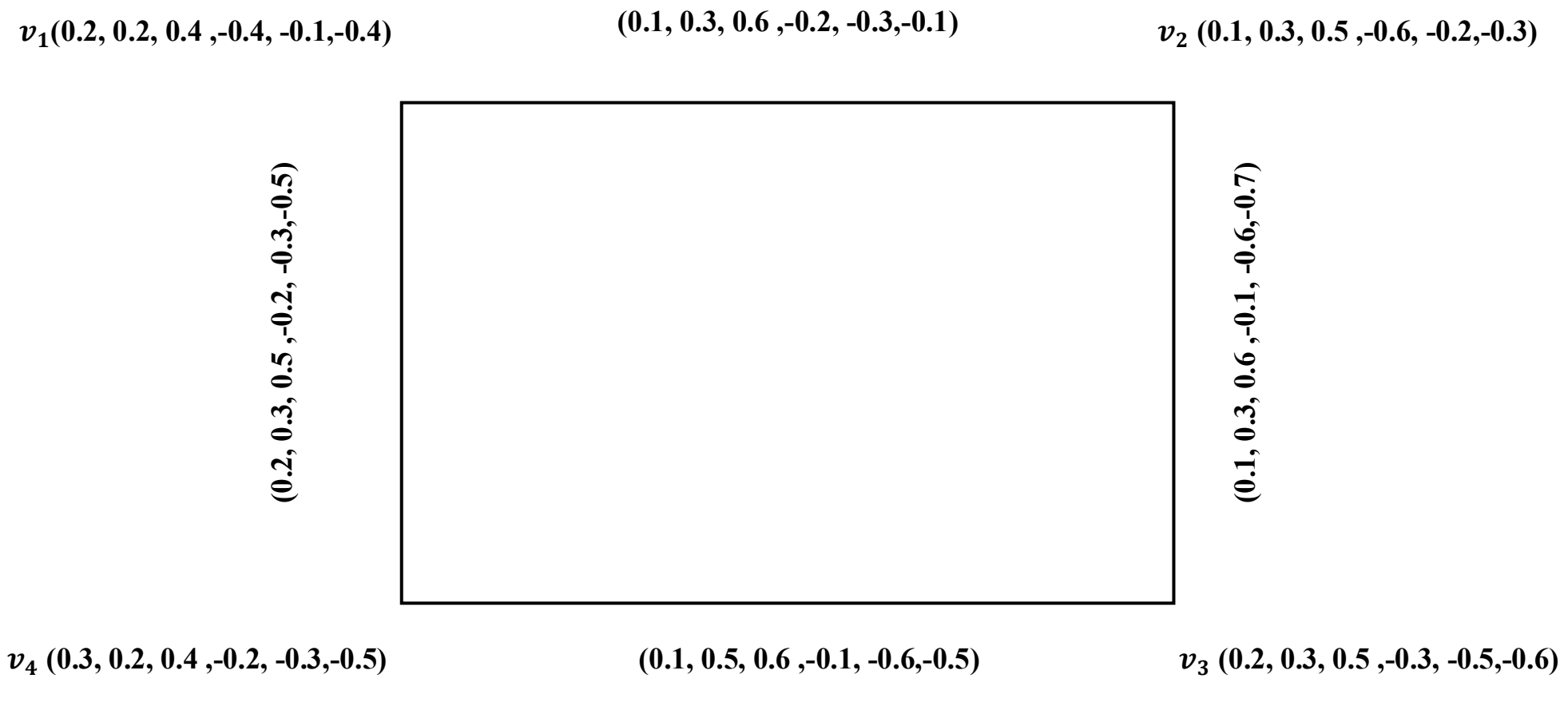

Figure 8: *Degree of a bipolar single valued neutrosophic graph G.*

In this example, the degree of $v_1$ is (0.3, 0.6, 1.1, -0.4, -0.6, -0.6). the degree of $v_2$ is (0.2, 0.6, 1.2, -0.3, -0.9, -0.8). the degree of $v_3$ is (0.2, 0.8, 1.2, -0.2, -1.2, -1.2). the degree of $v_4$ is (0.3, 0.8, 1.1, -0.3, -0.9, -1)

Order and size of a bipolar single valued neutrosophic graph is an important term in bipolar single valued neutrosophic graph theory. They are defined below.

**Definition 3.22:** Let G = (V, E) be a BSVNG. The order of G, denoted O(G) is defined as O(G)= $(O_T^p(G), O_I^p(G), O_F^p(G), O_T^N(G), O_I^N(G), O_F^N(G))$, where

$O_T^p(G)= \sum_{v\in V} T_1^p$ (v) denotes the positive T- order of a vertex v,

$O_I^p(G)= \sum_{v\in V} I_1^p$ (v) denotes the positive I- order of a vertex v,

$O_F^p(G)= \sum_{v\in V} F_1^p$ (v) denotes the positive F- order of a vertex v,

$O_T^N(G)= \sum_{v\in V} T_1^N$ (v) denotes the negative T- order of a vertex v,

$O_I^N(G)= \sum_{v\in V} I_1^N$ (v) denotes the negative I- order of a vertex v,

$O_F^N(G)= \sum_{v\in V} F_1^N$ (v) denotes the negative F- order of a vertex v.

**Definition 3.23**: Let G = (V, E) be a BSVNG. The size of G, denoted S(G) is defined as

S(G)= $(S_T^p(G), S_I^p(G), S_F^p(G), S_T^N(G), S_I^N(G), S_F^N(G))$, where

$S_T^p(G)= \sum_{u\neq v} T_2^p$ (u, v) denotes the positive T- size of a vertex v,

$S_I^p(G)= \sum_{u\neq v} I_2^p$ (u, v) denotes the positive I- size of a vertex v,

$S_F^p(G)= \sum_{u\neq v} F_2^p$ (u, v) denotes the positive F- size of a vertex v,

$S_T^N(G)= \sum_{u\neq v} T_2^N$ (u, v) denotes the negative T- size of a vertex v,

$S_I^N(G)= \sum_{u\neq v} I_2^N$ (u, v) denotes the negative I- size of a vertex v,

$S_F^N(G)= \sum_{u\neq v} F_2^N$ (u, v) denotes the negative F- size of a vertex v.

**Definition 3.24** A bipolar single valued neutrosophic graph G = (V, E) is called constant if degree of each vertex is k = $(k_1, k_2, k_3, k_4, k_5, k_6)$. That is, d $(v)$ = $(k_1, k_2, k_3, k_4, k_5, k_6)$ for all $v \in$ V.

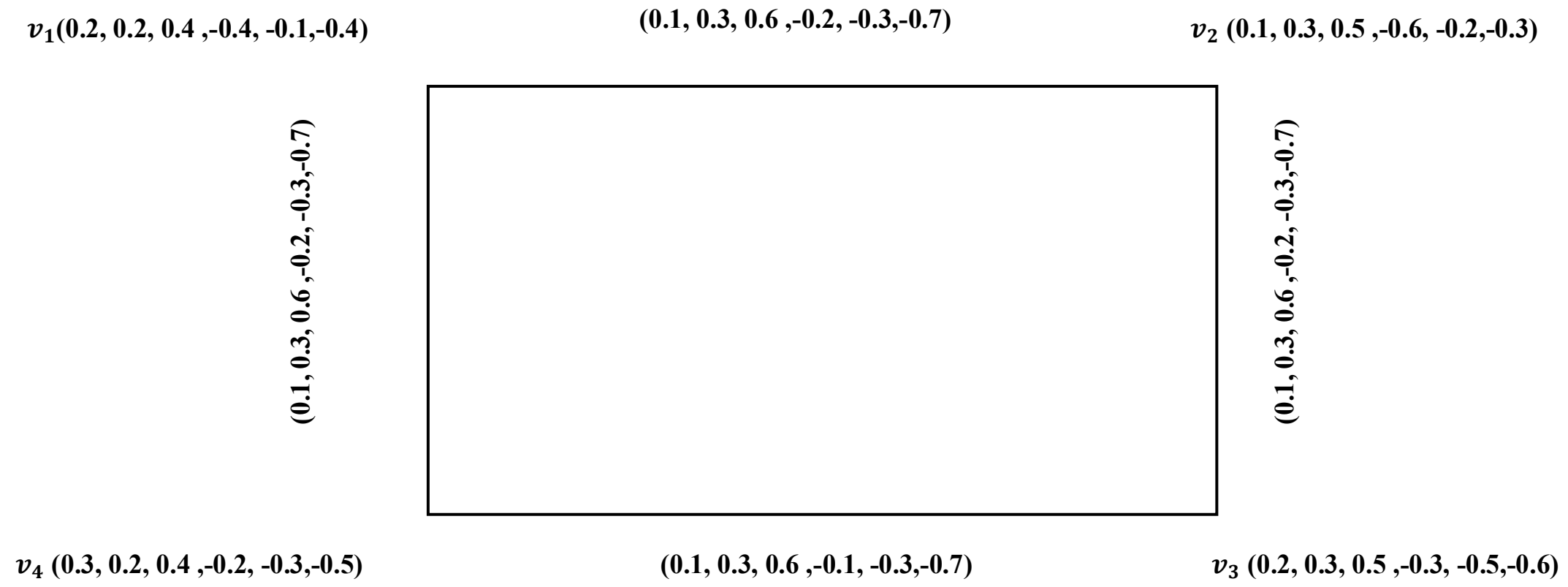

Figure 9: *Constant bipolar single valued neutrosophic graph G.*

In this example, the degree of $v_1, v_2, v_3, v_4$ is (0.2, 0.6, 1.2, -0.4, -0.6, -1.4).

O(G)= (0.8, 1, 1.8, -1.5, -1.1, -1.8)

S(G) = (0.4, 1.2, 2.4, -0.7, -1.2, -2.8)

**Remark 3.25**. G is a ($k_i$, $k_j$, $k_l$, $k_m$, $k_n$, $k_o$)-constant BSVNG iff $\delta$ = $\Delta$= k, where k = $k_i$+ $k_j$+ $k_l$ +$k_m$+ $k_n$+ $k_o$.

**Definition 3.26.** A bipolar single valued neutrosophic graph G= (A, B) is called strong bipolar single valued neutrosophic graph if

$T_B^P(u, v)$ =min ($T_A^P(u)$, $T_A^P(v)$),

$I_B^P(u, v)$ =max ($I_A^P(u)$, $I_A^P(v)$),

$F_B^P(u, v)$ =max ($F_A^P(u)$, $F_A^P(v)$),

$T_B^N(u, v)$ =max ($T_A^N(u)$, $T_A^N(v)$),

$I_B^N(u, v)$ =min ($I_A^N(u)$, $I_A^N(v)$),

$F_B^N(u, v)$ = min ($F_A^N(u)$, $F_A^N(v)$) for all (u, v) ∈ E

**Example 3.27.** Consider a strong BSVN-graph G such that V = {$v_1, v_2, v_3, v_4$}and E = {($v_1, v_2$), ($v_2, v_3$), ($v_3, v_4$), ($v_4, v_1$)}

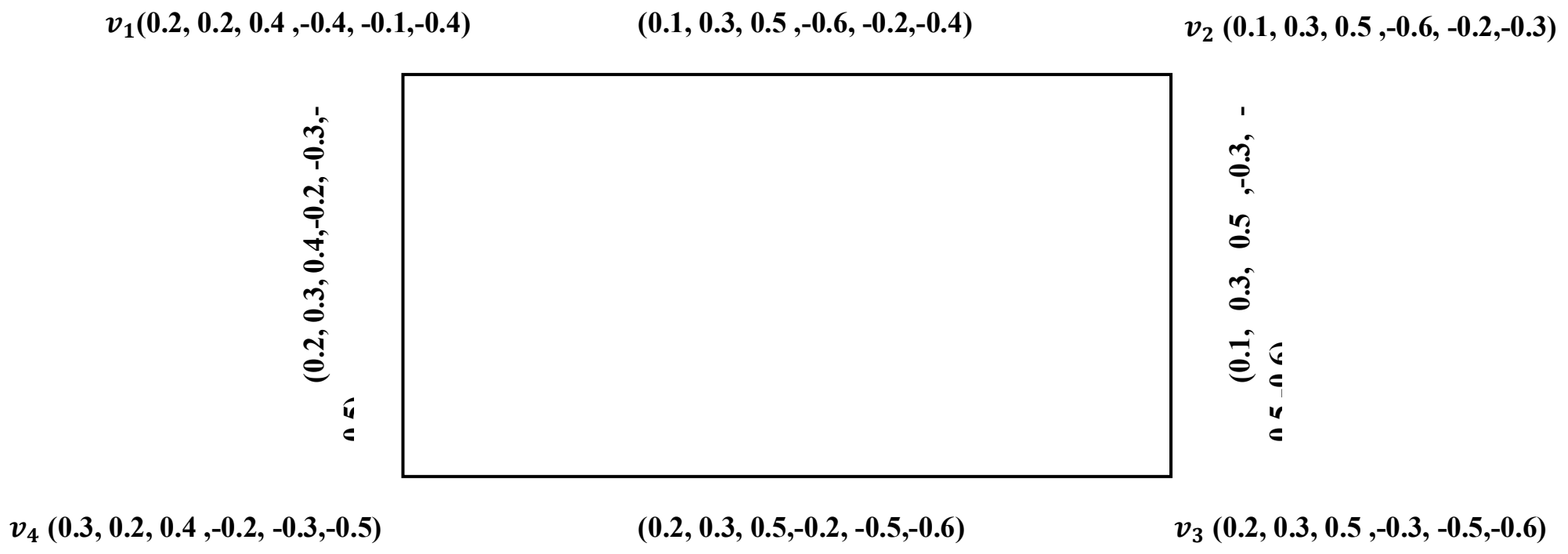

Figure 10: *Strong bipolar single valued neutrosophic graph G.*

**Definition 3.28.** A bipolar single valued neutrosophic graph G= (A, B) is called complete if

$T_B^P(u, v)$ =min ($T_A^P(u)$, $T_A^P(v)$),

$I_B^P(u, v)$ =max ($I_A^P(u)$, $I_A^P(v)$),

$F_B^P(u, v)$ =max ($F_A^P(u)$, $F_A^P(v)$),

$T_B^N(u,v)$ =max $(T_A^N(u), T_A^N(v))$,

$I_B^N(u,v)$ =min $(I_A^N(u), I_A^N(v))$,

$F_B^N(u,v)$ = min $(F_A^N(u), F_A^N(v))$ for all u, v ∈ V.

**Example 3.29.** Consider a complete BSVN-graph G such that V = $\{v_1, v_2, v_3, v_4\}$and E = $\{(v_1, v_2), (v_2, v_3), (v_3, v_4), (v_4, v_1), (v_1, v_3), (v_2, v_4)\}$

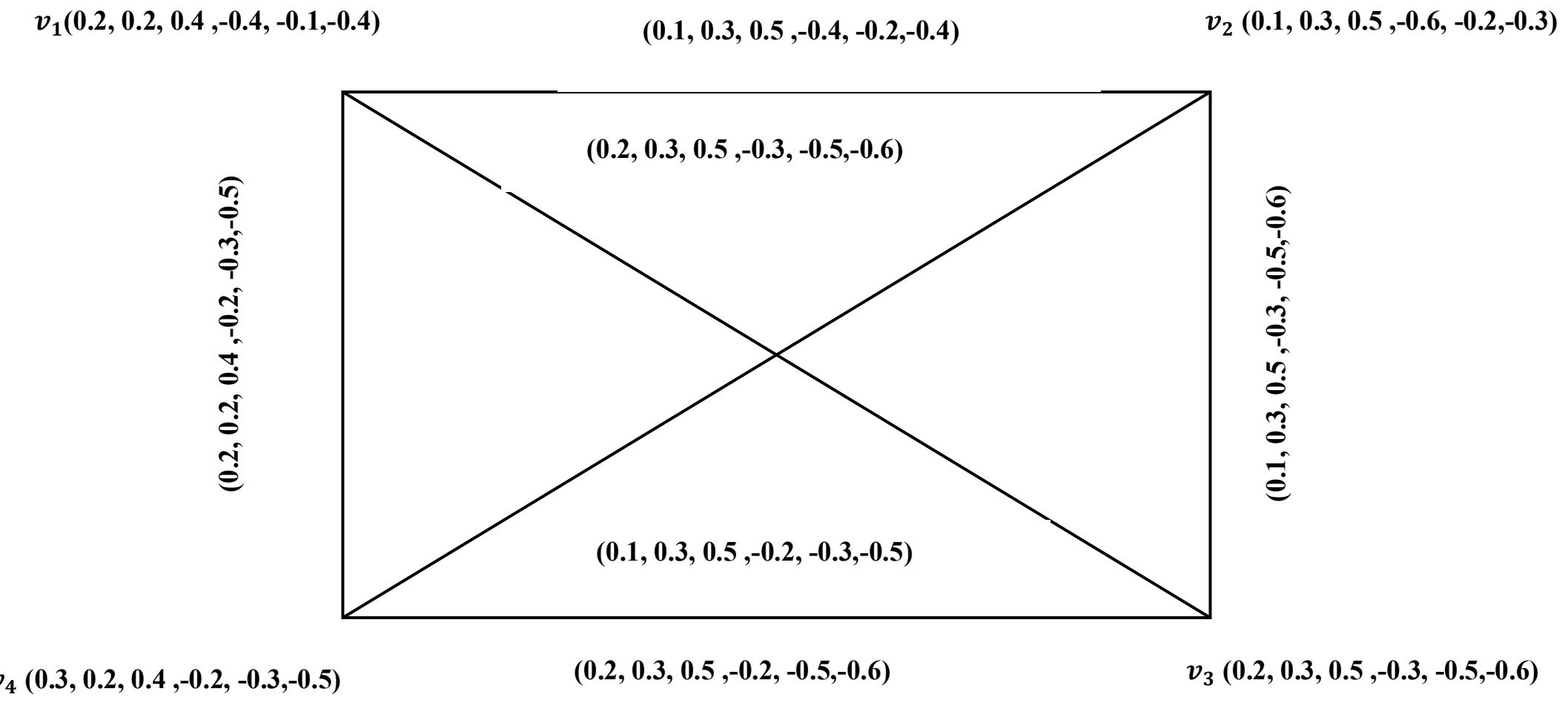

Figure 11: *Complete bipolar single valued neutrosophic graph G.*

$d(v_1)$= (0.5, 0.8, 1.4, -0 .9, -1, -1.5)
$d(v_2)$= (0.4, 0.9, 1.5, -1 .2, -1, -1.6)
$d(v_3)$= (0.4, 0.9, 1.5, -0 .7, -1.3, -1.7)
$d(v_4)$= (0.5, 0.8, 1.4, -0 .6, -1.1, -1.6)

**Definition 3.30.** The complement of a bipolar single valued neutrosophic graph G = (A, B) of a graph $G^*$= (V, E) is a bipolar single valued neutrosophic graph $\bar{G} = (\bar{A}, \bar{B})$ of $\overline{G^*}$ = (V,V ×V), where $\bar{A}$ = A = $(T_A^P, I_A^P, F_A^P, T_A^N, I_A^N, F_A^N)$ and $\bar{B} = (\overline{T_B^P}, \overline{I_B^P}, \overline{F_B^P}, \overline{T_B^N}, \overline{I_B^N}, \overline{F_B^N})$
is defined by

$\bar{T}_B^P$(u, v) = $\min(T_A^P(u), T_A^P(v))$ - $T_B^P(u,v)$ for all $u, v$ ∈ V, uv ∈ $\tilde{V}^2$
$\bar{I}_B^P$(u, v) = $\max(I_A^P(u), I_A^P(v))$ - $I_B^P(u,v)$ for all $u, v$ ∈ V, uv ∈ $\tilde{V}^2$
$\bar{F}_B^P$(u, v) = $\max(F_A^P(u), F_A^P(v))$ - $F_B^P(u,v)$ for all $u, v$ ∈ V, uv ∈ $\tilde{V}^2$
$\bar{T}_B^N$(u, v) = $\max(T_A^N(u), T_A^N(v))$ - $T_B^N(u,v)$ for all $u, v$ ∈ V, uv ∈ $\tilde{V}^2$
$\bar{I}_B^N$(u, v) = $\min(I_A^N(u), I_A^N(v))$ - $I_B^N(u,v)$ for all $u, v$ ∈ V, uv ∈ $\tilde{V}^2$
$\bar{F}_B^N$(u, v) = $\min(F_A^N(u), F_A^N(v))$ - $F_B^N(u,v)$ for all $u, v$ ∈ V, uv ∈ $\tilde{V}^2$

**Proposition 3.31:** The complement of complete BSVN-graph is a BSVN-graph with no edge. Or if G is a complete then in $\bar{G}$ the edge is empty.

**Proof**

Let G= (V, E) be a complete BSVN-graph. $T_B^P(u,v)$ =min $(T_A^P(u), T_A^P(v))$,
So $T_B^P(u,v)$ =min $(T_A^P(u), T_A^P(v))$, $T_B^N(u,v)$ =max $(T_A^N(u), T_A^N(v))$,

$I_B^P(u,v)$ =max $(T_A^P(u), T_A^P(v))$, $I_B^N(u,v)$ =min $(I_A^N(u), I_A^N(v))$,

$F_B^P(u,v)$ =max $(T_A^P(u), T_A^P(v))$, $F_B^N(u,v)$ =min $(F_A^N(u), F_A^N(v)$ for all $u, v \in$ V

Hence in $\bar{G}$,
$\bar{T}_B^P$= $\min(T_A^P(u), T_A^P(v))$ - $T_B^P(u,v)$ for all $u, v \in$ V
= $\min(T_A^P(u), T_A^P(v))$ - $\min(T_A^P(u), T_A^P(v))$ for all $u, v \in$ V
= 0 for all $u, v \in$ V
and
$\bar{I}_B^P$= $\max(I_A^P(u), I_A^P(v))$ - $I_B^P(u,v)$ for all $u, v \in$ V
= $\max(I_A^P(u), I_A^P(v))$ - $\max(I_A^P(u), I_A^P(v))$ for all $u, v \in$ V
= 0 for all $u, v \in$ V
Also
$\bar{F}_B^P$= $\max(F_A^P(u), F_A^P(v))$ - $F_B^P(u,v)$ for all $u, v \in$ V
= $\max(F_A^P(u), F_A^P(v))$ - $\max(F_A^P(u), F_A^P(v))$ for all $u, v \in$ V
= 0 for all $u, v \in$ V
Similarly
$\bar{T}_B^N$= $\max(T_A^N(u), T_A^N(v))$ - $T_B^N(u,v)$ for all $u, v \in$ V
= $\max(T_A^N(u), T_A^N(v))$ - $\max(T_A^N(u), T_A^N(v))$ for all $u, v \in$ V
= 0 for all $u, v \in$ V
and
$\bar{I}_B^P$= $\min(I_A^N(u), I_A^N(v))$ - $I_B^N(u,v)$ for all $u, v \in$ V
= $\min(I_A^N(u), I_A^N(v))$ - $\min(I_A^N(u), I_A^N(v))$ for all $u, v \in$ V
= 0 for all $u, v \in$ V
Also
$\bar{F}_B^N$= $\min(F_A^N(u), F_A^N(v))$ - $F_B^N(u,v)$ for all $u, v \in$ V
= $\min(F_A^N(u), F_A^N(v))$ - $\min(F_A^N(u), F_A^N(v))$ for all $u, v \in$ V
= 0 for all $u, v \in$ V
$(\bar{T}_B^P, \bar{I}_B^P, \bar{F}_B^P, \bar{T}_B^N, \bar{I}_B^N, \bar{F}_B^N)$
Thus $(\bar{T}_B^P, \bar{I}_B^P, \bar{F}_B^P, \bar{T}_B^N, \bar{I}_B^N, \bar{F}_B^N)$= (0, 0, 0, 0, 0)
Hence the edge set of $\bar{G}$ is empty if G is a complete BSVNG.

**Definition 3.32:** A regular BSVN-graph is a BSVN-graph where each vertex has the same number of open neighbors degree. $d_N$(v)= $(d_{NT}^P(v), d_{NI}^P(v), d_{NF}^P(v), d_{NT}^N(v), d_I^N(v), d_{NF}^N(v))$.

The following example shows that there is no relationship between regular BSVN-graph and a constant BSVN-graph

**Example 3.33.** Consider a graph $G^*$ such that V= $\{v_1, v_2, v_3, v_4\}$, E = $\{v_1v_2, v_2v_3, v_3v_4, v_4v_1\}$. Let A be a single valued neutrosophic subset of V and le B a single valued neutrosophic subset of E denoted by

| | $v_1$ | $v_2$ | $v_3$ | $v_4$ |
|---|---|---|---|---|
| $T_A^P$ | 0.2 | 0.2 | 0.2 | 0.2 |
| $I_A^P$ | 0.2 | 0.2 | 0.2 | 0.2 |

| | $v_1v_2$ | $v_2v_3$ | $v_3v_4$ | $v_4v_1$ |
|---|---|---|---|---|
| $T_B^P$ | 0.1 | 0.1 | 0.1 | 0.2 |
| $I_B^P$ | 0.3 | 0.3 | 0.5 | 0.3 |
| $F_A^P$ | 0.4 | 0.4 | 0.4 | 0.4 |
| $T_A^N$ | -0.4 | -0.4 | -0.4 | -0.4 |
| $I_A^N$ | -0.1 | -0.4 | -0.1 | -0.1 |
| $F_A^N$ | -0.4 | -0.4 | -0.4 | -0.4 |
| $F_B^P$ | 0.6 | 0.6 | 0.6 | 0.5 |
| $T_B^N$ | -0.2 | -0.1 | -0.1 | -0.2 |
| $I_B^N$ | -0.3 | -0.6 | -0.6 | -0.3 |
| $F_B^N$ | -0.5 | -0.7 | -0.7 | -0.5 |

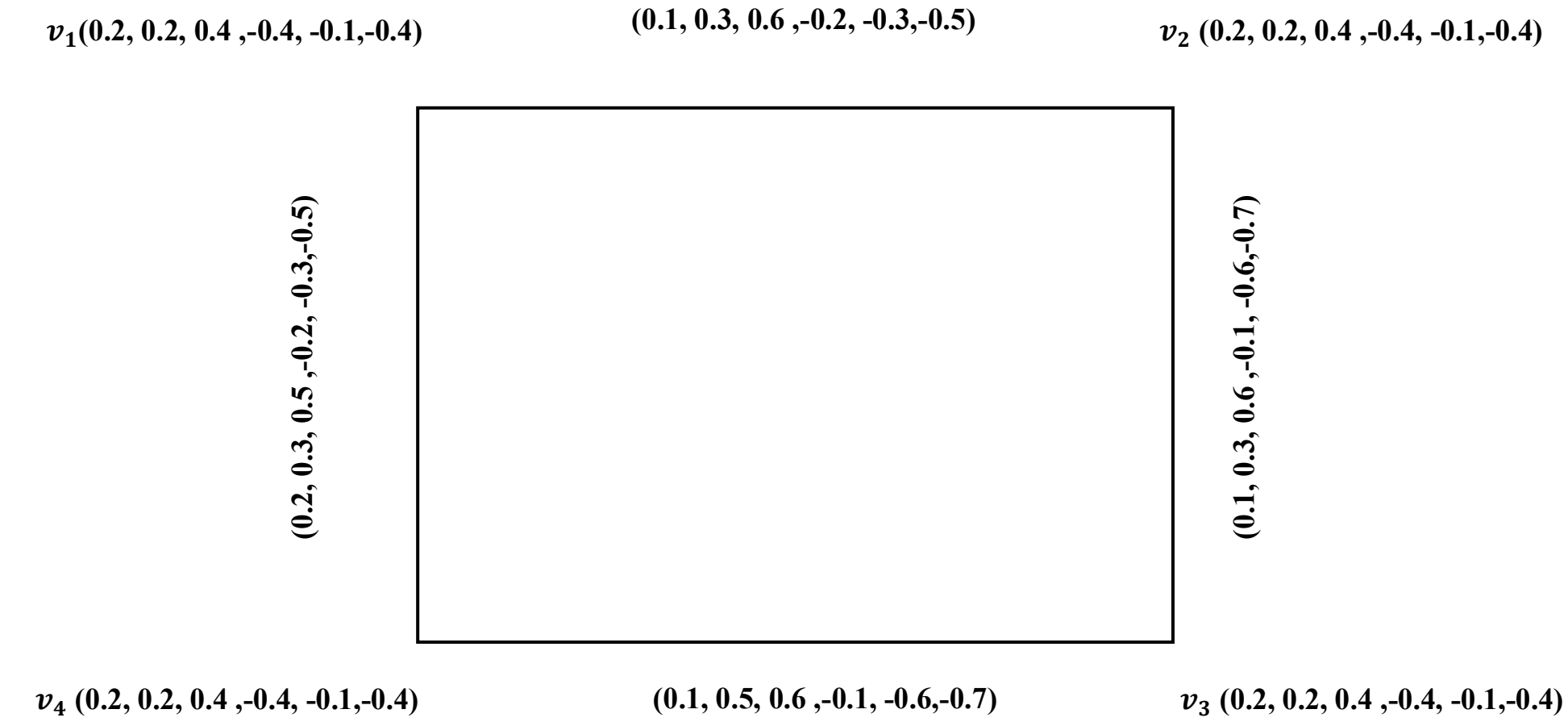

Figure 12: *Regular bipolar single valued neutrosophic graph G.*

By routing calculations show that G is regular BSVN-graph since each open neighbors degree is same, that is (0.4, 0.4, 0.8, -0.8, -0.2, -0.8). But it is not constant BSVN-graph since degree of each vertex is not same.

**Definition 3.34:** Let G = (V, E) be a bipolar single valued neutrosophic graph. Then the totally degree of a vertex $v \in V$ is defined by

td(v)= $(td_T^P(v), td_I^P(v), td_F^P(v), td_T^N(v), td_I^N(v), td_F^N(v))$ where

$td_T^P(v) = \sum_{u \neq v} T_B^P(u,v) + T_A^P(v)$ denotes the totally positive T- degree of a vertex v,

$td_I^P(v) = \sum_{u \neq v} I_B^P(u,v) + I_A^P(v)$ denotes the totally positive I- degree of a vertex v,

$td_F^P(v) = \sum_{u \neq v} F_B^P(u,v) + F_A^P(v)$ denotes the totally positive F- degree of a vertex v,

$td_T^N(v) = \sum_{u \neq v} T_B^N(u,v) + T_A^N(v)$ denotes the totally negative T- degree of a vertex v,

$td_I^N(v) = \sum_{u \neq v} I_B^N(u,v) + I_A^N(v)$ denotes the totally negative I- degree of a vertex v,

$td_F^N(v) = \sum_{u \neq v} F_B^N(u,v) + F_A^N(v)$ denotes the totally negative F- degree of a vertex v

If each vertex of G has totally same degree **m = (**$m_1, m_2, m_3, m_4, m_5, m_6$**)**, then G is called a **m**-totally constant BSVN-Graph.

**Example 3.35**. Let us consider a bipolar single valued neutrosophic graph G= (A, B) of $G^* =$ (V, E), such that V = $\{v_1, v_2, v_3, v_4\}$, E = $\{(v_1, v_2), (v_2, v_3), (v_3, v_4), (v_4, v_1)\}$

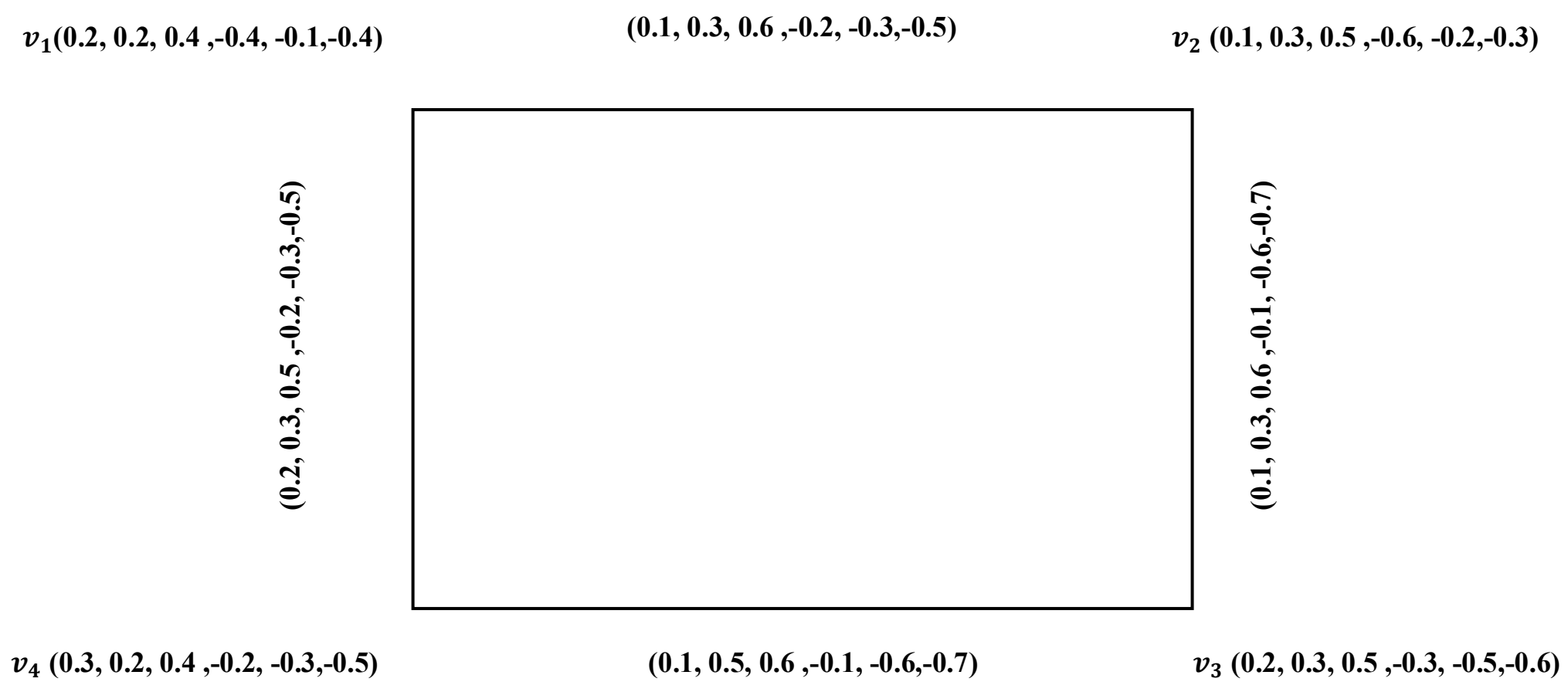

Figure 13: *Totally degree of a bipolar single valued neutrosophic graph G.*

In this example, the totally degree of $v_1$ is (0.5, 0.8, 1.4, -0.8, -0.7, -1.4). The totally degree of $v_2$ is (0.3, 0.9, 1.7, -0.9, -1.1, -1.5). The totally degree of $v_3$ is (0.4, 1.1, 1.7, -0.5, -1.7, -2). The totally degree of $v_4$ is (0.6, 1, 1.5, -0.5, -1.1, -1,7).

**Definition 3.36:** A totally regular BSVN-graph is a BSVN-graph where each vertex has the same number of closed neighbors degree, it is noted d[v]

**Example 3.37**. Let us consider a BSVN-graph G= (A, B) of $G^*$ = (V, E), such that V = $\{v_1, v_2, v_3, v_4\}$ and E = {$(v_1, v_2)$, $(v_2, v_3)$, $(v_3, v_4)$, $(v_4, v_1)$}

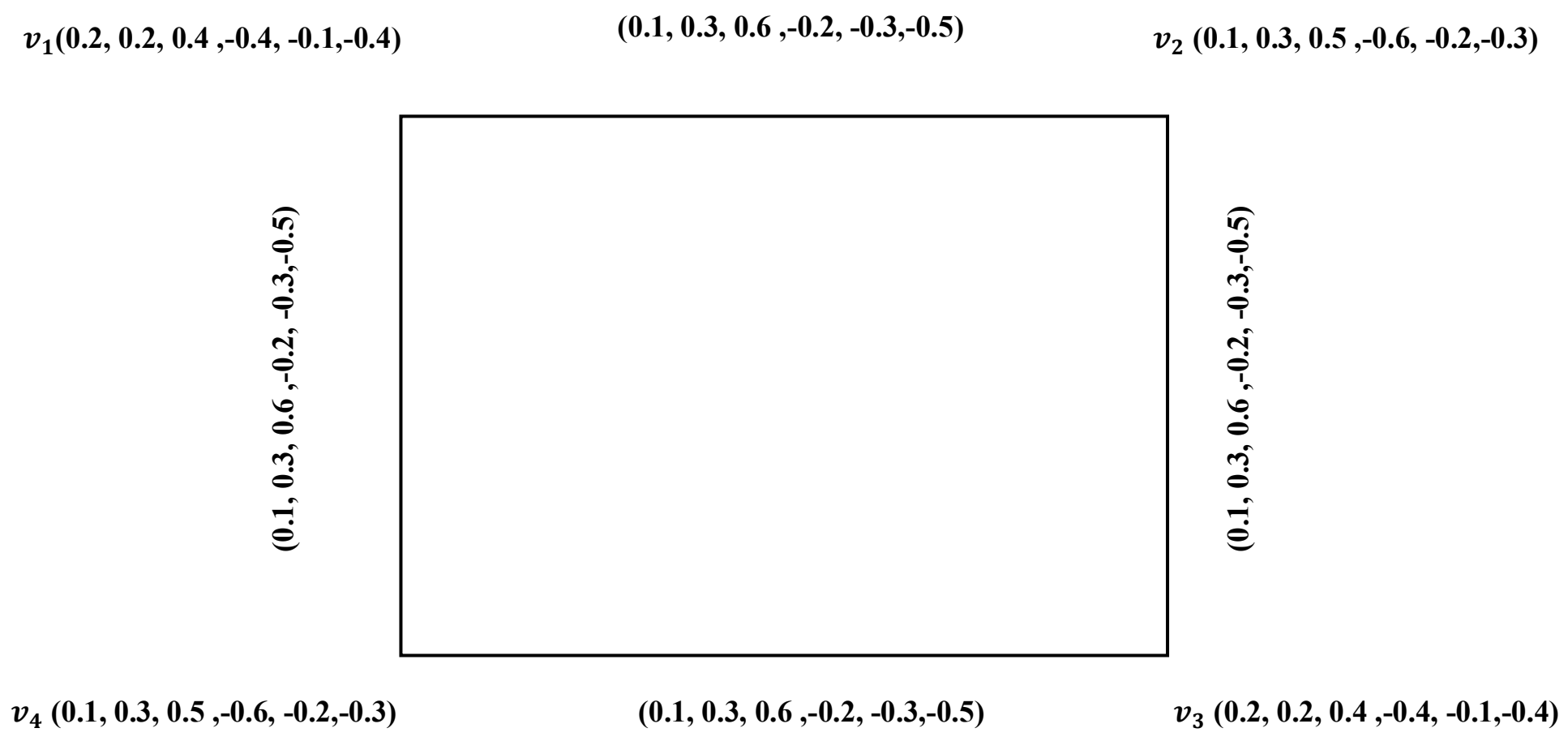

Figure 14: *Degree of a bipolar single valued neutrosophic graph G.*

By routing calculations, we show that G is regular BSVN-graph since the degree of $v_1, v_2, v_3$, and $v_4$ is (0.2, 0.6, 1.2, -0.4, -0.6, -1). It is neither totally regular BSVN-graph not constant BSVN-graph.

## 4. Conclusion

In this paper, we have introduced the concept of bipolar single valued neutrosophic graphs and described degree of a vertex, order, size of bipolar single valued neutrosophic graphs, also we have introduced the notion of complement of a bipolar single valued neutrosophic graph, strong bipolar single valued neutrosophic graph, complete bipolar single valued neutrosophic graph, regular bipolar single valued neutrosophic graph. Further, we are going to study some types of single valued neutrosophic graphs such irregular and totally irregular single valued neutrosophic graphs and bipolar single valued neutrosophic graphs.

### Acknowledgement

The authors are greatful to the editor of "Journal of New Theory" for giving permission to publish the article "On bipolar single valued neutrosophic graphs. *Journal of New Theory, 1(2016), (84-102).*" in the book **"New Trends in Neutrosophic Theories and Applications''.**

## References

1. A. Q. Ansari, R. Biswas & S. Aggarwal, (2012). Neutrosophic classifier: An extension of fuzzy classifier. Elsevier- AppliedSoft Computing, 13, p.563-573 (2013) http://dx.doi.org/10.1016/j.asoc.2012.08.002.
2. A. Q. Ansari, R. Biswas & S. Aggarwal. (Poster Presentation) Neutrosophication of Fuzzy Models, IEEE Workshop On Computational Intelligence: Theories, Applications and Future Directions (hostedby IIT Kanpur), 14th July'13.
3. A. Q. Ansari, R. Biswas & S. Aggarwal Extension to fuzzy logic representation: Moving towards neutrosophic logic - A new laboratory rat, Fuzzy Systems (FUZZ), 2013 IEEE International Conference, p.1 –8, (2013) DOI:10.1109/FUZZ-IEEE.2013.6622412.
4. A. Nagoor Gani and M. Basheer Ahamed, Order and Size in Fuzzy Graphs, Bulletin of Pure and Applied Sciences, Vol 22E (No.1), p.145-148 (2003).
5. A. Nagoor Gani. A and S. Shajitha Begum, Degree, Order and Size in Intuitionistic Fuzzy Graphs, International Journal of Algorithms, Computing and Mathematics, (3)3 (2010).
6. A. Nagoor Gani and S.R Latha, On Irregular Fuzzy Graphs, Applied Mathematical Sciences, Vol.6, no.11,517-523, (2012).
7. A. V. Devadoss, A. Rajkumar & N. J. P . Praveena,.A Study on Miracles through Holy Bible using Neutrosophic Cognitive Maps (NCMS).International Journal of Computer Applications, 69(3) (2013).
8. A. Mohamed Ismayil and A. Mohamed Ali, On Strong Interval-Valued Intuitionistic Fuzzy Graph, International Journal of Fuzzy Mathematics and Systems.Volume 4, Number 2 pp. 161-168 (2014).
9. D. Ezhilmaran & K. Sankar, Morphism of bipolar intuitionistic fuzzy graphs, Journal of Discrete Mathematical Sciences and Cryptography, 18:5, 605-621(2015), DOI:10.1080/09720529.2015.1013673.
10. F. Smarandache. Refined Literal Indeterminacy and the Multiplication Law of Sub-Indeterminacies, Neutrosophic Sets and Systems, Vol. 9, 58-(2015),
11. F. Smarandache, Types of Neutrosophic Graphs and neutrosophic Algebraic Structures together with their Applications in Technology, seminar, Universitatea Transilvania din Brasov, Facultatea de Design de Produs si Mediu, Brasov, Romania 06 June 2015.
12. F. Smarandache, Neutrosophic set - a generalization of the intuitionistic fuzzy set, Granular Computing, 2006 IEEE International Conference, p. 38 – 42 (2006), DOI: 10.1109/GRC.2006.1635754.
13. F. Smarandache, A geometric interpretation of the neutrosophic set — A generalization of the intuitionistic fuzzy set Granular Computing (GrC), 2011 IEEE International Conference, p.602 – 606 (2011), DOI 10.1109/GRC.2011.6122665.
14. F. Smarandache, Symbolic Neutrosophic Theory, Europanova asbl, Brussels, 195p.

15. G. Garg, K. Bhutani, M. Kumar and S. Aggarwal, Hybrid model for medical diagnosis using Neutrosophic Cognitive Maps with Genetic Algorithms, FUZZ-IEEE, 6page 2015(IEEE International conference on fuzzy systems).
16. H .Wang, Y. Zhang, R. Sunderraman, Truth-value based interval neutrosophic sets, Granular Computing, 2005 IEEE International Conference, vol. 1, p. 274 - 277 (2005), DOI: 10.1109/GRC.2005.1547284.
17. H. Wang, F. Smarandache, Y. Zhang, and R. Sunderraman, Single valued Neutrosophic Sets, Multisspace and Multistructure 4, p. 410-413 (2010).
18. H. Wang, F. Smarandache, Zhang, Y.-Q. and R. Sunderraman,"Interval Neutrosophic Sets and Logic: Theory and Applications in Computing", Hexis, Phoenix, AZ, (2005).
19. H. Y Zhang, J. Wang, X. Chen, An outranking approach for multi-criteria decision-making problems with interval-valued neutrosophic sets, Neural Computing and Applications, pp 1-13 (2015).
20. H.Y. Zhang , P. Ji, J. Q.Wang & X. HChen, An Improved Weighted Correlation Coefficient Based on Integrated Weight for Interval Neutrosophic Sets and its Application in Multi-criteria Decision-making Problems, International Journal of Computational Intelligence Systems ,V8, Issue 6 (2015), DOI:10.1080/18756891.2015.1099917.
21. H. D. Cheng and Y. Guo, A new neutrosophic approach to image thresholding, New Mathematics and Natural Computation, 4(3) (2008) 291–308.
22. H.J, Zimmermann, Fuzzy Set Theory and its Applications, Kluwer-Nijhoff, Boston, 1985.
23. A. Deli, M. Ali, F. Smarandache, Bipolar neutrosophic sets and their application based on multi-criteria decision making problems, Advanced Mechatronic Systems (ICAMechS), 2015 International Conference, p.249 - 254 (2015), DOI: 10.1109/ICAMechS.2015.7287068.
24. A. Turksen, Interval valued fuzzy sets based on normal forms, Fuzzy Sets and Systems, vol. 20, p. 191-210 (1986).
25. J. Ye, vector similarity measures of simplified neutrosophic sets and their application in multicriteria decision making, *International Journal of Fuzzy Systems, Vol. 16, No. 2*, p.204-211 (2014).
26. J. Ye, Single-Valued Neutrosophic Minimum Spanning Tree and Its Clustering Method, Journal of Intelligent Systems 23(3): 311–324, (2014).
27. J. Ye, Trapezoidal neutrosophic set and its application to multiple attribute decision making, Neural Computing and Applications, (2014) DOI: 10.1007/s00521-014-1787-6.
28. J. Chen, S. Li, S. Ma, and X. Wang, m-Polar Fuzzy Sets: An Extension of Bipolar Fuzzy Sets, The Scientific World Journal, (2014) http://dx.doi.org/10.1155/2014/416530.
29. K. Atanassov, "Intuitionistic fuzzy sets," Fuzzy Sets and Systems, vol. 20, p. 87-96 (1986).
30. K. Atanassov and G. Gargov, "Interval valued intuitionistic fuzzy sets," Fuzzy Sets and Systems, vol.31, pp. 343-349 (1989).
31. K. Atanassov. Intuitionistic fuzzy sets: theory and applications. Physica, New York, 1999.
32. L. Zadeh Fuzzy sets, Inform and Control, 8, 338-353(1965)
33. M. Akram and W. A. Dudek, Interval-valued fuzzy graphs, Computers & Mathematics with Applications, vol. 61, no. 2, pp. 289–299 (2011).
34. M. Akram, Interval-valued fuzzy line graphs, Neural Computing and Applications, vol. 21, pp. 145–150 (2012).
35. M. Akram and B. Davvaz, Strong intuitionistic fuzzy graphs, Filomat, vol. 26, no. 1, pp. 177–196 (2012).
36. M. Akram, Bipolar fuzzy graphs, Information Sciences, vol. 181, no. 24, pp. 5548–5564 (2011).
37. M. Akram, Bipolar fuzzy graphs with applications, Knowledge Based Systems, vol. 39, pp. 1–8 (2013).
38. M. Akram, K. H. Dar, On N- graphs, Southeast Asian Bulletin of Mathematics, (2012)
39. Akram, M and A. Adeel, m- polar fuzzy graphs and m-polar fuzzy line graphs, Journal of Discrete Mathematical Sciences and Cryptography, 2015.
40. M. Akram, W. A. Dudek, Regular bipolar fuzzy graphs, Neural Computing and Applications, Vol 21, (2012)197-205.

41. M. Ali, and F. Smarandache, Complex Neutrosophic Set, Neural Computing and Applications, Vol. 25, (2016), 1-18. DOI: 10.1007/s00521-015-2154-y.
42. M. Ali, I. Deli, F. Smarandache, The Theory of Neutrosophic Cubic Sets and Their Applications in Pattern Recognition, Journal of Intelligent and Fuzzy Systems, (In press), 1-7, DOI:10.3233/IFS-151906.
43. P. Bhattacharya, Some remarks on fuzzy graphs, Pattern Recognition Letters 6: 297-302 (1987).
44. R. Parvathi and M. G. Karunambigai, Intuitionistic Fuzzy Graphs, Computational Intelligence, Theory and applications, International Conference in Germany, p.18 -20 (2006).
45. S.Broumi, M. Talea, F. Smarandache, Single Valued Neutrosophic Graphs: Degree, Order and Size, (2016) submitted
46. S. Broumi, M. Talea. A. Bakkali and F. Samarandache, Single Valued Neutrosophic Graph, Journal of New theory, N 10 (2016) 86-101.
47. S.Broumi, M. Talea, A.Bakali, F. Smarandache, Interval Valued Neutrosophic Graphs, critical review (2016) in presse
48. S.Broumi, M. Talea, A.Bakali, F. Smarandache, Operations on Interval Valued Neutrosophic Graphs, (2016) submitted.
49. S.Broumi, M. Talea, A.Bakali, F. Smarandache, On Strong Interval Valued Neutrosophic Graphs, (2016) submitted.
50. S. Broumi, F. Smarandache, New distance and similarity measures of interval neutrosophic sets, Information Fusion (FUSION), 2014 IEEE 17th International Conference, p. 1 – 7 (2014).
51. S. Aggarwal, R. Biswas, A. Q. Ansari, Neutrosophic modeling and control , Computer and Communication Technology (ICCCT), 2010 International Conference, p. 718 – 723 (2010) DOI:10.1109/ICCCT.2010.5640435.
52. S. N. Mishra and A. Pal, Product of Interval Valued Intuitionistic fuzzy graph, Annals of Pure and Applied Mathematics Vol. 5, No.1, 37-46 (2013).
53. Y. Hai-Long, G. She, Yanhonge , L. Xiuwu, On single valued neutrosophic relations, Journal of Intelligent & Fuzzy Systems, vol. Preprint, no. Preprint, p. 1-12 (2015).
54. W. B. Vasantha Kandasamy and F. Smarandache, Fuzzy Cognitive Maps and Neutrosophic Congtive Maps,2013.
55. W. B. Vasantha Kandasamy, K. Ilanthenral and Florentin Smarandache, Neutrosophic Graphs: A New Dimension to Graph Theory, Kindle Edition, 2015.
56. W.B. Vasantha Kandasamy and F. Smarandache “Analysis of social aspects of migrant laborers living with HIV/AIDS using Fuzzy Theory and Neutrosophic Cognitive Maps”, Xiquan, Phoenix (2004).
57. W.R. Zhang, "Bipolar fuzzy sets and relations: a computational framework for cognitive modeling and multiagent decision analysis," in Fuzzy Information Processing Society Biannual Conference, 1994. Industrial Fuzzy Control and Intelligent Systems Conference, and the NASA Joint Technology Workshop on Neural Networks and Fuzzy Logic (1994) pp.305-309, doi: 10.1109/IJCF.1994.375115

**SHIMAA FATHI[1], HEWAYDA ELGHAWALBY[2], A.A. SALAMA[3]**

1,2Egypt, Port Said University, Faculty of Engineering, Physics and Engineering Mathematics Department
1Email:Shaimaa_ f_a@eng.psu.edu.eg, 2Email:hewayda2011@eng.psu.edu.eg
3 Egypt, Port Said University, Faculty of Science, Department of Mathematics and Computer Science
3Email:drsalama44@gmail.com

# A Neutrosophic Graph Similarity Measures

## Abstract

This paper is devoted for presenting new neutrosophic similarity measures between neutrosophic graphs. We proposetwo ways to determine the neutrosophic distance between neutrosophic vertex graphs. The two neutrosophic distances are based on the Haussdorff distance, and a robust modified variant of the Haussdorff distance, moreover we show that they both satisfy the metric distance measure axioms. Furthermore, a similarity measure between neutrosophic edge graphs that is based on a probabilistic variant of Haussdorff distance is introduced. The aim is to use those measures for the purpose of matching neutrosophic graphs whose structure can be described in the neutrosophic domain.

## 1. Introduction

Graphs are essential for encoding information, which may serve in several fields ranging from computational biology to computer vision. The notion of graph theory was first introduced by Euler in 1736, given a graph where vertices and edges represent pairwise interactions between entities [2, 5].The past years have witnessed a high development in the areas of the applications of graphs of pattern recognition and computer vision, where graphs are the most powerful and handy tool used in representing both objects and concepts. The invariance properties, as well as the fact that graphs are well suited to model objects in terms of parts and their relations, make them very attractive for various applications. Hence, the theory of graph became an extremely useful tool for solving combinatorial problems in different areas such as geometry, algebra, number theory, topology, operations research, optimization and computer science [1]. In 1975, a fuzzy graph theory as a generalization of Euler's graph theory was introduced by Rosenfeld [7], based on the concepts of fuzzy set theory proposed by Zadeh in 1965 [19].

In a world full of indeterminacy, traditional crisp set with its boundaries of truth and false has not infused itself with the ability of reflecting the reality. Therefore, neutrosophic found its place into contemporary research as an alternative representation of the real world Established by Florentin Smarandache [16], Neutrosophy was presented as the study of "the origin, nature, and scope of neutralities, as well as their interactions with different ideational spectra". The main idea was to consider an entity ”A” in relation to its opposite ”Non-A”, and to that which is neither "A” nor" Non-A", denoted by "Neut-A". From then on, Neutrosophy became the basis of Neutrosophic Logic, Neutrosophic Probability, Neutrosophic Set Theory, and Neutrosophic Statistics. According to this theory every idea "A" tends to be neutralized and balanced by "neut-A" and "non- A" ideas - as a state of equilibrium. In a classical way "A", "neut-A", "anti-A" are disjoint two by two. But, since in many cases the borders between notions are vague, imprecise or sorties, it is possible that "A", "neut-A" and "anti-A" have common parts two by two, or even all three of them as well. In [16, 17], Smarandache introduced the fundamental concepts of neutrosophic set, that had led Salama and Smarandache [15], to provide a mathematical treatment for the neutrosophic phenomena which already existed in our real world. Moreover the work of Salama and Smarandache [15, 16, 17] formed a starting point to construct new branches of neutrosophic mathematics. Hence, Neutrosophic set theory turned out to be a generalization of both the classical and fuzzy counterparts.

In [6, 11, 12, 13], the authors gave a new dimension for the graph theory using the concept of neutrosophy, some study for different types of neutrosophic graphs were presented and some of their properties were investigated. The aim of this paper is to compute the dissimilarity between two graphs, our methodology is based on the Haussdorff distance, which is invariant to rotation. Whereas several neutrosophic distances where introduced in [4, 14], the authors constructed the neutrosophic distance between neutrosophic sets. The remaining of the paper is structured as follows: definitions of neutrosophic sets and graphs are presented in §2 and §3. Whereas, §4 introduces the idea behind the Haussdorff distance between two crisp sets. In §5.2 and §5.3, we propose two new neutrosophic dissimilarity measures between neutrosophicvertex graphs based on the classical and the modified Haussdorff distances. Furthermore, we investigate the metric axioms for the obtained distances. A neutrosophic similarity measure between neutrosophic edge graphs, based on a probabilistic variant of Haussdorff distance, is introduced in §5.3.

## 1. Neutrosophic Sets

let X be a space of points (objects), with a generic element in X denoted by $x$, a neutrosophic set A in X is characterized by a truth-membership function T, a indeterminacy-membership function I and a falsity-membership function F [15, 18], That is: T, I, F: $x \to ]^{-}0, 1^{+}[$.

Where T ($x$), I($x$) and F ($x$) are real standard or non-standard subsets of $]^{-}0, 1^{+}[$.

In general if there is no restriction on the sum of T ($x$), I($x$) and F ($x$), so $0^{-} \leq$ T($x$) + I ($x$) + F ($x$) $\leq 3^{+}$.T, I, F are called neutrosophic components.

In this paper we will restrict our work to use the standard unit interval [0, 1].

## 3. Neutrosophic Graphs

In [6], the authors defined the neutrosophic graph, to be a graph G < V, E > combined with six mappings, written in the form $G_N = < V,\ E,\ T_e,\ I_e,\ F_e,\ T_v,\ I_v,\ F_v >$, where

$T_v$:V→ [0, 1] , $I_v$:V→ [0, 1], $F_v$:V→ [0, 1] denoting the degree of membership ,degree of indeterminacy and non- membership of the element $v_i \in V$ respectively and $0 \leq T_v(v_i) + I_v(v_i) + F_v(v_i) \leq 3$ for every $v_i \in V$, (i = 1, 2, ….. , n) , and

$T_e$: V× V →[0, 1], $I_e$: V× V → [0, 1] and $F_e$: V× V →[0, 1] are such that $T_e(v_i,\ v_j) \leq \min(T_v(v_i),\ T_v(v_j))$,$I_e(v_i,\ v_j) \leq \min(I_v(v_i),\ I_v(v_j))$ and $F_e(v_i,\ v_j) \leq \min(F_v(v_i),\ F_v(v_j))$

and $0 \leq T_e(v_i,\ v_j) + I_e(v_i,\ v_j) + F_e(v_i,) \leq 3$ for every $(v_i,\ v_j) \in$ E (i, j =1, 2, 3, …., n).

The concept of neutrosophic graph was used by several authors; nevertheless they took different points of view when describing the interpretation of graph neutrosophy.

We constructed the following structure depending on the one given in [6, 12].

**3.1. Neutrosophic Edge Graphs:**

A neutrosophic graph is defined as a graph combined with three mappings, written as $G = (V,\ E\ ,\ T_e,\ I_e\ ,\ F_e), where\ \ T_e$:V× V →[0, 1], $I_e$: V× V → [0, 1] and $F_e$: V× V →[0, 1] are such that $T_e(v_i, v_j)\leq$ min $(T_v(v_i),\ T_v(v_j))$,$I_e(v_i, v_j)\leq$ min $(I_v(v_i),\ I_v(v_j))$ and $F_e(v_i, v_j) \leq$ min $(F_v(v_i), F_v(v_j))$ and 0≤ $T_e(v_i, v_j)$+$I_e(v_i, v_j)$+$F_e(v_i, v_j) \leq 3$ for every $(v_i,\ v_j)$ ∈ E (i , j =1,2,3, …., n).

**3.2. Neutrosophic Vertex Graphs**:

The term neutrosophic vertex graph was used to definea graphof the form:

$G = (V\ ,\ E\ ,\ T_v,\ I_v\ ,\ F_v\ )$ combined with three mappings, written as $T_v$:V→ [0, 1], $I_v$:V→ [0, 1], $F_v$:V→ [0, 1] denoting the degree of membership, degree of indeterminacy and non-membership of the element $v_i$∈ V respectively and $0 \leq T_v$ ($v_i$) + $I_v$($v_i$) + $F_v$ ($v_i$) $\leq 3$ for every $v_i$∈ V, (i = 1, 2, ….. , n).

## 4. Haussdorff distance

Since first introduced by Haussdorff in 1914 [8], the Haussdorff distance has been used in several areas including matching and recognition problems. It provides a means of computing the distance between sets of unordered observations when the correspondences between the individual items are unknown. In its most general setting, the Haussdorff distance measures how far two subsets of a metric space are from each other. It turns the set of non-empty compact subsets of a metric space into a metric space in its own right. Given two such sets, the closest point in the second set for each point in the first set is considered. Hence, the Haussdorff distance is the maximum over all these values. More formally, the classical Haussdorff distance (H D) [4, 10], between two finite point sets A and B is given by:

$$\text{H(A, B)} = \max\big(h(A,B), h(B,A)\big)$$

Where the directed Haussdorff distance from A to B is defined to be:

$$h(A,B) = \max_{a\in A} \min_{b\in B} \|a-b\|$$

And $\|.\|$ is some underlying norm on the points of A and B (e.g., the $L_2$ or Euclidean norm). Regardless of the norm, the Haussdorff metric captures the notion of the worst match between two objects. The computed value is the largest distance between a point in one set and a point in the other one. Several variants of the Haussdorff distance have been proposed as alternatives to the maximum of the minimum approach in the classical one; such as Haussdorff fraction, Haussdorff quintile [10] and Spatially Coherent Matching [3].

A robust modified Haussdorff distance (MHD) based on the average distance value instead of the maximum value was proposed by Dubuisson and Jain [7], in this sense they defined the directed distance of the MHD as:

$$MH(A,B) = \frac{1}{N_A} \sum_{a\in \text{A}} \min_{\text{b}\in\text{B}} \|\text{a}-\text{b}\|$$

## 5. Neutrosophic Graph Similarity Measures

In this section, we introduce neutrosophic graph similarity measures, based on the concept of Haussdorffdistance and some of its variants.

Firstly, we propose two new neutrosophic dissimilarity measures based on the classical and the modified Haussdorff distances [4, 6, 14]. Basically the neutrosophic dissimilarity measure is a triple: the first part is a dissimilarity measure of the true value of the neutrosophic object, the second part is a dissimilarity measure of the indeterminate value of the neutrosophic object, and the third part is a dissimilarity measure of the false value of the neutrosophicobject; that is the opposite of the neutrosophic object. Secondly, we propose a new neutrosophic similarity measure based on the probabilistic Haussdorff distance [9]. With a similar structure, the neutrosophic similarity measure is also a triple as the explained in the neutrosophic dissimilarity measure. Obviously, if the indeterminate part does not exist (its measure is zero) and if the measure of the opposite object is ignored the suggested neutrosophic dissimilarity measure is reduced to the concept of Haussdorff distance in the fuzzy sense.

**5.1 NeutrosophicHaussdorff Distance:**

To commence, we consider two neutrosophic vertex graphs

$G_1 = (V_1, E_1, T_{v1}, I_{v1}, F_{v1})$ *and* $G_2 = (V_2, E_2, T_{v2}, I_{v2}, F_{v2})$, where $V_i$, $i$ =1, 2 are the sets of nodes, $E_i$, where i =1,2 are the sets of edges and $T_{vi}, I_{vi}, F_{vi}$, where $i$ =1, 2 are the matrices whose elements are the true, indeterminate and false values defined for each element of $V_i$, $i$ = 1, 2, respectively. We can now write the distances between the two neutrosophic vertex graphs$G_1$, $G_2$ as follows:

$$NGD(G_1, G_2) = (T_{NGD}(G_1, G_2), I_{NGD}(G_1, G_2), F_{NGD}(G_1, G_2))$$

Where,

$$T_{NGD}(G_1, G_2) = \max(T_{NGd}(G_1, G_2), T_{NGd}(G_2, G_1))$$
$$I_{NGD}(G_1, G_2) = \max(I_{NGd}(G_1, G_2), I_{NGd}(G_2, G_1))$$
$$F_{NGD}(G_1, G_2) = \max(F_{NGd}(G_1, G_2), F_{NGd}(G_2, G_1))$$

And

$$T_{NGd}(G_1, G_2) = \max_{i\in V_1}\max_{j\in V_1}\min_{I\in V_2}\min_{J\in V_2}\left\|T_{v_2}(I,J) - T_{v_1}(i,j)\right\|$$
$$I_{NGd}(G_1, G_2) = \max_{i\in V_1}\max_{j\in V_1}\frac{1}{|V_2|\times|V_2|}\sum_{I\in V_2}\sum_{J\in V_2}\left\|I_{v_2}(I,J) - I_{v_1}(i,j)\right\|$$
$$F_{NGd}(G_1, G_2) = \min_{i\in V_1}\min_{j\in V_1}\max_{I\in V_2}\max_{J\in V_2}\left\|F_{v_2}(I,J) - F_{v_1}(i,j)\right\|$$

$NGd(G_2, G_1)$can be computed in a similar way.

**Proposition1**:

The Neutrosophic vertex graph distance NGD satisfies the metric distance measure axioms:

A1) (**Symmetry**): NGD $(G_1, G_2)$ = NGD $(G_2, G_1)$,

A2) (**Non-negativity**): NGD $(G_1, G_2) \geq 0$,

A3) (**Coincidence**): if NGD $(G_1, G_2)$ = 0 then $G_1 = G_2$,

A4) (**Triangle Inequality**): for any three neutrosophic vertex graphs $G_1$, $G_2$ and $G_3$ we have: NGD $(G_1, G_2) \leq$ : NGD $(G_1, G_2)$ + NGD $(G_2, G_3)$.

**Poof**: A1 and A2 can easily be proven.

A3): When NGD $(G_1, G_2)$=$(T_{NGD}(G_1, G_2), I_{NGD}(G_1, G_2), F_{NGD}(G_1, G_2))$ = (0, 0, 0), that is every component of the triple which is the maximum of two positive values is zero, the values of $T_{NGd}(G_i, G_j)$, $I_{NGd}(G_i, G_j)$and $F_{NGd}(G_i, G_j)$ for $i$, j =1, 2 are all zeros. Namely the maximum distance among the nearest nodes in both $G_1$, $G_2$ is zero.That means that the distance between each element of $V_1$ and its nearest element in the set $V_2$ is zero. That is each element in $V_1$ coincides with an element in $V_2$ and vice versa; hence $V_1 = V_2$.

A4): Consider any three neutrosophic graphs$G_1 = (V_1, E_1, T_1, I_1, F_1)$,

$G_2 = (V_2, E_2, T_2, I_2, F_2)$*and* $G_3 = (V_3, E_3, T_3, I_3, F_3)$. For any $i_k, j_k \in V_k$, k =1, 2, 3, we can easily see that:

$$\|T_3(i_3, j_3) - T_1(i_1, j_3)\| \leq \|T_3(i_3, j_3) - T_2(i_2, j_2)\| + \|T_2(i_2, j_2) - T_1(i_1, j_1)\|$$

Where the values $T_K(i_K,\ j_K)$, K=1, 2, 3, lye in the interval [0, 1]. Consequently, one can show that:

$$\max_{i_1\in V_1}\max_{j_1\in V_1}\min_{i_3\in V_3}\min_{j_3\in V_3}\|T_3(i_3,j_3)-T_1(i_1,j_3)\| \le \max_{i_2\in V_2}\max_{j_2\in V_2}\min_{i_3\in V_3}\min_{j_3\in V_3}\|T_3(i_3,j_3)-T_2(i_2,j_2)\| + \max_{i_1\in V_1}\max_{j_1\in V_1}\min_{i_2\in V_2}\min_{j_2\in V_2}\|T_2(i_2,j_2)-T_1(i_1,j_1)\|$$

That is: $T_{NGd}(G_1,G_3) \le T_{NGd}(G_2,G_3) + T_{NGd}(G_1,G_2)$

and similarly $T_{NGd}(G_3,G_1) \le T_{NGd}(G_3,G_2) + T_{NGd}(G_2,G_1)$

Hence, max $(T_{NGd}(G_1,G_3),T_{NGd}(G_3,G_1)) \le$ max $(T_{NGd}(G_2,G_3),T_{NGd}(G_3,G_2))$ + max $(T_{NGd}(G_1,G_2),T_{NGd}(G_2,G_1))$. Then, $T_{NGD}(G_1,G_3) \le T_{NGD}(G_1,G_2) + T_{NGD}(G_2,G_3)$.

The same procedure goes for both $I_{NGD}$ and $F_{NGD}$. That leads to

NGD $(G_1,G_3) \le$ NGD $(G_1,G_2)$ + NGD $(G_2,G_3)$.

**5.2 Modified Neutrosophic Haussdorff Distance:**

Consider two neutrosophic vertex graphs $G_1 = (V_1,\ E_1,\ T_{v1},\ I_{v1},\ F_{v1})$ and $G_2 = (V_2,\ E_2,\ T_{v2},\ I_{v2},\ F_{v2})$, where $V_i$, i = 1, 2 are the sets of nodes, $E_i$, where *i* =1, 2 are the sets of edges and $T_{vi}, I_{vi}, F_{vi}$, where i =1, 2 are the matrices whose elements are the true, indeterminate and false values defined for each element of $V_i$, $i$ = 1, 2, respectively. We can now write the distances between the two neutrosophic vertex graphs $G_1$, $G_2$ as follows:

MNGD $(G_1, G_2) = (T_{MNGD}(G_1,G_2), I_{MNGD}(G_1,G_2), F_{MNGD}(G_1,\ G_2))$

Where,

$T_{\mathrm{MNGD}}(G_1,G_2)$ = max $(T_{MNGd}(G_1, G_2),T_{MNGd}(G_2, G_1)($

$I_{MNGD}(G_1,G_2)$ = max $(I_{MNGd}(G_1, G_2),I_{MNGd}(G_2, G_1))$

$F_{\mathrm{MNGD}}(G_1, G_2)$ = max $(F_{MNGd}(G_1, G_2),F_{MNGd}(G_2, G_1))$

And,

$$\mathrm{T_{MNGd}}(G_1,G_2) = \frac{1}{|V_1|\times|V_1|}\sum_{i\in v_1}\sum_{j\in v_1}\min_{i\in v_2}\min_{j\in v_2}\|T_2(I,J)-T_1(i,j)\|$$

$$\mathrm{I_{MNGd}}(G_1,G_2) = \frac{1}{|V_1|\times|V_1|}\sum_{i\in v_1}\sum_{j\in v_1}\frac{1}{|V_2|\times|V_2|}\sum_{I\in V_2}\sum_{J\in V_2}\|T_2(I,J)-T_1(i,j)\|.$$

$$\mathrm{F_{MNGd}}(\mathrm{G_1},\mathrm{G_2}) = \frac{1}{|\mathrm{V_1}|\times|\mathrm{V_1}|}\sum_{i\in \mathrm{V_1}}\sum_{j\in V_1}\max_{I\in V_2}\max_{J\in V_2}\|F_2(I,J)-F_1(i,j)\|$$

Similarly, we can find MNGd $(G_2, G_1)$.

**Proposition 2:** The Modified Neutrosophic vertex graph distance MNGD satisfies the metric distance measure axioms:

AA1) (symmetry): MNGD $(G_1,\ G_2)$ = MNGD $(G_2,\ G_1)$,

AA2) (non-negativity): MNGD $(G_1,\ G_2) \ge 0$,

AA3) (coincidence): if MNGD $(G_1, G_2) = 0$ then $G_1 = G_2$,

AA4) (triangle inequality): for any three neutrosophic vertex graphs $G_1$,$G_2$ and $G_3$ we have:

MNGD $(G_1, G_3) \le$ MNGD $(G_1, G_2)$ + M NGD $(G_2, G_3)$.

**Proof:** Similar to the procedure used to prove Proposition **1.**

**5.3 ProbabilisticNeutrosophic Haussdorff Distance:**

To overcome the robustness of both the classical and the modified Haussdorff distance, Hue and Hancock [9] have developed a probabilistic variant of the Haussdorff distance. This measure the similarity of the set of attributes rather than using defined set based distance measures. To commence, we recall two edgegraphs $G_1 = (V_1,\ E_1,\ T_{e_1},\ I_{e_1},\ F_{e_1})$, $G_2 = (V_2,\ E_2,\ T_{e_2},\ I_{e_2},$ $F_{e_2}$ ) as mentioned before, the set of all nodes connected to the node I$\in G_2$ by an edge is defined as:

$C_I^2 = \{J|(I,\ J) \in E_2\}$, and the corresponding set of nodes connected to the node $i \in G_1$ by an edge $C_i^1 = \{j|(i,\ j) \in E_1\}$. A measure for the match of the graph $G_2$ onto $G_1$ is:

$$PNGD(G_1, G_2) = (T_{PNGD}(G_1, G_2),\ I_{PNGD}(G_1, G_2), F_{PNGD}(G_1, G_2))$$

where

$$T_{PNGD}(G_1, G_2) = \max(T_{PNGd}(G_1, G_2),\ T_{PNGd}(G_2,\ G_1))$$

$$I_{PNGD}(G_1, G_2) = \max(I_{PNGd}(G_1, G_2),\ I_{PNGd}(G_2,\ G_1))$$

$$F_{\mathrm{PNGD}}(\mathrm{G}_1, \mathrm{G}_2) = \max(F_{PNGd}(\mathrm{G}_1, \mathrm{G}_2),\ \mathrm{F}_{\mathrm{PNGd}}(\mathrm{G}_2,\ \mathrm{G}_1))$$

and

$$T_{PNGd}(G_1, G_2) = \frac{1}{|V_2| \times |V_1|} \sum_{i \in V_1} \sum_{j \in C_i^1} \max_{I \in V_2} \max_{J \in C_I^2} P((i,j) \to (I,J)\ |T_{e_2}(I,J), T_{e_1}(i,j))$$

$$I_{PNGd}(G_1, G_2) = \frac{1}{|V_2| \times |V_1|} \sum_{i \in V_1} \sum_{j \in C_i^1} \max_{I \in V_2} \max_{J \in C_I^2} P((i,j) \to (I,J)\ |I_{e_2}(I,J), I_{e_1}(i,j))$$

$$F_{PNGd}(G_1, G_2) = \frac{1}{|V_2| \times |V_1|} \sum_{i \in V_1} \sum_{j \in C_i^1} \min_{I \in V_2} \min_{J \in C_I^2} P((i,j) \to (I,J)\ |F_{e_2}(I,J), F_{e_1}(i,j))$$

In this formula the posteriori probability $P\ ((i,\ j) \to (I,\ J) \to (I,\ J)|T_{e_2}(I\ ,\ J), T_{e_1}(i,\ j))$ represents the true value for the match of the $G_2$ edge (I, J) onto the $G_1$ edge (*i, j*) provided by the corresponding pair of $T_{e_2}(I,\ J)$ and $T_{e_1}(i,\ j)$. This similarity measure works as follows, it commence with finding the maximum probability over the nodes in $C_I^2$ then averaging the edge compatibilities over the nodes $C_i^1$. Similarly we consider the maximum probability over the nodes in the graph $G_2$ followed by averaging over the nodes in$G_1$.It worth mentioned here that unlike Neutrosophic Haussdorff distance this similarity measure does not satisfy the distance axioms. Moreover, while the true components of the Neutrosophic Haussdorff distance measures the maximum distance between two sets of observations, our measures here returns the maximum similarity. Back to the rest formulae of the posteriori probability which represent the indeterminacy value and the false value for the match of the $G_2$ edge (I, J) onto the $G_1$ edge (*i, j*) using similar procedure to the true value.We still need to compute the probabilities$P((i,\ j) \to (I,J)|T_{e_2}(I,\ J), T_{e_1}(i,\ j))$,

$P\left((i,\ j) \to (I,\ J)\middle|I_{e_2}(I,\ J), I_{e_1}(i,\ j)\right)$ and $P((i,\ j) \to (I,\ J)|F_{e_2}(I,\ J), F_{e_1}(i,\ j))$. For that purpose we will use a robust weighting function:

$$P((i,j) \to (I,J)|T_{e_2}(I,J), T_{e_1}(i,j)) = \frac{\Gamma_\sigma(\|T_{e_2}(I,J), T_{e_1}(i,j)\|)}{\sum_{(I,J)\epsilon E_2} \Gamma_\sigma(\|T_{e_2}(I,J), T_{e_1}(i,j)\|)}$$

$$P((i,j) \to (I,J)|I_{e_2}(I,J), I_{e_1}(i,j)) = \frac{\Gamma_\sigma(\|I_{e_2}(I,J), I_{e_1}(i,j)\|)}{\sum_{(I,J)\epsilon E_2} \Gamma_\sigma(\|I_{e_2}(I,J), I_{e_1}(i,j)\|)}$$

$$P((i,j) \to (I,J)|TF_{e_2}(I,J), TF_{e_1}(i,j)) = \frac{\Gamma_\sigma(\|F_{e_2}(I,J), F_{e_1}(i,j)\|)}{\sum_{(I,J)\epsilon E_2} \Gamma_\sigma(\|F_{e_2}(I,J), F_{e_1}(i,j)\|)}$$

Where$\Gamma_\sigma(.)$ is a distance weighting function. There are several alternative robust weighting functions. For instance, one may consider the Gaussian of the form $\Gamma_\sigma(p) = \exp(\frac{-\rho^2}{2\sigma^2})$ where $\rho^2 = \left(T_{e_2}(I,J) - T_{e_1}(i,j)\right)^2$ according to the true part, $\rho^2 = \left(I_{e_2}(I,J) - I_{e_1}(i,j)\right)^2$ according to the indeterminacy part and $\rho^2 = \left(F_{e_2}(I,J) - F_{e_1}(i,j)\right)^2$ according to the false part, where$\sigma$ is the standard deviation. The similarity measure can be viewed as an average pairwise attribute consistency measure.

## 6. Conclusion and Future Work

Graphs are the most powerful and handy tool used in representing objects and concepts. This paper is dedicated for presenting new neutrosophic similarity and dissimilarity measures between neutrosophic graphs. The proposed distance measures are based on the Haussdorff distance, a modified and a probabilistic variant of the Haussdorff distance, additionally we proved that the given Neutrosophic Haussdorff and the Neutrosophic Modified Haussdorff distances satisfy the metric distance measure axioms. The aim is to use those measures for the purpose of matching graphswhose structure is described in the neutrosophic domain.In our plan for the future we will consider using the deduced measurements in image processing applications, such as image clustering and segmentation.

## References

1. Arora, S., Rao, S. and Vazirani, U. Expander Flows, Geometric Embeddings and Graph Partitioning, In Symposium on Theory of Computing, 2004.

2. Battista, G. D., Eades, P., Tamassia, R. and Tollis, I., Graph Drawing Algorithms for the Visualization of Graphs, Prentice Hall, 1999.

3. Boykov, Y. and Huttenlocher, D.,A New Bayesian Framework for Object Recognition, Proceeding of IEEE Computer Society Conference on CVPR, Vol. 2, pp. 517–523, 1999.

4. Broumi, S. and, Smarandache, F. Several Similarity Measures of Neutrosophic Sets. Neutrosophic Sets and Systems, Vol. 1.1, pp. 54-62, 2013.

5. Chung, F. R. K., Spectral Graph Theory, CBMS Vol. 92, 1997.

6. Dhavaseelan, R., Vikramaprasad, R. and Krishnaraj, V., Certain Types of Neutrosophic Graphs, Int. J. of Mathematical Sciences & Applications, Vol. 5, No. 2, pp. 333-339, July-December, 2015.

7. Dubuisson, M., and Jain, A., A Modified Haussdorff Distance for Object Matching, pp. 566–568.

8.Haussdorff, F., Grundzge der Mengenlehre, Leipzig: Veit and Company, 1914.

9. Heut, B., and Hancock, E. R., Relational Object Recognition From Large Structural Libraries Pattern Recognition, Vol.32, pp.1895-1915, 2002.

10. Huttenlocher, D. , Klanderman, G., and Rucklidge, W., Comparing Images Using the Haussdorff Distance, IEEE. Trans. Pattern Anal. Mach. Intell, Vol. 15, pp. 850–863, 1993.

11. Kandasamy, W. B. Vasantha, Ilanthenral, K. and Smarandache, F. Neutrosophic Graphs: A New Dimension to Graph Theory, 2015.

12. Kandasamy, W., B. Vasantha and Smarandache, F., Basic Neutrosophic Algebraic Structures and their Applications to Fuzzy and Neutrosophic Models, Hexis, 2004.

13. Rajeswari, V. and Parveen Banu, J., A Study on Neutrosophic Graphs. Int. J. Res. Ins., Vol. 2, Issue 2, pp. 8-16, Oct 2015.

14. Salama, A. A., Abdelfattah, M. and Eisa, M., Distances, Hesitancy Degree and Flexible Querying via Neutrosophic Sets, International Journal of Computer Applications, Vol.10, pp. 101, 2014.

15. Salama, A. A. and Smarandache, F. Neutrosophic Crisp Set Theory. Educational Publisher, Columbus, Ohio, USA., 2015.

16. Smarandache, F., A Unifying Field in Logics: Neutrosophic Logic, Neutrosophy, Neutrosophic Set, Neutrosophic Probability, American Research Press, Rehoboth, NM, 1999.

17. Smarandache, F., Neutrosophy and Neutrosophic Logic, First International Conference on Neutrosophy, Neutrosophic Logic, Set, Probability, and Statistics University of New Mexico, Gallup, NM 87301, USA, 2002.

18. Smarandache, F. Neutrosophic Set, A Generalization of The Intuitionistic Fuzzy Sets. Int. J. Pure Appl. Math, Vol. 24, pp.287- 297, 2005.
19. Zadeh, L. A., Fuzzy sets, Inform. Control,Vol. 8, pp. 338 -353, 1965.

**BROUMI, S[1,*], SMARANDACHE, F[2], TALEA, M[3], BAKALI, A[4]**

[1,3] Laboratory of Information Processing, University Hassan II, B.P 7955, Sidi Othman, Casablanca, Morocco.
2 Department of Mathematics, University of New Mexico,705 Gurley Avenue, Gallup, NM 87301, USA.
4 Ecole Royale Navale, Boulevard Sour Jdid, B.P 16303 Casablanca, Morocco.
Corresponding author's E-Mail: broumisaid78@gmail.com

# Operations on Interval Valued Neutrosophic Graphs

## Abstract

Combining the single valued neutrosophic set with graph theory, a new graph model emerges, called single valued neutrosophic graph. This model allows attaching the truth-membership (t), indeterminacy–membership (i) and falsity- membership degrees (f) both to vertices and edges. Combining the interval valued neutrosophic set with graph theory, a new graph model emerges, called interval valued neutrosophic graph. This model generalizes the fuzzy graph, intuitionistic fuzzy graph and single valued neutrosophic graph. In this paper, the authors define operations of Cartesian product, composition, union and join on interval valued neutrosophic graphs, and investigate some of their properties, with proofs and examples.

## 1. Introduction

The neutrosophy was pioneered by F. Smarandache (1995, 1998). It is a branch of philosophy which studies the origin, nature, and scope of neutralities, as well as their interactions with different ideational spectra. The neutrosophic set proposed by Smarandache is a powerful tool to deal with incomplete, indeterminate and inconsistent information in real world, being a generalization of fuzzy set ( Zadeh 1965; Zimmermann 1985), intuitionistic fuzzy set (Atanassov 1986; Atanassov 1999),interval valued fuzzy set (Turksen 1986) and interval valued intuitionistic fuzzy sets (Atanassov and Gargov 1989).The neutrosophic set is characterized by a truth-membership degree (t), an indeterminacy-membership degree (i) and a falsity-membership degree (f) independently, which are within the real standard or nonstandard unit interval $]^{-}0, 1^{+}[$. If the range is restrained within the real standard unit interval [0, 1], the neutrosophic set easily applies to engineering problems. For this purpose, Wang et al. (2010) introduced the concept of single valued neutrosophic set (SVNS) as a subclass of the neutrosophic set. The same author introduced the notion of interval valued neutrosophic sets (Wang et al. 2005b, 2010) as subclass of neutrosophic sets in which the value of truth-membership, indeterminacy-membership and falsity-membership degrees are intervals of numbers instead of real numbers. The single valued neutrosophic set and the interval valued neutrosophic set have been applied in a wide variety of fields, including computer science, engineering, mathematics, medicine and economics (Ansari 2013a, 2013b, 2013c;Aggarwal 2010;Broumi 2014;Deli 2015;Hai-Long 2015;Liu and Shi 2015;Şahin 2015; Wang et al. 2005b;Ye 2014a, 2014b,2014c).

Graph theory has now become a major branch of applied mathematics and it is generally regarded as a branch of combinatorics. Graph is a widely used tool for solving combinatorial problems in different areas, such as geometry, algebra, number theory, topology, optimization and computer science. To be noted that, when there is uncertainty regarding either the set of vertices or edges, or both, the model becomes a fuzzy graph. Many works on fuzzy graphs, intuitionistic fuzzy graphs and interval valued intuitionistic fuzzy graphs (Antonios K et al. 2014; Bhattacharya 1987; Mishra and Pal 2013; Nagoor Gani and Shajitha Begum 2010; Nagoor Gani and Latha 2012; Nagoor Gani and Basheer Ahamed 2003;Parvathi and Karunambigai 2006; Shannon and Atanassov 1994) have been carried out and all of them have considered the vertex sets and edge sets as fuzzy and /or intuitionistic fuzzy sets. But, when the relations between nodes (or vertices) are indeterminate, the fuzzy graphs and intuitionistic fuzzy graphs fail to work. For this purpose, Smarandache (2015a, 2015b, 2015c) defined four main categories of neutrosophic graphs. Two are based on literal indeterminacy (I): I-edge neutrosophic graph and I-vertex neutrosophic graph. The two categories were deeply studied and gained popularity among the researchers (Garg et al. 2015,Vasantha Kandasamy2004, 2013, 2015) due to their applications via real world problems. The other neutrosophic graph categories are based on (t, i, f) components and are called:(t, i, f)-edge neutrosophic graph and (t, i, f)-vertex neutrosophic graph. These two categories are not developed at all.

Further on, Broumi et al. (2016b) introduced a new neutrosophic graph model, called single valued neutrosophic graph (SVNG), and investigated some of its properties as well. This model allows attaching the membership (t), indeterminacy (i) and non-membership degrees (f) both to vertices and edges. The single valued neutrosophic graph is a generalization of fuzzy graph and intuitionistic fuzzy graph. Broumi et al. (2016a) also introduced neighborhood degree of a vertex and closed neighborhood degree of a vertex in single valued neutrosophic graph, as a generalization of neighborhood degree of a vertex and closed neighborhood degree of a vertex in fuzzy graph and intuitionistic fuzzy graph. Moreover, Broumi et al. (2016c) introduced the concept of interval valued neutrosophic graph, as a generalization of single valued neutrosophic graph, and discussed some properties, with proofs and examples. In addition, Broumi et al.(2016c) introduced the concept of bipolar single valued neutrosophic graph, as a generalization of fuzzy graphs, intuitionistic fuzzy graph, N-graph, bipolar fuzzy graph and single valued neutrosophic graph, and studied some related properties.

In this paper, researchers' objective is to define some operations on interval valued neutrosophic graphs, and to investigate some properties.

## 2. Preliminaries

In this section, the authors mainly recall some notions related to neutrosophic sets, single valued neutrosophic sets, interval valued neutrosophic sets, fuzzy graphs, intuitionistic fuzzy graphs, interval valued intuitionistic fuzzy graphs, single valued neutrosophic graphs and interval valued neutrosophic graphs, relevant to the present work. The readers are referred for further details to (Broumi et al. 2016b;Mishra and Pal 2013;Nagoor Gani and Basheer Ahamed 2003;Parvathi and Karunambigai 2006;Smarandache 2006;Wang et al. 2010;Wang et al. 2005a).

**Definition 1** (Smarandache 2006)

Let X be a space of points (objects) with generic elements in X denoted by x; then the neutrosophic set A (NS A) is an object having the form A = {< x: $T_A(x)$, $I_A(x)$, $F_A(x)$>, x ∈X}, where the functions T, I, F: X→$]^-0,1^+[$ define respectively a truth-membership function, an indeterminacy-membership function, and a falsity-membership function of the element x ∈X to the set A with the condition:

$$^-0 \leq T_A(x)+ I_A(x)+ F_A(x)\leq 3^+. \quad (1)$$

The functions $T_A(x)$, $I_A(x)$ and $F_A(x)$ are real standard or nonstandard subsets of $]^-0,1^+[$.

Since it is difficult to apply NSs to practical problems, Wang et al. 2010 introduced the concept of a SVNS, which is an instance of a NS and can be used in real scientific and engineering applications.

**Definition 2** (Wang et al. 2010)

Let X be a space of points (objects) with generic elements in X denoted by x. A single valued neutrosophic set A (SVNS A) is characterized by truth-membership function $T_A(x)$, an indeterminacy-membership function $I_A(x)$, and a falsity-membership function $F_A(x)$. For each point x in X, $T_A(x)$, $I_A(x)$, $F_A(x)$∈ [0, 1]. A SVNS A can be written as

$$A = \{< x: T_A(x), I_A(x), F_A(x)>, x \in X\} \quad (2)$$

**Definition 3** (Wang et al. 2005a)

Let X be a space of points (objects) with generic elements in X denoted by x. An interval valued neutrosophic set (for short IVNS A) A in X is characterized by truth-membership function $T_A(x)$, indeteminacy-membership function $I_A(x)$ and falsity-membership function $F_A(x)$. For each point x in X, one has that

$T_A(x) = [T_{AL}(x), T_{AU}(x)]$,

$I_A(x) = [I_{AL}(x), I_{AU}(x)]$,

$F_A(x) = [F_{AL}(x), F_{AU}(x)] \subseteq [0, 1]$, and

$$0 \leq T_A(x)+ I_A(x)+ F_A(x)\leq 3. \quad (3)$$

**Definition 4** (Wang et al. 2005a)

An IVNS A is contained in the IVNS B, A ⊆ B, if and only if

$$T_{AL}(\mathrm{x}) \leq T_{BL}(\mathrm{x}),\ T_{AU}(\mathrm{x}) \leq T_{BU}(\mathrm{x}),$$
$$I_{AL}(\mathrm{x}) \geq I_{BL}(\mathrm{x}), I_{AU}(\mathrm{x}) \geq I_{BU}(\mathrm{x}),$$
$$F_{AL}(\mathrm{x}) \geq F_{BL}(\mathrm{x}), F_{AU}(\mathrm{x}) \geq F_{BU}(\mathrm{x}), \quad \text{for any x in X.} \qquad (4)$$

**Definition 5** (Wang et al. 2005a)

The union of two interval valued neutrosophic sets A and B is an interval neutrosophic set C, written as C = A ∪ B, whose truth-membership, indeterminacy-membership, and false membership are related to those A and B by

$$T_{CL}(\mathrm{x}) = \max(T_{AL}(\mathrm{x}),\ T_{BL}(\mathrm{x}))$$

$$T_{CU}(\mathrm{x}) = \max(T_{AU}(\mathrm{x}),\ T_{BU}(\mathrm{x}))$$

$$I_{CL}(\mathrm{x}) = \min(I_{AL}(\mathrm{x}),\ I_{BL}(\mathrm{x}))$$

$$I_{CU}(\mathrm{x}) = \min(I_{AU}(\mathrm{x}),\ I_{BU}(\mathrm{x}))$$

$$F_{CL}(\mathrm{x}) = \min(F_{AL}(\mathrm{x}),\ F_{BL}(\mathrm{x}))$$

$$F_{CU}(\mathrm{x}) = \min(F_{AU}(\mathrm{x}),\ F_{BU}(\mathrm{x})), \text{ for all x in X.} \qquad (5)$$

**Definition 6** (Wang et al 2005a)

Let X and Y be two non-empty crisp sets. An interval valued neutrosophic relation R(X, Y) is a subset of product space X × Y, and is characterized by the truth membership function $T_R$(x, y), the indeterminacy membership function $I_R$(x, y), and the falsity membership function $F_R$(x, y), where x ∈ X and y ∈ Y and $T_R$(x, y),$I_R$(x, y),$F_R$(x, y) ⊆ [0, 1].

**Definition 7** (Nagoor Gani and Basheer Ahamed 2003)

A fuzzy graph is a pair of functions G = (σ, μ), where σ is a fuzzy subset of a non-empty set V and μ is a symmetric fuzzy relation on σ, i.e.σ: V → [ 0,1] and μ: VxV→[0,1], such that μ(uv) ≤ σ(u) ⋀σ(v), for all u, v ∈ V where uv denotes the edge between u and v and σ(u) ⋀σ(v) denotes the minimum of σ(u) and σ(v). σ is called the fuzzy vertex set of G andμ is called the fuzzy edge set of G.

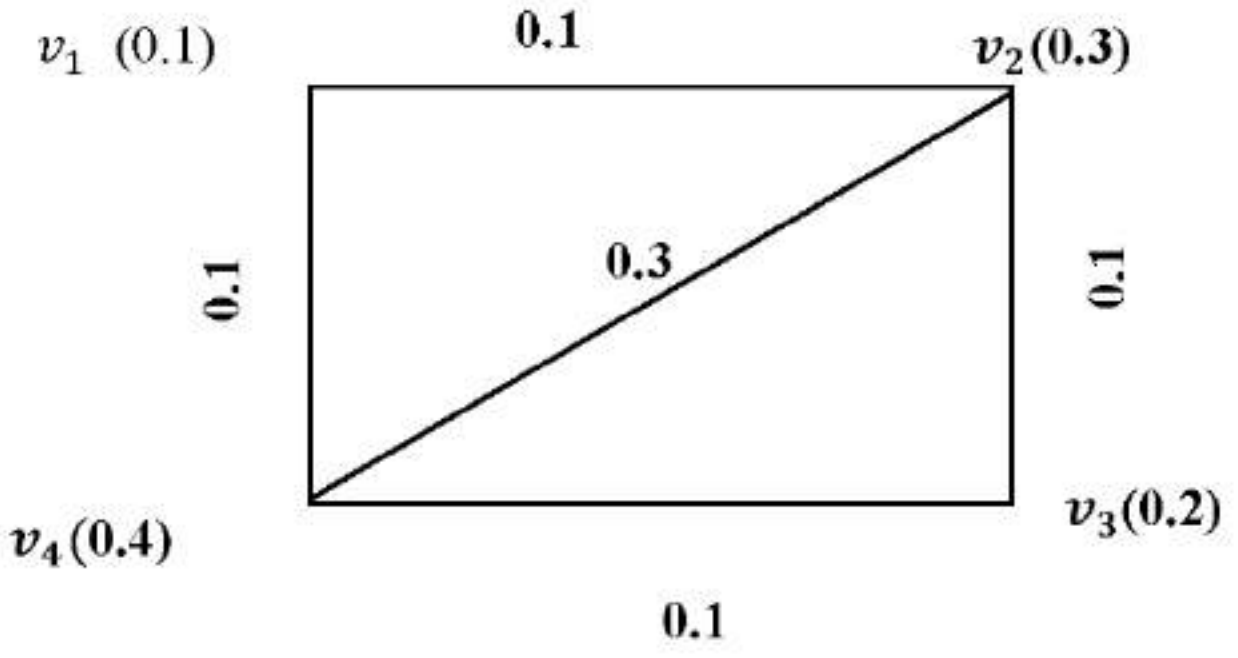

***Figure 1: Fuzzy Graph***

**Definition 8** (Nagoor Gani and Basheer Ahamed 2003)

The fuzzy subgraph H=(τ,ρ) is called a fuzzy subgraph of G =( σ, μ) if τ(u) ≤ σ(u) for all u ∈ V and ρ(u, v) ≤ μ(u, v) for all u, v ∈ V.

**Definition 9** (Parvathi and Karunambigai 2006)

An Intuitionistic fuzzy graph is of the form G=<V,E>,where V={$v_1$,$v_2$,….,$v_n$},such that $\mu_1$:V→ [0,1] and $\gamma_1$:V→ [0,1] denote the degree of membership and non-membership of the elementv$_i$ ∈ V, respectively, and

$$0\leq \mu_1(v_i)+\gamma_1(v_i))\leq 1,\text{forevery } v_i \in V,(i=1, 2,\ldots\ldots n), \quad (6)$$

E ⊆ VxVwhere $\mu_2$:VxV→[0,1]and $\gamma_2$:VxV→ [0,1] are such that $\mu_2(v_i,v_j)\leq \min[\mu_1(v_i),\mu_1(v_j)]$ and $\gamma_2(v_i,v_j)\geq \max[\gamma_1(v_i),\gamma_1(v_j)]$, and

$$0\leq\mu_2(v_i,v_j)+\gamma_2(v_i,v_j)\leq 1 \text{ for every } (v_i,v_j) \in E,(i,j =1,2,\ldots\ldots n) \quad (7)$$

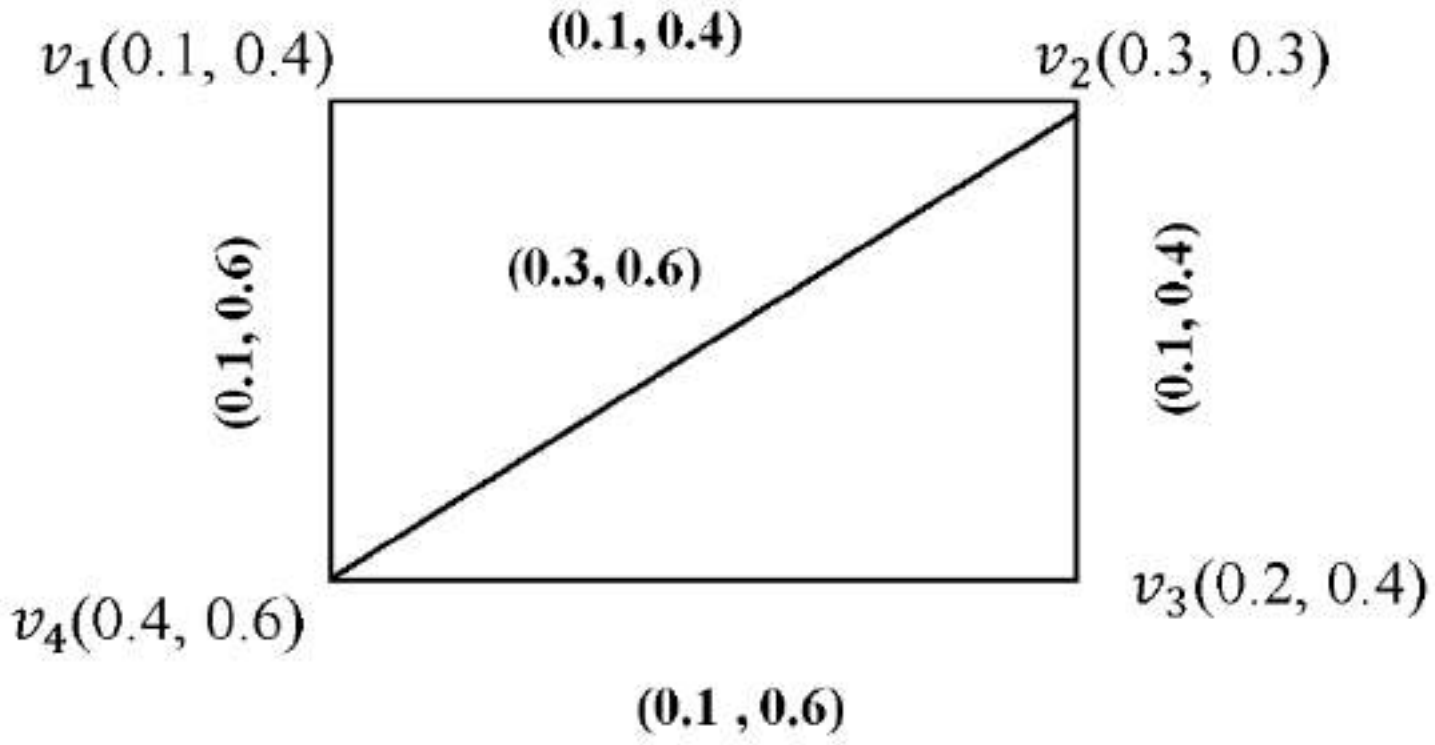

***Figure 2: Intuitionistic Fuzzy Graph***

**Definition 10** (Mishra and Pal 2013)

An interval valued intuitionistic fuzzy graph (IVIFG) G= (A, B) satisfies the following conditions:

1.V= $\{v_1, v_2, \ldots, v_n\}$ such that $M_{AL}$:V→[0, 1],$M_{AU}$:V→[0, 1] and $N_{AL}$:V→[0,1], $N_{AU}$:V→[0, 1] denote the degree of membership and non-membership of the element $y$ ∈ V, respectively, and

$$0 \le M_A(x) + N_A(x) \le 1 \text{ for every } x \in V \quad (8)$$

2.The functions $M_{BL}$:V x V →[0, 1],$M_{BU}$:V x V →[0, 1] and $N_{BL}$:V x V →[0,1], $N_{BU}$:V x V →[0, 1] are denoted by

$$M_{BL}(xy) \le \min\,[M_{AL}(x), M_{AL}(y)],\ M_{BU}(xy) \le \min\,[M_{AU}(x), M_{AU}(y)]$$

$$N_{BL}(xy) \ge \max\,[N_{BL}(x), N_{BL}(y)],\ N_{BU}(xy) \ge \max[N_{BU}(x), N_{BU}(y)]$$

such that $0 \le M_B(xy) + N_B(xy) \le 1$, for every $xy \in E$ (9)

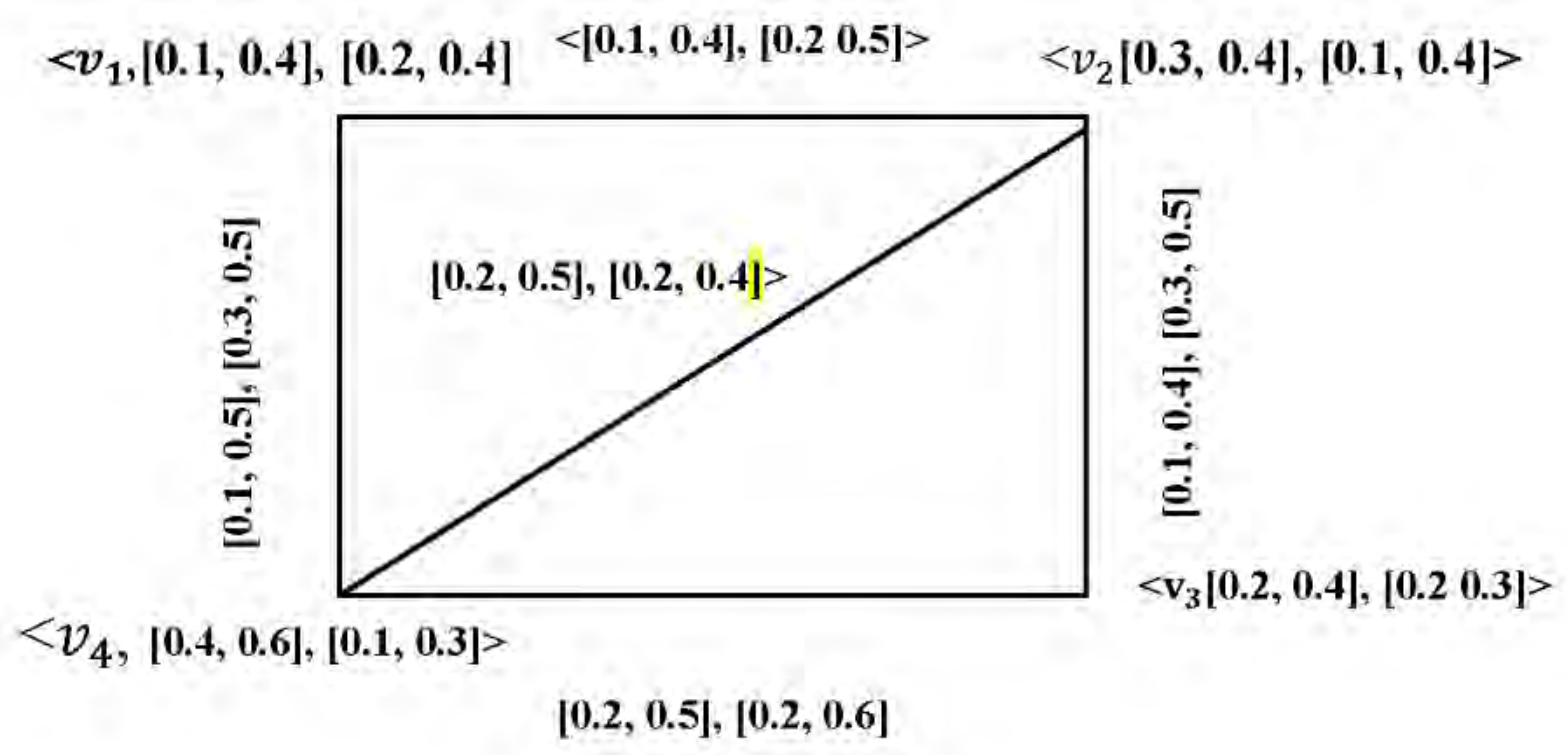

***Figure 3: Interval valued intuitionistic graph***

**Definition 11** (Broumi et al. 2016b)

A single valued neutrosophic graph (SVN-graph) with underlying set V is defined to be a pair G= (A, B), where:

1.The functions $T_A$:V→[0, 1], $I_A$:V→[0, 1] and $F_A$:V→[0, 1] denote the degree of truth-membership, degree of indeterminacy-membership and falsity-membership of the element $v_i$ ∈ V, respectively, and

$$0 \le T_A(v_i) + I_A(v_i) + F_A(v_i) \le 3, \text{ for all } v_i \in V \text{ (i=1, 2, …,n)} \quad (10)$$

2. The functions $T_B$: E ⊆ V x V →[0, 1],$I_B$: E ⊆ V x V →[0, 1] and $F_B$: E ⊆ V x V →[0, 1] are defined by

$T_B(\{v_i, v_j\}) \leq \min [T_A(v_i), T_A(v_j)]$,

$I_B(\{v_i, v_j\}) \geq \max [I_A(v_i), I_A(v_j)]$ and

$F_B(\{v_i, v_j\}) \geq \max [F_A(v_i), F_A(v_j)]$ (11)

and denote the degree of truth-membership, indeterminacy-membership and falsity-membership of the edge $(v_i, v_j) \in$ E respectively, where

$$0 \leq T_B(\{v_i, v_j\}) + I_B(\{v_i, v_j\}) + F_B(\{v_i, v_j\}) \leq 3,$$

for all $\{v_i, v_j\} \in$ E (i, j = 1, 2,…, n). (12)

"A"is called the single valued neutrosophic vertex set of V, "B" - the single valued neutrosophic edge set of E, respectively. B is a symmetric single valued neutrosophic relation on A. The notation $(v_i, v_j)$is used for an element of E. Thus, G = (A, B) is a single valued neutrosophic graph of $G^*$= (V, E), if :

$T_B(v_i, v_j) \leq \min [T_A(v_i), T_A(v_j)]$,

$I_B(v_i, v_j) \geq \max [I_A(v_i), I_A(v_j)]$ and

$F_B(v_i, v_j) \geq \max [F_A(v_i), F_A(v_j)]$, for all $(v_i, v_j) \in$ E (13)

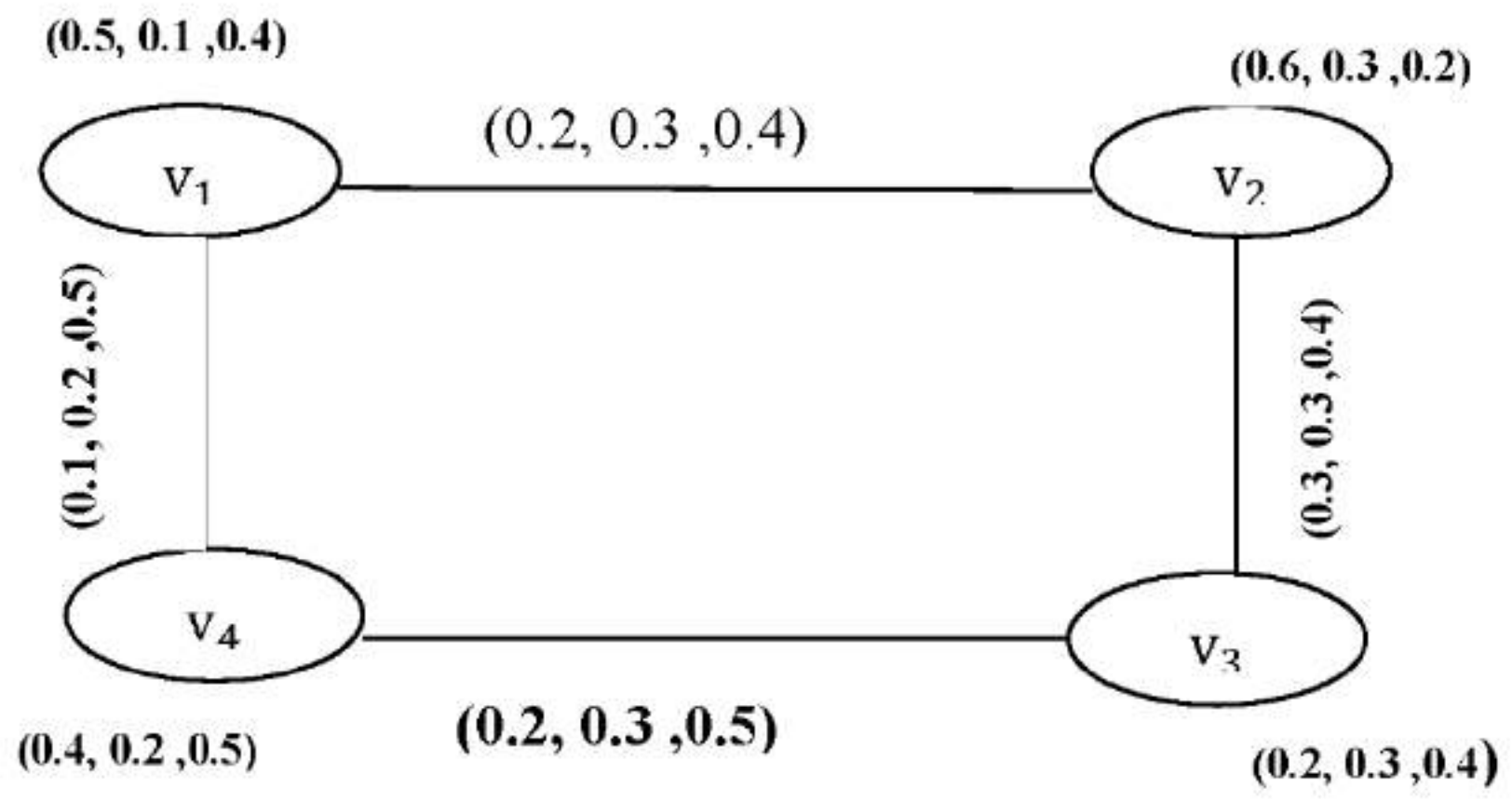

***Figure 4: Single valued neutrosophic graph***

**Definition 12** (Broumi et al. 2016b)

Let G = (A, B) be a single valued neutrosophic graph. Then the degree of a vertex v is defined by d(v)= $(d_T(v), d_I(v), d_F(v))$, where

$d_T(v) = \sum_{u \neq v} T_B(u, v)$, $d_I(v) = \sum_{u \neq v} I_B(u, v)$ and $d_F(v) = \sum_{u \neq v} F_B(u, v)$ (14)

**Definition 13** (Broumi et al. 2016b)

A single valued neutrosophic graph G= (A, B) and $G^*$ is called strong neutrosophic graph

$$T_B(v_i, v_j) = \min [T_A(v_i), T_A(v_j)]$$

$$I_B(v_i, v_j) = \max [I_A(v_i), I_A(v_j)]$$

$$F_B(v_i, v_j) = \max [F_A(v_i), F_A(v_j)] \text{ for all } (v_i, v_j) \in \text{E}. \quad (15)$$

**Definition 14** (Broumi et al. 2016b)

The complement of a strong single valued neutrosophic graph G on $G^*$ is strong single valued neutrosophic graph $\bar{G}$ on $G^*$ where

1.$\bar{V}$ =V

2.$\overline{T_A}(v_i)= T_A(v_i)$,$\overline{I_A}(v_i)= I_A(v_i)$,$\overline{F_A}(v_i)= F_A(v_i)$, $v_j \in$ V.

3.$\overline{T_B}(v_i, v_j)= \min \left[T_A(v_i), T_A(v_j)\right]-T_B(v_i, v_j)$

$\overline{I_B}(v_i, v_j)= \max \left[I_A(v_i), I_A(v_j)\right]-I_B(v_i, v_j)$ and

$\overline{F_B}(v_i, v_j)= \max \left[F_A(v_i), F_A(v_j)\right]-F_B(v_i, v_j)$, for all $(v_i, v_j) \in$ E. (16)

**Definition 15** (Broumi et al. 2016b)

A single valued neutrosophic graph G = (A, B) is called complete, if:

$T_B(v_i, v_j)= \min(T_A(v_i), T_A(v_j))$,

$I_B(v_i, v_j)= \max(I_A(v_i), I_A(v_j))$

and $F_B(v_i, v_j)= \max(F_A(v_i), F_A(v_j))$, for every $v_i, v_j \in$ V. (17)

**Example 1**

Consider a graph $G^*$= (V, E) such that V = {a, b, c, d} , E= {ab ,ac ,bc , cd}. Then, G= (A, B) is a single valued neutrosophic complete graph of $G^*$.

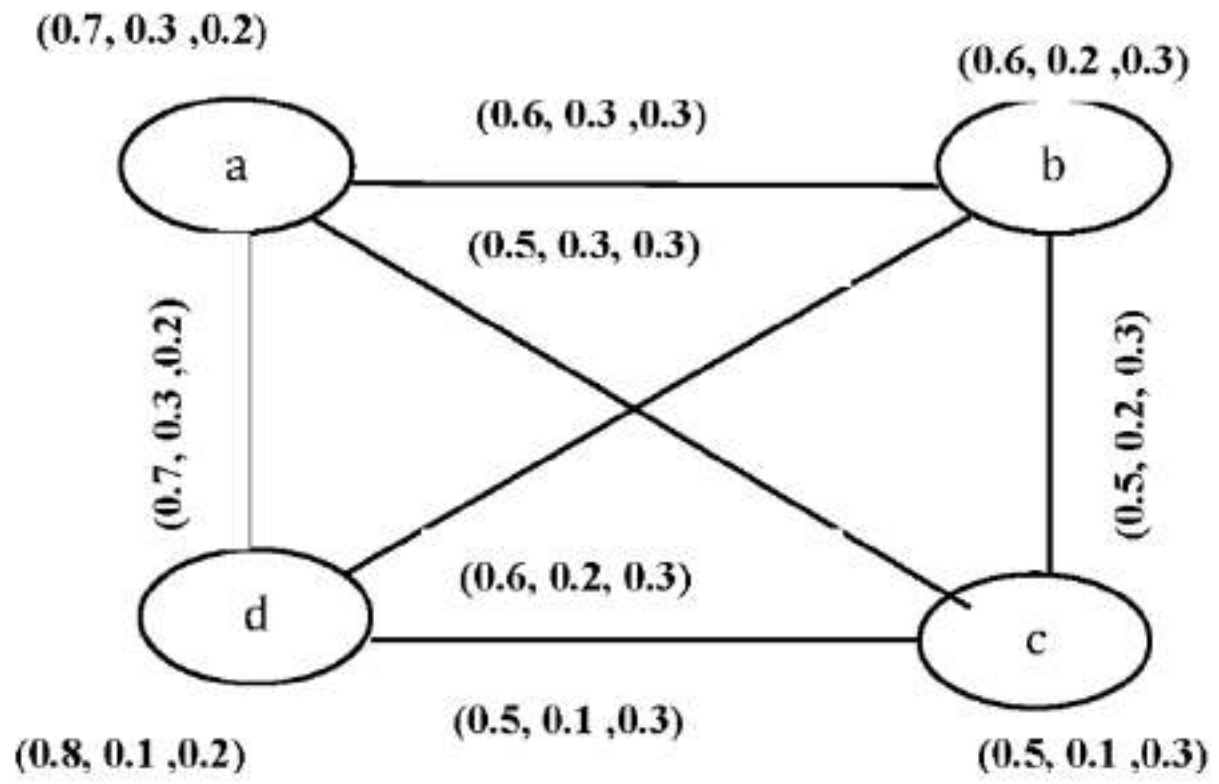

***Figure 5: Complete single valued neutrosophic graph***

## 3. Operations on Interval-Valued Neutrosophic Graphs

Throughout this section, $G^*$ = (V, E) denotes a crisp graph, and G - an interval valued neutrosophic graph.

### Definition 16

By an interval-valued neutrosophic graph of a graph $G^*$=(V,E)one means a pair G=(A,B), where A=< $[T_{AL}, T_{AU}]$, $[I_{AL}, I_{AU}]$, $[F_{AL}, F_{AU}]$>is an interval-valued neutrosophic set on V and B=< $[T_{BL}$, $T_{BU}]$, $[I_{BL}, I_{BU}]$, $[F_{BL}, F_{BU}]$ > is an interval-valued neutrosophic relation on E satisfying the following condition:

1.V= $\{v_1,v_2 ,\dots,v_n\}$ such that $T_{AL}$:V→[0, 1],$T_{AU}$:V→[0, 1], $I_{AL}$:V→[0,1],$I_{AU}$:V→[0, 1] and $F_{AL}$:V→[0,1], $F_{AU}$:V→[0, 1] denote the degree of truth-membership, the degree of indeterminacy-membership and falsity-membership of the element $y \in$ V, respectively, and

$$0\leq T_A(v_i)+ I_A(v_i) +F_A(v_i) \leq 3,$$

for every$v_i \in$ V. (18)

2. The functions $T_{BL}$:V x V →[0, 1],$T_{BU}$:V x V →[0, 1],$I_{BL}$:V x V →[0, 1],$I_{BU}$:V x V →[0, 1] and $F_{BL}$:V x V →[0,1], $F_{BU}$:V x V →[0, 1], such that

$$T_{BL}(v_i, v_j) \leq \min [T_{AL}(v_i), T_{AL}(v_j)]$$

$$T_{BU}(v_i, v_j) \leq \min [T_{AU}(v_i), T_{AU}(v_j)]$$

$$I_{BL}(v_i, v_j) \geq \max [I_{BL}(v_i), I_{BL}(v_j)]$$

$$I_{BU}(v_i, v_j) \geq \max [I_{BU}(v_i), I_{BU}(v_j)]$$

and

$$F_{BL}(v_i, v_j) \geq \max [F_{BL}(v_i), F_{BL}(v_j)]$$

$$F_{BU}(v_i, v_j) \geq \max [F_{BU}(v_i), F_{BU}(v_j)] \quad (19)$$

denote the degree of truth-membership, indeterminacy-membership and falsity-membership of the edge $(v_i,v_j) \in$ E respectively, where

$$0\leq T_B(v_i, v_j) + I_B(v_i, v_j)+ F_B(v_i, v_j) \leq 3,$$

for all $(v_i, v_j) \in$ E. (20)

### Example 2

Figure 5 is an example for IVNG, G = (A,B) defined on a graph $G^*$= (V, E)

such that V = {x, y, z}, E = {xy, yz, zx}, A is an interval valued neutrosophic set of V

A={ < x, [0.5, 0.7], [0.2, 0.3], [0.1, 0.3]>, <y, [0.6, 0.7],[0.2, 0.4], [0.1, 0.3]>, <z, [0.4, 0.6],[0.1, 0.3], [0.2, 0.4],>}, and B an interval valued neutrosophic set of $E \subseteq V \times V$

B={ <xy, [0.3, 0.6], [0.2, 0.4], [0.2, 0.4]>, <yz, [0.3, 0.5],[0.2, 0.5], [0.2, 0.4]>, <xz, [0.3, 0.5],[0.3, 0.5], [0.2, 0.4]>}.

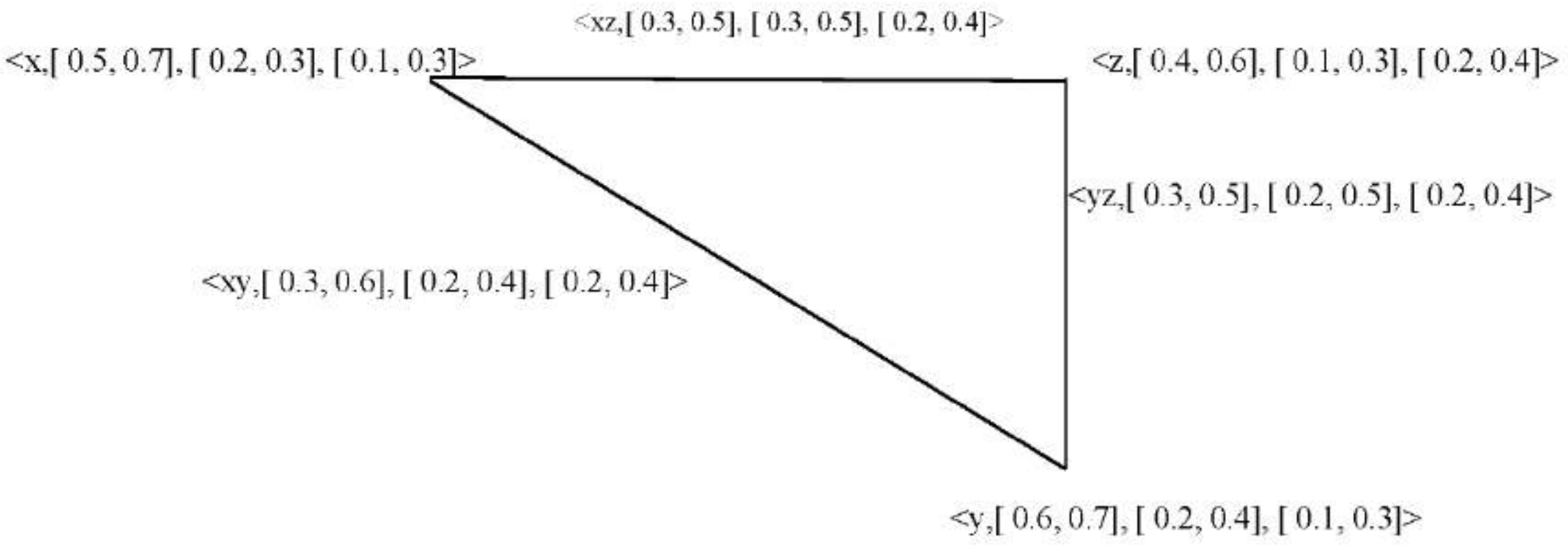

***Figure 6: Interval valued neutrosophic graph***

By routine computations, it is easy to see that G=(A,B) is an interval valued neutrosophic graph of $G^*$.

Here, the new concept of Cartesian product is given.

**Definition 17**

Let $G^* = G_1^* \times G_2^*$=(V, E) be the Cartesian product of two graphs where V = $V_1 \times V_2$and E= $\{(x, x_2)\,(x, y_2)\,/x \in V_1,\, x_2y_2 \in E_2\} \cup \{(x_1,\ z)\,(y_1, z)\,/z \in V_2,\, x_1y_1 \in E_1\}$; then, the Cartesian product G = $G_1 \times G_2$ = ( $A_1 \times A_2$, $B_1 \times B_2$ ) is an interval valued neutrosophic graph defined by

1) $(T_{A_1L} \times T_{A_2L})\,(x_1,x_2) = \min\,(T_{A_1L}(x_1),\, T_{A_2L}(x_2))$
   $(T_{A_1U} \times T_{A_2U})\,(x_1,x_2) = \min\,(T_{A_1U}(x_1),\, T_{A_2U}(x_2))$
   $(I_{A_1L} \times I_{A_2L})\,(x_1,x_2) = \max\,(I_{A_1L}(x_1),\, I_{A_2L}(x_2))$
   $(I_{A_1U} \times I_{A_2U})\,(x_1,x_2) = \max\,(I_{A_1U}(x_1),\, I_{A_2U}(x_2))$
   $(F_{A_1L} \times F_{A_2L})\,(x_1,x_2) = \max\,(F_{A_1L}(x_1),\, F_{A_2L}(x_2))$
   $(F_{A_1U} \times F_{A_2U})\,(x_1,x_2) = \max\,(F_{A_1U}(x_1),\, F_{A_2U}(x_2))$

for all ( $x_1$,$x_2$) $\in V$.

(21)

2) $(T_{B_1L} \times T_{B_2L})\,((x,x_2)(x,y_2)) = \min\,(T_{A_1L}(x),\, T_{B_2L}(x_2y_2))$
   $(T_{B_1U} \times T_{B_2U})\,((x,x_2)(x,y_2)) = \min\,(T_{A_1U}(x),\, T_{B_2U}(x_2y_2))$
   $(I_{B_1L} \times I_{B_2L})\,((x,x_2)(x,y_2)) = \max\,(I_{A_1L}(x),\, I_{B_2L}(x_2y_2))$
   $(I_{B_1U} \times I_{B_2U})\,((x,x_2)(x,y_2)) = \max\,(I_{A_1U}(x),\, I_{B_2U}(x_2y_2))$
   $(F_{B_1L} \times F_{B_2L})\,((x,x_2)\,(x,y_2)) = \max\,(F_{A_1L}(x),\, F_{B_2L}(x_2y_2))$

$(F_{B_1U} \times F_{B_2U})$ $((x, x_2)(x, y_2)) = \max(F_{A_1U}(x), F_{B_2U}(x_2 y_2))$,

$\forall$ x $\in V_1, \forall x_2 y_2 \in E_2$.

(22)

3) $(T_{B_1L} \times T_{B_2L})$ $((x_1, z)\ (y_1, z)) = \min\ (T_{B_1L}(x_1 y_1), T_{A_2L}(z))$

$(T_{B_1U} \times T_{B_2U})$ $((x_1, z)\ (y_1, z)) = \min\ (T_{B_1U}(x_1 y_1), T_{A_2U}(z))$

$(I_{B_1L} \times I_{B_2L})$ $((x_1, z)\ (y_1, z)) = \max\ (I_{B_1L}(x_1 y_1), I_{A_2L}(z))$

$(I_{B_1U} \times I_{B_2U})$ $((x_1, z)\ (y_1, z)) = \max\ (I_{B_1U}(x_1 y_1), I_{A_2U}(z))$

$(F_{B_1L} \times F_{B_2L})$ $((x_1, z)\ (y_1, z)) = \max\ (F_{B_1L}(x_1 y_1), F_{A_2L}(z))$

$(F_{B_1U} \times F_{B_2U})$ $((x_1, z)\ (y_1, z)) = \max\ (F_{B_1U}(x_1 y_1), F_{A_2U}(z))$

$\forall$ z $\in V_2$, $\forall x_1 y_1 \in E_1$. (23)

**Example 3**

Let $G_1^*$= $(A_1, B_1)$ and $G_2^*$= $(A_2, B_2)$ be two graphs where$V_1$ ={a, b}, $V_2$ ={c, d},$E_1$ ={a, b} and $E_2$ ={c, d}.Consider two interval valued neutrosophic graphs:

$A_1$={ <a, [0.5, 0.7], [0.2, 0.3], [0.1, 0.3]>, <b, [0.6, 0.7],[0.2, 0.4], [0.1, 0.3]>},

$B_1$={ <ab, [0.3, 0.6], [0.2, 0.4], [0.2, 0.4]>};

$A_2$={ <c, [0.4, 0.6], [0.2, 0.3], [0.1, 0.3]>, <d, [0.4, 0.7],[0.2, 0.4], [0.1, 0.3}>},

$B_2$={ <cd, [0.3, 0.5], [0.4, 0.5], [0.3, 0.5]>}.

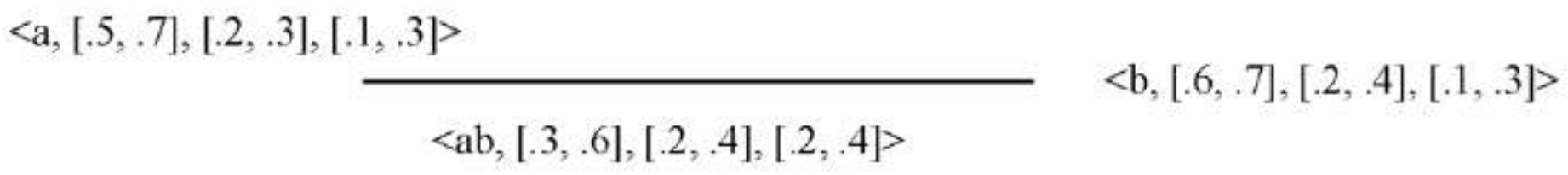

***Figure 7: Interval valued neutrosophic graph G₁***

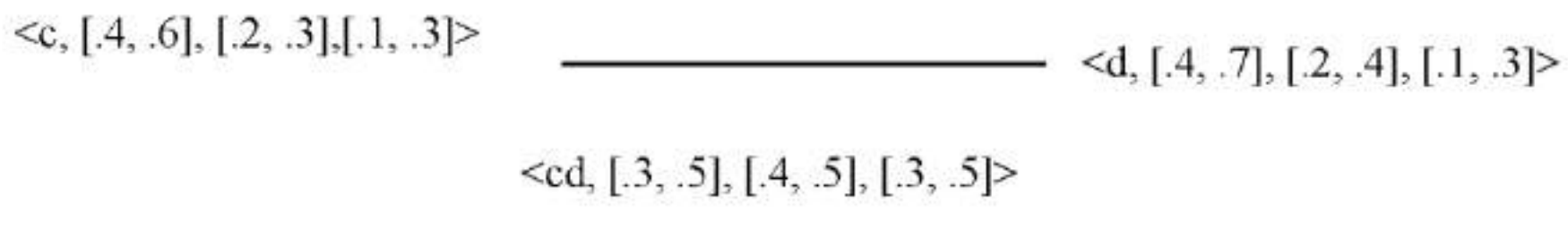

***Figure 8: Interval valued neutrosophic graph G2***

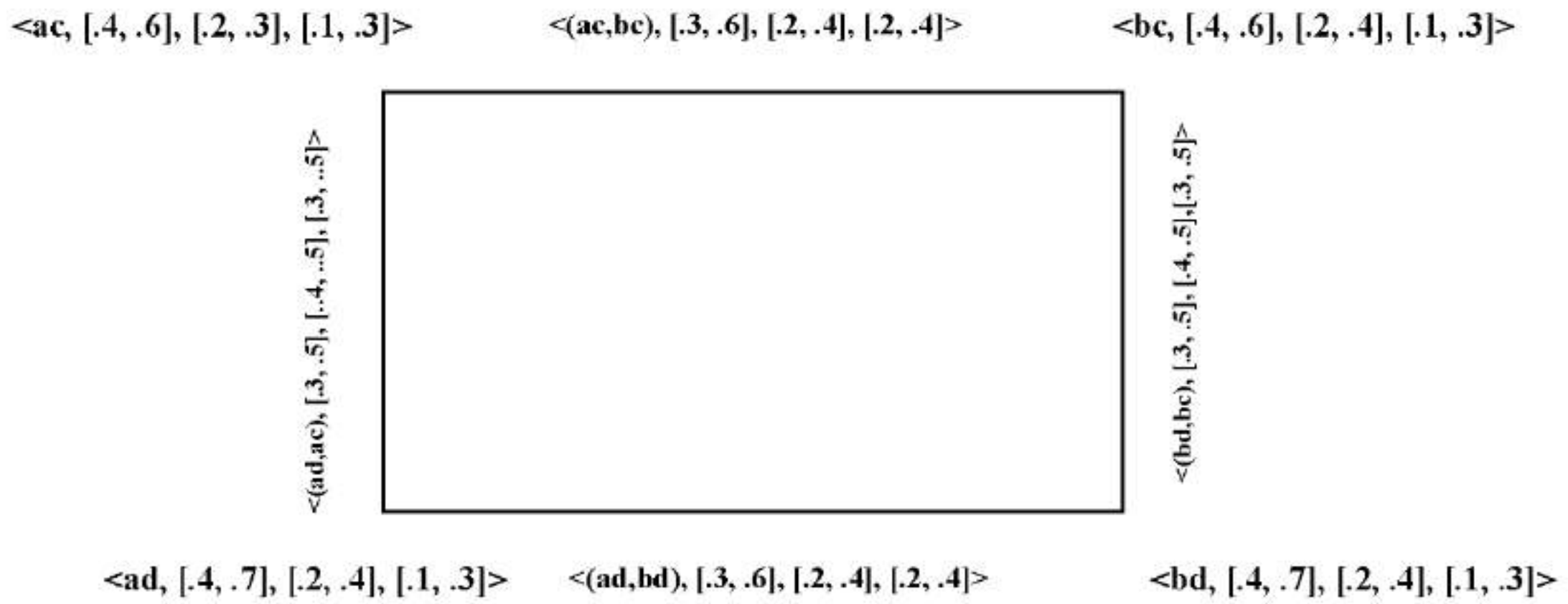

***Figure 9: Cartesian product of interval valued neutrosophic graph***

By routine computations, It is easy to see that $G_1 \times G_2$ is an interval-valued neutrosophic graph of $G_1^* \times G_2^*$.

### Proposition 1

The Cartesian product $G_1 \times G_2 = (A_1 \times A_2, B_1 \times B_2)$ of two interval valued neutrosophic graphs of the graphs $G_1^*$and$G_2^*$ is an interval valued neutrosophic graph of $G_1^* \times G_2^*$.

**Proof.** Verifying only conditions for $B_1 \times B_2$, because conditions for $A_1 \times A_2$ areobvious.

Let E= $\{(x,x_2)\,(x,y_2) / x \in V_1, x_2 y_2 \in E_2\} \cup \{(x_1, z)\,(y_1, z) / z \in V_2, x_1 y_1 \in E_1\}$

Considering $(x,x_2)\,(x,y_2) \in E$, one has:

$(T_{B_1L} \times T_{B_2L})\,((x,x_2)\,(x,y_2)) = \min\,(T_{A_1L}(x), T_{B_2L}(x_2y_2)) \leq \min\,(T_{A_1L}(x), \min(T_{A_2L}(x_2), T_{A_2L}(y_2))) = \min(\min\,(T_{A_1L}(x), T_{A_2L}(x_2)), \min\,(T_{A_1L}(x), T_{A_2L}(y_2)))$

$= \min\,((T_{A_1L} \times T_{A_2L})\,(x,x_2), (T_{A_1L} \times T_{A_2L})\,(x,y_2)),$ (24)

$(T_{B_1U} \times T_{B_2U})\,((x,x_2)\,(x,y_2)) = \min\,(T_{A_1U}(x), T_{B_2U}(x_2y_2)) \leq \min\,(T_{A_1U}(x), \min(T_{A_2U}(x_2), T_{A_2U}(y_2))) = \min(\min\,(T_{A_1U}(x), T_{A_2U}(x_2)), \min T_{A_1U}(x), T_{A_2U}(y_2))) = \min\,((T_{A_1U} \times T_{A_2U})\,(x,x_2), (T_{A_1U} \times T_{A_2U})\,(x,y_2)),$ (25)

$(I_{B_1L} \times I_{B_2L})\,((x,x_2)\,(x,y_2)) = \max\,(I_{A_1L}(x), I_{B_2L}(x_2y_2)) \geq \max\,(I_{A_1L}(x), \max(I_{A_2L}(x_2), I_{A_2L}(y_2))) = \max(\max\,(I_{A_1L}(x), I_{A_2L}(x_2)), \max\,(I_{A_1L}(x), I_{A_2L}(y_2))) = \max\,((I_{A_1L} \times I_{A_2L})\,(x,x_2), (I_{A_1L} \times I_{A_2L})\,(x,y_2)),$ (26)

$(I_{B_1U} \times I_{B_2U})\,((x,x_2)\,(x,y_2)) = \max\,(I_{A_1U}(x), I_{B_2U}(x_2y_2)) \geq \max\,(I_{A_1U}(x), \max(I_{A_2U}(x_2), I_{A_2U}(y_2))) = \max(\max\,(I_{A_1U}(x), I_{A_2U}(x_2)), \max\,(I_{A_1U}(x), I_{A_2U}(y_2))) = \max\,((I_{A_1U} \times I_{A_2U})\,(x,x_2), (I_{A_1U} \times I_{A_2U})\,(x,y_2)),$ (27)

$(F_{B_1L} \times F_{B_2L})$ $((x, x_2)\ (x, y_2))$ = max $(F_{A_1L}(x),\ F_{B_2L}(x_2y_2)) \geq$ max $(F_{A_1L}(x)$, max$(F_{A_2L}(x_2), F_{A_2L}(y_2)))$ = max(max $(F_{A_1L}(x), F_{A_2L}(x_2))$, max $(F_{A_1L}(x), F_{A_2L}(y_2)))$ = max $((F_{A_1L} \times F_{A_2L})\ (x, x_2), (F_{A_1L} \times F_{A_2L})\ (x, y_2))$, (28)

$(F_{B_1U} \times F_{B_2U})$ $((x, x_2)\ (x, y_2))$ = max $(F_{A_1U}(x),\ F_{B_2U}(x_2y_2)) \geq$ max $(F_{A_1U}(x)$, max$(F_{A_2U}(x_2), F_{A_2U}(y_2)))$ = max(max $(F_{A_1U}(x), F_{A_2U}(x_2))$, max $(F_{A_1U}(x), F_{A_2U}(y_2)))$= max $((F_{A_1U} \times F_{A_2U})\ (x, x_2), (F_{A_1U} \times F_{A_2U})\ (x, y_2))$. (29)

Similarly, for$(x_1, z)\ (y_1, z) \in E$, one has:

$(T_{B_1L} \times T_{B_2L})$ $((x_1, z)\ (y_1, z))$ = min $(T_{B_1L}(x_1y_1), T_{A_2L}(z)) \leq$ min $(min(T_{A_1L}(x_1), T_{A_1L}(y_1))), T_{A_2L}(z)))$ = min(min $(T_{A_1L}(x), T_{A_2L}(z))$, min $(T_{A_1L}(y_1), T_{A_2L}(z)))$= min $((T_{A_1L} \times T_{A_2L})\ (x_1, z), (T_{A_1L} \times T_{A_2L})\ (y_1, z))$, (30)

$(T_{B_1U} \times T_{B_2U})$ $((x_1, z)\ (y_1, z))$ = min $(T_{B_1U}(x_1y_1), T_{A_2U}(z)) \leq$ min $(min(T_{A_1U}(x_1), T_{A_1U}(y_1))), T_{A_2U}(z)))$= min(min $(T_{A_1U}(x), T_{A_2U}(z))$, min $(T_{A_1U}(y_1), T_{A_2U}(z)))$ = min $((T_{A_1U} \times T_{A_2U})\ (x_1, z), (T_{A_1U} \times T_{A_2U})\ (y_1, z))$, (31)

$(I_{B_1L} \times I_{B_2L})$ $((x_1, z)\ (y_1, z))$ = max $(I_{B_1L}(x_1y_1), I_{A_2L}(z)) \geq$ max$(max(I_{A_1L}(x_1), I_{A_1L}(y_1))), I_{A_2L}(z)))$ = max(max $(I_{A_1L}(x), I_{A_2L}(z))$, max $(I_{A_1L}(y_1), I_{A_2L}(z)))$ = max $((I_{A_1L} \times I_{A_2L})\ (x_1, z), (I_{A_1L} \times I_{A_2L})\ (y_1, z))$, (32)

$(I_{B_1U} \times I_{B_2U})$ $((x_1, z)\ (y_1, z))$ = max $(I_{B_1U}(x_1y_1), I_{A_2U}(z)) \geq$ max $(max(I_{A_1U}(x_1), I_{A_1U}(y_1))), I_{A_2U}(z)))$ = max(max $(I_{A_1U}(x), I_{A_2U}(z))$, max $(I_{A_1U}(y_1), I_{A_2U}(z)))$ = max $((I_{A_1U} \times I_{A_2U})\ (x_1, z), (I_{A_1U} \times I_{A_2U})\ (y_1, z))$, (33)

$(F_{B_1L} \times F_{B_2L})$ $((x_1, z)\ (y_1, z))$ = max$(F_{B_1L}(x_1y_1), F_{A_2L}(z)) \geq$ max $(max(F_{A_1L}(x_1), F_{A_1L}(y_1))), F_{A_2L}(z)))$ = max(max $(F_{A_1L}(x), F_{A_2L}(z))$, max $(F_{A_1L}(y_1), F_{A_2L}(z)))$ = max $((F_{A_1L} \times F_{A_2L})\ (x_1, z), (F_{A_1L} \times F_{A_2L})\ (y_1, z))$, (34)

$(F_{B_1U} \times F_{B_2U})$ $((x_1, z)\ (y_1, z))$ = max $(F_{A_1U}(x_1y_1), F_{B_2U}(z)) \geq$ max (max$(F_{A_1U}(x_1), F_{A_1U}(y_1))), F_{A_2U}(z)))$ = max(max $(F_{A_1U}(x), F_{A_2U}(z))$, max $(F_{A_1U}(y_1), F_{A_2U}(z)))$ = max $((F_{A_1U} \times F_{A_2U})\ (x_1, z), (F_{A_1U} \times F_{A_2U})\ (y_1, z))$. (35)

This completes the proof.

**Definition 18**

Let $G^* = G_1^* \times G_2^*$=$(V_1 \times V_2$, E) be the composition of two graphs where E= $\{(x, x_2)\ (x, y_2)\ /x \in V_1,\ x_2y_2 \in E_2\} \cup \{(x_1,\ z)\ (y_1, z)\ /z \in V_2,\ x_1y_1 \in E_1\} \cup \{(x_1, x_2)\ (y_1, y_2)\ |x_1y_1 \in E_1,\ x_2 \neq y_2\}$,then the composition of interval valued neutrosophic graphs $G_1[G_2] = (A_1 \circ A_2, B_1 \circ B_2)$ is an interval valued neutrosophic graphs defined by:

1. $(T_{A_1L} \circ T_{A_2L})\ (x_1, x_2) = \min\ (T_{A_1L}(x_1), T_{A_2L}(x_2))$ (36)
   $(T_{A_1U} \circ T_{A_2U})\ (x_1, x_2) = \min\ (T_{A_1U}(x_1), T_{A_2U}(x_2))$

$(I_{A_1L} \circ I_{A_2L})$ $(x_1,x_2)$ = max $(I_{A_1L}(x_1), I_{A_2L}(x_2))$
$(I_{A_1U} \circ I_{A_2U})$ $(x_1, x_2)$ = max $(I_{A_1U}(x_1), I_{A_2U}(x_2))$
$(F_{A_1L} \circ F_{A_2L})$ $(x_1, x_2)$ = max $(F_{A_1L}(x_1), F_{A_2L}(x_2))$
$(F_{A_1U} \circ F_{A_2U})$ $(x_1, x_2)$ = max $(F_{A_1U}(x_1), F_{A_2U}(x_2))$ $\forall x_1 \in V_1, x_2 \in V_2$;

2. $(T_{B_1L} \circ T_{B_2L})$ $((x,x_2)(x,y_2))$ = min $(T_{A_1L}(x), T_{B_2L}(x_2y_2))$ (37)
$(T_{B_1U} \circ T_{B_2U})$ $((x, x_2)(x, y_2))$ = min $(T_{A_1U}(x), T_{B_2U}(x_2y_2))$
$(I_{B_1L} \circ I_{B_2L})$ $((x, x_2)(x, y_2))$ = max $(I_{A_1L}(x), I_{B_2L}(x_2y_2))$
$(I_{B_1U} \circ I_{B_2U})$ $((x, x_2)(x, y_2))$ = max $(I_{A_1U}(x), I_{B_2U}(x_2y_2))$
$(F_{B_1L} \circ F_{B_2L})$ $((x,x_2)(x,y_2))$ = max $(F_{A_1L}(x), F_{B_2L}(x_2y_2))$
$(F_{B_1U} \circ F_{B_2U})$ $((x,x_2)(x,y_2))$ = max $(F_{A_1U}(x), F_{B_2U}(x_2y_2))$ $\forall x \in V_1, \forall x_2y_2 \in E_2$;

3. $(T_{B_1L} \circ T_{B_2L})$ $((x_1,z)(y_1,z))$ = min $(T_{B_1L}(x_1y_1), T_{A_2L}(z))$
(38)
$(T_{B_1U} \circ T_{B_2U})$ $((x_1,z)(y_1,z))$ = min $(T_{B_1U}(x_1y_1), T_{A_2U}(z))$
$(I_{B_1L} \circ I_{B_2L})$ $((x_1,z)(y_1,z))$ = max $(I_{B_1L}(x_1y_1), I_{A_2L}(z))$
$(I_{B_1U} \circ I_{B_2U})$ $((x_1,z)(y_1,z))$ = max $(I_{B_1U}(x_1y_1), I_{A_2U}(z))$
$(F_{B_1L} \circ F_{B_2L})$ $((x_1,z)(y_1,z))$ = max $(F_{B_1L}(x_1y_1), F_{A_2L}(z))$
$(F_{B_1U} \circ F_{B_2U})$ $((x_1,z)(y_1,z))$= max $(F_{B_1U}(x_1y_1), F_{A_2U}(z))$ $\forall$ z $\in V_2, \forall x_1y_1 \in E_1$;

4. $(T_{B_1L} \circ T_{B_2L})$ $((x_1,x_2)(y_1,y_2))$ = min $(T_{A_2L}(x_2), T_{A_2L}(y_2), T_{B_1L}(x_1y_1))$
(39)
$(T_{B_1U} \circ T_{B_2U})$ $((x_1,x_2)(y_1,y_2))$ = min $(T_{A_2U}(x_2), T_{A_2U}(y_2), T_{B_1U}(x_1y_1))$

$(I_{B_1L} \circ I_{B_2L})$ $((x_1,x_2)(y_1,y_2))$ = max $(I_{A_2L}(x_2), I_{A_2L}(y_2), I_{B_1L}(x_1y_1))$

$(I_{B_1U} \circ I_{B_2U})$ $((x_1,x_2)(y_1,y_2))$ = max $(I_{A_2U}(x_2), I_{A_2U}(y_2), I_{B_1U}(x_1y_1))$

$(F_{B_1L} \circ F_{B_2L})$ $((x_1,x_2)(y_1,y_2))$ = max $(F_{A_2L}(x_2), F_{A_2L}(y_2), F_{B_1L}(x_1y_1))$

$(F_{B_1U} \circ F_{B_2U})((x_1, x_2)(y_1, y_2))$ = max $(F_{A_2U}(x_2), F_{A_2U}(y_2), F_{B_1U}(x_1y_1))$,
$\forall (x_1,x_2)(y_1,y_2) \in E^0$-E,where $E^0$= E $\cup$ $\{(x_1,x_2)(y_1,y_2) \mid x_1y_1 \in E_1, x_2 \neq y_2\}$.

**Example 4**

Let $G_1^*$= $(V_1, E_1)$ and $G_2^*$= $(V_2, E_2)$ be two graphs such that $V_1$ ={a, b}, $V_2$ ={c, d},$E_1$ ={a, b} and $E_2$ ={c, d}. Consider two interval-valued neutrosophic graphs:

$A_1$={ < a, [0.5, 0.7], [0.2, 0.3], [0.1, 0.3]>, <b, [0.6, 0.7],[0.2, 0.4], [0.1, 0.3},

$B_1$={ < ab, [0.3, 0.6], [0.2, 0.4], [0.2, 0.4]>};

$A_2$={ < c, [0.4, 0.6], [0.2, 0.3], [0.1, 0.3]>, <d, [0.4, 0.7],[0.2, 0.4], [0.1, 0.3},

$B_2$={ < cd, [0.3, 0.5], [0.2, 0.5], [0.3, 0.5]>}.

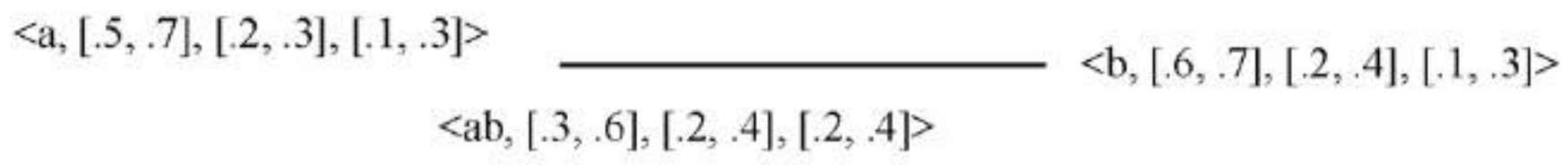

***Figure 10: Interval valued neutrosophic graph G₁***

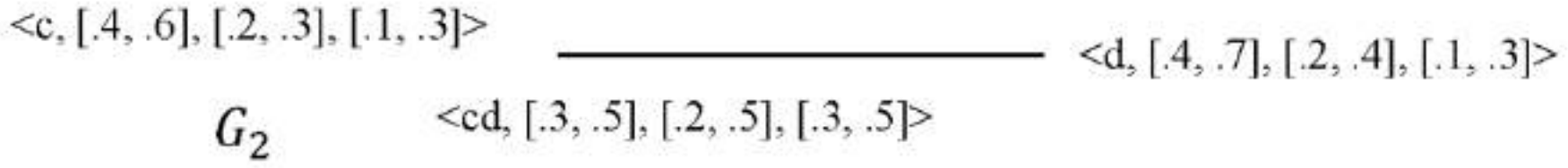

***Figure 11: Interval valued neutrosophic graph $G_2$***

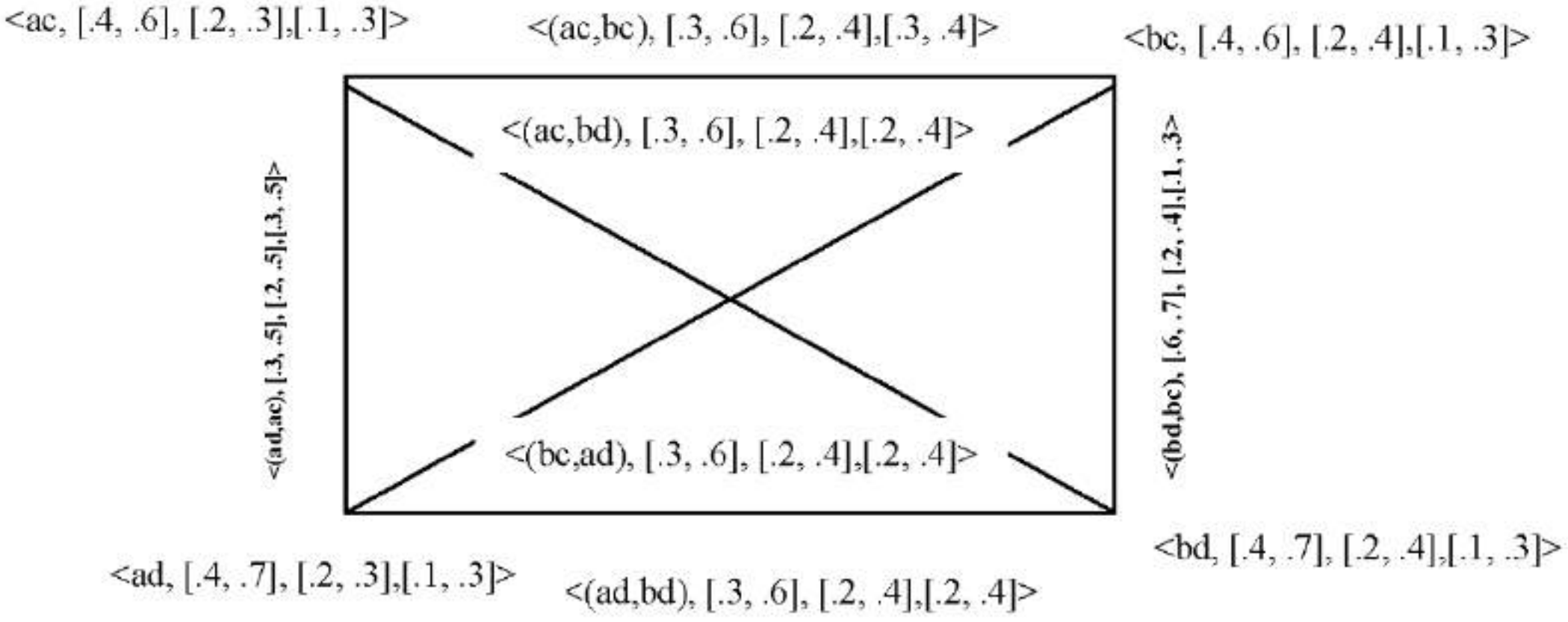

***Figure 12: Composition of interval valued neutrosophic graph.***

**Proposition2**

The composition$G_1[\,G_2]$ =($A_1 \circ A_2$, $B_1 \circ B_2$) of two interval valued neutrosophic graphs of the graphs $G_1^*$and $G_2^*$ is an interval valued neutrosophic graph of $G_1^*[G_2^*]$.

Proof**.** Verifying only conditions for $B_1 \circ B_2$, because conditions for $A_1 \circ A_2$ are obvious. Let E= $\{(x,x_2)\ (x,y_2)\ /x_1 \in V_1,\ x_2y_2 \in E_2\} \cup \{(x_1, z)\ (y_1, z)\ /z \in V_2, x_1y_1 \in E_1\}$.Considering$(x,x_2)$ $(x,y_2) \in E$, one has:

$(T_{B_1L} \circ T_{B_2L})\ ((x, x_2)\ (x, y_2)) = \min\ (T_{A_1L}(x),\ T_{B_2L}(x_2y_2)) \leq \min\ (T_{A_1L}(x),$ $\min(T_{A_2L}(x_2), T_{A_2L}(y_2)))= \min(\min\ (T_{A_1L}(x), T_{A_2L}(x_2)),\ \min\ (T_{A_1L}(x), T_{A_2L}(y_2)))= \min$ $((T_{A_1L} \circ T_{A_2L})\ (x,x_2), (T_{A_1L} \circ T_{A_2L})\ (x,y_2))$, (40)

$(T_{B_1U} \circ T_{B_2U})$ $((x, x_2)$ $(x, y_2))$ = min $(T_{A_1U}(x), T_{B_2U}(x_2y_2)) \leq$ min $(T_{A_1U}(x)$, min$(T_{A_2U}(x_2),T_{A_2U}(y_2)))$= min(min $(T_{A_1U}(x),T_{A_2U}(x_2))$, min $(T_{A_1U}(x),T_{A_2U}(y_2)))$= min $((T_{A_1U} \circ T_{A_2U})$ $(x,x_2),(T_{A_1U} \circ T_{A_2U})$ $(x,y_2))$, (41)

$(I_{B_1L} \circ I_{B_2L})$ $((x, x_2)$ $(x, y_2))$ = max $(I_{A_1L}(x), I_{B_2L}(x_2y_2)) \geq$ max $(I_{A_1L}(x)$, max$(I_{A_2L}(x_2), I_{A_2L}(y_2)))$ = max(max $(I_{A_1L}(x), I_{A_2L}(x_2))$, max $(I_{A_1L}(x), I_{A_2L}(y_2)))$ = max$((I_{A_1L} \circ I_{A_2L})$ $(x,x_2),(I_{A_1L} \circ I_{A_2L})$ $(x,y_2))$, (42)

$(I_{B_1U} \circ I_{B_2U})$ $((x, x_2)$ $(x, y_2))$ = max $(I_{A_1U}(x), I_{B_2U}(x_2y_2)) \geq$ max $(I_{A_1U}(x)$, max$(I_{A_2U}(x_2),I_{A_2U}(y_2)))$ = max(max $(I_{A_1U}(x),I_{A_2U}(x_2))$, max $(I_{A_1U}(x),I_{A_2U}(y_2)))$ = max $((I_{A_1U} \circ I_{A_2U})$ $(x,x_2),(I_{A_1U} \circ I_{A_2U})$ $(x,y_2))$, (43)

$(F_{B_1L} \circ F_{B_2L})$ $((x, x_2)$ $(x, y_2))$ = max $(F_{A_1L}(x), F_{B_2L}(x_2y_2)) \geq$ max $(F_{A_1L}(x)$, max$(F_{A_2L}(x_2),F_{A_2L}(y_2)))$ = max(max $(F_{A_1L}(x),F_{A_2L}(x_2))$, max $(F_{A_1L}(x),F_{A_2L}(y_2)))$ = max $((F_{A_1L} \circ F_{A_2L})$ $(x,x_2),(F_{A_1L} \circ F_{A_2L})$ $(x,y_2))$, (44)

$(F_{B_1U} \circ F_{B_2U})$ $((x, x_2)$ $(x, y_2))$ = max $(F_{A_1U}(x), F_{B_2U}(x_2y_2)) \geq$ max $(F_{A_1U}(x)$, max$(F_{A_2U}(x_2),F_{A_2U}(y_2)))$ = max(max $(F_{A_1U}(x),F_{A_2U}(x_2)$, max $(F_{A_1U}(x),F_{A_2U}(y_2))$ = max$((F_{A_1U} \circ F_{A_2U})$ $(x,x_2),(F_{A_1U} \circ F_{A_2U})$ $(x,y_2))$. (45)

In the case $(x_1, z)$ $(y_1, z) \in E$, the proof is similar.

In the case $(x_1,x_2)$ $(y_1,y_2) \in E^0$-E.

$(T_{B_1L} \circ T_{B_2L})((x_1,x_2)$ $(y_1,y_2))$ = min $(T_{A_2L}(x_2), T_{A_2L}(y_2), T_{B_1L}(x_1y_1)) \leq$ min $(T_{A_2L}(x_2)$, $T_{A_2L}(y_2)$,min $(T_{A_1L}(x_1), T_{A_1L}(y_1)))$ = min(min $(T_{A_1L}(x_1), T_{A_2L}(x_2))$, min $(T_{A_1L}(y_1),T_{A_2L}(y_2)))$ = min $((T_{A_1L} \circ T_{A_2L})$ $(x_1,x_2),(T_{A_1L} \circ T_{A_2L})$ $(y_1,y_2))$, (46)

$(T_{B_1U} \circ T_{B_2U})$ $((x_1,x_2)$ $(y_1,y_2))$ = min $(T_{A_2U}(x_2), T_{A_2U}(y_2), T_{B_1L}(x_1y_1)) \leq$ min $(T_{A_2U}(x_2), T_{A_2U}(y_2)$,min $(T_{A_1U}(x_1), T_{A_1U}(y_1)))$ = min(min $(T_{A_1U}(x_1),T_{A_2U}(x_2))$, min $(T_{A_1U}(y_1),T_{A_2U}(y_2)))$ = min $((T_{A_1U} \circ T_{A_2U})$ $(x_1,x_2),(T_{A_1U} \circ T_{A_2U})$ $(y_1,y_2))$,(47)

$(I_{B_1L} \circ I_{B_2L})$ $((x_1,x_2)$ $(y_1,y_2))$ = max $(I_{A_2L}(x_2), I_{A_2L}(y_2), I_{B_1L}(x_1y_1)) \geq$ max $(I_{A_2L}(x_2)$, $I_{A_2L}(y_2)$,max $(I_{A_1L}(x_1), I_{A_1L}(y_1)))$ = max(max $(I_{A_1L}(x_1), I_{A_2L}(x_2))$, max $(I_{A_1L}(y_1),I_{A_2L}(y_2)))$ = max $((I_{A_1L} \circ I_{A_2L})$ $(x_1,x_2),(I_{A_1L} \circ I_{A_2L})$ $(y_1,y_2))$, (48)

$(I_{B_1U} \circ I_{B_2U})$ $((x_1,x_2)$ $(y_1,y_2))$ = max $(I_{A_2U}(x_2), I_{A_2U}(y_2), I_{B_1L}(x_1y_1)) \geq$ max $(I_{A_2U}(x_2)$, $I_{A_2U}(y_2)$,max $(I_{A_1U}(x_1), I_{A_1U}(y_1)))$ = max(max $(I_{A_1U}(x), I_{A_2U}(x_2))$, max $(I_{A_1U}(y_1),I_{A_2U}(y_2)))$ = max $((I_{A_1U} \circ I_{A_2U})$ $(x_1,x_2),(I_{A_1U} \circ I_{A_2U})$ $(y_1,y_2))$, (49)

$(F_{B_1L} \circ F_{B_2L})$ $((x_1,x_2)$ $(y_1,y_2))$ = max $(F_{A_2L}(x_2), F_{A_2L}(y_2), F_{B_1L}(x_1y_1)) \geq$ max $(F_{A_2L}(x_2), F_{A_2L}(y_2)$,max $(F_{A_1L}(x_1), F_{A_1L}(y_1)))$ = max(max $(F_{A_1L}(x),F_{A_2L}(x_2))$, max $(F_{A_1L}(y_1),F_{A_2L}(y_2)))$= max $((F_{A_1L} \circ F_{A_2L})$ $(x_1,x_2),(F_{A_1L} \circ F_{A_2L})$ $(y_1,y_2))$, (50)

$(F_{B_1U} \circ F_{B_2U})$ $((x_1, x_2)$ $(y_1, y_2))$ = max$(F_{A_2U}(x_2)$ , $F_{A_2U}(y_2)$, $F_{B_1L}(x_1y_1)) \geq$ max $(F_{A_2U}(x_2)$, $F_{A_2U}(y_2)$,max $(F_{A_1U}(x_1)$, $F_{A_1U}(y_1)))$ = max(max $(F_{A_1U}(x), F_{A_2U}(x_2))$, max $(F_{A_1U}(y_1), F_{A_2U}(y_2)))$ = max $((F_{A_1U} \circ F_{A_2U})$ $(x_1, x_2), (F_{A_1U} \circ F_{A_2U})$ $(y_1, y_2))$. (51)

This completes the proof.

**Definition 19**

The union $G_1 \cup G_2 = (A_1 \cup A_2, B_1 \cup B_2)$ of two interval valued neutrosophic graphs of the graphs $G_1^*$ and $G_2^*$ is an interval-valued neutrosophic graph of $G_1^* \cup G_2^*$.

1) $(T_{A_1L} \cup T_{A_2L})(x) = T_{A_1L}(x)$ if x ∈ $V_1$ and x ∉ $V_2$,
   $(T_{A_1L} \cup T_{A_2L})(x) = T_{A_2L}(x)$ if x ∉ $V_1$ and x ∈ $V_2$,
   $(T_{A_1L} \cup T_{A_2L})(x) = \max(T_{A_1L}(x), T_{A_2L}(x))$ if x ∈ $V_1 \cap V_2$,
   (52)
2) $(T_{A_1U} \cup T_{A_2U})(x) = T_{A_1U}(x)$ if x ∈ $V_1$ and x ∉ $V_2$,
   $(T_{A_1U} \cup T_{A_2U})(x) = T_{A_2U}(x)$ if x ∉ $V_1$ and x ∈ $V_2$,
   $(T_{A_1U} \cup T_{A_2U})(x) = \max(T_{A_1U}(x), T_{A_2U}(x))$ if x ∈ $V_1 \cap V_2$,
   (53)
3) $(I_{A_1L} \cup I_{A_2L})(x) = I_{A_1L}(x)$ if x ∈ $V_1$ and x ∉ $V_2$,
   $(I_{A_1L} \cup I_{A_2L})(x) = I_{A_2L}(x)$ if x ∉ $V_1$ and x ∈ $V_2$,
   $(I_{A_1L} \cup I_{A_2L})(x) = \min(I_{A_1L}(x), I_{A_2L}(x))$ if x ∈ $V_1 \cap V_2$, (54)
4) $(I_{A_1U} \cup I_{A_2U})(x) = I_{A_1U}(x)$ if x ∈ $V_1$ and x ∉ $V_2$,
   $(I_{A_1U} \cup I_{A_2U})(x) = I_{A_2U}(x)$ if x ∉ $V_1$ and x ∈ $V_2$,
   $(I_{A_1U} \cup I_{A_2U})(x) = \min(I_{A_1U}(x), I_{A_2U}(x))$ if x ∈ $V_1 \cap V_2$,
   (55)
5) $(F_{A_1L} \cup F_{A_2L})(x) = F_{A_1L}(x)$ if x ∈ $V_1$ and x ∉ $V_2$,
   $(N_{A_1L} \cup N_{A_2L})(x) = F_{A_2L}(x)$ if x ∉ $V_1$ and x ∈ $V_2$,
   $(N_{A_1L} \cup N_{A_2L})(x) = \min(F_{A_1L}(x), F_{A_2L}(x))$ if x ∈ $V_1 \cap V_2$,
   (56)
6) $(F_{A_1U} \cup F_{A_2U})(x) = F_{A_1U}(x)$ if x ∈ $V_1$ and x ∉ $V_2$,
   $(F_{A_1U} \cup F_{A_2U})(x) = F_{A_2U}(x)$ if x ∉ $V_1$ and x ∈ $V_2$,
   $(F_{A_1U} \cup F_{A_2U})(x) = \min(F_{A_1U}(x), F_{A_2U}(x))$ if x ∈ $V_1 \cap V_2$,
   (57)
7) $(T_{B_1L} \cup T_{B_2L})(xy) = T_{B_1L}(xy)$ if xy ∈ $E_1$ and xy ∉ $E_2$,
   $(T_{B_1L} \cup T_{B_2L})(xy) = T_{B_2L}(xy)$ if xy ∉ $E_1$ and xy ∈ $E_2$,
   $(T_{B_1L} \cup T_{B_2L})(xy) = \max(T_{B_1L}(xy), T_{B_2L}(xy))$ if xy ∈ $E_1 \cap E_2$, (58)
8) $(T_{B_{1U}} \cup T_{B_2U})(xy) = T_{B_1U}(xy)$ if xy ∈ $E_1$ and xy ∉ $E_2$,
   $(T_{B_1U} \cup T_{B_2U})(xy) = T_{B_2U}(xy)$ if xy ∉ $E_1$ and xy ∈ $E_2$,
   $(T_{B_1U} \cup T_{B_2U})(xy) = \max(T_{B_1U}(xy), T_{B_2U}(xy))$ if xy ∈ $E_1 \cap E_2$,
   (59)
9) $(I_{B_1L} \cup I_{B_2L})(xy) = I_{B_1L}(xy)$ if xy ∈ $E_1$ and xy ∉ $E_2$,
   $(I_{B_1L} \cup M_{B_2L})(xy) = I_{B_2L}(xy)$ if xy ∉ $E_1$ and xy ∈ $E_2$,

$(I_{B_1L} \cup I_{B_2L})\,(xy) = \min\,(I_{B_1L}(xy), I_{B_2L}(xy))$ if $xy \in E_1 \cap E_2$, (60)

10) $(I_{B_1U} \cup I_{B_2U})\,(xy) = I_{B_1U}(xy)$ if $xy \in E_1$ and $xy \notin E_2$,
$(I_{B_1U} \cup I_{B_2U})\,(xy) = I_{B_2U}(xy)$ if $xy \notin E_1$ and $xy \in E_2$,
$(I_{B_1U} \cup I_{B_2U})\,(xy) = \min\,(I_{B_1U}(xy), I_{B_2U}(xy))$ if $xy \in E_1 \cap E_2$, (61)

11) $(F_{B_1L} \cup F_{B_2L})\,(xy) = F_{B_1L}(xy)$ if $xy \in E_1$ and $xy \notin E_2$,
$(F_{B_1L} \cup F_{B_2L})\,(xy) = F_{B_2L}(xy)$ if $xy \notin E_1$ and $xy \in E_2$,
$(F_{B_1L} \cup F_{B_2L})\,(xy) = \min\,(F_{B_1L}(xy), F_{B_2L}(xy))$ if $xy \in E_1 \cap E_2$, (62)

12) $(F_{B_1U} \cup F_{B_2U})\,(xy) = F_{B_1U}(xy)$ if $xy \in E_1$ and $xy \notin E_2$,
$(F_{B_1U} \cup F_{B_2U})\,(xy) = F_{B_2U}(xy)$ if $xy \notin E_1$ and $xy \in E_2$,
$(F_{B_1U} \cup F_{B_2U})\,(xy) = \min\,(F_{B_1U}(xy), F_{B_2U}(xy))$ if $xy \in E_1 \cap E_2$. (63)

**Proposition 3**

Let $G_1$ and $G_2$ are two interval valued neutrosophic graphs, then $G_1 \cup G_2$ is an interval valued neutrosophic graph.

P**roof**. Verifying only conditions for $B_1 \circ B_2$, because conditions for $A_1 \circ A_2$ are obvious.

Let x y $\in E_1 \cap E_2$.

Then:

$(T_{B_1L} \cup T_{B_2L})(xy) = \max(T_{B_1L}(xy), T_{B_2L}(xy)) \leq \max(\min(T_{A_1L}(x), T_{A_1L}(y)), \min(T_{A_2L}(x), T_{A_2L}(y))) = \min(\max(T_{A_1L}(x), T_{A_2L}(x)), \max(T_{A_1L}(y), T_{A_2L}(y))) = \min((T_{A_1L} \cup T_{A_2L})\,(x), (T_{A_1L} \cup T_{A_2L})\,(y)))$; (64)

$(T_{B_1U} \cup T_{B_2U})(xy) = \max(T_{B_1U}(xy), T_{B_2U}(xy)) \leq \max(\min(T_{A_1U}(x), T_{A_1U}(y)), \min(T_{A_2U}(x), T_{A_2U}(y))) = \min(\max(T_{A_1U}(x), T_{A_2U}(x)), \max(T_{A_1U}(y), T_{A_2U}(y))) = \min((T_{A_1U} \cup T_{A_2U})\,(x), (T_{A_1U} \cup T_{A_2U})\,(y)))$; (65)

$(I_{B_1L} \cup I_{B_2L})(xy) = \min(I_{B_1L}(xy), I_{B_2L}(xy)) \geq \min(\max(I_{A_1L}(x), I_{A_1L}(y)), \max(I_{A_2L}(x), I_{A_2L}(y))) = \min(\min(I_{A_1L}(x), I_{A_2L}(x)), \min(I_{A_1L}(y), I_{A_2L}(y))) = \max((I_{A_1L} \cup I_{A_2L})\,(x), (I_{A_1L} \cup I_{A_2L})\,(y)))$; (66)

$(I_{B_1U} \cup I_{B_2U})(xy) = \min(I_{B_1U}(xy), I_{B_2U}(xy)) \geq \min(\max(I_{A_1U}(x), I_{A_1U}(y)), \max(I_{A_2U}(x), I_{A_2U}(y))) = \max(\min(I_{A_1U}(x), I_{A_2U}(x)), \min(I_{A_1U}(y), I_{A_2U}(y))) = \max((I_{A_1U} \cup I_{A_2U})\,(x), (I_{A_1U} \cup I_{A_2U})\,(y)))$; (67)

$(F_{B_1L} \cup F_{B_2L})(xy) = \min(F_{B_1L}(xy), F_{B_2L}(xy)) \geq \min(\max(F_{A_1L}(x), F_{A_1L}(y)), \max(F_{A_2L}(x), F_{A_2L}(y))) = \min(\min(F_{A_1L}(x), F_{A_2L}(x)), \min(F_{A_1L}(y), F_{A_2L}(y))) = \max((F_{A_1L} \cup F_{A_2L})\,(x), (F_{A_1L} \cup F_{A_2L})\,(y)))$; (68)

$(F_{B_1U} \cup F_{B_2U})(xy) = \min(F_{B_1U}(xy), F_{B_2U}(xy)) \geq \min(\max(F_{A_1U}(x), F_{A_1U}(y)), \max(F_{A_2U}(x), F_{A_2U}(y))) = \max(\min(F_{A_1U}(x), F_{A_2U}(x)), \min(F_{A_1U}(y), F_{A_2U}(y))) = \max((F_{A_1U} \cup F_{A_2U})\,(x), (F_{A_1U} \cup F_{A_2U})\,(y)))$. (69)

This completes the proof.

**Example 5**

Let $G_1^*= (V_1, E_1)$ and $G_2^*= (V_2, E_2)$ be two graphs such that $V_1 =\{v_1, v_2, v_3, v_4, v_5\}$. $V_2 =\{v_1, v_2, v_3, v_4\}$, $E_1 =\{v_1v_2, v_1v_5, v_2v_3, v_5v_3, v_5v_4, v_4v_3\}$ and $E_2 =\{v_1v_2, v_2v_3, v_2v_4, v_3v_{34}, v_4v_1\}$. Consider two interval valued neutrosophic graphs $G_1= (A_1, B_1)$ and $G_2= (A_2, B_2)$.

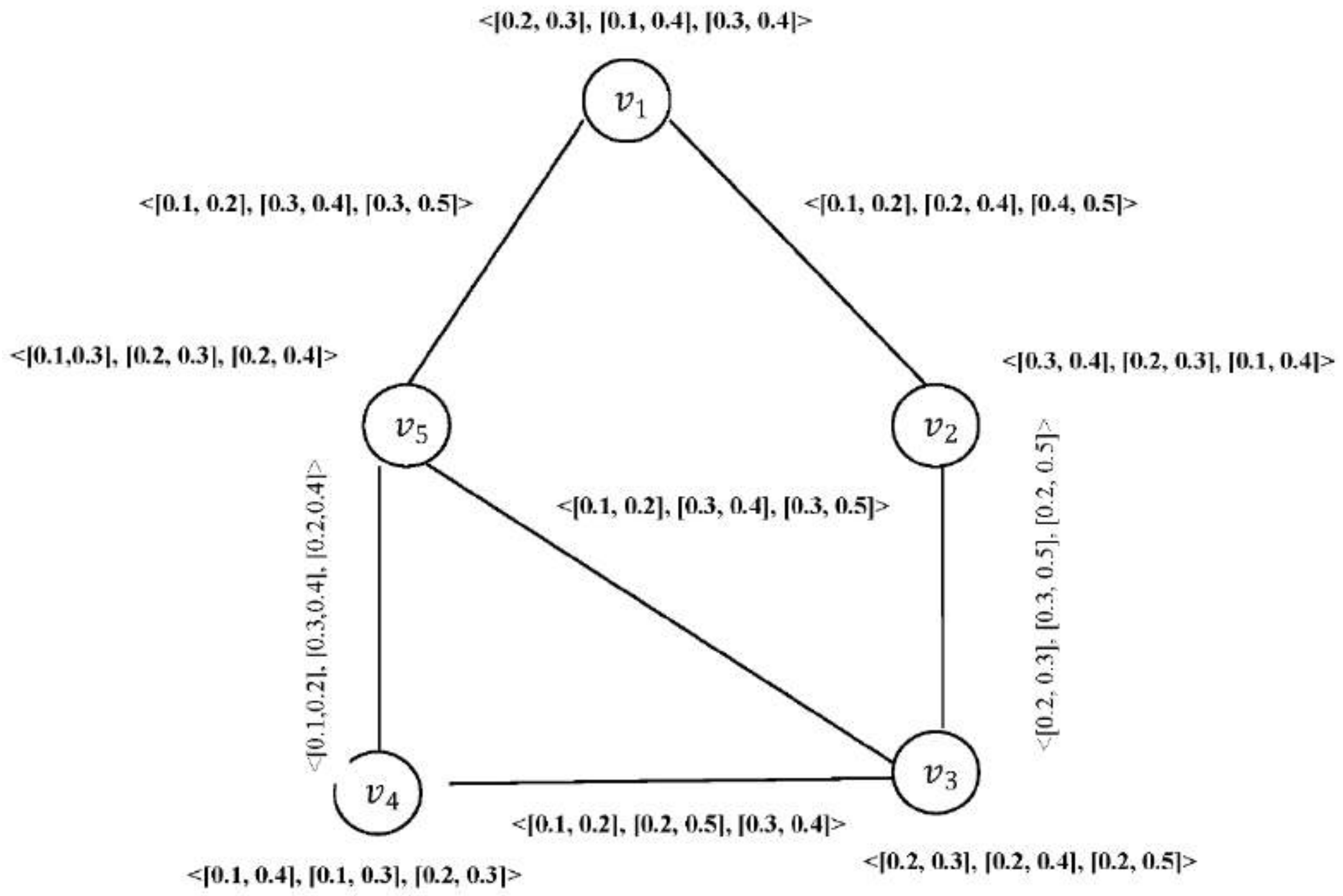

***Figure 13: Interval valued neutrosophic graph G₁***

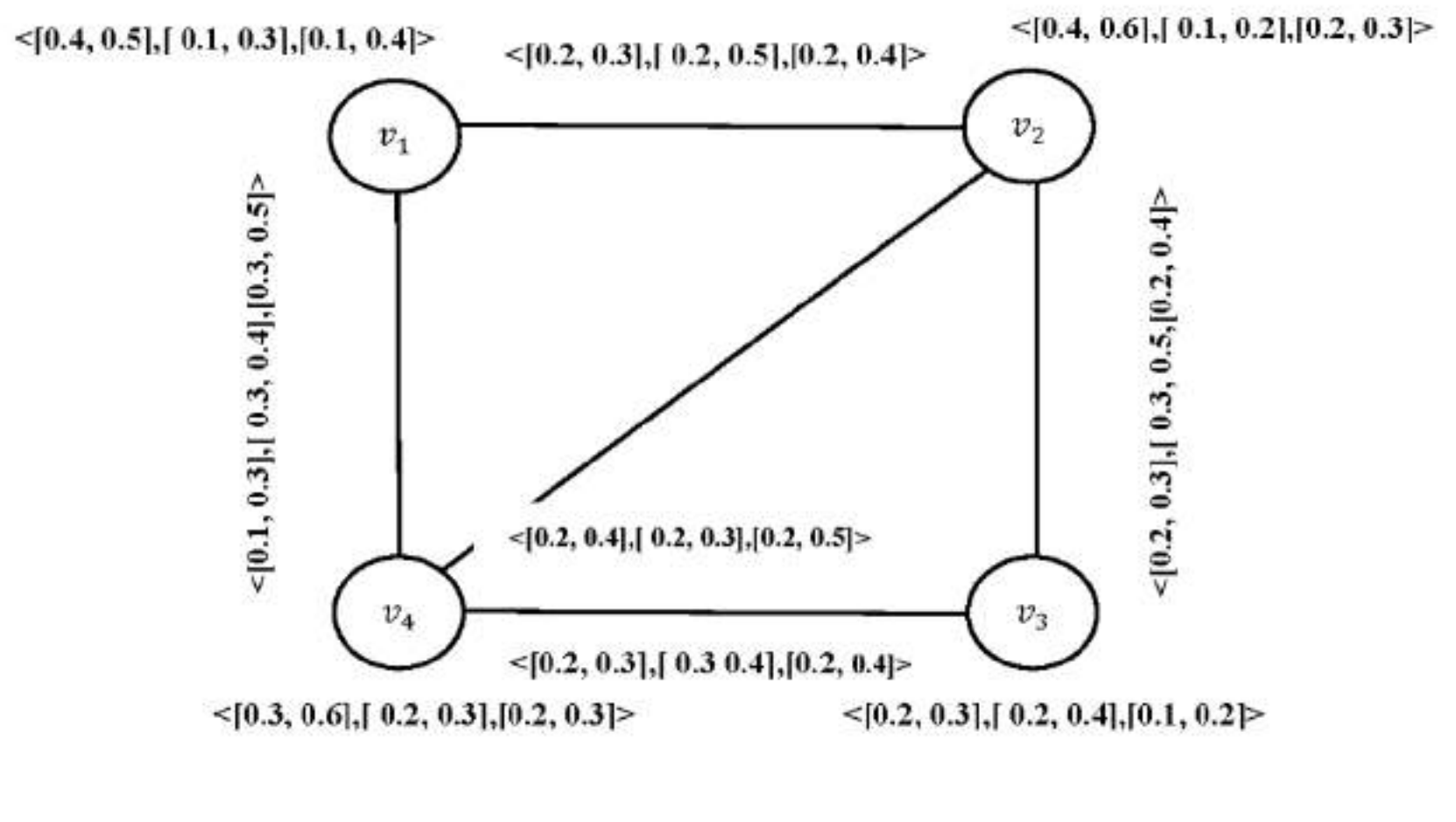

***Figure 14: Interval valued neutrosophic graph G₁***

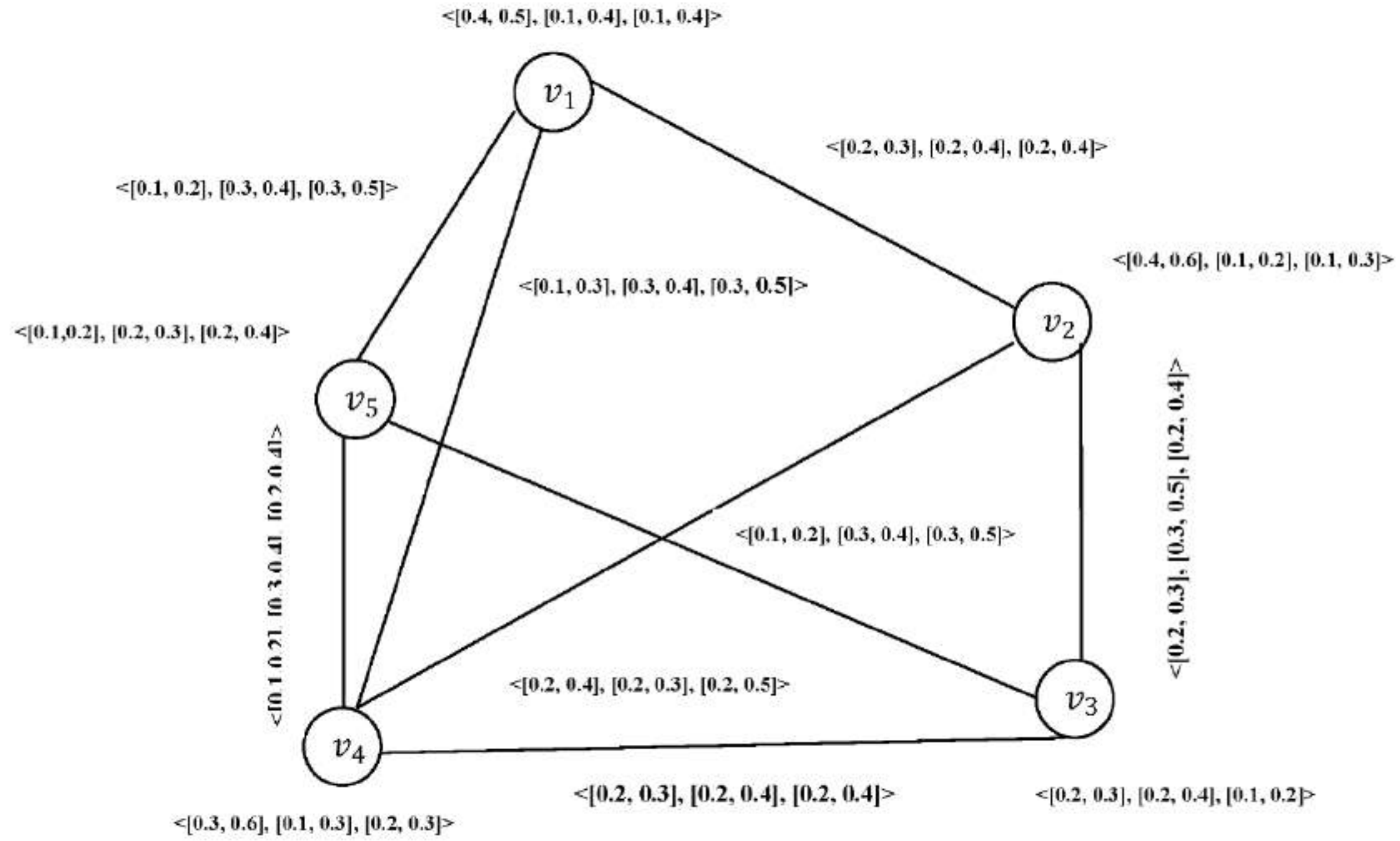

***Figure 15: Interval valued neutrosophic graph $G_1 \cup G_2$***

### Definition 20

The join of $G_1 + G_2 = (A_1 + A_2\,,\, B_1 + B_2)$ interval valued neutrosophic graphs $G_1$ and $G_2$ of the graphs $G_1^*$ and $G_2^*$ is defined as follows:

1) $$(T_{A_1L} + T_{A_2L})(x) = \begin{cases} (T_{A_1L} \cup T_{A_2L})(x) & \text{if } x \in V_1 \cup V_2 \\ T_{A_1L}(x) & \text{if } x \in V_1 \\ T_{A_2L}(x) & \text{if } x \in V_2 \end{cases} \qquad (70)$$

$$(T_{A_1U} + T_{A_2U})(x) = \begin{cases} (T_{A_1U} \cup T_{A_2U})(x) & \text{if } x \in V_1 \cup V_2 \\ T_{A_1U}(x) & \text{if } x \in V_1 \\ T_{A_2U}(x) & \text{if } x \in V_2 \end{cases}$$

$$(I_{A_1L} + I_{A_2L})\,(x) = \begin{cases} (I_{A_1L} \cap I_{A_2L})(x) & \text{if } x \in V_1 \cup V_2 \\ I_{A_1L}(x) & \text{if } x \in V_1 \\ I_{A_2L}(x) & \text{if } x \in V_2 \end{cases}$$

$$(I_{A_1U} + I_{A_2U})\,(x) = \begin{cases} (I_{A_1U} \cap I_{A_2U})(x) & \text{if } x \in V_1 \cup V_2 \\ I_{A_1U}(x) & \text{if } x \in V_1 \\ I_{A_2U}(x) & \text{if } x \in V_2 \end{cases}$$

$$(F_{A_1L} + F_{A_2L})\,(x) = \begin{cases} (F_{A_1L} \cap F_{A_2L})(x) & \text{if } x \in V_1 \cup V_2 \\ F_{A_1L}(x) & \text{if } x \in V_1 \\ F_{A_2L}(x) & \text{if } x \in V_2 \end{cases}$$

$$(F_{A_1U} + F_{A_2U})\,(x) = \begin{cases} (F_{A_1U} \cap F_{A_2U})(x) & \text{if } x \in V_1 \cup V_2 \\ F_{A_1U}(x) & \text{if } x \in V_1 \\ F_{A_2U}(x) & \text{if } x \in V_2 \end{cases}$$

2) $$(T_{B_1L} + T_{B_2L})\,(xy) = \begin{cases} (T_{B_1L} \cup T_{B_2L})(xy) & \text{if } xy \in E_1 \cup E_2 \\ T_{B_1L}(xy) & \text{if } xy \in E_1 \\ T_{B_2L}(xy) & \text{if } xy \in E_2 \end{cases} \qquad (71)$$

$$(T_{B_1U} + T_{B_2U})\,(xy) = \begin{cases} (T_{B_1U} \cup T_{B_2U})(xy) & \text{if } xy \in E_1 \cup E_2 \\ T_{B_1U}(xy) & \text{if } xy \in E_1 \\ T_{B_2U}(xy) & \text{if } xy \in E_2 \end{cases}$$

$$(I_{B_1L} + I_{B_2L})\,(xy) = \begin{cases} (I_{B_1L} \cap I_{B_2L})(xy) & \text{if } xy \in E_1 \cup E_2 \\ I_{B_1L}(xy) & \text{if } xy \in E_1 \\ I_{B_2L}(xy) & \text{if } xy \in E_2 \end{cases}$$

$$(I_{B_1U} + I_{B_2U})\,(xy) = \begin{cases} (I_{B_1U} \cap I_{B_2U})(xy) & \text{if } xy \in E_1 \cup E_2 \\ I_{B_1U}(xy) & \text{if } xy \in E_1 \\ I_{B_2U}(xy) & \text{if } xy \in E_2 \end{cases}$$

$$(F_{B_1L} + F_{B_2L})\,(xy) = \begin{cases} (F_{B_1L} \cap F_{B_2L})(xy) & \text{if } xy \in E_1 \cup E_2 \\ F_{B_1L}(xy) & \text{if } xy \in E_1 \\ F_{B_2L}(xy) & \text{if } xy \in E_2 \end{cases}$$

$$(F_{B_1U} + F_{B_2U})\,(x\,y) = \begin{cases} (F_{B_1U} \cap F_{B_2U})(xy) & \text{if } xy \in E_1 \cup E_2 \\ F_{B_1U}(xy) & \text{if } xy \in E_1 \\ F_{B_2U}(xy) & \text{if } xy \in E_2 \end{cases}$$

3) $(T_{B_1L} + T_{B_2L})\,(x\,y) = \min\,(T_{B_1L}(x), T_{B_2L}(x))$ (72)

$(T_{B_1U} + T_{B_2U})\,(xy) = \min\,(T_{B_1U}(x), T_{B_2U}(x))$

$(I_{B_1L} + I_{B_2L})\,(xy) = \max\,(I_{B_1L}(x), I_{B_2L}(x))$

$(I_{B_1U} + I_{B_2U})\,(x\,y) = \max\,(I_{B_1U}(x), I_{B_2U}(x)$

$(F_{B_1L} + F_{B_2L})\,(xy) = \max\,(F_{B_1L}(x), F_{B_2L}(x))$

$(F_{B_1U} + F_{B_2U})\,(x\,y) = \max\,(F_{B_1U}(x), F_{B_2U}(x)) if xy \in E'$,

where $E'$ is the set of all edges joining the nodes of $V_1$ and $V_2$, assuming $V_1 \cap V_2 = \emptyset$.

**Example 6**

Let $G_1^*= (V_1, E_1)$ and $G_2^*= (V_2, E_2)$ be two graphs such that $V_1 = \{u_1, u_2, u_3\}$, $V_2 = \{v_1, v_2, v_3\}$, $E_1 = \{u_1u_2,\, u_2u_3\}$ and $E_2 = \{v_1v_2,\, v_2v_3\}$. Consider two interval valued neutrosophic graphs $G_1 = (A_1, B_1)$ and $G_2 = (A_2, B_2)$.

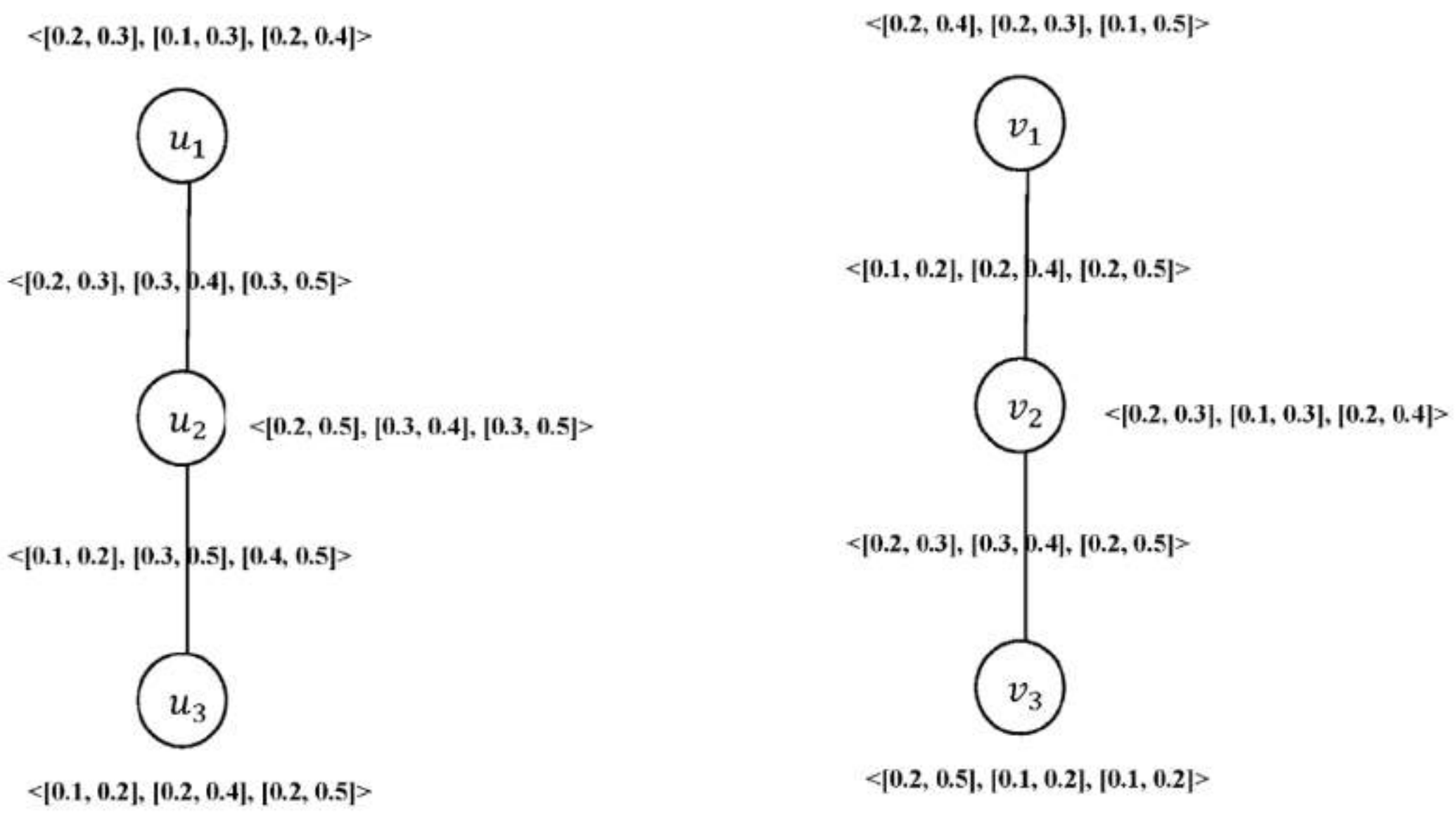

***Figure 16: Interval valued neutrosophic graph ofG*** $_1$ ***and G*** $_2$

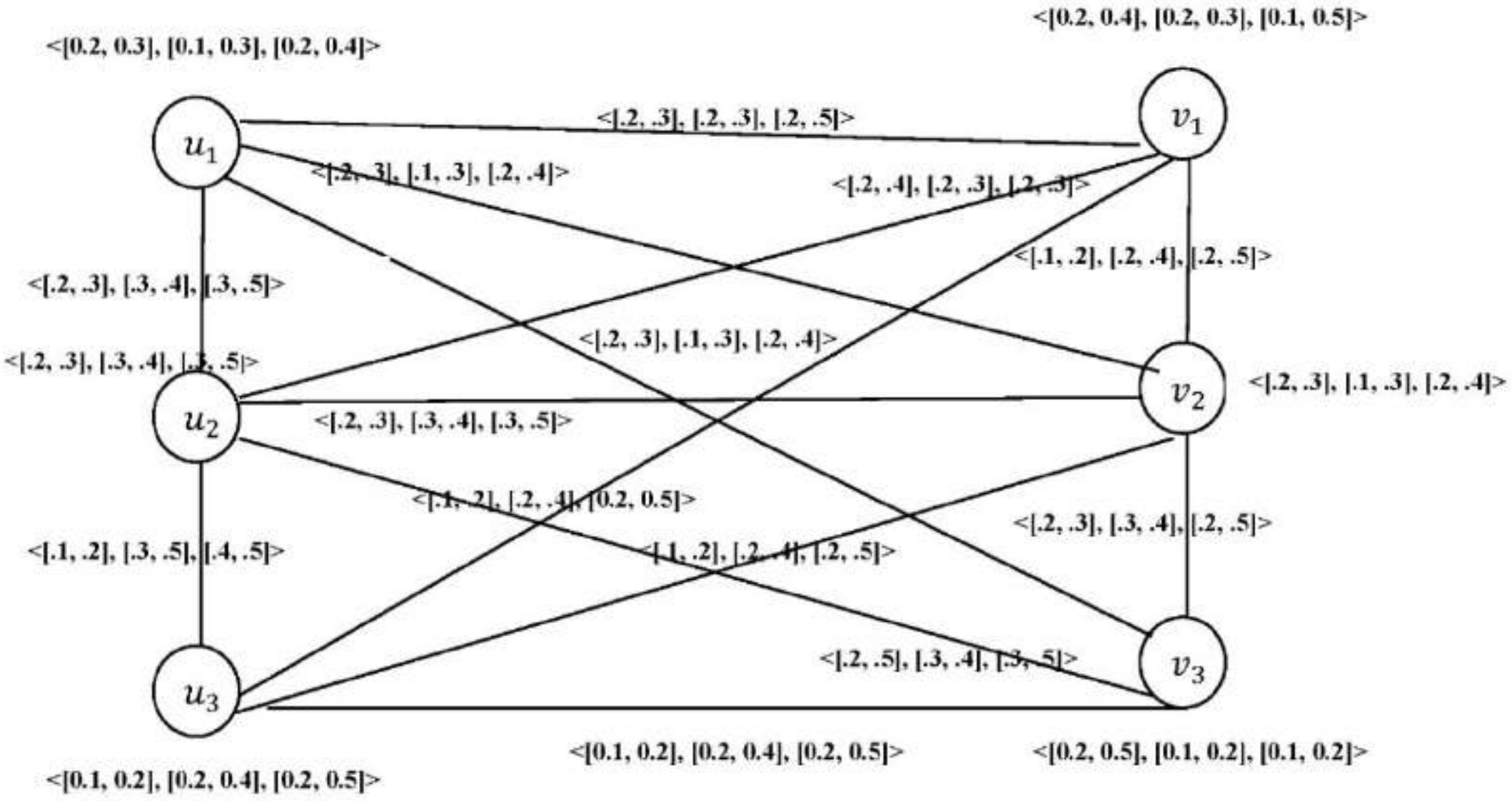

***Figure 17: Interval valued neutrosophic graph of G*** $_1$ ***+ G*** $_2$

## 5. Conclusion

The interval valued neutrosophic models give more precision, flexibility and compatibility to the system as compared to the classical, fuzzy, intuitionistic fuzzy and neutrosophic models. In this paper, the authors introduced some operations: Cartesian product, composition, union and join on interval valued neutrosophic graphs, and investigated some of their properties. In the future, the authors plan to study others operations, such as: tensor product and normal product of two interval valued neutrosophic graphs.

## References

1. Ansari, A. Q., Biswas,R., &Aggarwal S.(2013a).Neutrosophic classifier: An extension of fuzzy classifier. Elsevier, Applied Soft Computing, 13, 563-573, http://dx.doi.org/10.1016/ j.asoc.2012.08.002.
2. Ansari, A. Q., Biswas,R., &Aggarwal, S.(2013b). (Presentation) Neutrosophication of Fuzzy Models, In: Proceedings of the IEEE Workshop On Computational Intelligence: Theories, Applications and Future Directions (hostedby IIT Kanpur), IEEE, 14th July 2013.
3. Ansari, A. Q., Biswas,R.,&Aggarwal, S. (2013c).Extension to fuzzy logic representation: Moving towards neutrosophic logic - A new laboratory rat. In: Proceedings of the IEEE Workshop On Fuzzy Systems (FUZZ), IEEE: 1 –8.DOI:10.1109/FUZZ-IEEE.2013.6622412.
4. Atanassov, K. (1999).Intuitionistic fuzzy sets: theory and applications, Physica, New York.
5. Atanassov, K. (1986). Intuitionistic fuzzy sets, Fuzzy Sets and Systems, vol. 20: 87-96.
6. Atanassov, K., Gargov, G. (1989). Interval valued intuitionistic fuzzy sets, Fuzzy Sets and Systems, vol.31:343-349.
7. Antonios, K., Stefanos, S., Lazaros, I., &Elias, P. (2014).Fuzzy graphs: algebraic structure and syntactic recognition, Artificial Intelligence Review, vol 42, Issue 3:479-490.
8. Aggarwal, S., Biswas, R., &Ansari, A. Q.(2010). Neutrosophic modeling and control, Computer and Communication Technology (ICCCT), International Conference:718–723 DOI:10.1109/ICCCT.2010.5640435.
9. Broumi, S., Talea, M.,& Smarandache, F. (2016a) Single Valued Neutrosophic Graphs: Degree, Order and Size. IEEE World Congress on Computational Intelligence, 8 pages, accepted
10. Broumi, S., Talea, M., Bakali, A. and Smarandache, F. (2016b). Single Valued Neutrosophic Graphs, Journal of New Theory, N 10:86-101.
11. Broumi, S., Talea, M., Bakali, A and Smarandache, F. Interval Valued Neutrosophic Graphs. In press.
12. Broumi, S., Talea, M., Bakali, A., &Smarandache, F. (2016c). On Bipolar Single Valued Neutrosophic Graphs, Journal of New Theory, no. 11: 84-102.
13. Broumi, S., Smarandache, F., Talea, M. & Bakali, A. (2016). An Introduction to Bipolar Single Valued Neutrosophic Graph Theory. Applied Mechanics and Materials, vol.841:184-191, doi:10.4028/www.scienti_c.net/AMM.841.184
14. Broumi, S., Smarandache, F. (2014). New distance and similarity measures of interval neutrosophic sets.Information Fusion (FUSION), IEEE 17th International Conference:1 – 7.
15. Bhattacharya, P. (1987). Some remarks on fuzzy graphs, Pattern Recognition Letters, 6: 297-302.
16. Deli, I., Mumtaz, A. &Smarandache, F.(2015).Bipolar neutrosophic sets and their application based on multi-criteria decision making problems, Advanced Mechatronic Systems (ICAMechS), 2015 International Conference:249 – 254, DOI: 10.1109/ICAMechS.2015.7287068.
17. Garg, G., Bhutani, K., Kumar,M., & Aggarwal, S. (2015).Hybrid model for medical diagnosis using Neutrosophic Cognitive Maps with Genetic Algorithms, FUZZ-IEEE 2015 (IEEE International conference on fuzzy systems).
18. Hai-Long, Y., She, G., Yanhonge, & Xiuwu, L. (2015). On single valued neutrosophic relations, Journal of Intelligent & Fuzzy Systems:1-12.
19. Liu, P., &Shi, L. (2015).The generalized hybrid weighted average operator based on interval neutrosophic hesitant set and its application to multiple attribute decision making, Neural Computing and Applications, 26 (2):457-471.
20. Mishra S. N. &Pal, A. (2013). Product of Interval Valued Intuitionistic fuzzy graph, Annals of Pure and Applied Mathematics, Vol. 5, No.1:37-46.
21. Nagoor, G.A., & Shajitha Begum, S. (2010). Degree, Order and Size in Intuitionistic Fuzzy Graphs, International Journal of Algorithms, Computing and Mathematics, (3)3:11-16.
22. Nagoor, G.A., & Latha, S.R. (2012). On Irregular Fuzzy Graphs. Applied Mathematical Sciences, Vol.6, no.11:517-523.

23. Nagoor, G. A., & Basheer Ahamed, M. (2003). Order and Size in Fuzzy Graphs, Bulletin of Pure and Applied Sciences, Vol 22E (No.1):145-148.
24. Parvathi, R., Karunambigai, M. G. (2006).Intuitionistic Fuzzy Graphs. Computational Intelligence, Theory and applications, International Conference in Germany,18 -20.
25. Shannon, A., & Atanassov, K. (1994). A first step of the intuitionistic fuzzy graph, Proceeding of the first Workshop on fuzzy based Expert Systems, Sofia, sept. 28-30: 59-61.
26. Şahin, R. (2015). Cross-entropy measure on interval neutrosophic sets and its applications in multicriteria decision making, Neural Computing and Applications:1-11.
27. Smarandache, F. (2015a).Symbolic Neutrosophic Theory, Brussels, Belgium: Europanova, 195p.
28. Smarandache, F. (2015b). Refined Literal Indeterminacy and the Multiplication Law of Sub-Indeterminacies, Neutrosophic Sets and Systems, Vol. 9:58-63.
29. Smarandache, F. (2015c). Types of Neutrosophic Graphs and Neutrosophic Algebraic Structures together with their Applications in Technology. Seminar, Universitatea Transilvania din Brasov, Facultatea de Design de Produs si Mediu, Brasov, Romania.
30. Smarandache, F. (2006). Neutrosophic set - a generalization of the intuitionistic fuzzy set.Granular Computing, 2006 IEEE International Conference:38–42. DOI: 10.1109/GRC.2006.1635754.
31. Smarandache, F. (2011). A geometric interpretation of the neutrosophic set - A generalization of the intuitionistic fuzzy set.Granular Computing (GrC), 2011 IEEE International Conference: 602 – 606.
32. Turksen, I. (1986). Interval valued fuzzy sets based on normal forms.Fuzzy Sets and Systems, vol. 20:191-210.
33. Vasantha Kandasamy, W. B., Smarandache, F.(2013).Fuzzy Cognitive Maps and Neutrosophic Cognitive Maps, Kindle Edition.
34. Vasantha Kandasamy, W. B., Ilanthenral. K and Smarandache, F.(2015).Neutrosophic Graphs: A New Dimension to Graph Theory. Kindle Edition.
35. Vasantha Kandasamy, W.B., Smarandache,F. (2004). Analysis of social aspects of migrant laborers living with HIV/AIDS using Fuzzy Theory and Neutrosophic Cognitive Maps, Phoenix, AZ, USA: Xiquan.
36. Wang, H., Zhang, Y., & Sunderraman, R. (2005a).Truth-value based interval neutrosophic sets. Granular Computing, 2005 IEEE International Conference, Vol.1:274–277. DOI: 10.1109/GRC.2005.1547284.
37. Wang, H., Smarandache, F., Zhang,Y., &Sunderraman,R.(2010). Single valued Neutrosophic Sets, Multispace and Multistructure, 4:410-413.
38. Wang, H., Smarandache, F., Zhang,Y.Q and Sunderraman, R. (2005b).Interval Neutrosophic Sets and Logic: Theory and Applications in Computing. Phoenix, AZ, USA: Hexis.
39. Ye, J. (2014a). Some aggregation operators of interval neutrosophic linguistic numbers for multiple attribute decision making, Journal of Intelligent & Fuzzy Systems (27):2231-2241.
40. Ye, J. (2014b). Vector similarity measures of simplified neutrosophic sets and their application in multicriteria decision making, International Journal of Fuzzy Systems. Vol. 16, No. 2: 204-211.
41. Ye, J. (2014c). Similarity measures between interval neutrosophic sets and their applications in Multi-criteria decision-making. Journal of Intelligent and Fuzzy Systems, 26:165-172.
42. Zimmermann, H.J. (1985). Fuzzy Set Theory and its Applications. Boston, MA, USA: Kluwer-Nijhoff.
43. Zhang, H., Wang J, Chen, X. (2015a). An outranking approach for multi-criteria decision-making problems with interval-valued neutrosophic sets. Neural Computing and Applications, 1-13.
44. Zhang, H.Y., Ji, P., Wang, Q.J, Chen, H. X.(2015b). An Improved Weighted Correlation Coefficient Based on Integrated Weight for Interval Neutrosophic Sets and its Application in Multi-Criteria Decision-making Problems, International Journal of Computational Intelligence Systems, vol 8, Issue 6.
45. Zhang, H.Y., Wang, J.Q. & Chen, X.H. (2014). Interval neutrosophic sets and their application in multicriteria decision making problems. The Scientific World Journal,DOI:10.1155/2014/ 645953.
46. Zadeh, L. (1965).Fuzzy sets, Information and Control, 8:338-353.

# MEDICAL DIAGNOSIS

**DEEPIKA KOUNDAL[1A], SAVITA GUPTA[2B], SUKHWINDER SINGH[2C]**

*[a]koundal@gmail.com, [b]savita2k8@yahoo.com, [c]sukhdalip@yahoo.com*
*[1]Chitkara University, Rajpura, Punjab*
*[2]University Institute of Engineering & Technology, Panjab University, Chandigarh, India*

# Applications of Neutrosophic Sets in Medical Image Denoising and Segmentation

## Abstract

In medical science, diagnosis and prognosis is one of the most difficult and challenging task because of restricted subjectivity of the experts and presence of fuzziness in medical images. In observing the severity of several diseases, different professional experts may result in wrong diagnosis. In order to perform diagnosis intuitively in the medical images, different image processing methods have been explored in terms of neutrosophic theory to interpret the inherent uncertainty, ambiguity and vagueness. This paper demonstrates the use of neutrosophic theory in medical image denoising and segmentation where the performance is observed to be much better.

## 1. Introduction

Generally medical images are consisted of fuzziness and imprecision information, therefore segmentation, feature extraction and classification are difficult to perform [1]. Since fuzzy sets are widely used for processing fuzziness and uncertainty in a wide range of fields such as control science and image processing [2]. But the limitation of this method is that it does not consider the spatial context of the pixels due to noise and artifacts [3]. The generalization of fuzzy set in form of neutrosophic set is becoming more popular in image processing tasks to overcome the limitations of fuzzy based approaches. The concept of Neutrosophy is introduced by Smarandache [4]. Neutrosophy is the foundation of neutrosophic probability, neutrosophic statistics, neutrosophic logic and neutrosophic set [4]. *Neutrosophic set* generalizes the concept of the classic set, fuzzy set, interval valued fuzzy set [5], intuitionistic fuzzy set [6], paraconsistent set, paradoxist set, tautological set, dialetheist set [3]. *Neutrosophy theory* takes into account every theory, concept, or entity *<A>* in relation to its opposite, *<Anti-A>* and *<Non-A>*. The neutralities <Neut-A> which is not *A,* and that which is neither *<A>* nor *<Anti-A>* are referred to as *<Non-A>*. In *neutrosophic logic*, three neutrosophic components: *T, I, F* are defined to estimate the truth membership degree, the false membership degree, and the indeterminacy membership degree (neither true nor false) in <*A*>. Unlike fuzzy logic, neutrosophic logic introduces the extra domain *I* which provides a more efficient way to handle higher degrees of uncertainty that is very difficult for fuzzy logic to be handled [7]. The major difference between a Neutrosophic Set (NS) and a Fuzzy Set (FS) is that there is no limit on the sum *m* in a NS, while in a FS *m (m=t+f)* must be equal to 1 [8]. The Neutrosophic image domain is shown in Fig. 1.

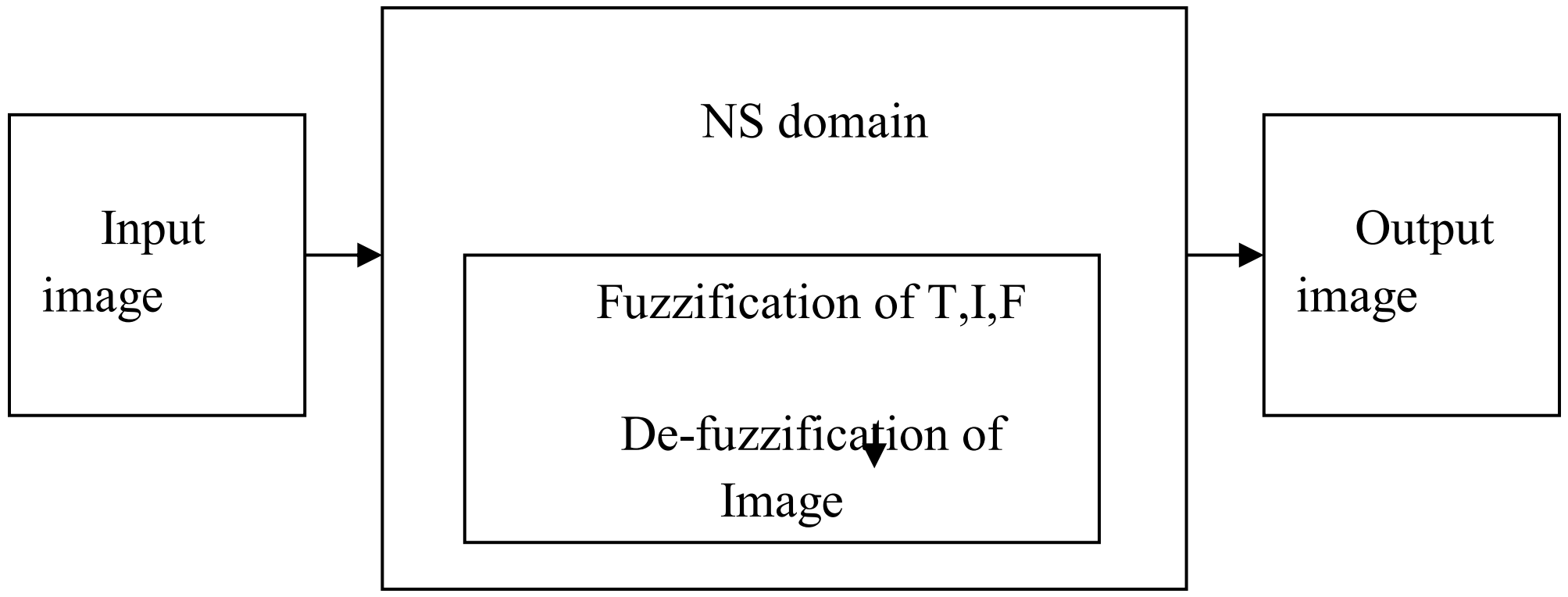

Figure 1. *Neutrosophic Image Domain*

A neutrosophic image is characterized by three subsets *T, I* and *F.* A pixel *P* in neutrosophic image is described as *P(i ,j ), { T(i ,j), I(i ,j ), F(i ,j)}.* Thus, for each pixel in the neutrosophic image, the truth degree *T,* false degree *F* and indeterminacy degree *I* is required to be computed. In general, a NS is symbolized as <*T, I, F*>. In case of determining the tumor in image, tumor can be considered as <*A*>, boundaries as <*Neut-A*> and background as <*Anti-A*>. *T, I,* and *F* are the neutrosophic components to represent <*A*>, <*Neut-A*> and <*Anti-A*>, <*A*> and <*Anti-A*> contain region information, while <*Neut-A*> has boundary information [9, 10].

A pixel in the neutrosophic image can be represented as $A\{t, i, f\}$, where $t\%$ represents true (tumor), $i\%$ represents indeterminate (boundaries) and $f\%$ represents false (background), where $t \in T$, $i \in I$ and $f \in F$ [7]. In the FS, $i = 0, 0 \leq t, f \leq 100$. In the NS, $0 \leq t, i, f \leq 100$ [11,12]. An element $x(t, i, f)$ belongs to the set in the following way: it is $t$ true in the set, $i$ indeterminate in the set, and $f$ false, where $t, i,$ and $f$ are real numbers taken from the sets $T, I,$ and $F$ with no restriction on $T, I, F$ nor on their sum $m = t + i + f$. In literature, number of neutrosophic based denoising and segmentation methods are given [13, 14, 15, 21, 23, 29].

The rest of paper is organized in four sections. Section 2 describes the neutrosophic based image denoising and segmentation methods. Section 3 discusses the results of various neutrosophic domain methods. Finally, the conclusion is summarized in Section 4.

## 2. Neutrosophic Based Image Processing

### 1.1. Transformation of Image in Neutrosophic Domain

$TM$, $IM$ and $FM$ are the neutrosophic components to represent $< A >$, $< Neut - A >$ and $< Anti - A >$ respectively in neutrosophic domain. Every neutrosophic pixel can be represented as $P_{NI} = \{TM, IM, FM\}$, where $TM$ is the set of white pixels, $IM$ is the set of indeterminate pixels and $FM$ is the set of non-white pixels respectively [16, 17]. The membership functions $TM$, $IM$ and $FM$ are computed as

$$TM = \frac{\hat{f}_{ij} - \hat{f}_{min}}{\hat{f}_{max}} \tag{1}$$

where $i$ differs from 0 to *n-1, j* differs from 0 to $m$-*1*, $\hat{f}_{ij}$ is local mean obtained using window, $\hat{f}_{min}$ is minimum intensity value and $\hat{f}_{max}$ is the maximum intensity value.

$$\hat{f}_{ij} = \frac{1}{w \times w} \sum_{m=i-\frac{w}{2}}^{i+\frac{w}{2}} \sum_{n=j-\frac{w}{2}}^{j+\frac{w}{2}} f_{mn} \tag{2}$$

where $w$ is a window size, $f_{mn}$ is the noisy image and $\hat{f}_{ij}$ is local mean of pixels on $w$.

$$IM = \frac{\delta_{ij} - \delta_{min}}{\delta_{max}} \tag{3}$$

$$\delta_{ij} = abs(f_{ij} - \hat{f}_{ij}) \tag{4}$$

where $\delta_{ij}$ is absolute difference value between local mean value $\hat{f}_{ij}$ and intensity $f_{ij}$, $\delta_{max}$ is the maximum absolute difference value and $\delta_{min}$ is minimum absolute difference value. The false membership is computed as

$$FM = 1 - TM \tag{5}$$

The true subset, $TM$, is computed by normalizing the intensity values in $[0,1]$ as given in Eq.(1). In ultrasound images, pixels belonging to speckle and texture are hard to differentiate, hence, $\hat{f}_{ij}$, is calculated to ascertain the neighborhood mean of pixels in a kernel. Absolute difference is used to determine the indeterminate component and False subset, $FM$, is determined as the complement of $TM$ [18].

### 2.2. Related Work on Neutrosophic Domain image denoising

Several denoising methods based on neutrosophic set have been proposed in the literature to remove Speckle noise, Gaussian and Rician noise [19-28]. Various notions and theories based on NS are defined and applied for denoising of images. The image is converted into the NS domain and γ-median-filtering operation is used to decrease the image indeterminacy. The experiments have been carried out on natural images with various levels of noise for better image denoising [16].

A wiener filter in neutrosophic domain has been introduced in literature for removal of Rician noise. The wiener filtering operation is employed on true and false subsets for the reduction of the noise and indeterminacy. Experiments have been performed on simulated MRI from Brainweb database and clinical MR images, which are affected by Rician noise [22]. It has been found that wiener filter in neutrosophic domain is able to preserve edges with the suppression of Rician noise.

In [25, 26], LEE and KUAN filter were implemented in neutrosophic domain for the reduction of speckle noise [27]. The Neutrosophic Nonconvex Regularizer Speckle Noise Removal (NNRSNR) method based on Gamma statistics in neutrosophic domain is presented in [28]. Another method based on Nakagami distribution statistics (NTV) which is presented in [29] is further explored in neutrosophic domain. Neutrosophic Nakagami Total Variation method (NNTV) is presented to exploit the Nakagami statistics in neutrosophic domain [30].

### 2.3. Related work on Neutrosophic Domain Image Segmentation

Recently, neutrosophic based methods have been attracted attention in solving image segmentation problems due to their high performance and indeterminacy handling capability. In literature, several authors have reported number of segmentation methods based on NS [31-40].

Zhang et al. [7] introduced an algorithm, which used the region merge method in NS for the segmentation of natural images to resolve over-segmentation problem. The region merge algorithm started with initial seeds and merged the two regions until a stopping criterion is satisfied. The cluster center is selected on the basis of histogram features in fuzzy domain and the region merge criterion is defined in intensity domain based on edge value and standard deviation features.

Cheng et al. [31] introduced the NS approach with image thresholding for the segmentation of artificial and natural images with indeterminacy handling capability. However, selection of particular threshold value is a critical task as well as it ignores the spatial information and is noise sensitive. Guo et al.[32] presented the fuzzy c-means clustering in NS domain. In this method, entropy in NS domain is used to estimate the indeterminacy of image and $\alpha$-mean operation is presented to decrease the indeterminacy to make the image more homogenous. Then, image is segmented using a fuzzy c-means clustering. The membership value in the fuzzy clustering is updated as per the indeterminacy value. The experimental analysis demonstrated that the method performed better on both clean and noisy images. Another NS based image segmentation method is presented in which two new operations are defined to reduce the indeterminacy of the image. Zhang et al. [33] presented a watershed segmentation approach in NS domain. In the first phase, image is mapped to NS domain and then, neutrosophic logic and thresholding is used to get a binary image. Final segmentation result is obtained from watershed method. The Neutrosophic Watershed (NW) method has better performance on non-uniform as well as on noisy images.

Further NS is integrated with Improved Fuzzy C-Means (IFCM) for image segmentation [34]. In this, membership degree and convergence criterion of clustering are redefined accordingly. Experimental results demonstrated that the method segmented the images effectively and accurately. Another method named as Neutrosophic C-Means (NCM) clustering is introduced for uncertain data clustering, which is inspired from fuzzy c-means and the NS framework [35]. In this method, the clustering problem is derived as an objective function and is minimized with both ambiguity rejection and distance rejection. These measures are able to manage uncertainty due to imprecise definition of the clusters.

Another automatic segmentation approach is presented by Sengur et al. [36] which is the combination of texture information with color information in NS and wavelet domain. The method is used for the segmentation of natural color image using $\gamma$-$K$-means clustering. The cluster number $K$ is ascertained with cluster validity analysis. Experiments demonstrated that it segmented the natural images very effectively even if the texture and color of each region does not have homogeneous statistical characteristics. Shan *et al.* [37] presented a clustering method named as Neutrosophic L-Means (NLM) clustering for segmentation of breast ultrasound images. The method achieved the best accuracy with a fairly rapid processing speed. The main limitation of the method is that it is not able to segment multiple-lesions and failed under severe shadowing effect.

Karabatak et al.[38] has given a color image segmentation method in neutrosophic domain. Firstly, the image is transformed into NS domain by defining three membership sets. Then α-mean and β-enhancement operations were used to reduce the indeterminacy. The method suffered from over-segmentation and fixed parameters. An Iterative Neutrosophic Lung Segmentation (INLS) method has been introduced which is based on Expectation-Maximization (EM) analysis and Morphological operations (EMM) for the segmentation of ribs and lungs [39]. The results have shown that the images without or with lung diseases are segmented out more properly.

Guo *et al*. [40] has introduced a method for image based on the NS filter and level set. In First the image is transformed into NS domain by true, false and indeterminacy membership sets. Subsequently, a filter is applied for reduction of noise and level set for image segmentation. Further, a Neutrosophic Edge Detection (NSED) method is presented for edges detection with a new directional $\alpha$-mean operation [41]. The experiments have been performed using artificial and real images which demonstrated that it is able to detect edges accurately.

Recently a clustering algorithm named as Neutrosophic Evidential C-Means (NECM) with Dezert–Smarandache Theory (DSmT) is proposed for natural image segmentation [42]. The DSmT combination rule and decision has been utilized to achieve the final result. The NECM method is tested on both data clustering and image segmentation applications. Further, a Neutrosophic Similarity Score (NSS) method and level set algorithm is introduced for breast segmentation in ultrasound images [43]. First, the breast ultrasound is transformed to the NS domain via three membership subsets and then NSS is defined and used to determine the membership degree of the tumor region. Finally, the level set is employed for tumor segmentation in the NSS image. The results have shown that the method can segment the breast tissue in ultrasound images effectively and accurately.

Another neutrosophic domain segmentation method named as Spatial Neutrosophic Distance Regularizer Level Set (SNDRLS) method is presented for automated delineation of nodules in thyroid ultrasound images [44].

## 3. Experimental results and discussion

### 3.1. Results of denoising on synthetic images

This section demonstrates the qualitative and quantitative results to evaluate the effectiveness of the neutrosophic domain speckle reduction methods. In experiments, performance of the neutrosophic domain speckle reduction methods NLEE, NKUAN, NNTV and NNRSNR methods are compared with LEE, KUAN, NTV and NRSNR to study the impact of neutrosophic domain in speckle reduction and edge preservation [45]. Several quantitative measures like Signal to Noise ratio (SNR) and Edge Preservation Index (EPI) have been used for the evaluation of aforementioned methods [45, 46].

For quantitative evaluation of despeckling methods, the experiments are conducted on synthetic images, in which image is corrupted by speckle noise using speckle simulation procedure [30]. The performance of speckle reduction methods have been measured on the speckle simulated images at various noise levels ($\sigma = 0.3, 0.4, \ldots .0.9$). Table 1 represents SNR values of noisy image, KUAN, LEE, NKUAN and NLEE methods at various noise levels from $\sigma = 0.3$ to $0.9$ for synthetic image. From quantitative results, it has been noticed that the neutrosophic domain methods outperformed the spatial domain methods by achieving higher SNR values. The NKUAN outperformed the KUAN filter by gaining higher values of SNR. Similarly, NLEE has also outperformed the LEE filter by achieving higher SNR values.

Table 1

| SNR (dB) | | | | | |
|---|---|---|---|---|---|
| Noise Level | Noisy Image | KUAN | NKUAN | LEE | NLEE |
| 0.3 | 21.11 | 22.11 | 23.42 | 23.16 | 23.97 |
| 0.4 | 19.04 | 20.72 | 21.41 | 21.62 | 22.73 |
| 0.5 | 17.9 | 18.73 | 19.39 | 20.25 | 21.88 |
| 0.6 | 16.38 | 17.18 | 18.86 | 19.21 | 20.24 |
| 0.7 | 15.21 | 15.88 | 16.31 | 17.56 | 18.96 |
| 0.8 | 13.99 | 14.64 | 15.98 | 16.94 | 17.51 |
| 0.9 | 3.63 | 5.71 | 6.04 | 7.79 | 8.53 |

Figure 2 illustrates EPI of different speckle reduction techniques in which neutrosophic domain methods have high edge preservation as compared to spatial domain methods. From graphical representation as shown in Fig. 2, it is observed that NLEE is able to preserve edges better in neutrosophic domain as compared to LEE method. NLEE method has also been found to be performed better than NKUAN in terms of edge preservation. The results have been compared with spatial-domain speckle reduction filters such as LEE and KAUN. Figure 3 illustrates the results of speckle simulated synthetic image. Fig. 3(a) is the original image, and Fig. 3(b) is the image simulated with speckle noise at 0.5 noise level. Figure 3(c) and Fig. 3(d) are the despeckling results of LEE and KUAN filter, respectively. Figure 3(e) and Fig. 3(f) are the results of the proposed methods i.e. Neutrosophic KUAN (NKUAN) filter and Neutrosophic LEE (NLEE) filter.

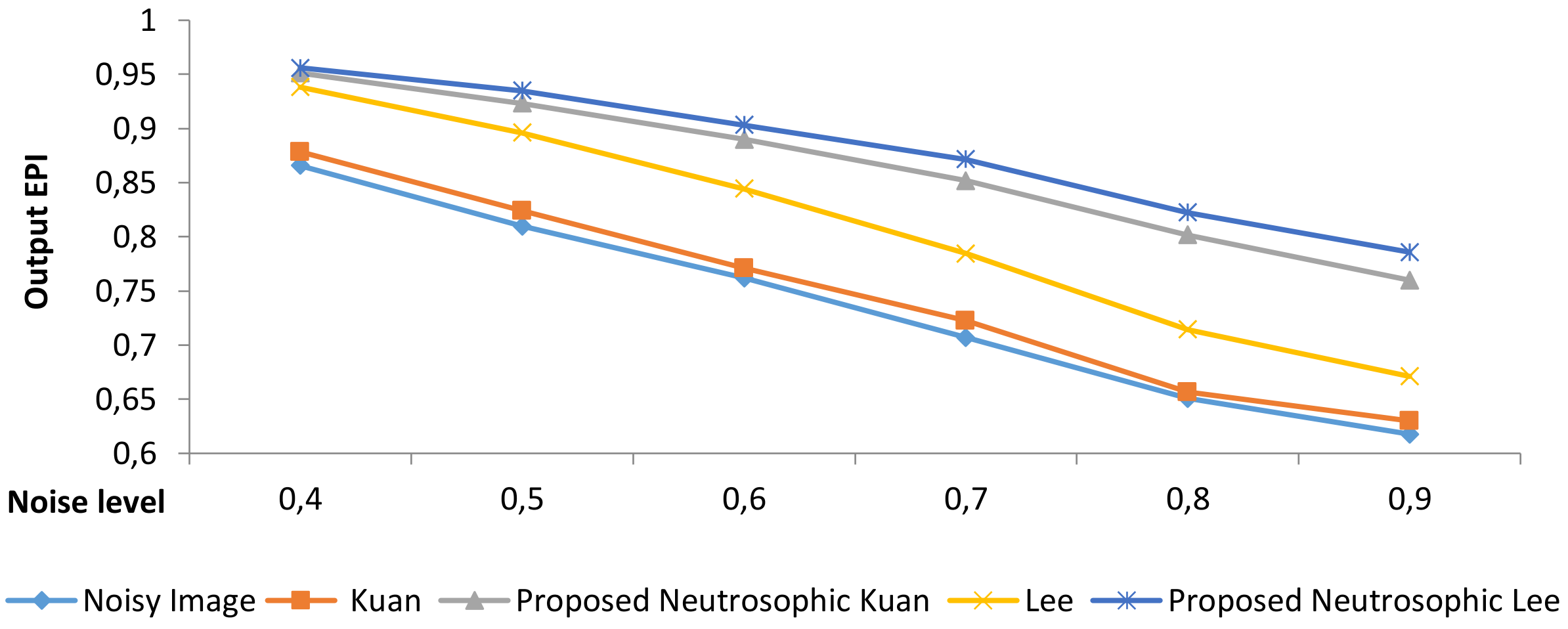

Figure 2: EPI comparison of different techniques on speckle simulated synthetic image.

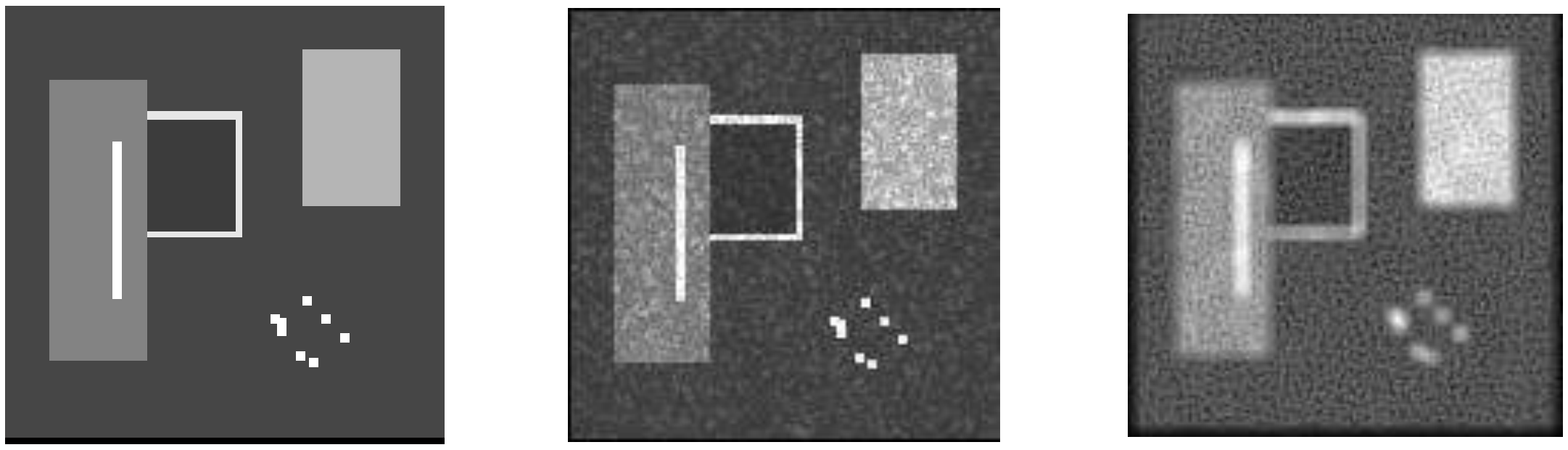

(a) (b) (c)

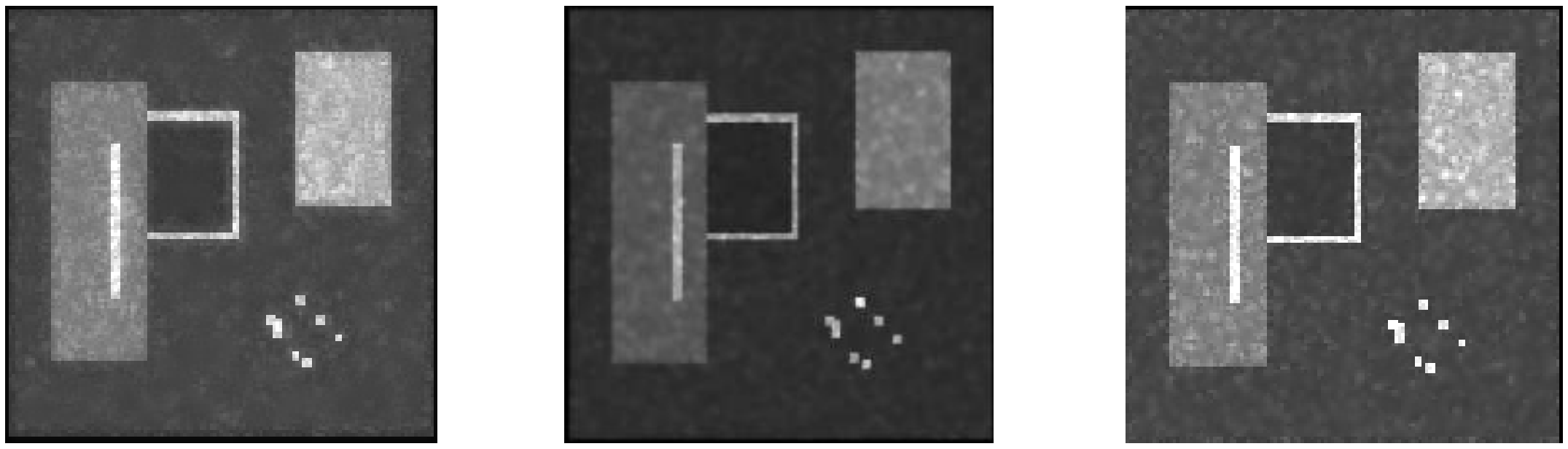

(d) (e) (f)

Figure 3: (a) Original synthetic image (b) Speckle simulated Synthetic image
(c) LEE [25] (d) KUAN [26] (e) NKUAN [28] (f) NLEE [28].

Table 2 lists the comparison of different speckle reduction methods such as Nakagami Total Variation (NTV) [29], Neutrosophic Nakagami Total variation (NNTV) [30], Non convex Sparse Regularizer Speckle Noise removal (NRSNR) [50] and Neutrosophic Nonconvex Regularizer Speckle Noise removal (NNRSNR) [27] methods in terms of SNR values at various noise levels from $\sigma = 0.3$ to 0.9. From quantitative results, it has been observed that the neutrosophic domain NNRSNR method outperformed the NRSNR method by achieving higher SNR values. Similarly, neutrosophic domain NNTV outperformed the NTV and other methods by gaining higher SNR value. It is clear from the Table 2 that both neutrosophic domain methods performed better as compared to their counterparts even at high noise levels by achieving maximum SNR values.

Table 2: SNR comparison of different methods at different noise levels ($\sigma = 0.3$ to 0.9).
SNR is given in dB.

| Methods / Variance | Noisy image | NRSNR [50] | NNRSNR [27] | NTV [29] | NNTV [30] |
|---|---|---|---|---|---|
| 0.3 | 21.11 | 22.73 | 24.22 | 25.35 | 26.89 |
| 0.4 | 19.04 | 22.12 | 23.03 | 23.73 | 24.32 |
| 0.5 | 17.9 | 22.46 | 23.73 | 24.26 | 25.86 |
| 0.6 | 16.38 | 21.64 | 21.98 | 22.75 | 23.07 |
| 0.7 | 15.21 | 18.71 | 19.66 | 20.99 | 21.75 |
| 0.8 | 13.99 | 17.20 | 18.37 | 20.15 | 20.89 |
| 0.9 | 3.63 | 6.85 | 8.75 | 9.66 | 10.33 |
| Average | 15.32 | 18.81 | 19.96 | 20.98 | 21.87 |

Similar type of observations could be made from Fig. 4 with the visual comparison of NRSNR, NTV, NNRSNR and NNTV on speckle simulated phantom image (img1). Figure 4(a) shows an original image and Fig. 4(b) displays the speckle simulated image. Whereas Fig. 4(c) reveals that the NRSNR blurred the image information such as edges. Figure 4(d) illustrates that the neutrosophic domain NNRSNR method performs well in speckle suppression. However, some of the pixels are advertantly suppressed and blurred near the boundaries. Similar type of observation could be made by Fig. 4(e) and Fig. 4(f) that the neutrosophic domain NNTV method has better visual result as compared to its counterpart in terms of speckle reduction and edge preservation.

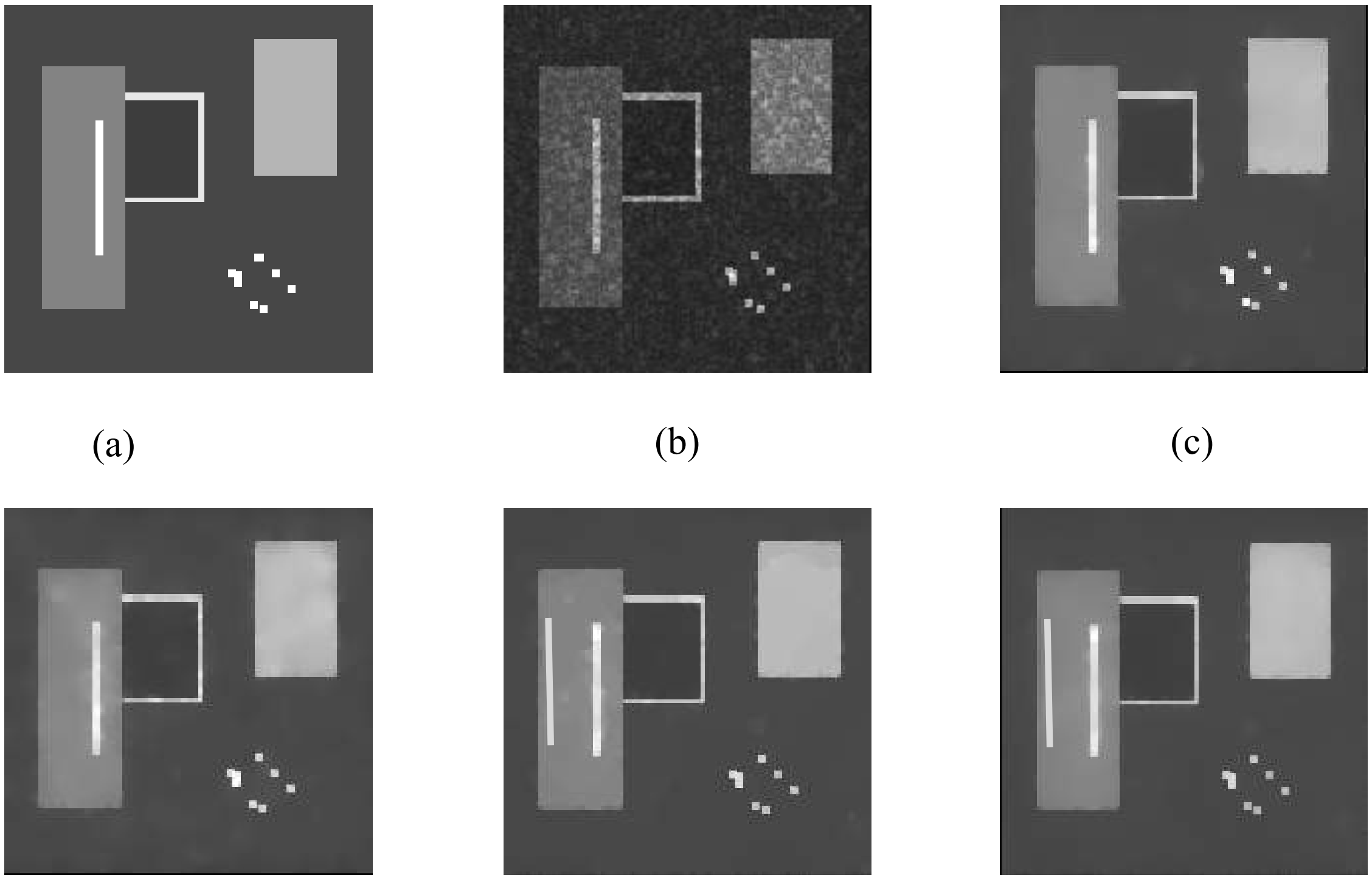

(a) (b) (c)

(d) (e) (f)

Figure 4: Visual comparison of different methods on speckle-simulated synthetic image (img2) σ=0.5. (a) Original image (b) Speckle simulated image. Image processed by (c) NRSNR (d) NNRSNR (e) NTV (f) NNTV.

### 3.2 Results of Denoising on Real Images

Figure 5 shows the results of KUAN, NKUAN, LEE and NLEE methods on medical images. The original image is shown in Fig. 5(a). The NKUAN and NLEE methods have outperformed the KUAN and LEE methods in spatial domain by removing speckle noise as illustrated in Fig. 5.

Figure 5: Visual comparison of various methods on (a) test image (b) KUAN (c) NKUAN (d) LEE (e) NLEE.

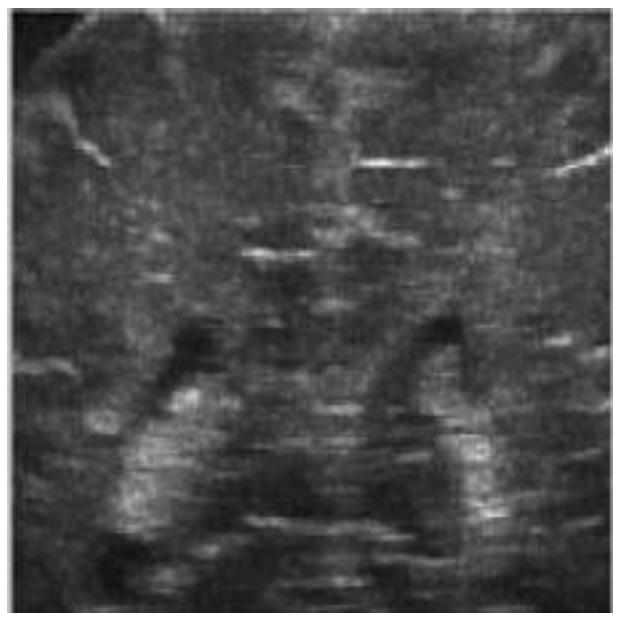

(a)

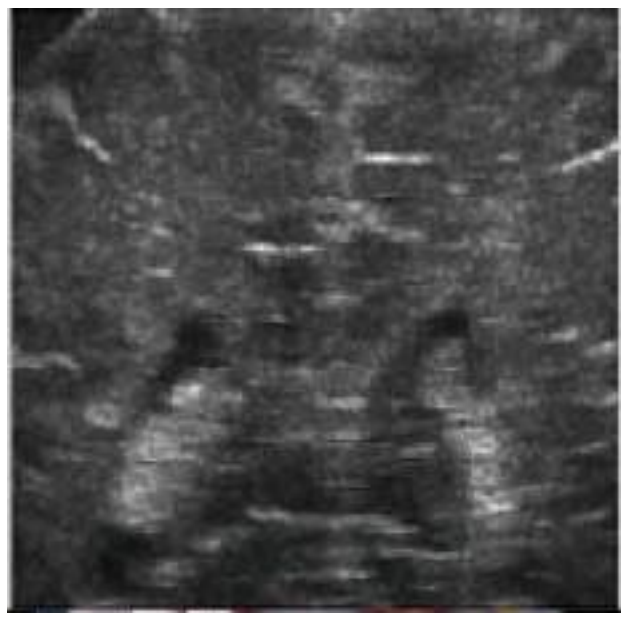

(b)

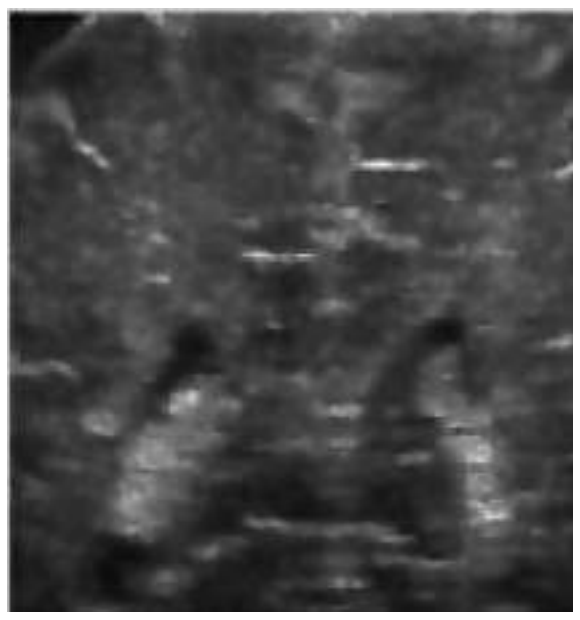

(c)

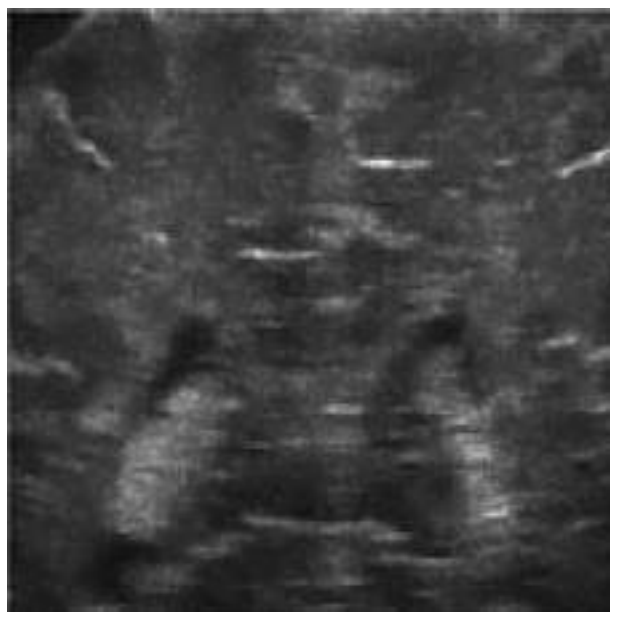

(d)

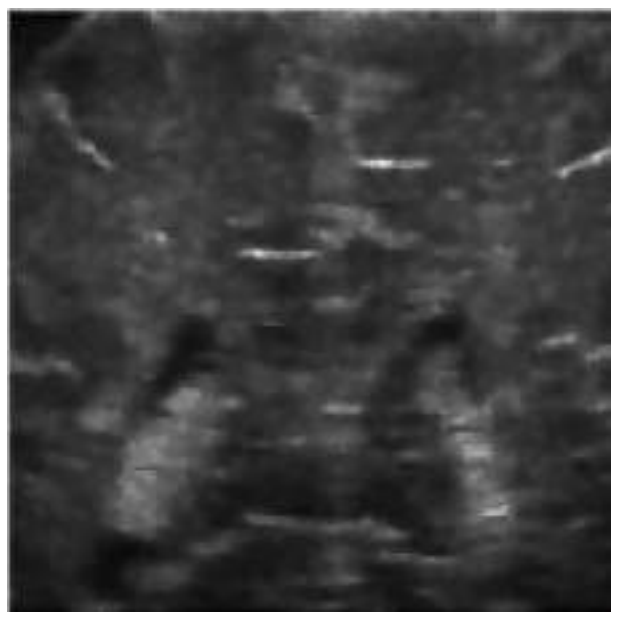

(d)

Figure 6 shows the results of NRSNR, NTV, NNTV and NNRSNR methods on thyroid ultrasound images. The original ultrasound image is given in Fig. 6(a). The NRSNR over-smoothed and blurred the images while speckle removal as illustrated in Fig. 6(b). It caused loss of important details and information of an image. The NNRSNR method has given better results but artifacts can be noticed in Fig. 6(c). The NNTV method effectively removed the speckle noise and preserved the nodule structure as illustrated in Fig. 6(e). Therefore, NNRSNR and NNTV method in neutrosophic domain can lead to efficient nodule detection in the ultrasound image. Small structures which are obscured by speckle noise become visible after processing by neutrosophic domain speckle reduction methods. The NNTV is able to remove speckle pattern, preserve anatomical structures, resolvable details and boundaries. All these results demonstrate the superiority of the neutrosophic domain methods in handling indeterminacy.

These visual outcomes are also evaluated via their line profiles shown in Fig. 7, along the line in the original image. Further, a closer glance in Fig. 7(d) and Fig. 7(f), it is observed that the NNRSNR and NNTV methods surpass the other methods by clearly highlighting the edges of thyroid nodule with the suppression of speckle noise as well as with the preservation of edges and corners in the thyroid gland ultrasound image. The methods in neutrosophic domain are able to preserve the corners, boundaries and sharp features of the image as shown in Fig. 7. Also the minute subtle details which are hidden by speckle, become noticeable in despeckled image processed by NNTV method.

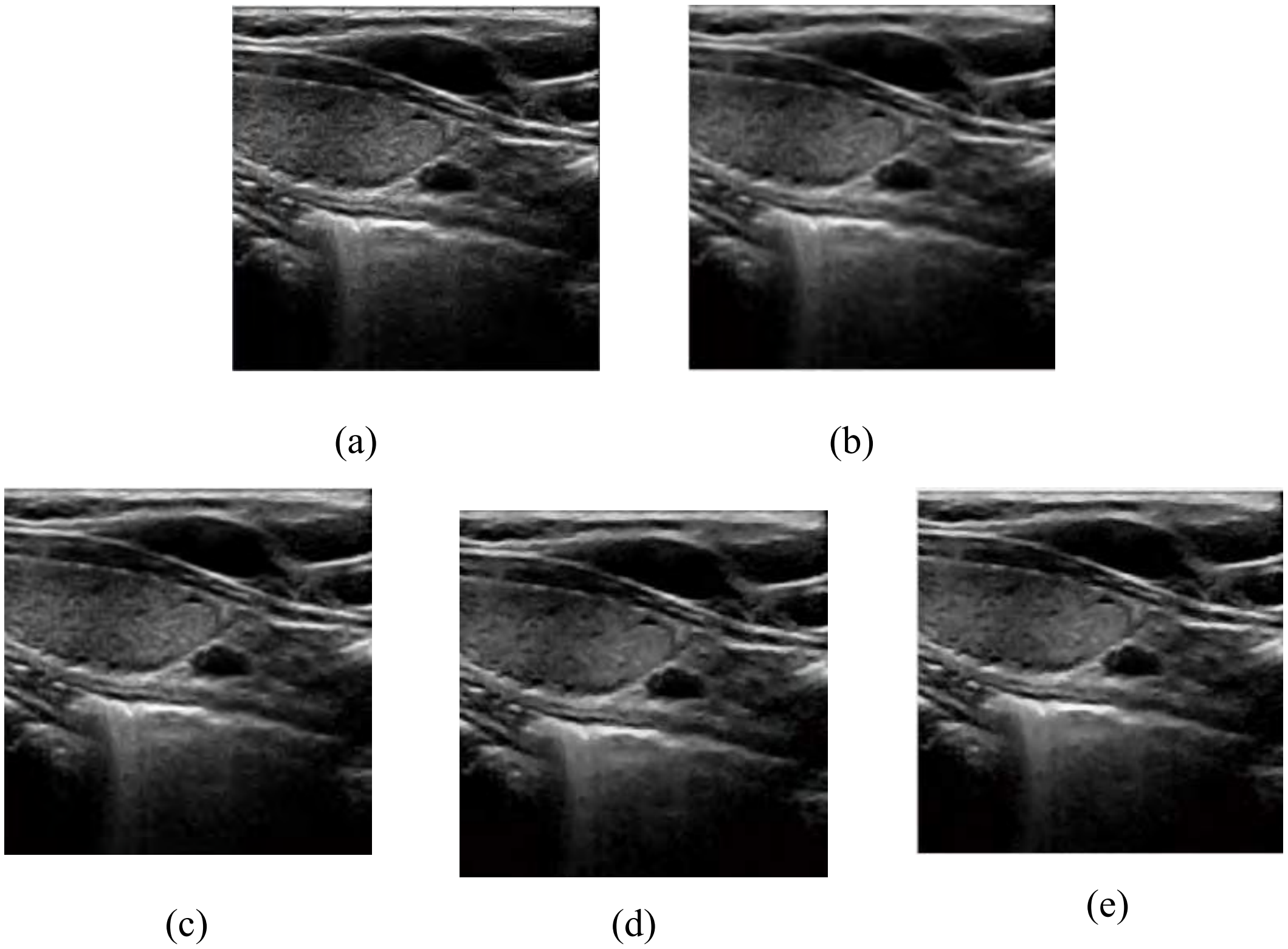

Figure 6: Visual results on the thyroid ultrasound image (img6) (a) Original image. Image processed by (b) NRSNR (c) NNRSNR (d) NTV (e) NNTV.

Further, comparison of NNRSNR and NNTV methods on real ultrasound image is illustrated in Fig. 8. Figure 8(c) shows the despeckled image and its intensity profile is shown in Fig. 8(d) along the highlighted line which revealed that the NNRSNR has lose some important information while removing speckle noise in the filtered image and has changed the contrast of the resultant image. It is observed that the NNTV method using Nakagami distribution can preserved the nodule boundaries better in ultrasound images while the degree of speckle suppression is high as compared to NNRSNR method.

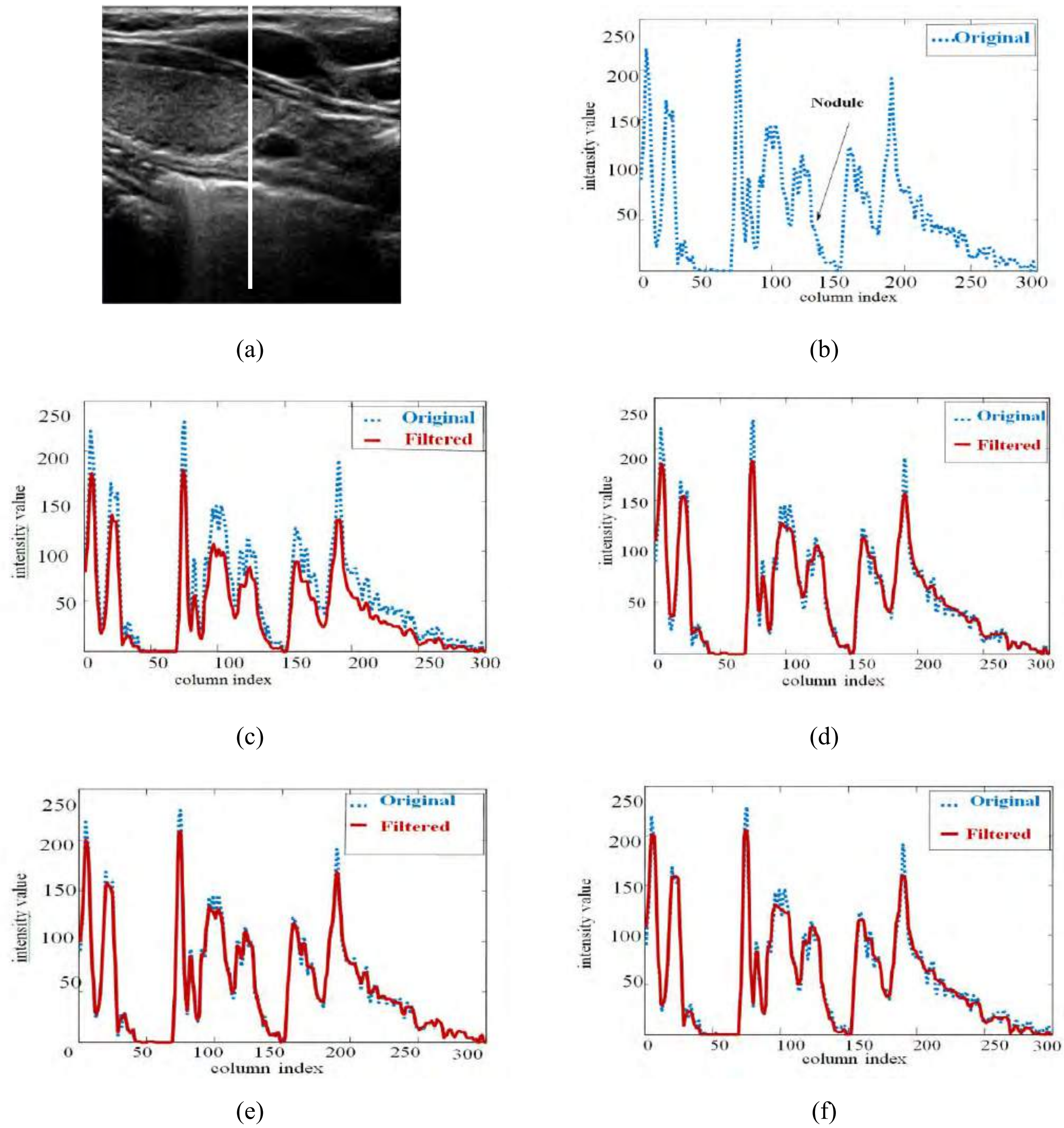

Figure 7: Line profiles for the thyroid ultrasound image (img6). (a) Original image. (b) Line profile of original image. Line profiles of (c) NRSNR (d) NNRSNR (e) NTV (f) NNTV with original image along the highlighted line.

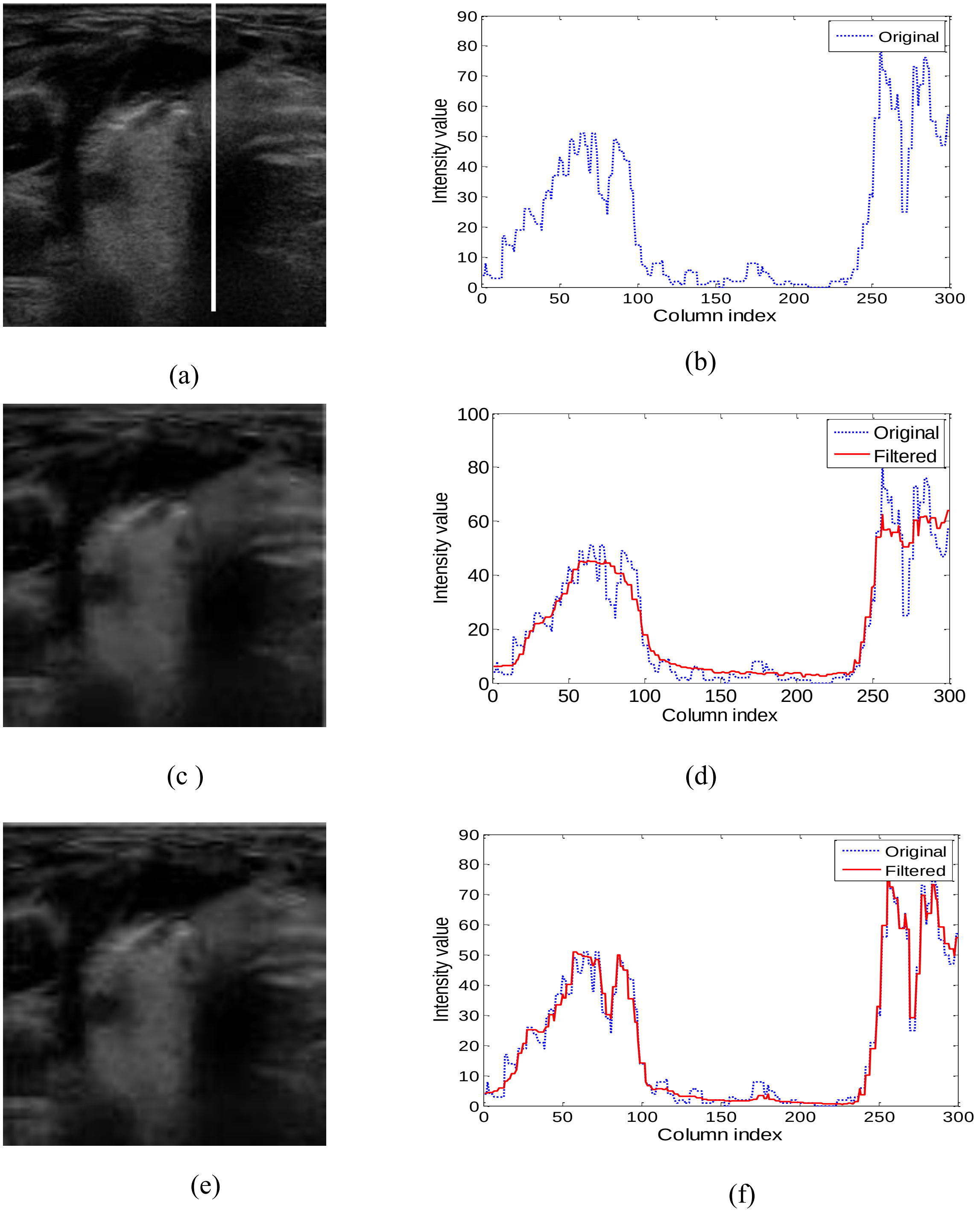

Figure 8: Denoising results on the thyroid ultrasound image (img8) (a) Original image (b) Line profile of original image (c) Image processed by NNRSNR (d) Line profile of NNRSNR (e) Image processed by NNTV (f) Line profile of NNTV.

### 1.3. Results of segmentation on real images

Further, the performance of segmentation methods in neutrosophic domain is compared on real ultrasound images [47]. Various performance metrics such as area-based and boundary-based are used to compute how much nodule pixels are correctly covered and to measure the possible disagreement over two curves [48, 49]. Area based metrics which are used in this work are True Positive (TP), False Positive (FP), Dice Coefficient (DC) and Hausdorff Distance (HD).

Table 3 lists the values of all quality metrics. As evident from results, it is observed that SNDRLS outperforms all other neutrosophic domain methods by achieving high values in terms of performance measures. The larger values of area based metrics produced by SNDRLS method assure more similarity between ground truth and the region extracted by automated segmentation method. The SNLM also reveals an improvement in FP value and HD values than other methods as listed in Table 3. The results have shown that more area is achieved by the SNDRLS method in comparison to NCM and NLM.

Table 3: Comparison of segmentation methods

| Metrics / Methods | TP (%) | DC (%) | FP (%) | HD (pixels) |
|---|---|---|---|---|
| NCM [35] | 88.5±6.2 | 78.50±18.4 | 10.93±10.9 | 20.1±19.7 |
| NLM [37] | 89.0±5.9 | 88.00 ±3.9 | 13.41±13.3 | 4.3±4.01 |
| SNLM [44] | 93.45±2.5 | 92.8±4.6 | 4.07±4.8 | 3.23±0.9 |
| SNDRLS [44] | 95.92±3.70 | 93.88±2.59 | 7.04±4.21 | 0.52±0.20 |

The quantitative results of proposed method are also supplemented with subjective outcomes. Figure 9 shows the comparison of proposed SNDRLS method with all aforementioned methods. Figure 9(a) illustrates the original thyroid ultrasound image and Fig. 9(b) shows the ground truth image. It is observed that the contour segmented by Neutrosophic Watershed (NW) is passed through the weak boundaries as shown in Fig. 9(c).

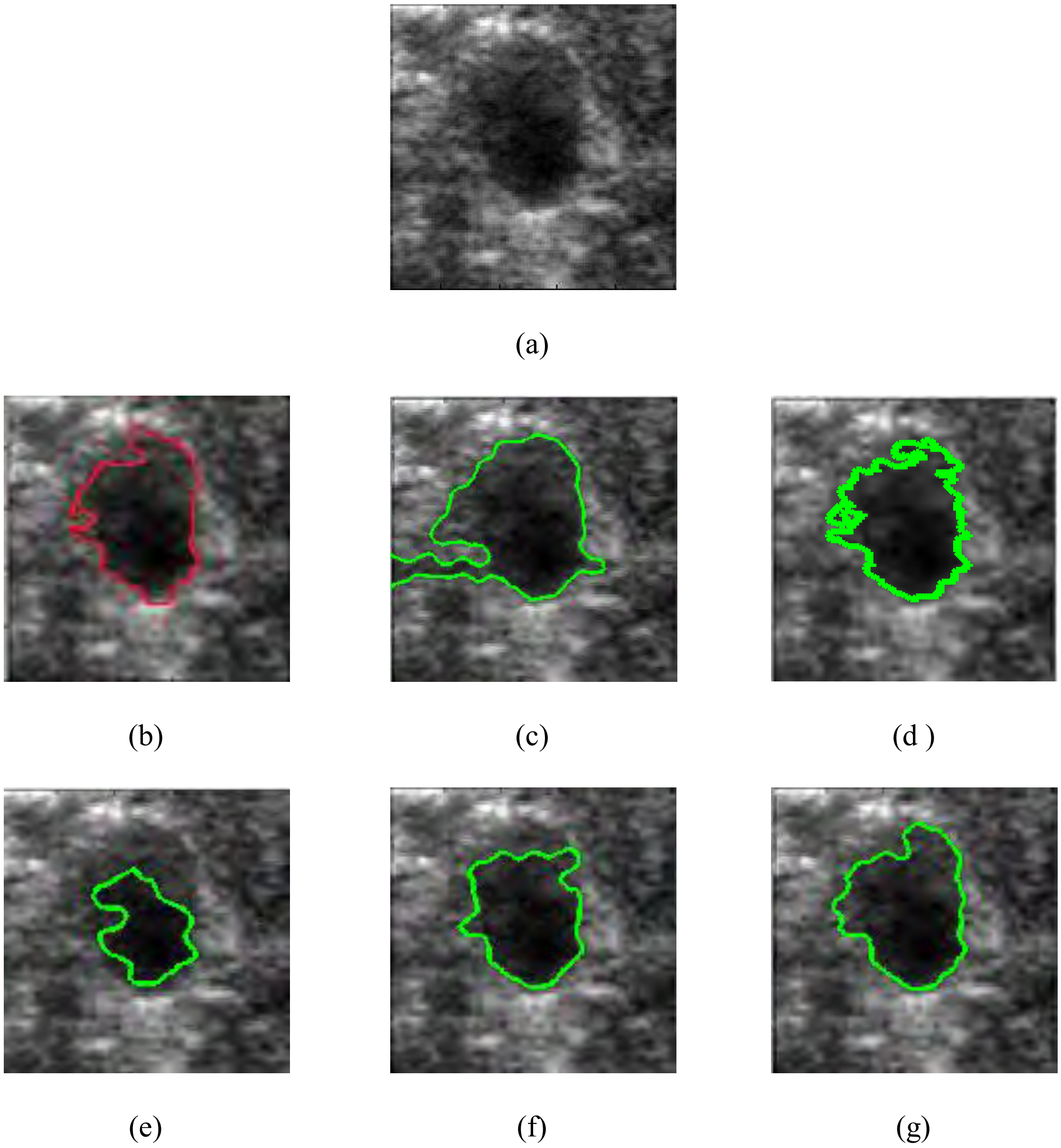

Figure 9: (a) Ultrasound image (img23) (b) Ground Truth. Segmentation results by (c) Neutrosophic Watershed (NW) [33] (d) NCM [35] (e) NLM [37] (f) SNLM [44] (g) SNDRLS [44].

Neutrosophic C Means (NCM) is affected as it is easily trapped into inappropriate local minima due to similar intensities as illustrated in Fig. 9(d). As evident from Fig. 9(e), NLM method is not able to segment the entire nodule properly. In addition, Fig. 9(f) illustrates the visual outcome of SNLM, which shows that the boundary of segmented nodule is not close to the boundary marked by an expert. It is found that the results of SNDRLS are very close to the manual segmentation as shown in Fig. 9(g). The SNDRLS is able to handle indeterminacy, fuzziness and uncertainty of pixels. From visual results, it has been noticed that the SNDRLS method is effective and accurate in nodule segmentation using ultrasound images. Figure 10(a) shows the original ultrasound image and Fig. 10(b) illustrates the ground truth image. While from Fig. 10(c), it has been noticed that the nodule is not properly segmented out due to low contrast and weak boundaries. The image segmented by NLM is able to attain delineate nodule regions with non-nodule regions also as shown in Fig. 10(d).

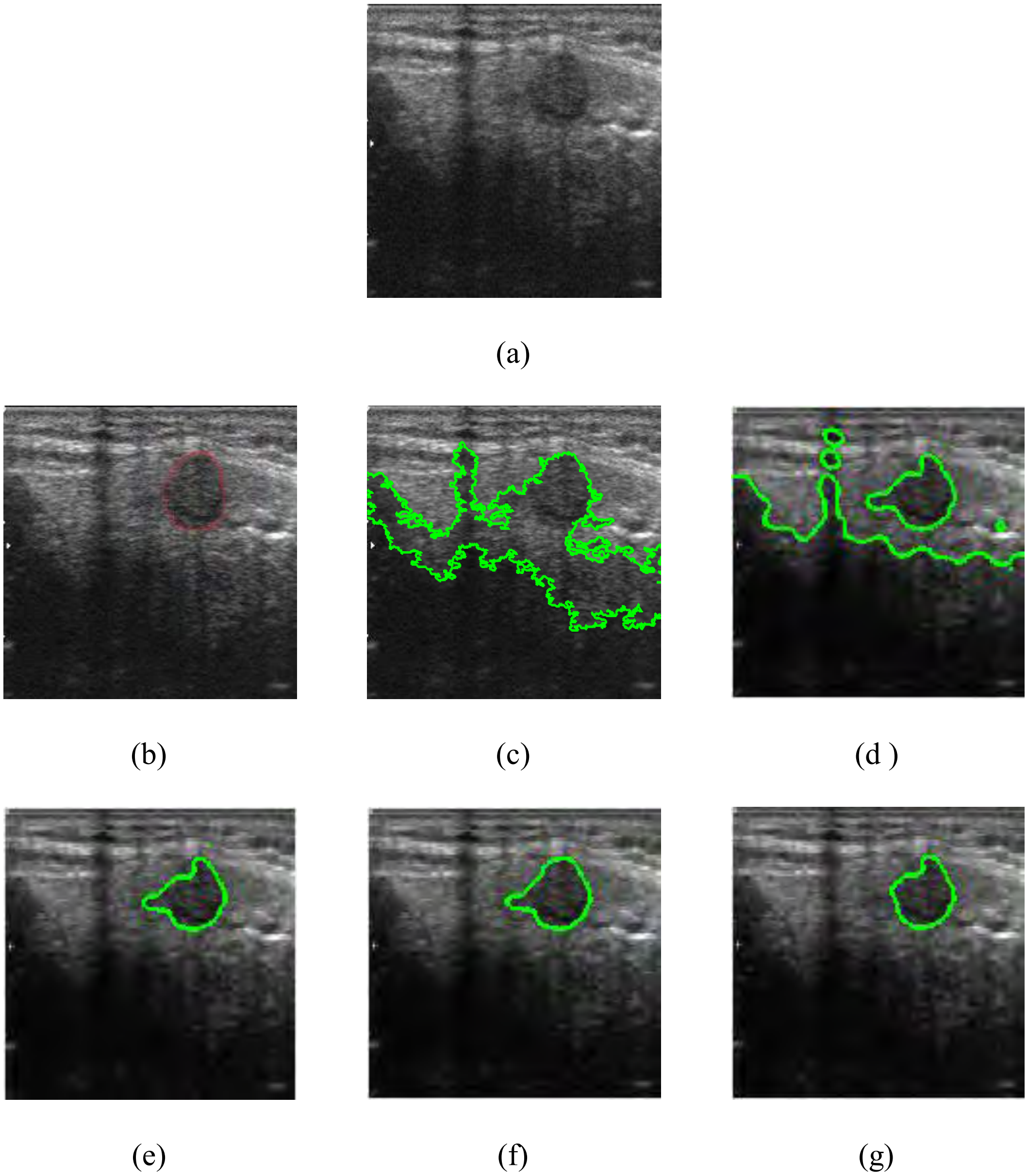

Figure 10: Ultrasound image (img318) (a) Original image (b) Ground Truth. Segmentation results by (c) NW [33] (d) NLM [37] (e) SNLM [44] (f) NCM [35] (g) SNDRLS [44].

The NCM and SNLM are able to segment the nodule in neutrosophic domain but the obtained boundary is not much close to the ground truth boundary as illustrated in Fig. 10(e) and Fig. 10(f). The best segmentation of nodule is achieved by SNDRLS as the attained delineations are very smooth and completely adapted to the thyroid nodule boundaries as shown in Fig. 10(g). Additionally, SNDRLS can prevent leakage through weak edges resulting in accurate extraction of nodule boundaries by handling the intensity in-homogeneity well.

## 4. Conclusion

Neutrosophic logic gives a powerful tool that can be used to describe the image with uncertain information. This paper provides the usefulness of neutrosophic theory in medical image denoising and segmentation. It is observed that the results using neutrosophic set are much better than the fuzzy/non fuzzy set theory because Neutrosophic set can consider more number of uncertainties by its indeterminacy handling capability. Neutrosophic set gives better result even in low contrasted images with vague region/boundaries. Through the work discussed above shows that neutrosophic based approaches can be utilized for more image processing and pattern recognition applications. It also helps in solving the problems where membership function is not defined accurately due to the lack of personal error.

## References

1. Mondal K, Dutta P, Bhattacharyya S. Fuzzy logic based gray image extraction and segmentation. International Journal of Scientific & Engineering Research. 2012;3(4):1-14.
2. Yang Y, Huang S. Image segmentation by fuzzy C-means clustering algorithm with a novel penalty term. Computing and Informatics. 2012;26(1):17-31.
3. Koundal D, Gupta S, Singh S. Applications of neutrosophic and intuitionistic fuzzy set on Image processing. National Conference on Green Technologies: Smart and Efficient Management (GTSEM-2012). 2012.
4. Smarandache F. A Unifying Field in Logics: Neutrosophic Logic. Neutrosophy, Neutrosophic Set, Neutrosophic Probability: Neutrosophic Logic. Neutrosophy, Neutrosophic Set, Neutrosophic Probability. Infinite Study; 2005.
5. Zhang M, Zhang L, Cheng HD. Segmentation of ultrasound breast images based on a neutrosophic method. Optical Engineering. 2010;49(11): 117001-117001.
6. Atanassov KT. Intuitionistic fuzzy sets. Fuzzy sets and Systems. 1986;20(1):87-96.
7. Zhang L, Zhang Y. A novel region merge algorithm based on neutrosophic logic. International Journal of Digital Content Technology and its Applications. 2011;5(7):381-7.
8. Smarandache F. A Geometric Interpretation of the Neutrosophic Set-A Generalization of the Intuitionistic Fuzzy Set. arXiv preprint math/0404520. 2004.
9. Zhang M. *Novel approaches to image segmentation based on neutrosophic logic*. (Doctoral dissertation, Utah State University). 2010.
10. Ju W. *Novel Application of Neutrosophic Logic in Classifiers Evaluated under Region-Based Image Categorization System* (Doctoral dissertation, Utah State University). 2011.
11. Smarandache F. Neutrosophic Logic-Generalization of the Intuitionistic Fuzzy Logic. arXiv preprint math/0303009. 2003.
12. Wang H, Smarandache F, Sunderraman R, Zhang YQ. Interval Neutrosophic Sets and Logic: Theory and Applications in Computing: Theory and Applications in Computing. Infinite Study. 2005(5).
13. Cheng HD, Shan J, Ju W, Guo Y, Zhang L. Automated breast cancer detection and classification using ultrasound images: A survey. Pattern Recognition. 2010;43(1):299-317.
14. Eisa M. A New Approach for Enhancing Image Retrieval using Neutrosophic Sets. International Journal of Computer Applications. 2014;95(8):12-20.

15. Shan J. A fully automatic segmentation method for breast ultrasound images. (Doctoral dissertation, Utah State University). 2011.
16. Guo Y, Cheng HD, Zhang Y. A new neutrosophic approach to image denoising. New Mathematics and Natural Computation. 2009 Nov;5(3):653-62.
17. Guo Y, Şengür A. A novel image segmentation algorithm based on neutrosophic filtering and level set. Neutrosophic Sets and Systems. 2013;1:46-49.
18. Mohan J, Chandra AT, Krishnaveni V, Guo Y. Evaluation of Neutrosophic Set Approach Filtering Technique For Image Denoising. The International Journal of Multimedia & Its Applications (IJMA). 2012;4(4):73-81.
19. Mohan J, Chandra AT, Krishnaveni V, Guo Y. Image Denoising Based on Neutrosophic Wiener Filtering. In Advances in Computing and Information Technology. Springer Berlin Heidelberg. 2013;861-869.
20. Mohan J, Krishnaveni V, Guo Y. Performance analysis of neutrosophic set approach of median filtering for MRI denoising. Int. J Elec. & Commn. Engg & Tech. 2012;3:148-163.
21. Mohan J, Krishnaveni V, Guo Y. MRI denoising using nonlocal neutrosophic set approach of Wiener filtering. Biomedical Signal Processing and Control. 2013;8(6):779-791.
22. Mohan J, Krishnaveni V, Guo Y. A new neutrosophic approach of wiener Filtering for MRI denoising. Measurement Science Review. 2013;13(4):177-186.
23. Qi X, Liu B, Xu J. A Neutrosophic Filter for High-Density Salt and Pepper Noise Based on Pixel-Wise Adaptive Smoothing Parameter. Journal of Visual Communication and Image Representation. 2016;36:1-10.
24. Guo Y, Cheng HD, Zhao W, Zhang Y. A novel image segmentation algorithm based on fuzzy c-means algorithm and neutrosophic set. Proceeding of the 11th Joint Conference on Information Sciences, Atlantis Press. 2008.
25. Lee JS. Digital image enhancement and noise filtering by use of local statistics. Pattern Analysis and Machine Intelligence, IEEE Transactions on. 1980;(2):165-168.
26. Kuan DT, Sawchuk AA, Strand TC, Chavel P. Adaptive restoration of images with speckle. In 26th Annual Technical Symposium of International Society for Optics and Photonics. 1983: 28-38.
27. Koundal D, Gupta S, Singh S. Speckle reduction method for thyroid ultrasound images in neutrosophic domain. IET Image Processing. 2016;10(2):167-75.
28. Koundal D, Gupta S, Singh S. Speckle reduction filter in neutrosophic domain. In Int. Conf. of Biomedical Engineering and Assisted Technologies. 2012:786-790.
29. Koundal D, Gupta S, Singh S. Nakagami-based total variation method for speckle reduction in thyroid ultrasound images. Proceedings of the Institution of Mechanical Engineers, Part H: Journal of Engineering in Medicine. 2016; 230(2):97-110.
30. Koundal D. Automated system for delineation of thyroid nodules in ultrasound images. 2016 (Doctoral dissertation, Panjab University, Chandigarh, India).
31. Cheng HD, Guo Y. A new neutrosophic approach to image thresholding. New Mathematics and Natural Computation. 2008;4(03):291-308.
32. Guo Y, Cheng HD. New neutrosophic approach to image segmentation. Pattern Recognition. 2009;42(5):587-595.
33. Zhang M, Zhang L, Cheng HD. A neutrosophic approach to image segmentation based on watershed method. Signal Processing. 2010;90(5):1510-1517.
34. Guo Y, Sengur A. A novel color image segmentation approach based on neutrosophic set and modified fuzzy c-means. Circuits, Systems, and Signal Processing. 2013;32(4):1699-1723.
35. Guo Y, Sengur A. NCM: Neutrosophic c-means clustering algorithm. Pattern Recognition. 2015;48(8):2710-2724.
36. Sengur A, Guo Y. Color texture image segmentation based on neutrosophic set and wavelet transformation. Computer Vision and Image Understanding. 2011;115(8):1134-1144.
37. Shan J, Cheng HD, Wang Y. A novel segmentation method for breast ultrasound images based on neutrosophic l-means clustering. Medical physics. 2012;39(9):5669-5682.

38. Karabatak E, Guo Y, Sengur A. Modified neutrosophic approach to color image segmentation. Journal of Electronic Imaging. 2013;22(1):013005(1-11).
39. Guo Y, Zhou C, Chan HP, Chughtai A, Wei J, Hadjiiski LM, Kazerooni EA. Automated iterative neutrosophic lung segmentation for image analysis in thoracic computed tomography. Medical physics. 2013;40(8):1-11.
40. Guo Y, Şengür A. A novel image segmentation algorithm based on neutrosophic filtering and level set. Neutrosophic Sets and Systems. 2013;1:46-9.
41. Guo Y, Şengür A. A novel image edge detection algorithm based on neutrosophic set. Computers & Electrical Engineering. 2014;40(8):3-25.
42. Guo Y, Sengur A. NECM: Neutrosophic evidential c-means clustering algorithm. Neural Computing and Applications. 2015;26(3):561-71.
43. Guo Y, Şengür A, Ye J. A novel image thresholding algorithm based on neutrosophic similarity score. Measurement. 2014;58:175-86.
44. Koundal D, Gupta S, Singh S. Automated delineation of thyroid nodules in ultrasound images using spatial neutrosophic clustering and level set. Applied Soft Computing. 2016;40:86-97.
45. Mazzetta J, Caudle D, Wageneck B. Digital camera imaging evaluation. Electro Optical Industries. 2005:8.
46. Chumning H, Huadong G, Changlin W. Edge preservation evaluation of digital speckle filters. IEEE International in Geoscience and Remote Sensing Symposium. IGARSS. 2002; 4, 2471-2473.
47. http://cimlaboratory.com/?lang=en&sec=proyecto&id=31
48. Clinton N, Holt A, Scarborough J, Yan LI, Gong P. Accuracy assessment measures for object-based image segmentation goodness. Photogrammetric Engineering and remote sensing. 2010;76(3):289-99.
49. Arbelaez P, Maire M, Fowlkes C, Malik J. Contour detection and hierarchical image segmentation. IEEE Transactions on Pattern Analysis and Machine Intelligence. 2011;33(5):898-916.
50. Han Y, Feng XC, Baciu G, Wang WW. Nonconvex sparse regularizer based speckle noise removal. Pattern Recognition. 2013;46(3):989-1001.

# NEUTROSOPHIC MODEL IN SOCIOLOGY

SANTANU KU. PATRO

Department of Mathematics, Berhampur University, Bhanja Bihar - 760007, Berhampur, Odisha, India.
Email: ksantanupatro@gmail.com

# On a model of Love dynamics: A Neutrosophic analysis

## Abstract

This study is an application of neutrosophy to the dynamics of love, the most interesting social phenomena. The love dynamics were studied earlier by Strogatz (Strogatz, 1994), Radzicki (Radzicki, 1993), Rapport (Rapport, 1960), etc. Although Strogatz's model (Strogatz, 1994) was originally intended only to motivate students, it makes several interesting and plausible predictions, and suggests extensions that produce even wider range of behavior. This paper has been written in the Strogatz's spirit, and it has extended Romeo & Juliet model (Sprott, 2004) to the neutrosophic domain. A love impact factor (LIF) has been proposed, and analyzed using neutrosophic logic.

## 1. Introduction

In present days, the human behavior has become an interesting issue to study. Researchers (concerning to dynamics) were looking for some new techniques to study it accurately. Apparently, studying it isn't a difficult task, and obviously it can be easily performed by the psychologists. Though, studying it accurately or near to accuracy, that is a difficult task. It may be achieved by mathematical analysis; but since there always appear indeterminacies, a more detailed analysis is required, and that is the main goal of this paper. In order to do this, we need to define human behavior in terms of equations (with indeterminacy) and we need to form a refined model, based on different feelings, taking into account different conditions. This paper deals with the refinement of love dynamics, a subject that falls in the field of social psychology, where interpersonal relationship are a topic of major concern. The feelings of love transpose in different forms; but here we opt to consider it as partner's love. One may say that romantic relationships are somehow a simpler case, since they involve only two individuals. The analysis has been performed following the modeling approach, with the induction of neutrosophic logic. An obvious difficulty in any model of love is defining what is meant by love and quantifying it in some meaning including intimacy, passion, and commitment (Strogatz, 1988); each type consists of complex mixtures of feelings. In addition to love for another person, there is love for oneself, love of life, love of humanity, and so forth. Furthermore, according to neutrosophy (Smarandache, 1998), the opposite of love may not be hate, since those two feelings can coexists, and one love some things about one's partner and hate others at the same time. Actually, the feelings in an individual can fluctuate depending on life, position, humanity or partner (Sprott, 2001). These feelings vary from person to person and from time to time. Even if everyone have the same in his/her hearts, the ratio or percentage differs. The feeling in human being varies according to different conditions. For this, different conditions and assumptions have to be applied and therefore we need to move towards an interpretative world or to think for a model that can give the complete dynamics of human feelings. It is obviously unrealistic to suppose that one's love is only influenced by only his/her own feelings and of the other related person, independent of external influences. The parameters that characterize the interactions are unchanged by excluding the possibility of learning (Scharfe & Bartholonew, 1994). However, the major goal in this research is to apply neutrosophic logic in the love model with the form of coupled ordinary differential equations.

This paper has been organized as follows: In section 2, we recall definition of neutrosophy and neutrosophic logic and preliminaries of neutrosophy. Section 3 is devoted to represent neutrosophic love model. Section 4 states open problems. Section 5 presents conclusion.

## 2. Neutrosophy & Neutrosophic logic

According to Prof. Florentin Smarandache (Smarandache, 1998), "*Neutrosophy is a branch of philosophy that studies the origin, nature and scope of neutralities as well as their interaction with different ideational spectra*". Prof. Florentin Smarandache is regarded as the father of neutrosophy, and Prof. Cheng-Gui Huang (Huang, n.d.) claims that neutrosophy is a deep thought on human culture, giving advantage to break mechanical understanding. Neutrosophic theory has been applied in many fields in order to solve problems related to indeterminacy. Neutrosophy is a generalization of Hegel's dialectics. It considers every entity < A > together with its opposite or negation < anti A > idea, refered to together as < non A >.

**Definition** (Smarandache, 1998)

A logic in which each proposition is estimated to have the percentage of truth in a subset T, the percentage of indeterminacy in a subset I, and the percentage of falsity in a subset F, where T, I, F are defined above, is called neutrosophic logic.

Actually, neutrosophic logic is a formal description frame trying to measure the truth, indeterminacy and falsehood. For detailed study of neutrosophic logic, researchers may consult the first book on neutrosophy authored by Florentin Smarandache (Smarandache, 1998). Neutrosophic logic was invented by F. Smarandache in 1995, which is an extension of fuzzy logic, intuitionistic fuzzy logic, paraconsistent logic. It deals with indeterminacy. In neutrosophic logic, every logical variable 'x' is described by an ordered triple x = (t, i, f), where t → degree of truth,
i → level of indeterminacy,
f → degree of false.

To maintain consistency with classical and fuzzy logic and with probability, there is a special case where $t + i + f = 1$. But to refer to intuitionistic logic, which means incomplete information on a variable, proposition or event, one has $t + i + f < 1$. Analogically, referring to paraconsistent logic, which means contradictory sources of information about some logical variable, proposition or event, we have $t + i + f > 1$. Florentin Smarandache (Smarandache,1998) defined neutrosophic components. Assume that T, I, F be standard or non-standard real subset of $\|-^{-}0,1^{+}-\|$. Florentin Smarandache (Smarandache, 1998) presented neutrosophic components as follows:

supT = t_sup, infT = t_inf,
supI = i_sup, infI = i_inf,
supF = f_sup, infF = f_inf, and
n_sup = t_sup + i_sup + f_sup,
n_inf = t_inf + i_inf + f_inf.

The sets T, I, F are not necessarily intervals; but may be any real sub- unitary subsets, discrete or continuous; single element, finite or (countable or uncountable) infinite; union or intersection of various subsets, etc.

## 3. Neutrosophic Love model

In this section, we present a neutrosophic love model, which is linear. The classical version was studied earlier by Strogatz (Strogatz, 1994). Generally, we classify by

a) Linear love model, and
b) Non-linear love model.

Here, we confine our discussion to linear model only.

### 3. A. Necessity of Neutrosophy in Love dynamics

It is well known that, in society, there is no single factor that affects 'love affairs'. There are so many other external factors (families, relatives, friends, enemies, situations etc.) including indeterminacy that can affect the love affairs, which are not described in previous studies (Bartholomew & Horowitz, 199; Carnelly & Janoff-Bulman, 1991; Gottman, Murray, Swanson, Tyson, & Swanson, 2002, Gragnani, Rinaldi, Feichtinger, 1997; Gragnani, Rinaldi, & Feichtinger, 1997; Griffin & Bartholomew, 1994; Kobak &Hazan, 1991; Radzicki, 1993; Rinaldi,1998a; Rinaldi,1998b; Rinaldi & Gragnani, 1998; Scharfe & Bartholomew, 1994; Sternberg, 1986; Stenberg & Barnes, 1988; Strogatz, 1988; Strogatz, 1994; Wauer, Schwarzer, Cai,&Lin,2007*)*. For an example, let us suppose that a boy, Dushmanta, is forced to love an unknown girl, Sakuntala. It should be noted that the persons (who forced Dushmanta) prisoned his sister, so that the boy acts with the girl, as a lover, only for his sister. In this case, the boy neither loves nor hates the girl. We can conclude that there is some indeterminacy in love dynamics. There are so many examples like this. Therefore, we apply neutrosophic logic to Romeo-Juliet model (Sprott, 2004).

### 3. B. Neutrosophic Linear love model (NLL model)

Let's consider a love affair between Romeo and Juliet, where

R(t) = Romeo's love (or hate, if –ve) for Juliet at a particular time 't'
J(t) = Juliet's love (or hate, if –ve) for Romeo at a particular time 't'.

The simplest neutrosophic linear love model is

$$\frac{dR}{dt} = (a+bI)R + (c+dI)J$$
$$\frac{dJ}{dt} = (e+fI)R + (g+hI)J \qquad \text{(i)}$$

Simplifying, we have

$$\frac{dR}{dt} = (aR+cJ) + (bR+dJ)I$$
$$\frac{dJ}{dt} = (eR+gJ) + (fR+hJ)I \qquad \text{(ii)}$$

where I→ level of indeterminacy, and a, b, c, d, e, f ∈ R.

**3. B. i. Features of NLL model**

The parameters 'a', 'b', 'c', 'd' in NLL model specify Romeo's situational styles, and the parameters 'e', 'f', 'g', 'h', specify Juliet's situational feelings. Overall, we can say that the parameter 'a' describes the extent to which Romeo is encouraged by his own feelings, and 'c' is the extent to which Romeo is encouraged by Juliet's feelings, 'b' describes the extent to which Romeo is encouraged or discouraged by his family or other sources, and 'd' is the extent to which Romeo is encouraged or discouraged by Juliet's family.

Now, we are going to present the characteristics of this NLL model.

**3. B. ii. Characteristics of NLL model**

We may describe the situational behavior of Romeo in this NLL model by portioning our universe of discourse $U_N$ into two parts. These are:

a) Independent indeterminacy model ($U_N^{I=0}$),

b) Dependent indeterminacy model ($U_N^{I\neq 0}$).

**3. B. ii. a. Independent Indeterminacy model ($U_N^{I=0}$)**

Here, Romeo can exhibit one of the nine romantic styles, depending upon the signs of 'a' and 'c'.

1. **Eager Behavior: [if a > 0, c > 0]** i.e. Romeo is encouraged by his own feelings as well as Juliet's.
2. **Narcissistic nerd: [if a > 0 and c < 0]** i.e. Romeo wants more of what he feels; but retreats from Juliet's feelings.
3. **Secure lover: [if a < 0, b > 0]** i.e. Romeo retreats from his own feelings; but is encouraged by Juliet's.
4. **Hermit: [if a < 0 and b < 0]** i.e. Romeo retreats from his own feelings as well as Juliet's.
5. **X-inertia: [if a > 0, b = 0]** i.e. Romeo is encouraged by his own feelings; but doesn't get any reply from Juliet's.
6. **Y-inertia: [if a = 0 and b > 0]** i.e. Romeo is encouraged by Juliet's feelings; but act as a neutral person.

7. **Juliet's Hate: [if a < 0 and b = 0]** i.e. Romeo retreats from his own feelings; but doesn't get any (positive) reply from Juliet, which ultimately leads to Juliet's hate.
8. **Romeo's hate: [if a = 0, b < 0**] i.e. Romeo retreats from Juliet's feelings, but doesn't give any positive reply, which ultimately leads to his hate towards Juliet.
9. **Not love at all: [if a = 0 and b = 0]** i.e. Both Romeo and Juliet has no reaction w.r.t each other.

**3. B. ii. b. Dependent Indeterminacy model**

In this case, there is a positive value of indeterminacy in which there exists an external factor, by means of which the love of Romeo and Juliet is affected.

1. **Limit touches the sky: [if a, b, c, d>0**] i.e. Romeo and Juliet encouraged by themselves as well as their families.
2. **Up-Romeo: [if a > 0, c > 0, b > 0, d < 0]** i.e. Romeo encouraged by himself, Juliet and his family; but the family of Juliet doesn't accept this proposal.
3. **Up-Juliet: [if a > 0, c > 0, b < 0, d>0**] i.e. Romeo & Juliet are encouraged by themselves; but Romeo's family doesn't cooperate for this love affairs.
4. **Unsecured love: [if a > 0, c > 0, b < 0, d<0]** i.e. Both Romeo and Juliet are encouraged by their love; but neither Romeo's family nor Juliet's family agree for this affair.
5. **Forced Juliet: [if a > 0, c < 0, b > 0, d>0**] i.e. both the families of Romeo and Juliet are correlated and agreed in this love affair. And Romeo is encouraged by his own love affair; but retreats from Juliet's feelings i.e. Juliet is forced to love or suppress her love.
6. **Failed Romeo: [if a > 0, c < 0, b > 0, d<0**] i.e. Romeo is encouraged by himself as well as his family; but retreats from Juliet's feelings and her family.
7. **Harassed Romeo: [if a > 0, c < 0, b < 0, d<0]** i.e. Romeo is encouraged by himself only; but has no support from both Juliet and their families.
8. **Crossed Love: [if a>0, c<0, b<0, d>0]** i.e. Romeo is encouraged by himself. Romeo's family agrees with the affair; but neither Juliet nor her family accept this affair.
9. **Suspected Love: [if a < 0, c > 0, b > 0, d>0**] i.e. Romeo retreats from his behavior and encouraged from both Juliet's behavior and her family. In this case, either Romeo suppresses his love or loves any other girl.
10. **Crossed Love [if a < 0, c > 0, b >0, d<0]** i.e. when Romeo isn't agreed, his family is agreed; but it is opposite for Juliet.
11. **Fickle Love: [if a < 0, c > 0, b < 0, d<0]** i.e. Romeo retreats from his own behavior as well as families, but encouraged by Juliet.
12. **One sided: [if a < 0, c > 0, b < 0, d > 0]** i.e. Romeo & his family retreats from the behavior of Juliet as well as her family.
13. **Family love: [if a < 0, c < 0, b >0, d > 0**] i.e. Romeo & Juliet aren't encouraged by themselves. Only their families are agreed.
14. **Not love: [if a < 0, c <0, b < 0, d < 0]** i.e. No factors are interested in this affairs.
15. **Fluctuated R-family: [if a < 0, c < 0, b > 0, d < 0]** i.e. Only Romeo's family is interested in this affairs.
16. **Fluctuated J-family: [if a < 0, c < 0, b < 0, d > 0]** i.e. Only Juliet's family is interested in this affair.

17. **Neutral Juliet: [if a > 0, c = 0, b > 0, d > 0**] i.e. Juliet is neutral in this affairs.
18. **Single Romeo: [if a > 0, c = 0, b < 0, d < 0]** i.e. only Romeo encouraged from his behavior.
19. **Lonely Romeo: [if a > 0, c = 0, b < 0, d < 0]** i.e. only Romeo is encouraged by his behavior; neither Juliet, nor their families.
20. **Moderate Romeo: [if a >0, c = 0, b< 0, d> 0]** i.e. the love is moderate, that is only Romeo is encouraged by his behavior.
21. **Neutral Romeo: [if a = 0, c > 0, b > 0, d > 0**] i.e. Juliet and her family are encouraged by themselves.
22. **Single Juliet: [if a = 0, c > 0, b < 0, d < 0]** i.e. only Juliet is agreed and Romeo is encouraged by Juliet's feelings.
23. **Moderate Juliet: [if a = 0, c > 0, b > 0, d < 0]** i.e. only Romeo is encouraged by the feelings of Juliet.
24. **Moderate J-family: [if a = 0, c > 0, b < 0, d > 0**] i.e. only Juliet is agreed in this proposal.
25. **Unarranged J-love: [if a < 0, c = 0, b > 0,d >0]** i.e. only families of the lovers are agreed in this proposal.
26. **No love: [if a < 0, c < 0, b < 0, d <0**] i.e. there exists no love.
27. **Failed J-love: [if a < 0, c = 0, b <0, d > 0**] i.e. only Juliet's family show their interest; but there is no interest from Romeo and Juliet.
28. **Failed R-Love: [if a < 0, c = 0, b > 0, d < 0]** i.e. only Romeo's family show their interest in this proposal.
29. **Unarranged R-love: [if a = 0, c < 0, b > 0, d > 0]** i.e. Families of the lovers are agreed in this issue.
30. **Family J-love: [if a = 0, c < 0, b < 0, d > 0]** i.e. only the family of Juliet agrees.
31. **Family R-love: [if a = 0, c < 0, b > 0, d <0]** i.e. only the family of Romeo agrees.
32. **No love: [if a=0, c<0, b<0, d<0]** i.e. none factors agreed and Juliet kept her behavior as neutral.
33. **One sided R-family: [if a = 0, c = 0, b > 0, d < 0**] i.e. only the family of Romeo is agreed in this proposal.
34. **One sided J-family: [if a = 0, c = 0, b < 0, d > 0]** i.e. the family of Juliet is agreed and there is no interest of others.
35. **Neutral lovers: [if a = 0, c = 0, b > 0, d > 0]** i.e. the lovers are kept as neutral and their families are interested in this issue.
36. **Never love: [if a = 0, c = 0, b < 0, d < 0]** i.e. all factors show the uninterested intention for this issue.

**3. C. Impact factor of Love Definition:** Let $U_N$ be the universe of discourse. Let 'LIF' be the impact factor of a love affair, which is defined as the index of affection of the love affair, whether it is going to succeed or to fail, or in between them.

The love impact factor is denoted as 'LIF' and defined as follows:

$$LIF = \delta_{N_{TL}, N_{IFL}} \begin{cases} < 0: \text{if } N_{TL}, N_{IFL} < 0 \\ 1. \quad = 0: \text{if } N_{TL}, N_{IFL} = 0 \\ = 1: \text{if } N_{TL}, N_{IFL} \leq \ \geq 0 \\ > 1: \text{if } N_{TL}, N_{IFL} > 1 \\ \geq 1: \text{if } N_{TL}, N_{IFL} \geq 1 \\ \leq 0: \text{if } N_{TL}, N_{IFL} \leq 0 \end{cases}$$

where, $N_{TL}$ & $N_{IFL}$ are the love functions of the parameters (a, c) and (b, d) respectively, and it is defined as follows:

$$N_{TL} = f_N(a, c) = \begin{cases} < 0 : \text{if } a, c < 0 \\ = 0 : \text{if } a , c = 0 \\ = 1 : \text{if } a \text{ or } c=0 \\ > 1 : \text{if } a > 0, c > 0 \\ \geq 1 : \text{if } a \geq 1 \text{ and } c \geq 1 \\ \leq 0: \text{if } a, c \leq 0 \end{cases}$$

and

$$N_{IFL} = f_N(b, d) = \begin{cases} = 1 : \text{if } b \text{ or } d = 0 \\ > 1 : \text{if } b > 0, d > 0 \\ \geq 1 : \text{if } b \geq 1 \text{ and } d \geq 0 \\ \leq 0: \text{if } b, d \leq 0 \\ = 0 : \text{if } b , d = 0 \\ < 0 : \text{if } b < 0, d < 0 \end{cases}$$

Now we consider some cases as follows:

**Examples: (Regarding LIF)**

1. Let's consider the case of 'limit touches the sky'. In this case, a >0, b > 0, c >0, d > 0 . So $N_{TL} > 1$ and, $N_{IFL} > 1$ this implies LIF > 1, which implies that it is very much effective love.
2. Let's consider the case of 'Up-Romeo'. In this case, $N_{TL} > 1, N_{IFL} = 1$, this implies 1- $\in$ $\leq$ LIF $\leq 1$, for all very small positive $\in$. It is like a fluctuating love, leading to success.
3. Let's consider the case of 'unsecured love'; in this case,

2. $$N_{TL} = 1 \ \& \ N_{IFL} > 1$$
$$\Rightarrow 1 - \in \ \leq \ \text{LIF} \leq 1, \forall \in > 0$$

So it is a case of fluctuating love, leading to success.

Like this, we can study any case described in Section 3, case, and can find the 'love index factor' for accuracy.

For inquiring minds, we suggest some research level open problems, as following.

## 4. Open problems

1. Extend the love dynamics to neutrosophic love triangles.
2. Propose new neutrosophic love models based upon ancient / modern society.
3. Create a neutrosophic model on attachment process.
4. Propose a neutrosophic love model and analyze it for a secure individual, etc.

## 5. Conclusions

The aim of this paper was to present a new neutrosophic love model, which would be able to describe the whole love features. Also, we proposed here for the first time the love impact factor (LIF). Due to insufficiencies of previous works, we decided to apply the neutrosophy to love dynamics, since 'love' involves indeterminacy. 'Love dynamics' being a very interesting and open topic for research, the present study may open up new avenue of research for current neutrosophic research arena.

### Acknowledgement

The author is very grateful to Prof. (Dr.) Florentin Smarandache, Mathematics & Science Department, University of New Mexico, USA and Dr. Surapati Pramanik, Department of Mathematics, Nandalal Ghosh B.T. College, Panpur, Narayanpur, West Bengal, India, for their insightful and constructive comments and suggestions, which have been very helpful in improving the paper.

## References

1. Bartholomew, K., Horowitz, L. M., Attachment styles among young adults: a test of a four-category model, *Journal of Personality and Social Psychology*, 61 (2), 226-244, 1991.
2. Carnelly, K. B., Janoff-Bulman, R., Optimism about love relationships: general vs. specific lessons from one's personal experiences, *Journal of Social and Personal Relationships*, 9, 5-20, 1992.
3. Gottman, J. M., Murray, J. D., Swanson, C. C., Tyson, R., Swanson, K. R., *The mathematics of marriage,* Cambridge, MA: MIT Press, 2002.
4. Gragnani, A., Rinaldi, S., Feichtinger, G., Cyclic dynamics in romantic relationships, *International Journal of Bifurcation and Chaos*, 7, 2611-2619, 1997.
5. Griffin, D. W., Bartholomew, K,, Models of the self and other. Fundamental dimensions underlying measures of adult attachment, *Journal of Personality and Social Psychology,* 67, 430-445, 1994.
6. Huang, C. G., A note on neutrosophy and Buddhism (n.d.), http://www.gallup.unm.edu/~smarandache/Huang-Neutrosophy.htm. Retrieved on September 15, 2016.
7. Jones, F. J., *The structure of Petrarch's Canzoniere*: A chronological, psychological, and stylistic analysis, Cambridge: Brewer, 1995.
8. Kobak, R. R., Hazan, C., Attachment in marriage: the effect of security and accuracy of working models, *Journal of Personality and Social Psychology,* 60, 861-869, 1991.
9. Radzicki, M. J., Dyadic processes, tempestuous relationships, and system dynamics, *System Dynamics Review*, 9, 79-94, 1993.

10. Rapoport, A., *Fights, games and debates,* Ann Arbor, University of Michigan Press. NDPLS, 8(3), 1960.
11. Rinaldi, S., Love dynamics: the case of linear couples, *Applied Mathematics and Computation*, 95, 181-192, 1998a.
12. Rinaldi, S., Laura and Petrarch: An intriguing case of cyclical love dynamics, *SIAM Journal on Applied Mathematics,* 58, 1205-1221, 1998b.
13. Rinaldi, S., Gragnani, A., Love dynamics between secure individuals: A modeling approach, *Nonlinear Dynamics, Psychology, and Life Sciences*, 2, 283-301, 1998.
14. Scharfe, E., Bartholomew, K., Reliability and stability of adult attachment patterns, *Personal Relationships*, 1, 23-43, 1994.
15. Smarandache, F., *A unifying field in logic. Neutrosopy, neutrosophic set, neutrosophic probability & statistics,* American Research Press, Rehoboth, 1998.
16. Sprott, J. C., *Dynamics of Love and Happiness,* Chaos and Complex Systems Seminar in Madison, Wisconsin 2001.
17. Sprott, J. C., *Chaos and time-series analysis,* Oxford: Oxford University Press, 2003.
18. Sprott, J. C., Dynamical model of love, nonlinear dynamics, *Psychology & Life Sciences*, 8 (3), 303-313, 2004.
19. Sternberg, R. J, The triangular theory of love, *Psychological Review*, 93, 119-135, 1986.
20. Stenberg, R. J., Barnes, M. L. (Eds.), *The psychology of love,* New Haven, CT: Yale University Press, 1988.
21. Strogatz, S. H.*, Love affairs and differential equations, Mathematics Magazine*, 61(1), 35, 1988.
22. Strogatz, S. H., *Nonlinear dynamics and chaos with applications to physics*, *biology, chemistry and engineering*. Addison-Wesley, Reading, M.A. 1994.
23. *Wauer J., Schwarzer D, Cai, G.Q. Lin, Y. K.,* Dynamical models of love with time-varying fluctuations*, Applied Mathematics and Computation,* 188, 1535-1448, 2007.

# PROBABILITY THEORY

A. A. SALAMA[1], FLORENTIN SMARANDACHE[2]

[1]Department of Mathematics and Computer Science, Faculty of Sciences, Port Said University, Egypt. Email: drsalama44@gmail.com
[2]Department of Mathematics, University of New Mexico Gallup, NM, USA. Email: smarand@unm.edu

# Neutrosophic Crisp Probability Theory & Decision Making Process

## Abstract

Since the world is full of indeterminacy, the neutrosophics found their place into contemporary research. In neutrosophic set, indeterminacy is quantified explicitly and truth-membership, indeterminacy-membership and falsity-membership are independent. So it is natural to adopt for that purpose the value from the selected set with highest degree of truth-membership, indeterminacy membership and least degree of falsity-membership on the decision set. These factors indicate that a decision making process takes place in neutrosophic environment. In this paper, we introduce and study the probability of neutrosophic crisp sets. After given the fundamental definitions and operations, we obtain several properties and discussed the relationship between them. These notions can help researchers and make great use of it in the future in making algorithms to solving problems and manage between these notions to produce a new application or new algorithm of solving decision support problems. Possible applications to mathematical computer sciences are touched upon.

## 1. Introduction

Neutrosophy has laid the foundation for a whole family of new mathematical theories generalizing both their classical and fuzzy counterparts [1, 2, 3, 22, 23, 24, 25, 26, 27, 28, 29, 30, 31, 32, 33, 34, 35, 36, 42] such as a neutrosophic set theory. The fundamental concepts of neutrosophic set, introduced by Smarandache in [48, 49, 50, 51], and Salama et al. in [4, 5, 6, 7, 8, 9, 10, 11, 12, 13, 14, 15, 16, 17, 18, 19, 20, 21], provides a natural foundation for treating mathematically the neutrosophic phenomena which exist pervasively in our real world and for building new branches of neutrosophic mathematics. In this paper is to introduce and study the probability of neutrosophic crisp sets. After given the fundamental definitions and operations, we obtain several properties, and discussed the relationship between neutrosophic crisp sets and others.

## 2. Terminologies

We recollect some relevant basic preliminaries, and in particular, the work of Smarandache in [37, 38, 39, 40], and Salama et al. [4, 5, 6, 7, 8, 9, 10, 11, 12, 13, 14, 15, 16, 17, 18, 19, 20, 21]. Smarandache introduced the neutrosophic components T, I, F which represent the membership, indeterminacy, and non-membership values respectively, where $\left]^{-}0, 1^{+}\right[$ is nonstandard unit interval.

**Example 2.1** [37, 39]

Let us consider a neutrosophic set a collection of possible locations (position) of particle x and Let A and B two neutrosophic sets. One can say, by language abuse, that any particle x neutrosophically belongs to any set, due to the percentages of truth/indeterminacy/falsity involved, which varies between $^{-}0$ and $1^{+}$.For example :x(0.5,0.2,0.3) belongs to A (which means, the probability of 50% particle x is in a poison of A, with a probability of 30% x is not in A, and the rest is undecidable); or y(0,0,1) belongs to A( which normally means y is not for sure in A );or z(0,1,0) belongs to A (which means one does know absolutely nothing about z affiliation with A). More general, x((0.2-0.3),(0.4—0.45) $\cup$ [0.50-0.51],{0.2,0.24,0.28}) belongs to the seta, which means: With a probability in between 20-30% particle x is in a position of A ( one cannot find an exact approximate because of various sources used ); With a probability of 20% or 24% or 28% x is not in A; The indeterminacy related to the appurtenance of x to A is in between 40-45% or between 50-51% ( limits included ). The subsets representing the appurtenance, indeterminacy, and falsity may overlap, and n-sup = 30%+51%+28% > 100 in this case.

**Definition 2.1** [14, 15, 21]

A neutrosophic crisp set (NCS for short) $A = \langle A_1, A_2, A_3 \rangle$ can be identified to an ordered triple $\langle A_1, A_2, A_3 \rangle$ are subsets on X, and every crisp set in X is obviously an NCS having the form $\langle A_1, A_2, A_3 \rangle$,

**Definition 2.2** [21]

The object having the form $A = \langle A_1, A_2, A_3 \rangle$ is called

**(Neutrosophic Crisp Set with Type I)** If satisfying $A_1 \cap A_2 = \phi$ , $A_1 \cap A_3 = \phi$ and $A_2 \cap A_3 = \phi$. (NCS-Type I for short).

**(Neutrosophic Crisp Set with Type II)** If satisfying $A_1 \cap A_2 = \phi$ , $A_1 \cap A_3 = \phi$ and $A_2 \cap A_3 = \phi$ and $A_1 \cup A_2 \cup A_3 = X$. (NCS-Type II for short).

**(Neutrosophic Crisp Set with Type III)** If satisfying, $A_1 \cap A_2 \cap A_3 = \phi$ and $A_1 \cup A_2 \cup A_3 = X$. (NCS-Type III for short).

**Definition 2.3**

1) **(Neutrosophic Set** [7]): Let X be a non-empty fixed set. A neutrosophic set ( NS for short) $A$ is an object having the form $A = \langle \mu_A(x), \sigma_A(x), \nu_A(x) \rangle$ where $\mu_A(x), \sigma_A(x)$ and $\nu_A(x)$ which represent the degree of member ship function (namely $\mu_A(x)$), the degree of indeterminacy (namely $\sigma_A(x)$), and the degree of non-member ship (namely $\nu_A(x)$) respectively of each element $x \in X$ to the set $A$ where $0^- \leq \mu_A(x), \sigma_A(x), \nu_A(x) \leq 1^+$ and $0^- \leq \mu_A(x) + \sigma_A(x) + \nu_A(x) \leq 3^+$.

**2) (Neutrosophic Intuitionistic Set of Type 1** [8]**)**: Let X be a non-empty fixed set. A neutrosophic intuitionistic set of type 1 (NIS1 for short) set $A$ is an object having the form $A = \langle \mu_A(x), \sigma_A(x), \nu_A(x) \rangle$ where $\mu_A(x), \sigma_A(x)$ and $\nu_A(x)$ which represent the degree of member ship function (namely $\mu_A(x)$), the degree of indeterminacy (namely $\sigma_A(x)$), and the degree of non-membership (namely $\nu_A(x)$) respectively of each element $x \in X$ to the set $A$ where $0^- \leq \mu_A(x), \sigma_A(x), \nu_A(x) \leq 1^+$ and the functions satisfy the condition $\mu_A(x) \wedge \sigma_A(x) \wedge \nu_A(x) \leq 0.5$ and $0^- \leq \mu_A(x) + \sigma_A(x) + \nu_A(x) \leq 3^+$.

**3) (Neutrosophic Intuitionistic Set of Type 2** [41]**)**. Let X be a non-empty fixed set. A neutrosophic intuitionistic set of type 2 $A$ (NIS2 for short) is an object having the form $A = \langle \mu_A(x), \sigma_A(x), \nu_A(x) \rangle$ where $\mu_A(x), \sigma_A(x)$ and $\nu_A(x)$ which represent the degree of member ship function (namely $\mu_A(x)$), the degree of indeterminacy (namely $\sigma_A(x)$), and the degree of non-membership (namely $\nu_A(x)$) respectively of each element $x \in X$ to the set $A$ where $0.5 \leq \mu_A(x), \sigma_A(x), \nu_A(x)$ and the functions satisfy the condition $\mu_A(x) \wedge \sigma_A(x) \leq 0.5$, $\mu_A(x) \wedge \nu_A(x) \leq 0.5$, $\sigma_A(x) \wedge \nu_A(x) \leq 0.5$, and $^-0 \leq \mu_A(x) + \sigma_A(x) + \nu_A(x) \leq 2^+$. A neutrosophic crisp with three types the object $A = \langle A_1, A_2, A_3 \rangle$ can be identified to an ordered triple $\langle A_1, A_2, A_3 \rangle$ are subsets on X, and every crisp set in X is obviously a NCS having the form $\langle A_1, A_2, A_3 \rangle$. Every neutrosophic set $A = \langle \mu_A(x), \sigma_A(x), \nu_A(x) \rangle$ on $X$ is obviously on NS having the form $\langle \mu_A(x), \sigma_A(x), \nu_A(x) \rangle$.

Salama et al. in [14, 15, 21] constructed the tools for developed neutrosophic crisp set, and introduced the NCS $\phi_N, X_N$ in X.

**Remark 2.1**

i) The neutrosophic intuitionistic set is a neutrosophic set but the neutrosophic set is not in general a neutrosophic intuitionistic set in general.

ii)Neutrosophic crisp sets with three types are neutrosophic crisp set.

## 3. The Probability of Neutrosophic Crisp Sets

If an experiment produces indeterminacy, that is called a neutrosophic experiment. Collecting all results, including the indeterminacy, we get the neutrosophic sample space (or the neutrosophic probability space) of the experiment. The neutrosophic power set of the neutrosophic sample space is formed by all different collections (that may or may not include the indeterminacy) of possible results. These collections are called neutrosophic events. In classical experimental the probability is $\left(\frac{\text{number of times event A occurs}}{\text{totel number of trials}}\right)$. Similarly, Smarandache [16, 17, 18] introduced neutrosophic experimental probability as follows:

$$\left(\frac{\text{number of times event A occurs}}{\text{total number of trials}}, \frac{\text{number of times indeterminacy occurs}}{\text{total number of trials}}, \frac{\text{number of times event A does not occur}}{\text{total number of trials}}\right)$$

Probability of NCS is a generalization of the classical probability in which the chance that event $A=\langle A_1, A_2, A_3\rangle$ occurs is: $P(A_1)$ *true*, $P(A_2)$ indeterminate, $P(A_3)$ *false*, on a sample space X, or $NP(A)=\langle P(A_1), P(A_2), P(A_3)\rangle$.

A subspace of the universal set, endowed with a neutrosophic probability defined for each of its subset, forms a probability neutrosophic crisp space.

### Definition 3.1

Let X be a non- empty set and A be any type of neutrosophic crisp set on a space X, then the probability is a mapping $NP: X \to [0,1]^3$, $NP(A)=\langle P(A_1), P(A_2), P(A_3)\rangle$ that is the probability a neutrosophic crisp set has the property that, $\text{NP(A)} = \begin{cases} (p_1,p_2,p_3) & \text{where } p_{1,2,3} \in [0,1] \\ 0 & \text{if } p_1,p_2,p_3 < o \end{cases}$,

### Remark 3.1

i) In case if $A=\langle A_1, A_2, A_3\rangle$ is NCS then $^-0 \le P(A_1)+P(A_2)+P(A_3) \le 3^+$

ii) In case if $A=\langle A_1, A_2, A_3\rangle$ is NCS-Type I then $0 \le P(A_1)+P(A_2)+P(A_3) \le 2$ .

iii) The probability of NCS-Type II is a neutrosophic crisp set where $^-0 \le P(A_1)+P(A_2)+P(A_3) \le 2^+$

iv) The probability of NCS-Type III is a neutrosophic crisp set where $^-0 \le P(A_1)+P(A_2)+P(A_3) \le 3^+$.

**Probability Axioms of NCS**

**Axioms:**

1- The probability of neutrosophic crisp set and NCS-Type III A on X
$NP(A)=\langle P(A_1), P(A_2), P(A_3)\rangle$ where $P(A_1)\ge 0, P(A_2)\ge 0, P(A_3)\ge 0$ or

$$NP(A) = \begin{cases} (p_1,p_2,p_3) & \text{where } p_{1,2,3} \in [0,1] \\ 0 & \text{if } p_1,p_2,p_3 < o \end{cases}$$

2- The probability of neutrosophic crisp set and NCS-Type IIIs A on X
$NP(A)=\langle P(A_1), P(A_2), P(A_3)\rangle$ where $^-0 \le p(A_1)+p(A_2)+p(A_3) \le 3^+$.

3- Bonding the probability of neutrosophic crisp set and NCS-Type IIIs
$NP(A)=\langle P(A_1), P(A_2), P(A_3)\rangle$ where $1 \ge P(A_1) \ge 0, P(A_2)\ge 0, P(A_3)\ge 0.$

4- Addition law for any two neutrosophic crisp sets or NCS-Type III

i) $$NP(A\cup B) =< (P(A_1)+P(B_1)-P(A_1\cap B_1), (P(A_2)+P(B_2)-P(A_2\cap B_2), (P(A_3)+P(B_3)-P(A_3\cap B_3) >$$

if $A \cap B = \phi_N$, then $NP(A \cap B) = NP(\phi_N)$.

$$NP(A \cup B) =< NP(A_1) + NP(B_1) - NP(\phi_{N_1}), NP(A_2) + NP(B_2) - NP(\phi_{N_2}),$$
$$NP(A_3) + NP(B_3) - NP(\phi_{N_3}).$$

Since our main purpose is to construct the tools for developing probability of neutrosophic crisp sets, we must introduce the following:

1) Probability of neutrosophic crisp empty set with three types ( $NP(\phi_N)$ for short) may be defined as four types:

i) Type 1: $NP(\phi_N) = \langle P(\phi), P(\phi), P(X) \rangle =< 0,0,1 >$

ii) Type 2: $NP(\phi_N) = \langle P(\phi), P(X), P(X) \rangle =< 0,1,1 >$

i)Type 3: $NP(\phi_N) = \langle P(\phi), P(\phi), P(\phi) \rangle =< 0,0,0 >$

ii) Type 4: $NP(\phi_N) = \langle P(\phi), P(X), P(\phi) \rangle =< 0,1,0 >$

2) Probability of neutrosophic crisp universal and NCS-Type III universal sets ( $NP(X_N)$ ) may be defined as four types:

i)Type 1: $NP(X_N) = \langle P(X), P(\phi), P(\phi) \rangle =< 1,0,0 >$

ii) Type 2: $NP(X_N) = \langle P(X), P(X), P(\phi) \rangle =< 1,1,0 >$

iii) Type 3: $NP(X_N) = \langle P(X), P(X), P(X) \rangle =< 1,1,1 >$

iv) Type 4: $NP(X_N) = \langle P(X), P(\phi), P(X) \rangle =< 1,0,1 >$

**Remark 3.1**

1) $NP(X_N) = 1_N$, $NP(\phi_N) = O_N$ .Where $1_N, O_N$ are in Definition 2.1 [6], or equals any type for $1_N$ .

2) The probability of neutrosophic crisp set is a neutrosophic set.

**Definition 3.2 (Monotonicity)**

Let $X$ be a non-empty set, and NCSS $A$ and $B$ in the form $A = \langle A_1, A_2, A_3 \rangle$, $B = \langle B_1, B_2, B_3 \rangle$ with $NP(A) = \langle P(A_1), P(A_2), P(A_3) \rangle$, $NP(B) = \langle P(B_1), P(B_2), P(B_3) \rangle$ then we may consider two possible definitions for subsets ( $A \subseteq B$ )

( $A \subseteq B$ ) may be defined as two types:

1) Type1: $NP(A) \le NP(B) \Leftrightarrow P(A_1) \le P(B_1), P(A_2) \le P(B_2)$ and $P(A_3) \ge P(B_3)$ or

2) Type2: $NP(A) \le NP(B) \Leftrightarrow P(A_1) \le P(B_1), P(A_2) \ge P(B_2)$ and $P(A_3) \ge P(B_3)$ .

**Definition 3.3**

Let χ be a non-empty set, and NCSs $A$ and $B$ in the form $A=\langle A_1, A_2, A_3\rangle$, $B=\langle B_1, B_2, B_3\rangle$ are NCSs. Then

1. $NP(A\cap B)$ may be defined two types as:

i) Type1: $NP(A\cap B)=\langle P(A_1\cap B_1), P(A_2\cap B_2), P(A_3\cup B_3)\rangle$ or

ii) Type2: $NP(A\cap B)=\langle P(A_1\cap B_1), P(A_2\cup B_2), P(A_3\cup B_3)\rangle$

2. $NP(A\cup B)$ may be defined two types as:

i) Type1: $NP(A\cup B)=\langle P(A_1\cup B_1), P(A_2\cap B_2), P(A_3\cap B_3)\rangle$ or

ii) Type 2: $NP(A\cup B)=\langle P(A_1\cup B_1), P(A_2\cup B_2), P(A_3\cap B_3)\rangle$

3. $NP(A^c)$ may be defined by three types

i) Type1: $NP(A^c)=\left\langle P(A_1{}^c), P(A_2^c), P(A_3^c)\right\rangle = <(1-A_1),(1-A_2),(1-A_3)>$ or

ii) Type2: $NP(A^c)=\langle P(A_3), P(A_2^c), P(A_1)\rangle$ or

iii) Type3: $NP(A^c)=\langle P(A_3), P(A_2), P(A_1)\rangle$.

**Proposition 3.1**

Let $A$ and $B$ in the form $A=\langle A_1, A_2, A_3\rangle$, $B=\langle B_1, B_2, B_3\rangle$ are NCSs on a non-empty set χ. Then

1) $NP(A)^c + NP(A) =< (1,1,1>$ or Type (iii) of $NP(X_N)=1_N$ or = any types for $1_N$.

2) $NP(A-B) = NP(A-B) =< (P(A_1)-P(A_1\cap B_1), (P(A_2)-P(A_2\cap B_2),$
$(P(A_3)-P(A_3\cap B_3)>$

3) $NP(A/B) =< \dfrac{NP(A_1)}{NP(A_1\cap B_1)}, \dfrac{NP(A_2)}{NP(A_2\cap B_2)}, \dfrac{NP(A_3)}{NP(A_3\cap B_3)} >$

**Proposition 3.1**

Let $A$ and $B$ in the form $A=\langle A_1, A_2, A_3\rangle$, $B=\langle B_1, B_2, B_3\rangle$ are NCSs on a non-empty set χ. And $p$, $p_N$ are NCSs Then

i) $$NP(p)=\left\langle \frac{1}{n(X)}, \frac{1}{n(X)}, \frac{1}{n(X)}\right\rangle$$

ii) $$NP(p_N)=\left\langle 0, \frac{1}{n(X)}, 1-\frac{1}{n(X)}\right\rangle$$

**Example 3.1**

1) Let $X = \{a,b,c,d\}$ and $A$ , $B$ are two neutrosophic crisp events on X defined by $A = \langle \{a\},\{b,c\},\{c,d\}\rangle$ , $B = \langle \{a,b\},\{a,c\},\{c\}\rangle$ , $p = \langle \{a\},\{c\},\{d\}\rangle$ then see that $NP(A) = \langle 0.25,0.5,0.5\rangle$, $NP(B) = \langle 0.5,0.5,0.25\rangle$, $NP(p) = \langle 0.25,0.25,0.25\rangle$, one can compute all probabilities from definitions.

2) If $A = \langle \{\phi\},\{b,c\},\{\phi\}\rangle$ and $B = \langle \{\phi\},\{d\},\{\phi\}\rangle$ are neutrosophic crisp sets on X then : $A \cap B = \langle \{\phi\},\{\phi\},\{\phi\}\rangle$ and $NP(A \cap B) = \langle 0,0,0\rangle = 0_N$ ,

$A \cap B = \langle \{\phi\},\{b,c,d\},\{\phi\}\rangle$ and $NP(A \cap B) = \langle 0,0.75,0\rangle \neq 0_N$ .

**Example 3.2**

Let $X = \{a,b,c,d,e,f\}$ , $A = \langle \{a,b,c,d\},\{e\},\{f\}\rangle$ , $D = \langle \{a,b\},\{e,c\},\{f,d\}\rangle$ be a NCS-Type 2,

$B = \langle \{a,b,c\},\{d\},\{e\}\rangle$ be a NCT-Type I but not NCS-Type II, III, $C = \langle \{a,b\},\{c,d\},\{e,f,a\}\rangle$ be a NCS-Type III, but not NCS-Type I,II, $E = \langle \{a,b,c,d,e\},\{c,d\},\{e,f,a\}\rangle$,
$F = \langle \{a,b,c,d,e\},\phi,\{e,f,a,d,c,b\}\rangle$

We can compute the probabilities for NCSs by the following:

$$NP(A) = \left\langle \frac{4}{6},\frac{1}{6},\frac{1}{6}\right\rangle,\ NP(D) = \left\langle \frac{2}{6},\frac{2}{6},\frac{2}{6}\right\rangle,\ NP(B) = \left\langle \frac{3}{6},\frac{1}{6},\frac{1}{6}\right\rangle,\ NP(C) = \left\langle \frac{2}{6},\frac{2}{6},\frac{3}{6}\right\rangle,$$
$$NP(E) = \left\langle \frac{4}{6},\frac{2}{6},\frac{3}{6}\right\rangle,\ NP(F) = \left\langle \frac{5}{6},0,\frac{6}{6}\right\rangle,$$

**Remark 3.2**

The probabilities of a neutrosophic crisp set are neutrosophic sets.

**Example 3.3**

Let $X = \{a,b,c,d\}$ , $A = \langle \{a,b\},\{c\},\{d\}\rangle$ , $B = \langle \{a\},\{c\},\{d,b\}\rangle$ are NCS-Type I on X and $U_1 = \langle \{a,b\},\{c,d\},\{a,d\}\rangle$, $U_2 = \langle \{a,b,c\},\{c\},\{d\}\rangle$ are NCS-Type III on X, then we can find the following operations

1) Union, intersection, complement, deference and its probabilities

a)Type1: $A \cap B = \langle \{a\},\{c\},\{d,b\}\rangle$ , $NP(A \cap B) = \langle 0.25,0.25,0.5\}\rangle$ and Type 2,3: $A \cap B = \langle \{a\},\{c\},\{d,b\}\rangle$ , $NP(A \cap B) = \langle 0.25,0.25,0.5\}\rangle$.

2) $NP(A - B)$ may be equals

Type1: $NP(A - B) =< 0.25,0,0 >$ , Type 2: $NP(A - B) =< 0.25,0,0 >$ , Type 3: $NP(A - B) =< 0.25,0,0 >$ ,

b) Type 2: $A \cup B = \langle \{a,b\},\{c\},\{d\} \rangle$, $NP(A \cup B) = \langle 0.5,0.25,0.25\} \rangle$ and Type 2: $A \cup B = \langle \{a.b\},\{c\},\{d\} \rangle$ $NP(A \cup B) = \langle 0.5,0.25,0.25\} \rangle$.

c) Type1: $A^c = \langle \{c,d\} \ ,\{a,b,d\},\{a,b,c\} \rangle$ NCS-Type III set on X, $NP(A^c) = \langle 0.5,0.75,0.75 \rangle$.

Type2: $A^c = \langle \{d\} \ ,\{a,b,d\},\{a,b\} \rangle$ NCS-Type III on X, $NP(A^c) = \langle 0.25,0.75,0.5 \rangle$.

Type3: $A^c = \langle \{d\} \ ,\{c\},\{a,b\} \rangle$ NCS-Type III on X, $NP(A^c) = \langle 0.75,0.75,0.5 \rangle$.

d) Type1: $B^c = \langle \{b,c,d\},\{a,b,d\},\{a,c\} \rangle$ be NCS-Type III on X , $NP(B^c) = \langle 0.75,0.75,0.5 \rangle$

Type2: $B^c = \langle \{b,d\},\{c\},\{a\} \rangle$ NCS-Type I on X, and $NP(B^c) = \langle 0.5,0.25,0.25 \rangle$.

Type3: $B^c = \langle \{b,d\},\{a,b,d\},\{a\} \rangle$ NCS-Type III on X and $NP(B^c) = \langle 0.5,0.75,0.25 \rangle$.

e) Type 1: $U_1 \cup U_2 = \langle \{a,b,c\},\{c,d\},\{a,d\} \rangle$, NCS-Type III, $NP(U_1 \cup U_2) = \langle \{0.75,0.5,0.5 \rangle$,

Type2: $U_1 \cup U_2 = \langle \{a,b,c\},\{c\},\{a,d\} \rangle$, $NP(U_1 \cup U_2) = \langle \{0.75,0.25,0.5 \rangle$,

f) Type1: $U_1 \cap U_2 = \langle \{a,b\},\{c,d\},\{a,d\} \rangle$, NCS-Type III, $NP(U_1 \cap U_2) = \langle 0.5,0.5,0.5 \rangle$,

Type2: $U_1 \cap U_2 = \langle \{a,b\},\{c\},\{a,d\} \rangle$, NCS-Type III, and $NP(U_1 \cap U_2) = \langle 0.5,0.25,0.5 \rangle$,

g) Type 1: ${U_1}^c = \langle \{c,d\},\{a,b\},\{c,b\} \rangle$, NCS-Type III and $NP({U_1}^c) = \langle 0.5,0.5,0.5 \rangle$

Type 2: ${U_1}^c = \langle \{a,d\},\{c,d\},\{a,b\} \rangle$, NCS-Type III and $NP({U_1}^c) = \langle 0.5,0.5,0.5 \rangle$

Type3: ${U_1}^c = \langle \{a,d\},\{a,b\},\{a,b\} \rangle$, NCS-Type III and $NP({U_1}^c) = \langle 0.5,0.5,0.5 \rangle$.

h) Type1: ${U_2}^c = \langle \{d\},\{a,b,d\},\{a,b,c\} \rangle$ NCS-Type III and $NP({U_2}^c) = \langle 0.25,0.75,0.75 \rangle$, Type2: $U^c{}_2 = \langle \{d\},\{c\},\{a,b,c\} \rangle$ NCS-Type III and $NP({U_2}^c) = \langle 0.25,0.25,0.75 \rangle$, Type3: $U^c{}_2 = \langle \{d\},\{a,b,d\},\{a,b,c\} \rangle$ NCS-Type III. $NP({U_2}^c) = \langle 0.25,0.75,0.75 \rangle$.

2) Probabilities for events: $NP(A) = \langle 0.5,0.25,0.25 \rangle$, $NP(B) = \langle 0.25,0.25,0.5 \rangle$, $NP(U_1) = \langle 0.5,0.5,0.5 \rangle$, $NP(U_2) = \langle 0.75,0.25,0.25 \rangle$

, $NP({U_1}^c) = \langle 0.5,0.5,0.5 \rangle$, $NP({U_2}^c) = \langle 0.25,0.75,0.75 \rangle$

e) $(A \cap B)^c =$ = $\langle \{b,c,d\},\{a,b,d\},\{a,c\} \rangle$ be a NCS-Type III. $NP(A \cap B)^c = \langle 0.75,0.75,0.25 \rangle$ be a neutrosophic set.

f) $NP(A)^c \cap NP(B)^c = \langle 0.5,0.75,0.75 \rangle$, $NP(A)^c \cup NP(B)^c = \langle 0.75,0.75,0.5 \rangle$

g) $NP(A \cup B) = NP(A) + NP(B) - NP(A \cap B) = \langle 0.5,0.25,0.25\} \rangle$

s) $NP(A)=\langle 0.5,0.25,0.25\rangle$, $NP(A)^c=\langle 0.5,0.75,0.75\rangle$, $NP(B)=\langle 0.25,0.25,0.5\rangle$, $NP(B^c)=\langle 0.75,0.75,0.5\rangle$

Probabilities for Products

1) The product of two events given by

$A\times B=\langle\{(a,a),(b,a)\},\{(c,c)\},\{(d,d),(d,b)\}\rangle$, and $NP(A\times B)=\langle 2/16, 1/16, 2/16\rangle$

$B\times A=\langle\{(a,a),(a,b)\},\{(c,c)\},\{(d,d),(b,d)\}\rangle$ and $NP(B\times A)=\langle 2/16, 1/16, 2/16\rangle$

$A\times U_1=\langle\{(a,a),(b,a),(a,b),(b,b)\},\{(c,c),(c,d)\},\{(d,d),(d,a)\}\rangle$, and $NP(A\times U_1)=\langle 4/16, 2/16, 2/16\rangle$

$U_1\times U_2=\langle\{(a,a),(b,a),(a,b),(b,b),(a,c),(b,c)\},\{(c,c),(d,c)\},\{(d,d),(a,d)\}\rangle$ and $NP(U_1\times U_2)=\langle 6/16, 2/16, 2/16\rangle$

**Remark 3.3**

The following diagram represents the relation between neutrosophic crisp concepts and neutrosphic sets

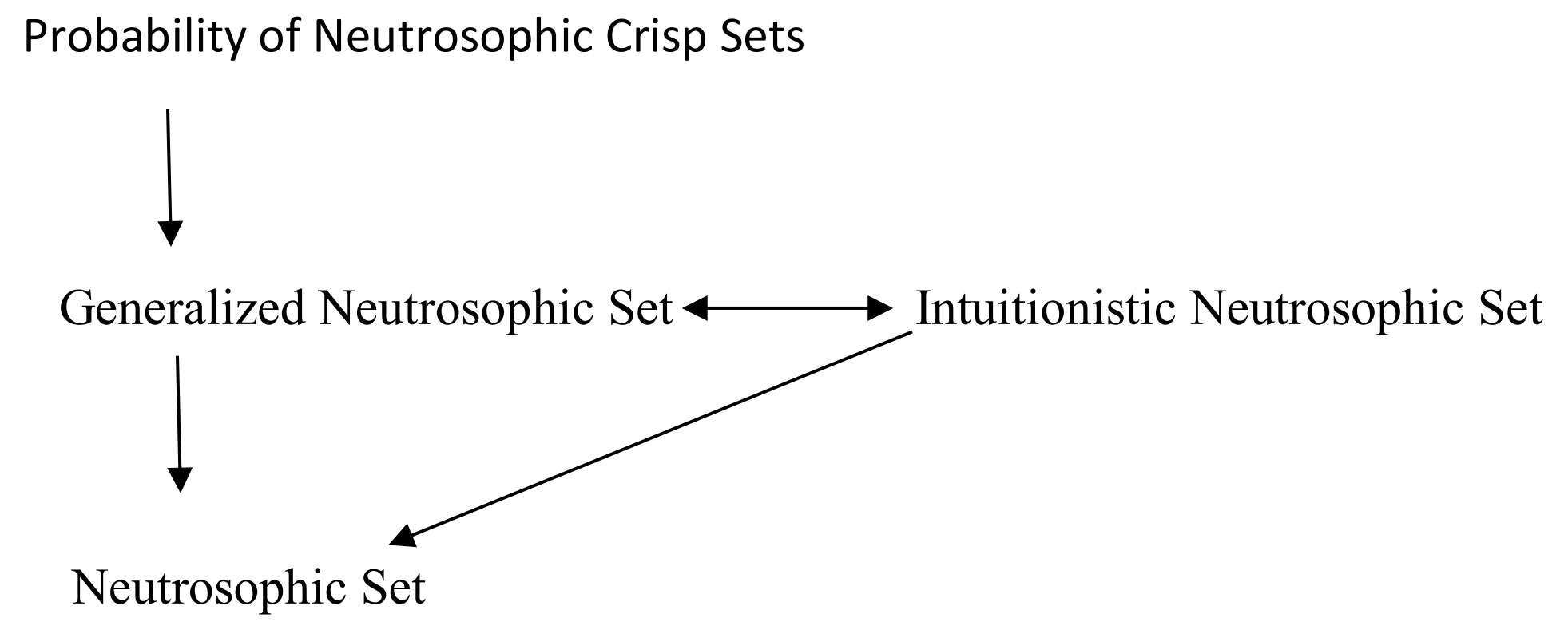

# References

1. K. Atanassov, Intuitionistic fuzzy sets, in V. Sgurev, ed.,Vii ITKRS Session, Sofia(June 1983 central Sci. and Techn. Library, Bulg. Academy of Sciences (1984).
2. K. Atanassov, Intuitionistic fuzzy sets, Fuzzy Sets and Systems 20, 87-96, (1986)
3. K. Atanassov, Review and new result on intuitionistic fuzzy sets, preprint IM-MFAIS-1-88, Sofia, (1988).
4. A. A. Salama, Basic structure of some classes of neutrosophic crisp nearly open sets and possible application to GIS topology, Neutrosophic Sets and Systems, 7 18-22, (2015).
5. A. A. Salama, M.Eisa, S. A. ELhafeez and M. M. Lotfy, Review of recommender systems algorithms utilized in social networks based e-Learning systems & neutrosophic system, Neutrosophic Sets and Systems, 8, 35-44, (2015).
6. A. A. Salama and S. Broumi, Roughness of neutrosophic sets, Elixir Appl. Math.,74, 26833-26837, (2014).
7. A. A. Salama, M. Abdelfattah and M. Eisa, Distances, hesitancy degree and flexible querying via neutrosophic sets, International Journal of Computer Applications, 101(10), 1-12, (2014)
8. M. M. Lofty, A. A. Salama, H. A. El-Ghareeb and M. A. El-dosuky, Subject recommendation using Ontology for computer science ACM curricula, International Journal of Information Science and Intelligent System, 3, 199-205, (2014).

9. A.A. Salama, H. A. El-Ghareeb, Ayman. M. Maine and F. Smarandache. Introduction to develop some software programs for dealing with neutrosophic sets, Neutrosophic Sets and Systems, 4, 51-52, (2014).
10. A. A. Salama, F. Smarandache, and M. Eisa. Introduction to image processing via neutrosophic technique, Neutrosophic Sets and Systems, 5, 59-63, (2014).
11. S. A. Alblowi, A.A. Salama and M. Eisa, New concepts of neutrosophic sets, International Journal of Mathematics and Computer Applications Research (IJMCAR), 4 (1), 59-66, (2014).
12. A. A. Salama, M. Eisa and M. M. Abdelmoghny, Neutrosophic relations database, International Journal of Information Science and Intelligent System, 3(1),33-46, (2014).
13. A. A. Salama, H.A. El-Ghareeb, A.M. Manie and M. M. Lotfy, Utilizing neutrosophic set in social network analysis e-Learning systems, International Journal of Information Science and Intelligent System, 3(2), 61-72, (2014).
14. I. M. Hanafy, A.A. Salama and K. Mahfouz, Correlation of neutrosophic data, International Refereed Journal of Engineering and Science (IRJES) , 1(2), 39-33, (2012).
15. I.M. Hanafy, A.A. Salama and K.M. Mahfouz,, Neutrosophic classical events and its probability, International Journal of Mathematics and Computer Applications Research(IJMCAR), 3(1), 171-178, (2013).
16. A. A. Salama and S.A. Alblowi, Generalized neutrosophic set and generalized neutrosophic topological spaces,Journal Computer Sci. Engineering, 2 (7), 129-132, (2012).
17. A. A. Salama and S. A. Alblowi, Neutrosophic set and neutrosophic topological spaces, ISOR J. Mathematics, 3 (3),31-35, (2012).
18. A. A. Salama, Neutrosophic crisp point & neutrosophic crisp ideals, Neutrosophic Sets and Systems, 1(1), 50-54, (2013).
19. A. A. Salama and F. Smarandache, Filters via neutrosophic crisp sets, Neutrosophic Sets and Systems, 1(1),34-38, (2013).
20. A.A. Salama, and H. Elagamy, Neutrosophic filters, International Journal of Computer Science Engineering and Information Technology Reseearch (IJCSEITR), 3 (1), 307-312, (2013).
21. A. A. Salama, F.Smarandache and Valeri Kroumov, Neutrosophic crisp Sets & Neutrosophic crisp Topological Spaces, Neutrosophic Sets and Systems, Vol.(2), pp25-30. (2014)
22. A. A. Salama, Mohamed Eisa and M. M. Abdelmoghny, Neutrosophic relations database, International Journal of Information Science and Intelligent System, 3(2): 33-46, (2014).
23. A. A. Salama , Florentin Smarandache and S. A. ALblowi, New Neutrosophic crisp topological concepts, Neutrosophic Sets and Systems, 2, 25-30 (2014) .
24. A. A. Salama, Said Broumi and Florentin Smarandache, Neutrosophic Crisp Open Set and Neutrosophic Crisp Continuity via Neutrosophic Crisp Ideals, I.J. Information Engineering and Electronic Business, 3, 1-8, (2014).
25. A. A. Salama, S. Broumi and F. Smarandache, Some types of neutrosophic crisp sets and neutrosophic crisp relations, I.J. Information Engineering and Electronic Business, (2014).
26. A.A.Salama, Haithem A. El-Ghareeb, Ayman. M. Maine and F. Smarandache. Introduction to develop some software programes for dealing with neutrosophic sets, Neutrosophic Sets and Systems, 3,51-52, (2014).
27. A.A. Salama, F. Smarandache and S.A. Alblowi. The characteristic function of a neutrosophic set, Neutrosophic Sets and Systems, 3,14-18, (2014).
28. A. A. Salama, M. Abdelfattah and M. Eisa, Distances, hesitancy degree and flexible querying via neutrosophic sets, International Journal of Computer Applications, 4(3), 2014.
29. A. A. Salama , F. Smarandache and Valeri Kroumov, Neutrosophic closed set and continuouse functions, Neutrosophic Sets and Systems, (2014) (Accepted).
30. A. A. Salama, S. Broumi and F. Smarandache, Some types of neutrosophic crisp sets and neutrosophic crisp relations, I.J. Information Engineering and Electronic Business, 2014
31. A. A. Salama, Florentin Smarandache, Neutrosophic ideal theory neutrosophic local function and generated neutrosophic topology, In Neutrosophic Theory and Its Applications. Collected Papers, 1, EuropaNova, Bruxelles, 213-218, (2014).

32. M. E. Abd El-Monsef, A.A. Nasef, A. A. Salama, Extensions of fuzzy ideals, Bull. Calcutta Math. Soc. 92 (3), 181-188 (2000).
33. M.E. Abd El-Monsef, A.A. Nasef, A.A. Salama, Some fuzzy topological operators via fuzzy ideals, Chaos Solitons Fractals, 12 (13), 2509-2515 (2001).
34. M. E. Abd El-Monsef, A. A. Nasef, A. A. Salama, Fuzzy L-open sets and fuzzy L-continuous functions, Analele Universitatii de Vest din Timisoara, Seria Matematica-Informatica, 40 (2), 3-13, (2002).
35. I. M. Hanafy and A.A. Salama,"A unified framework including types of fuzzy compactness" Conference Topology and Analysis in Applications Durban, 12-16 July, 2004. School of Mathematical Sciences, UKZN.
36. A.A. Salama," Fuzzy Hausdorff spaces and fuzzy irresolute functions via fuzzy ideals" V Italian-Spanish Conference on General Topology and its Applications June 21-23, 2004 Almeria, Spain
37. M.E. Abdel Monsef, A. Kozae, A. A. Salama and H. Elagamy, "Fuzzy Ideals and Bigranule Computing" 20th conference of topology and its Applications 2007, Port Said, Univ., Egypt .
38. A.A. Salama," Intuitionistic Fuzzy Ideals Theory and Intuitionistic Fuzzy Local Functions" CTAC'08 the 14th Biennial Computational Techniques and Applications Conference13–16th July 2008. Australian National University, Canberra, ACT, Australia.
39. A.A. Salama, Fuzzy Bitopological Spaces Via Fuzzy Ideals, Blast 2008, August 6-10, (2008), University of Denver, Denver, CO, USA.
40. A.A. Salama, A new form of fuzzy compact spaces and related topics via fuzzy idealization, Journal of fuzzy System and Mathematics, 24 (2),33-39, (2010).
41. A. A. Salama and A. Hassan, On fuzzy regression model, the Egyptian Journal for commercial Studies, 34(4), 305-319 (2010).
42. A.A. Salama and S.A. Alblowi, Neutrosophic set theory and neutrosophic topological ideal spaces The First International Conference on Mathematics and Statistics (ICMS'10) to be held at the American University
43. A.A. Salama, A new form of fuzzy Hausdorff space and related topics via fuzzy idealization, IOSR Journal of Mathematics (IOSR-JM ), 3(5), 01-04, (2012) .
44. A . A. Salama and Smarandache, Neutrosophic crisp set theory, 2015 USA Book , Educational. Education Publishing 1313 Chesapeake, Avenue, Columbus, Ohio 43212,
45. M. E. Abd El-Monsef, A. M. Kozae, A.A. Salama and H. Elagamy, Fuzzy biotopolgical ideals theory", IOSR Journal of Computer Engineering( IOSRJCE), 6 (4), 01-05, (2012).
46. I. M. Hanafy, A.A. Salama , M. Abdelfattah and Y. Wazery, Security in Mant based on Pki using fuzzy function, IOSR Journal of Computer Engineering, 6(3), 54-60, (2012).
47. M. E. Abd El-Monsef, A. Kozae, A.A. Salama, and H. M. Elagamy, Fuzzy pairwise L-Open sets andfuzzy pairwise L-continuous functions, International Journal of Theoretical and Mathematical Physics, 3(2), 69-72, (2013).
48. F. Smarandache, Neutrosophy and neutrosophic logic, First International Conference on Neutrosophy , Neutrosophic Logic, Set, Probability, and Statistics University of New Mexico, Gallup, NM 87301, USA (2002).
49. F. Smarandache, A unifying field in logics: Neutrosophic logic. Neutrosophy, neutrosophic crisp Set, neutrosophic probability. American Research Press, Rehoboth, NM, (1999).
50. F. Smarandache, Neutrosophic set, a generalization of the intuituionistics fuzzy sets, Inter. J. Pure Appl. Math., 24, 287 – 297, (2005).
51. Florentin Smarandach (2013), INTRODUCTION TONEUTROSOPHIC MEASURE, NEUTROSOPHIC INTEGRAL, AND NEUTROSOPHIC PROBABILITY http://fs.gallup.unm.edu/eBooks-otherformats.htm EAN: 9781599732534.
52. M. Bhowmik and M. Pal. Intuitionistic Neutrosophic Set Relations and Some of its Properties, Journal of Information and Computing Science, 5(3), 183-192, ((2010).
53. L.A. Zadeh, Fuzzy sets, Inform and Control , 8, 338-353, .(1965).

# TOPOLOGY

**FRANCISCO GALLEGO LUPIAÑEZ**

Dept. of Mathematics, Univ. Complutense, 28040 Madrid, SPAIN. E-mail: fg lupianez@mat.ucm.es

# On neutrosophic sets and topology

**Abstract:**

Recently, F.Smarandache generalized the Atanassov's intuitionistic fuzzy sets and other kinds of sets to neutrosophic sets. Also, this author defined the notion of neutrosophic topology on the non-standard interval. One can expect some relation between the intuitionistic fuzzy topology on an IFS and the neutrosophic topology. We show in this work that this is false.

## 1 ON NEUTROSOPHIC TOPOLOGY

### 1.1. Introduction.

The neutrosophic logic is a formal frame trying to measure the truth, indeterminacy, and falsehood.

Smarandache [36] remarks the differences between neutrosophic logic (NL) and intuitionistic fuzzy logic (IFL) and the corresponding neutrosophic sets and intuitionistic fuzzy sets. The main differences are:

a) Neutrosophic Logic can distinguish between absolute truth (that is an unalterable and permanent fact), and relative truth (where facts may vary depending on the circumstances), because

NL(absolute truth)=$1^+$ while NL(relative truth)=1. This has obvious application in philosophy. That's why the unitary standard interval $[0,1]$ used in IFL has been extended to the unitary non-standard interval $]^-0,1^+[$ in NL.

Similar distinctions for absolute or relative falsehood, and absolute or relative indeterminacy are allowed in NL.

b) In NL there is no restriction on $T, I, F$ other than they are subsets of $]^-0,1^+[$, thus:

$^-0 \leq \inf T + \inf I + \inf F \leq \sup T + \sup I + \sup F \leq 3^+$.

This non-restriction allows paraconsistent, dialetheist, and incomplete information to be characterized in NL (i.e. the sum of all three components if they are defined as points, or sum of superior limits of all three components if they are defined as subsets can be $> 1$, for paraconsistent information coming from different sources, or $< 1$ for incomplete information), while that information can not be described in IFL because in IFL the components $T$ (truth), $I$ (indeterminacy), $F$ (falsehood) are restricted either to $t + i + f = 1$ if $T, I, F$ are all reduced to the points $t$,$i$, $f$ respectively, or to $\sup T + \sup I + \sup F = 1$ if $T, I, F$ are subsets of $[0,1]$.

c) In NL the components $T, I, F$ can also be non-standard subsets included in the unitary non-standard interval $]^-0,1^+[$, not only standard subsets, included in the unitary standard interval $[0,1]$ as in IFL.

In various recent papers [35,38,39,40], F. Smarandache generalizes intuitionistic fuzzy sets (IFSs) and other kinds of sets to neutrosophic sets (NSs). In [39] some distinctions between NSs and IFSs are underlined.

The notion of intuitionistic fuzzy set defined by K.T. Atanassov [1] has been applied by Çoker [8] for study intuitionistic fuzzy topological spaces. This concept has been developed by many authors (Bayhan and Çoker[6], Çoker, [7,8], Çoker and Eş [9], Eş and Çoker[12], Gürçay, Çoker and Eş[13], Hanafy [14], Hur, Kim and Ryou [15], Lee and Lee [16]; Lupiáñez [17-21], Turanh and Çoker [41]).

A few years ago raised some controversy over whether the term "intuitionistic fuzzy set" was appropriate or not (see [11] and [4]). At present, it is customary to speak of "Atanassov' intuitionistic fuzzy set"

F. Smarandache also defined the notion of neutrosophic topology on the non-standard interval [35].

One can expect some relation between the inuitionistic fuzzy topology on an IFS and the neutrosophic topology. We show in this chapter that this is false. Indeed, the complement of an IFS $A$ is not the complement of $A$ in the neutrosophic operation, the union and the intersection of IFSs do not coincide with the corresponding operations for NSs, and finally an intuitionistic fuzzy topology is not necessarily a neutrosophic topology.

Clearly, for their various applications to many areas of knowledge, including philosophy, religion, sociology, .. (see [5,40,42]), the Atanassov' intuitionistic fuzzy sets and the neutrosophic sets are notions that use knowledge-based techniques to support human decision-making, learning and action.

### 1.2. Basic definitions.

First, we present some basic definitions:

**Definition 1** *Let $X$ be a non-empty set. An intuitionistic fuzzy set (IFS for short) $A$, is an object having the form $A = \{< x, \mu_A, \gamma_A > /x \in X\}$ where the functions $\mu_A : X \to I$ and $\gamma_A : X \to I$ denote the degree of membership (namely $\mu_A(x)$) and the degree of nonmembership (namely $\gamma_A(x)$) of each element $x \in X$ to the set $A$, respectively, and $0 \leq \mu_A(x) + \gamma_A(x) \leq 1$ for each $x \in X$. [1].*

**Definition 2** *Let $X$ be a non-empty set, and the IFSs $A = \{< x, \mu_A, \gamma_A > |x \in X\}$, $B = \{< x, \mu_B, \gamma_B > |x \in X\}$. Let*

$\overline{A} = \{< x, \gamma_A, \mu_A > |x \in X\}$

$A \cap B = \{< x, \mu_A \wedge \mu_B, \gamma_A \vee \gamma_B > |x \in X\}$

$A \cup B = \{< x, \mu_A \vee \mu_B, \gamma_A \wedge \gamma_B > |x \in X\}$.*[3].*

**Definition 3** *Let $X$ be a non-empty set. Let $0_\sim = \{< x, 0, 1 > |x \in X\}$ and $1_\sim = \{< x, 1, 0 > |x \in X\}$.[8].*

**Definition 4** *An intuitionistic fuzzy topology (IFT for short) on a non-empty set $X$ is a family $\tau$ of IFSs in $X$ satisfying:*

*(a) $0_\sim$,$1_\sim \in \tau$,*

*(b) $G_1 \cap G_2 \in \tau$ for any $G_1, G_2 \in \tau$,*

*(c) $\cup G_j \in \tau$ for any family $\{G_j|j \in J\} \subset \tau$.*

*In this case the pair $(X, \tau)$ is called an intuitionistic fuzzy topological space (IFTS for short) and any IFS in $\tau$ is called an intuitionistic fuzzy open set (IFOS for short) in $X$. [8].*

**Definition 5** *Let $T$, $I$,$F$ be real standard or non-standard subsets of the non-standard unit interval $]^-0, 1^+[$, with*

$\sup T = t_{\sup}$ , $\inf T = t_{\inf}$

$\sup I = i_{\sup}$ , $\inf I = i_{\inf}$

$\sup F = f_{\sup}$ , $\inf F = f_{\inf}$ *and* $n_{\sup} = t_{\sup} + i_{\sup} + f_{\sup}$ $n_{\inf} = t_{\inf} + i_{\inf} + f_{\inf}$,

*$T$, $I$,$F$ are called neutrosophic components. Let $U$ be an universe of discourse, and $M$ a set included in $U$. An element $x$ from $U$ is noted with respect to the set $M$ as $x(T, I, F)$ and belongs to $M$ in the following way: it is t% true in the set, i% indeterminate (unknown if it is) in the set, and f% false, where t varies in $T$, $i$ varies in $I$, $f$ varies in $F$. The set $M$ is called a neutrosophic set (NS). [40].*

***Remark.*** *All IFS is a NS.*

**Definition 6** *Let $S_1$ and $S_2$ be two (unidimensional) real standard or non-standard subsets, then we define:*

$S_1 \oplus S_2 = \{x|x = s_1 + s_2$, *where* $s_1 \in S_1$ *and* $s_2 \in S_2\}$,

$S_1 \ominus S_2 = \{x|x = s_1 - s_2$, *where* $s_1 \in S_1$*and* $s_2 \in S_2\}$,

$S_1 \odot S_2 = \{x|x = s_1 \cdot s_2$, *where* $s_1 \in S_1$*and* $s_2 \in S_2\}$. *[36].*

**Definition 7** *One defines, with respect to the sets $A$ an $B$ over the universe $U$:*
*1. Complement: if* $x(T_1, I_1, F_1) \in A$, *then*
$x(\{1^+\} \ominus T_1, \{1^+\} \ominus I_1, \{1^+\} \ominus F_1) \in C(A)$.
*2. Intersection: if* $x(T_1, I_1, F_1) \in A$, $x(T_2, I_2, F_2) \in B$, *then*
$x(T_1 \odot T_2, I_1 \odot I_2, F_1 \odot F_2) \in A \cap B$.
*3.Union: if* $x(T_1, I_1, F_1) \in A$, $x(T_2, I_2, F_2) \in B$, *then*
$x(T_1 \oplus T_2 \ominus T_1 \odot T_2, I_1 \oplus I_2 \ominus I_1 \odot I_2, F_1 \oplus F_2 \ominus F_1 \odot F_2) \in A \cup B$.
*[40].*

**1.3. Results.**

**Proposition 1.** Let $A$ be an IFS in $X$, and $j(A)$ be the corresponding NS. We have that the complement of $j(A)$ is not necessarily $j(\overline{A})$.

**Proof.** If $A =< x, \mu_A, \gamma_A >$ is $x(\mu_A(x), 1 - \mu_A(x) - \nu_A(x), \nu_A(x)) \in j(A)$.
Then ,
for $0_\sim =< x, 0, 1 >$ is $x(0, 0, 1) \in j(0_\sim)$
for $1_\sim =< x, 1, 0 >$ is $x(1, 0, 0) \in j(1_\sim)$
and for $\overline{A} =< x, \gamma_A, \mu_A >$ is $x(\gamma_A(x), 1 - \mu_A(x) - \nu_A(x), \mu_A(x)) \in j(\overline{A})$.
Thus, $1_\sim = \overline{0_\sim}$ and $j(1_\sim) \neq C(j(0_\sim))$ because $x(1, 0, 0, ) \in j(1_\sim)$ but $x(\{1^+\}, \{1^+\}, \{0^+\}) \in C(j(0_\sim))$.

**Proposition 2.** Let $A$ and $B$ be two IFSs in $X$,and $j(A)$ and $j(B)$ be the corresponding NSs. We have that $j(A) \cup j(B)$ is not necessarily $j(A \cup B)$, and $j(A) \cap j(B)$ is not necessarily $j(A \cap B)$ .

**Proof.** Let $A =< x, 1/2, 1/3 >$ and $B =< x, 1/2, 1/2 >$ (i.e. $\mu_A$, $\nu_A$, $\mu_B$, $\nu_B$ are constant maps).
Then, $A \cup B =< x, \mu_A \vee \mu_B, \gamma_A \wedge \gamma_B >=< x, 1/2, 1/3 >$ and $x(1/2, 1/6, 1/3) \in j(A \cup B)$. On the other hand, $x(1/2, 1/6, 1/3) \in j(A), x(1/2, 0, 1/2) \in j(B), x(1, 1/6, 5/6) \in j(A) \oplus j(B), x(1/4, 0, 1/6) \in j(A) \odot j(B)$ and $x(3/4, 1/6, 2/3) \in j(A) \cup j(B)$ .Thus $j(A \cup B) \neq j(A) \cup j(B)$.
Analogously, $A \cap B =< x, \mu_A \wedge \mu_B, \gamma_A \vee \gamma_B >=< x, 1/2, 1/2 >$ and $x(1/2, 0, 1/2) \in j(A \cap B)$, but $x(1/4, 0, 1/6) \in j(A) \cap j(B)$.Thus, $j(A \cap B) \neq j(A) \cap j(B)$.

**Definition 8** *Let's construct a neutrosophic topology on* $NT = ]^-0, 1^+[$, *considering the associated family of standard or non-standard subsets included in $NT$, and the empty set which is closed under set union and finite intersection neutrosophic. The interval $NT$ endowed with this topology forms a neutrosophic topological space. [35].*

**Proposition 3.** Let $(X, \tau)$ be an intuitionistic fuzzy topological space. Then, the family $\{j(U) | U \in \tau\}$ is not necessarily a neutrosophic topology.

**Proof.** Let $\tau = \{1_\sim, 0_\sim, A\}$ where $A =< x, 1/2, 1/2 >$ then $x(1, 0, 0) \in j(1_\sim)$, $x \in (0, 0, 1) \in j(0_\sim)$ and $x(1/2, 0, 1/2) \in j(A)$. Thus $\{j(1_\sim), j(0_\sim), j(A)\}$ is not a neutrosophic topology, because this family is not closed by finite intersections, indeed, $x(1/2, 0, 0) \in j(1_\sim) \cap j(A)$, and this neutrosophic set is not in the family.

## 2 OTHER NEUTROSOPHIC TOPOLOGIES

### 2.1. Introduction.

F. Smarandache also defined various notions of neutrosophic topologies on the non-standard interval [35,40].

One can expect some relation between the intuitionistic fuzzy topology on an IFS and the neutrosophic topology. We show in this chapter that this is false. Indeed, the union and the intersection of IFSs do not coincide with the corresponding operations for NSs, and an intuitionistic fuzzy topology is not necessarilly a neutrosophic topology on the non-standard interval, in the various senses defined by Smarandache.

### 2.2. Basic definitions.

First, we present some basic definitions:

**Definition 9** *Let $J \in \{T, I, F\}$ be a component. Most known N-norms are:*
*The* ***algebraic product N-norm****:* $N_{n-a\lg ebraic}J(x,y) = x \cdot y$
*The* ***bounded N-norm****:* $N_{n-bounded}J(x,y) = \max\{0, x+y-1\}$
*The* ***default (min) N-norm****:* $N_{n-\min}J(x,y) = \min\{x,y\}$
$N_n$ *represent the intersection operator in neutrosophic set theory. Indeed* $x \wedge y = (T_\wedge, I_\wedge, F_\wedge)$*.*
*[40]*

**Definition 10** *Let $J \in \{T, I, F\}$ be a component. Most known N-conorms are:*
*The* ***algebraic product N-conorm****:* $N_{c-a\lg ebraic}J(x,y) = x + y - x \cdot y$
*The* ***bounded N-conorm****:* $N_{c-bounded}J(x,y) = \min\{1, x+y\}$
*The* ***default (max) N-conorm****:* $N_{c-\max}J(x,y) = \max\{x,y\}$
$N_c$ *represent the union operator in neutrosophic set theory.Indeed* $x \vee y = (T_\vee, I_V, F_\vee)$
*[40]*

### 2.3. Results.

**Proposition 1.** Let $A$ and $B$ be two IFSs in $X$, and $j(A)$ and $j(B)$ be the corresponding NSs. We have that $j(A)\cup j(B)$ is not necessarily $j(A \cup B)$, and $j(A)\cap j(B)$ is not necessarily $j(A \cap B)$, for any of three definitions of intersection of NSs.

**Proof.** Let $A =< x, 1/2, 1/3 >$ and $B =< x, 1/2, 1/2 >$ (i.e. $\mu_A$, $\nu_A$, $\mu_B$, $\nu_B$ are constant maps).

Then, $A \cup B =< x, \mu_A \vee \mu_B, \gamma_A \wedge \gamma_B >=< x, 1/2, 1/3 >$ and $x(1/2, 1/6, 1/3) \in j(A \cup B)$. On the other hand, $x(1/2, 1/6, 1/3) \in j(A), x(1/2, 0, 1/2) \in j(B)$.

Then, we have that:

1) for the union operator defined by the algebraic product N-conorm $x(3/4, 1/6, 2/3) \in j(A)\cup j(B)$ .
2) for the union operator defined by the bounded N-conorm $x(1, 1/6, 5/6) \in j(A)\cup j(B)$ .
3) for the union operator defined by the default (max) N-conorm $x(1/2, 1/6, 1/2) \in j(A)\cup j(B)$ .

Thus $j(A \cup B) \neq j(A)\cup j(B)$,with the three definitions.

Analogously, $A \cap B =< x, \mu_A \wedge \mu_B, \gamma_A \vee \gamma_B >=< x, 1/2, 1/2 >$ and $x(1/2, 0, 1/2) \in j(A \cap B)$.

And, we have that:

1) for the intersection operator defined by the algebraic product N-norm $x(1/4, 0, 1/6) \in j(A)\cap j(B)$ .
2) for the intersection operator defined by the bounded N-norm $x(0, 0, 0) \in j(A)\cap j(B)$ .
3) for the intersection operator defined by the default (min) N-norm $x(1/2, 0, 1/3) \in j(A)\cap j(B)$ .

Thus $j(A \cap B) \neq j(A)\cap j(B)$,with the three definitions.

**Definition 11** *Let's construct a neutrosophic topology on $NT = ]^-0, 1^+[$, considering the associated family of standard or non-standard subsets included in $NT$, and the empty set which is closed under set union and finite intersection neutrosophic. The interval $NT$ endowed with this topology forms a neutrosophic topological space. There exist various notions of neutrosophic topologies on $NT$ , defined by using various N-norm/N-conorm operators. [35, 40].*

**Proposition 2.** Let $(X, \tau)$ be an intuitionistic fuzzy topological space. Then, the family $\{j(U)|U \in \tau\}$ is not necessarily a neutrosophic topology on $NT$ (in the three defined senses).

**Proof.** Let $\tau = \{1_\sim, 0_\sim, A\}$ where $A =< x, 1/2, 1/2 >$ then $x(1, 0, 0) \in j(1_\sim)$, $x \in (0, 0, 1) \in j(0_\sim)$ and $x(1/2, 0, 1/2) \in j(A)$. Thus $\tau^* = \{j(1_\sim), j(0_\sim), j(A)\}$ is not a neutrosophic topology, because this family is not closed by finite intersections, for any neutrosophic topology on $NT$. Indeed,

1) For the intersection defined by the algebraic product N-norm, we have that $x(1/2, 0, 0) \in j(1_\sim) \cap j(A)$, and this neutrosophic set is not in the family $\tau^*$.

2) For the intersection defined by the bounded N-norm, we have also that $x(1/2, 0, 0) \in j(1_\sim) \cap j(A)$, and this neutrosophic set is not in the family $\tau^*$.

3) For the intersection defined by the default (min) N-norm, we have also that $x(1/2, 0, 0) \in j(1_\sim) \cap j(A)$, and this neutrosophic set is not in the family $\tau^*$.

# 3 INTERVAL NEUTROSOPHIC SETS AND TOPOLOGY

## 3.1. Introduction.

Also, Wang, Smarandache, Zhang, and Sunderraman [42] introduced the notion of interval neutrosophic set, which is an instance of neutrosophic set and studied various properties. We study in this chapter relations between interval neutrosophic sets and topology.

## 3.2. Basic definitions.

First, we present some basic definitions. For definitions on non-standard Analysis, see [33] :

**Definition 12** *Let $X$ be a space of points (objects) with generic elements in $X$ denoted by $x$. An interval neutrosophic set (INS) $A$ in $X$ is characterized by thuth-membership function $T_A$, indeteminacy-membership function $I_A$ and falsity-membership function $F_A$. For each point $x$ in $X$, we have that $T_A(x)$, $I_A(x)$, $F_A(x) \in [0,1]$. [42].*

***Remark.*** *All INS is clearly a NS.*

*When $X$ is continuous, an INS $A$ can be written as*
$A = \int_X \langle T(x), I(x), F(x)\rangle /x, \quad x \in X$
*When $X$ is discrete, an INS $A$ can be written as*
$A = \sum_{i=1}^{n} \langle T(x_i), I(x_i)F(x_i)\rangle /x_i \ , \quad x_i \in X$

**Definition 13** *a) An interval neutrosophic set $A$ is empty if* $\inf T_A(x) = \sup T_A(x) = 0$, $\inf I_A(x) = \sup I_A(x) = 1$, $\inf F_A(x) = \sup F_A(x) = 0$ *for all $x$ in $X$.*
*b) Let $\underline{0} =< 0,1,1 >$ and $\underline{1} =< 1,0,0 >$ .[42].*

**Definition 14** *(Complement) Let $C_N$ denote a neutrosophic complement of $A$.*
*Then $C_N$ is a function $C_N : N \to N$ and $C_N$ must satisfy at least the following three axiomatic requirements:*
*1. $C_N(\underline{0}) = \underline{1}$ and $C_N(\underline{1}) = \underline{0}$ (boundary conditions).*
*2. Let $A$ and $B$ be two interval neutrosophic sets defined on $X$, if $A(x) \leq B(x)$, then $C_N(A(x)) \geq C_N(B(x))$, for all $x$ in $X$. (monotonicity).*
*3. Let $A$ be an interval neutrosophic set defined on $X$, then $C_N(C_N(A(x))) = A(x)$, for all $x$ in $X$. (involutivity).[42].*
***Remark.*** *There are many functions which satisfy the requirement to be the complement operator of interval*
*neutrosophic sets. Here we give one example.*

**Definition 15** *(Complement $C_{N_1}$) The complement of an interval neutrosophic set $A$ is denoted by $\bar{A}$ and is defined by*
$T_{\bar{A}}(x) = F_A(x)$;
$\inf I_{\bar{A}}(x) = 1 - \sup I_A(x)$;
$\sup I_{\bar{A}}(x) = 1 - \inf I_A(x)$;
$F_{\bar{A}}(x) = T_A(x)$; *for all $x$ in $X$.*

**Definition 16** *(N-norm) Let $I_N$ denote a neutrosophic intersection of two interval neutrosophic sets $A$ and $B$. Then $I_N$ is a function $I_N : N \times N \to N$ and $I_N$ must satisfy at least the following four axiomatic requirements:*
*1. $I_N(A(x), \underline{1}) = A(x)$, for all $x$ in $X$. (boundary condition).*
*2. $B(x) \leq C(x)$ implies $I_N(A(x), B(x)) \leq I_N(A(x), C(x))$, for all $x$ in $X$. (monotonicity).*
*3. $I_N(A(x), B(x)) = I_N(B(x), A(x))$, for all $x$ in $X$. (commutativity).*
*4. $I_N(A(x), I_N(B(x), C(x))) = I_N(I_N(A(x), B(x)), C(x))$, for all $x$ in $X$. (associativity).[42].*
***Remark.*** *Here we give one example of intersection of two interval neutrosophic sets which satis es above N-norm axiomatic requirements. Other diferent definitions can be given for different applications*

**Definition 17** *(Intersection $I_{N_1}$) The intersection of two interval neutrosophic sets $A$ and $B$ is an interval neutrosophic set $C$, written as $C = A \cap B$, whose truth-membership, indeterminacy-membership, and false-membership are related to those of $A$ and $B$ by*
$\inf T_C(x) = \min(\inf T_A(x); \inf T_B(x))$,
$\sup T_C(x) = \min(\sup T_A(x); \sup T_B(x))$,
$\inf I_C(x) = \max(\inf I_A(x); \inf I_B(x))$,
$\sup I_C(x) = \max(\sup I_A(x); \sup I_B(x))$,
$\inf F_C(x) = \max(\inf F_A(x); \inf F_B(x))$,
$\sup F_C(x) = \max(\sup F_A(x); \sup F_B(x))$; *for all $x$ in $X$.*

**Definition 18** *(N-conorm) Let $U_N$ denote a neutrosophic union of two interval neutrosophic sets $A$ and $B$. Then $U_N$ is a function $U_N : N \times N \to N$*
*and $U_N$ must satisfy at least the following four axiomatic requirements:*
*1. $U_N(A(x), \underline{0}) = A(x)$, for all $x$ in $X$. (boundary condition).*
*2. $B(x) \leq C(x)$ implies $U_N(A(x), B(x)) \leq U_N(A(x), C(x))$, for all $x$ in $X$. (monotonicity).*
*3. $U_N(A(x), B(x)) = U_N(B(x), A(x))$, for all $x$ in $X$. (commutativity).*
*4. $U_N(A(x), U_N(B(x), C(x))) = U_N(U_N(A(x), B(x)), C(x))$, for all $x$ in $X$. (associativity). [42].*
***Remark.*** *Here we give one example of union of two interval neutrosophic sets which satis es above N-conorm axiomatic requirements. Other different definitions can be given for different applications.*

**Definition 19** *(Union $U_{N_1}$) The union of two interval neutrosophic sets $A$ and $B$ is an interval neutrosophic set $C$, written as $C = A \cup B$, whose truth-membership, indeterminacy-membership, and false-membership are related to those of $A$ and $B$ by*
$\inf T_C(x) = \max(\inf T_A(x); \inf T_B(x))$,
$\sup T_C(x) = \max(\sup T_A(x); \sup T_B(x))$,
$\inf I_C(x) = \min(\inf I_A(x); \inf I_B(x))$,
$\sup I_C(x) = \min(\sup I_A(x); \sup I_B(x))$,
$\inf F_C(x) = \min(\inf F_A(x); \inf F_B(x))$,
$\sup F_C(x) = \min(\sup F_A(x); \sup F_B(x))$, *for all $x$ in $X$.*

### 3.3. Results.

**Proposition 1.** Let $A$ be an IFS in $X$, and $j(A)$ be the corresponding INS. We have that the complement of $j(A)$ is not necessarily $j(\overline{A})$.
**Proof.** If $A =< x, \mu_A, \gamma_A >$ is $j(A) =< \mu_A, 0, \gamma_A >$ .
Then ,
for $0_\sim =< x, 0, 1 >$ is $j(0_\sim) = j(< x, 0, 1 >) =< 0, 0, 1 > \neq \underline{0} =< 0, 1, 1 >$
for $1_\sim =< x, 1, 0 >$ is $j(1_\sim) = j(< x, 1, 0 >) =< 1, 0, 0 >= \underline{1}$
Thus, $1_\sim = \overline{0_\sim}$ and $j(1_\sim) = \underline{1} \neq C_N(j(0_\sim))$ because $C_N\ (\underline{1}) = \underline{0} \neq j(0_\sim)$.

**Definition 20** *Let's construct a neutrosophic topology on $NT = ]^-0, 1^+[$, considering the associated family of standard or non-standard subsets included in $NT$, and the empty set which is closed under set union and finite intersection neutrosophic. The interval $NT$ endowed with this topology forms a neutrosophic topological space. [35].*

**Proposition 2.** Let $(X, \tau)$ be an intuitionistic fuzzy topological space. Then, the family of INSs $\{j(U)|U \in \tau\}$ is not necessarily a neutrosophic topology.
**Proof.** Let $\tau = \{1_\sim, 0_\sim, A\}$ where $A =< x, 1/2, 1/2 >$ then $j(1_\sim) = \underline{1}$, $j(0_\sim) =< 0, 0, 1 > \neq \varnothing$ and $j(A) =< 1/2, 0, 1/2 >$. Thus $\{j(1_\sim), j(0_\sim), j(A)\}$ is not a neutrosophic topology, because the empty INS is not in this family.

## 4 NEUTROSOPHIC PARACONSISTENT TOPOLOGY

The history of paraconsistent logic is not very long. It was designed by S. Jaskowski in 1948. Without knowing the work of this author, N.C. A. da Costa, from 1958, using different methods and ideas, began to make statements about this type of logic. After other logicians have developed independently, new systems of paraconsistent logic, as Routley, Meyer, Priest, Asenjo, Sette, Anderson and Benalp, Wolf (with da Costa himself), .... At present there is a thriving movement dedicated to the study of paraconsistent logic in several countries. In the philosophical aspect has meant, in some cases, a real opening of horizons, for example, in the treatment of the paradoxes, in efforts to treat rigorously dialectical thinking, in fact possible to develop a set theory inconsistent. .. Because of this, there is growing interest in understanding the nature and scope.

Jaskowski deductive logic led her to refer to several problems that caused the need for paraconsistent logic:

1) The problem of organizing deductive theories that contain contradictions, as in the dialectic: "The principle that no two contradictory statements are both true and false is the safest of all."

2) To study theories that there are contradictions engendered by vagueness: "The contemporary formal approach to logic increases the accuracy of research in many fields, but it would be inappropriate to formulate the principle of contradiction of Aristotle thus:"*Two contradictory propositions are not true*". We need to add:"*in the same language*"or "*if the words that are part of those have the same meaning*". This restriction is not always found in daily use, and also science, we often use terms that are more or less vague.

3) To study directly some postulates or empirical theories whose basic meanings are contradictory. This applies, for example, the physics at the present stage.

Objectives and method of construction of paraconsistent logics can be mentioned, besides those mentioned by Jaskowski:

1) To study directly the logical and semantic paradoxes, for example, if we directly study the paradoxes of set theory (without trying to avoid them, as it normally is), we need to construct theories of sets of such paradoxes arising, but without being formal antinomies. In this case we need a paraconsistent logic.

2) Better understand the concept of negation.

3) Have logic systems on which to base the paraconsistent theories. For example, set up logical systems for different versions and possibly stronger than standard theories of sets, of dialectics, and of certain physical theories that , perhaps, are inconsistent (some versions of quantum mechanics).

Various authors [31] worked on "paraconsistent Logics", that is, logics where some contradiction is admissible. We remark the theories exposed by Da Costa [10], Routley and other [34], and Peña [29,30].

Smarandache defined also the neutrosophic paraconsistent sets [Sm5] and he proposed a natural definition of neutrosophic paraconsistent topology.

A problem that we consider is the possible relation between this concept of neutrosophic paraconsistent topology and the previous notions of general neutrosophic topology and intuitionistic fuzzy topology. We show in this chapter that neutrosophic paraconsistent topology is not an extension of intuitionistic fuzzy topology.

First, we present some basic definitions:

**Definition 21** *Let $M$ be a non-empty set. A* ***general neutrosophic topology*** *on $M$ is a family $\Psi$ of neutrosophic sets in $M$ satisfying the following axioms:*

*(a)* $0_\sim = x(0,0,1)$ ,$1_\sim = x(1,0,0) \in \Psi$

*(b) If* $A, B \in \Psi$ *, then* $A \cap B \in \Psi$

*(c) If a family* $\{A_j | j \in J\} \subset \Psi$*,then* $\cup A_j \in \Psi$.

*[40]*

**Definition 22** *A neutrosophic set $x(T,I,F)$ is called* ***paraconsistent*** *if* $inf(T) + inf(I) + inf(F) > 1$.*[39]*

**Definition 23** *For neutrosophic paraconsistent sets* $0_- = x(0,1,1)$ *and* $1_- = x(1,1,0)$.*(Smarandache).*

***Remark.*** *If we use the unary neutrosophic negation operator for neutrosophic sets [40],* $n_N(x(T,I,F)) = x(F,I,T)$ *by interchanging the thuth $T$ and falsehood $F$ components, we have that* $n_N(0_-) = 1_-$ .

**Definition 24** *Let $X$ be a non-empty set. A family $\Phi$ of neutrosophic paraconsistent sets in $X$ will called a **neutrosophic paraconsistent topology** if:*

*(a) $0_{-}$ and $1_{-} \in \Phi$*

*(b) If $A, B \in \Phi$, then $A \cap B \in \Phi$*

*(c) Any union of a subfamily of paraconsistent sets of $\Phi$ is also in $\Phi$.*

*(Smarandache).*

***Results.***

**Proposition 1.** The neutrosophic paraconsistent topology is not an extension of intuitionistic fuzzy topology.

**Proof.** We have that $0_{\sim} =< x, 0, 1 >$ and $1_{\sim} =< x, 1, 0 >$ are members of all intuitionistic fuzzy topology, but

$x(0, 0, 1) \in j(0_{\sim}) \neq 0_{-}$, and, $x(1, 0, 0) \in j(1_{\sim}) \neq 1_{-}$.

**Proposition 2.** A neutrosophic paraconsistent topology is not a general neutrosophic topology.

**Proof.** Let the family $\{1_{-}, 0_{-}\}$ . Clearly it is a neutrosophic paraconsistent topology, but $0_{\sim}$,$1_{\sim}$ are not in this family.

**REFERENCES**

[1] K.T.Atanassov :"Intuitionistic fuzzy sets", paper presented at the VII ITKR's Session, Sofia (June 1983).

[2] K.T.Atanassov :" Intuitionistic fuzzy sets", *Fuzzy Sets and Systems* **20** (1986),87-96.

[3] K.T.Atanassov : "Review and new results on intuitionistic fuzzy sets", preprint IM-MFAIS-1-88 (1988), Sofia.

[4] K.T.Atanassov: "Answer to D.Dubois, S.Gottwald, P.Hajek, J. Kacprzyk and H.Prade's paper "Terminological difficulties in fuzzy set theory-The case of "Intuitionistic Fuzzy Sets"", *Fuzzy Sets and Systems* **156** (2005), 496-499.

[5] K.T. Atanassov: *On Intuitionistic Fuzzy Sets Theory,* Springer (Berlin, 2012)

[6] S.Bayhan and D.Çoker : "On $T_1$ and $T_2$ separation axioms in intuitionistic fuzy topological spaces", , *J. Fuzzy Math.***11** (2003), 581-592.

[7] D.Çoker : "An introduction to fuzzy subspaces in intuitionistic fuzzy topological spaces", *J. Fuzzy Math.* **4** (1996), 749-764.

[8] D.Çoker : "An introduction to intuitionistic fuzzy topologial spaces", *Fuzzy Sets and Systems* **88** (1997), 81-89.

[9] D.Çoker and A.H. Eş :"On fuzzy compactness in intuitionistic fuzzy topological spaces", *J. Fuzzy Math.* **3** (1995), 899-909.

[10] N.C.A. Da Costa: "Nota sobre o conceito de contradição", *Soc. Paranense Mat. Anuário* (2) **1(**1958), 6-8.

[11] D.Dubois, S.Gottwald, P.Hajek, J. Kacprzyk, H.Prade: "Terminological difficulties in fuzzy set theory-The case of "Intuitionistic Fuzzy Sets"",
*Fuzzy Sets and Systems* **156** (2005), 485-491.

[12] A.H. Eş and D.Çoker : "More on fuzzy compactness in intuitionistic fuzzy topological spaces", *Notes IFS,* **2** , no1(1996), 4-10.

[13] H.Gürçay, D.Çoker and A.H.Eş : "On fuzzy continuity in intuitionistic fuzzy topological spaces", *J. Fuzzy Math.* **5** (1997), 365-378.

[14] J.H.Hanafy :"Completely continuous functions in intuitionistic fuzzy topological spaces", *Czech, Math. J.***53 (128)** (2003),793-803.

[15] K.Hur, J.H.Kim,and J.H.Ryou : "Intuitionistic fuzzy topologial spaces", *J. Korea Soc. Math. Educ., Ser B* **11** (2004),243-265.

[16] S.J.Lee and E.P. Lee : "The category of intuitionistic fuzzy topological spaces", *Bull. Korean Math. Soc.* **37** (2000), 63-76.

[17] F.G.Lupiañez : "Hausdorffness in intuitionistic fuzzy topological spaces", *J. Fuzzy Math.* **12,** (2004), 521-525.

[18] F.G.Lupiañez : "Separation in intuitionistic fuzzy topological spaces", *Intern. J. Pure Appl. Math.***17** (2004), 29-34.

[19] F.G.Lupiañez : "Nets and filters in intuitionistic fuzzy topological spaces", *Inform. Sci.* **176** (2006), 2396-2404.

[20] F.G.Lupiañez: "On intuitionistic fuzzy topological spaces", *Kybernetes* **35** (2006),743-747.

[21] F.G.Lupiañez : "Covering properties on intuitionistic fuzzy topological spaces", *Kybernetes* **36** (2007), 749-753.

[22] F.G.Lupiañez: "On neutrosophic Topology", *Kybernetes* **37** (2008), 797-800.

[23] F.G. Lupiañez: "Interval neutrosophic sets and Topology", in *Applied and Computational Mathematics*, WSEAS Press (Athens, 2008), pp.110-112.

[24] F.G.Lupiañez: "Interval neutrosophic sets and Topology", *Kybernetes* **38** (2009), 621-624.

[25] F.G.Lupiañez: "On various neutrosophic topologies", in *Recent advances in in Fuzzy Systems*, WSEAS Press (Athens, 2009),pp. 59-62.

[26] F.G.Lupiañez: "On various neutrosophic topologies", *Kybernetes* **38** (2009), 1009-1013.

[27] F.G.Lupiañez: "On neutrosophic paraconsistent topology", *Kybernetes* **39** (2010), 598-601.

[28] F. Miró Quesada: "La Lógica paraconsistente y el problema de la racionalidad de la Lógica", in *Antología de la Lógica en América Latina*, Fundación Banco Exterior (Madrid, 1988), pp. 593-622.

[29] L.Peña: "Dialectical arguments, matters of degree, and paraconsistent Logic", in *Argumentation: perspectives and approaches* (ed. by F.H. van Eemeren and other), Foris Publ.(Dordrecht, 1987), pp. 426-433.

[30] L.Peña: "La defendibilidad lógico-filosófica de teorías contradictorias", in *Antología de la Lógica en América Latina*, Fundación Banco Exterior (Madrid, 1988), pp. 643-676.

[31] G.Priest, R. Routley, and J.Norman (eds.): *Paraconsistent Logic: Essays on the inconsistent*, Philosophia Verlag, (Munich, 1989).

[32] G. Priest and K. Tanaka: "Paraconsistent Logic", in *Stanford Encyclopedia of Philosophy*, http://plato.stanford.edu/entries/logic-paraconsistent/

[33] A. Robinson :*Non-standard Analysis*, North-Holland Publ. (Amsterdam, 1966).

[34] R.Routley, V.Plumwood, R.K.Meyer, and R.T. Brady: *Relevant Logics and their rivals*, Ridgeview, (Atascadero,CA, 1982).

[35] F.Smarandache: "A unifying field in Logics: Neutrosophic Logic", *Multiple-Valued Logic* **8** (2002),385-438.

[36] F.Smarandache: "An introduction to Neutrosophy, neutrosophic Logic, neutrosophic set, and neutrosophic probability and Statistics", in *Proc. First Intern. Conf. Neutrosophy, neutrosophic Logic, neutrosophic set, and neutrosophic probability and Statistics*, Univ. New Mexico, Gallup, 2001, Xiquan (Phoenix, AZ 2002), pp. 5-21.

[37] F. Smarandache: "Definitions derived from neutrosophics (Addenda)", Idem, pp. 63-74.

[38] F.Smarandache: "Definition of Neutrosophic Logic. A generalization of the intuitionistic fuzzy Logic", *Proc. 3rd Conf. Eur. Soc. Fuzzy Logic Tech.* [EUSFLAT, 2003], pp.141-146.

[39] F.Smarandache : "Neutrosophic set. A generalization of the intuitionistic fuzzy set", *Intern. J. Pure Appl. Math.* **24** (2005), 287-297.

[40] F.Smarandache : *A Unifying Field in Logics: Neutrosophic Logic. Neutrosophy, Neutrosophic Set, Neutrosophic Probability and Statistics* (4th ed.) American Research Press (Rehoboth, NM, 2005).

[41] N.Turanh and D.Çoker :"Fuzzy connectedness in intuitionistic fuzzy topological spaces", *Fuzzy Sets and Systems* **116** (2000), 369-375.

[42] H. Wang, F. Smarandache, Y-Q. Zhang, R. Sunderraman: *Interval Neutrosophic Sets and Logic: Theory and Applications in Computing*, Hexis (Phoenix, AZ, 2005)

A.A.SALAMA[1], I.M.HANAFY[1], HEWAYDA ELGHAWALBY[3], M.S.DABASH[4]

1,2,4 Department of Mathematics and Computer Science, Faculty of Sciences, Port Said University, Egypt.
Emails: drsalama44@gmail.com, ihanafy@hotmail.com, majdedabash@yahoo.com
3 Faculty of Engineering, port-said University, Egypt. Email: hewayda2011@eng.psu.edu.eg

# Some GIS Topological Concepts via Neutrosophic Crisp Set Theory

## Abstract

In this paper we introduce and study the neutrosophic crisp pre-open, semi-open, $\beta$- open set, neutrosophic crisp continuity and neutrosophic crisp compact spaces are introduced. Furthermore, we investigate some of their properties and characterizations. Possible application to GIS topology rules are touched upon.

## 1. Introduction

Smarandache [26, 27] introduced the notion of neutrosophic sets, which is a generalization of Zadeh's fuzzy set [28]. In Zadah's sense, there is no precise definition for the set. Later on, Atanassov presented the idea of the intuitionistic fuzzy set [1], where he goes beyond the degree of membership introducing the degree of non-membership of some element in the set. The new presented concepts attracted several authors to develop the classical mathematics. For instance, Chang [2] and Lowen [6] started the discipline known as "Fuzzy Topology", where they forwarded the concepts from fuzzy sets to the classical topological spaces. Furthermore, Salama et al. [14, 17, 20] established several notations for what they called, "Neutrosophic topological spaces"].

In this paper, we study in more details some weaker and stronger structures constructed from the neutrosophic crisp topology introduced in [7], as well as the concepts neutrosophic crisp interior and the neutrosophic closure.

The remaining of this paper is structured as follows: in **§2**, some basic definitions are presented, while the new concepts of neutrosophic crisp nearly open sets are introduced in **§3**, in addition to providing a study of some of its properties. The neutrosophic crisp continuous function and neutrosophic crisp compact spaces are presented in **§4** and **§5**, respectively.

## 2. Terminologies

We recollect some relevant basic preliminaries, in particular, the work introduced by We recollect some relevant basic preliminaries, in particular, the work introduced by Smarandache and Salama [7], Salama et al. [8] and Smarandache [25,26,27]. The neutrosophic components T, I, F: X⟶$]0^-, 1^+[$to represent the membership, indeterminacy, and non-membership values of some universe X, respectively, where $]0^-, 1^+[$is the non-standard unit Interval.

**Definition 2.1** [7]

Let $X$ be a non-empty fixed sample space. A neutrosophic crisp set ($NCS$ for short) $A$ is an object having the form $A = (A_1, A_2, A_3)$ where $A_1$, $A_2$ and $A_3$ are subsets of $X$. Where $A_1$ contains all those members of the space X that accept the event A and $A_3$ contains all those members of the space X that rejected the event A, while $A_2$ contains those who stand in a distance from accepting or rejecting A.

**Definition 2.2**

Salama [7] defined the object having the form $A = (A_1, A_2, A_3)$ to be

1) ***(Neutrosophic Crisp Set with Type 1)***,if satisfying $A_1 \cap A_2 = \emptyset$ , $A_1 \cap A_3 = \emptyset$ and $A_2 \cap A_3 = \emptyset$. ($NCS$ -Type 1 ).
2) ***(Neutrosophic Crisp Set with Type 2)***, if satisfying $A_1 \cap A_2 = \emptyset$ , $A_1 \cap A_3 = \emptyset$ and $A_2 \cap A_3 = \emptyset$ and $A_1 \cup A_2 \cup A_3 = X$ ($NCS$ -Type 2 ).
3) ***(Neutrosophic Crisp Set with Type 3)*** if satisfying $A_1 \cap A_2 \cap A_3 = \emptyset$ and $A_1 \cup A_2 \cup A_3 = X$. ($NCS$ -Type3 for short) .

Every neutrosophic crisp set $A$ of a non-empty set $X$ is obviously a$NCS$ having the form $A = (A_1, A_2, A_3)$.

**Definition 2.3** [7]

Let $A = (A_1, A_2, A_3)$ a $NCS$ on $X$, then the complement of the set $A$ , ($A^c$ for short ) was presented in [7], to have one of the following forms:

($C_1$) $A^c = (A_1^c, A_2^c, A_3^c)$ or

($C_2$) $A^c = (A_3, A_2, A_1)$ or

($C_3$) $A^c = (A_3, A_2^c, A_1)$.

Several relations and operations between $NCS$ were defined in [7], which we are introducing in the following:

**Definition 2.4** [7]

Let $X$ be a non-empty set, and $NCSA$ and $B$ in the form $A = (A_1, A_2, A_3)$, $B = (B_1, B_2, B_3)$, then we may consider two possible definitions for subsets ($A \subseteq B$).

The concept of ($A \subseteq B$) may be defined as two types:

Type 1. $A \subseteq B \Leftrightarrow A_1 \subseteq B_1$, $A_2 \subseteq B_2$ and $A_3 \supseteq B_3$ or

Type 2. $A \subseteq B \Leftrightarrow A_1 \subseteq B_1$, $A_2 \supseteq B_2$ and $A_3 \supseteq B_3$

**Proposition 2.5**[7]

For any neutrosophic crisp set $A$ the following are hold

$\phi_N \subseteq A$, $\phi_N \subseteq \phi_N$

$A \subseteq X_N$, $X_N \subseteq X_N$

**Definition 2.6**[7]

Let $X$be a non-empty set, and the two $NCSsA$ and $B$given in the form $A = (A_1, A_2, A_3)$ , $B = (B_1, B_2, B_3)$, then :

1) $A \cap B$ may be defined as two types:

i)Type 1. $A \cap B = \langle A_1 \cap B_1, A_2 \cap B_2, A_3 \cup B_3 \rangle$

ii) Type 2. $A \cap B = \langle A_1 \cap B_1, A_2 \cup B_2, A_3 \cup B_3 \rangle$

2) $A \cup B$ may be defined as two types:

i) Type 1. $A \cup B = \langle A_1 \cup B_1, A_2 \cup B_2, A_3 \cap B_3 \rangle$

ii) Type 2. $A \cup B = \langle A_1 \cup B_1, A_2 \cap B_2, A_3 \cap B_3 \rangle$

**Definition 2.7**[7]

A neutrosophic crisp topology ($NCT$ ) on a non-empty set $X$ is a family $\Gamma$of neutrosophic crisp subsets of$X$ satisfying the following axioms:

i) $\emptyset_N, X_N \in \Gamma$.

ii) $A_1 \cap A_2 \in \Gamma$, $\forall A_1, A_2 \in \Gamma$.

iii) $\cup A_j \in \Gamma$, $\forall \{A_j : j \in J\} \subseteq \Gamma$.

In this case, the pair $(X, \Gamma)$ is called a neutrosophic crisp topological space ($NCTS$) in $X$. The elements of $\Gamma$are called neutrosophic crisp open sets ($NCOSs$) in $X$. A neutrosophic crisp set F is closed if and only if its complement $F^c$ is an open neutrosophic crisp set.

**Definition 2.8**[7]

Let $(X, \Gamma)$ be $NCTS$ and $A = \langle A_1, A_2, A_3 \rangle$ be a $NCS$ in $X$. Then the neutrosophic crisp closure of $A$ ($NCcl(A)$) and neutrosophic interior crisp ($NCint(A)$ ) of $A$ are defined by

$NCcl(A) = \cap\{K : K$ is an $NCCS$ in $X$ and $A \subseteq K\}$

$NCint(A) = \cup\{G : G$ is an $NCOS$ in $X$ and $G \subseteq A)$ ,

Where $NCS$ is a neutrosophic crisp set and $NCOS$ is a neutrosophic crisp open set. It can be also shown that $NCcl(A)$ is a $NCCS$ (neutrosophic crisp closed set) and $NCint(A)$ is a $NCOS$ (neutrosophic crisp open set) in $X$ .

## 3. Neutrosophic Crisp Nearly Open Sets

**Definition 3.1**

Let $(X, \Gamma)$ be a $NCTS$ and $A = \langle A_1, A_2, A_3 \rangle$ be a $NCS$ in $X$, then $A$ is called: Neutrosophic crisp $\alpha$-open set iff$A \subseteq NCint(NCcl(NCint(A))$. [24]

i) Neutrosophic crisp pre-open set iff$A \subseteq NCint(NCcl(A))$ .

ii) Neutrosophic crisp semi-open set iff$A \subseteq NCcl(NCint(A))$ .

iii) Neutrosophic crisp $\beta$- open set iff$A \subseteq (NCcl(NCint(NCcl(A))$.

We shall denote the class of all neutrosophic crisp $\alpha$- open sets as$NC\Gamma^{\alpha}$, and the class of all neutrosophic crisp pre-open sets as $NC\Gamma^{p}$, and the class of all neutrosophic crisp semi-open sets as $NC\Gamma^{S}$, and the class of all neutrosophic crisp $\beta$- open sets as $NC\Gamma^{\beta}$.

**Definition 3.2**

Let $(X, \Gamma)$ be a $NCTS$ and $B = \langle B_1, B_2, B_3 \rangle$ be a $NCS$ in $X$, then $B$ is called:

i) Neutrosophic crisp $\alpha$-closed set iff $(NCcl(NCint(NCcl(B)) \subseteq B$.

ii) Neutrosophic crisp pre- closed set iff $NCcl(NCint(B)) \subseteq B$.

iii) Neutrosophic crisp semi- closed set iff $NCint(NCcl(B)) \subseteq B$.

iv) Neutrosophic crisp $\beta$- closed set iff $NCint(NCcl(NCint(B)) \subseteq B$.

One can easily show that, the complement of a neutrosophic crisp ($\alpha$, pre, semi, $\beta$)- open set is a neutrosophic crisp ($\alpha$, pre, semi, $\beta$)- closed set,respectively.

**Remark 3.3**

For the class consisting of exactly all a $NC\alpha$- structure and $NC\beta$- structure, evidently, $NC\Gamma \subseteq NC\Gamma^{\alpha} \subseteq NC\Gamma^{\beta}$.

We notice that every non-empty $NC\beta$- open has $NC\alpha$-open non-empty interior.

If all neutrosophic crisp sets the family $\{B_i\}_{i\in I}$, are $NC\ \beta$- open sets, then

**Proposition 3.4**

Consider, $\{\cup B_i\}_{i\in I}$, is a family of $NC\beta$- open sets, then $\{\cup B_i\}_{i\in I} \subset NCcl(NCint(B_i)) \subset NCcl(NCint(B_i))$ , that is A $NC\beta$- structure is a neutrosophic closed with respect to arbitrary neutrosophic crisp unions .

We shall now characterize $NC\Gamma^{\alpha}$ in terms of$NC\Gamma^{\beta}$ .

**Definition 3.5**

Let $(X, \Gamma)$ be a $NCTS$ and $A = \langle A_1, A_2, A_3\rangle$ be a $NCS$ in $X$, then:

$NCcl\alpha(A) = \cap\{$ G:G$\supseteq A$ and G is $NC\alpha$-closed$\}$

$NCint\alpha\ (A) = \cup\{$G:G$\subseteq A$ and G is $NC\alpha$- open$\}$

$NCcl$ pre $(A) = \cap\{$ G:G$\supseteq A$ and G is $NC$pre-closed$\}$

$NCint$ pre $(A) = \cup\{$G:G$\subseteq A$ and G is $NC$pre- open$\}$

**Definition 3.6**

$NCcl$ semi $(A) = \cap\{$ G:G$\supseteq A$ and G is $NC$semi-closed$\}$

$NCint$ semi $(A) = \cup\{$G:G$\subseteq A$ and G is $NC$semi- open$\}$

$NCcl\beta(A) = \cap\{$ G:G$\supseteq A$ and G is $NC\beta$-closed$\}$

$NCint\beta(A) = \cup\{$G:G$\subseteq A$ and G is $NC\beta$- open$\}$

**Theorem 3.7**

Let $(X, \Gamma)$ be a $NCTS$. $NC\Gamma^{\alpha}$ Consists of exactly those $NCSA$ for which $A\cap B\in NC\Gamma^{\beta}$ for $B\in NC\Gamma^{\beta}$.

**Proof**

Let $A\in NC\Gamma^{\alpha}$, $B\in NC\Gamma^{\beta}$, $P\in A\cap B$ and $U$ be a neutrosophic crisp neighborhood (for short $NCnbd$) of p.

Clearly $U \cap NCint(NCcl(NCint(A))$, too is a neutrosophic crisp open neighborhood of $P$, so $V$=$(U\cap\ NCint(NCcl(NCint(A)))) \cap NCint(B)$ is non-empty . Since $V\subset NCcl(NCint(A))$ this implies

$(U\cap NCint(A) \cap NCint(B) = V\cap NCint(A) = \emptyset_N$ .

It follows that

Conversely, $A\cap B \subset NCcl(NCint(A) \cap NCint(B)) = NCcl(NCint(A\cap B))$ i.e. $A \cap B\in NC\Gamma^{\beta}$. Let $A\cap B\in NC\Gamma^{\beta}$for all $B\in NC\Gamma^{\beta}$. then in particular $A\in NC\Gamma^{\beta}$. Assume that

$P\in A\cap(NCint(NCcl(A)\cap(NCint(A)))^c$. Then $P\in NCcl(B)$, where $(NCcl(NCint(A)))^c$ Clearly $\{P\}\cup B\in NC\Gamma^{\beta}$ and consequently $A\cap\{\{P\}\cup B\}\in NC\Gamma^{\beta}$. But $A\cap\{\{P\}\cup B\}=\{P\}$. Hence $\{P\}$ is a neutrosophic crisp open. $P\in(NCcl(NCint(A)))$ implies $P\in)NCint(NCcl(NCint(A)()),$ contrary to assumption. Thus $P\in A$ implies $P\in(NCcl(NCint(A))$ and $A\in NC\Gamma^{\alpha}$. Thus we have found that $NC\Gamma^{\alpha}$ is complete determined by $NC\Gamma^{\beta}$ i.e. all neutrosophic crisp topologies with the same $NC\beta$- structure also determined the same $NC\alpha$-structure, explicitly given Theorem 3.1.

We shall prove that conversely all neutrosophic crisp topologies with the same $NC\alpha$-structure, so that $NC\Gamma^{\beta}$, is completely determined by $NC\Gamma^{\alpha}$

**Theorem 3.8**

Every $NC\alpha$-structure is a $NC\Gamma$.

**Proof**

$NC\Gamma^{\beta}$ Contains the neutrosophic crisp empty set and is closed with respect to arbitrary unions. A standard result gives the class of those neutrosophic crisp sets $A$ for which $A \cap B \in NC\Gamma^{\beta}$ for all $B \in NC\Gamma^{\beta}$ constitutes a neutrosophic crisp topology, hence the theorem.

We may now characterize $NC\Gamma^{\beta}$, in terms of $NC\Gamma^{\alpha}$ in the following way.

**Proposition 3.9**

Let $(X, \Gamma)$ be a $NCTS$. Then $NC\Gamma^{\beta} = NC\Gamma^{\alpha\beta}$ and hence $NC\alpha$ -equivalent topologies determine the same $NC\beta$ -structure.

**Proof**

Let $NC\alpha\text{–}cl$ and$\alpha$–int denote neutrosophic closure and Neutrosophic crisp interior with respect to $NC\Gamma^{\alpha}$. If $P \in B \in NC\Gamma^{\beta}$ and $P \in B \in NC\Gamma^{\alpha}$, then

$(NCint(NCcl(NCint(A))) \cap NCint(B)) \neq \emptyset_N$.

Since $(NCint(NCcl(NCint(A)))$ is a crisp neutrosophic neighbor-hood of point p, so certainly $NCint(B)$ meets $NCcl(NCint(A))$ and therefore (big neutrosophic open) meets $NCint(A)$, proving $A \cap NCint(B) \neq \emptyset_N$. This means $B \subset NC\alpha cl(NCint(B))$ .i.e. $B \in NC\Gamma^{\alpha\beta}$ on the other hand let $A \in NC\Gamma^{\alpha\beta}$, $P \in A$. and $P \in V \in NC\Gamma$. As $V \in NC\Gamma^{\alpha}$, and $P \in NCcl(NCint(A))$, we have $V \cap NCint(A) \neq \emptyset_N$ and there exist a neutrosophic trip set $W \in \Gamma$ such that $W \subset V \cap NC\alpha int(A) \subset A$.

In other words $V \cap (NCint(A)) \neq \emptyset_N$ and $P \in NCcl(NCint(A))$. Thus we have verified $NC\Gamma^{\alpha\beta} \subset NC\Gamma^{\alpha}$, and the proof is complete combining Theorem 3.1 and Proposition 3.1. and we get $NC\Gamma^{\alpha\alpha} = NC\Gamma^{\alpha}$.

**Corollary 3.10**

A neutrosophic crisp topology $NC\Gamma$is a $NC\alpha$ - topology iff $NC\Gamma = NC\Gamma^{\alpha}$. Evidently $NC\Gamma^{\beta}$ is a neutrosophic crisp topology iff$NC\Gamma^{\alpha} = NC\Gamma^{\beta}$. In this case $NC\Gamma^{\beta\beta} = NC\Gamma^{\alpha\beta} = NC\Gamma^{\beta}$.

**Corollary 3.11**

$NC\beta$-Structure $B$ is a neutrosophic crisp topology, then $B = B\alpha = B\beta$.

We proceed to give some results on the neutrosophic structure of neutrosophic crisp $NC\alpha$–topology

**Proposition 3.12**

The $NC\alpha$-open with respect to a given neutrosophic crisp topology are exactly those sets which may be written as a difference between a neutrosophic crisp open set and a neutrosophic crisp nowhere dense set. If $A \in NC\Gamma^{\alpha}$ we have $A = NCint(NCcl(NCint(A))) \cap (NCint(NCcl(NCint(A)) \cap A^c)^c$, where $(NCint(NCcl(NCint(A)) \cap A^c)$ clearly is neutrosophic crisp nowhere dense set, we easily see that

$B \subset NCcl(NCint(A))$ and consequently

$A \subset B \subset NCint(NCcl(NCint(A))$ so the proof is complete.

**Corollary 3.13**

A neutrosophic crisp topology is a $NC\alpha$- topology iff all neutrosophic crisp nowhere dense sets are neutrosophic crisp closed. For a neutrosophic crisp $NC\alpha$-topology may be characterized as neutrosophic crisp topology where the difference between neutrosophic crisp open and neutrosophic crisp nowhere dense set is again a neutrosophic crisp open, and this evidently is equivalent to the condition stated.

**Proposition 3.14**

Neutrosophic crisp topologies which are $NC\alpha$- equivalent, determine the same class of neutrosophic crisp nowhere dense sets.

**Proposition 3.16**

If a $NC\alpha$ -Structure $B$, is a neutrosophic crisp topology, then all neutrosophic crisp topologies $\Gamma$ for which $\Gamma^{\beta} = B$ are neutrosophic crisp extremely disconnected.

In particular: Either all or none of the neutrosophic crisp topologies of a $NC\alpha$ – class are extremely disconnected.

**Proof**

Let $\Gamma^{\beta} = B$, and suppose there is $A \in \Gamma$ such that $NCcl(A) \notin \Gamma$. Let $P \in NCcl(A) \cap NCint(NCcl(A))^{c}$ with $B = \{P\} \cup NCint(NCcl(A))$, $M = NCint(NCcl(A))^{c}$

We have $\{P\} \subset M = (NCint(NCcl(\mathrm{A}))^{c} = NCcl(NCint(M))$,

$\{P\} \subset NCcl(A) = NCcl(NCint(NCcl(A)) \subset NCcl(NCint(B))$. Hence both $B$ and $M$ are in $\Gamma^{\beta}$. The intersection $B \cap M = \{P\}$ is not neutrosophic crisp open, since $P \in NCcl(A) \cap M^{c}$ hence not $NC\beta$- open. So, $\Gamma^{\beta} = B$ is not a neutrosophic crisp topology. Now suppose $B$ is not a topology, and $\Gamma^{\beta} = B$ There is a $B \in \Gamma^{\beta}$ such that $B \notin \Gamma^{\alpha}$. Assume that $NCcl(NCint(B)) \in \Gamma$. Then

$B \subset NCcl(NCint(B)) = NCint(NCcl(NCint(B))$ i.e. $B \in \Gamma^{\alpha}$, contrary to assumption. Thus we have produced an open set whose closure is not open, which completes the proof.

**Corollary 3.17**

A neutrosophic crisp topology $\Gamma$ is a neutrosophic crisp extremally disconnected if and only if $\Gamma^{\beta}$ is a neutrosophic crisp topology.

**Remark 3.18**

The following diagram represents the relation between neutrosophic crisp nearly open sets:

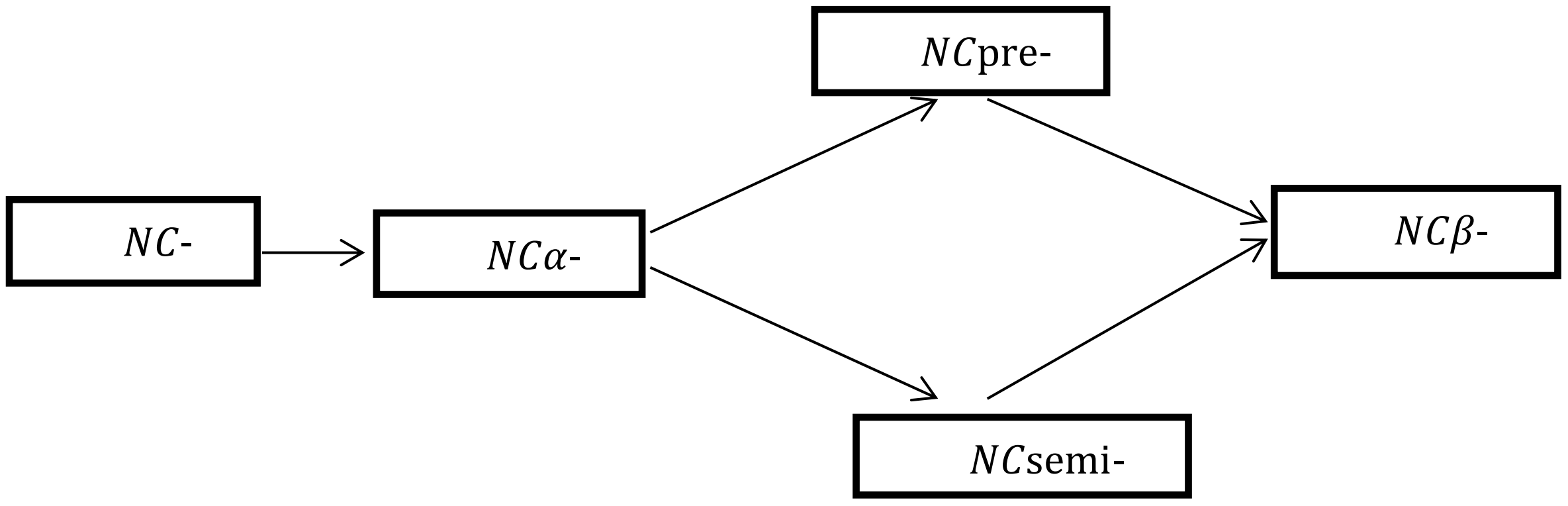

## 4. Neutrosophic Crisp Continuity

We, introduce and study of neutrosophic crisp continuous function and we obtain some characterizations of neutrosophic continuity. Here come the basic definitions first:

**Definition 4.1**

Let $(X, \Gamma)$ be a $NCTS$ and $A = \langle A_1, A_2, A_3 \rangle$ be a $NCS$ in $X$, and $f: X \longrightarrow X$ then:

1) If $f NC\alpha$-continuous $\Longrightarrow$ inverse image of $NC\alpha$ open set is $NC\alpha$- open set
2) If $f$ $NC$pre-continuous $\Longrightarrow$ inverse image of $NC$pre-open set is $NC$pre- open set
3) If $f$ $NC$semi-continuous $\Longrightarrow$ inverse image of $NC$semi-open set is $NC$semi- open set
4) If $f$ $NC\beta$-continuous $\Longrightarrow$ inverse image of $NC\beta$-open set is $NC\beta$- open set

**Definition 4.2**

The following was given in [24]

(a) If $A = \langle A_1, A_2, A_3 \rangle$ is a NCS in X, then the neutrosophic crisp image of A under $f$,

denoted by $f(A)$, is the a NCS in Y defined by $f(A)=\langle f(A_1), f(A_2), f(A_3)\rangle$.

(b) If $f$ is a bijective map then $f^{-1}: Y \longrightarrow X$ is a map defined such that:
for any NCS $B=\langle B_1, B_2, B_3\rangle$ in Y, the neutrosophic crisp preimage of B, denoted by $f^{-1}(B)$, is a $NCS$ in X defined by $f^{-1}(B)=\langle f^{-1}(B_1), f^{-1}(B_2), f^{-1}(B_3)\rangle$.

**Definition 4.3**

Let $(X, \Gamma_1)$ , and $(Y, \Gamma_2)$ be two $NCTSs$, and let $f: X \longrightarrow Y$ be a function. Then $f$ is said to be continuous if $f$ the preimage of each $NCS$ in $\Gamma_2$ is a $NCS$ in $\Gamma_1$.

**Definition 4.4**

Let $(X, \Gamma_1)$ , and $(Y, \Gamma_2)$ be two $NCTSs$ and let $f: X \longrightarrow Y$ be a function. Then $f$ is said to be open iff the image of each $NCS$ in $\Gamma_1$, is a $NCS$ in $\Gamma_2$.

**Proposition4.5**

Let $(X, \Gamma_o)$ and $(Y, \psi_o)$ be two $NCTSs$.

If $f: X \to Y$ is continuous in the usual sense, then in this case, $f$ is continuous in the sense of Definition 4.3 too.

**Proof**

Here we consider the $NCTSs$ on X and Y, respectively, as follows: $\Gamma_1 = \{\langle G, \phi, G^c\rangle : G \in \Gamma_o\}$ and $\Gamma_2 = \{\langle H, \phi, H^c\rangle : H \in \Psi_o\}$,

In this case we have, for each $\langle H, \phi, H^c\rangle \in \Gamma_2$ , $H \in \Psi_o$ ,

$f^{-1}\langle H, \phi, H^c\rangle = \langle f^{-1}(H), f^{-1}(\phi), f^{-1}(H^c)\rangle = \langle f^{-1}H, \phi, (f^{-1}(H))^c\rangle \in \Gamma_1$ .

Now we obtain some characterizations of neutrosophic continuity.

**Proposition 4.6**

Let $f$: $(X, \Gamma_1) \longrightarrow (Y, \Gamma_2)$. Then f is neutrosophic continuous iff the preimage of each neutrosophic crisp closed set ($NCCS$) in $\Gamma_2$ is a $NCCS$ in $\Gamma_1$.

**Proposition 4.7**

The following are equivalent to each other:

(a) $f$: $(X, \Gamma_1) \longrightarrow (Y, \Gamma_2)$ is neutrosophic continuous.

(b) $f^{-1}(NCint(B) \subseteq NCint(f^{-1}(B)))$ for each $NCSB$ in $Y$.

(c) $(NCcl\ f^{-1}(B)) \subseteq f^{-1}(NCcl(B))$. for each $NCSB$ in $Y$.

**Corollary 4.8**

Consider $(X, \Gamma_1)$ and $(Y, \Gamma_2)$ to be two $NCTSs$, and let $f: X \to Y$ be a function.

if $\Gamma_1 = \{f^{-1}(H) : H \in \Gamma_2\}$. Then $\Gamma_1$ will be the coarsest *NCT* on X which makes the function $f: X \to Y$ continuous. One may call it the initial neutrosophic crisp topology with respect to $f$.

## 5. Neutrosophic Crisp Compact Space

First we present the basic concepts:

**Definition5.1**

Let $(X, \Gamma)$ be an *NCTS*.

(a) If a family $\{\langle G_{i_1}, G_{i_2}, G_{i_3}\rangle : i \in J\}$ of *NCOSs* in X satisfies the condition $\cup \{\langle G_{i_1}, G_{i_2}, G_{i_3}\rangle : i \in J\} = X_N$, then it is called an neutrosophic open cover of X.

(b) A finite subfamily of an open cover $\{\langle G_{i_1}, G_{i_2}, G_{i_3}\rangle : i \in J\}$ on X, which is also a

neutrosophic open cover of X , is called a neutrosophic crisp finite open subcover.

**Definition5.2**

A neutrosophic crisp set $A=\langle A_1, A_2, A_3\rangle$ in a NCTS $(X,\Gamma)$ is called neutrosophic crisp compact iff every neutrosophic crisp open cover of A has a finite neutrosophic crisp open subcover.

**Definition5.3**

A family $\{\langle K_{i1}, K_{i2}, K_{i3}\rangle : i \in J\}$ of neutrosophic crisp compact sets in X satisfies the finite intersection property (FIP ) iff every finite subfamily $\{\langle K_{i_1}, K_{i_2}, K_{i_3}\rangle : i = 1,2,\ldots,n\}$ of the family satisfies the condition $\cap\{\langle K_{i_1}, K_{i_2}, K_{i_3}\rangle : i = 1,2,\ldots,n\} \neq \phi_N$ .

**Definition5.4**

A NCTS $(X,\Gamma)$ is called neutrosophic crisp compact iff each neutrosophic crisp open cover of X has a finite open subcover.

**Corollary5.5**

A *NCTS* $(X,\Gamma)$ is a neutrosophic crisp compact iff every family $\{\langle G_{i_1}, G_{i_2}, G_{i_3}\rangle : i \in J\}$ of neutrosophic crisp compact sets in X having the finite intersection properties has nonempty intersection.

**Corollary5.6**

Let $(X,\Gamma_1)$, $(Y,\Gamma_2)$ be NCTSs and $f: X \to Y$ be a continuous surjection. If $(X,\Gamma_1)$ is a neutrosophic crisp compact, then so is $(Y,\Gamma_2)$.

**Definition5.7**

If a family $\{\langle G_{i1}, G_{i2}, G_{i3}\rangle : i \in J\}$ of neutrosophic crisp compact sets in X satisfies the condition $A \subseteq \cup\{\langle G_{i1}, G_{i2}, G_{i3}\rangle : i \in J\}$, then it is called a neutrosophic crisp open cover of A.

Let's consider a finite subfamily of a neutrosophic crisp open subcover of $\{\langle G_{i1}, G_{i2}, G_{i3}\rangle : i \in J\}$.

**Corollary5.8**

Let $(X,\Gamma_1)$, $(Y,\Gamma_2)$ be *NCTSs* and $f: X \to Y$ be a continuous surjection. If A is a neutrosophic crisp compact in $(X,\Gamma_1)$, then so is $f(A)$ in $(Y,\Gamma_2)$.

## 6. Conclusion

In this paper, we presented a generalization of the neutrosophic topological space. The basic definitions of the neutrosophic crisp topological space and the neutrosophic crisp compact space with some of their characterizations were deduced. Furthermore, we constructed a neutrosophic crisp continuous function, with a study of a number its properties.

## References

1. K. Atanassov, intuitionistic fuzzy sets, Fuzzy Sets and Systems 20, 87-96,(1986).
2. C.L.Chang, Fuzzy Topological Spaces, 1. Math. Anal-Appl. 245,182-190(1968).
3. I.M. Hanafy, A.A. Salama and K.M. Mahfouz, Neutrosophic Crisp Events and Its Probability, International Journal of Mathematics and Computer Applications Research(IJMCAR) Vol.(3), Issue 1, pp.171-178,Mar (2013).
4. I. M. Hanafy, A.A. Salama and K. Mahfouz, Correlation of Neutrosophic Data, International Refereed Journal of Engineering and Science (IRJES), Vol.(1), Issue 2 . PP.39-33. (2012).

5. I. M. Hanafy, A. A. Salama, O. M.Khaled and K. M. Mahfouz Correlation of Neutrosophic Sets in Probability Spaces, JAMSI,Vol.10,No.(1), pp45-52,(2014).
6. R. Lowen, Fuzzy topological spaces and compactnees, J.Math. Anal. Appl.56, 621- 633(1976).
7. A. A. Salama and F. Smarandache Neutrosophic Crisp Set Theory, Columbus, Ohio,USA,(2015)
8. A. A. Salama, S. Broumi and F. Smarandache, Introduction to Neutrosophic Topological Spatial Region, Possible Application to GIS Topological Rules, I.J. Information engineering and electronic business, vol 6, pp15-21, (2014) .
9. A. Salama, Basic Structure of Some Classes of Neutrosophic Crisp Nearly Open Sets and Possible Application to GIS Topology, Neutrosophic Sets and Systems, Vol. (7) pp18-22,(2015).
10. A. Salama and F. Smarandache, Neutrosophic Ideal Theory Neutrosophic Local Function and Generated Neutrosophic Topology, In Neutrosophic Theory and Its Applications. Collected Papers, Volume 1, EuropaNova, Bruxelles, pp. 213-218,(2014).
11. A. Salama, Neutrosophic Crisp Points & Neutrosophic Crisp Ideals, Neutrosophic Sets and Systems, Vol.1, No. 1,pp 50-54,(2013).
12. A. Salama and F. Smarandache and S. A. Alblowi, The Characteristic Function of a Neutrosophic Set, Neutrosophic Sets and Systems, Vol.3pp14-17,(2014).
13. A. Salama, Smarandache and ValeriKroumov, Neutrosophic Crisp Sets & Neutrosophic Crisp Topological Spaces, Neutrosophic Sets and Systems, Vol. (2), pp25-30,(2014).
14. A. Salama, Florentin Smarandache and ValeriKroumov. Neutrosophic Closed Set and Neutrosophic Continuous Functions, Neutrosophic Sets and Systems, Vol. (4) pp4-8,(2014).
15. A. Salama, O. M. Khaled, and K. M. Mahfouz. Neutrosophic Correlation and Simple Linear Regression, Neutrosophic Sets and Systems, Vol. (5) pp3-8,(2014)
16. A. Salama, and F. Smarandache. Neutrosophic Crisp Set Theory, Neutrosophic Sets and Systems, Vol. (5) pp27-35,(2014).
17. A. Salama, Florentin Smarandache and S. A. ALblowi, New Neutrosophic Crisp Topological Concepts, Neutrosophic Sets and Systems, Vol(4)pp50-54,(2014).
18. A. Salama and F. Smarandache, Filters via Neutrosophic Crisp Sets, Neutrosophic Sets and Systems, Vol.1, No. 1, pp. 34-38,(2013).
19. A. Salama and S.A. Alblowi, Generalized Neutrosophic Set and Generalized Neutrosophic Spaces, Journal Computer Sci. Engineering, Vol. (2) No. (7) pp129-132,(2012) .
20. A. Salama and S. A. Alblowi,"Neutrosophic Set and Neutrosophic Topological Spaces", ISOR J. Mathematics, Vol.(3), Issue(3) pp-31-35,(2012).
21. A. Salama, S. Broumi and F. Smarandache, Neutrosophic Crisp Open Set and Neutrosophic Crisp Continuity via Neutrosophic Crisp Ideals, I.J. Information Engineering and Electronic Business, Vol.(3), pp. 1-8,(2014).
22. A. Salama, Said Broumi and S. A. Alblowi, Introduction to Neutrosophic Topological Spatial Region, Possible Application to GIS Topological Rules,I.J. Information Engineering and Electronic Business, 2014, 6, 15-21 ,(2014) .
23. A. Salama and H. Elghawalby, *Neutrosophic Crisp Set and *Neutrosophic Crisp relations, Neutrosophic Sets and Systems, Vol.(5),pp. 13-17,(2014).
24. A.A.Salama, I.M.Hanafy, H. ElGhawalby and M.S. Dabash, Neutrosophic Crisp α-Topological Spaces, accepted for Neutrosophic Sets and Systems, (2016)
25. F. Smarandache, A Unifying Field in Logics: Neutrosophic Logic. Neutrosophy, Neutrosophic Set, Neutrosophic Probability. American Research Press, Rehoboth, NM, (1999).
26. F. Smarandache, Neutrosophic set, a generalization of the intuitionistic fuzzy sets, Inter. J. Pure Appl. Math, vol. 24 ,pp. 287 — 297, (2005).
27. F. Smarandache, Introduction To Neutrosophic Measure,Neutrosophic Integral and Neutrosophic Probability,http://Fs.Gallup.Unntedu/Ebooks-Other formats.htm (2013).
28. L.A. Zadeh, Fuzzy Sets. Inform. Control, vol 8, pp 338-353, (1965).

# OTHER THEORETICAL PAPERS

J. MARTINA JENCY, I. AROCKIARANI

Department of Mathematics, Nirmala College for Women, Coimbatore, Tamilnadu, India.

# Hausdorff Extensions in Single Valued Neutrosophic $S^*$ Centered Systems

## Abstract

This paper explores the concept of single valued neutrosophic $S^*$ open sets in single valued neutrosophic $S^*$ centered system. Also the characterization of Hausdorff extensions of spaces in single valued neutrosophic $S^*$ centered systems are established.

## 1. Introduction

Florentin Smarandache [8, 9] combined the non-standard analysis with a tri component logic/set, probability theory with philosophy and proposed the term neutrosophy which means knowledge of neutral thoughts. This neutral represents the main distinction between fuzzy and intuitionistic fuzzy logic set. In 1998, Florentin Smarandache defined the neutrosophic set [8, 9]. Florentin Smarandache and his colleagues [5] presented an instance of neutrosophic set, called single valued neutrosophic set. Alexandrov [1] developed a method of centered systems for studying compact extensions of topological spaces. The method of centered systems in topological spaces was studied by Iliadis [6] and in fuzzy topological spaces by Uma et al. [10]. We extend the same in single valued neutrosophic topological spaces.

## 2. Preliminaries

### Definition 2.1. [5]

Let $X$ be a space of points (objects), with a generic element in $X$ denoted by $x$. A single valued neutrosophic set (SVNS) $A$ in $X$ is characterized by truth-membership function $T_A$, indeterminacy-membership function $I_A$ and falsity-membership function $F_A$.

For each point $x$ in $X$, $T_A(x)$, $I_A(x)$, $F_A(x) \in [0,1]$. When $X$ is continuous, a SVNS $A$ can be written as $A$, $\int_X \langle T_A(x), I_A(x), F_A(x)\rangle / x, x \in X$ .

When $X$ is discrete, a SVNS $A$ can be written as

$$A = \sum_{i=1}^{n} \langle T(x_i), I(x_i), F(x_i)\rangle / x_i, x_i \in X$$

**Definition 2.2: [10]**

Let $R$ be a fuzzy Hausdorff space. A system $p = \{\lambda_\alpha\}$ of fuzzy open sets of $R$ is called fuzzy centered if any finite collection of the fuzzy sets of the system has a non-empty intersection. The system $p$ is called a maximal fuzzy centered system or a fuzzy end if it cannot be included in any larger fuzzy centered system of fuzzy open sets.

**Definition 2.3: [10]**

Let $\theta(R)$ denote the collection of all fuzzy ends belonging to a given fuzzy Hausdorff space $R$. A fuzzy topology introduces into $\theta(R)$ in the following way. Let $P_\lambda$ be the set of all fuzzy ends that contain λ as an element, where λ is a fuzzy open set of $R$. Therefore, $P_\lambda$ is a fuzzy neighbourhood of each fuzzy end contained in $P_\lambda$.

## 3. Single valued neutrosophic S* Hausdorff extension spaces

**Definition 3.1**

Let $X$ be a non-empty set and $S$ be a collection of all single valued neutrosophic sets of X. A single valued neutrosophic S*structure on $S$ is a collection S* of subsets of $S$ having the following properties:

1. φ and $S$ are in S*.

2. The union of the elements of any sub collection of S* is in S*.

3. The intersection of the elements of any finite sub collection of S* is in S*.

The collection S together with the structure S* is called single valued neutrosophic S* structure space. The members of S* are called single valued neutrosophic S* open sets. The complement of single valued neutrosophic S* open set is said to be a single valued neutrosophic S* closed set.

**Example 3.2:**

Let $X = \{a,b\}$ , $S = \left\{\frac{a}{\langle 0.8,0.3,0.5\rangle}, \frac{b}{\langle 0.7,0.4,0.6\rangle}\right\}$, $S^* = \{S, \phi, S_1, S_2, S_3, S_4\}$ where,

$S_1 = \left\{ \frac{a}{\langle 0.6,0.1,0.7 \rangle}, \frac{b}{\langle 0.5,0.2,0.8 \rangle} \right\}$, $S_2 = \left\{ \frac{a}{\langle 0.4,0.2,0.6 \rangle}, \frac{b}{\langle 0.5,0.3,0.9 \rangle} \right\}$, $S_3 = \left\{ \frac{a}{\langle 0.4,0.1,0.7 \rangle}, \frac{b}{\langle 0.5,0.2,0.9 \rangle} \right\}$,
$S_4 = \left\{ \frac{a}{\langle 0.6,0.2,0.6 \rangle}, \frac{b}{\langle 0.5,0.3,0.8 \rangle} \right\}$.

Here $(S, S^*)$ is a structure space.

**Definition 3.3:**

Let $A$ be a member of $S$. A single valued neutrosophic S$^*$ open set U in $(S, S^*)$ is said to be a single valued neutrosophic S$^*$ open neighbourhood of $A$ if $A \in G \subset U$ for some single valued neutrosophic S$^*$ open set $G$ in $(S, S^*)$.

**Example 3.4:**

Let $X = \{a,b\}$ , $S = \left\{ \frac{a}{\langle 0.8,0.3,0.5 \rangle}, \frac{b}{\langle 0.7,0.4,0.6 \rangle} \right\}$, $S^* = \{S, \phi, S_1, S_2, S_3, S_4\}$ where,

$S_1 = \left\{ \frac{a}{\langle 0.6,0.1,0.7 \rangle}, \frac{b}{\langle 0.5,0.2,0.8 \rangle} \right\}$, $S_2 = \left\{ \frac{a}{\langle 0.4,0.2,0.6 \rangle}, \frac{b}{\langle 0.5,0.3,0.9 \rangle} \right\}$ ,

$S_3 = \left\{ \frac{a}{\langle 0.4,0.1,0.7 \rangle}, \frac{b}{\langle 0.5,0.2,0.9 \rangle} \right\}$, $S_4 = \left\{ \frac{a}{\langle 0.6,0.2,0.6 \rangle}, \frac{b}{\langle 0.5,0.3,0.8 \rangle} \right\}$.

Let $A = \left\{ \frac{a}{\langle 0.4,0.1,0.8 \rangle}, \frac{b}{\langle 0.3,0.1,0.9 \rangle} \right\}$.

Here $A \in S_1 \subset S_4$. $S_4$ is the single valued neutrosophic S* open neighbourhood of $A$ .

**Definition 3.5:**

Let $(S, S^*)$ be a single valued neutrosophic S$^*$ structure space and $A = \langle x, T_A, I_A, F_A \rangle$ be a single valued neutrosophic set in $X$. Then the single valued neutrosophic S$^*$ closure of $A$ (briefly $SV\,N\,S^*cl(A)$) and single valued neutrosophic S$^*$ interior of $A$ (briefly $SVN\,S^*int(A)$) are respectively defined by

$SVNS^*cl(A) = \bigcap$ {$K$: $K$ is a single valued neutrosophic S$^*$ closed sets in $S$ and $A \subseteq K$}

$SVNS^*int(A) = \bigcup$ {$G$: $G$ is a single valued neutrosophic $S^*$ open sets in $S$ and $G \subseteq A$}.

**Example 3.6:**

Let $X=\{a,b\}$ , $S=\left\{\frac{a}{\langle 0.8,0.3,0.5\rangle},\frac{b}{\langle 0.7,0.4,0.6\rangle}\right\}$, $S^{*}=\{S,\phi,S_1,S_2,S_3,S_4\}$ where,

$S_1=\left\{\frac{a}{\langle 0.6,0.1,0.7\rangle},\frac{b}{\langle 0.5,0.2,0.8\rangle}\right\}$, $S_2=\left\{\frac{a}{\langle 0.4,0.2,0.6\rangle},\frac{b}{\langle 0.5,0.3,0.9\rangle}\right\}$ ,

$S_3=\left\{\frac{a}{\langle 0.4,0.1,0.7\rangle},\frac{b}{\langle 0.5,0.2,0.9\rangle}\right\}$, $S_4=\left\{\frac{a}{\langle 0.6,0.2,0.6\rangle},\frac{b}{\langle 0.5,0.3,0.8\rangle}\right\}$.

$S_1^c=\left\{\frac{a}{\langle 0.7,0.9,0.6\rangle},\frac{b}{\langle 0.8,0.8,0.5\rangle}\right\}$, $S_2^c=\left\{\frac{a}{\langle 0.6,0.8,0.4\rangle},\frac{b}{\langle 0.9,0.7,0.5\rangle}\right\}$,

$S_3^c=\left\{\frac{a}{\langle 0.7,0.9,0.4\rangle},\frac{b}{\langle 0.9,0.8,0.5\rangle}\right\}$, $S_4^c=\left\{\frac{a}{\langle 0.6,0.8,0.6\rangle},\frac{b}{\langle 0.8,0.7,0.5\rangle}\right\}$.

Let $A=\left\{\frac{a}{\langle 0.5,0.3,0.6\rangle},\frac{b}{\langle 0.7,0.4,0.9\rangle}\right\}$.Then $SVN\,S^{*}\operatorname{int}(A)=\{S_3\}$.

$SVN\,S^{*}cl(A)=\{S_4^c\}$.

**Definition 3.7:**

The ordered pair (S, S*) is called a single valued neutrosophic S* Hausdorff space if for each pair $A_1$, $A_2$ of disjoint members of S, there exist disjoint single valued neutrosophic S* open sets $U_1$ and $U_2$ such that $A_1 \subseteq U_1$ and $A_2 \subseteq U_2$ .

**Example 3.8:**

Let $X=\{a,b\}$ , $S=\left\{\frac{a}{\langle 1,1,0\rangle},\frac{b}{\langle 1,1,0\rangle}\right\}$, $S^{*}=\{S,\phi,S_1,S_2,S_3\}$ where,

$S_1=\left\{\frac{a}{\langle 0.5,0,1\rangle},\frac{b}{\langle 0,0.3,0.4\rangle}\right\}$, $S_2=\left\{\frac{a}{\langle 0.5,0.2,0.5\rangle},\frac{b}{\langle 0.7,0.3,0.4\rangle}\right\}$,

$S_3=\left\{\frac{a}{\langle 0,0.2,0.5\rangle},\frac{b}{\langle 0.7,0,1\rangle}\right\}$.

Let $A_1 = \left\{ \frac{a}{\langle 0.3,0,1\rangle}, \frac{b}{\langle 0,0.1,1\rangle} \right\}$, $A_2 = \left\{ \frac{a}{\langle 0,0.1,0.6\rangle}, \frac{b}{\langle 0.5,0,1\rangle} \right\}$.

Here $A_1$ and $A_2$ are disjoint members of $S$ and $S_1, S_2$ are disjoint single valued neutrosophic S* open sets such that $A_1 \subseteq S_1$ *and* $A_2 \subseteq S_2$.

Hence the ordered pair $(S, S^*)$ is a single valued neutrosophic S* Hausdorff space.

**Definition 3.9:**

Let $(S_1, S_1^*)$ and $(S_2, S_2^*)$ be any two single valued neutrosophic *S** structure spaces and let $f:(S_1, S_1^*) \to (S_2, S_2^*)$ be a function. Then $f$ is said to be single valued neutrosophic *S** continuous iff the pre image of each single valued neutrosophic $S_2^*$ open set in $(S_2, S_2^*)$ is a single valued neutrosophic $S_1^*$ open set in $(S_1, S_1^*)$.

**Definition 3.10:**

Let $(S_1, S_1^*)$ and $(S_2, S_2^*)$ be any two single valued neutrosophic *S** structure spaces and let $f:(S_1, S_1^*) \to (S_2, S_2^*)$ be a bijective function. If both the functions $f$ and the inverse function $f^{-1}:(S_2, S_2^*) \to (S_1, S_1^*)$ are single valued neutrosophic *S** continuous then $f$ is called single valued neutrosophic *S** homeomorphism.

**Definition 3.11:**

Let $f$ be a function from a single valued neutrosophic *S** structure space $(S_1, S_1^*)$ into a single valued neutrosophic *S** structure space $(S_2, S_2^*)$ with $f(A_1) = f(A_2)$ where $A_1 \in (S_1, S_1^*)$ and $A_2 \in (S_2, S_2^*)$.Then $f$ is called a single valued neutrosophic $S^*\theta-$continuous at $A_1$ if for every neighbourhood $O_{A_2}$ of $A_2$, there exists a neighbourhood $O_{A_1}$ of $A_1$ such that $f(SVNS^*cl(O_{A_1})) \subset SVNS^*cl(O_{A_2})$.The function is called single valued neutrosophic $S^*\theta-$continuous if it is single valued neutrosophic $S^*\theta-$continuous at every member of $S_1$.

**Definition 3.12:**

A function is called a single valued neutrosophic $S^*\theta-$ homeomorphism if it is single valued neutrosophic $S^*$ one to one and single valued neutrosophic $S^*\theta-$ continuous in both directions.

**Definition 3.13:**

Let $(S, S^*)$ be a single valued neutrosophic $S^*$ Hausdorff space. A system $p = \{U_\alpha: \alpha = 1, 2, 3, ...n\}$ of single valued neutrosophic $S^*$ open sets is called a single valued neutrosophic $S^*$ centered system if any finite collection of the sets of the system has a non-empty intersection.

**Definition 3.14:**

The single valued neutrosophic $S^*$ centered system $p$ is called a maximal single valued neutrosophic S$^*$ centered system or a single valued neutrosophic $S^*$ end if it cannot be included in any larger single valued neutrosophic $S^*$ centered system of single valued neutrosophic $S^*$ open sets.

**Example 3.15:**

In Example 3.8 let us consider the system $p_1 = \{S_\alpha, \alpha = 1,2,3\}$. $p_1$ is a fuzzy neutrosophic $S^*$ centered system since $S_1, S_2$ has a non -empty intersection.

Let $p_2 = \{S_\alpha : \alpha = 1,2\}$ is also a fuzzy neutrosophic $S^*$ centered system.

Here $p_1$ is a maximal fuzzy neutrosophic $S^*$ centered system.

**Note 3.16:**

Throughout this paper $\{U_\alpha : \alpha = 1, 2, 3, ...n\}$ be a single valued neutrosophic $S^*$ open set in (S, S$^*$).

**Proposition 3.17:**

Let $(S, S^*)$ be a single valued neutrosophic $S^*$ Hausdorff space and $p = \{U_\alpha\}$ is a single valued neutrosophic $S^*$ centered system in $(S, S^*)$.Then the following properties hold.

1.If $U_i \in p \ (i = 1,2,3,....n)$ then $\bigcap_{i=1}^{n} U_i \in p$.

2.If $\phi \neq U \subseteq H, U \in p$ and $H$ is single valued neutrosophic $S^*$ open set, then $H \in p$.

3.If $H$ is single valued neutrosophic $S^*$ open set, then $H \notin p$ iff there exists $U \in p$ such that $U \in p$ such that $U \cap H = \phi$.

4.If $U_1 \cup U_2 = U_3 \in p, U_1$ and $U_2$ are single valued neutrosophic $S^*$ open sets and $U_1 \cap U_2 = \phi$, then either $U_1 \in p$ or $U_2 \in p$.

5.If $SVNS * cl(U) = S$ then $\phi \neq U \in p$ for any single valued neutrosophic $S^*$ end $p$.

**Proof:**

1.If $U_i \in p \ (i = 1,2,3,....n)$ then $\bigcap_{i=1}^{n} U_i \neq \phi$ .As a contrary, suppose that $\bigcap_{i=1}^{n} U_i \notin p$ , then $p \cup \left\{ \bigcap_{i=1}^{n} U_i \right\}$ will be a larger single valued neutrosophic $S^*$end than $p$ .This contradicts the maximality of $p$ .Therefore $\bigcap_{i=1}^{n} U_i \in p$.

2. If $H \notin p$, then $p \cup H$ will be a larger single valued neutrosophic $S^*$ end than $p$. This contradicts the maximality of $p$. Therefore $H \in p$.

3. Suppose that $H \notin p$. If there exists no $U \in p$ such that $H \cap U = \phi$ then by Definition 3.13 and Definition 3.14, $H \in p$. This contradicts the maximality of $p$, since $p \cup \{H\}$ will be a larger single valued neutrosophic $S^*$ end than $p$. Conversely, suppose that there exists $U \in p$ such that $H \cap U = \phi$. If $H \in p$ then $H \cap U \neq \phi$, which is a contradiction. Hence $H \notin p$.

4. If $U_1 \notin p, U_2 \notin p$, then $U_1 \cap U_3 = U_1 \notin p$ and $U_2 \cap U_3 = U_2 \notin p$. It follows that $U_3 \notin p$, which is a contradiction. Hence either $U_1 \in p$ or $U_2 \in p$.

5. $U \cap SVNS^*cl(U) = U$ and $SVNS^*cl(U) = S \in p$ for all single valued neutrosophic $S^*$ ends $p$. By (3) $U \cap SVNS^*cl(U) = U \neq \phi$. Therefore $\phi \neq U \in p$ for all single valued neutrosophic $S^*$ end $p$.

**Definition 3.18:**

Let $\theta(S)$ denote the collection of all single valued neutrosophic $S^*$ ends belonging to $S$. A single valued neutrosophic $S^*$ topology is introduced into $\theta(S)$ in the following way. Let $O_U$ be the set of all single valued neutrosophic $S^*$ ends that contains $U$ as an element, where $U$ is a single valued neutrosophic $S^*$ open set of $S$. Therefore $O_U$ is a single valued neutrosophic $S^*$ neighbourhood of each single valued neutrosophic $S^*$ end contained in $O_U$.

**Definition 3.19:**

A subset $A$ of a single valued neutrosophic $S^*$ structure space $(S, S^*)$ is said to be an everywhere single valued neutrosophic $S^*$ dense subset in $(S, S^*)$ if $SVNS^*cl(A) = S$.

**Definition 3.20:**

A subset of a single valued neutrosophic $S^*$ structure space $(S, S^*)$ is said to be a nowhere single valued neutrosophic $S^*$ dense subset in $(S, S^*)$ if $X \setminus A^c$ is everywhere single valued neutrosophic $S^*$ dense subset.

**Definition 3.21:**

Let $(S, S^*)$ be a single valued neutrosophic $S^*$ structure space and $Y$ be a single valued neutrosophic $S^*$ open set in $(S, S^*)$. Then the single valued neutrosophic $S^*$ relative topology $T_Y = \{G \cap Y : G \in S^*\}$ is called the single valued neutrosophic $S^*$ relative (or induced or subspace ) topology on $Y$. The ordered pair $(Y, T_Y)$ is called a single valued neutrosophic $S^*$ subpace of the single valued neutrosophic $S^*$ space $(S, S^*)$.

**Definition 3.22:**

Let $(S,S^*)$ be a single valued neutrosophic $S^*$ structure space.

1.If a family $\{U_\alpha : i \in \Lambda\}$ of single valued neutrosophic $S^*$ open sets in $(S,S^*)$ satisfies the condition $S = \cup\{U_\alpha : i \in \Lambda\}$, then it is called a single valued neutrosophic $S^*$ open cover of $S$. A finite subfamily of the single valued neutrosophic $S^*$ open cover $\{U_\alpha : i \in \Lambda\}$ of $S$, which is also a single valued neutrosophic $S^*$ open cover of $S$, is called a single valued neutrosophic $S^*$ finite subcover .

2.A single valued neutrosophic $S^*$ structure space $(S,S^*)$ is called single valued neutrosophic $S^*$ compact iff every single valued neutrosophic $S^*$ open cover of $S$ has a single valued neutrosophic $S^*$ finite subcover.

**Definition 3.23:**

A single valued neutrosophic $S^*$ Hausdorff space $\delta(S)$ is called an extension of a single valued neutrosophic $S^*$ Hausdorff space $S$ is contained in $\delta(S)$ as an everywhere single valued neutrosophic $S^*$ dense subset.

**Definition 3.24:**

A single valued neutrosophic $S^*$ Hausdorff space $S$ is called single valued neutrosophic $S^*H-$ closed if every extension coincides with $S$ itself.

**Definition 3.25:**

An extension $\delta(S)$ is called a single valued neutrosophic $S^*H-$closed if $\delta(S)$ is single valued neutrosophic $S^*H-$closed and single valued neutrosophic $S^*$ compact if $\delta(S)$ is single valued neutrosophic $S^*$ compact.

**Definition 3.26:**

Let $(S,S^*)$ be a single valued neutrosophic $S^*$ structure space. A system **B** of single valued neutrosophic $S^*$ open sets of a single valued neutrosophic $S^*$ structure space $S$ is called a single valued neutrosophic $S^*$base (or basis) for $(S,S^*)$ if each member of $(S,S^*)$ is a union of members of **B.**A member of **B** is called a single valued neutrosophic $S^*$ basic open set.

**Definition 3.27:**

Let $(S,S^*)$ be a single valued neutrosophic $S^*$ structure space .A system of single valued neutrosophic $S^*$ open sets of a single valued neutrosophic $S^*$structure space $S$ is called a single valued neutrosophic $S^*$ sub base if it together with all possible finite intersections of members of the system form a base of $S$ .

**Lemma 3.28:**

A single valued neutrosophic $S^*$ structure space $S$ is single valued neutrosophic $S^*H-$closed if and only if any single valued neutrosophic $S^*$ centered system $\{U_\alpha\}$ of single valued neutrosophic $S^*$ open sets of $S$ satisfies the condition $\bigcap_\alpha SVNS^*cl(U_\alpha) \neq \phi$.

**Proof:**

**Necessity:** If $p = \{U_\alpha\}$ is single valued neutrosophic $S^*$ centered system with $\bigcap_\alpha SVNS^*cl(U_\alpha) = \phi$ then it can be constructed the following single valued neutrosophic $S^*$ extensions $\delta(S)$ which does not coincide with $S$ and a new member $p$ .The single valued neutrosophic $S^*$ neighbourhoods of each member $A \in S$ in $\delta(S)$ are the same as in $S$ .Any set $U_\alpha$ together with the member $p$ is a single valued neutrosophic $S^*$ neighbourhood of $p$. Because of the condition $\bigcap_\alpha SVNS^*cl(U_\alpha) = \phi$, a single valued neutrosophic $S^*$ structure space $\delta(S)$ is single valued neutrosophic $S^*$Hausdorff and since $\{U_\alpha\}$ is a single valued neutrosophic $S^*$ centered system , it contains $S$ as an everywhere single valued neutrosophic $S^*$dense subset. Therefore $S$ is not a single valued neutrosophic $S^*H-$closed, which is a contradiction.

**Sufficiency:** Let $S$ be a proper everywhere single valued neutrosophic $S^*$ dense subset of $\delta(S)$ .Assume that $\delta(S)$ consists of all single valued neutrosophic $S^*$ neighbourhoods of some member $p \in \delta(S) \setminus S$ .Let this be the system $\{U_\alpha\}$.This system is single valued neutrosophic $S^*$ centered for otherwise $p$ would be an isolated member in $\delta(S)$ and $S$ would not be everywhere single valued neutrosophic $S^*$ dense subset of $\delta(S)$ , since $\delta(S)$ is single valued neutrosophic $S^*$ Hausdorff space then $\bigcap_\alpha SVNS^*cl(V_\alpha^{\delta(S)}) = p$ .But the system $\{V_\alpha = U_\alpha \cap S\}$ is single valued neutrosophic $S^*$ centered and $\bigcap_\alpha SVNS^*cl(V_\alpha^S) = \phi$, which contradicts the condition of the Lemma.

**Lemma 3.29:**

A single valued neutrosophic $S^*$ structure space $S$ is single valued neutrosophic $S^*H-$closed if and only if any maximal single valued neutrosophic $S^*$ centered system $\{U_\alpha\}$ of single valued neutrosophic $S^*$ open sets of $S$ contains all the single valued neutrosophic $S^*$ neighbourhoods of some member.

The proof follows easily from Lemma 3.28.

**Lemma 3.30:**

The single valued neutrosophic $S^*$ structure space $S$ is single valued neutrosophic $S^*H-$ closed if and only if from any single valued neutrosophic $S^*$ cover $\{U_\alpha\}$ of $S$ a finite subsystem $U_i (i = 1,2,3,...n)$ may be chosen such that $\bigcup_{i=1}^{n} SVNS^*cl(U_i) = S$ .The proof follows from Lemma 3.28.

## 4. Single valued neutrosophic $S^*$ centered systems

**Definition 4.1:**

Let $\{q\}$ be a collection of single valued neutrosophic $S^*$ centered (not necessarily maximal) systems of single valued neutrosophic $S^*$ open sets of $S$ .A single valued neutrosophic $S^*$ topology may be defined on this collection.

For if $U$ is a single valued neutrosophic $S^*$ open set of $S$ .Let $O_U$ denote the collection of all single valued neutrosophic $S^*$centered systems $q \in \{q\}$ containing $U$ as an element. All sets of the form $O_U$ form a sub base.

**Definition 4.2:**

Let $\delta(S)$ be an arbitrary single valued neutrosophic $S^*$ extension of $S$ . Every member $A \in \delta(S)$ in particular. A may belong to $S$ defines a certain single valued neutrosophic $S^*$ centered system in $S$ , namely $\left\{V_\alpha^A = S \cap U_\alpha^A\right\}$ where $U_\alpha^A$ runs through all neighbouhoods of $A$ in $\delta(S)$ .

**Note 4.3:**

Every extension of an arbitrary single valued neutrosophic $S^*$ Hausdorff space $S$ can be realized as a single valued neutrosophic $S^*$ structure space of centered systems of single valued neutrosophic $S^*$ open sets of $S$ with an appropriately chosen single valued neutrosophic $S^*$ topology.

**Lemma 4.4:**

For any single valued neutrosophic $S^*$extension $\delta(S)$ , the single valued neutrosophic $S^*$ structure space $\delta_\sigma(S)$ is a single valued neutrosophic $S^*$ extension of $S$ and single valued neutrosophic $S^*\theta-$ homeomorphic to $\delta(S)$ , where $\delta_\sigma(S)$ denote the single valued neutrosophic $S^*$ structure space that is obtained by introducing a single valued neutrosophic $S^*$ topology into a set of single valued neutrosophic $S^*$ centered systems $\left\{V_\alpha^A\right\}$ by the mentioned above.

**Proof:**

Since if $\{V_\alpha^{A_1}\}$ and $\{V_\alpha^{A_2}\}$ are two single valued neutrosophic $S^*$ centered systems constructed relative to single valued neutrosophic sets $A_1$ and $A_2$, $\delta_\sigma(S)$ is single valued neutrosophic $S^*$ Hausdorff space. Since $S$ is a single valued neutrosophic $S^*$ Hausdorff space, by the above procedure, they contain disjoint elements. The relation $A \in U$ and $q_A \in O_U$ are equivalent so that $S$ is single valued neutrosophic $S^*$ homeomorphic to a subset of $\delta_\sigma(S)$ .Since $O_U \cap O_V = O_{U \cap V}$ and since $O_U$ contains all the $q_A$ for which $A \in U$, it follows that $S$ is everywhere single valued neutrosophic $S^*$ dense in $\delta_\sigma(S)$, that is $\delta_\sigma(S)$ is a single valued neutrosophic $S^*$ extension of $S$.

Next to prove that $\delta_\sigma(S)$ and $\delta(S)$ are single valued neutrosophic $S^*\theta$ -homeomorphic. There is a single valued neutrosophic $S^*$ one-to-one correspondence between the members of $\delta(S)$ and $\delta_\sigma(S)$ which is denoted by $i$ .Thus $i(A) = A$ if $A \in S$ .Let $A' \in \delta_\sigma(S)$ , $A' \in O_V, i(A) = A'$ and let $U$ be a single valued neutrosophic $S^*$ neighbourhood of $A$ in $\delta(S)$ such that $U \cap S = V$ .We prove that $i(U) = O_V$ .This shows that the function $i$ is single valued neutrosophic $S^*$continuous and hence it is single valued neutrosophic $S^*\theta-$ continuous. But this is obvious because if $A_1 \in U$ then $V \in i(A_1)$ and hence $i(A_1) \in O_V$ .

To prove that the inverse function is single valued neutrosophic $S^*\theta$-continuous. Let $U$ be a single valued neutrosophic $S^*$ neighbourhood of $i(A)$, where $V = S \cap U$ .To show that $i^{-1}(SVNS^*cl(O_V)) \subset SVNS^*cl(U)$. Let $A' \in SVNS^*cl(O_V)$. This means that an arbitrary single valued neutrosophic $S^*$ neighbourhood $O_G$ of $A'$ meets $O_V$, that is $G \cap V \neq \phi$ and this in turns means that an arbitrary single valued neutrosophic $S^*$ neighbourhood of $i^{-1}(A')$ meets $V$ that is $i^{-1}(A') \in SVNS^*cl(V) = SVNS^*cl(U)$ .Thus $i^{-1}(SVNS^*cl(O_V)) \subset SVNS^*cl(U)$ and the Lemma is proved.

**Definition 4.5:**

A single valued neutrosophic $S^*$ extension $\delta(S)$ is of type $\sigma$ if the function $i$ (one – to-one correspondence between the members of $\delta(S)$ and $\delta_\sigma(S)$ ) is a single valued neutrosophic $S^*\theta-$ homeomorphism.

**Definition 4.6:**

A single valued neutrosophic $S^*$ extension $\delta(S)$ is of type $\tau$ if the set $\delta(S) \setminus S$ is discrete in the single valued neutrosophic $S^*$ relative topology.

**Proposition 4.7:**

Every single valued neutrosophic $S^*$ extension of $S$ is a single valued neutrosophic $S^*\theta-$ homeomorphic to some extension of type $\sigma$ of the same space.

**Proof:**

The proof follows from the fact that the single valued neutrosophic $S^*$ extension $\delta_\sigma(S)$ in Lemma 4.4 is of type $\sigma$.

Now, let $\delta(S)$ be any single valued neutrosophic $S^*$ extension. Let $\delta_\tau(S)$ denote the single valued neutrosophic $S^*$ structure space obtained as follows. The members of $\delta_\tau(S)$ are those of $\delta(S)$ .The single valued neutrosophic $S^*$ neighbourhoods of members of $A \in S$ are same as in S ,but for members $A \in \delta(S) \setminus S$ the single valued neutrosophic $S^*$ neighbourhoods are obtained from those of $A$ in $\delta(S)$ by rejecting the set $\delta(S) \setminus S \cup A$ .Clearly $\delta_\tau(S)$ is asingle valued neutrosophic $S^*$ Hausdorff space.

**Definition 4.8:**

Let $(S_1, S_1^*)$ and $(S_2, S_2^*)$ be two single valued neutrosophic $S^*$ structure spaces. A single valued neutrosophic $S^*$ structure space $(S_1, S_1^*)$ is said to be topologically embedded in another single valued neutrosophic $S^*$ structure space $(S_2, S_2^*)$ if $(S_1, S_1^*)$ is a single valued neutrosophic $S^*$ homeomorphic to a single valued neutrosophic $S^*$ subspace of $(S_2, S_2^*)$.

**Lemma 4.9:**

For any single valued neutrosophic $S^*$ extension $\delta(S)$ , the single valued neutrosophic $S^*$ structure space $\delta_\tau(S)$ is a single valued neutrosophic $S^*$ extension of $S$ , single valued neutrosophic $S^*\theta-$ homeomorphic to $\delta(S)$ and of type $\tau$.

**Proof:**

It is clear that $S$ is single valued neutrosophic $S^*$ topologically embedded in $\delta_\tau(S)$ as an everywhere single valued neutrosophic $S^*$ dense subset , that is , $\delta_\tau(S)$ is a single valued neutrosophic $S^*$ extension of $S$ .

From the construction of $\delta_\tau(S)$, $\delta_\tau(S) \setminus S$ is discrete and hence $\delta_\tau(S)$ is of type $\tau$ .It remains to show that $\delta_\tau(S)$ and $\delta(S)$ are single valued neutrosophic $S^*\theta-$ homeomorphic.

This follows from the fact that if $U$ is single valued neutrosophic $S^*$ open set in $S$ , then $(SVNS^*cl(U))^{\delta(S)} = (SVNS^*cl(U))^{\delta_\tau(S)}$ .Then the single valued neutrosophic $S^*$ structure space $\delta_\tau(S)$ is mapped continuously onto $\delta(S)$ .

**Note 4.10:**

From Lemma 4.9 each single valued neutrosophic $S^*$ extension $\delta(S)$ of $S$ is associated with single valued neutrosophic $S^*$ extensions $\delta_\sigma(S)$ and $\delta_\tau(S)$, of types $\sigma$ and $\tau$ respectively and single valued neutrosophic $S^*\theta-$ homeomorphic to each other and also single valued neutrosophic $S^*\theta-$ homeomorphic to the original single valued neutrosophic $S^*$ extension $\delta(S)$.

**Definition 4.11:**

Let **G** be a single valued neutrosophic $S^*$ base of single valued neutrosophic $S^*$ open sets in a single valued neutrosophic $S^*$ structure space $S$ and $\sigma_G(S)$, the single valued neutrosophic $S^*$ structure space whose elements are the members of $S$ itself and all the maximal single valued neutrosophic $S^*$ centered systems $\{U_\alpha\}$ consisting of single valued neutrosophic $S^*$ open sets belonging to **G** , none of which contains as a subsystem of the single valued neutrosophic $S^*$ neighbourhoods of any single valued neutrosophic $S^*$ open set of $S$ belonging to **G** (Clearly this condition is equivalent to the following : $\bigcap_\alpha SVNS^*cl(U_\alpha)=\phi$).

**Definition 4.12:**

A single valued neutrosophic $S^*$ topology is defined in $\sigma_G(S)$ as follows. If $U\in$**G,** $O_U$ denotes the set of all $A\in U$ and all maximal single valued neutrosophic $S^*$ centered system in a $\sigma_G(S)$ that contains $U$ as an element .Since in $\sigma_G(S)$ each member $A\in S$ can be replaced by the single valued neutrosophic $S^*$ centered system of all its single valued neutrosophic $S^*$ neighbourhoods belonging to **G** (with the single valued neutrosophic $S^*$ topologization : $\{U_\alpha\}\in O_U$ if $U\in\{U_\alpha\}$).It is clear that each $\sigma_G(S)$ is a single valued neutrosophic $S^*$ Hausdorff extension of type $\sigma$ of the original single valued neutrosophic $S^*$ structure space $S$ .

**Definition 4.13:**

A single valued neutrosophic $S^*$ centered system $\{U_\alpha\}$ of single valued neutrosophic $S^*$ open sets of **G** is called a single valued neutrosophic $S^*$ Hausdorff system if for every $B\in S$ not belonging to $U\in\{U_\alpha\}$ there exists a $U'\in\{U_\alpha\}$ such that $B\notin SVNS^*cl(U')$.

**Definition 4.14:**

A maximal single valued neutrosophic $S^*$Hausdorff system (that is, one which cannot be extended while remaining single valued neutrosophic $S^*$centered system and a single valued neutrosophic $S^*$ Hausdorff space) is called a single valued neutrosophic $S^*$ Hausdorff end.

**Note 4.15:**

A single valued neutrosophic $S^*$structure space $\sigma_G(S)$ associated with the base containing all the single valued neutrosophic $S^*$ open sets of $S$ will simply be denoted by $\sigma(S)$.

**Proposition 4.16:**

A single valued neutrosophic $S^*$ extension $\sigma(S)$ is a single valued neutrosophic $S^*$ $H-$closed extension of $S$.

**Proof:**

Let $\{U_\alpha\}$ be an arbitrary single valued neutrosophic $S^*$ centered system of single valued neutrosophic $S^*$ open sets of $\sigma(S)$.Let $V_\alpha = U_\alpha \cap S$.

Since $S$ is everywhere single valued neutrosophic $S^*$ dense in $\sigma(S)$, $(SVNS*cl(V_\alpha))^{\sigma(S)} = (SVNS*cl(U_\alpha))^{\sigma(S)}$ .Hence it is enough to show that $\bigcap_\alpha (SVNS*cl(V_\alpha))^{\sigma(S)} \neq \phi$.If $\bigcap_\alpha (SVNS*cl(V_\alpha))^{S} \neq \phi$, then by Lemma 3.28, there is nothing to prove.

If $\bigcap_\alpha (SVNS*cl(U_\alpha))^{S} = \phi$, then there exists a single valued neutrosophic $S^*$ Hausdorff end $p$ containing all the sets $V_\alpha$, and hence $p \in (SVNS*cl(V_\alpha))^{\sigma(S)}$ for all $\alpha$.

**Note 4.17:**

Let **G** be any single valued neutrosophic $S^*$ base of $S$ .If $U \in$ **G** then **G** is called single valued neutrosophic $S^*$ algebraically closed.

**Remark 4.18:**

If **G** is called single valued neutrosophic $S^*$ algebraically closed base of $S$, then $\sigma_G(S)$ is a single valued neutrosophic $S^*H-$closed extension of $S$.

The proof is same as that of Proposition 4.16.

**Note 4.19:**

Each single valued neutrosophic $S^*$ extension $\delta(S)$ is associated with single valued neutrosophic $S^*\,\theta-$homeomorphic extension $\delta_\tau(S)$ of type $\tau$ ,the single valued neutrosophic $S^*$structure space $\sigma_\tau(S)$ which is associated with $\sigma(S)$ is denoted by $\tau(S)$ and is called a single valued neutrosophic $S^*$Katetov extension of $S$.

**Lemma 4.20:**

A single valued neutrosophic $S^*\theta-$continuous image of a single valued neutrosophic $S^*H-$ closed space is a single valued neutrosophic $S^*H-$ closed.

**Proof:**

Let $f$ be a single valued neutrosophic $S^*\theta-$continuous function from a single valued neutrosophic $S_1^*H-$closed space $S_1$ onto single valued neutrosophic $S_2^*H-$closed space $S_2$.Suppose that $S_2$ is not a single valued neutrosophic $S_2^*H-$closed ,then by Lemma 3.30 there exists a single valued neutrosophic $S_2^*$ covering $\{U_\alpha\}$ of $S_2$ from which a finite number of single valued neutrosophic $S_2^*$ open sets cannot be extracted whose single valued neutrosophic $S^*$closures cover $S_2$.Let $A_1 \in S_1$ and $V_\beta$ be a single valued neutrosophic $S_1^*$open set of $S_1$ such that $A_1 \in V_\beta$and $f(SVNS_1^*cl(V_\beta)) \subset \left(SVNS_1^*cl(U_\alpha)\right)$.

Choosing such a set for each member of $S_1$ ,the collection $\{V_\beta\}$ of this single valued neutrosophic $S_1^*$ structure space is obtained. A finite number of sets $V_1, V_2, ....V_n$ is picked such that $\bigcup_{i=1}^{n} SVNS_1^*cl(V_i) = S_1$.But then the union $\bigcup_{i=1}^{n} SVNS_2^*cl(U_{\alpha_i}) \supseteq \bigcup_{i=1}^{n} f(SVNS_1^*cl(V_i)) = f(S_i) = S_2$. But in general $\bigcup_{i=1}^{n} SVNS_2^*cl(U_{\alpha_i}) \subseteq S_2$ implies that $\bigcup_{i=1}^{n} f(SVNS_1^*cl(V_i)) = S_2$ is the whole of $S_2$,which is impossible by hypothesis.

**Remark 4.21:**

The single valued neutrosophic $S^*$structure space $\tau(S)$is a single valued neutrosophic $S^*H-$ closed extension of $S$ .

The proof follows from Proposition 4.16 and Lemma 4.20.

**Note 4.22:**

A single valued neutrosophic $S^*$ structure space $\tau(S)$ has the following maximal properties.

**Proposition 4.23:**

If $\delta(S)$ is any (not necessarily single valued neutrosophic $S^*H-$ closed) single valued neutrosophic $S^*$ extension of $S$ then there exists a subset $\tau_\delta(S) \subseteq \tau(S)$ containing $S$ and a single valued neutrosophic $S^*$ continuous function $f_\delta$ of this subset onto $\delta(S)$ such that $f_\delta(A) = A$, where $A \in S$. Here if $\delta(S)$ is a single valued neutrosophic $S^*H-$ closed extension, it may be assumed that $\tau_\delta(S) = \tau(S)$.

**Proof:** Let $\delta(S)$ be a single valued neutrosophic $S^*$ extension of $S$ .Each member $q \in \delta(S) \setminus S$ defines a single valued neutrosophic $S^*$ centered system of single valued neutrosophic $S_1^*$ open sets in $S$ ,namely $q$ defines $\{V_\alpha\} = \{U_\alpha \cap S\}$ where $U_\alpha$ is the set of all single valued neutrosophic $S_1^*$ neighbourhoods of $q$ in $\delta(S)$ .It can be further identified each member of $\delta(S)$ with the corresponding single valued neutrosophic $S^*$ centered system $\{V_\alpha\}$.Because $\delta(S)$ is a fuzzy neutrosophic $S^*$ Hausdorff space the system $\{V_\alpha\}$has the property $\bigcap_\alpha (SVNS * cl(V_\alpha))^S = \phi$.

Consider in $\tau(S)$,the subset $\tau_\delta(S)$ consisting of all members of $S$ and all single valued neutrosophic $S_2^*$ ends containing at least one system $\{V_\alpha\}$corresponding to some $q \in \delta(S)$.The function $f_\delta$ is constructed as follows: if $A \in S$,put $f_\delta(A) = A$,while if $p \in \tau(S) \setminus S$,then $p$ contains some $q$. As $q$ is unique put $f_\delta(p) = q$.

Clearly, $f$ is a single valued neutrosophic $S^*$ continuity at every $A \in S$.Because $S$ is a single valued neutrosophic $S_2^*$ open in $\tau(S)$ (by definition of the single valued neutrosophic $S^*$ topology of $\tau(S)$), and hence also in $\tau_\delta(S)$.Let $p \in \tau_\delta(S) \setminus S$ and $f_\delta(p) = q$.

Let $U_\alpha$ be a single valued neutrosophic $S^*$neighbourhood of $q$ in $\delta(S)$.Then the set $V_\alpha \cup p$ is a single valued neutrosophic $S^*$ neighbourhood of $p$ in $\tau_\delta(S)$,where $V_\alpha = U_\alpha \cap S$ with $f_\delta(V_\alpha \cup p) \subseteq U_\alpha$, that is, $f_\delta$ is single valued neutrosophic $S^*$ continuous at $p$.

Suppose that $\delta(S)$ is a single valued neutrosophic $S^*H-$ closed extension. Let $p \in \tau(S) \setminus S$,and let $\{U_\alpha\}$be the system of all single valued neutrosophic $S_2^*$ neighbourhoods of $p$ in $\tau(S)$and let $V_\alpha = U_\alpha \cap S$.Let $H_\alpha$ denote a single valued neutrosophic $S_1^*$ open set in $\delta(S)$ such that $V_\alpha = H_\alpha \cap S$.

The system $\{H_\alpha\}$is a single valued neutrosophic $S_1^*$centered system and since $\delta(S)$ is a single valued neutrosophic $S_1^*H-$ closed ,then by Lemma 3.28, $\bigcap_\alpha (SVNS_1^* cl(H_\alpha)) \neq \phi$ .Let $q \in \bigcap_\alpha (SVNS_1^* cl(H_\alpha))$.If $G$ is a single valued neutrosophic $S_1^*$ neighbourhood of $q$ in $\delta(S)$,we have $G \cap V_\alpha \neq \phi$ for every $\alpha$ ,that is , $(G \cap S) \in \{V_\alpha\}$.This means that $p$ contains the single valued neutrosophic $S_1^*$ centered system $q$ and $\tau(S) \subseteq \tau_\delta(S)$,that is $\tau_\delta(S) = \tau(S)$.

**Remark 4.24:**

$\tau_{\delta\sigma}(S)$ denotes the single valued neutrosophic $S^*$ structure space obtained from $\tau_\delta(S)$ by the procedure described in section 4. It is easy to see that $\tau_{\delta\sigma}(S)$ is a single valued neutrosophic $S^*\theta-$ homeomorphic to a subset of single valued neutrosophic $S^*$ extension $\sigma(S)$.As $\tau_{\delta\sigma}(S)$ is a single valued neutrosophic $S^*\theta-$homeomorphic to $\tau_\delta(S)$, Proposition 4.23 holds if $\tau(S)$ is replaced by $\sigma(S)$ and single valued neutrosophic $S^*$ continuity by single valued neutrosophic $S^*\theta-$continuity.

**Remark 4.25:**

A single valued neutrosophic $S^*$ structure space $\sigma(S)$ can be mapped single valued neutrosophic $S^*\theta-$continuity onto any single valued neutrosophic $S^*H-$ closed extension of $S$ in such a way that the members of $S$ remain fixed.

Now, the classes of single valued neutrosophic $S^*$ Hausdorff extensions of $S$ are discussed.

**Lemma 4.26:**

A single valued neutrosophic $S^*$ open set $\tau_\delta(S)$ is the largest subset of $\tau(S)$ that can be continuously mapped onto $\delta(S)$ in such a way that the members of $S$ remain fixed. In other words, if a set $\tau_\delta^{'}(S)$ is continuously mapped onto $\delta(S)$ in such a way that the members of $S$ remain fixed, then $\tau_\delta^{'}(S)\subseteq\tau_\delta(S)$.

**Proof:**

Let $p\in\tau_\delta^{'}(S)\setminus S$ and let $f_\delta^{'}(p)=q$,where $f_\delta^{'}$ is a single valued neutrosophic $S^*$ continuous function of $\tau_\delta^{'}(S)$ onto $\delta(S)$ .Let $U$ be a single valued neutrosophic $S^*$ neighbourhood of $q$ in $\delta(S)$ .There exists a single valued neutrosophic $S^*$ neighbourhood $H$ of $p$ in $\tau(S)$ such that $f_\delta^{'}(H)\subseteq U$ .Then ,$H\cap S\subseteq U\cap S$ that is , $p$ contains $U\cap S$ and since $U$ is any single valued neutrosophic $S^*$ neighbourhood of $q$, $p$ contains the system $q$,that is, $p\in\tau_\delta(S)$.

**Note 4.27:**

Thus, all single valued neutrosophic $S^*$ extensions of $S$ fall into classes, where $\delta(S)$ and $\delta'(S)$ are in the same class if and only if $\tau_\delta(S)=\tau_\delta^{'}(S)$. All single valued neutrosophic $S^*H-$ closed extensions belong to the same class, by Lemma 4.20 contains only single valued neutrosophic $S^*H-$closed extensions.

**Lemma 4.28:**

If single valued neutrosophic $S^*$ extension $\delta(S)$ and $\gamma(S)$ are single valued neutrosophic $S^*\theta-$homeomorphic, then they belong to the same class, that is $\tau_\delta(S)=\tau_\gamma(S)$.

**Proof:**

Let $i$ be a single valued neutrosophic $S^*\theta-$ homeomorphism between $\delta(S)$ and $\gamma(S)$ such that $i(A)=A$ for $A\in S$ .Let $\{U_\alpha\}=\{V_\alpha\cap S\}$ ,where $V_\alpha$ is a single valued neutrosophic $S^*$ neighbourhood of $p\in\tau(S)\setminus S$ .Let $\{H_\alpha\}=\{G_\alpha\cap S\}$,where $G_\alpha$ is a single valued neutrosophic $S^*$ neighbourhood of $i(p)=q$ in $\gamma(S)$ .If some single valued neutrosophic $S^*$ end $d$ of $S$ contains all the single valued neutrosophic sets $U_\alpha$,then it also contains all the $H_\alpha$.Choose some $H_\alpha$ and a $G_\alpha$ such that $G_\alpha \cap S = H_\alpha$ ,and in $\delta(S)$ choose $V_\beta$ such that $i(SVNS * cl(V_\beta)) \subseteq SVNS * cl(G_\alpha)$ .Then , $SVNS * cl(V_\beta) \cap S \subseteq SVNS * cl(G_\alpha) \cap S$ .That is , $V_\beta \cap S = U_\beta \subseteq SVNS * \text{int}(SVNS * cl(G_\alpha) \cap S)$ .Hence ,if $SVNS * \text{int}(SVNS * cl(G_\alpha) \cap S) \in d$ , then $G_\alpha \cap S = V_\alpha$ as the everywhere single valued neutrosophic $S^*$ dense subset, $SVN\,\text{int}(SVNcl(G_\alpha) \cap S)$ also belongs to *d*.

Thus, $\tau_\delta(S) \subset \tau_\gamma(S)$.Similarly, $\tau_\delta(S) \supseteq \tau_\gamma(S)$.That is $\tau_\delta(S) = \tau_\gamma(S)$.

## References

1. Alexandrov P. S and Urysohn P.S, On compact topological spaces, Trudy Mat. Inst. Steklov, 31(1950).
2. I. Arockiarani, J. Martina Jency, More on single valued neutrosophic sets and fuzzy neutrosophic topological spaces, International journal of innovative research and studies, 3 (5) (2014), 643-652.
3. D. Coker and A.H. Es, On fuzzy compactness in intuitionistic fuzzy topological spaces, J. Fuzzy Math., 3(1996), 899-909.
4. Gleason A.M, Projective topological spaces, Illinois J. Math., 2(1958).
5. Wang, F. Smarandache, Y. Zhang and R. Sunderraman. Single valued Neutrosophic Sets, Multi-space and Multi-structure, 4 (2010), 410-413.
6. Iliadis.S and Fomin. S, The method of centered systems in the theory of topological spaces, UMN,21, 1996,47-66.
7. Ponomarev V.I., Paracompacta, their projective spectra and continuous mappings, Mat. Sb. (N.S.), (1963) 60 (120), volume (1), 89-119.
8. F. Smarandache. A unifying field of logics. Neutrosophy: neutrosophic probability, set and logic, American Research Press, Rehoboth, 1998
9. Smarandache, F. Neutrosophic set, a generalization of the intuitionistic fuzzy sets, Inter. J. Pure Appl.Math.,24 (2005),287-297.
10. Uma M.K, Roja. E. and Balasubramanian. G, The method of centered systems in fuzzy topological spaces, The Journal of Fuzzy Mathematics, 15 (2007),1-7.

**SAEID JAFARI[1], I. AROCKIARANI[2], J. MARTINA JENCY[3]**

1 College of Vestsjaelland South,Herrestraede 11,4200, Slagelse, Denmark.
2,3* Department of Mathematics, Nirmala College for women, Coimbatore, Tamilnadu, India.
3* E-mail: martinajency@gmail.com

# The Alexandrov-Urysohn Compactness On Single Valued Neutrosophic $S^*$ Centered Systems

## Abstract

In this paper we present the notion of the single valued neutrosophic $S^*$ maximal compact extension in single valued neutrosophic $S^*$ centered system. Moreover, the concept of single valued neutrosophic $S^*$ absolute is applied to establish the Alexandrov -Urysohn compactness criterion. Some of the basic properties are characterized.

## 1. Introduction

Florentin Smarandache [9] combined the non- standard analysis with a tri component logic/set, probability theory with philosophy and proposed the term neutrosophic which means knowledge of neutral thoughts. This neutral represents the main distinction between fuzzy and intuitionistic fuzzy logic set. In 1998, Florentin Smarandache [6] defined the single valued neutrosophic set involving the concept of standard analysis. Stone [10, 11] applied the apparatus of Boolean rings to investigate spaces more general than completely regular ones, related to some extent to the function-theoretic approach. Using these methods tone [10, 11] obtained a number of important results on Hausdorff spaces and in fact introduced the important topological construction that was later called the absolute. The first proof of Alexandrov-Urysohn compactness criterion without any axiom of countability was given by Stone [10, 11].Cech extension in topological spaces and Alexandrov-Urysohn compactness criterion were constructed by Iliadis and Fomin[7].

In this paper, the concept of absolute in single valued neutrosophic $S^*$ structure space and the single valued neutrosophic $S^*$ maximal compact extension β(S) (single valued neutrosophic $S^*$ cech extension) of an arbitrary single valued neutrosophic $S^*$ completely regular space is introduced. Further, the Alexandrov -Urysohn compactness criterion on single valued neutrosophic $S^*$ structure has been studied.

## 2. Preliminaries

**Definition 2.1: [6]**

Let $X$ be a space of points (objects), with a generic element in $X$ denoted by $x$. A single valued neutrosophic set (SVNS) $A$ in $X$ is characterized by truth-membership function $T_A$, indeterminacy-membership function $I_A$ and falsity-membership function $F_A$.

For each point $x$ in $X$, $T_A(x), I_A(x), F_A(x) \in [0,1]$. When $X$ is continuous, a SVNS $A$ can be written as $A, \int_X \langle T_A(x), I_A(x), F_A(x)\rangle / x, x \in X$ .

When $X$ is discrete, a SVNS $A$ can be written as

$$A = \sum_{i=1}^{n} \langle T(x_i), I(x_i), F(x_i)\rangle / x_i, x_i \in X$$

**Definition 2.2: [3]**

Let X be a non- empty set and S a collection of all single valued neutrosophic sets of X.A single valued neutrosophic $S^*$ structure on S is a collection $S^*$ of subsets of S having the following properties

1. φ and S are in $S^*$.
2. The union of the elements of any sub-collection of $S^*$ is in $S^*$.
3. The intersection of the elements of any finite sub-collection of $S^*$ is in $S^*$.

The collection S together with the structure $S^*$ is called single valued neutrosophic $S^*$ structure space. The members of $S^*$ are called single valued neutrosophic $S^*$ open sets. The complement of single valued neutrosophic $S^*$ open set is said to be a single valued neutrosophic $S^*$ closed set.

**Example 2.3: [3]**

Let $X = \{a,b\}$ , $S = \left\{ \frac{a}{\langle 0.8,0.3,0.5\rangle}, \frac{b}{\langle 0.7,0.4,0.6\rangle} \right\}$, $S^* = \{S, \phi, S_1, S_2, S_3, S_4\}$ where,

$S_1 = \left\{ \frac{a}{\langle 0.6,0.1,0.7 \rangle}, \frac{b}{\langle 0.5,0.2,0.8 \rangle} \right\}$, $S_2 = \left\{ \frac{a}{\langle 0.4,0.2,0.6 \rangle}, \frac{b}{\langle 0.5,0.3,0.9 \rangle} \right\}$ ,

$S_3 = \left\{ \frac{a}{\langle 0.4,0.1,0.7 \rangle}, \frac{b}{\langle 0.5,0.2,0.9 \rangle} \right\}, S_4 = \left\{ \frac{a}{\langle 0.6,0.2,0.6 \rangle}, \frac{b}{\langle 0.5,0.3,0.8 \rangle} \right\}$.

Here $(S, S^*)$ is a structure space.

**Definition 2.4: [3]**

Let $A$ be a member of $S$. A single valued neutrosophic S* open set U in $(S, S^*)$ is said to be a single valued neutrosophic S* open neighborhood of $A$ if $A \in G \subset U$ for some single valued neutrosophic S* open set $G$ in $(S, S^*)$.

**Example 2.5: [3]**

Let $X = \{a,b\}$ , $S = \left\{ \frac{a}{\langle 0.8,0.3,0.5 \rangle}, \frac{b}{\langle 0.7,0.4,0.6 \rangle} \right\}$, $S^* = \{S, \phi, S_1, S_2, S_3, S_4\}$ where,

$S_1 = \left\{ \frac{a}{\langle 0.6,0.1,0.7 \rangle}, \frac{b}{\langle 0.5,0.2,0.8 \rangle} \right\}$, $S_2 = \left\{ \frac{a}{\langle 0.4,0.2,0.6 \rangle}, \frac{b}{\langle 0.5,0.3,0.9 \rangle} \right\}$ ,

$S_3 = \left\{ \frac{a}{\langle 0.4,0.1,0.7 \rangle}, \frac{b}{\langle 0.5,0.2,0.9 \rangle} \right\}, S_4 = \left\{ \frac{a}{\langle 0.6,0.2,0.6 \rangle}, \frac{b}{\langle 0.5,0.3,0.8 \rangle} \right\}$.

Let $A = \left\{ \frac{a}{\langle 0.4,0.1,0.8 \rangle}, \frac{b}{\langle 0.3,0.1,0.9 \rangle} \right\}$.

Here, $A \in S_1 \subset S_4$ . $S_4$ is the single valued neutrosophic S* open neighbourhood of $A$ .

**Definition 2.6: [3]**

Let $(S, S^*)$ be a single valued neutrosophic S* structure space and $A = \langle x, T_A, I_A, F_A \rangle$ be a single valued neutrosophic set in X. Then the single valued neutrosophic S* closure of A (briefly SV N S*cl(A)) and single valued neutrosophic S* interior of A (briefly SVN S*int(A)) are respectively defined by

SVN S*cl(A) $= \bigcap$ {K: K is a single valued neutrosophic S* closed sets in S and A $\subseteq$ K}

SVN S*int(A) = $\bigcup$ {G: G is a single valued neutrosophic S* open sets in S and G $\subseteq$ A}.

**Example 2.7: [3]**

Let $X=\{a,b\}$ , $S=\left\{\frac{a}{\langle 0.8,0.3,0.5\rangle},\frac{b}{\langle 0.7,0.4,0.6\rangle}\right\}$, $S^{*}=\{S,\phi,S_1,S_2,S_3,S_4\}$ where,

$S_1=\left\{\frac{a}{\langle 0.6,0.1,0.7\rangle},\frac{b}{\langle 0.5,0.2,0.8\rangle}\right\}$, $S_2=\left\{\frac{a}{\langle 0.4,0.2,0.6\rangle},\frac{b}{\langle 0.5,0.3,0.9\rangle}\right\}$ ,

$S_3=\left\{\frac{a}{\langle 0.4,0.1,0.7\rangle},\frac{b}{\langle 0.5,0.2,0.9\rangle}\right\}$, $S_4=\left\{\frac{a}{\langle 0.6,0.2,0.6\rangle},\frac{b}{\langle 0.5,0.3,0.8\rangle}\right\}$.

$S_1^c=\left\{\frac{a}{\langle 0.7,0.9,0.6\rangle},\frac{b}{\langle 0.8,0.8,0.5\rangle}\right\}$, $S_2^c=\left\{\frac{a}{\langle 0.6,0.8,0.4\rangle},\frac{b}{\langle 0.9,0.7,0.5\rangle}\right\}$,

$S_3^c=\left\{\frac{a}{\langle 0.7,0.9,0.4\rangle},\frac{b}{\langle 0.9,0.8,0.5\rangle}\right\}$, $S_4^c=\left\{\frac{a}{\langle 0.6,0.8,0.6\rangle},\frac{b}{\langle 0.8,0.7,0.5\rangle}\right\}$.

Let $A=\left\{\frac{a}{\langle 0.5,0.3,0.6\rangle},\frac{b}{\langle 0.7,0.4,0.9\rangle}\right\}$. Then $SVNS^{*}\mathrm{int}(A)=\{S_3\}$.

$SVNS^{*}cl(A)=\{S_4^c\}$.

**Definition 2.8: [3]**

The ordered pair (S, S*) is called a single valued neutrosophic S* Hausdorff space if for each pair $A_1, A_2$ of disjoint members of S, there exist disjoint single valued neutrosophic S* open sets $U_1$ and $U_2$ such that $A_1 \subseteq U_1$ and $A_2 \subseteq U_2$ .

**Example 2.9: [3]**

Let $X=\{a,b\}$ , $S=\left\{\frac{a}{\langle 1,1,0\rangle},\frac{b}{\langle 1,1,0\rangle}\right\}$, $S^{*}=\{S,\phi,S_1,S_2,S_3\}$ where,

$S_1=\left\{\frac{a}{\langle 0.5,0,1\rangle},\frac{b}{\langle 0,0.3,0.4\rangle}\right\}$, $S_2=\left\{\frac{a}{\langle 0.5,0.2,0.5\rangle},\frac{b}{\langle 0.7,0.3,0.4\rangle}\right\}$, $S_3=\left\{\frac{a}{\langle 0,0.2,0.5\rangle},\frac{b}{\langle 0.7,0,1\rangle}\right\}$.

Let $A_1=\left\{\frac{a}{\langle 0.3,0,1\rangle},\frac{b}{\langle 0,0.1,1\rangle}\right\}$, $A_2=\left\{\frac{a}{\langle 0,0.1,0.6\rangle},\frac{b}{\langle 0.5,0,1\rangle}\right\}$.

Here $A_1$ and $A_2$ are disjoint members of $S$ and $S_1, S_2$ are disjoint single valued neutrosophic S* open sets such that $A_1 \subseteq S_1$ *and* $A_2 \subseteq S_2$.

Hence the ordered pair $(S,S^{*})$ is a single valued neutrosophic S* Hausdorff space.

**Definition 2.10: [3]**

Let $(S_1,S_1^*)$ and $(S_2,S_2^*)$ be any two single valued neutrosophic S* structure spaces and let $f:(S_1,S_1^*)\rightarrow(S_2,S_2^*)$ be a function. Then $f$ is said to be single valued neutrosophic S* continuous iff the pre image of each single valued neutrosophic $S_2^*$ open set in $(S_2,S_2^*)$ is a single valued neutrosophic $S_1^*$ open set in $(S_1,S_1^*)$.

**Definition 2.11: [3]**

Let $(S_1,S_1^*)$ and $(S_2,S_2^*)$ be any two single valued neutrosophic S* structure spaces and let $f:(S_1,S_1^*)\rightarrow(S_2,S_2^*)$ be a bijective function. If both the functions $f$ and the inverse function $f^{-1}:(S_2,S_2^*)\rightarrow(S_1,S_1^*)$ are single valued neutrosophic S* continuous then $f$ is called single valued neutrosophic S* homeomorphism.

**Definition 2.12: [4]**

Let $f$ be a function from a single valued neutrosophic S* structure space $(S_1,S_1^*)$ into a single valued neutrosophic S* structure space $(S_2,S_2^*)$ with $f(A_1)=f(A_2)$ where $A_1\in(S_1,S_1^*)$ and $A_2\in(S_2,S_2^*)$.Then $f$ is called a single valued neutrosophic $S^*\theta$ continuous at $A_1$ if for every neighbourhood $O_{A_2}$ of $A_2$, there exists a neighbourhood $O_{A_1}$ of $A_1$ such that $f(SVNS*cl(O_{A_1}))\subset SVNS*cl(O_{A_2})$.The function is called single valued neutrosophic $S^*\theta-$ continuous if it is single valued neutrosophic $S^*\theta-$ continuous at every member of $S_1$.

**Definition 2.13: [3]**

A function is called a single valued neutrosophic $S^*\theta-$ homeomorphism if it is single valued neutrosophic $S^*$ one to one and single valued neutrosophic $S^*\theta-$ continuous in both directions.

**Definition 2.14: [3]**

Let $(S, S^*)$ be a single valued neutrosophic $S^*$ Hausdorff space. A system $p=\{U_\alpha : \alpha=1,2,3,...n\}$ of single valued neutrosophic $S^*$ open sets is called a single valued neutrosophic $S^*$ centered system if any finite collection of the sets of the system has a non- empty intersection .

**Example 2.15: [3]**

Let $X=\{a,b\}$ , $S=\left\{\frac{a}{\langle 1,1,0\rangle},\frac{b}{\langle 1,1,0\rangle}\right\}$, $S^*=\{S,\phi,S_1,S_2,S_3\}$ where,

$S_1 = \left\{ \frac{a}{\langle 0.5,0,1 \rangle}, \frac{b}{\langle 0,0.3,0.4 \rangle} \right\}$, $S_2 = \left\{ \frac{a}{\langle 0.5,0.2,0.5 \rangle}, \frac{b}{\langle 0.7,0.3,0.4 \rangle} \right\}$, $S_3 = \left\{ \frac{a}{\langle 0,0.2,0.5 \rangle}, \frac{b}{\langle 0.7,0,1 \rangle} \right\}$.

Let $A_1 = \left\{ \frac{a}{\langle 0.3,0,1 \rangle}, \frac{b}{\langle 0,0.1,1 \rangle} \right\}$ , $A_2 = \left\{ \frac{a}{\langle 0,0.1,0.6 \rangle}, \frac{b}{\langle 0.5,0,1 \rangle} \right\}$ .Let us consider the system $p_1 = \{S_\alpha, \alpha = 1,2,3\}$. $p_1$ is a single valued neutrosophic S* centered system since $S_1, S_2$ has a non -empty intersection.

Let $p_2 = \{S_\alpha : \alpha = 1,2\}$ is also a single valued neutrosophic S* centered system.

Here $p_1$ is a maximal single valued neutrosophic S* centered system.

**Definition 2.16: [3]**

The single valued neutrosophic $S^*$ centered system $p$ is called a maximal single valued neutrosophic $S^*$ centered system or a single valued neutrosophic $S^*$ end if it cannot be included in any larger single valued neutrosophic $S^*$ centered system of single valued neutrosophic $S^*$ open sets.

**Definition 2.17: [3]**

A subset A of a single valued neutrosophic $S^*$ structure space $(S, S^*)$ is said to be an everywhere single valued neutrosophic $S^*$ dense subset in $(S, S^*)$ if $SVNS^*cl(A) = S$.

**Definition 2.18: [3]**

A subset of a single valued neutrosophic $S^*$ structure space $(S, S^*)$ is said to be a nowhere single valued neutrosophic $S^*$ dense subset in $(S, S^*)$ if $X \setminus A^c$ is everywhere single valued neutrosophic $S^*$ dense subset.

## 3. Single valued neutrosophic Cech extension

**Definition 3.1:**

A single valued neutrosophic $S^*$ centered system $p = \{U_\alpha\}$ of single valued neutrosophic $S^*$ open sets of $S$ is called a single valued neutrosophic $S^*$ completely regular system if for any $U_\alpha \in p$ there exists a $V_\alpha \in p$ and a single valued neutrosophic $S^*$ continuous function $f$ on $S$ such that $f(A) = 1$ for $A \in S \setminus U_\alpha$, $f(A) = 0$ for $A \in V_\alpha$ and $0 \leq f(A) \leq 1$ for any $A \in S$. In this case $V_\alpha$. is a single valued neutrosophic $S^*$ completely regularly contained in $U_\alpha$.

**Definition 3.2:**

A single valued neutrosophic $S^*$ completely regular system is called a single valued neutrosophic $S^*$ completely regular end if it is not contained in any larger single valued neutrosophic $S^*$ completely regular system.

**Definition 3.3:**

Let $(S_1, S_1^*)$ and $(S_2, S_2^*)$ be any two single valued neutrosophic $S^*$ structure spaces. A function $f:(S_1, S_1^*) \to (S_2, S_2^*)$ of a single valued neutrosophic $S^*$ structure space $(S_1, S_1^*)$ onto a single valued neutrosophic $S^*$ structure space $(S_2, S_2^*)$ is a quotient function (or natural function) if, whenever $V$ is a single valued neutrosophic $S_2^*$ open set in $(S_2, S_2^*)$, $f^{-1}(V)$ is a single valued neutrosophic $S_1^*$ open set in $(S_1, S_1^*)$ and conversely.

**Note 3.4:**

The maximal single valued neutrosophic $S^*$ centered systems of single valued neutrosophic $S^*$ open sets ( single valued neutrosophic $S^*$ ends) regarded as elements of the single valued neutrosophic $S^*$ space $\theta(S)$ , fall into two classes: those single valued neutrosophic $S^*$ ends each of which contains all the single valued neutrosophic $S^*$ open neighbourhoods of one (obviously only one) member of S, and the single valued neutrosophic $S^*$ ends not containing such systems of single valued neutrosophic $S^*$ open neighbourhoods. The single valued neutrosophic $S^*$ ends of the first type can be regarded as representing the members of the original single valued neutrosophic $S^*$ space S and those of the second type as corresponding to holes in S.

**Definition 3.5:**

The collection of all single valued neutrosophic $S^*$ ends of the first type in $\theta(S)$ is a single valued neutrosophic $S^*$ completely regular space and it is also called the single valued neutrosophic $S^*$ absolute of $S$ which is denoted by $w(S)$.

In $w(S)$, each member $V \in S$ is represented by single valued neutrosophic $S^*$ ends containing all single valued neutrosophic $S^*$ open neighbourhoods of $S$. It is obvious that $w(S) = \bigcup_{V \in S} B(V)$ where $B(V)$ are the single valued neutrosophic $S^*$ ends $p$ of $S$ that contain all the single valued neutrosophic $S^*$ open neighbourhoods of V .The subset $w(S)$ is mapped in a natural way onto $S$ .If $p \in w(S)$ , then by definition $\pi_S(p) = V$ , where V is the member whose single valued neutrosophic $S^*$ open neighbourhoods all belong to $p$ and $\pi_S$ is the natural function of *w(S)* onto $S$.

**Lemma 3.6:**

A single valued neutrosophic $S^*$ centered system $\{U_\alpha\}$ of all single valued neutrosophic $S^*$ open neighbourhoods of a member A in a single valued neutrosophic $S^*$ completely regular space $S$ is a single valued neutrosophic $S^*$ completely regular end.

**Proof:**

Here, $\{U_\alpha\}$ is a single valued neutrosophic $S^*$ completely regular centered system .The Lemma will be proved if it is possible to show that $\{U_\alpha\}$ is not contained in any other single valued neutrosophic $S^*$ completely regular system. As a contrary, suppose that $\{V_\alpha\}$ is a single valued neutrosophic $S^*$ centered completely regular system containing $\{U_\alpha\}$ with $V_{\alpha_1} \notin \{U_\alpha\}$. Since $V_{\alpha_1}$ meets every single valued neutrosophic $S^*$ open neighbourhood of $A$ , $A \in SVNS^*cl(V_{\alpha_1})\backslash V_{\alpha_1}$ .Let $V_{\alpha_2}$ be an element of $\{V_\alpha\}$ such that $SVNS^*cl(V_{\alpha_2})$ $\subseteq V_{\alpha_1}$.But then $A \in SVNS^*cl(V_{\alpha_2})$.It follows that $V_{\alpha_2}$ does not meet any of the single valued neutrosophic $S^*$ open neighbourhoods of $A$ , so $\{V_\alpha\}$ cannot be a single valued neutrosophic $S^*$ centered system containing $\{U_\alpha\}$.

Now we construct a single valued neutrosophic $S^*$ structure space which is denoted by $\alpha'(S)$ .Itsmembers are all single valued neutrosophic $S^*$ completely regular ends of $S$, and its single valued neutrosophic $S^*$ topology is defined as follows: Choose an arbitrary single valued neutrosophic $S^*$ open set U in S and the collection $O_U$ of all single valued neutrosophic $S^*$ centered completely regular ends of $S$ that contain $U$ as a member is to be a single valued neutrosophic $S^*$ open neighbourhood of each of them.

**Lemma 3.7:**

A single valued neutrosophic $S^*$ completely regular end $p = \{U_\alpha\}$ of a single valued neutrosophic $S^*$ structure space $S$ has the following properties:

1. If $U_\beta \supseteq U_\alpha \in p$, then $U_\beta \in p$.
2. The intersection of any finite number of members of $p$ belongs to $p$.

**Proof:**

Assertion (1) is obvious.

(2) Let $U_{\alpha 1}, U_{\alpha 2},...U_{\alpha n} \in p, U'_{\alpha 1}, U'_{\alpha 2},....U'_{\alpha n} \in p$ and $f_1, f_2,......f_n$ be functions such that $f_i(A)=0$ on $SVNS*cl(U'_{\alpha i}), f_i(A)=1$ on $S \setminus U_{\alpha i}$. Then the function $f_1(A)+f_2(A)+....+f_n(A)$ is zero on $SVNS*cl(U'_{\alpha 1}) \cap ..... \cap SVNS*cl(U'_{\alpha n})$ and a fortiori on $SVNS*cl((U'_{\alpha 1}) \cap ... \cap (U'_{\alpha n}))$.

Since $S \setminus (\cap U_{\alpha i}) = \cup (S \setminus U_{\alpha i})$, then $f_1(A)+.........+f_n(A) \geq 1$ at each member of $S \setminus (\cap U_{\alpha i})$. Putting $f(A)=1$ whenever $f_1(A)+.........+f_n(A) \geq 1$ and $f(A)=f_1(A)+.........+f_n(A)$. When this sum is less than 1, it may be obtained a function $f(A)$ such that $0 \leq f(A) \leq 1, f(A)=0$ on $\cap SVNS*cl(U'_{\alpha n})$ and $f(A)=1$ on $S \setminus (\cap U_{\alpha})$. Hence the single valued neutrosophic $S^*$ system $p$ must contain $\cap U_{\alpha i}$ and $\cap U'_{\alpha i}$, otherwise it would not be maximal single valued neutrosophic $S^*$ completely regular system.

**Corollary 3.8:**

$O_U \cap O_V = O_{U \cap V}$. For if $p \in O_U \cap O_V$, then $U \in p$ and $V \in p$. By Lemma 3.7, $U \cap V \in p$, that is, $p \in O_{U \cap V}$. Therefore, $O_U \cap O_V \subseteq O_{U \cap V}$. If $p \in O_{U \cap V}$, then $U \cap V \in p$ and by the same Lemma $U \in p$ and $V \in p$. Therefore, $p \in O_U$ and $p \in O_V$. That is, $p \in O_U \cap O_V$. Hence, $O_{U \cap V} \subseteq O_U \cap O_V$. Thus, $O_U \cap O_V = O_{U \cap V}$.

**Lemma 3.9:**

A single valued neutrosophic $S^*$ structure space $\alpha'(S)$ is a single valued neutrosophic $S^*$ Hausdorff extension of $S$.

**Proof:**

The single valued neutrosophic $S^*$ structure space $\alpha'(S)$ is a single valued neutrosophic $S^*$ Hausdorff space. Let $p$ and $q$ be any two disjoint members of $\alpha'(S)$. Then it is easy to find $U \in p$ and $V \in p$ such that $U \cap V = \varphi$, for otherwise the single valued neutrosophic $S^*$ centered system consisting of all the members of $p$ and all the members of $q$ would be a single valued neutrosophic $S^*$ centered completely regular system containing $p$ and $q$, which is impossible. $O_U$ and $O_V$, associated with this $U$ and $V$, are disjoint single valued neutrosophic $S^*$ open neighbourhoods of $p$ and $q$ in $\alpha'(S)$.

It shall be identified that the member $A \in S$ with the single valued neutrosophic S $^*$ end $p_A$ $= \{U_\alpha\}$ consisting of all the single valued neutrosophic S$^*$ open neighbourhood of $A$ in $S$. Then $O_U \cap S = U$, which shows that $S$ is single valued neutrosophic $S^*$ topologically embedded in $\alpha'(S)$, and since it is easy to see that $S$ is everywhere single valued neutrosophic S$^*$ dense in $\alpha'(S)$. Therefore, $\alpha'(S)$ is a single valued neutrosophic S$^*$extension of $S$. This proves the Lemma.

**Note 3.10:**

In a single valued neutrosophic S$^*$ completely regular space, the single valued neutrosophic S$^*$canonical neighbourhood forms a base.

**Lemma 3.11:**

A single valued neutrosophic S$^*$ structure space $\alpha'(S)$can be continuously mapped onto every single valued neutrosophic S$^*$ compact extension of $S$ in such a way that the members of $S$ remain fixed.

**Proof:**

Let $b(S)$ be any single valued neutrosophic S$^*$ compact extension of $S$. Each member A $\in$ b(S) determines the single valued neutrosophic S$^*$ centered system $p_A$ $= \{U_\alpha\}$ consisting of all single valued neutrosophic S$^*$ open neighbourhoods of $A$ in $b(S)$ .By Lemma 3.6 , this is a single valued neutrosophic S$^*$ completely regular system and a single valued neutrosophic S$^*$ maximal .The single valued neutrosophic S$^*$ centered system $q_A$ $= \{V_\alpha = U_\alpha \cap S\}$ is a single valued neutrosophic S$^*$ completely regular system .If $d = \{H_\alpha\} \in \alpha'(S)$ contains a single valued neutrosophic S$^*$ centered system $q_A$, then we define φ on $\alpha'(S)$ as φ(d) $= q_A$ .Since $b(S)$ is a single valued neutrosophic S$^*$ Hausdorff extension , $d$ can contain only one such single valued neutrosophic S$^*$ centered system $q_A$ .Hence the function φ is well-defined .Since an arbitrary single valued neutrosophic S$^*$ completely regular system can be extended to a single valued neutrosophic S$^*$ completely regular end , φ is onto. It is easy to see that if $A \in S$ , then $\varphi(A) = A$ .φ is defined on the whole of $\alpha'(S)$.For if $d = \{H_\alpha\} \in \alpha'(S)$, then $\bigcap_\alpha \left(SVNS^*cl(H_\alpha)\right)^{b(S)} \neq \phi$ (because b(S) is a single valued neutrosophic S$^*$compact space).Let $A \in \bigcap_\alpha \left(SVNS^*cl(H_\alpha)\right)^{b(S)}$.Then the single valued neutrosophic $S^*$centered system $d \cup q_A$ consisting of all members $H_\alpha \in d$ and all members $U_\alpha \in q_A$ is a single valued neutrosophic $S^*$completely regular and since $d$ is a single valued neutrosophic $S^*$ maximal completely regular system , $d \cup q_A = d$, that is , $q_A \subseteq d$ , so that $\varphi(d) = A$. Let $A \in b(S)$ and $U_\alpha$ be any single valued neutrosophic $S^*$ open neighbourhood of $A$ in $b(S)$. Assuming $U_\alpha$ is a canonical single valued neutrosophic $S^*$ open neighbourhood. Put $V_\alpha = U_\alpha \cap S$. Let $d \in \alpha'(S)$ and $\varphi(d)$=A. Then $O_{V_\alpha}$ is a single valued neutrosophic $S^*$ open neighbourhood of $d$ in $\alpha'(S)$.To show that $\varphi(O_{V_\alpha}) \subseteq SVNS^*cl(U_\alpha)$.For this, it is clear that $V_\alpha \in q'_A$ if and only if $A' \in U_\alpha$. Now , if $d' \in O_{V_\alpha}$ , then $V_\alpha \in d'$.If $\varphi(d') = A' \notin SVNS^*cl(U_\alpha)$ then some single valued neutrosophic $S^*$ open neighbourhood of $A'$ which does not contained in $U_\alpha$, but then $V_\alpha \notin q_{A'}$, so that $V_\alpha \notin d'$ , that is $d' \notin O_{V_\alpha}$, This contradicts our assumption .Since b(S) is single valued neutrosophic $S^*$ regular space ,φ is a single valued neutrosophic $S^*$ continuous and the Lemma is proved .The single valued neutrosophic set $O_U$ , where $U$ is a canonical single valued neutrosophic $S^*$ open set of $S$ forms a base in $\alpha'(S)$.

**Lemma 3.12:**

The single valued neutrosophic $S^*$ structure space $\alpha'(S)$ is a single valued neutrosophic $S^*$ completely regular space.

**Proof:**

Let $p = \{U_\alpha\} \in \alpha'(S)$ and let $U_2$ be any single valued neutrosophic $S^*$ completely regular contained in the canonical single valued neutrosophic $S^*$ open set $U_1$.Assume that $SVNS^*cl(O_{U_2}) \not\subset (\alpha'(S) \setminus O_{U_1})$.Then there is a member $q = \{SVNS^*cl(V_\alpha)\}$ such that $q \in SVNS^*cl(O_{U_2}) \cap (\alpha'(S) \setminus O_{U_1})$.The relation $q \in SVNS^*cl(O_{U_2})$ means that every $V_\alpha \in q$ meets U2 and the relation $q \in \alpha'(S) \setminus O_{U_1}$ , equivalent to $q \notin O_{U_1}$ means that every $V_\alpha$ meets $S \setminus U1$ .Since U1 is a canonical single valued neutrosophic $S^*$ open set , it follows that $V_\alpha$ meets $S \setminus SVNS^*cl(U1)$.

If V1 and V2 are single valued neutrosophic $S^*$ open sets such that V1 is single valued neutrosophic $S^*$ completely regularly contained in $S \setminus SVNS^*cl(V2)$ and $V = V1 \cap V2 \in q$, then either $V_1 \in q$ or $V_2 \in q$. Now , let *f(B)* be a function that is zero on SVNS*cl($U_2$) and 1 on $S \backslash U_1$.Such a function exists , since $U_2$ is single valued neutrosophic S* completely regularly contained in $U_1$ .Also let 0 < a < b < 1 and let *Γ(a, b)* be the single valued neutrosophic S* open set {B : a < f(B) < b}.By assumption , every $V_\alpha \in q$ has non- empty intersection with *Γ(a, b)*.For otherwise , $V_\alpha$ splits into two single valued neutrosophic S* open sets $V_{\alpha 1}$ and $V_{\alpha 2}$ such that $V_{\alpha 1}$ is single valued neutrosophic S* completely regularly contained in S\SVNS*cl($V_{\alpha 2}$) and $V_{\alpha 2} \cap U_2 = \varphi$ , $V_{\alpha 1} \cap$ (S\SVNS*cl($U_1$)) = φ. The last equation contradicts the fact that q ∈ SVNS*cl ($O_{U_2}$) ∩ (α'(S)\ $O_{U_1}$ ).

Consider the single valued neutrosophic S* open sets *Γ(a, b)* where $0 < a < a_0 < b_0 < b < 1$ and $a_0$ and $b_0$ are fixed .They form a single valued neutrosophic S* completely regular system which must be contained in *q* .But $\Gamma(a, b) \subset U_1$ that is , *Γ(a, b)*∩*Γ(S\SVN S*cl(U1))* = *φ* and hence , *q* ∉(α'(S)\ $O_{U_1}$ ). This contradiction shows that SVNS*cl ($O_{U_2}$ ) ⊆ $O_{U_1}$ , from which it follows that α'(S) is a single valued neutrosophic S* regular space . To prove that it is a single valued neutrosophic S* completely regular space. Let $\Gamma_t$, $0 \le t \le 1$, denote the set of all B ∈ S for which $f(B) < t$. It has shown that if $t_1 < t_2$, then $SVNS^*cl(O_{\Gamma_{t_1}}) \subseteq O_{\Gamma_{t_2}}$ .Hence it follows that $O_{U_2}$ is single valued neutrosophic S* completely regularly contained in $O_{U_1}$ .

**Lemma 3.13:**

The single valued neutrosophic S* structure space α'(S) is a single valued neutrosophic S* compact.

**Proof:**

If *H* is a single valued neutrosophic S* open set of α'(S), then there exists a single valued neutrosophic S* open set *U(H)* of S such that $H \subseteq O_{U(H)} \subseteq SVNS^*cl(H)$ then $U(H) = \bigcup_\alpha U_\alpha$.If *H* is single valued neutrosophic S* completely regularly embedded in *G* , then $O_{U(H)}$ is clearly single valued neutrosophic S* completely regularly embedded in $O_{U(G)}$.Suppose that α'(S) is not a single valued neutrosophic S* compact space. Then by Tychonoff's theorem, there exists a single valued neutrosophic $S^*$ completely regular space $\alpha'(S) \cup \xi$, containing $\alpha'(S)$ as an everywhere single valued neutrosophic $S^*$ dense set. Let $H_\alpha$ be the set of all single valued neutrosophic $S^*$ open sets of $\alpha'(S)$ for which $H_\alpha \cup \xi$ is a single valued neutrosophic $S^*$ open neighbourhood of $\xi$ in $\alpha'(S) \cup \xi$. Then, $\{H_\alpha\}$ is a single valued neutrosophic $S^*$ completely regular system in $\alpha'(S)$. Hence $O_{U(H_\alpha)}$ is also a single valued neutrosophic $S^*$ completely regular space. Thus, $\cap O_{U(H_\alpha)} = \varphi$. Since $\alpha'(S)$ is a single valued neutrosophic $S^*$ centered system $p = \{\cup (H_\alpha)\}$ of single valued neutrosophic $S^*$ completely regular too. But $p \in O_{U(H_\alpha)}$ for every $\alpha \in \Lambda$, that is $\bigcap_\alpha O_{U(H_\alpha)} = \varphi$. This contradiction proves the lemma.

**Proposition 3.14:**

For any single valued neutrosophic $S^*$ completely regular space *S*, the single valued neutrosophic $S^*$ structure space $\alpha'(S)$ coincides with the Cech extension $\beta(S)$ upto a single valued neutrosophic $S^*$ homeomorphism leaving the members of *S* fixed.

**Proof:**

The proof follows immediately from Lemma 3.11 and Lemma 3.13 and the uniqueness of a maximal single valued neutrosophic $S^*$ compact extension.

## 4. The Alexandrov - Urysohn compactness

In this section, the concept of single valued neutrosophic $S^*$ absolute is applied to establish the Alexandrov - Urysohn compactness.

**Property 4.1:**

If $F_1 \subset F_2 \subset F_3 \subset .... \subset F_n = S$ with $F_1$ non -empty, then $\bigcap_{i=1}^{n} \widetilde{F}_i \neq \phi$ (in particular, if $F$ is non-empty, so is $\widetilde{F}$ ).

**Proof:**

Let $B \in F_1$ and let $q' = \{G'\}$ be a single valued neutrosophic $S^*$ end of $F_1$ containing a single valued neutrosophic $S^*$ centered system of single valued neutrosophic $S^*$ open sets $G'$, in $F$, such that $B \in SVNS^*int(SVNS^*cl(G'))$. It may be assumed that it has been constructed systems $q^i = \{G^i\}$ of $F_i$ such that $q^i$ contains all the single valued neutrosophic $S^*$ open sets $G^i \subseteq F_i$ for which B $\in$ SVNS$^*int$(SVNS$^*$cl($G^i$)) and all single valued neutrosophic $S^*$ open sets whose intersection with $F_{i-1}$ is some $G^{i-1}$.

Now construct $q^{i+1}$. By definition, $q^{i+1}$, consists of all sets $G^{i+1} \subseteq F_{i+1}$ for which B $\in$ SVNS$^*int$(SVNS$^*cl$($G^{i+1}$)) and of all single valued neutrosophic $S^*$ open sets whose intersection with $F_i$ is some $G^i$.It is easy to show that $q^{i+1}$ is a single valued neutrosophic $S^*$ centered system. Thus for each $i$ construct a single valued neutrosophic $S^*$ centered system $q^i$. Let $p = \{H\}$ denote the single valued neutrosophic $S^*$ end of S containing $q^n$ .To show that $p \in \bigcap_{i=1}^{n} \widetilde{F}_i$ .It follows from the construction of $p$ that if H $\cap F_i \in q^i$ for some $i$ and some single valued neutrosophic $S^*$ open set $H$ in $S$ , then $H \in p$.To show that $p \in \widetilde{F}_i$.Let $H$ be a single valued neutrosophic $S^*$ open set of $S$ such that B $\in$ SVNS$^*$int(SVNS$^*$cl(H $\cap$ F$_i$)) .Then H $\cap$ F$_i$ $\in$ q$^i$ and hence H $\in$ p , that is , $p \in \widetilde{F}_i$.Hence the proof.

**Property 4.2:**

If $F$ is a single valued neutrosophic $S^*$H$-$ closed, then $\widetilde{F}$ is single valued neutrosophic $S^*$compact (and hence single valued neutrosophic $S^*$ closed in θ(S)).

**Proof:**

Let $\{H_\alpha\}$ be any single valued neutrosophic $S^*$ covering of $\widetilde{F}$ by single valued neutrosophic $S^*$ open sets in $\widetilde{F}$ .They may be extended to single valued neutrosophic $S^*$open in *w(S)*. It may assume that each of that each of the extended sets has the form $O_U$ , where $U$ is a single valued neutrosophic $S^*$ open set in *S*. Otherwise $\{H_\alpha\}$ may be replaced by a finer covering for which this condition holds .So it may be assumed that $\{H_\alpha\}$ is a single valued neutrosophic $S^*$ covering of $F$ by sets single valued neutrosophic $S^*$ open in *w(S)* of the form $O_{U\alpha}$ , where $U_\alpha$ is single valued neutrosophic $S^*$ open in *S*. Let B $\in$ F. Let $H_\beta^B$ denote the union of a finite number of single valued neutrosophic $S^*$ open sets $H_\alpha$ covering the single valued neutrosophic $S^*$ compact set $\pi_s^{-1}(B)$.It is clear that $H_\beta^B$ has the form $O_{U_\beta^B}$, where $U_\beta^B$ is a single valued neutrosophic $S^*$ open set in $S$ and is maximal among the single valued neutrosophic $S^*$ open sets $H$ for which $O_H = O_{U_\beta^B}$ .From the above, it follows that the single valued neutrosophic $S^*$centered system $\{SVNS^* \operatorname{int}(U_\beta^B \cap F)\}$is a single valued neutrosophic $S^*$ covering of F .Since F is single valued neutrosophic $S^*H-$ closed, choose a finite number of members of this single valued neutrosophic $S^*$ covering such that $\bigcup_{i=1}^{n} SVNS^* cl(SVNS^* \operatorname{int}(SVNS^* cl(U_{\beta i}^B \cap F))) = F$,where the closure is taken in $F$ in both cases. To show that $\bigcup_{i=1}^{n} O_{U_{\beta i}^B} \supseteq \widetilde{F}$ .Since the union $\bigcup_{i=1}^{n} O_{U_{\beta i}^B} = U$ has the property that $B \in SVNS^* \operatorname{int}(SVNS^* cl(F \cap U))$for any B,then an arbitrary single valued neutrosophic $S^*$ end $p \in \widetilde{F}$ contains U and hence belongs to some $O_{U_\beta^B}$. Thus, for only those $H\alpha$ that make up $O_{U_{\beta i}^B}$ and take their intersections with $\widetilde{F}$ , the required finite covering is obtained.

**Definition 4.3:**

A single valued neutrosophic $S^*$ Hausdorff space $S$ is a single valued neutrosophic $S^*$ compact space if and only if every (not necessarily countable) well-ordered decreasing sequence of non-empty single valued neutrosophic $S^*$ closed sets has a non-empty intersection.

**Theorem 4.4: (Alexandrov - Urysohn compactness)**

A single valued neutrosophic $S^*$ Hausdorff space S is a single valued neutrosophic $S^*$ compact space if and only if each of its single valued neutrosophic $S^*$ closed subset is single valued neutrosophic $S^*H-$ closed.

**Proof:**

**Necessity:**

The necessity of this condition follows from Property 4.2. Since in a single valued neutrosophic $S^*$ compact space every single valued neutrosophic $S^*$ closed subset is a single valued neutrosophic $S^*$ compact space and hence single valued neutrosophic $S^*H-$ closed.

**Sufficiency:**

Let $S$ be a single valued neutrosophic $S^*$ Hausdorff space, *w(S)* be its single valued neutrosophic $S^*$ absolute and $\pi_S$ be a single valued neutrosophic $S^*$ natural function of *w(S)* onto $S$. Also let $F$ be any single valued neutrosophic $S^*$ subset of S. It can be associated it with a certain single valued neutrosophic $S^*$ subset $\widetilde{F}$ of *w(S)*, defined by saying that the member $p \in \pi_s^{-1}(B), B \in S$ , belongs to $\widetilde{F}$ if $p \in O_U$ for every $U$ satisfying the condition $B \in SVNS^*\operatorname{int}(SVNS^*cl(U \cap F))$.By the construction of $\widetilde{F}$ it is contained in the complete single valued neutrosophic S* inverse image $\pi_s^{-1}(F)$ of $F$ in *w(S)*. $\widetilde{F}$ is called as the reduced inverse image of $F$ in *w(S)*. The proof of the Alexandrov - Urysohn compactness in single valued neutrosophic S* topology is based on the properties discussed above. For suppose that the conditions of the theorem are satisfied and that $\{F_\alpha\}$ is a well-ordered decreasing system of single valued neutrosophic S* closed sets of $S$. Then by Property 4.1, the set $\widetilde{F}_\alpha$ form a single valued neutrosophic S* centered system in *w(S)*. Also, since all the $F_\alpha^{'}$s are single valued neutrosophic S* compact space (Property 4.2), hence, $\bigcap_\alpha \widetilde{F}_\alpha \neq \phi$. Let $C \in \widetilde{F}$. Then $\pi_S(C) \in F_\alpha$ for every α, that is, $\bigcap_\alpha F_\alpha = \varphi$, as required.

**Property 4.5:**

Any well-ordered sequence of decreasing single valued neutrosophic S*H− closed sets in a single valued neutrosophic S* Hausdorff space has a non-empty intersection.

**Proof:**

From the proof of Property 4.1 it is easy to see that $\pi_S(\widetilde{F}) = F$. However, in general $\widetilde{F}$ does not coincide with $\pi_s^{-1}(F)$. Also, in the proof of Theorem 4.4, it cannot be taken $\pi_s^{-1}(F)$ instead of $\widetilde{F}$, since the complete inverse image of a single valued neutrosophic S* *H*- closed set need not be single valued neutrosophic S* compact . In fact , let $S$ be a single valued neutrosophic S* Hausdorff space and $F$ a single valued neutrosophic S*$H-$ closed subset such that there is a member $A \in S \backslash F$ for which there does not exist disjoint single valued neutrosophic S* open neighbourhoods of $A$ and $F$. Note that in a single valued neutrosophic S* Hausdorff space two disjoint single valued neutrosophic S* compact sets have disjoint single valued neutrosophic S* open neighbourhoods. If $\pi_s^{-1}(F)$ were single valued neutrosophic S* compact, then the single valued neutrosophic S* compact sets $\pi_s^{-1}(F)$ and $\pi_s^{-1}(A)$ would have disjoint single valued neutrosophic $S^*$ open neighbourhoods in *w(S)* , say $U$ and V. Then it follows from the proof of the theorem that SVN S*int(SVN S*cl(πS (U)) and SV N S*int(SV N S*cl(πS (V ))) would be disjoint single valued neutrosophic S* open neighbourhoods of $F$ and $A$ in $S$, which contradicts our assumption.

## References

1. Alexandrov, P. S and Urysohn, P. S. On compact topological spaces. Trudy Mat. Inst. Steklov 31, 1950.
2. Arockiarani, I., and Martina Jency, J. More on Fuzzy neutrosophic sets and fuzzy neutrosophic topological spaces. International journal of innovative research and studies 3 (5) (2014), 643-652.
3. Arockiarani, I., and Martina Jency, J. Hausdorff extensions in single valued neutrosophic S∗ centered systems. (Communicated)
4. Coker, D. and Es, A.H. On fuzzy compactness in intuitionistic fuzzy topological spaces J. Fuzzy math. 3(1996), 899-909.
5. Gleason, A.M. Projective topological spaces. Illinois J. Math. 2(1958), 482-489.
6. Wang, H., Smarandache, F., Zhang, Y., and Sunderraman, R. Single valued neutrosophic sets. Technical sciences and applied Mathematics. October 2012, 10-14.
7. Iliadis. S and Fomin. S. The method of centered systems in the theory of topological spaces, UMN, 21(1996), 47-66.
8. Ponomarev.V.I. Paracompacta, their projective spectra and continuous mappings, Mat. Sb. 60(1963), 89-119.
9. Smarandache, F. Neutrosophic set, a generalization of the intuitionistic fuzzy sets, Inter. J. Pure Appl. Math., 24,2005, 287-297 .
10. Stone. M.H. The theory of representations for Boolean algebra. Trans. Amer .Math.Soc.40(1936), 37-111.
11. Stone. M .H, Application of Boolean algebras to topology, Trans. Amer. Math. Soc. 41(1937), 375-481.
12. Uma M. K, Roja. E and Balasubramanian, G. The method of centered systems in fuzzy topological spaces, The journal of fuzzy mathematics 15(2007), 1- 7.

**EMAN.M.EL-NAKEEB[1], H. ELGHAWALBY[2], A.A.SALAMA[3], S.A.EL-HAFEEZ[4]**

1,2 Port Said University, Faculty of Engineering, Physics and Engineering Mathematics Department, Egypt.
E-mails: emanmarzouk1991@gmail.com, hewayda2011@eng.psu.edu.eg
3,4 Port Said University, Faculty of Science, Department of Mathematics and Computer Science, Egypt.
E-mails: drsalama44@gmail.com, samyabdelhafeez@yahoo.com

# Foundation for Neutrosophic Mathematical Morphology

## Abstract

The aim of this paper is to introduce a new approach to Mathematical Morphology based on neutrosophic set theory. Basic definitions for neutrosophic morphological operations are extracted and a study of its algebraic properties is presented. In our work we demonstrate that neutrosophic morphological operations inherit properties and restrictions of Fuzzy Mathematical Morphology.

## 1. Introduction

Established in 1964, Mathematical Morphology was firstly introduced by Georges Matheron and Jean Serra, as a branch of image processing [12]. As morphology is the study of shapes, Mathematical Morphology mostly deals with the mathematical theory of describing shapes using set theory. In image processing, the basic morphological operators dilation, erosion, opening and closing form the fundamentals of this theory [12]. A morphological operator transforms an image into another image, using some structuring element which can be chosen by the user. Mathematical Morphology stands somewhat apart from traditional linear image processing, since the basic operations of morphology are non-linear in nature, and thus make use of a totally different type of algebra than the linear algebra. At first, the theory was purely based on set theory and operators which defined for binary cases only. Later on the theory was extended to the grayscale images as the theory of lattices was introduced, hence, a representation theory for image processing was given [7]. As a scientific branch, Mathematical Morphology expanded worldwide during the 1990's. It is also during that period, different models based on fuzzy set theory were introduced [3, 4]. Today, Mathematical Morphology remains a challenging research field [6, 7].

In 1995, Samarandache initiated the theory of neutrosophic set as new mathematical tool for handling problems involving imprecise indeterminacy, and inconsistent data [14]. Later on, several researchers such as Bhowmik and Pal [2], and Salama [11], studied the concept of neutrosophic set. Neutrosophy introduces a new concept which represents indeterminacy with respect to some event, which can solve certain problems that cannot be solved by fuzzy logic.

This work is devoted for introducing the neutrosophic concepts to Mathematical Morphology. The rest of the paper is structured as follows: In §2, we introduce the fundamental definitions from the Mathematical Morphology whereas, the concepts of Fuzzy Morphology are introduced in §3. The basic definitions for Neutrosophic Morphological operations are extracted and a study of its algebraic properties is presented in §4.

## 2. Mathematical Morphology [5]

Basically, Mathematical Morphology describes an image's regions in the form of sets. Where the image is considered to be the universe with values are pixels in the image, hence, standard set notations can be used to describe image operations [7]. The essential idea, is to explore an image with a simple, pre-defined shape, drawing conclusions on how this shape fits or misses the shapes in the image [12]. This simple pre-defined shape is called the "structuring element", and it is usually small relative to the image.

In the case of digital images, a simple binary structuring elements like a cross or a square is used. The structuring elements can be placed at any pixel in the image, nevertheless, the rotation is not allowed. In this process, some reference pixel whose position defines where the structuring element is to be placed. The choice of this reference pixel is often arbitrary.

### 2.1. Binary Morphology

In binary morphology, an image is viewed as a subset of an Euclidean space $R^n$ or the integer grid $Z^n$, for some dimension *n*. The structuring element is a binary image (i.e., a subset of the space or the grid). In this section we briefly review the basic morphological operations, the dilation, the erosion, the opening and the closing.

#### 2.1.1. Binary Dilation: (Minkowski addition)

Dilation is one of the basic operations in Mathematical Morphology, which originally developed for binary images [15]. The dilation operation uses a structuring element for exploring and expanding the shapes contained in the input image. In binary morphology, dilation is a shift-invariant (translation invariant) operator, strongly related to the Minkowski addition.

For any Euclidean space E and a binary image A in E, the dilation of A by some structuring element B is defined by: $A \oplus B = \underset{b \in B}{\cup} A_b$ where $A_b$ is the translate of the set A along the vector $b$, i.e., $A_b$=$\{a + b \in E | a \in A , b \in B\}$.

The dilation is commutative, and may also be given by: $A \oplus B = B \oplus A = \underset{a \in A}{\cup} B_a$ .

An interpretation of the dilation of A by B can be understood as, if we put a copy of B at each pixel in A and union all of the copies, then we get $A \oplus B$.

The dilation can also be obtained by: $A \oplus B = \{b \in E \,|(-B) \cap A \neq \emptyset\}$, where (–B) denotes the reflection of B, that is, $-B = \{x \in E | -x \in B\}$.

Where the reflection satisfies the following property: $-(A \oplus B) = (-A) \oplus (-B)$

**2.1.2. Binary Erosion: (Minkowski subtraction)**

Strongly related to the Minkowski subtraction, the erosion of the binary image $A$ by the structuring element $B$ is defined by: $A \ominus B = \bigcap_{b \in B} A_{-b}$.

Unlike dilation, erosion is not commutative, much like how addition is commutative while subtraction is not [8, 15]. An interpretation for the erosion of $A$ by $B$ can be understood as, if we again put a copy of B at each pixel in A, this time we count only those copies whose translated structuring elements lie entirely in A; hence $A \ominus B$ is all pixels in A that these copies were translated to. The erosion of $A$ by $B$ is also may be given by the expression: $\mathrm{A} \ominus \mathrm{B} = \{\mathrm{p} \in \mathrm{E} \,|\mathrm{B_p} \subseteq \mathrm{A}\}$, where $B_p$ is the translation of $B$ by the vector $p$, i.e.,

$B_p = \{b + p \in E \,|b \in B\},\ \forall\, p \in E.$

**2.1.3. Binary Opening [15]**

The opening of $A$ by $B$ is obtained by the erosion of $A$ by $B$, followed by dilation of the resulting image by $B$: $A \circ B = (A \ominus B) \oplus B$.

The opening is also given by $A \circ B = \bigcup_{B_x \subseteq A} B_x$, which means that, an opening can be consider to be the union of all translated copies of the structuring element that can fit inside the object. Generally, openings can be used to remove small objects and connections between objects.

**2.1.4. Binary Closing [6]**

The closing of $A$ by $B$ is obtained by the dilation of $A$ by $B$, followed by erosion of the resulting structure by $B$:$\mathrm{A} \bullet \mathrm{B} = (A \oplus B) \ominus B$.

The closing can also be obtained by $\mathrm{A} \bullet \mathrm{B} = (\mathrm{A^c} \circ (-\mathrm{B}))^{\mathrm{c}}$, where $\mathrm{A^c}$ denotes the complement of $A$ relative to $E$ (that is, $\mathrm{A^c} = \{a \in E \,|a \notin A\}$). Whereas opening removes all pixels where the structuring element won't fit inside the image foreground, closing fills in all places where the structuring element will not fit in the image background, that is opening removes small objects, while closing removes small holes.

**2.2. Properties of the Basic Binary Operations**

Here are some properties of the basic binary morphological operations (dilation, erosion, opening and closing[8]). We define the power set of X, denoted by P(X), to be the set of all crisp subset of X.

For all A, B, C $\in P(X)$, the following properties hold:

- $A \oplus B = B \oplus A$,
- $A \subseteq B \Longrightarrow (A \oplus C) \subseteq (B \oplus C)$,
- $A \subseteq (A \oplus B)$,
- $(A \oplus B) \oplus C = A \oplus (B \oplus C)$, and $(A \ominus B) \ominus C = A \ominus (B \ominus C)$,
- Erosion and dilation satisfy the duality that is:

  $\mathrm{A} \oplus \mathrm{B} = (\mathrm{A^c} \ominus (-\mathrm{B}))^{\mathrm{c}}$, and $\mathrm{A} \ominus \mathrm{B} = (\mathrm{A^c} \oplus (-\mathrm{B}))^{\mathrm{c}}$,
- $A \subseteq B \Longrightarrow (A \circ C) \subseteq (B \circ C)$,

- $A \circ B \subseteq A$,
- Opening and closing satisfy the duality that is:

$$A \bullet B = (A^c \circ (-B))^c, \text{ and } A \circ B = \left(A^c \bullet (-B)\right)^c.$$

## 3. Fuzzy Mathematical Morphology

When operations are expressed in algebraic or logical terms, one powerful approach leading to good properties consists of formally replacing the classical symbols in the equations by their fuzzy equivalent. This framework led to an infinity of fuzzy Mathematical Morphologies, which are constructed in families with specific properties described in [3, 13].

### 3.1. Fuzzy Set

Since introduced by Zadeh [16], fuzzy sets have received a great deal of interest [17]. For an ordinary set, a given element either belongs or does not belong to the set, whereas for a fuzzy set the membership of an element is determined by the value of a given membership function, which assigns to each element a degree of membership ranging between zero and one.

#### 3.1.1. Definition [16]

Let $X$ be a fixed set. A fuzzy set $A$ of $X$ is an object having the form $A = \langle \mu_A \rangle$, where the function $\mu_A : X \to [0,1]$ defines the degree of membership of the element $x \in X$ to the set $A$. The set of all fuzzy subset of $X$ is denoted by $\mathcal{F}(X)$.

The fuzzy empty set in X is denoted by $0_f = \langle \underline{0} \rangle,\ where\ \underline{0} : X \longrightarrow [0,1]$ and $\underline{0}(x) = 0,\ \forall x \in X$. Moreover, the fuzzy universe set in X is denoted by $1_f = \langle \underline{1} \rangle,\ where\ \underline{1} : X \longrightarrow [0,1]\ \ and\ \ \underline{1}(x) = 1,\ \forall x \in X$.

### 3.2. Fuzzy Mathematical Operations [4]

The fuzziness concept was introduced to the morphology by defining the degree to which the structuring element fits into the image. The operations of dilation and erosion of a fuzzy image by a fuzzy structuring element having a bounded support, are defined in terms of their membership functions.

#### 3.2.1. Fuzzy Dilation [4]

Let us consider the notion of dilation within the original formulation of Mathematical Morphology in Euclidean space E. For any two n-dimensional gray-scale images, A and B, the fuzzy dilation, $A \oplus B = \langle \mu_{A \oplus B} \rangle$, of A by the structuring element B is an n-dimensional gray-scale image, that is: $\mu_{A \oplus B} : Z^2 \longrightarrow [0,1]$, and

$$\mu_{A \oplus B}(v) = \sup_{u \in Z^2} \min[\mu_A(v+u),\ \mu_B(u)]$$

Where $u, v \in Z^2$ are the spatial co-ordinates of pixels in the image and the structuring element; while $\mu_A$, $\mu_B$ are the membership functions of the image and the structuring element, respectively.

#### 3.2.2. Fuzzy Erosion [4]

For any two n-dimensional gray-scale image, A and B, the fuzzy erosion $A \ominus B = \langle \mu_{A \ominus B} \rangle$ of A by the structuring element B is an n-dimensional gray-scale image, that is:

$\mu_{A\ominus B}: Z^2 \longrightarrow [0,1]$, and

$$\mu_{A\ominus B}(v) = \inf_{u\in Z^2} max[\mu_A(v+u), 1-\mu_B(u)]$$

where $u$, $v \in Z^2$are the spatial co-ordinates of pixels in the image and the structuring element; while $\mu_A$, $\mu_B$ are the membership functions of the image and the structuring element respectively.

### 3.2.3. Fuzzy Closing and Fuzzy Opening [3]

In a similar way the two fuzzy operations for closing and opening for any two n-dimensional gray-scale images, A and B, are defined as follows:

$$\mu_{A\bullet B}(v) = \inf_{u\in Z^2} max\left(\sup_{w\in Z^2} min\ (\mu_A(v-u+w), \mu_B(u)), 1-\mu_B(u)\right)$$

$$\mu_{A\circ B}(v) = \sup_{u\in Z^2} min\left(\inf_{w\in Z^2} max\ (\mu_A(v-u+w), \mu_B(u)), 1-\mu_B(u)\right)$$

where $u$, $v$, w $\in Z^2$ are the spatial co-ordinates of pixels in the image and the structuring element; while $\mu_A, \mu_B$ are the membership functions of the image and the structuring element respectively.

### 3.3. Properties of the Basic Operations

Here are some properties of the basic fuzzy morphological operations (dilation, erosion, opening and closing [4]). We define the power set of X, denoted by $\mathcal{F}(Z^2)$, to be the set of all fuzzy subset of X,

For all $A, B, C \in \mathcal{F}(Z^2)$ the following properties hold:

i. Monotonicity (increasing in both argument)

$A \subseteq B \Longrightarrow A \oplus C \subseteq B \oplus C$

$A \subseteq B \Longrightarrow C \oplus A \subseteq C \oplus B$

ii. Monotonicity (increasing in the first and decreasing in the argument)

$A \subseteq B \Longrightarrow A \ominus C \subseteq B \ominus C$

$A \subseteq B \Longrightarrow C \ominus A \supseteq C \ominus B$

iii. Monotonicity (increasing in the first argument)

$A \subseteq B \Longrightarrow A \bullet C \subseteq B \bullet C$

iv. Monotonicity (increasing in the first argument)

$A \subseteq B \Longrightarrow A \circ C \subseteq B \circ C$

For any family $(A_i | i \in I)\ in \mathcal{F}(Z^2)\ and\ B \in \mathcal{F}(Z^2)$,

i. $\bigcap_{i\in I} A_i \oplus B \subseteq \bigcap_{i\in I} (A_i \oplus B)$ and $B \oplus \bigcap_{i\in I} A_i \subseteq \bigcap_{i\in I} (B \oplus A_i)$

ii. $\bigcap_{i\in I} A_i \ominus B \subseteq \bigcap_{i\in I} (A_i \ominus B)$ and $B \ominus \bigcap_{i\in I} A_i \supseteq \bigcap_{i\in I} (B \ominus A_i)$

iii. $\bigcap_{i\in I} A_i \bullet B \subseteq \bigcap_{i\in I} (A_i \bullet B)$

iv. $\bigcap_{i\in I} A_i \circ B \subseteq \bigcap_{i\in I} (A_i \circ B)$

For any family $(A_i | i \in I)\ in \mathcal{F}(Z^2)\ and\ B \in \mathcal{F}(Z^2)$,

i. $\bigcup_{i\in I} A_i \oplus B \supseteq \bigcup_{i\in I} (A_i \oplus B)$ and $B \oplus \bigcup_{i\in I} A_i \supseteq \bigcup_{i\in I} (B \oplus A_i)$

ii. $\bigcup_{i\in I} A_i \ominus B \supseteq \bigcap_{i\in I} (A_i \ominus B)$ and $B \ominus \bigcap_{i\in I} A_i \subseteq \bigcap_{i\in I} (B \ominus A_i)$

iii. $\bigcup_{i\in I} A_i \bullet B \supseteq \bigcap_{i\in I} (A_i \bullet B)$

iv. $\bigcup_{i\in I} A_i \circ B \supseteq \bigcap_{i\in I} (A_i \circ B)$.

## 4. Neutrosophic Approach to Mathematical Morphology

Smarandache [14] introduced the neutrosophic components ($T$, $I$, $F$) which represent the membership, indeterminacy, and non-membership values respectively, $T, I, F : X \to ]^-0,1^+[$ where $]^-0,1^+[$ is non-standard unit interval. Let $\varepsilon > 0$ be some infinitesimal number, hence, $1^+ = 1+\varepsilon$ and $^-0 = 0-\varepsilon$.

### 4.1. Neutrosophic Sets [1]

We denote the set of all neutrosophic subset of X by $\mathcal{N}(X)$. In [1, 14], the authors gave different definition for the concept of the neutrosophic sets. For more convenience we are choosing the following definitions to follow up our work for neutrosophic morphology. In the following definitions, we consider a space E and two neutrosophic subsets of X; $A, B \in \mathcal{N}(X)$.

#### 4.1.1. Definition [11, 14]

A neutrosophic set $A$ on the universe of discourse X is defined as:

$A = \langle T_A, I_A, F_A \rangle$, where $T_A, I_A, F_A : X \to [0,1]$.

#### 4.1.2. Definition [11]

The complement of a neutrosophic set $A$ is denoted by $A^c$ and is defined as:

$A^c = \langle T_{A^c}, I_{A^c}, F_{A^c} \rangle$, where $T_{A^c}, I_{A^c}, F_{A^c} : X \to [0,1]$ and for all $x$ in $X$.

$T_{A^c}(x) = 1 - T_A(x)$, $I_{A^c}(x) = 1 - I_A(x)$ and $F_{A^c}(x) = 1 - F_A(x)$

The neutrosophic empty Set of X is the triple, $0_{\mathcal{N}} = \langle \underline{0}, \underline{0}, \underline{1} \rangle$, where

$$\underline{1}(x) = 1 \text{ and } \underline{0}(x) = 0, \quad \forall x \in X.$$

The neutrosophic universe set of X is the triple, $1_{\mathcal{N}} = \langle \underline{1}, \underline{1}, \underline{0} \rangle$, where

$$\underline{1}(x) = 1 \text{ and } \underline{0}(x) = 0 \quad \forall x \in X.$$

### 4.2. Neutrosophic Mathematical Morphology

In this section we introduce the concept of neutrosophic morphology based on the fact that the basic morphological operators make use of fuzzy set operators, or equivalently, crisp logical operators. Hence, such expressions can easily be extended using the context of neutrosophic sets.

#### 4.2.1. Definition

The reflection of the structuring element B mirrored in its origin is defined as:

- $-B = \langle -T_B, -I_B, -F_B \rangle$, where

  $-T_B(u) = T_B(-u)$, $\quad -I_B(u) = I_B(-u)$ and $-F_B(u) = F_B(-u)$

- For every $p$ in E, Translation of A by $p \in Z^2$ is $A_p = \langle T_{A_p}, I_{A_p}, F_{A_p} \rangle$, Where $T_{A_p}(u) = T_{A_p}(u+p)$, $I_{A_p}(u) = I_{A_p}(u+p)$ and $F_{A_p}(u) = F_{A_p}(u+p)$

Most morphological operations on neutrosophic can be obtained by combining neutrosophic set theoretical operations with two basic operations, dilation and erosion.

### 4.3 Neutrosophic Morphological Operations

The neutrosophy concept is introduced to morphology by a triple degree to which the structuring element fits into the image in the three levels of trueness, indeterminacy, and falseness. The operations of neutrosophic erosion, dilation, opening and closing of the neutrosophic image by neutrosophic structuring element, are defined in terms of their membership, indeterminacy and non-membership functions; which is defined for the first time as far as we know.

#### 4.3.1. The Operation of Dilation

The process of the structuring element B on the image A and moving it across the image in a way like convolution is defined as dilation operation. The two main inputs for the dilation operator [7] are the image which is to be dilated and a set of coordinate points known as a structuring element which may be considered as a kernel. The exact effect of the dilation on the input image is determined by this structuring element [6].

**4.3.1.1. Definition:** (Neutrosophic Dilation)

let A and B are two neutrosophic sets; then the neutrosophic dilation is given as $(\mathrm{A} \widetilde{\oplus} \mathrm{B}) = \langle \mathrm{T}_{\mathrm{A}\widetilde{\oplus}\mathrm{B}}, \mathrm{I}_{\mathrm{A}\widetilde{\oplus}\mathrm{B}}, \mathrm{F}_{\mathrm{A}\widetilde{\oplus}\mathrm{B}} \rangle$, where for each u and v $\in Z^2$.

$$T_{A\widetilde{\oplus}B}(v) = \sup_{u \in Z^n} \min(T_A(v+u), T_B(u))$$

$$I_{A\widetilde{\oplus}B}(v) = \sup_{u \in Z^n} \min(I_A(v+u), I_B(u))$$

$$F_{A\widetilde{\oplus}B}(v) = \inf_{u \in Z^n} \max(1 - F_A(v+u), 1 - F_B(u))$$

#### 4.3.2. The Operation of Erosion

The erosion process is as same as dilation, but the pixels are converted to 'white', not 'black'. The two main inputs for the erosion operator [12], are the image which is to be eroded and a structuring element. The exact effect of the erosion on the input image is determined by this structuring element. The following steps are the mathematical definition of erosion for gray-scale images.

**4.3.2.1. Definition:** (Neutrosophic Erosion)

let A and B are two neutrosophic sets , then the neutrosophic erosion is given

$(\mathrm{A} \widetilde{\ominus} \mathrm{B}) = \langle \mathrm{T}_{\mathrm{A}\widetilde{\ominus}\mathrm{B}}, \mathrm{I}_{\mathrm{A}\widetilde{\ominus}\mathrm{B}}, F_{A\widetilde{\ominus}B} \rangle$ ; where for each u and v $\in Z^2$

$$T_{A\widetilde{\ominus}B}(v) = \inf_{u \in Z^n} \max(T_A(v+u), 1 - T_B(u))$$

$$I_{A\widetilde{\ominus}B}(v) = \inf_{u \in Z^n} \max(I_A(v+u), 1 - I_B(u))$$

$$F_{A\widetilde{\ominus}B}(v) = \sup_{u \in Z^n} \min\big(1 - F_A(v+u), F_B(u)\big)$$

#### 4.3.3. The Operation of Opening and Closing

The combination of the two main operations, dilation and erosion, can produce more complex sequences. Opening and closing are the most useful of these for morphological filtering [8]. An opening operation is defined as erosion followed by a dilation using the same structuring element for both operations. The basic two inputs for opening operator are an image to be opened, and a structuring element. Gray-level opening consists simply of gray-level erosion followed by gray-level dilation. The morphological opening $\circ$ and closing $\bullet$ are defined by:

$$A \tilde{\circ} B = (A \widetilde{\ominus} B) \widetilde{\oplus} B$$

$$A \tilde{\bullet} B = (A \widetilde{\oplus} B) \widetilde{\ominus} B$$

From a granularity perspective, opening and closing provide coarser descriptions of the set A. The opening describes A as closely as possible using not the individual pixels but by fitting (possibly overlapping) copies of E within A. The closing describes the complement of A by fitting copies of $E^*$ outside A. The actual set is always contained within these two extremes: $A \tilde{\circ} B \subseteq A \subseteq A \tilde{\bullet} B$ and the informal notion of fitting copies of E, or of $E^*$, within a set is made precise in these equations:

The operator $\mathcal{N}(E) \rightarrow \mathcal{N}(E) : A \rightarrow A \tilde{\circ} B$ is called the opening by B; it is the composition of the erosion $\ominus$, followed by the dilation $\oplus$. On the other hand, the operator $\mathcal{N}(E) \rightarrow \mathcal{N}(E) : A \rightarrow A \tilde{\bullet} B$ is called the closing.

To understand what *e.g.,* a closing operation does: imagine the closing applied to a set; the dilation will expand object boundaries, which will be partly undone by the following erosion. Small, (*i.e.*, smaller than the structuring element) holes and thin tubelike structures in the interior or at the boundaries of objects will be filled up by the dilation, and not reconstructed by the erosion, inasmuch as these structures no longer have a boundary for the erosion to act upon. In this sense the term 'closing' is a well-chosen one, as the operation removes holes and thin cavities. In the same sense the opening opens up holes that are near (with respect to the size of the structuring element) a boundary, and removes small object protuberances.

##### 4.3.3.1. Neutrosophic Opening

let A and B are two neutrosophic sets it's defined as the flowing:

$(A \tilde{\circ} B) = \langle T_{A\tilde{\circ}B}, I_{A\tilde{\circ}B}, F_{A\tilde{\circ}B} \rangle$, where u, v, w$\in Z^2$

$$T_{A\tilde{\circ}B}(v) = \sup_{u \in Z^n} \min \left[ \inf_{z \in R^n} \max\big(T_A(v-u+w), 1 - T_B(w)\big), T_B(u) \right]$$

$$I_{A\tilde{\circ}B}(v) = \sup_{u \in Z^n} \min \left[ \inf_{z \in R^n} \max\big(I_A(v-u+w), 1 - I_B(w)\big), I_B(u) \right]$$

$$F_{A\tilde{\circ}B}(v) = \inf_{u \in Z^n} \max \left[ \sup_{z \in R^n} \min\big(1 - F_A(v-u+w), F_B(w)\big), 1 - F_B(u) \right]$$

##### 4.3.3.2. Neutrosophic Closing

let A and B be two neutrosophic sets it's defined as the flowing:

$(A \tilde{\bullet} B) = \langle T_{A\tilde{\bullet}B}, I_{A\tilde{\bullet}B}, F_{A\tilde{\bullet}B} \rangle$, where u, v, w$\in Z^2$

$$T_{A\tilde{\bullet}B}(v) = \inf_{u\in Z^2} \max\left[\sup_{w\in Z^2} \min\big(T_A(v-u+w), T_B(w)\big), 1-T_B(u)\right]$$

$$I_{A\tilde{\bullet}B}(v) = \inf_{u\in Z^2} \max\left[\sup_{w\in Z^2} \min\big(I_A(v-u+w), I_B(w)\big), 1-I_B(u)\right]$$

$$F_{A\tilde{\bullet}B}(v) = \sup_{u\in Z^2} \min\left[\inf_{w\in Z^2} \max\big(1-F_A(v-u+w), 1-F_B(w)\big), F_B(u)\right]$$

**4.4. Algebraic Properties in Neutrosophic**

The algebraic properties for Neutrosophic Mathematical Morphology erosion and dilation, as well as for neutrosophic opening and closing operations are now considered.

**4.4.1. Proposition Duality Theorem of Dilation**

let A and B be two neutrosophic sets. neutrosophic erosion and dilation are dual operations i.e. $(A^c \tilde{\oplus} B)^c = \langle T_{(A^c \tilde{\oplus} B)^c}, I_{(A^c \tilde{\oplus} B)^c}, F_{(A^c \tilde{\oplus} B)^c}\rangle$; where for each u, v $\in Z^2$

$$T_{(A^c\tilde{\oplus}B)^c}(v) = 1 - T_{(A^c\tilde{\oplus}B)}(v)$$

$$= 1 - \sup_{u\in Z^2} \min(T_{A^c}(v+u), T_B(u)) = \inf_{u\in Z^2}[1-\min(T_{A^c}(v+u), T_B(u))]$$

$$= \inf_{u\in Z^2}[\max(1-T_{A^c}(v+u), 1- T_B(u))]$$

$$= \inf_{u\in Z^2}[\max(T_A(v+u), 1- T_B(u))] \quad = T_{A\tilde{\ominus}B}(v)$$

$$I_{(A^c\tilde{\oplus}B)^c}(v) = 1 - I_{(A^c\tilde{\oplus}B)}(v) \quad = 1 - \sup_{x\in R^n} \min(A^c(v+u), I_B(u))$$

$$= \inf_{u\in Z^2}[1-\min(I_{A^c}(v+u), I_B(x))]$$

$$= \inf_{u\in Z^2}[\max(1-I_{A^c}(v+u), 1- I_B(u))]$$

$$= \inf_{u\in Z^2}[\max(I_A(v+u), 1- I_B(u))] \quad = I_{A\tilde{\ominus}B}(v)$$

$$F_{(A^c\tilde{\oplus}B)^c}(v) = 1 - F_{(A^c\tilde{\oplus}B)}(v)$$

$$= 1 - \inf_{x\in R^n} \max(1-F_{A^c}(v+u), 1-F_B(u))$$

$$= \sup_{u\in Z^2}[1-\max(1-F_{A^c}(v+u), 1-F_B(u))]$$

$$= \sup_{u\in Z^2}[\min(1-F_A(v+u), F_B(u))] \quad = F_{A\tilde{\ominus}B}(v)$$

$$\langle T_{(A^c \tilde{\oplus} B)^c}, I_{(A^c \tilde{\oplus} B)^c}, F_{(A^c \tilde{\oplus} B)^c}\rangle = \langle T_{A\tilde{\ominus}B}, I_{A\tilde{\ominus}B}, F_{A\tilde{\ominus}B}\rangle.$$

**4.4.2. Proposition the Duality Theorem Closing**

let A and B be two neutrosophic sets, neutrosophic opening and neutrosophic closing are also dual operation i.e.

$(A^c \tilde{\bullet} B)^c = \langle T_{(A^c \tilde{\bullet} B)^c}, I_{(A^c \tilde{\bullet} B)^c}, F_{(A^c \tilde{\bullet} B)^c}\rangle$, where for all $x \in X$

$T_{(A^c \tilde{\bullet} B)^c}(v) = 1 - T_{A^c \tilde{\bullet} B}(v)$

$$T_{(A^c \tilde{\bullet} B)^c}(v) = 1 - \inf_{u \in Z^2} max \left[\sup_{z \in R^n} min\big(T_{A^c}(v-u+w), T_{B(w)}\big), 1 - T_B(u)\right]$$

$$= \sup_{u \in Z^2} min \left[1 - \sup_{z \in R^n} min\big(T_{A^c}(v-u+w), T_{B(w)}\big), T_B(u)\right]$$

$$= \sup_{u \in Z^2} min \left[\inf_{z \in R^n} max\big(1 - T_{A^c}(v-u+w), 1 - T_{B(w)}\big), T_B(u)\right]$$

$$= \sup_{u \in Z^2} min \left[\inf_{z \in R^n} max\big(T_{A^c}(v-u+w), 1 - T_{B(w)}\big), T_B(u)\right] \quad = T_{A \tilde{\circ} B}(v)$$

$$I_{(A^c \tilde{\bullet} B)^c}(v) = 1 - I_{A^c \tilde{\bullet} B}(v)$$

$$I_{(A^c \tilde{\bullet} B)^c}(v) = 1 - \inf_{u \in Z^2} max \left[\sup_{z \in R^n} min\big(I_{A^c}(v-u+w), I_{B(w)}\big), 1 - I_B(u)\right]$$

$$= \sup_{u \in Z^2} min \left[1 - \sup_{z \in R^n} min\big(I_{A^c}(v-u+w), I_{B(w)}\big), I_B(u)\right]$$

$$= \sup_{u \in Z^2} min \left[\inf_{z \in R^n} max\big(1 - I_{A^c}(v-u+w), 1 - I_{B(w)}\big), I_B(u)\right]$$

$$= \sup_{u \in Z^2} min \left[\inf_{z \in R^n} max\big(I_A(v-u+w), 1 - I_{B(w)}\big), I_B(u)\right] \quad = I_{A \tilde{\circ} B}(v)$$

$$F_{(A^c \tilde{\bullet} B)^c}(v) = 1 - F_{A^c \tilde{\bullet} B}$$

$$F_{(A^c \tilde{\bullet} B)^c}(v) = 1 - \sup_{u \in Z^2} min \left[\inf_{z \in R^n} max\big(1 - F_A(v-u+w), 1 - F_{B(w)}\big), F_B(u)\right]$$

$$F_{(A^c \tilde{\bullet} B)^c}(v) = \inf_{u \in Z^2} max \left[1 - \inf_{z \in R^n} max\big(1 - F_A(v-u+w), 1 - F_{B(w)}\big), F_B(u)\right]$$

$$F_{(A^c \tilde{\bullet} B)^c}(v) = \inf_{u \in Z^2} max \left[\sup_{z \in R^n} min\big(1 - F_A(v-u+w), F_{B(w)}\big), 1 - F_B(u)\right]$$

$$= F_{A \tilde{\circ} B}(v)$$

$$\left\langle T_{(A^c \tilde{\bullet} B)^c}, I_{(A^c \tilde{\bullet} B)^c}, F_{(A^c \tilde{\bullet} B)^c} \right\rangle = \langle T_{A \tilde{\circ} B}, I_{A \tilde{\circ} B}, F_{A \tilde{\circ} B} \rangle.$$

.

**Lemma 1:** for any $A \in \mathcal{N}(X)$, and the neutrosophic universal set $1_{\mathcal{N}}$, we have that $A \,\widetilde{\oplus}\, 1_{\mathcal{N}} \subseteq A$, $A \,\widetilde{\oplus}\, 1_{\mathcal{N}} = \left\langle T_{A \widetilde{\oplus} 1_{\mathcal{N}}}, I_{A \widetilde{\oplus} 1_{\mathcal{N}}}, F_{A \widetilde{\oplus} 1_{\mathcal{N}}} \right\rangle$

**Proof:**

$$T_{A \widetilde{\oplus} 1_{\mathcal{N}}}(v) = \sup_{u \in Z^2} min(T_A(v+u), 1) \qquad = \sup_{u \in Z^2}\big(T_A(y+x)\big) = T_A(v)$$

$$I_{A \widetilde{\oplus} 1_{\mathcal{N}}}(v) = \sup_{u \in Z^2} min(I_A(v+u), 1) \qquad = \sup_{u \in Z^2}\big(I_A(y+x)\big) = I_A(v)$$

$$F_{A \widetilde{\oplus} 1_{\mathcal{N}}}(v) = \inf_{u \in Z^2} max(1 - F_A(v+u), 1 - 0) \qquad = \underline{1}(v)$$

$$\langle T_A, I_A, \underline{1} \rangle \subseteq \langle T_A, I_A, F_A \rangle = A$$

***Lemma 2:*** for any $A \in \mathcal{N}(X)$, and the neutrosophic empty set $0_{\mathcal{N}}$, we have that

$$\mathbf{A} \widetilde{\oplus} 0_{\mathcal{N}} \subseteq A^c, \quad \mathrm{A} \widetilde{\oplus} 0_{\mathcal{N}} = \left\langle T_{\mathrm{A}\widetilde{\oplus} 0_{\mathcal{N}}}, I_{\mathrm{A}\widetilde{\oplus} 0_{\mathcal{N}}}, F_{A\widetilde{\oplus} 0_{\mathcal{N}}} \right\rangle$$

**Proof:**

$$T_{A\widetilde{\oplus}0_{\mathcal{N}}}(v) = \sup_{u\in Z^2} \min(T_A(v+u), 0) \qquad = \underline{0}(v)$$

$$I_{A\widetilde{\oplus}0_{\mathcal{N}}}(v) = \sup_{u\in Z^2} \min(I_A(v+u), 0) \qquad = \underline{0}(v)$$

$$F_{A\widetilde{\oplus}0_{\mathcal{N}}}(v) = \inf_{u\in Z^2} \max(1 - F_A(v+u), 1-1) \quad = \inf_{u\in Z^2}\left(1 - F_A(v+u)\right) = F_{A^c}(v)$$

$$\left\langle \underline{0}, \underline{0}, \mathrm{F}_{\mathrm{A}^c} \right\rangle \subseteq \langle \mathrm{T}_{\mathrm{A}^c}, \mathrm{I}_{\mathrm{A}^c}, \mathrm{F}_{\mathrm{A}^c} \rangle = \mathrm{A}^c$$

### 4.5. Properties of the Neutrosophic Morphological Operations

In this section, we investigate the basic properties of the neutrosophic morphological operation (dilation, erosion, opening and closing), which we defined in §4.

#### 4.5.1. Properties of the Neutrosophic Dilation

**Proposition 1**

The neutrosophic dilation satisfies the following properties: $\forall\ \mathrm{A}, \mathrm{B} \in \mathcal{N}(\mathrm{Z}^2)$

*i.* ***Commutativity:*** $\mathrm{A}\widetilde{\oplus}\mathrm{B} = \mathrm{B} \widetilde{\oplus} \mathrm{A}$

*ii.* ***Associativity:*** $\left(\mathrm{A} \widetilde{\oplus} \mathrm{B}\right) \widetilde{\oplus} \mathrm{C} = \mathrm{A} \widetilde{\oplus} \left(\mathrm{B} \widetilde{\oplus} \mathrm{C}\right)$.

iii. ***Monotonicity:*** (increasing in both arguments):

a) $\mathrm{A} \subseteq \mathrm{B} \Longrightarrow \langle \mathrm{T}_{\mathrm{A}\widetilde{\oplus}\mathrm{C}}, I_{\mathrm{A}\widetilde{\oplus}\mathrm{C}}, F_{\mathrm{A}\widetilde{\oplus}\mathrm{C}} \rangle \subseteq \langle \mathrm{T}_{\mathrm{B}\widetilde{\oplus}\mathrm{C}}, I_{\mathrm{B}\widetilde{\oplus}\mathrm{C}}, F_{\mathrm{B}\widetilde{\oplus}\mathrm{C}} \rangle$

$$\mathrm{T}_{\mathrm{A}\widetilde{\oplus}\mathrm{C}} \subseteq \mathrm{T}_{\mathrm{B}\widetilde{\oplus}\mathrm{C}}, \quad \mathrm{I}_{\mathrm{A}\widetilde{\oplus}\mathrm{C}} \subseteq \mathrm{I}_{\mathrm{B}\widetilde{\oplus}\mathrm{C}} \ \ and\ \ \mathrm{F}_{\mathrm{A}\widetilde{\oplus}\mathrm{C}} \supseteq \mathrm{F}_{\mathrm{B}\widetilde{\oplus}\mathrm{C}}$$

b) $\mathrm{A} \subseteq \mathrm{B} \Longrightarrow \langle \mathrm{T}_{\mathrm{C}\widetilde{\oplus}\mathrm{A}}, I_{\mathrm{C}\widetilde{\oplus}\mathrm{A}}, F_{\mathrm{C}\widetilde{\oplus}\mathrm{A}} \rangle \subseteq \langle \mathrm{T}_{\mathrm{C}\widetilde{\oplus}\mathrm{B}}, I_{\mathrm{C}\widetilde{\oplus}\mathrm{B}}, F_{\mathrm{C}\widetilde{\oplus}\mathrm{B}} \rangle$

$$\mathrm{T}_{\mathrm{C}\widetilde{\oplus}\mathrm{A}} \subseteq \mathrm{T}_{\mathrm{C}\widetilde{\oplus}\mathrm{B}}, \quad \mathrm{I}_{\mathrm{C}\widetilde{\oplus}\mathrm{A}} \subseteq \mathrm{I}_{\mathrm{C}\widetilde{\oplus}\mathrm{B}} \ \ and\ \ \mathrm{F}_{\mathrm{C}\widetilde{\oplus}\mathrm{A}} \supseteq \mathrm{F}_{\mathrm{C}\widetilde{\oplus}\mathrm{B}}$$

**Proof*:***

*i), ii), iii)* Obvious.

**Proposition2:** for any family $(A_i | i \in I)\ in\ \mathcal{N}(\mathrm{Z}^2)\ and\ B \in \mathcal{N}(\mathrm{Z}^2)$

a) $\langle \mathrm{T}_{\bigcap_{i\in I} \mathrm{A_i}\widetilde{\oplus}\mathrm{B}}, \mathrm{I}_{\bigcap_{i\in I} \mathrm{A_i}\widetilde{\oplus}\mathrm{B}}, \mathrm{F}_{\bigcap_{i\in I} \mathrm{A_i}\widetilde{\oplus}\mathrm{B}} \rangle \subseteq \langle \mathrm{T}_{\bigcap_{i\in I}(A_i\widetilde{\oplus}\mathrm{B})}, \mathrm{I}_{\bigcap_{i\in I}(A_i\widetilde{\oplus}\mathrm{B})}, \mathrm{F}_{\bigcap_{i\in I}(A_i\widetilde{\oplus}\mathrm{B})} \rangle$

$$T_{\bigcap_{i\in I} A_i\widetilde{\oplus}\mathrm{B}} \subseteq T_{\bigcap_{i\in I}(A_i\widetilde{\oplus}\mathrm{B})}, \ I_{\bigcap_{i\in I} A_i\widetilde{\oplus}\mathrm{B}} \subseteq I_{\bigcap_{i\in I}(A_i\widetilde{\oplus}\mathrm{B})} \ \text{ and } F_{\bigcap_{i\in I} A_i\widetilde{\oplus}\mathrm{B}} \supseteq F_{\bigcup_{i\in I}(A_i\widetilde{\oplus}\mathrm{B})}$$

b) $\langle \mathrm{T}_{\mathrm{B}\widetilde{\oplus}\bigcap_{i\in I} A_i}, \mathrm{I}_{\mathrm{B}\widetilde{\oplus}\bigcap_{i\in I} A_i}, \mathrm{F}_{\mathrm{B}\widetilde{\oplus}\bigcap_{i\in I} A_i} \rangle \subseteq \langle \mathrm{T}_{\bigcap_{i\in I}(\mathrm{B}\widetilde{\oplus}A_i)}, \mathrm{I}_{\mathrm{T}_{\bigcap_{i\in I}(\mathrm{B}\widetilde{\oplus}A_i)}}, \mathrm{F}_{\mathrm{T}_{\bigcap_{i\in I}(\mathrm{B}\widetilde{\oplus}A_i)}} \rangle$

$$T_{B\widetilde{\oplus}\bigcap_{i\in I} A_i} \subseteq T_{\bigcap_{i\in I}(\mathrm{B}\widetilde{\oplus}A_i)}, \ I_{B\widetilde{\oplus}\bigcap_{i\in I} A_i} \subseteq I_{\bigcap_{i\in I}(\mathrm{B}\widetilde{\oplus}A_i)} \ \text{ and } \ F_{B\widetilde{\oplus}\bigcap_{i\in I} A_i} \supseteq F_{\bigcup_{i\in I}(\mathrm{B}\widetilde{\oplus}A_i)}$$

**Proof:** a)

$$\langle T_{\bigcap_{i\in I} A_i\widetilde{\oplus}B}, I_{\bigcap_{i\in I} A_i\widetilde{\oplus}B}, F_{\bigcap_{i\in I} A_i\widetilde{\oplus}B} \rangle \subseteq \langle T_{\bigcap_{i\in I}(A_i\widetilde{\oplus}B)}, I_{\bigcap_{i\in I}(A_i\widetilde{\oplus}B)}, F_{\bigcap_{i\in I}(A_i\widetilde{\oplus}B)} \rangle$$

$T_{\underset{i\in I}{\cap} A_i \widetilde{\oplus} B}(v) = \underset{u\in Z^n}{sup}\, min\left(T_{\underset{i\in I}{\cap} A_i}(v+u), T_B(u)\right) \quad = \underset{u\in Z^n}{sup}\, min\left(\underset{i\in I}{inf}\; T_{A_i}(v+u), T_B(u)\right)$

$= \underset{u\in Z^n}{sup}\; \underset{i\in I}{inf}\left(minT_{\boldsymbol{A_i}}(v+u), T_B(u)\right) \quad \le \underset{\boldsymbol{i\in I}}{inf}\; \underset{u\in Z^n}{sup}\left(minT_{\boldsymbol{A_i}}(v+ u), T_B(u)\right)$

$\le \underset{i\in I}{\cap} T_{(A_i \widetilde{\oplus} B)}(v) \quad \le T_{\underset{i\in I}{\cap}(A_i \widetilde{\oplus} B)}(v)$

$I_{\underset{i\in I}{\cap} A_i \widetilde{\oplus} B}(v) = \underset{u\in Z^n}{sup}\, min\left(I_{\underset{i\in I}{\cap} \boldsymbol{A_i}}(v+u), I_B(u)\right)$

$= \underset{u\in Z^n}{sup}\, min\left(\underset{i\in I}{inf}\; I_{\boldsymbol{A_i}}(v+u), I_B(u)\right) \quad = \underset{u\in Z^n}{sup}\; \underset{i\in I}{inf}\left(min\;\; I_{\boldsymbol{A_i}}(v+ u), I_B(u)\right)$

$\le \underset{\boldsymbol{i\in I}}{\cap}\; \underset{u\in Z^n}{sup}\left(min\;\; I_{\boldsymbol{A_i}}(v+u), I_B(u)\right) \quad \le \underset{i\in I}{\cap} I_{(A_i \widetilde{\oplus} B)}(v)$

$\le I_{\underset{i\in I}{\cap}(A_i \widetilde{\oplus} B)}(v)$

$F_{\underset{\boldsymbol{i\in I}}{\cap} \boldsymbol{A_i} \widetilde{\oplus} \boldsymbol{B}}(v) = \underset{u\in Z^n}{inf}\, max\left(1 - F_{\underset{i\in I}{\cap} \boldsymbol{A_i}}(v+u), 1 - F_B(u)\right)$

$= \underset{u\in Z^n}{inf}\, max\left(1 - \underset{i\in I}{inf}\; F_{\boldsymbol{A_i}}(v+u), 1 - F_B(u)\right)$

$= \underset{u\in Z^n}{inf}\, max\left(\underset{i\in I}{sup}\left(1 - \; F_{\boldsymbol{A_i}}(v+u)\right), 1 - F_B(u)\right)$

$= \underset{u\in Z^n}{inf}\; \underset{i\in I}{sup}\left(max\; F_{\boldsymbol{A_i}}(v+u), 1 - F_B(u)\right)$

$\ge \underset{i\in I}{sup}\; \underset{u\in Z^n}{inf}\left(max\; F_{\boldsymbol{A_i}}(v+u), 1 - F_B(u)\right)$

$\ge \underset{i\in I}{\cup}\; \underset{u\in Z^n}{inf}\left(maxF_{A_i}(v+u), 1 - F_B(u)\right) \quad \ge F_{\underset{i\in I}{\cup}(A_i \widetilde{\oplus} B)}(v)$

b) The proof is similar to a).

**Proposition 3:** for any family $(A_i | i \in I)\; in\; \mathcal{N}(\mathrm{Z}^2)\;\; and\;\; B \in \mathcal{N}(\mathrm{Z}^2)$

*a)* $\langle T_{\underset{i\in I}{\cup} \mathrm{A}_i \widetilde{\oplus} \mathrm{B}}, I_{\underset{i\in I}{\cup} \mathrm{A}_i \widetilde{\oplus} \mathrm{B}}, F_{\underset{i\in I}{\cup} \mathrm{A}_i \widetilde{\oplus} \mathrm{B}}\rangle \supseteq \langle T_{\underset{i\in I}{\cup}(\mathrm{A}_i \widetilde{\oplus} \mathrm{B})}, I_{\underset{i\in I}{\cup}(\mathrm{A}_i \widetilde{\oplus} \mathrm{B})}, F_{\underset{i\in I}{\cup}(\mathrm{A}_i \widetilde{\oplus} \mathrm{B})}\rangle$

$T_{\underset{i\in I}{\cup} \mathrm{A}_i \widetilde{\oplus} \mathrm{B}} \supseteq T_{\underset{i\in I}{\cup}(\mathrm{A}_i \widetilde{\oplus} \mathrm{B})},\; I_{\underset{i\in I}{\cup} \mathrm{A}_i \widetilde{\oplus} \mathrm{B}} \supseteq I_{\underset{i\in I}{\cup}(\mathrm{A}_i \widetilde{\oplus} \mathrm{B})}$ and $F_{\underset{i\in I}{\cup} \mathrm{A}_i \widetilde{\oplus} \mathrm{B}} \subseteq F_{\underset{i\in I}{\cap}(\mathrm{A}_i \widetilde{\oplus} \mathrm{B})}$

*b)* $\langle T_{B \widetilde{\oplus} \underset{i\in I}{\cup} \mathrm{A}_i}, I_{B \widetilde{\oplus} \underset{i\in I}{\cup} \mathrm{A}_i}, F_{B \widetilde{\oplus} \underset{i\in I}{\cup} \mathrm{A}_i}\rangle \supseteq \langle T_{\underset{i\in I}{\cup}(\mathrm{A}_i \widetilde{\oplus} \mathrm{B})}, I_{\underset{i\in I}{\cup}(\mathrm{A}_i \widetilde{\oplus} \mathrm{B})}, F_{\underset{i\in I}{\cup}(\mathrm{A}_i \widetilde{\oplus} \mathrm{B})}\rangle$

$T_{B \widetilde{\oplus} \underset{i\in I}{\cup} \mathrm{A}_i} \supseteq T_{\underset{i\in I}{\cup}(\mathrm{A}_i \widetilde{\oplus} \mathrm{B})},\; I_{B \widetilde{\oplus} \underset{i\in I}{\cup} \mathrm{A}_i} \supseteq I_{\underset{i\in I}{\cup}(\mathrm{A}_i \widetilde{\oplus} \mathrm{B})}\;\; and\;\; F_{B \widetilde{\oplus} \underset{i\in I}{\cup} \mathrm{A}_i} \subseteq F_{\underset{i\in I}{\cap}(\mathrm{A}_i \widetilde{\oplus} \mathrm{B})}$

**Proof:** **b)**

$\langle T_{B \widetilde{\oplus} \underset{i\in I}{\cup} A_i}, I_{B \widetilde{\oplus} \underset{i\in I}{\cup} A_i}, F_{B \widetilde{\oplus} \underset{i\in I}{\cup} A_i}\rangle \supseteq \langle T_{\underset{i\in I}{\cup}(B \widetilde{\oplus} A_i)}, I_{\underset{i\in I}{\cup}(B \widetilde{\oplus} A_i)}, F_{\underset{i\in I}{\cup}(B \widetilde{\oplus} A_i)}\rangle$

$$T_{B\widetilde{\oplus}\underset{i\in I}{\cup}A_i}(v) = \sup_{u\in Z^n} \min\left(T_B(v+u), T_{\underset{i\in I}{\cup}A_i}(u)\right) \quad = \sup_{u\in Z^n} \min\left(T_{\boldsymbol{B}}(v+u), \sup_{i\in I} T_{\boldsymbol{A_i}}(u)\right)$$

$$\geq \sup_{u\in Z^n}\left(\sup_{i\in I} \min T_{\boldsymbol{B}}(v+u), T_{\boldsymbol{A_i}}(u)\right) \quad \geq \underset{\boldsymbol{i\in I}}{\cup}\left(\sup_{u\in Z^n} \min T_{\boldsymbol{B}}(v+u), T_{\boldsymbol{A_i}}(u)\right)$$

$$\geq \underset{i\in I}{\cup} T_{(B\widetilde{\oplus}\boldsymbol{A_i})}(v+u) \quad \geq T_{\underset{i\in I}{\cup}(B\widetilde{\oplus}\boldsymbol{A_i})}(v)$$

$$I_{B\widetilde{\oplus}\underset{i\in I}{\cup}A_i}(v) = \sup_{u\in Z^2} \min\left(I_{\boldsymbol{B}}(v+u), I_{\underset{i\in I}{\cup}\boldsymbol{A_i}}(u)\right) \quad = \sup_{u\in Z^2} \min\left(I_{\boldsymbol{B}}(v+u), \sup_{i\in I} I_{\boldsymbol{A_i}}(u)\right)$$

$$\geq \sup_{u\in Z^2}\left(\sup_{i\in I} \min\ I_{\boldsymbol{A_i}}(v+u), I_{\boldsymbol{A_i}}(u)\right) \quad \geq \underset{\boldsymbol{i\in I}}{\cup}\left(\sup_{u\in Z^2} \min I_{\boldsymbol{A_i}}(v+u), I_{\boldsymbol{A_i}}(u)\right)$$

$$\geq \underset{i\in I}{\cup} I_{(B\widetilde{\oplus}A_i)}(v) \quad \geq I_{\underset{i\in I}{\cup}(B\widetilde{\oplus}A_i)}(v)$$

$$F_{\underset{i\in I}{\cup}\boldsymbol{A_i}\widetilde{\oplus}A_i}(v) = \inf_{u\in Z^n} \max\left(1 - F_{\boldsymbol{B}}(v+u), 1 - F_{\underset{i\in I}{\cup}A_i}(u)\right)$$

$$= \inf_{u\in Z^n} \max\left(1 - F_{\boldsymbol{B}}(v+u), 1 - \sup_{i\in I} F_{A_i}(u)\right)$$

$$= \inf_{u\in Z^n} \max\left(1 - F_{\boldsymbol{B}}(v+u), \inf_{i\in I}\left(1 - F_{A_i}(u)\right)\right)$$

$$\leq \inf_{u\in Z^n}\left(\inf_{i\in I} \max\left(1 - F_{\boldsymbol{B}}(v+u), 1 - F_{A_i}(u)\right)\right)$$

$$\leq \inf_{i\in I} \inf_{u\in Z^n}\left(\max\left(1 - F_{\boldsymbol{B}}(v+u), 1 - F_{A_i}(u)\right)\right)$$

$$\leq \underset{i\in I}{\cap} \inf_{u\in Z^n} \max\left(1 - F_B(v+u), 1 - F_{A_i}(u)\right) \quad \leq F_{\underset{i\in I}{\cap}(B\widetilde{\oplus}A_i)}(v)$$

a) The proof is similar to b).

**4.5.2. Proposition** (properties of the neutrosophic erosion)***:***

**Proposition 1:**

The neutrosophic erosion satisfies the monotonicity, $\forall A, B, C \in \mathcal{N}(Z^2)$.

a) $A \subseteq B \Longrightarrow \langle T_{A\widetilde{\ominus}C}, I_{A\widetilde{\ominus}C}, F_{A\widetilde{\ominus}C}\rangle \subseteq \langle T_{B\widetilde{\ominus}C}, I_{B\widetilde{\ominus}C}, F_{B\widetilde{\ominus}C}\rangle$

$$T_{\mathbf{A}\widetilde{\ominus}\mathbf{C}} \subseteq T_{\mathbf{B}\widetilde{\ominus}\mathbf{C}}\ ,\ I_{\mathbf{A}\widetilde{\ominus}\mathbf{C}} \subseteq I_{\mathbf{B}\widetilde{\ominus}\mathbf{C}}\ \textit{and}\ F_{\mathbf{A}\widetilde{\ominus}\mathbf{C}} \supseteq F_{\mathbf{B}\widetilde{\ominus}\mathbf{C}}$$

b) $A \subseteq B \Longrightarrow \langle T_{C\widetilde{\ominus}A}, I_{C\widetilde{\ominus}A}, F_{C\widetilde{\ominus}A}\rangle \supseteq \langle T_{C\widetilde{\ominus}B}, I_{C\widetilde{\ominus}B}, F_{C\widetilde{\ominus}B}\rangle$

$$T_{\mathbf{C}\widetilde{\ominus}\mathbf{A}} \supseteq T_{\mathbf{C}\widetilde{\ominus}\mathbf{B}}\ ,\ I_{\mathbf{C}\widetilde{\ominus}\mathbf{A}} \supseteq I_{\mathbf{C}\widetilde{\ominus}\mathbf{B}}\ \textit{and}\ F_{\mathbf{C}\widetilde{\ominus}\mathbf{A}} \subseteq F_{\mathbf{C}\widetilde{\ominus}\mathbf{B}}$$

Note that: dislike the dilation operator, the erosion does not satisfy commutativity and associativity.

**Proposition 2:**

for any family $(A_i | i \in I)\ in\ \mathcal{N}(Z^2) and\ B \in \mathcal{N}(Z^2)$

a) $\langle T_{\underset{i\in I}{\cap}A_i\widetilde{\ominus}B}, I_{\underset{i\in I}{\cap}A_i\widetilde{\ominus}B}, F_{\underset{i\in I}{\cap}A_i\widetilde{\ominus}B}\rangle \subseteq \langle T_{\underset{i\in I}{\cap}(A_i\widetilde{\ominus}B)}, I_{\underset{i\in I}{\cap}(A_i\widetilde{\ominus}B)}, F_{\underset{i\in I}{\cap}(A_i\widetilde{\ominus}B)}\rangle$

$$T_{\underset{i\in I}{\cap}\mathbf{A_i}\widetilde{\ominus}\mathbf{B}} \subseteq T_{\underset{i\in I}{\cap}(\mathbf{A_i}\widetilde{\ominus}\mathbf{B})},\ I_{\underset{i\in I}{\cap}\mathbf{A_i}\widetilde{\ominus}\mathbf{B}} \subseteq I_{\underset{i\in I}{\cap}(\mathbf{A_i}\widetilde{\ominus}\mathbf{B})} \text{ and} F_{\underset{i\in I}{\cap}\mathbf{A_i}\widetilde{\ominus}\mathbf{B}} \supseteq F_{\underset{i\in I}{\cup}(\mathbf{A_i}\widetilde{\ominus}\mathbf{B})}$$

b) $\langle T_{B\widetilde{\ominus}\underset{i\in I}{\cap}A_i}, I_{B\widetilde{\ominus}\underset{i\in I}{\cap}A_i} F_{B\widetilde{\ominus}\underset{i\in I}{\cap}A_i}\rangle \supseteq \langle T_{\underset{i\in I}{\cap}(B\widetilde{\ominus}A_i)}, I_{\underset{i\in I}{\cap}(B\widetilde{\ominus}A_i)}, F_{\underset{i\in I}{\cap}(B\widetilde{\ominus}A_i)}\rangle$

$$T_{B\widetilde{\oplus}\underset{i\in I}{\cap}A_i} \supseteq T_{\underset{i\in I}{\cup}(B\widetilde{\oplus}A_i)},\ I_{B\widetilde{\oplus}\underset{i\in I}{\cap}A_i} \supseteq I_{\underset{i\in I}{\cup}(B\widetilde{\oplus}A_i)} \text{ and} F_{B\widetilde{\oplus}\underset{i\in I}{\cap}A_i} \subseteq F_{\underset{i\in I}{\cap}(B\widetilde{\oplus}A_i)}$$

**Proof:** a)

$$\langle T_{\underset{i\in I}{\cap}A_i\widetilde{\ominus}B}, I_{\underset{i\in I}{\cap}A_i\widetilde{\ominus}B}, F_{\underset{i\in I}{\cap}A_i\widetilde{\ominus}B}\rangle \subseteq \langle T_{\underset{i\in I}{\cap}(A_i\widetilde{\ominus}B)}, I_{\underset{i\in I}{\cap}(A_i\widetilde{\ominus}B)}, F_{\underset{i\in I}{\cap}(A_i\widetilde{\ominus}B)}\rangle$$

$$\begin{aligned}
T_{\underset{i\in I}{\cap}A_i\widetilde{\ominus}B}(v) &= \inf_{u\in Z^n} \max\left(T_{\underset{i\in I}{\cap}A_i}(v+u), 1-T_B(u)\right)\\
&= \inf_{u\in Z^n} \max\left(\inf_{i\in I} T_{\boldsymbol{A_i}}(v+u), 1-T_B(u)\right)\\
&\le \inf_{u\in Z^n}\inf_{i\in I}\left(\max T_{\boldsymbol{A_i}}(v+u), 1-T_B(u)\right)\\
&\le \underset{\boldsymbol{i\in I}}{\cap}\inf_{u\in Z^n}\left(\max T_{\boldsymbol{A_i}}(v+u), 1-T_B(u)\right) \qquad \le \underset{i\in I}{\cap} T_{(A_i\widetilde{\ominus}B)}(v)
\end{aligned}$$

$$\begin{aligned}
I_{\underset{i\in I}{\cap}A_i\widetilde{\ominus}B}(v) &= \inf_{u\in Z^n} \max\left(I_{\underset{i\in I}{\cap}A_i}(v+u), 1-I_B(u)\right)\\
&= \inf_{u\in Z^n} \max\left(\min_{i\in I} I_{\boldsymbol{A_i}}(v+u), 1-I_B(u)\right)\\
&\le \inf_{u\in Z^n}\min_{i\in I}\left(\max I_{\boldsymbol{A_i}}(v+u), 1-I_B(u)\right)\\
&\le \underset{\boldsymbol{i\in I}}{\cap}\inf_{u\in Z^n}\left(\max I_{\boldsymbol{A_i}}(v+u), 1-I_B(u)\right) \qquad \le \underset{i\in I}{\cap} I_{(A_i\widetilde{\ominus}B)}(v)
\end{aligned}$$

$$\begin{aligned}
F_{\underset{i\in I}{\cap}\boldsymbol{A_i}\widetilde{\ominus}\boldsymbol{B}}(v) &= \sup_{u\in Z^n} \min\left(1-F_{\underset{i\in I}{\cap}\boldsymbol{A_i}}(v+u), F_B(u)\right)\\
&= \sup_{u\in Z^n} \min\left(1-\inf_{i\in I} F_{\boldsymbol{A_i}}(v+u), F_B(u)\right)\\
&= \sup_{u\in Z^n} \min\left(\sup_{i\in I}\left(1-F_{\boldsymbol{A_i}}(v+u)\right), F_B(u)\right)\\
&\ge \sup_{u\in Z^n}\sup_{i\in I}\left(\min 1-F_{\boldsymbol{A_i}}(v+u), F_B(u)\right)\\
&\ge \underset{i\in I}{\cup}\sup_{u\in Z^n}\left(\min 1-F_{\boldsymbol{A_i}\widetilde{\ominus}\boldsymbol{B}}(v+u), F_B(u)\right) \qquad \ge F_{\underset{i\in I}{\cup}(A_i\widetilde{\ominus}B)}(v)
\end{aligned}$$

b) The proof is similar to a).

**Proposition 3:** for any family $(A_i|i\in I)$ *in* $\mathcal{N}(Z^2)$ *and* $B\in\mathcal{N}(Z^2)$

*a)* $\langle T_{\underset{i\in I}{\cup}A_i\widetilde{\ominus}B}, I_{\underset{i\in I}{\cup}A_i\widetilde{\ominus}B}, F_{\underset{i\in I}{\cup}A_i\widetilde{\ominus}B}\rangle \supseteq \langle T_{\underset{i\in I}{\cup}(A_i\widetilde{\ominus}B)}, I_{\underset{i\in I}{\cup}(A_i\widetilde{\ominus}B)}, F_{\underset{i\in I}{\cup}(A_i\widetilde{\ominus}B)}\rangle$

$$T_{\underset{i\in I}{\cup}\boldsymbol{A_i}\widetilde{\ominus}\mathbf{B}} \supseteq T_{\underset{i\in I}{\cup}(\boldsymbol{A_i}\widetilde{\ominus}\mathbf{B})},\ I_{\underset{i\in I}{\cup}\boldsymbol{A_i}\widetilde{\ominus}\mathbf{B}} \supseteq I_{\underset{i\in I}{\cup}(\boldsymbol{A_i}\widetilde{\ominus}\mathbf{B})} \text{ and } F_{\underset{i\in I}{\cup}\boldsymbol{A_i}\widetilde{\ominus}\mathbf{B}} \subseteq F_{\underset{i\in I}{\cap}(\boldsymbol{A_i}\widetilde{\ominus}\mathbf{B})}$$

*b)* $\langle T_{B\widetilde{\ominus}\underset{i\in I}{\cup}A_i}, I_{B\widetilde{\ominus}\underset{i\in I}{\cup}A_i}, F_{B\widetilde{\ominus}\underset{i\in I}{\cup}A_i}\rangle \subseteq \langle T_{\underset{i\in I}{\cup}(B\widetilde{\ominus}A_i)}, I_{\underset{i\in I}{\cup}(B\widetilde{\ominus}A_i)}, F_{\underset{i\in I}{\cup}(B\widetilde{\ominus}A_i)}\rangle$

$$T_{B\widetilde{\ominus}\underset{i\in I}{\cup}A_i} \subseteq T_{\underset{i\in I}{\cap}(\boldsymbol{B}\widetilde{\ominus}\boldsymbol{A_i})},\ I_{B\widetilde{\ominus}\underset{i\in I}{\cup}A_i} \subseteq I_{\underset{i\in I}{\cap}(\boldsymbol{B}\widetilde{\ominus}\boldsymbol{A_i})} \textit{ and } F_{B\widetilde{\ominus}\underset{i\in I}{\cup}A_i} \supseteq F_{\underset{\boldsymbol{i\in I}}{\cup}(\boldsymbol{B}\widetilde{\ominus}\boldsymbol{A_i})}$$

**Proof: a)**

$$\langle T_{\underset{i\in I}{\cup} A_i \widetilde{\ominus} B}, I_{\underset{i\in I}{\cup} A_i \widetilde{\ominus} B}, F_{\underset{i\in I}{\cup} A_i \widetilde{\ominus} B} \rangle \supseteq \langle T_{\underset{i\in I}{\cup}(A_i \widetilde{\ominus} B)}, I_{\underset{i\in I}{\cup}(A_i \widetilde{\ominus} B)}, F_{\underset{i\in I}{\cup}(A_i \widetilde{\ominus} B)} \rangle$$

$$T_{\underset{i\in I}{\cup} A_i \widetilde{\ominus} B}(v) = \inf_{u\in Z^n} \max\left(T_{\underset{i\in I}{\cup} A_i}(v+u), T_B(u)\right)$$

$$= \inf_{u\in Z^n} \max\left(\sup_{i\in I} T_{\boldsymbol{A_i}}(v+u), T_B(u)\right) \quad = \inf_{u\in Z^n} \sup_{i\in I}\left(\max T_{\boldsymbol{A_i}}(v+u), T_B(u)\right)$$

$$\geq \underset{\boldsymbol{i\in I}}{\cup} \inf_{u\in Z^n}\left(\max T_{\boldsymbol{A_i}}(v+u), T_B(u)\right) \qquad \geq \underset{i\in I}{\cup} T_{(A_i \widetilde{\ominus} B)}(v)$$

$$\geq T_{\underset{i\in I}{\cup}(A_i \widetilde{\ominus} B)}(v)$$

$$I_{\underset{i\in I}{\cup} A_i \widetilde{\ominus} B}(v) = \inf_{u\in Z^n} \max\left(I_{\underset{i\in I}{\cup} A_i}(v+u), I_B(u)\right)$$

$$= \inf_{u\in Z^n} \max\left(\sup_{i\in I} I_{\boldsymbol{A_i}}(v+u), I_B(u)\right) \qquad = \inf_{u\in Z^n} \sup_{i\in I}\left(\max I_{\boldsymbol{A_i}}(v+u), I_B(u)\right)$$

$$\geq \underset{\boldsymbol{i\in I}}{\cup} \inf_{u\in Z^n}\left(\max I_{\,A_i}(v+u), I_B(u)\right) \qquad \geq \underset{i\in I}{\cup} I_{(A_i \widetilde{\ominus} B)}(v)$$

$$\geq I_{\underset{i\in I}{\cup}(A_i \widetilde{\ominus} B)}(v)$$

$$F_{\underset{i\in I}{\cup} \boldsymbol{A_i} \widetilde{\ominus} \boldsymbol{B}}(v) = \sup_{u\in Z^n} \min\left(1 - F_{\underset{i\in I}{\cup} \boldsymbol{A_i}}(v+u), F_B(u)\right)$$

$$= \sup_{u\in Z^n} \min\left(1 - \sup_{i\in I} F_{\boldsymbol{A_i}}(v+u), F_B(u)\right)$$

$$= \sup_{u\in Z^n} \min\left(\inf_{i\in I}\left(1 - F_{\boldsymbol{A_i}}(v+u)\right), F_B(u)\right)$$

$$= \sup_{u\in Z^n} \inf_{i\in I}\left(\min 1 - F_{\boldsymbol{A_i}}(v+u), F_B(u)\right)$$

$$\leq \inf_{i\in I} \sup_{u\in Z^n}\left(\min 1 - F_{\boldsymbol{A_i}}(v+u), F_B(u)\right)$$

$$\leq \underset{i\in I}{\cap} \sup_{u\in Z^n}\left(\min 1 - F_{A_I \widetilde{\ominus} B}(v+u), F_B(u)\right) \quad \leq F_{\underset{i\in I}{\cap}(A_i \widetilde{\ominus} B)}(v)$$

b) The proof is similar to a).

**4.5.3. Proposition** (properties of the neutrosophic closing)***:***

The neutrosophic closing satisfies the following properties

**Proposition 1:** The neutrosophic closing satisfies:

Monotonicity, $\forall\ A, B, C \in \mathcal{N}(Z^2)$

$$A \subseteq B \Longrightarrow \langle T_{A\tilde{\bullet}C}, I_{A\tilde{\bullet}C}, F_{A\tilde{\bullet}C} \rangle \subseteq \langle T_{B\tilde{\bullet}C}, I_{B\tilde{\bullet}C}, F_{B\tilde{\bullet}C} \rangle$$

$$T_{A\tilde{\bullet}C} \subseteq T_{B\tilde{\bullet}C} \ , \quad I_{A\tilde{\bullet}C} \subseteq I_{B\tilde{\bullet}C} \quad and \quad F_{A\tilde{\bullet}C} \supseteq F_{B\tilde{\bullet}C}$$

**Proposition 2:** For any family $(A_i | i \in I)$ $in\ \mathcal{N}(Z^2)$ $and$ $B \in \mathcal{N}(Z^2)$

$$\langle T_{\underset{i\in I}{\cap} A_i \tilde{\bullet} B}, I_{\underset{i\in I}{\cap} A_i \tilde{\bullet} B}, F_{\underset{i\in I}{\cap} A_i \tilde{\bullet} B} \rangle \subseteq \langle T_{\underset{i\in I}{\cap}(A_i \tilde{\bullet} B)}, I_{\underset{i\in I}{\cap}(A_i \tilde{\bullet} B)}, F_{\underset{i\in I}{\cap}(A_i \tilde{\bullet} B)} \rangle$$

$$T_{\underset{i\in I}{\cap} A_i \tilde{\bullet} B} \subseteq T_{\underset{i\in I}{\cap}(A_i \tilde{\bullet} B)},\ I_{\underset{i\in I}{\cap} A_i \tilde{\bullet} B} \subseteq I_{\underset{i\in I}{\cap}(A_i \tilde{\bullet} B)}\ \ and\ \ F_{\underset{i\in I}{\cap} A_i \tilde{\bullet} B} \supseteq F_{\underset{i\in I}{\cup}(A_i \tilde{\bullet} B)}$$

**Proposition 3:** For any family $(A_i | i \in I)\ in\ \mathcal{N}(Z^2)\ \ and\ \ B \in \mathcal{N}(Z^2)$

$$\langle T_{\underset{i\in I}{\cup} A_i \tilde{\bullet} B}, I_{\underset{i\in I}{\cup} A_i \tilde{\bullet} B}, F_{\underset{i\in I}{\cup} A_i \tilde{\bullet} B} \rangle \supseteq \langle T_{\underset{i\in I}{\cup}(A_i \tilde{\bullet} B)}, I_{\underset{i\in I}{\cup}(A_i \tilde{\bullet} B)}, F_{\underset{i\in I}{\cup}(A_i \tilde{\bullet} B)} \rangle$$

$$T_{\underset{i\in I}{\cup} A_i \tilde{\bullet} B} \supseteq T_{\underset{i\in I}{\cup}(A_i \tilde{\bullet} B)},\ I_{\underset{i\in I}{\cup} A_i \tilde{\bullet} B} \supseteq I_{\underset{i\in I}{\cup}(A_i \tilde{\bullet} B)}\ \ and\ F_{\underset{i\in I}{\cup} A_i \tilde{\bullet} B} \subseteq F_{\underset{i\in I}{\cap}(A_i \tilde{\bullet} B)}$$

**Proof:** The proof is similar to the procedure used in propositions §4.5.1 and §4.5.2.

**4.5.4. Proposition** (properties of the neutrosophic opening)**:**

The neutrosophic opening satisfies the following properties

**Proposition 1:** The neutrosophic opening satisfies:

Monotonicity: $\forall\ \ A, B, C\ \in \mathcal{N}(Z^2)$

$$A \subseteq B \Longrightarrow \langle T_{A\,\tilde{o}\,C}, I_{A\,\tilde{o}\,C}, F_{A\,\tilde{o}\,C} \rangle \subseteq \langle T_{B\,\tilde{o}\,C}, I_{B\,\tilde{o}\,C}, F_{B\,\tilde{o}\,C} \rangle$$

$$T_{A\tilde{o}C} \subseteq T_{B\tilde{o}C}\ \ ,\ \ I_{A\tilde{o}C} \subseteq I_{B\tilde{o}C}\ \ and\ \ F_{A\tilde{o}C} \supseteq F_{B\tilde{o}C}$$

**Proposition 2:** For any family $(A_i | i \in I)\ in\ \mathcal{N}(Z^2)\ and\ B \in \mathcal{N}(Z^2)$

$$\langle T_{\underset{i\in I}{\cap} A_i \tilde{o} B}, I_{\underset{i\in I}{\cap} A_i \tilde{o} B}, F_{\underset{i\in I}{\cap} A_i \tilde{o} B} \rangle \subseteq \langle T_{\underset{i\in I}{\cap}(A_i \tilde{o} B)}, I_{\underset{i\in I}{\cap}(A_i \tilde{o} B)}, F_{\underset{i\in I}{\cap}(A_i \tilde{o} B)} \rangle$$

$$T_{\underset{i\in I}{\cap} A_i \tilde{o} B} \subseteq T_{\underset{i\in I}{\cap}(A_i \tilde{o} B)},\ I_{\underset{i\in I}{\cap} A_i \tilde{o} B} \subseteq I_{\underset{i\in I}{\cap}(A_i \tilde{o} B)}\ \text{and}\ F_{\underset{i\in I}{\cap} A_i \tilde{o} B} \supseteq F_{\underset{i\in I}{\cup}(A_i \tilde{o} B)}$$

**Proposition 3:** For any family $(A_i | i \in I)\ in\ \mathcal{N}(Z^2)\ and\ B \in \mathcal{N}(Z^2)$

$$\langle T_{\underset{i\in I}{\cup} A_i \tilde{o} B}, I_{\underset{i\in I}{\cup} A_i \tilde{o} B}, F_{\underset{i\in I}{\cup} A_i \tilde{o} B} \rangle \supseteq \langle T_{\underset{i\in I}{\cup}(A_i \tilde{o} B)}, I_{\underset{i\in I}{\cup}(A_i \tilde{o} B)}, F_{\underset{i\in I}{\cup}(A_i \tilde{o} B)} \rangle$$

$$T_{\underset{i\in I}{\cup} A_i \tilde{o} B} \supseteq T_{\underset{i\in I}{\cup}(A_i \tilde{o} B)},\ I_{\underset{i\in I}{\cup} A_i \tilde{o} B} \supseteq I_{\underset{i\in I}{\cup}(A_i \tilde{o} B)}\ \text{and}\ F_{\underset{i\in I}{\cup} A_i \tilde{o} B} \subseteq F_{\underset{i\in I}{\cap}(A_i \tilde{o} B)}$$

**Proof** The proof is similar to the procedure used in propositions §4.5.1 and §4.5.2.

## 5. Conclusion

In this paper, our aim was to establish a foundation for what we called, Neutrosophic Mathematical Morphology. It is a new approach to Mathematical Morphology based on neutrosophic set theory. Several basic definitions for Neutrosophic Morphological operations were extracted and a study of its algebraic properties was presented. In addition, we were able to prove that Neutrosophic Morphological operations inherit properties and restrictions of fuzzy Mathematical Morphology. In future, we plane to apply the introduced concepts in Image Processing. For instance, Image Smoothing, Enhancement and Retrieval, as well as in medical imaging.

## References

1. Alblowi, S. A., Salama, A.A. and Eisa, M. "New Concepts of Neutrosophic Sets" International Journal of Mathematics and Computer Applications Research (IJMCAR),Vol. 4, Issue 1, pp. 59-6 (2014).
2. Bhowmik, M. and Pal, M. "Intuitionistic Neutrosophic Set". England, UK, Journal of Information and Computing Science, Vol. 4, Issue 2, pp. 142-152 (2009).
3. Bloch, I. and Maitre, H. "Fuzzy Mathematical Morphologies: a comparative study", *Pattern Recognit.*, Vol. 28, Issue 9, pp. 1341–1387 (1995).

4. De Baets, B., Kerre, E. and Gadan, M. "The Fundamentals of Fuzzy Mathematical Morphology Part 1: Basic Concept" Int. J. General System Reprint available directly from the publisher photocopying permitted by license only, Vol.23, pp. 155-171 (1995).
5. Gonzalez, R. C. and Woods, R. E." Digital Image Processing. Prentice Hall", (2001).
6. Hanbury, A. and Serra, J." Mathematical Morphology in the Cielab Space" in: Image Analysis and Stereology, Vol. 21, Issue 3, pp. 201-206 (2002).
7. Heijmans, H. "Morphological Image Operators", Advances in Electronics and Electron Physics. Boston Academic Press, (1994).
8. Heijmans, H. J. A. M., and Ronse, C. "the Algebraic Basic of Mathematical Morphology part I . Dilation and Erosion". comput. vision Graphics Image process, Vol. 50, pp. 245-295 (1989).
9. Ross, T. "Fuzzy Logic with Engineering Applications", McGraw-Hill, New York, (1995).
10. Ross, T. J., Booker, J. M., and Parkinson, W. J., eds., "Fuzzy Logic and Probability Applications", Bridging the Gap.
11. Salama, A. A. and Alblowi, S. A. "Neutrosophic Set and Neutrosophic Topological Spaces", ISOR J. Mathematics, Vol. 3, Issue 3, pp. 31-35 (2012)
12. Serra, J. " Image Analysis and Mathematical Morphology", Academic Press Inc., Landen, (1982).
13. Shinha, D. and Dougherty, E. R. "Fuzzy Mathematical Morphology", J. Vis. Commun. Image. Represent., Vol. 3, Issue 3, pp. 286–302 (1992).
14. Smarandach, F. "A Unifing Field in Logics: Neutrosophic Logic. Neutrosophy, Neutrosophic set, Neutrosophic Probability". American Research Press, Rehoboth, NM,(1991).
15. Soille, P. " Morphological Image Analysis; Principles and Applications" by pierre soille, ISBN 3-540-65671-5 (1999), 2nd edition, (2003).
16. Zadeh, L. A. " Fuzzy sets", Information and Control, Vol. 8, pp. 338-353(1965).
17. Zimmermann, H. J. "Fuzzy Set Theory and its Application", 2nd ed., Kluwer Academic Publishers, Publisher and Reprinted by Maw Chang Book Company, Taiwan, (1991).

**M. K. EL GAYYAR**

Physics and Mathematical Engineering Dept., Faculty of Engineering, Port-Said University, Egypt.
Email: mohamedelgayyar@hotmail.com

# Smooth Neutrosophic Preuniform Spaces

## Abstract

As a new branch of philosophy, the neutrosophy was presented by Smarandache in 1998. It was presented as the study of origin, nature, and scope of neutralities; as well as their interactions with different ideational spectra. The aim of this paper is to introduce the concepts of smooth neutrosophic preuniform space, smooth neutrosophic preuniform subspace, and smooth neutrosophic preuniform mappings. Furthermore, some properties of these concepts will be investigated.

## 1. Introduction

In 1984, R. Badard [[3], [4]] introduced the concept of a fuzzy preuniformity and he discussed the links between fuzzy preuniformity and fuzzy pretopology. In 1986, R. Badard [6] introduced the basic idea of smooth structure, Badard et al. [5] (1993) investigated some properties of smooth preuniform. Ramadan et al. [10] (2003) introduced smooth topologies induced by a smooth uniformity and investigated some properties of them. In 1983 the intuitionistic fuzzy set was introduced by Atanassov [[1], [2], [7]], as a generalization of fuzzy sets in Zadeh's sense [16], where besides the degree of membership of each element there was considered a degree of non-membership. Smarandache [[13], [14], [15]], defined the notion of neutrosophic set, which is a generalization of Zadeh's fuzzy sets and Atanassov's intuitionistic fuzzy set. Neutrosophic sets have been investigated by Salama et al. [[11], [12]]. The purpose of this paper is to introduce the concepts of smooth neutrosophic preunifrm space, smooth neutrosophic preuniform subspace, and smooth neutrosophic preuniform mappings. We also investigate some of their properties.

## 2. Preliminaries

In this section we use $X$ to denote a nonempty set, $I$ to denote the closed unit interval [0, 1], $I_0$ to denote the interval $(0,1]$, $I_1$ to denote the interval $[0,1)$, and $I^X$ to be the set of all fuzzy subsets defined on $X$. By $\underline{0}$ and $\underline{1}$ we denote the characteristic functions of $\phi$ and $X$, respectively. The family of all neutrosophic sets in $X$ will be denoted by $\aleph(X)$.

**2.1. Definition [14], [15]**. A neutrosophic set $A$ (NS for short) on a nonempty set $X$ is defined as: $A=\langle x,T_A(x),I_A(x),F_A(x)\rangle$ , $x\in X$ where $T,I,F:X\to[0\ ,\ 1]$ , and $0\le T_A(x)+I_A(x)+F_A(x)\le 3$ representing the degree of membership (namely $T_A(x)$), the degree of indeterminacy (namely, $I_A(x)$), and the degree of non-membership (namely $F_A(x)$); for each element $x\in X$ to the set $A$.

**2.2. Definition [13], [14].** The Null (empty) neutrosophic set $0_N$ and the absolute (universe) neutrosophic set $1_N$ are defined as follows:

$$\text{TypeI: } 0_N=\langle x,0,0,1\rangle,\ x\in X \quad , \quad 1_N=\langle x,1,1,0\rangle,\ x\in X$$

$$\text{TypeII: } 0_N=\langle x,0,1,1\rangle,\ x\in X \quad , \quad 1_N=\langle x,1,0,0\rangle,\ x\in X$$

**2.3. Definition [11], [12].** A neutrosophic set $A$ is a subset of a neutrosophic set $B$, ($A\subseteq B$), may be defined as:

$$\text{TypeI: } A\subseteq B\Leftrightarrow T_A(x)\le T_B(x),I_A(x)\le I_B(x),F_A(x)\ge F_B(x),\quad \forall x\in X$$

$$\text{TypeII: } A\subseteq B\Leftrightarrow T_A(x)\le T_B(x),I_A(x)\ge I_B(x),F_A(x)\ge F_B(x),\quad \forall x\in X$$

**2.4. Definition [11], [12].** The Complement of a neutrosophic set $A$, denoted by $coA$, is defined as:

$$\text{TypeI: } coA=\langle x,F_A(x),1-I_A(x),T_A(x)\rangle$$

$$\text{TypeII: } coA=\langle x,1-T_A(x),1-I_A(x),1-F_A(x)\rangle$$

**2.5. Definition [11], [12]**. Let $A,B\in\aleph(X)$ then:

$$\text{TypeI: } A\cup B=\langle x,\max(T_A(x),T_B(x)),\max(I_A(x),I_B(x)),\min(F_A(x),F_B(x))\rangle$$

$$\text{TypeII: } A\cup B=\langle x,\max(T_A(x),T_B(x)),\min(I_A(x),I_B(x)),\min(F_A(x),F_B(x))\rangle$$

$$\text{TypeI: } A\cap B=\langle x,\min(T_A(x),T_B(x)),\min(I_A(x),I_B(x)),\max(F_A(x),F_B(x))\rangle$$

$$\text{TypeII: } A\cap B=\langle x,\min(T_A(x),T_B(x)),\max(I_A(x),I_B(x)),\max(F_A(x),F_B(x))\rangle$$

$$[\ \ ]A=\langle x,T_A(x),I_A(x),1-T_A(x)\rangle \quad , \quad \langle\ \ \rangle A=\langle x,1-F_A(x),I_A(x),F_A(x)\rangle$$

**2.6. Definition [11], [12]**. Let $\{A_i\},i\in J$ be an arbitrary family of neutrosophic sets, then:

$$\text{TypeI: } \bigcup_{i\in J} A_i=\left\langle x,\sup_{i\in j} T_{A_i}(x),\sup_{i\in j} I_{A_i}(x),\inf_{i\in j} F_{A_i}(x)\right\rangle$$

$$\text{TypeII: } \bigcup_{i\in J} A_i=\left\langle x,\sup_{i\in j} T_{A_i}(x),\inf_{i\in j} I_{A_i}(x),\inf_{i\in j} F_{A_i}(x)\right\rangle$$

$$\text{Type I: } \bigcap_{i\in J} A_i = \left\langle x, \inf_{i\in j} T_{A_i}(x), \inf_{i\in j} I_{A_i}(x), \sup_{i\in j} F_{A_i}(x) \right\rangle$$

$$\text{Type II: } \bigcap_{i\in J} A_i = \left\langle x, \inf_{i\in j} T_{A_i}(x), \sup_{i\in j} I_{A_i}(x), \sup_{i\in j} F_{A_i}(x) \right\rangle$$

**2.7. Definition [11], [12].** The difference between two neutrosophic sets $A$ and $B$ defined as $A \setminus B = A \cap coB$.

**2.8. Definition [11], [12].** Every intuitionistic fuzzy set $A$ on $X$ is NS having the form $A = \langle x, T_A(x), 1-(T_A(x)+F_A(x)), F_A(x)\rangle$, and every fuzzy set $A$ on $X$ is NS having the form $A = \langle x, T_A(x), 0, 1-T_A(x)\rangle$, $x \in X$.

**2.9. Definition [8].** Let $Y$ be a subset of $X$ and $A \in I^X$; the restriction of $A$ on $Y$ is denoted by $A_{/Y}$. For each $B \in I^Y$, the extension of $B$ on $X$, denoted by $B_X$, is defined by:

$$B_X = \begin{cases} B(x) & \text{if } x \in Y \\ 0.5 & \text{if } x \in X - Y \end{cases}$$

**2.10. Definition [6].** A smooth topological space (STS) is an ordered pair $(X, \tau)$, where $X$ is a nonempty set and $\tau : I^X \to I$ is a mapping satisfying the following properties:

(O1) $\tau(\underline{0}) = \tau(\underline{1}) = 1$

(O2) $\forall A_1, A_2 \in I^X,\ \tau(A_1 \cap A_2) \geq \tau(A_1) \wedge \tau(A_2)$

(O3) $\forall A_i, i \in J,\ \tau(\bigcup_{i\in J} A_i) \geq \bigwedge_{i\in J} \tau(A_i)$

**2.11. Definition [5].** A fuzzy preuniform structure $U^*$ on $X$ is a family of fuzzy sets in $X \times X$, called entourages which satisfies:

($FP_1$) $\Delta \subseteq u$, for every $u \in U^*$, where $\Delta$ is the diagonal :

$$\Delta(x,y) = \begin{cases} 0 & ,\text{if } x \neq y \\ 1 & ,\text{if } x = y \end{cases} \quad ,\forall x, y \in X$$

($FP_2$) If $u \in U^*$ and $u \subseteq v$, then $v \in U^*$ $\quad ,\forall\ u, v \in I^{X\times X}$

The pair $(X, U^*)$ is said to be a fuzzy preuniform space.

**2.12. Definition [5].** Let $(X, U^*)$ be a fuzzy preuniform space, the following potential properties are considered:

($FP_3$) For every $u \in U^*$, we can assert that $u^{-1} \in U^*$, where $u^{-1}(x,y) = u(y,x)$.

In this case, $(X, U^*)$ is said to be symmetrical.

($FP_4$) For every $u, v \in U^*$, we can assert that $u \cap v \in U^*$.

In this case, $(X, U^*)$ is said to be of type D.

$(FP_5)$ For every $u \in U^*$, there exists $v \in U^*$ such that $v \Theta v \subseteq u$, where
$v\Theta v(x,y) = \sup\{v(x,z) \wedge v(z,y) : z \in X\}$

In this case, $(X, U^*)$ is said to be of type S.

**Note** that a fuzzy preunifom space which is symmetrical, of type D and of type S is a fuzzy uniform space as defined by Hutton [9].

**2.13. Definition [5].** A smooth preuniform structure $U$ on $X$ is a fuzzy set in the fuzzy sets in $X \times X$. $U$ is an element of $I^{I^{X\times X}}$ which satisfies:

$(SP_1)$ $\Delta \not\subseteq u \Rightarrow U(u) = 0$ for every $u \in I^{X\times X}$

$(SP_2)$ $u \subseteq v \Rightarrow U(v) \geq U(u)$, $\forall\ u, v \in I^{X\times X}$

$(SP_3)$ $U(X \times X) = 1$, where $(X \times X)(x,y) = 1$, for every $x, y \in X$

The pair $(X,U)$ is said to be a smooth preuniform space.

**2.14. Definition [5].** Let $(X,U)$ be a smooth preuniform space, the following potential properties are considered:

$(SP_4)$ For every $u \in I^{X\times X}$, we have $U(u) = U(u^{-1})$, where $u^{-1}(x,y) = u(y,x)$.

In this case, $(X,U)$ is said to be symmetrical.

$(SP_5)$ For every $u, v \in I^{X\times X}$, we have $U(u \cap v) \geq U(u) \wedge U(v)$,
but from $(SP_2)$ we can write $U(u \cap v) = U(u) \wedge U(v)$

In this case, $(X,U)$ is said to be of type D.

$(SP_6)$ If we have $\sup_{v \in I^{X\times X}} \{U(v) : v\Theta v \subseteq u\}) \geq U(u)$, for every $u \in I^{X\times X}$

In this case, $(X,U)$ is said to be of type S.

## 3. Smooth Neutrosophic Preuniform Spaces

Now, we will define two types of smooth neutrosophic preuniform spaces,

a smooth neutrosophic preuniform space (SNPS) take the form $(X, U^T, U^I, U^F)$ and the mappings $U^T, U^I, U^F : I^{X\times X} \to I$ represent the degree of membership, the degree of indeterminacy, and the degree of non-membership respectively.

### 3.1. Smooth Neutrosophic preuniform Spaces of type I

**3.1.1. Definition.** A smooth neutrosophic preuniformity $(U^T, U^I, U^F)$ of type I satisfying the following axioms:

$(SNPI_1)$ $\Delta \not\subseteq u \Rightarrow U^T(u) = U^I(u) = 0$ and $U^F(u) = 1$ , for every $u \in I^{X\times X}$

$(SNPI_2)$ $u \subseteq v \Rightarrow U^T(v) \geq U^T(u)$, $U^I(v) \geq U^I(u)$, and $U^F(v) \leq U^F(u)$, $\forall$ $u, v \in I^{X\times X}$ $(X, U^T, U^I, U^F)$

$(SNPI_3)$ $U^T(X\times X) = U^I(X\times X) = 1$ , and $U^F(X\times X) = 0$

is said to be a smooth neutrosophic preuniform space of type I

$(SNPI_4)$ For every $u \in I^{X\times X}$, we have:

$$U^T(u) = U^T(u^{-1}), U^I(u) = U^I(u^{-1}), \text{and } U^F(u) = U^F(u^{-1}).$$

In this case, $(X, U^T, U^I, U^F)$ is said to be symmetrical.

$(SNPI_5)$ For every $u, v \in I^{X\times X}$, we have:

$$U^T(u \cap v) \geq U^T(u) \wedge U^T(v), U^I(u \cap v) \geq U^I(u) \wedge U^I(v), \text{and}$$

$$U^F(u \cap v) \leq U^F(u) \vee U^F(v), \text{but from } (SNPI_2) \text{ we can write}$$

$$U^T(u \cap v) = U^T(u) \wedge U^T(v), U^I(u \cap v) = U^I(u) \wedge U^I(v), \text{and}$$

$$U^F(u \cap v) = U^F(u) \vee U^F(v).$$

In this case, $(X, U^T, U^I, U^F)$ is said to be of type D.

$(SNPI_6)$ If we have:

$$\sup_{v\in I^{X\times X}} \{U^T(v) : v\Theta v \subseteq u\}) \geq U^T(u), \quad \sup_{v\in I^{X\times X}} \{U^I(v) : v\Theta v \subseteq u\}) \geq U^I(u), \text{and}$$

$$\inf_{v\in I^{X\times X}} \{U^F(v) : v\Theta v \subseteq u\}) \leq U^F(u), \qquad \text{for every } u \in I^{X\times X}.$$

In this case, $(X, U^T, U^I, U^F)$ is said to be of type S.

**3.1.2. Example.** Let $X = \{a, b\}$. Define the mappings $U^T, U^I, U^F : I^{X\times X} \rightarrow I$ as:

$$U^T(u) = \begin{cases} 1 & , \text{if } u = X\times X \\ 0.5 & , \text{if } \Delta \subseteq u \\ 0 & , \text{otherwise} \end{cases}$$

$$U^I(u) = \begin{cases} 1 & , \text{if } u = X\times X \\ 0.6 & , \text{if } \Delta \subseteq u \\ 0 & , \text{otherwise} \end{cases}$$

$$U^F(u) = \begin{cases} 0 & , \text{if } u = X\times X \\ 0.3 & , \text{if } \Delta \subseteq u \\ 1 & , \text{otherwise} \end{cases}$$

Then $(X, U^T, U^I, U^F)$ is a smooth neutrosophic preuniform space of type I on $X$.

**3.1.3.Remark.** Both $U^T$ and $U^I$ with their conditions are smooth preuniformities.

**3.1.4.Proposition.** Let $\{(U_i^T, U_i^I, U_i^F)\}_{i\in J}$ be a family of smooth neutrosophic preuniformitiess on $X$. Then $\bigcup_{i\in J} (U_i^T, U_i^I, U_i^F)$ and $\bigcap_{i\in J} (U_i^T, U_i^I, U_i^F)$ are smooth neutrosophic preuniformities on $X$.

**Proof.** First, for $\bigcup_{i\in J}(U_i^T, U_i^I, U_i^F)$ :

(SNPI$_1$) Let $u \in I^{X\times X}$ and $\Delta \not\subseteq u$ . It follows that $U_i^T(u) = U_i^I(u) = 0$ and

$U_i^F(u) = 1$, for every $i \in J$, hence $\sup_{i\in J} U_i^T(u) = \sup_{i\in J} U_i^I(u) = 0$ and $\inf_{i\in J} U_i^F(u) = 1$.

(SNPI$_2$) Let $u, v \in I^{X\times X}$ such that $u \subseteq v$. Then for every $i \in J$ we have $U_i^T(v) \geq U_i^T(u)$,

$U_i^I(v) \geq U_i^I(u)$, and $U_i^F(v) \leq U_i^F(u)$, hence $\sup_{i\in J} U_i^T(v) \geq \sup_{i\in J} U_i^T(u)$,

$\sup_{i\in J} U_i^I(v) \geq \sup_{i\in J} U_i^I(u)$, and $\inf_{i\in J} U_i^F(v) \leq \inf_{i\in J} U_i^F(u)$.

(SNPI$_3$) $U_i^T(X\times X) = U_i^I(X\times X) = 1$ , and $U_i^F(X\times X) = 0$, for every $i \in J$. Then

$\sup_{i\in J} U_i^T(X\times X) = \sup_{i\in J} U_i^I(X\times X) = 1$ , and $\inf_{i\in J} U_i^F(X\times X) = 0$.

Second, the proof for $\bigcap_{i\in J}(U_i^T, U_i^I, U_i^F)$ is similar to the first

**3.1.5.Proposition.** Let $\{(U_i^T, U_i^I, U_i^F)\}_{i\in J}$ be a family of smooth neutrosophic preuniformitiess on $X$ .Then:

(i) If every $(U_i^T, U_i^I, U_i^F)$ is symmetrical, then $\bigcup_{i\in J}(U_i^T, U_i^I, U_i^F)$ and $\bigcap_{i\in J}(U_i^T, U_i^I, U_i^F)$ are also symmetrical.

(ii) If every $(U_i^T, U_i^I, U_i^F)$ is of type $D$ , then $\bigcap_{i\in J}(U_i^T, U_i^I, U_i^F)$ is also of type $D$

(iii) If every $(U_i^T, U_i^I, U_i^F)$ is of type $S$ , then $\bigcup_{i\in J}(U_i^T, U_i^I, U_i^F)$ is also of type $S$

Proof. (i) Let $u \in I^{X\times X}$ ,then $U_i^T(u) = U_i^T(u^{-1})$, $U_i^I(u) = U_i^I(u^{-1})$, and

$U_i^F(u) = U_i^F(u^{-1})$ $\forall$ $i \in J$, hence $\sup_{i\in J} U_i^T(u) = \sup_{i\in J} U_i^T(u^{-1})$, $\sup_{i\in J} U_i^I(u) = \sup_{i\in J} U_i^I(u^{-1})$,

and $\inf_{i\in J} U_i^F(u) = \inf_{i\in J} U_i^F(u^{-1})$, also $\inf_{i\in J} U_i^T(u) = \inf_{i\in J} U_i^T(u^{-1})$, $\inf_{i\in J} U_i^I(u) = \inf_{i\in J} U_i^I(u^{-1})$,

and $\sup_{i\in J} U_i^F(u) = \sup_{i\in J} U_i^F(u^{-1})$

(ii) Let $u, v \in I^{X\times X}$, then $U_i^T(u\cap v) = U_i^T(u)\wedge U_i^T(v), U_i^I(u\cap v) = U_i^I(u)\wedge U_i^I(v)$, and $U_i^F(u\cap v) = U_i^F(u)\vee U_i^F(v)\ \forall\ i\in J$, hence $\inf_{i\in J} U_i^T(u\cap v) = \inf_{i\in J} U_i^T(u)\wedge \inf_{i\in J} U_i^T(v)$, $\inf_{i\in J} U_i^I(u\cap v) = \inf_{i\in J} U_i^I(u)\wedge \inf_{i\in J} U_i^I(v)$, and $\sup_{i\in J} U_i^F(u\cap v) = \sup_{i\in J} U_i^F(u)\vee \sup_{i\in J} U_i^F(v)$.

(iii) Let $\sup_{v\in I^{X\times X}}\{U_i^T(v): v\Theta v\subseteq u\}) \geq U_i^T(u)$, $\sup_{v\in I^{X\times X}}\{U_i^I(v): v\Theta v\subseteq u\}) \geq U_i^I(u)$, and $\inf_{v\in I^{X\times X}}\{U_i^F(v): v\Theta v\subseteq u\}) \leq U_i^F(u), \forall u\in I^{X\times X}$, then $\sup_{v\in I^{X\times X}}\{\sup_{i\in J} U_i^T(v): v\Theta v\subseteq u\}) \geq \sup_{i\in J} U_i^T(u)$, $\sup_{v\in I^{X\times X}}\{\sup_{i\in J} U_i^I(v): v\Theta v\subseteq u\}) \geq \sup_{i\in J} U_i^I(u)$, and $\inf_{v\in I^{X\times X}}\{\inf_{i\in J} U_i^F(v): v\Theta v\subseteq u\}) \leq \inf_{i\in J} U_i^F(u)$.

Next, we will introduce a kind of subspace of a smooth neutrosophic preuniform space and some hereditary properties.

**3.1.6.Definition.** Let $A$ be a nonempty subset of $X$ and let $u\in I^{A\times A}$. We define the extension of $u$ to $X\times X$, denoted $u_{X\times X}$ by:

$$u_{X\times X}(x,y) = \begin{cases} u(x,y) & ,\text{if } x,y\in A \\ 0.5 & ,\text{ otherwise}\end{cases}$$

**3.1.7.Definition.** Let $A$ be a nonempty subset of $X$. We define the subdiagonal $\Delta_A \in I^{X\times X}$ by:

$$\Delta_A(x,y) = \begin{cases} 1 & ,\text{if } x=y\in A \\ 0 & ,\text{ otherwise}\end{cases}$$

One may notice that $(\Delta_A \cup \Delta_{coA}) = \Delta$

**3.1.8. Proposition.** Let $(X, U^T, U^I, U^F)$ be a smooth neutrosophic preuniform space and let $A$ be a nonempty subset of $X$, and the mappings $U_A^T, U_A^I, U_A^F : I^{A\times A} \to I$ defined by:

$$U_A^T(u) = \begin{cases} 1 & ,\text{ if } u = A\times A \\ U^T(u_{X\times X}\cup \Delta_{coA}) & ,\ \forall u\in I^{A\times A}\setminus\{A\times A\}\end{cases},$$

$$U_A^I(u) = \begin{cases} 1 & ,\text{ if } u = A\times A \\ U^I(u_{X\times X}\cup \Delta_{coA}) & ,\ \forall u\in I^{A\times A}\setminus\{A\times A\}\end{cases}, \text{ and}$$

$$U_A^F(u) = \begin{cases} 0 & ,\text{ if } u = A\times A \\ U^F(u_{X\times X}\cup \Delta_{coA}) & ,\ \forall u\in I^{A\times A}\setminus\{A\times A\}\end{cases}, \text{ where}$$

$(A\times A)(x,y) = 1$ for every $x,y\in A$. Then $(U_A^T, U_A^I, U_A^F)$ is a smooth neutrosophic preuniformity on $A$ .

Proof. $(SNPI_1)$ Let $\Delta^A \in I^{A\times A}$ be the diagonal in A.Then, $\forall\ u\in I^{A\times A}$ we have :

$$\Delta^A \not\subseteq u \Rightarrow \Delta \not\subseteq (u_{X\times X}\cup\Delta_{coA}) \Rightarrow U^T(u_{X\times X}\cup\Delta_{coA}) = U^I(u_{X\times X}\cup\Delta_{coA}) = 0$$
$$\text{and } U^F(u_{X\times X}\cup\Delta_{coA}) = 1 \Rightarrow U_A^T(u) = U_A^I(u) = 0 \text{ and } U_A^F(u) = 1.$$

$(SNPI_2)$ Let $u, v \in I^{A\times A}$ such that $u \subseteq v$. Then it follows that $(u_{X\times X} \cup \Delta_{coA}) \subseteq (v_{X\times X} \cup \Delta_{coA})$, hence $U^T(v_{X\times X} \cup \Delta_{coA}) \geq U^T(u_{X\times X} \cup \Delta_{coA})$, $U^I(v_{X\times X} \cup \Delta_{coA}) \geq U^I(u_{X\times X} \cup \Delta_{coA})$, and $U^F(v_{X\times X} \cup \Delta_{coA}) \leq U^F(u_{X\times X} \cup \Delta_{coA})$. So, $U_A^T(v) \geq U_A^T(u)$, $U_A^I(v) \geq U_A^I(u)$, and $U_A^F(v) \leq U_A^F(u)$. $(SNPI_3)$ The proof is straightforward from the definition.

**3.1.9. Definition.** The smooth neutrosophic preuniform space $(A, U_A^T, U_A^I, U_A^F)$ is called a subspace of $(X, U^T, U^I, U^F)$ and $(U_A^T, U_A^I, U_A^F)$ is called the smooth neutrosophic preuniformity on $A$ induced by $(U^T, U^I, U^F)$.

**3.1.10. Proposition.** Let $(U^T, U^I, U^F)$ be a smooth neutrosophic preuniformity on $X$, $A$ be a nonempty subset of $X$ and $(U_A^T, U_A^I, U_A^F)$ be the corresponding smooth neutrosophic preuniformity on $A$ induced by $(U^T, U^I, U^F)$. Then the properties $(SNPI_4)$ and $(SNTI_5)$ are hereditary.

Proof. $(SNPI_4)$ Let $u \in I^{A\times A}$. Then it follows:

(1) If $u = A\times A$, we find that $U_A^T(u) = U_A^T(u^{-1}) = U_A^I(u) = U_A^I(u^{-1}) = 1$, and $U_A^F(u) = U_A^F(u^{-1}) = 0$.

(2) If $u \neq A\times A$, we find that $U_A^T(u) = U^T(u_{X\times X} \cup \Delta_{coA}) = U^T((u_{X\times X} \cup \Delta_{coA})^{-1})$

$$= U^T((u_{X\times X})^{-1} \cup (\Delta_{coA})^{-1}) = U^T((u^{-1})_{X\times X} \cup \Delta_{coA}) = U_A^T(u^{-1}), \text{ because}$$

$$(u_{X\times X})^{-1}(x,y) = u_{X\times X}(y,x) = \begin{cases} u(y,x) & ,\text{if } x,y \in A \\ 0.5 & ,\text{ otherwise} \end{cases} = \begin{cases} u^{-1}(x,y) & ,\text{if } x,y \in A \\ 0.5 & ,\text{ otherwise} \end{cases}$$

$$= (u^{-1})_{X\times X}(x,y).$$

Similarly, we can prove that $U_A^I(u) = U_A^I(u^{-1})$ and $U_A^F(u) = U_A^F(u^{-1})$.

$(SNPI_5)$ Let $u, v \in I^{A\times A}$. Then we obtain successively:

$$U_A^T(u \cap v) = U^T((u \cap v)_{X\times X} \cup \Delta_{coA}) = U^T((u_{X\times X} \cup \Delta_{coA}) \cap (v_{X\times X} \cup \Delta_{coA}))$$

$$\geq U^T(u_{X\times X} \cup \Delta_{coA}) \wedge U^T(v_{X\times X} \cup \Delta_{coA})) = U_A^T(u) \wedge U_A^T(v).$$ Similarly, we can prove that $U_A^I(u \cap v) \geq U_A^I(u) \wedge U_A^I(v)$ and $U_A^F(u \cap v) \leq U_A^F(u) \vee U_A^F(v)$.

**3.1.11. Definition.** Consider two ordinary sets $X, Y$ and a mapping $f$ from $X$ into $Y$. The fuzzy product $f \otimes f$ is defined as the following mapping:

$$f \otimes f : I^{X\times X} \to I^{Y\times Y}$$

$$u \mapsto (f \otimes f)(u),\ \forall\, u \in I^{X\times X}$$

where $(f \otimes f)(u)$ is defined as the following fuzzy set in $Y\times Y$:

$(f \otimes f)(u) : Y\times Y \to I$

$$(y_1, y_2) \mapsto \begin{cases} \sup\{u(x_1, x_2) : x_1, x_2 \in X, f(x_1) = y_1 \text{ and } f(x_2) = y_2\} & ,\text{if } y_1, y_2 \in \text{rng}(f) \\ 0 & ,\text{otherwise} \end{cases}$$

**3.1.12. Definition.** Consider two ordinary sets $X, Y$ .Let $f$ be a mapping from $X$ to $Y$ and $v \in I^{Y\times Y}$ .Then the inverse image of $v$ under $(f \otimes f)$ is defined as the following fuzzy set in $X \times X$:

$$(f \otimes f)^{-1}(v) : X \times X \to I$$
$$(x_1, x_2) \mapsto v(f \otimes f)(x_1, x_2) = v\,(f(x_1), f(x_2)),\ \forall\ x_1, x_2 \in X$$

**3.1.13. Definition.** A map $f : X \to Y$ is called weakly smooth neutrosophic preuniform with respect to the smooth neutrosophic preuniformities $(U_1^T, U_1^I, U_1^F)$ on $X$ and $(U_2^T, U_2^I, U_2^F)$ on $Y$ iff for every $v \in I^{Y\times Y}$ we have:

$$U_2^T(v) > 0 \Rightarrow U_1^T((f \otimes f)^{-1}(v)) > 0, U_2^I(v) > 0 \Rightarrow U_1^I((f \otimes f)^{-1}(v)) > 0,$$
$$\text{and}\ U_2^F(v) < 1 \Rightarrow U_1^F((f \otimes f)^{-1}(v)) < 1.$$

**3.1.14. Definition.** A map $f : X \to Y$ is called smooth neutrosophic preuniform with respect to the smooth neutrosophic preuniformities $(U_1^T, U_1^I, U_1^F)$ on $X$ and $(U_2^T, U_2^I, U_2^F)$ on $Y$ iff for every $v \in I^{Y\times Y}$ we have:

$$U_1^T((f \otimes f)^{-1}(v)) \ge U_2^T(v), U_1^I((f \otimes f)^{-1}(v)) \ge U_2^I(v),\ \text{and}\ U_1^F((f \otimes f)^{-1}(v)) \le U_2^F(v).$$

**3.1.15. Definition.** A map $f : X \to Y$ is called smooth neutrosophic direct preuniform with respect to the smooth neutrosophic preuniformities $(U_1^T, U_1^I, U_1^F)$ on $X$ and $(U_2^T, U_2^I, U_2^F)$ on $Y$ iff for every $u \in I^{X\times X}$ we have:

$$U_2^T((f \otimes f)(u)) \ge U_1^T(u),\ U_2^I((f \otimes f)(u)) \ge U_1^I(u),\ \text{and}\ U_2^F((f \otimes f)(u)) \le U_1^F(u).$$

**3.1.16. Definition.** A map $f : X \to Y$ is called a (weakly) smooth neutrosophic homeomorphism with respect to the smooth neutrosophic preuniformities $(U_1^T, U_1^I, U_1^F)$ on $X$ and $(U_2^T, U_2^I, U_2^F)$ on $Y$ iff $f$ is bijective and $f$ , $f^{-1}$ are (weakly) smooth neutrosophic preuniform.

**3.1.17. Proposition.** Let $(X, U_1^T, U_1^I, U_1^F)$ and $(Y, U_2^T, U_2^I, U_2^F)$ be two smooth neutrosophic preuniform spaces and $f : X \to Y$ a bijective mapping. The following statements are equivalent:

(i) f is a smooth netrosophic homeomorphism.

(ii)f is smooth netrosophic preuniform and smooth netrosophic direct preuniform

**Proof.** (i) $\Rightarrow$ (ii) .Let $f$ be a smooth neutrosophic homeomorphism, then $f$ is smooth neutrosophic preuniform, and for every $u \in I^{X\times X}$ we have:

$$U_2^T((f^{-1} \otimes f^{-1})^{-1}(u)) \ge U_1^T(u), U_2^I((f^{-1} \otimes f^{-1})^{-1}(u)) \ge U_1^I(u),$$
$$\text{and}\ U_2^F((f^{-1} \otimes f^{-1})^{-1}(u)) \le U_1^F(u).$$

Applying the definitions and from the bijectivity of $f$ we obtain the following result for $y_1, y_2 \in Y$:

$$((f^{-1} \otimes f^{-1})^{-1}(u)(y_1, y_2) = u(f^{-1} \otimes f^{-1})(y_1, y_2) = u(f^{-1}(y_1), f^{-1}(y_2)) = u(x_1, x_2)$$
$$= ((f \otimes f)(u))(y_1, y_2).\ \text{So}\ U_2^T((f \otimes f)(u)) \ge U_1^T(u), U_2^I((f \otimes f)(u)) \ge U_1^I(u),\ \text{and}$$
$$U_2^F((f \otimes f)(u)) \le U_1^F(u),\ \text{hence } f \text{ is smooth neutrosoph ic direct preuniform}.$$

(ii) $\Rightarrow$(i) Let $f$ be smooth neutrosophic preuniform and smooth neutrosophic direct preuniform, then for every $u \in I^{X\times X}$ we have:

$U_2^T((f^{-1}\otimes f^{-1})^{-1}(u)) = U_2^T((f\otimes f)(u)) \ge U_1^T(u), U_2^I((f^{-1}\otimes f^{-1})^{-1}(u)) =$

$U_2^I((f\otimes f)(u)) \ge U_1^I(u)$, and $U_2^F((f^{-1}\otimes f^{-1})^{-1}(u)) = U_2^F((f\otimes f)(u)) \le U_1^F(u)$.

Because $f$ is bijective, then $f^{-1}$ is smooth neutrosophic preuniform, hence $f$ is a smooth neutrosophic homeomorphism.

**3.1.18. Proposition.** Let $f: X \to Y$ be a bijective and smooth neutrosophic direct preuniform mapping with respect to the smooth neutrosophic preuniformities $(U^T, U^I, U^F)$ on $X$ and $(U'^T, U'^I, U'^F)$ on $Y$ and let $A$ be a nonempty subset of $X$, then the restriction mapping

$$f_{/A} : (A, U_A^T, U_A^I, U_A^F) \to (f(A), U'^T_{f(A)}, U'^I_{f(A)}, U'^F_{f(A)})$$

Is smooth neutrosophic direct preuniform.

**Proof.** For every $u \in I^{A\times A}$ we have:

$$U'^T_{f(A)}(((f_{/A})\otimes(f_{/A}))(u)) = U'^T((((f_{/A})\otimes(f_{/A}))(u))_{Y\times Y} \cup \Delta_{cof(A)}),$$

$$U'^I_{f(A)}(((f_{/A})\otimes(f_{/A}))(u)) = U'^I((((f_{/A})\otimes(f_{/A}))(u))_{Y\times Y} \cup \Delta_{cof(A)}),$$

$$U'^F_{f(A)}(((f_{/A})\otimes(f_{/A}))(u)) = U'^F((((f_{/A})\otimes(f_{/A}))(u))_{Y\times Y} \cup \Delta_{cof(A)}),$$

$$U_A^T(u) = U^T(u_{X\times X} \cup \Delta_{coA}) \le U'^T((f\otimes f)(u_{X\times X} \cup \Delta_{coA})),$$

$$U_A^I(u) = U^I(u_{X\times X} \cup \Delta_{coA}) \le U'^I((f\otimes f)(u_{X\times X} \cup \Delta_{coA})), \text{ and}$$

$$U_A^F(u) = U^F(u_{X\times X} \cup \Delta_{coA}) \ge U'^F((f\otimes f)(u_{X\times X} \cup \Delta_{coA})).$$

Applying the definitions and from the bijectivity of $f$ we obtain the following result for $y_1, y_2 \in Y$:

$$((((f_{/A})\otimes(f_{/A}))(u))_{Y\times Y} \cup \Delta_{cof(A)})(y_1, y_2)$$

$$= \begin{cases} 1 & \text{,if } y_1 = y_2 \in cof(A) \\ ((f_{/A})\otimes(f_{/A}))(u)(y_1, y_2) & \text{,if } y_1, y_2 \in f(A) \\ 0.5 & \text{,otherwise} \end{cases}$$

$$= \begin{cases} 1 & \text{,if } f(x_1) = y_1, = f(x_2) = y_2 \text{ and } x_1 = x_2 \in coA \\ u(x_1, x_2) & \text{,if } f(x_1) = y_1, = f(x_2) = y_2 \text{ and } x_1, x_2 \in A \\ 0.5 & \text{,otherwise} \end{cases}$$

$$= ((f\otimes f)(u_{X\times X} \cup \Delta_{coA}))(y_1, y_2).$$

So, $U'^T_{f(A)}(((f_{/A})\otimes(f_{/A}))(u)) \ge U_A^T(u)$, $U'^I_{f(A)}(((f_{/A})\otimes(f_{/A}))(u)) \ge U_A^I(u)$, and $U'^F_{f(A)}(((f_{/A})\otimes(f_{/A}))(u)) \le U_A^F(u)$, hence $f_{/A}$ is smooth neutrosophic direct preuniform.

3.2. Smooth Neutrosophic Preuniform Spaces of type II

In this part we will consider the definitions of type II. In a similar way as in type I, we can state the following definitions and propositions. The proofs of the propositions in type II will be similar to the proofs of the propositions in type I.

**3.2.1. Definition.** A smooth neutrosophic preuniformity $(U^T, U^I, U^F)$ of type II satisfying the following axioms:

($SNPII_1$) $\Delta \not\subseteq u \Rightarrow U^T(u) = 0$ and $U^I(u) = U^F(u) = 1$, for every $u \in I^{X \times X}$

($SNPII_2$) $u \subseteq v \Rightarrow U^T(v) \geq U^T(u), U^I(v) \leq U^I(u)$, and $U^F(v) \leq U^F(u), \forall u, v \in I^{X \times X}$ $(X, U^T, U^I, U^F)$

($SNPII_3$) $U^T(X \times X) = 1$, and $U^I(X \times X) = U^F(X \times X) = 0$

is said to be a smooth neutrosophic preuniform space of type II. Also, for type II:

($SNPII_4$) For every $u \in I^{X \times X}$, we have:

$$U^T(u) = U^T(u^{-1}), U^I(u) = U^I(u^{-1}), \text{and } U^F(u) = U^F(u^{-1}).$$

In this case, $(X, U^T, U^I, U^F)$ is said to be symmetrical.

($SNPII_5$) For every $u, v \in I^{X \times X}$, we have:

$$U^T(u \cap v) \geq U^T(u) \wedge U^T(v), U^I(u \cap v) \leq U^I(u) \vee U^I(v), \text{and}$$
$$U^F(u \cap v) \leq U^F(u) \vee U^F(v), \text{but from } (SNPII_2) \text{ we can write}$$
$$U^T(u \cap v) = U^T(u) \wedge U^T(v), U^I(u \cap v) = U^I(u) \vee U^I(v), \text{and}$$
$$U^F(u \cap v) = U^F(u) \vee U^F(v).$$

In this case, $(X, U^T, U^I, U^F)$ is said to be of type D.

($SNPII_6$) If we have:

$$\sup_{v \in I^{X \times X}} \{U^T(v) : v \Theta v \subseteq u\}) \geq U^T(u), \inf_{v \in I^{X \times X}} \{U^I(v) : v \Theta v \subseteq u\}) \leq U^I(u), \text{and}$$
$$\inf_{v \in I^{X \times X}} \{U^F(v) : v \Theta v \subseteq u\}) \leq U^F(u), \quad \text{for every } u \in I^{X \times X}.$$

In this case, $(X, U^T, U^I, U^F)$ is said to be of type S.

**3.2.2. Example.** Let $X = \{a, b\}$. Define the mappings $U^T, U^I, U^F : I^{X \times X} \to I$ as:

$$U^T(u) = \begin{cases} 1 & , \text{if } u = X \times X \\ 0.5 & , \text{if } \Delta \subseteq u \\ 0 & , \text{otherwise} \end{cases}$$

$$U^I(u) = \begin{cases} 0 & , \text{if } u = X \times X \\ 0.4 & , \text{if } \Delta \subseteq u \\ 1 & , \text{otherwise} \end{cases}$$

$$U^F(u) = \begin{cases} 0 & , \text{if } u = X \times X \\ 0.7 & , \text{if } \Delta \subseteq u \\ 1 & , \text{otherwise} \end{cases}$$

Then $(X, U^T, U^I, U^F)$ is a smooth neutrosophic preuniform space of type II on $X$.

**3.2.3. Remark.** $U^T$ with its conditions is smooth preuniformity.

**Note** that: the propositions (3.1.4) and (3.1.5) are satisfied for type II.

Proposition.

Let $(X, U^T, U^I, U^F)$ be a smooth neutrosophic preuniform space and let $A$ be a nonempty subset of $X$, and the mappings $U_A^T, U_A^I, U_A^F : I^{A\times A} \to I$ defined by:

$$U_A^T(u) = \begin{cases} 1 & , \text{ if } u = A\times A \\ U^T(u_{X\times X} \cup \Delta_{coA}) & , \forall u \in I^{A\times A} \setminus \{A\times A\} \end{cases},$$

$$U_A^I(u) = \begin{cases} 0 & , \text{ if } u = A\times A \\ U^I(u_{X\times X} \cup \Delta_{coA}) & , \forall u \in I^{A\times A} \setminus \{A\times A\} \end{cases},$$

and $$U_A^F(u) = \begin{cases} 0 & , \text{ if } u = A\times A \\ U^F(u_{X\times X} \cup \Delta_{coA}) & , \forall u \in I^{A\times A} \setminus \{A\times A\} \end{cases}.$$

Then $(U_A^T, U_A^I, U_A^F)$ is a smooth neutrosophic preuniformity on $A$.

**Proof.** Similar to the procedure used to prove proposition (3.1.8).

Also, $(A, U_A^T, U_A^I, U_A^F)$ is a subspace of $(X, U^T, U^I, U^F)$ and $(U_A^T, U_A^I, U_A^F)$ is called the smooth neutrosophic preuniformity on $A$ induced by $(U^T, U^I, U^F)$.

**Note** that: the proposition (3.1.10) is satisfied for type II.

For smooth neutrosophic preuniform mappings in type II we can state the following definitions:

Definition.

A map $f: X \to Y$ is called weakly smooth neutrosophic preuniform with respect to the smooth neutrosophic preuniformities $(U_1^T, U_1^I, U_1^F)$ on $X$ and $(U_2^T, U_2^I, U_2^F)$ on $Y$ iff for every $v \in I^{Y\times Y}$ we have:

$$U_2^T(v) > 0 \Rightarrow U_1^T((f\otimes f)^{-1}(v)) > 0, U_2^I(v) < 1 \Rightarrow U_1^I((f\otimes f)^{-1}(v)) < 1, \text{ and}$$
$$U_2^F(v) < 1 \Rightarrow U_1^F((f\otimes f)^{-1}(v)) < 1.$$

**3.2.6. Definition.** A map $f: X \to Y$ is called smooth neutrosophic preuniform with respect to the smooth neutrosophic preuniformities $(U_1^T, U_1^I, U_1^F)$ on $X$ and $(U_2^T, U_2^I, U_2^F)$ on $Y$ iff for every $v \in I^{Y\times Y}$ we have:

$$U_1^T((f\otimes f)^{-1}(v)) \ge U_2^T(v), U_1^I((f\otimes f)^{-1}(v)) \le U_2^I(v), \text{ and } U_1^F((f\otimes f)^{-1}(v)) \le U_2^F(v).$$

**3.2.7. Definition.** A map $f: X \to Y$ is called smooth neutrosophic direct preuniform with respect to the smooth neutrosophic preuniformities $(U_1^T, U_1^I, U_1^F)$ on $X$ and $(U_2^T, U_2^I, U_2^F)$ on $Y$ iff for every $u \in I^{X\times X}$ we have:

$$U_2^T((f\otimes f)(u)) \ge U_1^T(u), U_2^I((f\otimes f)(u)) \le U_1^I(u), \text{ and } U_2^F((f\otimes f)(u)) \le U_1^F(u).$$

**Note** that the definition (3.1.16), and the propositions (3.1.17) , (3.1.18) are satisfied for type II.

## 4. Conclusion

In this paper, the concepts of smooth neutrosophic preuniform structures were introduced. In two different types we've presented the concepts of smooth neutrosophic preuniform space, smooth neutrosophic preuniform subspace, smooth neutrosophic preuniform mappings. Due to unawareness of the behaviour of the degree of indeterminacy, we've chosen for $U^I$ to act like $U^T$ in the first type, while in the second type we preferred that $U^I$ behaves like $U^F$.Therefore, the definitions given above can also be modified in several ways depending on the behaviour of $U^I$.

## References

1. K. T. Atanassov, Intuitionistic fuzzy sets: past, present and future, Proc. of the Third Conf. of the European Society for Fuzzy Logic and Technology EUSFLAT 2003, Zittau (2003) 12 – 19.
2. K. T. Atanassov, Intuitionistic fuzzy sets, Fuzzy Sets and Systems 20 (1986) 87 – 96.
3. R. Badard, Fuuzy preuniform structure and the structures they induce, Part I, J. Math.Anal. Appl.100(1984)530-548.
4. R. Badard, Fuuzy preuniform structure and the structures they induce, Part II, J.Math. Anal. Appl.100(1984)549-560.
5. R. Badard, A.A. Ramadan and A.S.Mashhour, smooth preuniform and preproximity spaces, Fuzzy Sets and Systems 59 (1993) 715 – 720.
6. R. Badard, Smooth axiomatics, 1st IFSA Congress, Palma de Mallorca, 1986.
7. C. Cornelis, K. T. Atanassov, E. E. Kerre, Intuitionistic fuzzy sets and interval-valued fuzzy sets: a critical comparison, EUSFLAT Conf. 2003: 159-163.
8. M.K. El-Gayyar, A.A. Ramadan, E.E. Kerre, Smooth pretopological spaces, J. Egypt. Math. Soc. 6 (1) (1998) 9-26.
9. B. Hutton, Uniformities on fuzzy topological spaces, J.Math.Anal. Appl.58(1977)559-571.
10. A.A. Ramadan, Y.C. Kim, M.K. El-Gayyar, On Fuzzy Uniform space , The Journal of Fuzzy Mathematics. Vol. 11, No. 2,2003.
11. A. A. Salama, S. Alblowi, Neutrosophic set and generalized neutrosophic spaces, Journal Computer Sci. Engineering 2(2012)129 132.
12. A. A. Salama, F. Smarandache, S. Alblowi, New neutrosophic crisp topological concepts, Neutrosophic Sets and Systems 2 (2014) 50 54.
13. F. Smarandache, Neutrosophy and neutrosophic logic, in: First International Conference on Neutrosophy, Neutrosophic Logic, Set, Probability, and Statistics, University of New Mexico, Gallup, NM 87301, USA.
14. F. Smarandache, A Unifying Field in Logics: Neutrosophic Logic. Neutrosophy, Neutrosophic crisp Set, Neutrosophic Probability, American Research Press, 1999.
15. F. Smarandache, Neutrosophic set, a generalization of the intuitionistic fuzzy sets, Inter. J. Pure Appl. Math., 24 (2005) 287 297.
16. L. A. Zadeh, Fuzzy sets, Inform. and Control 8 (1965) 338 – 353.

**HEWAYDA ELGHAWALBY**[1], **A. A. SALAMA**[2]

1 Faculty of Engineering, Port-Said University, Egypt. E-mail: hewayda2011@eng.psu.edu.eg
2 Faculty of Science, Port-Said University, Egypt. E-mail: drsalama44@gmail.com

# Ultra neutrosophic crisp sets and relations

## Abstract

In this paper we present a new neutrosophic crisp family generated from the three components' neutrosophic crisp sets presented by Salama [4]. The idea behind Salam's neutrosophic crisp set was to classify the elements of a universe of discourse with respect to an event "A" into three classes: one class contains those elements that are fully supportive to A, another class contains those elements that totally against A, and a third class for those elements that stand in a distance from being with or against A. Our aim here is to study the elements of the universe of discourse which their existence is beyond the three classes of the neutrosophic crisp set given by Salama. By adding more components we will get a four components' neutrosophic crisp sets called the Ultra Neutrosophic Crisp Sets. Four types of set's operations is defined and the properties of the new ultra neutrosophic crisp sets are studied. Moreover, a definition of the relation between two ultra neutrosophic crisp sets is given.

## 1 INTRODUCTION

Established by Florentin Smarandache, neutrosophy [10] was presented as the study of origin, nature, and scope of neutralities, as well as their interactions with different ideational spectra. The main idea was to consider an entity, "A" in relation to its opposite "Non-A", and to that which is neither "A" nor " Non-A ", denoted by "Neut-A". And from then on, neutrosophy became the basis of neutrosophic logic, neutrosophic probability, neutrosophic set, and neutrosophic statistics. According to this theory every idea "A" tends to be neutralized and balanced by "neutA" and "nonA" ideas - as a state of equilibrium. In a classical way "A", "neutA", and "antiA" are disjoint two by two. Nevertheless, since in many cases the borders between notions are vague and imprecise, it is possible that "A", "neutA", and "antiA" have common parts two by two, or even all three of them as well.

In [10], [11], [12] , Smarandache introduced the fundamental concepts of neutrosophic sets, that had led Salama et al.( see for instance [1], [2], [3], [4], [5], [6], [8], [9], and the references therein) to provide a mathematical treatment for the neutrosophic phenomena which already existed in our real world. Hence, neutrosophic set theory turned out to be a generalization of both the classical and fuzzy counterparts.

In [4], Salama introduced the concept of neutrosophic crisp sets as a triple structure of the form, $A_N = \langle A_1, A_2, A_3 \rangle$. The three components - $A_1, A_2$, and $A_3$ - refer to three classes of the elements of the universe $X$ with respect to an event $A$. Where $A_1$ is the class containing those elements that are fully supportive to $A$, $A_3$ for those elements that totally against $A$, and $A_2$ for those elements that stand in a distance from being with or against $A$. The three classes are subsets of $X$. Furthermore, in [7] the authors suggested three different types of such neutrosophic crisp sets depending on whither there is an overlap between the three classes or not, and whither their union covers the universe or not.

The purpose of this paper is to investigate the elements of the universe which have not been subjected to the classification; those elements belonging to the complement of the union of the three classes. Hence, a fourth component is to be added to the already existed three classes in Salama's sense.

For the purpose of this paper, we will categorize the neutrosophic crisp sets of some universe $X$ into two categories according to whither the union of the three classes covers the universe or not. The first category will contain all the neutrosophic crisp sets whose components do not cover the universe, whether they are mutually exclusive or they have some common parts in-between; two by two, or even all the three of them. While the second category will contain the remaining neutrosophic crisp sets whose components cover the whole universe.

The remaining of this paper is organized as follows: in (sec. 2) we introduce some basic definition necessary for this work. The concept of the ultra neutrosophic crisp sets is introduced in (sec. 3). Furthermore, four types of ultra neutrosophic crisp sets' operations and its properties are presented in (sec. 4) and (sec. 5). Hence, the product and relation between ultra neutrosophic crisp sets are defined in (sec. 6), (sec. 7) and (sec. 8). Finally, conclusions are drawn and future directions of research are suggested in (sec. 9).

## 2 PRELIMINARIES

### 2.1 Neutrosophic Crisp Sets

**2.1.1 Definition [4]** For any arbitrary universe $X$, a neutrosophic crisp set $A_N$ is a triple $A_N = \langle A_1, A_2, A_3 \rangle$, where $A_i \in P(X), i = 1, 2, 3$.
The three components of $A_N$ represent a classification of the elements of $X$ according to some event $A$; the subset $A_1$ contains all the elements of $X$ that are fully supportive to $A$, $A_3$ contains those elements that totally against $A$, and $A_2$ contains those elements that stand in a distance from being with or against $A$.

If we consider the event $A$ in the ordinary sense, each neutrosophic crisp set will be in the form $A_N = \langle A_1, \phi, A_1^c \rangle$, while in the fuzzy sense it will be in the form $A_N = \langle A_1, \phi, A_3 \rangle$, where $A_3 \subseteq A_1^c$. Moreover, in the intuitionistic fuzzy sense $A_N = \langle A_1, (A_1 \cup A_3)^c, A_3 \rangle$.

**2.1.2 Definition [4]** The complement of a neutrosophic crisp set is defined as:

$$coA_N = \langle coA_1, coA_2, coA_3 \rangle$$

**2.1.3 Definition [7]** A neutrosophic crisp set $A_N = \langle A_1, A_2, A_3 \rangle$ is called:

- A neutrosophic crisp set of type1, if satisfying that:
  $A_i \cap A_j = \phi$, where $i \neq j$ and $\bigcup_{i=1}^{3} A_i \subset X, \ \forall i, j = 1, 2, 3$
- A neutrosophic crisp set of type2, if satisfying that:
  $A_i \cap A_j = \phi$, where $i \neq j$ and $\bigcup_{i=1}^{3} A_i = X, \ \forall i, j = 1, 2, 3$
- A neutrosophic crisp set of type3, if satisfying that:
  $\bigcap_{i=1}^{3} A_i = \phi$, and $\bigcup_{i=1}^{3} A_i = X, \ \forall i, j = 1, 2, 3$

### 2.2 Neutrosophic Crisp Sets Operations of Type 1 [7]

For any two neutrosophic crisp sets $A_N$ and $B_N$, we have that:

$$A_N \subseteq B_N \text{ if } A_1 \subseteq B_1, A_2 \subseteq B_2, \text{ and } A_3 \supseteq B_3,$$
$$A_N = B_N \text{ if and only if } A_i = B_i \text{ for } i = 1, 2, 3.$$

Hence, we can define the following:

$$A_N \cup B_N = \langle A_1 \cup B_1, A_2 \cup B_2, A_3 \cap B_3 \rangle$$
$$A_N \cap B_N = \langle A_1 \cap B_1, A_2 \cap B_2, A_3 \cup B_3 \rangle$$

### 2.3 Neutrosophic Crisp Sets Operations of Type 2 [7]

For any two neutrosophic crisp sets $A_N$ and $B_N$, we have that:

$$A_N \subseteq B_N \text{ if } A_1 \subseteq B_1, A_2 \supseteq B_2, \text{ and } A_3 \supseteq B_3,$$
$$A_N = B_N \text{ if and only if } A_i = B_i \text{ for } i = 1, 2, 3.$$

Hence, we can define the following:

$$A_N \cup B_N = \langle A_1 \cup B_1, A_2 \cap B_2, A_3 \cap B_3 \rangle$$
$$A_N \cap B_N = \langle A_1 \cap B_1, A_2 \cup B_2, A_3 \cup B_3 \rangle$$

## 3 ULTRA NEUTROSOPHIC CRISP SETS

In this section we consider elements in $X$ which do not belong to any of the three classes of the neutrosophic crisp set defined in (2.1.1).

### 3.1 Definition

Let $X$ be any given universe, the ultra neutrosophic crisp set is defined as:

$$\breve{A} = \langle A_1, A_2, A_3, M_A \rangle, \text{ where } M_A = co(\bigcup_{i=1}^{3} A_i)$$

The family of all ultra neutrosophic crisp sets in $X$ will be denoted by $\breve{\mathfrak{U}}(X)$.

### 3.2 Definition

The complement of any ultra neutrosophic crisp set $\breve{A}$, is defined as:

$$co\breve{A} = \langle coA_1, coA_2, coA_3, coM_A \rangle$$

## 4 ULTRA NEUTROSOPHIC CRISP SETS OPERATIONS

### 4.1 Ultra Operations of Type I

For any two ultra neutrosophic crisp sets $\breve{A}$ and $\breve{B}$, we have that:

$$\breve{A} \subseteq_I \breve{B} \text{ if } A_1 \subseteq B_1, A_2 \subseteq B_2, A_3 \supseteq B_3, \text{ and } M_A \supseteq M_B,$$
$$\breve{A} = \breve{B} \text{ if and only if } A_i = B_i \text{ for} i = 1, 2, 3 \text{ and } M_A = M_B.$$

Hence, we can define the following:

$$\breve{A} \uplus_I \breve{B} = \langle A_1 \cup B_1, A_2 \cup B_2, A_3 \cap B_3, M_A \cap M_B \rangle$$
$$\breve{A} \Cap_I \breve{B} = \langle A_1 \cap B_1, A_2 \cap B_2, A_3 \cup B_3, M_A \cup M_B \rangle$$

### 4.2 Ultra Operations of Type II

For any two ultra neutrosophic crisp sets $\breve{A}$ and $\breve{B}$, we have that:

$$\breve{A} \subseteq_{II} \breve{B} \text{ if } A_1 \subseteq B_1, A_2 \supseteq B_2, A_3 \supseteq B_3, \text{ and } M_A \supseteq M_B,$$
$$\breve{A} = \breve{B} \text{ if and only if } A_i = B_i \text{ for} i = 1, 2, 3 \text{ and } M_A = M_B.$$

Hence, we can define the following:

$$\breve{A} \uplus_{II} \breve{B} = \langle A_1 \cup B_1, A_2 \cap B_2, A_3 \cap B_3, M_A \cap M_B \rangle$$
$$\breve{A} \Cap_{II} \breve{B} = \langle A_1 \cap B_1, A_2 \cup B_2, A_3 \cup B_3, M_A \cup M_B \rangle$$

### 4.3 Ultra Operations of Type III

For any two ultra neutrosophic crisp sets $\breve{A}$ and $\breve{B}$, we have that:

$$\breve{A} \subseteq_{III} \breve{B} \text{ if } A_1 \subseteq B_1, A_2 \subseteq B_2, A_3 \supseteq B_3 \text{ and } M_A \subseteq M_B,$$
$$\breve{A} = \breve{B} \text{ if and only if } A_i = B_i \text{ for} i = 1,2,3 \text{ and } M_A = M_B.$$

Hence, we can define the following:

$$\breve{A} \uplus_{III} \breve{B} = \langle A_1 \cup B_1, A_2 \cup B_2, A_3 \cap B_3, M_A \cup M_B \rangle$$
$$\breve{A} \Cap_{III} \breve{B} = \langle A_1 \cap B_1, A_2 \cap B_2, A_3 \cup B_3, M_A \cap M_B \rangle$$

### 4.4 Ultra Operations of Type IV

For any two ultra neutrosophic crisp sets $\breve{A}$ and $\breve{B}$, we have that:

$$\breve{A} \subseteq_{IV} \breve{B} \text{ if } A_1 \subseteq B_1, A_2 \supseteq B_2, A_3 \supseteq B_3 \text{ and } M_A \subseteq M_B,$$
$$\breve{A} = \breve{B} \text{ if and only if } A_i = B_i \text{ for} i = 1,2,3 \text{ and } M_A = M_B.$$

Hence, we can define the following:

$$\breve{A} \uplus_{IV} \breve{B} = \langle A_1 \cup B_1, A_2 \cap B_2, A_3 \cap B_3, M_A \cup M_B \rangle$$
$$\breve{A} \Cap_{IV} \breve{B} = \langle A_1 \cap B_1, A_2 \cup B_2, A_3 \cup B_3, M_A \cap M_B \rangle$$

### 4.5 Definition

- The difference between any two ultra neutrosophic crisp sets $\breve{A}$ and $\breve{B}$, is defined as $\breve{A} \backslash \breve{B} = \breve{A} \Cap co\breve{B}$.
- The symmetric difference between any two ultra neutrosophic crisp sets $\breve{A}$ and $\breve{B}$, is defined as $\breve{A} \bigoplus \breve{B} = \breve{A} \backslash \breve{B} \uplus \breve{B} \backslash \breve{A}$.

## 5 PROPERTIES OF ULTRA NEUTROSOPHIC CRISP SETS

Knowing that the four components $A_1, A_2, A_3$, and $M_A$ are crisp subsets of the universe $X$, we can prove that for all the Types (I, II, III, and IV) the ultra neutrosophic crisp sets operations verify the following properties:

1. Associative laws: $\breve{A} \uplus (\breve{B} \uplus \breve{C}) = (\breve{A} \uplus \breve{B}) \uplus \breve{C}$
   $\breve{A} \Cap (\breve{B} \Cap \breve{C}) = (\breve{A} \Cap \breve{B}) \Cap \breve{C}$
2. Commutative laws: $\breve{A} \uplus \breve{B} = \breve{B} \uplus \breve{A}$
   $\breve{A} \Cap \breve{B} = \breve{B} \Cap \breve{A}$
3. Distributive laws: $\breve{A} \uplus (\breve{B} \Cap \breve{C}) = (\breve{A} \uplus \breve{B}) \Cap (\breve{A} \uplus \breve{C})$
   $\breve{A} \Cap (\breve{B} \uplus \breve{C}) = (\breve{A} \Cap \breve{B}) \uplus (\breve{A} \Cap \breve{C})$
4. Idempotent laws: $\breve{A} \uplus \breve{A} = \breve{A}$
   $\breve{A} \Cap \breve{A} = \breve{A}$
5. Absorption laws: $\breve{A} \uplus (\breve{A} \Cap \breve{B}) = \breve{A}$
   $\breve{A} \Cap (\breve{A} \uplus \breve{B}) = \breve{A}$
6. Involution law: $co(co\breve{A}) = \breve{A}$
7. DeMorgan's laws: $co(\breve{A} \uplus \breve{B}) = co\breve{A} \Cap co\breve{B}$
   $co(\breve{A} \Cap \breve{B}) = co\breve{A} \uplus co\breve{B}$

**Proof**

For explanation, we will show the proof of the first associative law for type I, the proof of the first distributive law for type II, the proof of the first absorption law for type III, and the proof of the first DeMorgan's law for typeIV. using the definitions

$$
\begin{aligned}
i)\ \breve{A} \uplus_{\text{I}} (\breve{B} \uplus_{\text{I}} \breve{C}) &= \langle A_1 \cup (B_1 \cup C_1), A_2 \cup (B_2 \cup C_2), A_3 \cap (B_3 \cap C_3), M_A \cap (M_B \cap M_C)\rangle \\
&= \langle (A_1 \cup B_1) \cup C_1, (A_2 \cup B_2) \cup C_2, (A_3 \cap B_3) \cap C_3, (M_A \cap M_B) \cap M_C)\rangle \\
&= (\breve{A} \uplus_{\text{I}} \breve{B}) \uplus_{\text{I}} \breve{C}
\end{aligned}
$$

$$
\begin{aligned}
ii)\ \breve{A} \uplus_{\text{II}} (\breve{B} \Cap_{\text{II}} \breve{C}) &= \langle A_1 \cup (B_1 \cap C_1), A_2 \cap (B_2 \cup C_2), A_3 \cap (B_3 \cup C_3, M_A \cap (M_B \cup M_C)\rangle \\
&= \langle (A_1 \cup B_1) \cap (A_1 \cup C_1), (A_2 \cap B_2) \cup (A_2 \cap C_2), (A_3 \cap B_3) \cup (A_3 \cap C_3), \\
&\quad (M_A \cap M_B) \cup (M_A \cap M_C)\rangle \\
&= (\breve{A} \uplus_{\text{II}} \breve{B}) \Cap_{\text{II}} (\breve{A} \uplus_{\text{II}} \breve{C})
\end{aligned}
$$

$$
\begin{aligned}
iii)\ \breve{A} \uplus_{\text{III}} (\breve{A} \Cap_{\text{III}} \breve{B}) &= \langle A_1 \cup (A_1 \cap B_1), A_2 \cup (A_2 \cap B_2), A_3 \cap (A_3 \cup B_3), M_A \cup (M_A \cap M_B)\rangle \\
&= \langle A_1, A_2, A_3, M_A\rangle \\
&= \breve{A}
\end{aligned}
$$

$$
\begin{aligned}
iv)\ co(\breve{A} \uplus_{\text{IV}} \breve{B}) &= \langle co(A_1 \cup B_1), co(A_2 \cap B_2), co(A_3 \cap B_3), co(M_A \cup M_B)\rangle \\
&= \langle coA_1 \cap coB_1, coA_2 \cup coB_2, coA_3 \cup coB_3, coM_A \cap coM_B)\rangle \\
&= co\breve{A} \Cap_{\text{IV}} co\breve{B}
\end{aligned}
$$

**Note that:** the same procedure can be applied to prove any of the laws given in (5) for all types: I, II, II, and IV.

### 5.1 Proposition

Let $\breve{A}_i$, $i \in J$, be an arbitrary family of ultra neutrosophic crisp sets on $X$; then we have the following:

1. Type I: $\Cap_{i \in J_I} \breve{A}_i = \langle \bigcap_{i \in J} A_{i1}, \bigcap_{i \in J} A_{i2}, \bigcup_{i \in J} A_{i3}, \bigcup_{i \in J} M_{A_i}\rangle$
   $\biguplus_{i \in J} \breve{A}_i = \langle \bigcup_{i \in J} A_{i1}, \bigcup_{i \in J} A_{i2}, \bigcap_{i \in J} A_{i3}, \bigcap_{i \in J} M_{A_i}\rangle$
2. Type II: $\Cap_{i \in J} \breve{A}_i = \langle \bigcap_{i \in J} A_{i1}, \bigcup_{i \in J} A_{i2}, \bigcup_{i \in J} A_{i3}, \bigcup_{i \in J} M_{A_i}\rangle$
   $\biguplus_{i \in J} \breve{A}_i = \langle \bigcup_{i \in J} A_{i1}, \bigcap_{i \in J} A_{i2}, \bigcap_{i \in J} A_{i3}, \bigcap_{i \in J} M_{A_i}\rangle$
3. Type III: $\Cap_{i \in J} \breve{A}_i = \langle \bigcap_{i \in J} A_{i1}, \bigcap_{i \in J} A_{i2}, \bigcup_{i \in J} A_{i3}, \bigcap_{i \in J} M_{A_i}\rangle$
   $\biguplus_{i \in J} \breve{A}_i = \langle \bigcup_{i \in J} A_{i1}, \bigcup_{i \in J} A_{i2}, \bigcap_{i \in J} A_{i3}, \bigcup_{i \in J} M_{A_i}\rangle$
4. Type IV: $\Cap_{i \in J} \breve{A}_i = \langle \bigcap_{i \in J} A_{i1}, \bigcup_{i \in J} A_{i2}, \bigcup_{i \in J} A_{i3}, \bigcap_{i \in J} M_{A_i}\rangle$
   $\biguplus_{i \in J} \breve{A}_i = \langle \bigcup_{i \in J} A_{i1}, \bigcap_{i \in J} A_{i2}, \bigcap_{i \in J} A_{i3}, \bigcup_{i \in J} M_{A_i}\rangle$

## 6 THE ULTRA CARTESIAN PRODUCT OF ULTRA NEUTROSOPHIC CRISP SETS

Consider any two ultra neutrosophic crisp sets, $A$ on $X$ and $B$ on $Y$; where $\breve{A} = \langle A_1, A_2, A_3, M_A\rangle$ and $\breve{B} = \langle B_1, B_2, B_3, M_B\rangle$
The ultra cartesian product of $\breve{A}$ and $\breve{B}$ is defined as the quadruple structure:

$$\breve{A} \times \breve{B} = \langle A_1 \times B_1, A_2 \times B_2, A_3 \times B_3, M_A \times M_B\rangle$$

where each component is a subset of the cartesian product $X \times Y$;

$$
\begin{aligned}
A_i \times B_i &= \{(a_i, b_i) : a_i \in A_i \text{ and } b_i \in B_i\}, \forall i = 1, 2, 3 \text{ and} \\
M_A \times M_B &= \{(m_a, m_b) : m_a \in M_A \text{ and } m_b \in M_B\}
\end{aligned}
$$

### 6.1 Corollary

In general if $\breve{A} \neq \breve{B}$, then $\breve{A} \times \breve{B} \neq \breve{B} \times \breve{A}$

## 7 ULTRA NEUTROSOPHIC CRISP RELATIONS

An ultra neutrosophic crisp relation $\breve{R}$ from an ultra neutrosophic crisp set $\breve{A}$ to $\breve{B}$, namely $\breve{R} : \breve{A} \to \breve{B}$, is defined as a quadruple structure of the form $\breve{R} = \langle R_1, R_2, R_3, R_M \rangle$, where $R_i \subseteq A_i \times B_i, \forall i = 1, 2, 3$ and $R_M \subseteq M_A \times M_B$,
that is

$$R_i = \{(a_i, b_i) : a_i \in A_i \text{ and } b_i \in B_i\}$$
$$R_M = \{(m_a, m_b) : m_a \in M_A \text{ and } m_b \in M_B\}$$

### 7.1 Domain and Range of Ultra Neutrosophic Crisp Relations

For any ultra neutrosophic crisp relation $\breve{R} : \breve{A} \to \breve{B}$, we define the following:

- The ultra domain of $\breve{R}$, is defined as:
  $uDom(\breve{R}) = \langle dom(R_1), dom(R_2), dom(R_3), dom(R_M) \rangle$

- The ultra range of $\breve{R}$, is defined as:
  $uRng(\breve{R}) = \langle rng(R_1), rng(R_2), rng(R_3), rng(R_M) \rangle$

- The Domain of $\breve{R}$, is defined as:
  $Dom(\breve{R}) = dom(R_1) \cup dom(R_2) \cup dom(R_3) \cup dom(R_M)$

- The Range of $\breve{R}$, is defined as:
  $Rng(\breve{R}) = rng(R_1) \cup rng(R_2) \cup rng(R_3) \cup rng(R_M)$

### 7.2 Corollary

From the definitions given in 7.1, one may notice that for any ultra neutrosophic crisp relation $\breve{R} : \breve{A} \to \breve{B}$, we have:

- The domain of $\breve{R}$ is a crisp subset of $X$, namely, $Dom(\breve{R}) \subseteq X$.

- The range of $\breve{R}$ is a crisp subset of $Y$, namely, $Rng(\breve{R}) \subseteq Y$.

- The ultre domain of $\breve{R}$ is a quadruple structure whose components are crisp subsets of $X$; furthermore, $dom(R_i) \subseteq A_i, i = 1, 2, 3$ and $dom(R_M) \subseteq M_A$

- The ultre range of $\breve{R}$ is a quadruple structure whose components are crisp subsets of $Y$; furthermore, $rng(R_i) \subseteq B_i, i = 1, 2, 3$ and $rng(R_M) \subseteq M_B$

### 7.3 Definition

An ultra neutrosophic crisp inverse relation $\breve{R}^{-1}$ is an ultra neutrosophic crisp relation from an ultra neutrosophic crisp set $\breve{B}$ to $\breve{A}$, $\breve{R}^{-1} : \breve{B} \to \breve{A}$, and to be defined as a quadruple structure of the form:
$\breve{R}^{-1} = \langle R_1^{-1}, R_2^{-1}, R_3^{-1}, R_M^{-1} \rangle$, where $R_i^{-1} \subseteq B_i \times A_i, \forall i = 1, 2, 3$ and $R_M \subseteq M_B \times M_A$,
that is:

$$R_i^{-1} = \{(b_i, a_i) : (a_i, b_i) \in R_i\}$$
$$R_M^{-1} = \{(m_b, m_a) : (m_a, m_b) \in R_M\}$$

### 7.4 Corollary

For any ultra neutrosophic crisp relation $\breve{R} : \breve{A} \to \breve{B}$, we have that :

$$Dom(\breve{R}^{-1}) = Rng(\breve{R}) \quad Rng(\breve{R}^{-1}) = Dom(\breve{R})$$
$$uDom(\breve{R}^{-1}) = uRng(\breve{R}) \quad uRrng(\breve{R}^{-1}) = uDom(\breve{R})$$

## 8 COMPOSITION OF ULTRA NEUTROSOPHIC CRISP RELATIONS

Consider the three ultra neutrosophic crisp sets: $\breve{A}$ of $X$, $\breve{B}$ of $Y$ and $\breve{C}$ of $Z$; and, the two ultra neutrosophic crisp relations: $\breve{R} : \breve{A} \to \breve{B}$ and $\breve{S} : \breve{B} \to \breve{C}$; where $\breve{R} = \langle R_1, R_2, R_3, R_M \rangle$, and $\breve{S} = \langle S_1, S_2, S_3, S_M \rangle$. The composition of $\breve{R}$ and $\breve{S}$, is denoted and defined as:
$\breve{R} \odot \breve{S} = \langle R_1 \circ S_1, R_2 \circ S_2, R_3 \circ S_3, R_M \circ S_M \rangle$ such that,
$R_i \circ S_i : A_i \to C_i$, where, $R_i \circ S_i = \{(a_i, c_i) : \exists b_i \in B_i,\ (a_i, b_i) \in R_i \text{ and } (b_i, c_i) \in S_i\}$;
$R_M \circ S_M : M_A \to M_C$, where, $R_M \circ S_M = \{(m_a, m_c) : \exists m_b \in M_B,\ (m_a, m_b) \in R_M \text{ and } (m_b, m_c) \in S_M\}$

### 8.1 Corollary

For any two ultra neutrosophic crisp relations: $\breve{R} : \breve{A} \to \breve{B}$ and $\breve{S} : \breve{B} \to \breve{C}$;

$$uDom(\breve{R} \odot \breve{S}) \subseteq uDom(\breve{R})$$
$$uRng(\breve{R} \odot \breve{S}) \subseteq uRng(\breve{S})$$

### 8.2 Corollary

Consider the three ultra neutrosophic crisp relations: $\breve{R} : \breve{A} \to \breve{B}$, $\breve{S} : \breve{B} \to \breve{C}$, and $\breve{K} : \breve{C} \to \breve{D}$; we have that:

$$\breve{R} \odot (\breve{S} \odot \breve{K}) = (\breve{R} \odot \breve{S}) \odot \breve{K}$$

## 9 CONCLUSION AND FUTURE WORK

In this paper we have presented a new concept of neutrosophic crisp sets, called "The Ultra Neutrosophic Crisp Sets", as a quadrable structure. The first three components represent a classification of the universe of discourse with respect to some event; while the fourth component deals with the elements which have not been subjected to that classification. While the elements of the first and the third are considered to be well defined, there is a blurry about the behavior of the elements in both second and fourth components. Consequently, four types of set's operations were established and the properties of the new ultra neutrosophic crisp sets were studied according to different expectations about the performance of the second and the fourth components. Moreover, the definition of the relation between two ultra neutrosophic crisp sets were given. Finally, the concepts of product and composition of ultra neutrosophic crisp sets were introduced.

## References

1. Alblowi, S., Salama, A. A., and Eisa, M. New concepts of neutrosophic sets. *International Journal of Mathematics and Computer Applications Research (IJMCAR) vol. 4*, no. 1 (2014), 59 – 66.
2. Hanafy, I., Salama, A., and Mahfouz, K. Neutrosophic classical events and its probability. *International Journal of Mathematics and Computer Applications Research(IJMCAR) vol. 3*, no. 3 (2013), 171 – 178.
3. Salama, A., Khaled, O. M., and Mahfouz, K. Neutrosophic correlation and simple linear regression. *Neutrosophic Sets and Systems vol. 5* (2014), 3 – 8.

4. Salama, A. A. Neutrosophic crisp point & neutrosophic crisp ideals. *Neutrosophic Sets and Systems vol. 1*, no. 1 (2013), 50 – 54.
5. Salama, A. A., and Alblowi, S. Neutrosophic set and neutrosophic topological spaces. *ISOR J. Mathematics vol. 3*, no. 3 (2012), 31 – 35.
6. Salama, A. A., Alblowi, S. A., and Smarandache, F. Neutrosophic crisp open set and neutrosophic crisp continuity via neutrosophic crisp ideals. *I.J. Information Engineering and Electronic Business vol. 3* (2014), 1 – 8.
7. Salama, A. A., and Smarandache, F. *Neutrosophic Crisp Set Theory*. Educational Publisher, Columbus, Ohio, USA., 2015.
8. Salama, A. A., Smarandache, F., and Alblowi, S. A. The characteristic function of a neutrosophic set. *Neutrosophic Sets and Systems vol. 3* (2014), 14 – 18.
9. Salama, A. A., Smarandache, F., and Kroumov, V. Neutrosophic closed set and neutrosophic continuous functions. *Neutrosophic Sets and Systems vol. 4* (2014), 4 – 8.
10. Smarandache, F. *A Unifying Field in Logics: Neutrosophic Logic. Neutrosophy, Neutrosophic Set, Neutrosophic Probability*. American Research Press, Rehoboth, NM, 1999.
11. Smarandache, F. Neutrosophy and neutrosophic logic. In *First International Conference on Neutrosophy , Neutrosophic Logic, Set, Probability, and Statistics University of New Mexico, Gallup, NM 87301, USA* (2002).
12. Smarandache, F. Neutrosophic set, a generialization of the intuituionistics fuzzy sets. *Inter. J. Pure Appl. Math. vol. 24* (2005), 287 297.

**A.A.Salama[1], I.M.Hanafy[2], Hewayda Elghawalby[3], M.S.Dabash[4]**

[1,2,4]Department of Mathematics and Computer Sciences, Faculty of Sciences, Port Said University, Egypt. Emails:drsalama44@gmail.com, ihanafy@hotmail.com, majdedabash@yahoo.com
[3]Faculty of Engineering, port-said University, Egypt. Email: hewayda2011@eng.psu.edu.eg

# Neutrosophic Crisp Closed Region and Neutrosophic Crisp Continuous Functions

**Abstract**

In this paper, we introduce and study the concept of "neutrosophic crisp closed set "and "neutrosophic crisp continuous function. Possible application to GIS topology rules are touched upon.

## 1. Introduction

The idea of "neutrosophic crisp set" was first given by Salama and Smarandache [8]. Neutrosophic crisp operations have been investigated by Salama and Alblowi [4, 5], Salama [6], Salama and Smarandache [7, 8], Salama, and Elagamy [9], Salama et al. [10]. Neutrosophy has laid the foundation for a whole family of new mathematical theories, generalizing both their crisp and fuzzy counterparts [13]. Here we shall present the neutrosophic crisp version of these concepts. In this paper, we introduce and study the concept of "neutrosophic crisp closed set "and "neutrosophic crisp continuous function".

## 2. Terminologies

We recollect some relevant basic preliminaries, and in particular the work of Smarandache in [11,12], and Salama and Alblowi [4, 5], Salama [6], . Salama and Smarandache [7, 8], Salama, and Elagamy [9], Salama et al. [10].

**2.1 Definition:**

Let X be a non-empty fixed set. A generalized neutrosophic crisp set $(GNCS)$ A is an object having the form $A = \langle A ,A_2,A_3\rangle$, where $A_1, A_2, A_3 \subseteq X$ and $A_1 \cap A_2 \cap A_3 = \phi$.

**2.2 Definition** [8,10]

As defined in [10] a neutrosophic crisp topology (NCT) on a non-empty set $X$ is a family, $\tau$, of neutrosophic crisp subsets of $X$ satisfying the following axioms:

$(NCT_1)$ $\emptyset_N$, $X_N \in \tau$,

$(NCT_2)$ $G_1 \cap G_2 \in \tau$ for any $G_1, G_2 \in \tau$,

$(NCT_3)$ $\cup G_i \in \tau \, \forall \{G_i : i \in J\} \subseteq \tau$.

In this case the pair $(X, \tau)$ is called a neutrosophic crisp topological space (NCTS ) and the elements of $\tau$ are called neutrosophic crisp open sets, (NCOS). A neutrosophic crisp set F is said to be neutrosophic crisp closed if and only if its complement, $F^c$, is neutrosophic crisp open.

**2.3 Definition** [7]

Let $(X, \Gamma)$ be $NCTS$ and $A = \langle A_1, A_2, A_3 \rangle$ be a $NCS$ in $X$. Then the neutrosophic crisp closure of $A$ ($NCcl(A)$) and neutrosophic interior crisp ($NCint(A)$) of $A$ are defined by

$NCcl(A) = \cap \{K: K$ is an $NCCS$ in $X$ and $A \subseteq K\}$

$NCint(A) = \cup \{G: G$ is an $NCOS$ in $X$ and $G \subseteq A)$,

where $NCS$ is a neutrosophic crisp set and $NCOS$ is a neutrosophic crisp open set. It can be also shown that $NCcl(A)$ is a neutrosophic crisp closed set($NCCS$) and $NCint(A)$ is a neutrosophic crisp open set ($NCOS$) in $X$.

## 3. Neutrosophic Crisp Co-Topology

**3.1 Definition**

Let (X,T) be a neutrosophic crisp topological space, a neutrosophic crisp set A in (X,T) is said to be neutrosophic crisp closed ( NC-closed), if NCcl(A) $\subseteq$ G whenever A $\subseteq$ G and G is neutrosophic crisp open set.

**3.2 Proposition**

If A and B are neutrosophic crisp closed sets, then A$\cup$B is neutrosophic crisp closed set.

**3.3 Remark**

The intersection of two neutrosophic crisp closed (NC-closed ) sets need not be neutrosophic crisp closed set.

**3.4 Example**

Let X = {a,b,c,d,e,f,g} and that A = <{a,b},{b,c},{b,d}>, B = <{a,c},{d,c},{a,c}> are two neutrosophic crisp sets on X. Then $T = \{\phi_N, \chi_N, A, B\}$ is a neutrosophic crisp topology on X. Define the two neutrosophic crisp sets $A_1$ and $A_2$ as follows,

$A_1$ = <{b,d},{a,d,e,f,g},{a,b}>

$A_2$ = <{a,c},{a,b,e,f,g},{a,c}>

$A_1$ and $A_2$ are neutrosophic crisp closed set but $A_1 \cap A_2$ is not a neutrosophic crisp closed set.

**3.5 Proposition**

Let (X,T) be a neutrosophic crisp topological space. If B is neutrosophic crisp closed set and B $\subseteq$ A $\subseteq$ NCcl (B), then A is NC-closed.

**Definition**

(Defining NC topology by closed sets). A NC topology on a set X is given by defining " NC open set" of X. Since NC closed sets are just exactly the complement of NC open sets, it is possible to define NC topology by giving a collection of NC closed sets. Let K be a collection of NC subsets of *X* satisfying

$(NCT_1)$ $\emptyset_N$, $X_N \in$ K,

$(NCT_2)$ $G_1 \cup G_2 \in K$ for any $G_1, G_2 \in K$,

$(NCT_3)$ $\bigcap G_i \in K \, \forall \{G_i : i \in J\} \subseteq K$.

Then define *T* by: *T*:={X-C |C∈K}

Is a NC topology, i.e. it satisfyies Definition(2.2). On the other hand, if *T* is a NC topology, i.e. the collection of NC-open sets, then K:={X-U|U∈*T*}.

In this case the pair $(X, K)$ is called a neutrosophic crisp Co-topological space $(NCKS)$ and the elements of $K$ are called neutrosophic crisp closed sets, ( $NCCS$ for short).

**3.6 Proposition**

In a neutrosophic crisp topological space (X,T), T=$\Im$ (the family of all neutrosophic crisp closed sets) iff every neutrosophic crisp subset of (X,T) is a neutrosophic crisp closed set.

**Proof.**

Suppose that every neutrosophic crisp set of (X,T) is NC-closed, and let A$\in$T. Since A $\subseteq$ A and A is NC-closed, NCcl(A) $\subseteq$ A. However, we have that A $\subseteq$NCcl(A), for each set A. Hence, NCcl (A) = A. thus, A$\in\Im$. Therefore, T$\subseteq\Im$. Now, consider $B \in \Im$, then $B^c \in$T $\subseteq \Im$. Hence B$\in$T, That is, $\Im \subseteq$ T. Therefore T=$\Im$.

Conversely, suppose that A be a neutrosophic crisp set in (X,T), and B is a neutrosophic crisp open set in (X,T) such that A $\subseteq$ B. By hypothesis, B is NC-closed. By definition of any neutrosophic crisp closure set, we have that NCcl(A) $\subseteq$ B. Therefore A is NC-closed set.

**3.7 Proposition**

Let (X,*T*) be a neutrosophic crisp topological space. A neutrosophic crisp set A is neutrosophic crisp open iff B $\subset$ NCInt (A), whenever B is neutrosophic crisp closed and B $\subset$ A.

**Proof**

Let A a neutrosophic crisp open set and B be a NC-closed, such that B $\subset$ A. Now, $B \subset A \Rightarrow A^c \subset B^c$ and $A^c$ is a neutrosophic crisp closed set $\Rightarrow NCcl(A^c) \subset B^c$. That is, B = $(B^c)^c \subset (NCcl(A^c))^c$. But $(NCcl(A^c))^c$ = NCint (A). Thus, B $\subset$ NCint (A). Conversely, suppose that A be a neutrosophic crisp set, such that $B \subset NC\,\mathrm{int}(A)$ whenever B is neutrosophic crisp closed and B ⊂ A. Let $A^c \subset B \Rightarrow B^c \subset A$. Hence by assumption $B^c \subset NC\,\mathrm{int}(A)$. that is, $(NC\,\mathrm{int}(A))^c \subset B$. But $(NC\,\mathrm{int}(A))^c = NCcl(A^c)$.

Hence $NCcl(A^c) \subset B$. That is $A^c$ is neutrosophic crisp closed set. Therefore, A is neutrosophic crisp open set

**3.8 Proposition**

If (A) ⊆ B ⊆ NCcl(A) and if A is neutrosophic crisp closed set then B is also neutrosophic crisp closed set.

## 4. Neutrosophic Crisp Continuous Functions

*4.1 Definition*

(c) If $A = \langle A_1, A_2, A_3 \rangle$ is a NCS in X, then the neutrosophic crisp image of A under $f$, denoted by $f(A)$, is the a NCS in Y defined by $f(A) = \langle f(A_1), f(A_2), f(A_3) \rangle$.

(d) If $f$ is a bijective map then $f^{-1}: Y \longrightarrow X$ is a map defined such that: for any NCS $B = \langle B_1, B_2, B_3 \rangle$ in Y, the neutrosophic crisp preimage of B, denoted by $f^{-1}(B)$, is a NCS in X defined by $f^{-1}(B) = \langle f^{-1}(B_1), f^{-1}(B_2), f^{-1}(B_3) \rangle$.

Here we introduce the properties of images and preimages some of which we shall frequently use in the following sections .

*4.2 Corollary*

Consider, the two families of neutrosophic crisp sets;

A= $\{A_i: i \in I, A_i \subseteq X\}$ and B = $\{B_j: j \in J, B_j \subseteq Y\}$; and let $f$ be a function such that $f: X \rightarrow Y$.

(a) $A_1 \subseteq A_2 \Leftrightarrow f(A_1) \subseteq f(A_2)$, $B_1 \subseteq B_2 \Leftrightarrow f^{-1}(B_1) \subseteq f^{-1}(B_2)$,

(b) $A \subseteq f^{-1}(f(A))$ and if $f$ is injective, then $A = f^{-1}(f(A))$ .

(c) $f(f^{-1}(B)) \subseteq B$ and if $f$ is surjective, then $f(f^{-1}(B)) = B$, .

(d) $f^{-1}(\cup B_i) = \cup f^{-1}(B_i)$, $f^{-1}(\cap B_i) = \cap f^{-1}(B_i)$,

(e) $f(\cup A_i) = \cup f(A_i)$; $f(\cap A_i) \subseteq \cap f(A_i)$; and if $f$ is injective, then $f(\cap A_i) = \cap f(A_i)$;

(f) $f^{-1}(Y_N) = X_N$ $f^{-1}(\phi_N) = \phi_N$.

(g) $f(\phi_N) = \phi_N$, $f(X_N) = Y_N$ if $f$ is subjective.

**Proof**

Obvious.

*4.3 Proposition*

Consider the function $f : X \to Y$, then $f$ is said to be neutrosophic crisp continuous iff the preimage of each neutrosophic crisp closed set in Y is a neutrosophic crisp closed set in X.

4.4 Proposition

Consider the function $f : X \to Y$, then $f$ is said to be neutrosophic crisp continuous iff the image of each neutrosophic crisp closed set in X is a neutrosophic crisp closed set in Y.

*4.5 Proposition*

The following are equivalent to each other:

(a) $f : (X, \Gamma_1) \to (Y, \Gamma_2)$ is neutrosophic crisp continuous .

(b) $f^{-1}(NCInt(B)) \subseteq NCInt(f^{-1}(B))$ for each *NCCS* B in Y.

(c) $NCCl(f^{-1}(B)) \subseteq f^{-1}(NCCl(B))$ for each *N NCCS* B in Y.

*4.6 Example*

Let $(Y, \Gamma_2)$ be a *NCTS* and $f : X \to Y$ be a function. In this case $\Gamma_1 = \{f^{-1}(H) : H \in \Gamma_2\}$is a *NCT* on X. Indeed, it is the coarsest *NCT* on X which makes the function $f : X \to Y$ neutrosophic crisp continuous. One may call it the initial neutrosophic crisp topology with respect to $f$.

**4.7 Definition**

Let (X,T) and (Y,S) be two neutrosophic crisp topological space, then

(a) A bijective map $f$:(X,T)→ (Y,S) is called neutrosophic crisp irresolute if the inverse image of every neutrosophic crisp closed set in (Y,S) is neutrosophic crisp closed in (X,T). Equivalently if the inverse image of every neutrosophic crisp open set in (Y,S) is neutrosophic crisp open in (X,T).

(b) A map $f$:(X,T)→ (Y,S) is said to be strongly neutrosophic crisp continuous if $f$(A) is both neutrosophic crisp open and neutrosophic crisp closed in (Y,S) for each neutrosophic crisp set A in (X,T).

(c) A map $f$ : (X,T) → (Y,S) is said to be perfectly neutrosophic crisp continuous if $f^{-1}$(B) is both neutrosophic crisp open and neutrosophic crisp closed in (X,T) for each neutrosophic crisp open set B in (Y,S).

**4.8 Proposition**

Let (X,T) and (Y,S) be any two neutrosophic crisp topological spaces. Let $f$ : (X,T) → (Y,S) be neutrosophic crisp continuous. Then for every neutrosophic crisp set A in X, $f$(NCcl(A)) ⊆ NCcl($f$(A)).

**4.9 Proposition**

Let (X,T) and (Y,S) be any two neutrosophic crisp topological spaces. Let $f$ : (X,T) → (Y,S) be neutrosophic crisp continuous. Then for every neutrosophic crisp set A in Y, NCcl($f^{-1}$(A)) ⊆ $f^{-1}$(NCcl(A)).

**Definition**

Let $(X,\Gamma_1)$ and $(Y,\Gamma_2)$ be two NCTSs and let $f : X \to Y$ be a function. Then $f$ is said to be open iff the neutrosophic crisp image of each NCS in $\Gamma_1$ is a NCS in $\Gamma_2$.

**Definition**

Consider the two neutrosophic crisp co-topologios $(X,K_1)$, $(Y,K_2)$ and function $f : X \to Y$ the function $f$ is said to be neutrosophic crisp closed iff $f(A) \in K_2$, $\forall A \in K_1$.

Or equivalently, $f^{-1}(B) \in K_1$, $\forall B \in K_2$.

**Definition**

Consider the two neutrosophic crisp topologios $(X,T_1)$, $(Y,T_2)$ and function $f : X \to Y$ the function $f$ is said to be neutrosophic crisp closed iff $f(A) \in T_2$, $\forall A \in T_1$.

Or equivalently, $f^{-1}(B) \in T_1$, $\forall B \in T_2$.

**4.10 Proposition**

Let (X,T) and (Y,S) be any two neutrosophic crisp topological spaces. If A is a neutrosophic crisp closed set in (X,T) and if $f$ : (X,T) → (Y,S) is neutrosophic crisp continuous and neutrosophic crisp closed then $f$(A) is neutrosophic crisp closed in (Y,S).

**Proof.**

Let G be a neutrosophic crisp- open in (Y,S). If $f(A) \subseteq G$, then $A \subseteq f^{-1}(G)$ in (X,T). Since A is neutrosophic crisp closed and $f^{-1}(G)$ is neutrosophic crisp open in (X,T), $NCcl(A) \subset f^{-1}(G)$, (i.e) $f(NCcl(A) \subset G$. Now by assumption, $f(NCcl(A))$ is neutrosophic crisp closed and $NCcl(f(A)) \subseteq Ncl(f(NCcl(A))) = f(NCcl(A)) \subset G$. Hence, $f$(A) is NC-closed.

**4.11 Proposition**

If the function $f : X \to Y$ is neutrosophic crisp continuous, then it is neutrosophic crsip closed. Whereas, the converse need not be true, as shown in Example 4.12.

**4.12 Example**

Let X ={a,b,c,d,e,f,g} and Y ={a,b,c}. Define neutrosophic crisp sets A and B as follows:

A = <{d,a} ,{f,g}, {c,b}>

B = <{f,a}, {e,f},{d,c} >

Then the family T = $\{\phi_N, X_N, A\}$ is a neutrosophic crisp topology on X and S = $\{\phi_N, X_N, B\}$ is a neutrosophic crisp topology on Y. Thus (X,T) and (Y,S) are neutrosophic crisp topological spaces. Define

$f$ : (X,T) → (Y,S) as $f(a) = b$, $f(b) = a$, $f(c) = c$. Clearly $f$ is NC-closed continuous. Now $f$ is not neutrosophic crisp continuous, since $f^{-1}(B) \notin T$ for $B \in S$.

**Definition**

If the function $f : X \to Y$ *is neutrosophic crisp continuous, then it is* neutrosophic crisp open. *Whereas, the converse need not* be true, as shown in Example 4.13.

**4.13 Example**

Let X = {a,b,c,d,e,f,g}. Define the neutrosophic crisp sets A and B as follows.

A = <{f,g}, {d,a} , {c,b}>

B = <{f,a}, {d,c} , {e,f}> and

C = <{b,d}, {c,d} , {d,a}>

T = { $\phi_N$, $X_N$, A ,B} and

S = { $\phi_N$, $X_N$, C}

are neutrosophic crisp topologies on X. Thus (X,T) and (X,S) are neutrosophic crisp topological spaces. Define $f$: (X,T) → (X,S) as follows $f$(a) = b,

$f$(b) = b, $f$(c) = c. Clearly $f$ is NC-continuous. Since

D = <{d,a}, {c,f} , {g,e}>

is neutrosophic crisp open in (X,S), $f^{-1}$(D) is not neutrosophic crisp open in (X,T).

**4.14 Proposition**

Let (X,T) and (Y,S) be any two neutrosophic crisp topological space. If $f$ : (X,T) → (Y,S) is strongly NC-continuous then $f$ is neutrosophic crisp continuous.

The converse of Proposition 4.16 is not true. See Example 4.17

**4.15 Example**

Let X ={a,b,c}. Define the neutrosophic crisp sets A and B as follows.

A = <{d,e}, {f,h} , {c,a}>

B = <{g,e}, {q,z} , {b,a}>and

C = <{a,c}, {f,a} , {h,d}>

T = { $\phi_N$, $X_N$, A ,B} and S = { $\phi_N$, $X_N$, C} are neutrosophic crisp topologies on X. Thus (X,T) and (X,S) are neutrosophic crisp topological spaces. Also define $f$ :(X,T)→ (X,S) as follows $f$(a) = a, $f$(b) = c, $f$(c) = b. Clearly $f$ is neutrosophic crisp continuous. But $f$ is not strongly NC-continuous. Since D = <{c,f},{e,c},{b,g,d}>

Is a neutrosophic crisp open set in (X,S), $f^{-1}$(D) is not neutrosophic crisp open in (X,T).

**4.16 Proposition**

Let (X,T) and (Y,S) be any two neutrosophic crisp topological spaces. If $f$: (X,T) → (Y,S) is perfectly NC-continuous then $f$ is strongly NC-continuous.

The converse of Proposition 4.16 is not true. See Example 4.17

**4.17 Example**

Let X = {a,b,c,d,e,f,g}. Define the neutrosophic crisp sets A and B as follows.

A = <{f,g}, {d,a} , {c,b}>, B = <{f,a}, {d,c} , {e,f}> and

C = <{b,b}, {c,d} , {d,a}>

T = {$\phi_N$, $X_N$, A ,B} and S = {$\phi_N$ ,$X_N$, C} are neutrosophic crisp topologies space on X. Thus (X,T) and (X,S) are neutrosophic crisp topological spaces. Also define $f$ : (X,T) → (X,S) as follows $f$(a) = a, $f$(b) = $f$(c) = b. Clearly $f$ is strongly NC-continuous. But $f$ is not perfectly NC continuous. Since D = <{d,a},{b,b}, {c,d}> is aneutrosophic crisp open set in (X,S), $f^{-1}$(D) is neutrosophic crisp open and not neutrosophic crisp closed in (X,T).

**4.18 Proposition**

Let (X,T) and (Y,S) be any neutrosophic crisp topological spaces. If $f$: (X,T) → (Y,S) is strongly neutrosophic crisp continuous then $f$ is strongly NC-continuous.

The converse of proposition 4.20 is not true. See Example 4.21

**4.19 Example**

Let X = {a,b,c,d,e,f,g} and define the neutrosophic crisp sets A and B as follows.

A = <{sb,b}, {d,a} , {c,d}>

B = <{e,f }, {d,c} , {f,a}> and

C = <{f,g}, {c,b} , {d,a}>

T = {$\phi_N$, $X_N$, A ,B} and S = {$\phi_N$, $X_N$, C} are neutrosophic crisp topologies on X. Thus (X,T) and (X,S) are neutrosophic crisp topological spaces. Also define $f$ : (X,T) → (X,S) as follows: $f$(a) = a, $f$(b) = $f$(c) = b. Clearly $f$ is strongly NC-continuous. But $f$ is not strongly neutrosophic crisp continuous. Since

D = <{d,a}, {f,g} , {c,b}>

be a neutrosophic crisp set in (X,S), $f^{-1}$(D) is neutrosophic crisp open and not neutrosophic crisp closed in (X,T).

**4.20 Proposition**

Let (X,T),(Y,S) and (Z,R) be any three neutrosophic crisp topological spaces. Suppose $f$ : (X,T) → (Y,S), g : (Y,S) → (Z,R) be maps. Assume $f$ is neutrosophic crisp irresolute and g is NC-continuous then g o $f$ is NC-continuous.

**4.21 Proposition**

Let (X,T), (Y,S) and (Z,R) be any three neutrosophic crisp topological spaces. Let $f$ : (X,T) → (Y,S), g : (Y,S) → (Z,R) be map, such that $f$ is strongly NC-continuous and g is NC-continuous. Then the composition g o $f$ is neutrosophic crisp continuous.

**4.22 Definition**

A neutrosophic crisp topological space (X,T) is said to be neutrosophic crisp $T_{1/2}$ if every neutrosophic crisp closed set in (X,T) is neutrosophic crisp closed in (X,T).

**4.23 Proposition**

Let (X,T),(Y,S) and (Z,R) be any neutrosophic crisp topological spaces. Let $f$ : (X,T) → (Y,S) and g : (Y,S) → (Z,R) be mapping and (Y,S) be neutrosophic crisp $T_{1/2}$ if $f$ and g are NC-continuous then the composition g o $f$ is NC-continuous.

The proposition 4.11 is not valid if (Y,S) is not neutrosophic crisp $T_{1/2}$.

**4.24 Example**

Let X = {a,b,c,d,e,f,g}. Define the neutrosophic crisp sets A,B and C as follows.

A = <{d,c},{d,a} , {c,b}>

B = <{f,g},{b,b},{e,f}> and

C = < {f,a}, {c,d},{d,a}>

Then the family T={$\phi_N$, $X_N$, A}, S={$\phi_N$, $X_N$, B} and R={$\phi_N$,$X_N$,C} are neutrosophic crisp topologies on X. Thus (X,T),(X,S) and (X,R) are neutrosophic crisp topological spaces. Also define $f$ : (X,T) → (X,S) as $f$(a) = b, $f$(b) = a, $f$(c) = c and g : (X,S) → (X,R) as g(a) = b, g(b) = c, g(c) = b. Clearly $f$ and g are NC-continuous function. But g o $f$ is not NC-continuous. For $C^c$ is neutrosophic crisp closed in (X,R). $f^{-1}$($g^{-1}$ $C^c$ ) is not NC closed in (X,T).

G o $f$ is not NC-continuous.

## 5. Conclusion

In this paper, we presented a generalization of the neutrosophic crisp topological space. The basic definitions of neutrosophic crisp closed set "and "neutrosophic crisp continuous function. with some of their characterizations were deduced. Furthermore, we constructed a neutrosophic crisp open and closed functions, with a study of a number its properties.

## References

1. S.A. Alblowi, A.A. Salama and Mohmed Eisa, New Concepts of Neutrosophic Sets, International Journal of Mathematics and Computer Applications Research (IJMCAR), Vol. 3, Issue 3, Oct, 95-102, (2013).
2. A. Hanafy, A.A. Salama and K. Mahfouz, Correlation of neutrosophic Data, International Refereed Journal of Engineering and Science (IRJES), Vol. (1), Issue 2, PP.39-33. (2012).
3. I.M. Hanafy, A.A. Salama and K.M. Mahfouz," Neutrosophic Classical Events and Its Probability" International Journal of Mathematics and Computer Applications Research (IJMCAR) Vol. (3), Issue 1, pp171-178, Mar (2013).
4. A.A. Salama and S.A. Alblowi, "Generalized Neutrosophic Set and Generalized Neutrosophic Topological Spaces,"Journal Computer Sci. Engineering, Vol. (2) No. (7) pp 129-132(2012) .
5. A.A. Salama and S.A. Alblowi, Neutrosophic Set and Neutrosophic Topological Spaces, ISORJ. Mathematics, Vol. (3), Issue (3), pp-31-35(2012).
6. A. A. Salama, "Neutrosophic Crisp Points & Neutrosophic Crisp Ideals", Neutrosophic Sets and Systems, Vol.1, No. 1, pp. 50-54(2013).
7. A. A. Salama and F. Smarandache, "Filters via Neutrosophic Crisp Sets", Neutrosophic Sets and Systems, Vol.1, No. 1, pp. 34-38 (2013).
8. A.A. Salama and F. Smarandache, Neutrosophic crisp set theory, Educational Publisher, Columbus,USA, (2015).
9. A.A. Salama, and H. Elagamy, "Neutrosophic Filters" International Journal of Computer Science Engineering and Information Technology Reseearch (IJCSEITR), Vol.3, Issue (1), pp 307-312, Mar (2013).
10. A. A. Salama, F. Smarandache and Valeri Kroumov "Neutrosophic Crisp Sets & Neutrosophic Crisp Topological Spaces" Bulletin of the Research Institute of Technology (Okayama University of Science, Japan), in January-February (2014).

11. F. Smarandache, Neutrosophy and Neutrosophic Logic, First International Conference on Neutrosophy , Neutrosophic Logic , Set, Probability, and Statistics University of New Mexico, Gallup, NM 87301, USA(2002).
12. F. Smarandache. A Unifying Field in Logics: Neutrosophic Logic. Neutrosophy, Neutrosophic Set, Neutrosophic Probability. American Research Press, Rehoboth, NM, (1999).
13. L.A. Zadeh, Fuzzy Sets, Inform and Control 8, 338-353, (1965).

**FLORENTIN SMARANDACHE**[1], **HUDA E. KHALID**[2], **AHMED K. ESSA**[3]

[1]University of New Mexico, 705 Gurley Ave. Gallup, NM 87301, USA. E-mail: smarand@unm.edu
[2]University of Telafer, Mathematics Department, College of Basic Education, Telafer, Mosul, Iraq. E-mail: hodaesmail@yahoo.com
[3] University of Telafer, College of Basic Education Telafer, Mosul, Iraq. E-mail: ahmed.ahhu@gmail.com

# A New Order Relation on the Set of Neutrosophic Truth Values

## Abstract

In this article, we discuss all possible cases to construct an atom of matter, antimatter, or unmatter, and also the cases of contradiction (i.e. impossible case).

## 1. Introduction

Anti-particle in physics means a particle which has one or more opposite properties to its "original particle kind". If one property of a particle has the opposite sign to its original state, this particle is anti-particle, and it annihilates with its original particle.

The anti-particles can be electrically charged, color or fragrance (for quarks). Meeting each other, a particle and its anti-particle annihilate into gamma-quanta.

This formulation may be mistaken with the neutrosophic <antiA>, which is strong opposite to the original particle kind. The <antiA> state is the ultimate case of anti-particles [6].

In [7], F. Smarandache discusses the refinement of neutrosophic logic. Hence, <A>, <neutA> and <antiA> can be split into: <$A_1$>, <$A_2$>, ...; <neut$A_1$>, <neut$A_2$>, ...; <anti$A_1$>, <anti$A_2$>, ...; therefore, more types of matter, more types of unmatter, and more types of antimatter.

One may refer to <A>, <neutA>, <anti-A> as "matter", "unmatter" and "anti-matter".

Following this way, in analogy to anti-matter as the ultimate case of anti-particles in physics, the unmatter can be extended to "strong unmatter", where all properties of a substance or a field are unmatter, and to "regular unmatter", where just one of the properties of it satisfies the unmatter.

## 2. Objective

The aim is to check whether the indeterminacy component $I$ can be split to sub-indeterminacies $I_1, I_2, I_3$, and then justify that the below are all different:

$$I_1 \cap I_2 \cap I_3,\ I_1 \cap I_3 \cap I_2,\ I_2 \cap I_3 \cap I_1,\ I_2 \cap I_1 \cap I_3,\ I_3 \cap I_1 \cap I_2,\ I_3 \cap I_2 \cap I_1. \qquad (1)$$

## 3. Cases

Let $e$, $e^+$, $P$, $antiP$, $N$, $antiN$ be electrons, anti-electrons, protons, anti-protons, neutrons, anti-neutrons respectively, also $\cup$ means union/OR, while $\cap$ means intersection/AND, and suppose:

$$I = (e \cup e^+) \cap (P \cup antiP) \cap (N \cup antiN) \qquad (2)$$

The statement (2) shows indeterminacy, since one cannot decide the result of the interaction if it will produce any of the following cases:

1. $(e \cup e^{+}) \cap (\mathrm{P} \cup \mathrm{antiP}) \cap (\mathrm{N} \cup \mathrm{antiN}) \rightarrow e \cap \mathrm{P} \cap \mathrm{antiN}$,
   which is *unmatter* type (a), see reference [2];
2. $(e \cup e^{+}) \cap (\mathrm{N} \cup \mathrm{antiN}) \cap (\mathrm{P} \cup \mathrm{antiP}) \rightarrow e^{+} \cap \mathrm{N} \cap \mathrm{antiP}$,
   which is *unmatter* type (b), see reference [2];
3. $(\mathrm{P} \cup \mathrm{antiP}) \cap (\mathrm{N} \cup \mathrm{antiN}) \cap (e \cup e^{+}) \rightarrow \mathrm{P} \cap \mathrm{N} \cap e^{+} = contradiction$;
4. $(\mathrm{P} \cup \mathrm{antiP}) \cap (e \cup e^{+}) \cap (\mathrm{N} \cup \mathrm{anti\,N}) \rightarrow \mathrm{antiP} \cap \mathrm{e} \cap anti\mathrm{N} = contradiction$;
5. $(\mathrm{N} \cup \mathrm{antiN}) \cap (e \cup e^{+}) \cap (\mathrm{P} \cup \mathrm{antiP}) \rightarrow \mathrm{N} \cap \mathrm{e} \cap \mathrm{P}$,
   which is a *matter*;
6. $(\mathrm{N} \cup \mathrm{antiN}) \cap (\mathrm{P} \cup \mathrm{antiP}) \cap (e \cup e^{+}) \rightarrow \mathrm{antiN} \cap anti\mathrm{P} \cap e^{+}$,
   which is *antimatter*.

## 4. Comment

It is obvious that all above six cases are not equal in pairs; suppose:

$e \cup e^{+} = I_1 = uncertainty$,

$\mathrm{P} \cup \mathrm{antiP} = I_2 = uncertainty$,

$\mathrm{N} \cup \mathrm{antiN} = I_3 = uncertainty$.

Consequently, the statement (2) can be rewritten as:

$I = I_1 \cap I_2 \cap I_3$

but we cannot get the equality for any pairs in eq. (1).

## 5. Remark

This example is a response to the article [4], where Florentin Smarandache stated that "for each application we might have some different order relations on the set of neutrosophic truth values; (…) one can get one such order relation workable for all problems", and also to a commentary in [5], that "It would be very useful to define suitable order relations on the set of neutrosophic truth values".

## References

1. F. Smarandache: A new form of matter — unmatter, formed by particles and anti-particles. CERN CDS, EXT-2004-182, 2004.
2. F. Smarandache: Verifying Unmatter by Experiments, More Types of Unmatter, and a Quantum Chromodynamics Formula. In: "Progress in Physics", Vol. 2, July 2005, pp. 113-116.
3. F. Smarandache: (T, I, F)-Neutrosophic Structures, In: "Neutrosophic Sets and Systems", Vol. 8, 2015, pp. 3-10.
4. F. Smarandache: Neutrosophic Logic as a Theory of Everything in Logics. http://fs.gallup.unm.edu/NLasTheoryOfEverything.pdf.
5. U. Rivieccio: Neutrosophic logics: Prospects and problems. In: "Fuzzy Sets and Systems", Vol. 159, Issue 14, 2008, pp. 1860-1868.
6. Dmitri Rabounski, F. Smarandache, Larissa Borisova: Neutrosophic Methods in General Relativity. Hexis: Phoenix, Arizona, USA, 2005, 78 p.
7. F. Smarandache: Symbolic Neutrosophic Theory. EuropaNova, Brussels, Belgium, 2015, 194 p.

FU YUHUA

CNOOC Research Institute, Beijing, 100028, China. E-mail: fuyh1945@sina.com

# Expanding Comparative Literature into Comparative Sciences Clusters with Neutrosophy and Quad-stage Method

## Abstract

By using Neutrosophy and Quad-stage Method, the expansions of comparative literature include: comparative social sciences clusters, comparative natural sciences clusters, comparative interdisciplinary sciences clusters, and so on. Among them, comparative social sciences clusters include: comparative literature, comparative history, comparative philosophy, and so on; comparative natural sciences clusters include: comparative mathematics, comparative physics, comparative chemistry, comparative medicine, comparative biology, and so on. In addition, comparative literature itself can also be expanded. Under the two main categories of research and practice, comparative literature can be expanded into: comparative literature research, comparative literature practice (including comparative essay, comparative fiction, comparative poetry, comparative drama, and so on), comparative literature research and practice, and so on. This paper discusses the applications of comparative method in comparative sciences clusters and their various branches. Point out that in the existing fields of social sciences and natural sciences, many sprouts of comparative sciences clusters can be found, but a wide range of the achievements of comparative sciences clusters, still are the virgin lands to be developed.

## 1. Introduction

Comparative literature is the literary branch running comparative study (research) about the relationship between two or more kinds of literatures. It consists of influence study, parallel study, interdisciplinary study, and so on.

At present, the research method of comparative literature has expanded into other areas, and establish many disciplines such as comparative sociology, comparative jurisprudence, and so on. But the expansion is not enough. In this paper, we try to expand comparative literature into comparative sciences clusters (including comparative social sciences clusters, comparative natural sciences clusters, comparative interdisciplinary sciences clusters, and so on).

## 2. Basic Contents of Neutrosophy and Basic Contents of Quad-stage

Neutrosophy is proposed by Prof. Florentin Smarandache in 1995. Neutrosophy is a new branch of philosophy that studies the origin, nature, and scope of neutralities, as well as their interactions with different ideational spectra.

This theory considers every notion or idea <A> together with its opposite or negation <Anti-A> and the spectrum of "neutralities" <Neut-A> (i.e. notions or ideas located between the two extremes, supporting neither <A> nor <Anti-A>). The <Neut-A> and <Anti-A> ideas together are referred to as <Non-A>.

Neutrosophy is the base of neutrosophic logic, neutrosophic set, neutrosophic probability and statistics used in engineering applications (especially for software and information fusion), medicine, military, cybernetics, and physics.

Neutrosophic Logic is a general framework for unification of many existing logics, such as fuzzy logic (especially intuitionistic fuzzy logic), paraconsistent logic, intuitionistic logic, etc. The main idea of NL is to characterize each logical statement in a 3D Neutrosophic Space, where each dimension of the space represents respectively the truth (T), the falsehood (F), and the indeterminacy (I) of the statement under consideration, where T, I, F are standard or non-standard real subsets of ]-0, 1+[ without necessarily connection between them.

More information about Neutrosophy can be found in references [1, 2].

Quad-stage (Four stages) is presented in reference [3], it is the expansion of Hegel's triad-stage (triad thesis, antithesis, synthesis of development). The four stages are "general theses", "general antitheses", "the most important and the most complicated universal relations", and "general syntheses". They can be stated as follows.

The first stage, for the beginning of development (thesis), the thesis should be widely, deeply, carefully and repeatedly contacted, explored, analyzed, perfected and so on; this is the stage of general theses. It should be noted that, here the thesis will be evolved into two or three, even more theses step by step. In addition, if in other stage we find that the first stage's work is not yet completed, then we may come back to do some additional work for the first stage.

The second stage, for the appearance of opposite (antithesis), the antithesis should be also widely, deeply, carefully and repeatedly contacted, explored, analyzed, perfected and so on; this is the stage of general antitheses. It should be also noted that, here the antithesis will be evolved into two or three, even more antitheses step by step.

The third stage is the one that the most important and the most complicated universal relations, namely the seedtime inherited from the past and carried on for the future. Its purpose is to establish the universal relations in the widest scope. This widest scope contains all the regions related and non-related to the "general theses", "general antitheses", and the like. This stage's foundational works are to contact, grasp, discover, dig, and even create the opportunities, pieces of information, and so on as many as possible. The degree of the universal relations may be different, theoretically its upper limit is to connect all the existences, pieces of information and so on related to matters, spirits and so on in the universe; for the cases such as to create science fiction, even may connect all the existences, pieces of information and so on in the virtual world. Obviously, this stage provides all possibilities to fully use the complete achievements of nature and society, as well as all the humanity's wisdoms in the past, present and future. Therefore this stage is shortened as "universal relations" (for other stages, the universal relations are also existed, but their importance and complexity cannot be compared with the ones in this stage).

The fourth stage, to carry on the unification and synthesis regarding various opposites and the suitable pieces of information, factors, and so on; and reach one or more results which are the best or agreed with some conditions; this is the stage of "general syntheses". The results of this stage are called "synthesized second generation theses", all or partial of them may become the beginning of the next quad-stage.

For realizing the innovations in the areas such as science and technology, literature and art, and the like, it is a very useful tool to combine neutrosophy with quad-stage method. For example, in reference [4], expanding Newton mechanics with neutrosophy and quad-stage method, and establishing New Newton Mechanics taking law of conservation of energy as unique source law; in reference [5], negating four color theorem with neutrosophy and quad-stage method, and "the two color theorem" and "the five color theorem" are derived to replace "the four color theorem"; in reference [6], expanding Hegelian triad thesis, antithesis, synthesis with Neutrosophy and Quad-stage Method; in reference [7], interpretating and expanding Laozi's governing a large country is like cooking a small fish with Neutrosophy and Quad-stage Method; in reference [8], interpretating and expanding the meaning of "Yi" with Neutrosophy and Quad-stage Method; in reference [9], Creating Generalized and Hybrid Set and Library with Neutrosophy and Quad-stage Method.

Applying Neutrosophy and Quad-stage Method, will significantly help us to consider all possible situations. Therefore, Neutrosophy and Quad-stage Method can play a very important role to expand comparative literature.

## 3. Expanding Comparative Literature with Neutrosophy and Quad-stage Method

The process of expanding comparative literature can be divided into four stages.

The first stage (stage of "general theses"), for the beginning of development, the thesis (namely "comparative literature") should be widely, deeply, carefully and repeatedly contacted, explored, analyzed, perfected and so on.

Currently, "comparative literature" has become a complex subject. Its research achievements absorb the research results of traditional world literature, as well as a variety of other areas even including natural science research; in fact, the inherent discipline bounds have been broken, and beyond the limitations of region and time, put the Asian-African literature, European-American literature, and so on, as well as classical literature, modern literature, and so on, into one or more overall structures or frames.

For example, in the research (study) of comparative literature, the literature can be compared with social sciences (philosophy, psychology, linguistics, history, sociology, anthropology, and so on), and the natural sciences (mathematical statistics, computer technology, system theory, information theory, and so on), as well as other artistic disciplines (painting, sculpture, architecture, music, film, and so on).

Of course, we should also see that different scholars may have different viewpoints and interpretations for "comparative literature" and the related problems, and the different opinions and arguments will be endless from generation to generation.

In the second stage (the stage of "general antitheses"), the opposites (antitheses) should be discussed carefully. Obviously, there are more than one opposites (antitheses) of comparative literature here.

For example, according to the viewpoint of Neutrosophy, if "comparative literature" is considered as the concept <A>, the opposite <Anti-A> may be: "non-comparative literature" (such as comparative sociology, comparative jurisprudence, and so on); while the neutral (middle state) fields <Neut-A> including: "undetermined comparative literature" (neither "comparative literature", nor "non-comparative literature"; or, sometimes it is "comparative literature", and sometimes it is "non-comparative literature; and so on".

In the third stage, considering the most important and the most complicated universal relations to link with "comparative literature". The purpose of this provision stage is to establish the universal relations in the widest scope.

For "comparative literature", different people will have different research methods and findings; even if for the same person, at different times and in different situations, he or she may also apply different research methods and reach different research results. Therefore, pursuing the unique right research method and research result do not seem to make sense. So the advisable method of work is to collect all people's research methods and research results from ancient times to modern times, and plus own research methods and research results, to form the so-called "full research methods and research results", and to store up them as Think Tank; while once we need to apply them, then immediately the one or several best research methods and research results can be elected, or according to the information in Think Tank and the reality to obtain one or several best programmes temporarily, thus we can be invincible.

Now we list some specific research methods and results.

The first school of comparative literature in the world is France school. Characterized by respecting the facts, and emphasizing the textual studies; and the research achievements occupy a glorious page in the history of world literature.

Later, United States school is appeared and takig "parallel study" as the symbol, the scholars of this school consider that literature as a discipline should compare with other disciplines.

At present, in United Kingdom, Russia, China and other countries, comparative literature studies have achieved fruitful results.

In the fourth stage, the comprehensive results of the front three stages can be used to expand "comparative literature" with a variety of ways and means. Here we mainly according to Neutrosophy and Quad-stage method to seek expanded results.

According to Neutrosophy and Quad-stage method, if the social sciences can be considered as <A>, then the natural sciences can be considered as the opposite <Anti-A>, and the interdisciplinary sciences can be considered as <Neut-A> (neutral A).

Firstly, link to social sciences, "comparative literature" should be expanded into "comparative social sciences", or "comparative social sciences clusters" including comparative literature, comparative history, comparative philosophy, and so on.

Secondly, link to natural sciences, "comparative literature" should be expanded into "comparative natural sciences", or "comparative natural sciences clusters" including comparative mathematics, comparative physics, chemistry, comparative medicine, comparative biology, and so on.

Thirdly, link to interdisciplinary sciences, "comparative literature" should be expanded into "comparative interdisciplinary sciences", or "comparative interdisciplinary sciences clusters" including comparative mathematical medicine, comparative mathematical biology, and so on.

In addition, the "comparative literature" itself can also be expanded. In addition, comparative literature itself can also be expanded. In references [10], under the two main categories of research and practice, comparative literature can be expanded into: comparative literature research, comparative literature practice (Including comparative essay, comparative fiction, comparative poetry, comparative drama, and so on), comparative literature research and practice, and so on. For the sake of convenience of classification, and to distinguish with other forms of work, naming the essay created by comparative method as comparative essay, the fiction created by comparative method as comparative fiction, the poetry created by comparative method as comparative poetry, the drama created by comparative method as comparative drama, and so on.

## 4. Applications of comparative method in comparative sciences clusters and their branches

"Comparison" means: according to the certain standards and methodologies, to identify advantages and disadvantages, same and different, beauty and ugliness, and so on between two or more things.

The principle of comparison: there shall be the object to be compared with, as well as the common comparative foundation, and the certain standards and methods, and so on; as comparing, we should try to consider all possible situations.

Based on the above concepts and principles, comparative method can be widely used in comparative sciences clusters and their various branches, and provide a variety of ways and broad space for development.

Firstly we discuss the comparative objects. In comparative sciences clusters and their various branches, the comparative objects can be selected within the large range, the medium range, and the small range.

Secondly we discuss the comparative standards. The comparative standards can be selected as: advantages and disadvantages, same and different, beauty and ugliness, and so on. As taking advantages and disadvantages as the comparative standard, the comparative result can be decided by experts, by ordinary scholars and readers, and by all the people (including experts, ordinary scholars and readers).

As for the methods and ways for comparison, they are also numerous. For example, to compare according to the time sequence, according to the different spatial locations; or according to the longitudinal direction and transverse direction; as well as qualitative comparison, quantitative comparison, macro-comparison, micro-comparison; and the combination of different methods and ways.

It needs to be emphasized that, when comparing, we should try to consider all possible situations. This is also the great feature of comparative sciences clusters and their various branches.

As for how to consider all possible situations, we will discuss this problem in another paper.

It should be noted that, in the existing fields of social sciences and natural sciences, many sprouts of comparative sciences clusters can be found, but a wide range of the achievements of comparative sciences clusters, still are the virgin lands to be developed.

## 5. Conclusions

Applying Neutrosophy and Quad-stage Method, as well as comparative methods and ways in comparative sciences clusters and their various branches, will play an extremely important role to promote the development of social sciences, natural sciences, interdisciplinary sciences, and so on; and continue to make new achievements.

## References

1. Florentin Smarandache, A Unifying Field in Logics: Neutrosophic Logic. Neutrosophy, Neutrosophic Set, Neutrosophic Probability and Statistics, third edition, Xiquan, Phoenix, 2003
2. Fu Yuhua, Neutrosophic Examples in Physics, Neutrosophic Sets and Systems, Vol. 1, 2013
3. Fu Yuhua, Quad general theses, general antitheses, universal relations, general syntheses in development —Expansion of Hegelian triad thesis, antithesis, synthesis, Matter Regularity, No.1, 2011, 61-64
4. Fu Yuhua, Expanding Newton Mechanics with Neutrosophy and Quad-stage Method——New Newton Mechanics Taking Law of Conservation of Energy as Unique Source Law, Neutrosophic Sets and Systems, Vol.3, 2014
5. Fu Yuhua, Negating Four Color Theorem with Neutrosophy and Quad-stage Method, Neutrosophic Sets and Systems, Vol.8, 2015
6. Fu Yuhua, Expanding Hegelian Triad Thesis, Antithesis, Synthesis with Neutrosophy and Quad-stage Method, China Preprint Service System (in Chinese) http://www.nstl.gov.cn/preprint/main.html?action=showFile&id=2c9282824a765e950150e9b4a3930677
7. Fu Yuhua, Interpretating and Expanding Laozi's Governing A Large Country Is Like Cooking A Small Fish with Neutrosophy and Quad-stage Method, China Preprint Service System (in Chinese) http://www.nstl.gov.cn/preprint/main.html?action=showFile&id=2c9282824a765e95014fef149c610588
8. Fu Yuhua, Interpretating and Expanding the Meaning of "Yi" with Neutrosophy and Quad-stage Method, China Preprint Service System (in Chinese) http://prep.istic.ac.cn/main.html?action=showFile&id=2c928282510e4d730151799c24a40070
9. Fu Yuhua, Creating Generalized and Hybrid Set and Library with Neutrosophy and Quad-stage Method, Chapter 15, Handbook of Research on Generalized and Hybrid Set Structures and Applications for Soft Computing, Sunil Jacob John. © 2016. 480 pages.
10. Fu Yuhua, Comparative method for literature creative work, China Preprint Service System (in Chinese) http://prep.istic.ac.cn/main.html?action=showFile&id=2c928282510e4d730153ad416f2b0346

# Contributors

**I. Arockiarani**
Department of Mathematics, Nirmala College for Women, Coimbatore, Tamilnadu, India.

**A. Bakali**
Ecole Royale Navale, Boulevard Sour Jdid, B.P 16303 Casablanca, Morocco,
E-mail: broumisaid78@gmail.com

**Durga Banerjee**
Ranaghat Yusuf Institution, P. O. Ranaghat, Dist. Nadia, India-741201
E-mail: dbanerje3@gmail.com

**Kajla Basu**
Department of Mathematics, NIT Durgapur, West Bengal 713209, India.
E-mail: kajla.basu@gmail.com

**Mangal G. Bhatt**
Shantilal Shah Govt. Engineering College, Bhavnagar, Gujarat, India- 364060
E-mail: mangalbhatt15@gmail.com

**Pranab Biswas**
Department of Mathematics, Jadavpur University, West Bengal, Pin 700032, India.
E-mail: paldam2010@gmail.com

**Said Broumi**
Laboratory of Information processing, Faculty of Science Ben M'Sik, University Hassan II, B.P 7955, Sidi Othman, Casablanca, Morocco.
E-mail: broumisaid78@gmail.com,

**M.S. Dabash**
Department of Mathematics and Computer Science, Faculty of Sciences, Port Said University, Egypt.
E-mail: majdedabash@yahoo.com

**Shyamal Dalapati**
Indian Institute of Engineering Science and Technology, Department of Mathematics, Shibpur, Pin-711103, West Bengal, India.
E-mail: dalapatishyamal30@gmail.com

**Partha Pratim Dey**
Department of Mathematics, Jadavpur University, Kolkata-700032, West Bengal, India
E-mail: parsur.fuzz@gmail.com

**M. K. El Gayyar**
Physics and Mathematical Engineering Department, Faculty of Engineering, Port-Said University, Egypt.
E-mail: mohamedelgayyar@hotmail.com

**S. A. El-Hafeez**
Port Said University, Faculty of Science, Department of Mathematics and Computer Science, Egypt.
E-mail: samyabdelhafeez@yahoo.com

**Eman. M. El-Nakeeb**
Port Said University, Faculty of Engineering, Physics and Engineering Mathematics Department, Egypt.
E-mail: emanmarzouk1991@gmail.com

**Hewayda Elghawalby**
Faculty of Engineering, port-said University, Egypt.
E-mail: hewayda2011@eng.psu.edu.eg

**Ahmed K. Essa**
University of Telafer, College of Basic Education Telafer, Mosul, Iraq.
E-mail: ahmed.ahhu@gmail.com

**Shimaa Fathi**
Egypt, Port Said University, Faculty of Engineering, Physics and Engineering Mathematics Department
E-mail: shaimaa_ f_a@eng.psu.edu.eg.

**Bibhas C. Giri**
Department of Mathematics, Jadavpur University, West Bengal, India, Kolkata-700032
E-mail: bcgiri.jumath@gmail.com

**Savita Gupta**
University Institute of Engineering & Technology, Panjab University, Chandigarh, India.
E-mail: savita2k8@yahoo.com

**I.M. Hanafy**
Department of Mathematics and Computer Science, Faculty of Sciences, Port Said University, Egypt.
E-mail: ihanafy@hotmil.com

**Saeid Jafari**
College of Vestsjaelland South, Herrestraede 11,4200, Slagelse, Denmark.

**J. Martina Jency**
Department of Mathematics, Nirmala College for women, Coimbatore, Tamilnadu, India.
E-mail: martinajency@gmail.com

**Huda E. Khalid**
University of Telafer, Mathematics Department, College of Basic Education, Telafer, Mosul, Iraq.
E-mail: hodaesmail@yahoo.com

**Peide Liu**
School of Management Science and Engineering, Shandong University of Finance and Economics, Jinan 250014, Shandong, China.
E-mail: peide.liu@gmail.com

**Francisco Gallego Lupiañez**
Dept. Mathematics, Univ. Complutense, 28040 Madrid, Spain.
E-mail: fg_lupianez@mat.ucm.es

**Kanika Mandal**
Department of Mathematics, NIT Durgapur, West Bengal 713209, India.
E-mail: boson89@yahoo.com.

**Kalyan Mondal**
Department of Mathematics, Jadavpur University, West Bengal, India.
E-mail: kalyanmathematic@gmail.com

**Nital P. Nirmal**
Production Engineering Department, Shantilal Shah Govt. Engineering College, Bhavnagar, Gujarat, India-364060
E-mail: nirmal_nital@gtu.edu.in

**Santanu Ku. Patro**
Department of Mathematics, Berhampur University, Bhanja Bihar - 760007, Berhampur, Odisha, India.
E-mail: ksantanupatro@gmail.com

**Surapati Pramanik**
Department of Mathematics, Nandalal Ghosh B.T. College, Panpur, PO-Narayanpur, and District: North 24 Parganas, Pin Code: 743126, West Bengal, India
E-mail: sura_pati@yahoo.co.in

**Nouran M. Radwan**
Information Systems Dept., Mansoura university, Egypt
E-mail: radwannouran@yahoo.com

**Tapan Kumar Roy**

Indian Institute of Engineering Science and Technology, Shibpur, Pin-711103, West Bengal, India
E-mail: roy_t_k@yahoo.co.in

**A.A. Salama**

Department of Mathematics and Computer Science, Faculty of Science, Port Said University, Egypt

**Florentin Smarandache**

Mathematics & Science Department, University of New Mexico, 705 Gurley Ave., Gallup, NM 87301, USA
E-mail: fsmarandache@gmail.com

**Sukhwinder Singh**

University Institute of Engineering & Technology, Panjab University, Chandigarh, India
E-mail: sukhdalip@yahoo.com

**Mohamed Talea**

Laboratory of Information processing, Faculty of Science Ben M'Sik, University Hassan II, B.P 7955, Sidi Othman, Casablanca, Morocco.
E-mail: broumisaid78@gmail.com

**Mirela Teodorescu**

Neutrosophic Science International Association, New Mexico, USA
E-mail: mirela.teodorescu@yahoo.co.uk

**Fu Yuhua**

CNOOC Research Institute, Beijing, 100028, China.
E-mail: fuyh1945@sina.com

Neutrosophic theory and applications have been expanding in all directions at an astonishing rate especially after the introduction the journal entitled "Neutrosophic Sets and Systems". New theories, techniques, algorithms have been rapidly developed. One of the most striking trends in the neutrosophic theory is the hybridization of neutrosophic set with other potential sets such as rough set, bipolar set, soft set, hesitant fuzzy set, etc. The different hybrid structure such as rough neutrosophic set, single valued neutrosophic rough set, bipolar neutrosophic set, single valued neutrosophic hesitant fuzzy set, etc. are proposed in the literature in a short period of time. Neutrosophic set has been a very important tool in all various areas of data mining, decision making, e-learning, engineering, medicine, social science, and some more.

The Book "New Trends in Neutrosophic Theories and Applications" focuses on theories, methods, algorithms for decision making and also applications involving neutrosophic information. Some topics deal with data mining, decision making, e-learning, graph theory, medical diagnosis, probability theory, topology, and some more.